\documentclass[smallcondensed]{svjour3}
\pdfoutput=1

\smartqed  
\usepackage{graphicx}
\usepackage{amsmath,amssymb,amsfonts}
\usepackage{textcomp}
\usepackage{xcolor}
\usepackage{booktabs}
\usepackage{caption}
\usepackage{subcaption}
\usepackage{enumitem}
\usepackage[ruled,linesnumbered]{algorithm2e}
\usepackage{placeins}
\usepackage{natbib}
\usepackage{hyperref}
\usepackage{multirow}
\usepackage{array}
\usepackage{eqparbox}
\usepackage{etoolbox}  
\usepackage{tabularx,stackengine,collcell}
\usepackage{pifont}
\usepackage{enumitem}

\newcommand{\xmark}{\ding{55}}%

\usepackage[acronym, nonumberlist, hyperfirst=false, nopostdot]{glossaries}

\usepackage[top=1in,bottom=1in,right=1in,left=1in]{geometry}

\usepackage[utf8]{inputenc}
\usepackage[english]{babel}
\usepackage[english=nohyphenation]{hyphsubst}

\SetNlSty{}{}{}
\let\oldnl\nl
\newcommand{\nonl}{\renewcommand{\nl}{\let\nl\oldnl}}

\let\endminwd\relax
\newcolumntype{L}[1]{>{\collectcell\xminwd l{#1}}l<{\endminwd\endcollectcell}}
\newcolumntype{C}[1]{>{\collectcell\xminwd c{#1}}c<{\endminwd\endcollectcell}}
\newcolumntype{R}[1]{>{\collectcell\xminwd r{#1}}r<{\endminwd\endcollectcell}}
\newcolumntype{P}[1]{>{\raggedleft\let\newline\\\arraybackslash\hspace{0pt}}m{#1}}

\def\minwd#1#2#3\endminwd{\stackengine{0pt}{#3}{\rule{#2}{0pt}}{O}{#1}{F}{F}{L}}
\newcommand\xminwd[1]{\minwd#1}

\journalname{Machine Learning}

\makenoidxglossaries
\newacronym{smote}{SMOTE}{Synthetic Minority Oversampling Technique}
\newacronym{sota}{SOTA}{State-Of-The-Art}

\newacronym{kue}{KUE}{Kappa Updated Ensemble}
\newacronym{arf}{ARF}{Adaptative Random Forest}
\newacronym{arfr}{ARFR}{Adaptative Random Forest with Resampling}
\newacronym{lb}{LB}{Leveraging Bagging Adwin}
\newacronym{oba}{OBA}{Online Bagging with ADWIN}
\newacronym{srp}{SRP}{Streaming Random Patches}
\newacronym{csarf}{CSARF}{Cost Sensitive Adaptative Random Forest}
\newacronym{hdvfdt}{HDVFDT}{Hellinger Distance Very Fast Decision Tree}
\newacronym{ghvfdt}{GHVFDT}{Gaussian Hellinger Very Fast Decision Tree}
\newacronym{oob}{OOB}{Oversampling Online Bagging}
\newacronym{uob}{UOB}{Undersampling Online Bagging}
\newacronym{rose}{ROSE}{Robust Online Self-Adjusting Ensemble}
\newacronym{osmote}{OSMOTE}{Online Continuous Synthetic Minority Oversampling Bagging}
\newacronym{ouob}{OUOB}{Online Undersampling and Oversampling Bagging}
\newacronym{csmote}{C-SMOTE}{Continuous Synthetic Minority Oversampling}
\newacronym{oada}{OADA}{Online AdaBoost}
\newacronym{oadaC2}{OADAC2}{Online AdaC2}
\newacronym{orus}{ORUB}{Online RUSBoost}
\newacronym{irl}{IRL}{Incremental Rebalancing Learning}
\newacronym{smoteob}{SMOTE-OB}{Synthetic Minority Oversampling Technique with Online Bagging}
\newacronym{vfcsmote}{VFC-SMOTE}{Very fast continuous synthetic minority oversampling}
\newacronym{esoselm}{ESOS-ELM}{Ensemble of subset online sequential extreme learning machine}
\newacronym{calmid}{CALMID}{Comprehensive active learning method\\ \hspace*{2.7cm} for multiclass imbalanced streaming data with concept drift }
\newacronym{micfoal}{MICFOAL}{Multiclass Imbalanced and Concept Drift\\ \hspace*{2.8cm}Network Traffic Classification Framework Based on Online Active Learning}
\newacronym{dwse}{DWSE}{Dynamic Weighting Selective Ensemble}

\usepackage{amssymb}

\renewcommand{\theenumii}{\theenumi.\arabic{enumii}.}

\begin{document}

\title{A survey on learning from imbalanced data streams: taxonomy, challenges, empirical study, and reproducible experimental framework}
\subtitle{}
\titlerunning{A survey on learning from imbalanced data streams}

\author{Gabriel Aguiar \and Bartosz Krawczyk \and Alberto Cano*}

\institute{G. Aguiar \at
          Department of Computer Science, Virginia Commonwealth University, 401 W. Main St. ERB2334, Richmond, VA 23284\\
          \email{aguiargj@vcu.edu}
          \and
          B. Krawczyk \at
          Department of Computer Science, Virginia Commonwealth University, 401 W. Main St. ERB2316, Richmond, VA 23284\\
          \email{bkrawczyk@vcu.edu}
          \and
          * A. Cano (Corresponding author)\at
          Department of Computer Science, Virginia Commonwealth University, 401 W. Main St. ERB2314, Richmond, VA 23284\\
          \email{acano@vcu.edu}
}

\date{}

\maketitle

\begin{abstract}
Class imbalance poses new challenges when it comes to classifying data streams. Many algorithms recently proposed in the literature tackle this problem using a variety of data-level, algorithm-level, and ensemble approaches. However, there is a lack of standardized and agreed-upon procedures and benchmarks on how to evaluate these algorithms. This work proposes a standardized, exhaustive, and comprehensive experimental framework to evaluate algorithms in a collection of diverse and challenging imbalanced data stream scenarios. The experimental study evaluates 24 state-of-the-art data streams algorithms on 515 imbalanced data streams that combine static and dynamic class imbalance ratios, instance-level difficulties, concept drift, real-world and semi-synthetic datasets in binary and multi-class scenarios. This leads to a large-scale experimental study comparing state-of-the-art classifiers in the data stream mining domain. We discuss the advantages and disadvantages of state-of-the-art classifiers in each of these scenarios and we provide general recommendations to end-users for selecting the best algorithms for imbalanced data streams. Additionally, we formulate open challenges and future directions for this domain. Our experimental framework is fully reproducible and easy to extend with new methods. This way, we propose a standardized approach to conducting experiments in imbalanced data streams that can be used by other researchers to create complete, trustworthy, and fair evaluation of newly proposed methods. Our experimental framework can be downloaded from \url{https://github.com/canoalberto/imbalanced-streams}. 


\end{abstract}



\vspace*{-0.4cm}
\section{Introduction}

Recent advancements in our ability to collect, integrate, store, and analyze big amounts of data led to the emergence of new challenges for machine learning methods. Traditional algorithms were designed to discover knowledge from static datasets. Contrary, contemporary data sources generate information characterized by both volume and velocity. Such a scenario is known as data streams \citep{gama2010knowledge,Bahri2021,read2022learning} and traditional methods lack the speed, adaptability, and robustness to succeed.

One of the biggest challenges, when compared to learning from static data, lies in the need of adapting to the evolving nature of data, where concepts are non-stationary and may change over time. This phenomenon is called concept drift \citep{krawczyk2017ensemble, khamassi2018discussion} and leads to degradation of the classifier, as knowledge learned on previous concepts may not be useful anymore for the recent instances. Recovering from concept drift requires either the presence of explicit detectors or implicit adaptation mechanisms.

Another vital challenge in data stream mining lies in the need for algorithms to display robustness to class imbalance \citep{krawczyk2016learning,fernandez2018learning}. Despite almost three decades of research, handling skewed class distributions is still a crucial domain of machine learning. This becomes even more challenging in the streaming scenario, where imbalance happens simultaneously with concept drift. Not only do the definitions of classes change but also the imbalance ratio becomes dynamic and class roles may switch. Solutions that assume fixed data properties cannot be applied here, as streams may oscillate between varying degrees of imbalance and periods of balance among classes. Furthermore, imbalanced streams can have other underlying difficulties, such as small sample size, borderline and rare instances, overlapping among classes, or noisy labels \citep{santos2022joint}. Imbalanced data streams are usually handled via class resampling \citep{Korycki2020,Bernardo2020,Bernardo2021}, algorithm adaptation mechanism \citep{Loezer2020,Lu2020}, or ensembles \citep{Zyblewski2021,cano2021rose}. This problem is motivated by a plethora of real-world problems where data is both streaming and skewed, such as Twitter streams \citep{Shah:2022}, fraud detection \citep{DeLa:2022}, abuse and hate speech detection \citep{Marwa:2021}, Internet of Things \citep{Sudharsan:2021}, or intelligent manufacturing \citep{Lee:2018}. While there are several works on how to handle imbalanced data streams, there are no agreed-upon standards, benchmarks, or good practices that are necessary for fully reproducible, transparent, and impactful research. 

\smallskip
\noindent\textbf{Research goal.} \ To create a standardized, exhaustive, and informative experimental framework for binary and multi-class imbalanced data streams, and conduct an extensive comparison of state-of-the-art classifiers.

\smallskip
\noindent\textbf{Motivation.} While there are many algorithms for drifting and imbalanced data streams in the literature, there is a lack of standardized procedures and benchmarks on how to evaluate these algorithms holistically. Existing studies are often limited to a selection of algorithms and data difficulties, typically only considering binary class data, and do not provide insights into what aspects of imbalanced data streams must be considered and translated into meaningful benchmark problems. There is a need for a unified and holistic evaluation framework for imbalanced data streams that could be used as a template for researchers to evaluate their newly proposed algorithms against the relevant methods in the literature. Additionally, in-depth experimental comparison of state-of-the-art methods would allow to gain valuable insights into what classifiers and learning mechanisms work under different conditions. Therefore, we propose an evaluation framework and perform a large-scale empirical study to obtain insights into the performance of the methods under an extensive and varied set of data difficulties.

\smallskip
\noindent\textbf{Overview and contributions.} This paper proposes a complete and holistic framework for benchmarking and evaluating classifiers for imbalanced data streams. We summarize existing works and organize them according to established taxonomies dedicated to skewed and streaming problems. We distill the most crucial and insightful problems that appear in this domain and use them to design a set of benchmark problems that capture distinctive learning difficulties and challenges. We compile these benchmarks into a framework embedding various metrics, statistical tests, and visualization tools. Finally, we showcase our framework by comparing 24 state-of-the-art algorithms, which allows us to choose the best-performing ones, discover in what specific areas they excel and formulate recommendations for end-users. The main contributions of the paper are summarized as follows:

\begin{itemize}
\item \textbf{Taxonomy of algorithms for imbalanced data streams.} We organize the methods in the state of the art according to established taxonomies that summarize recent progress in learning from imbalanced data streams and provide a survey of the most important contributions.  
\item \textbf{Holistic and reproducible evaluation framework.} We propose a complete and holistic framework for evaluating classifiers for two-class and multi-class imbalanced data streams that standardizes metrics, statistical tests, and visualization tools to be used for transparent and reproducible research.
\item \textbf{Diverse benchmark problems.} We formulate a set of benchmark problems to be used within our framework. We capture the most vital and challenging problems that are present in imbalance data streams, such as dynamic imbalance ratio, instance-level difficulties (borderline, rare, and subconcepts), or number of classes. Furthermore, we include real-world and semi-synthetic imbalanced problems, leading to a total of 515 data stream benchmarks.
\item \textbf{Comparison among state-of-the-art classifiers.} We conduct an extensive, comprehensive, and reproducible comparative study among 24 state-of-the-art stream mining algorithms based on the proposed framework and 515 benchmark problems.
\item \textbf{Recommendations and open challenges.} Based on the results from the exhaustive experimental study, we formulate recommendations for end-users that will allow to understand the strengths and weaknesses of the best-performing classifiers. Furthermore, we formulate open challenges in learning from imbalanced data streams that should be addressed by researchers in the years to come.
\end{itemize}

\smallskip
\noindent\textbf{Comparison with most related experimental works.} In recent years, several survey papers and works with large experimental studies touching on joint areas of class imbalance and data streams were published. Therefore, it is important to understand the key differences between them and this work, as well as how our survey provides new insights into this topic that were not touched upon in the previous works. Wang et al. \citep{wang2018systematic} proposed an overview of several existing techniques, both drift detectors and adaptive classifiers, and experimentally compared their predictive accuracy. While being the first dedicated study in this area, it was limited by not evaluating computational complexities of compared algorithms, using a very small selection of datasets (7 benchmarks), and investigating only limited properties of imbalanced data streams (not touching upon instance-level characteristics or multi-class problems). Brzezi{\'n}ski et al. \citep{Brzezinski2021} proposed a follow-up study that focused on data-level properties of imbalanced streams, such as instance difficulties (borderline and rare instances) and the presence of subconcepts. However, the study was done for a limited number of algorithms (5 classifiers) and focused only on two-class problems. Bernardo et al. \citep{Bernardo:2021ie} proposed an experimental comparison of methods for imbalanced data streams. They extended Brzezi{\'n}ski et al. \citep{Brzezinski2021} benchmarks using different levels of imbalance ratio and three drift speeds. However, their study analyzed a limited number of algorithms (11 classifiers) and only three real-world datasets. Cano and Krawczyk \citep{cano2021rose} presented a large comparison of 30 algorithms focusing on ensemble approaches but 21 of them were general-purpose ensembles rather than imbalanced specific classifiers. These four works address only binary class imbalanced data streams. This paper extends the benchmark evaluation from all previous studies, proposes new benchmark scenarios, extends the number of real-world datasets, and evaluates both two-class and multi-class imbalanced data streams. We also extend the comparison to 24 classifiers, 19 of them specifically designed for imbalanced data streams.
Table~\ref{tab:works_comparison} summarizes the main differences in the experimental evaluations of these works. 
This allows us to conclude that while these works are an important first step, there is a need for a unified, comprehensive, and holistic study of learning from imbalanced data streams that could be used as a template for researchers to evaluate their newly proposed algorithms.

\begin{table}[h!]
\centering
\caption{Comparison of the number of algorithms and benchmarks evaluated in most related works.}
\label{tab:works_comparison}
\resizebox{\textwidth}{!}{%
\begin{tabular}{@{}l
>{\raggedleft\arraybackslash}p{1.8cm}%
>{\raggedleft\arraybackslash}p{2.5cm}%
>{\raggedleft\arraybackslash}p{2.5cm}%
>{\raggedleft\arraybackslash}p{2.5cm}%
>{\raggedleft\arraybackslash}p{2cm}%
@{}}
\toprule
 & \citep{wang2018systematic} & \citep{Brzezinski2021} & \citep{Bernardo:2021ie} & \citep{cano2021rose} & This paper\\
\midrule
Algorithms & & & & & \\
\ \ General purpose & \xmark & 1 & 2 & 21 & 5 \\
\ \ Imbalanced specific & 10 & 4 & 9 & 9 & 19 \\
Benchmarks & & & & & \\
\ \ Binary class generators & 4 & 385 & 232 & 99 & 406\\
\ \ Binary class datasets & 3 & 4 & 3 & 16 & 19 \\
\ \ Multi-class generators & \xmark & \xmark & \xmark & \xmark & 72 \\
\ \ Multi-class datasets & \xmark & \xmark & \xmark & \xmark & 18\\
\midrule
Total algorithms  & 10 & 5 & 11 & 30 & 24\\
Total benchmarks & 7 & 389 & 235 & 115 & 515\\
\bottomrule
\end{tabular}%
}
\end{table}

This paper is organized as follows. Section~\ref{sec:data_streams} provides a background on data streams. Section\ref{sec:imb_data} discusses the main challenges of imbalanced data. Section~\ref{sec:imb_streams} presents the specific difficulties of imbalanced streams. Section~\ref{sec:ens_streams} describes the approaches for tackling imbalanced steams with ensembles. Section~\ref{sec:exp_setup} introduces the experimental setup and methodology. Section~\ref{sec:results} presents and analyzes the results of our study. Section~\ref{sec:lessons} summarizes the lessons learned. Section~\ref{sec:rec} formulates recommendations to end-users for selecting the best algorithms for imbalanced data streams. Section~\ref{sec:open} discusses the open challenges and future directions. Finally, Section~\ref{sec:conclusion} covers the conclusions.

\vspace{-0.5cm}
\section{Data streams}
\vspace{-0.5cm}
\label{sec:data_streams}

In this section we present the preliminaries of data stream characteristics, learning approaches, and the concept drift properties.

\vspace{-0.75cm}
\subsection{Data stream characteristics}
\vspace{-0.5cm}

\begin{figure}[h!]
    \centering
    \includegraphics[width=0.8\textwidth]{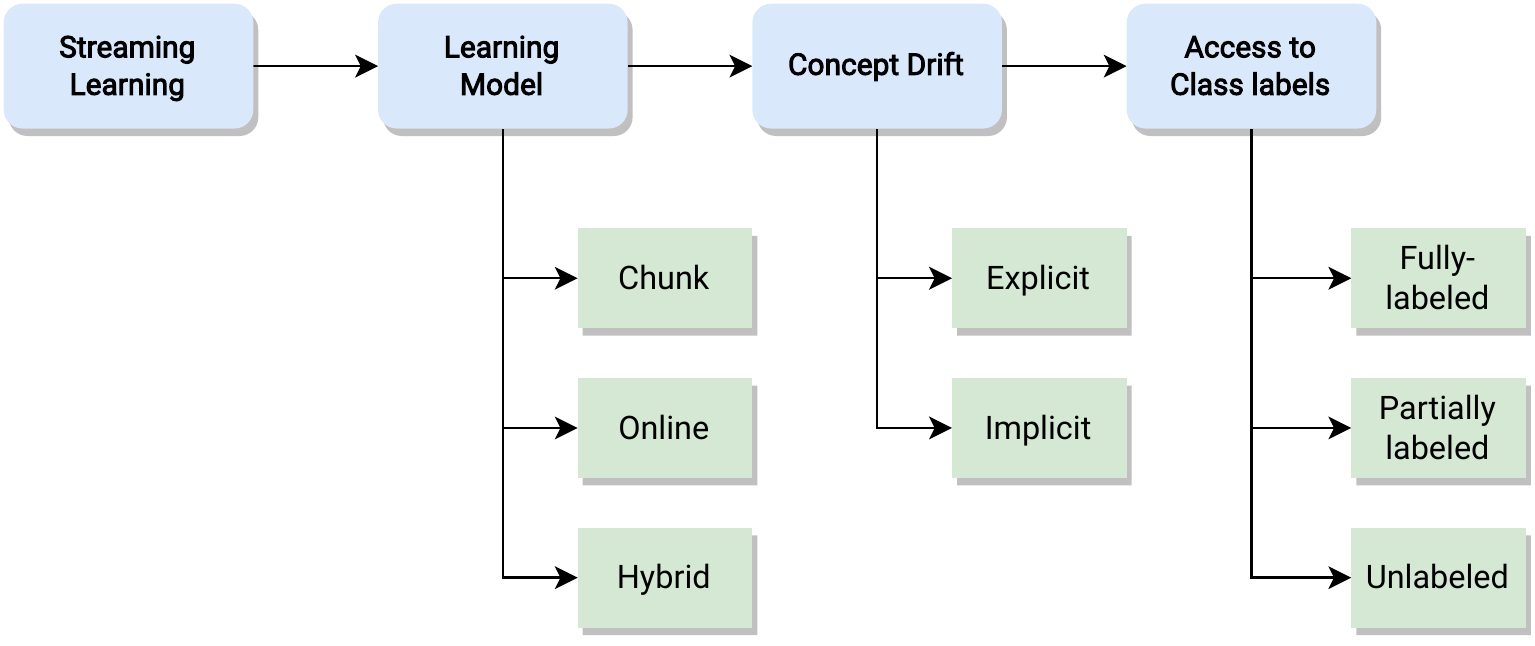}
    \caption{Streaming learning taxonomy.}
    \label{fig:str_tax}
\end{figure}

The main characteristics of data streams can be summarized as follows \citep{gama2010knowledge,krempl2014open,Bahri2021}:

\begin{itemize}
    \item \textbf{Volume.} Streams are potentially unbounded collections of data that constantly flood the system and thus they are impossible to be stored and must be processed incrementally. The volume also imposes limitations on the computational resources, which are magnitudes smaller than the actual size of data would call for.
    \item \textbf{Velocity.} Streaming data sources are in constant motion. New data is being generated continuously and often in rapid bursts, leading to high-speed data streams. These force learning systems to work in real-time, must be analyzed and incorporated into the learning system to model the current state of the stream.
    \item \textbf{Non-stationarity.} Data streams are subject to change over time, which is known as concept drift. This phenomenon may affect feature distributions, class boundaries, but also lead to changes in class proportions, or emergence of new classes (or disappearance of old ones).
    \item \textbf{Veracity.} Data arriving from the stream can be uncertain and affected by various problems, such as noise, injection of adversarial patterns, or missing values. Having access to fully labeled stream is often impossible due to cost and time requirements, leading to need for learning from weakly labeled instances.   
\end{itemize}
We can define a stream \textit{S} as a sequence $<s_1, s_2, s_3, \dots, s_{\infty}>$. We consider a supervised scenario $s_i = (X, y)$, where $X = [x_1, x_2, \dots, x_{f}]$ with $f$ as the dimensionality of the feature space, and $y$ as the target variable, which may or may not be available on arrival. Each instance in the stream is independent and randomly drawn from a stationary probability distribution. Figure \ref{fig:str_tax} illustrates the workflow to learn from data streams and approaches to tackle related challenges \citep{gama2012survey,nguyen2015survey,ditzler2015learning,wares2019data}. 

\subsection{Learning model}
Due to both the volume and velocity of data streams, algorithms need to be capable of incremental processing of the continuously arriving information. Instances from the data stream are provided either online, or in the form of data chunks (portions, blocks).
\begin{itemize}
    \item \textbf{Online.} Algorithms will process each single instance one by one. The main advantage of this approach is a low response time and adaptivity to changes in the stream. The main drawback lies in their limited view of the current state of the stream, as a single instance can be either a poor representation of a larger concept or may be susceptible to noise. 
    \item \textbf{Chunk.} Instances are processed in windows called data chunks or blocks. Chunk-based approaches offer a better estimation of the current concept due to a larger training sample size. The main drawback is the delayed response to changes in some settings because the construction, evaluation, or updating of classifiers is done when all instances from a new block are available. Additionally, in case of rapid changes chunks may consist of instances coming from multiple concepts, further harming the adaptation capabilities.
    \item \textbf{Hybrid.} Hybrid approaches can combine the previous methodologies to address their shortcomings. One of the most popular approaches is to use online learning, while maintaining chunks of data to extract statistics and useful knowledge about the stream for additional periodical classifier updates.
\end{itemize}

\vfill

\subsection{Concept drift}
Data streams are subject to a phenomenon called concept drift \citep{krawczyk2017ensemble, lu2018learning}. Each instance arrives at a time $t$ and is generated by a probabilistic distribution $\Phi^{t} (X,y)$ where $X$ corresponds to the feature vector and $y$ to the class label. If the same probability distribution generates all instances in the stream, data is stationary, i.e., originating from the same concept. On the other hand, if two separate instances, arriving at times $t$ and $t + C$, are generated by $\Phi^{t} (X,y)$ and $\Phi^{t+C} (X,y)$. If $\Phi^{t} \neq \Phi^{t + C}$, then a concept drift occurred. 
When analyzing and understanding concept drift, following factors are considered:
\begin{itemize}
    \item \textbf{Influence of the decision boundaries.} Here we distinguish: (i) virtual; and (ii) real types of drift. Virtual drift can be defined as a change in the unconditional probability distribution $P(x)$, meaning it does not affect the learned decision boundaries. Such drift, while not having a deteriorating influence on learning models, must be detected as it may trigger false alarms and force unnecessary, yet costly adaptation. Real concept drift affects the decision boundaries, making them worthless to the current concept. Detecting it and adapting to new distribution is crucial for maintaining predictive performance.
    \item \textbf{Speed of change.} Here we can distinguish three types of concept drift \citep{webb2016characterizing}: (i) incremental; (ii) gradual; and (iii) sudden. Incremental drift generates a sequence of intermediate states between the old and new concept that are often. This requires detection of the stabilization moment when new concept becomes fully formed and relevant. Gradual drift oscillates between instances coming from both old and new concepts, with new concept becoming more and more frequent over time. Sudden drift instantaneously switches between old and new concept, leading to an instant degradation of the underlying learning algorithm. 
  \item \textbf{Recurrence}. Changes in the stream can be either unique or recurring. In the latter case the previously seen concept may reemerge over time, allowing us to recycle previously learned knowledge. This calls for having a repository of models that can be utilized for faster adaptation to previously seen changes. With more relaxed assumptions, one can extend recurrence to appearance of concepts similar to the ones seen in the past. Here, the past knowledge can be used as initialization point for the drift recovery. 
\end{itemize}

There are two strategies to tackle concept drift: explicit and implicit \citep{lu2018learning,Han2022}:
\begin{itemize}
    \item \textbf{Explicit}. Here drift adaptation is managed by an external tool, called drift detector \citep{barros2018large}. They are used for continuous monitoring of the stream properties (e.g. statistics) or classifier performance (e.g. error rates). Drift detectors raise a warning signal when there are signs of upcoming drift, and alarm signal when the concept drift has taken place. When drift is detected, the classifier is replaced with a new one trained on recent instances. The pitfall of drift detectors is the need for labeled instances (semi-supervised and unsupervised detectors also exist but are less accurate) and false alarms that replaces competent classifiers.
    \item \textbf{Implicit}. Here drift adaptation is managed by learning mechanisms embedded in the classifier, assuming that it can adjust itself to new instances from the latest concept and gradually forget outdated information \citep{ditzler2015learning, da2018strict}. This requires establishing proper learning and forgetting rates, use of adaptive sliding windows, or continual hyperparameter tuning.
\end{itemize}

\subsection{Access to labels}
Obtaining the ground truth (e.g. class labels) in a data stream setting relates to significant time and cost requirements. As instances arrive continuously and in large volumes, domain experts may not be able to label a significant portion of the data or may not be able to provide labels fast enough. In the case of applications where labels can be obtained at no cost (e.g. weather prediction), a significant delay between instance and label arrival must be considered. Data streams can be divided into three groups concerning ground truth availability:

\begin{itemize}
    \item \textbf{Fully-labeled}. For every instance $x$ in the stream the label $y$ is known and can be used for training. This scenario assumes no need for explicit label query and is the most common one for evaluating stream learning algorithms. However, the assumption of a fully labeled stream may not be feasible for many real-world applications.
    \item \textbf{Partially labeled}. Only a subset of instances in the stream are labeled on arrival. The ratio between labeled and unlabeled instances can change overtime. This scenario requires either active learning for selecting most valuable instances for labelling \citep{vzliobaite2013active} or semi-supervised mechanisms for extending the knowledge from labeled instances unto unlabeled ones \citep{Bhowmick2022,suverySSL}.
    \item \textbf{Unlabeled}. Every instance arrives without label and one cannot obtain it upon request, or it will arrive with a significant delay. This forces approximation mechanisms that can either generate pseudo-labels, look for evolving structures in data, or use delayed labels to approximate future concepts.   
\end{itemize}
In this work, only fully labeled streams were used, but some of the algorithms evaluated possess mechanisms to deal with partially labeled or unlabeled streams.

\vspace{-0.25cm}
\section{Imbalanced data}
\label{sec:imb_data}
In this section we will discuss shortly the main challenges present when learning from imbalanced data. Almost three decades of developments in this field allowed us to gain deeper insights into what inhibits the performance of classifier training procedures under skewed distributions \citep{fernandez2018learning}. 
\begin{itemize}
    \item \textbf{Imbalance ratio.} The most obvious and well-studied property of imbalanced datasets is their imbalance ratio, i.e., the disproportion between majority and minority classes. It is commonly assumed that the higher the imbalance ratio, the more difficulty it poses to a classifier. This is justified by the fact that most classifier training procedures are driven by 0-1 loss functions that assume uniform importance of every instance. Therefore, the more predominant the majority class is, the more classifier becomes biased towards it. However, many recent studies have pointed out that the imbalance ratio is not the sole source of learning difficulties \citep{he2013imbalanced}. As long as classes are well-separated and sufficiently represented in the training set, even very high imbalance ratio will not significantly impair the classifier. Therefore, we must look into instance-level properties to find other sources of classifier bias. 
    \item \textbf{Small sample size.} The imbalance ratio is often accompanied by the fact that minority class is appearing infrequently and collecting sufficient number of instances may be costly, time-consuming, or simply impossible. This leads to an issue of small sample size, where minority class does not have big enough training set to allow classifiers to correctly capture its characteristics \citep{Wasikowski:2010}. This, combined with high imbalance ratio, can significantly affect the training procedure, leading to poor generalization capabilities and classification bias. Furthermore, small sample size cannot guarantee that the training set is representative of the actual distribution – problem known as data shift \citep{Rabanser:2019}. 
    \item \textbf{Class overlapping.} Another challenge in imbalanced learning comes from the topology of classes, as often minority and majority classes overlap significantly. Class overlap poses difficulty for standard machine learning problems \citep{Galar:2014}, while presence of skewed distribution makes it even more challenging \citep{Vuttipittayamongkol:2021}. Overlapping regions can be seen as uncertainty regions for classifiers. In such case, the majority class will dominate the training procedure, leading to decision boundary ignoring the minority class in the overlapping area. This problem becomes even more difficult when dealing with multiple classes overlapping with each other. 
   \item \textbf{Instance-level difficulties.} The problem of class overlapping points out to the importance of analyzing the properties of minority class instances and their individual difficulties. Minority classes often form small disjuncts, creating subconcepts that further reduce the minority class sample size in given area \citep{Garcia:2015}. When looking at individual properties of each instance, one can analyze its neighborhood in order to determine how challenging it will be for the classifier. A popular taxonomy divides minority instances into safe, borderline, rare, and outliers based on how homogeneous are the class labels of their nearest neighbors \citep{Napierala:2016}. This information can be utilized to either obtain more effective resampling approaches or guide the classifier training procedure.  
\end{itemize}

\vspace{-0.25cm}
\section{Imbalanced data streams}
\label{sec:imb_streams}
Class imbalance is one of the most vital problems in contemporary machine learning \citep{fernandez2018learning,wang2019learning}. It deals with a disproportion among the number of instances in each class, where some of the classes are significantly underrepresented. As most classifiers are driven by 0-1 loss, they get biased towards the easier to model majority classes. The underrepresented minority classes are usually the more important ones, thus one needs to alter either the dataset or learning procedure to create balanced decision boundaries that do not favor any of the classes. 

Class imbalance is a common problem in the data stream mining domain \citep{wu2014classifying, aminian2019study}. Here streams can have a fixed imbalance ratio, or it may evolve over time \citep{komorniczak2021prior}. Furthermore, class imbalance combined with concept drift poses novel and unique challenges \citep{brzezinski2017prequential,sun2021cost}. Class roles may switch (majority becomes the minority and vice versa), several classes may change (new classes appearing or old disappearing), or instance level difficulties may emerge (evolving class overlapping or clusters/sub-concepts)  \citep{krawczyk2016learning}. Changes in the imbalance ratio can be independent or connected with concept drift, where class definitions ($P(y\mid x)$) will change over time \citep{Wang2020}. Henceforth, monitoring each class for changes in its properties is not enough, as one also needs to track per-class frequencies of arriving new instances. 

In most real-life scenarios streams are not predefined as balanced or imbalanced and they may become imbalanced only temporarily \citep{wang2018systematic}. Users' interests over time (where new topics emerge and old ones lose relevance) \citep{wang2014user}, social media analysis \citep{liu2020event}, or medical data streams \citep{al2019effective} are examples of such cases. Therefore, a robust data stream mining algorithm should display high predictive performance regardless of the underlying class distributions \citep{fernandez2018learning}. Most algorithms dedicated to imbalanced data streams do not perform as well on balanced problems as their canonical counterparts \citep{Cano2020}. On the other hand, these canonical algorithms display low robustness to high imbalance ratios. There exist but few algorithms that can handle both scenarios with satisfactory performance \citep{ Cano2020, cano2021rose}.

There are two main approaches dedicated to handling imbalanced data:

\begin{itemize}
\item \textbf{Data-level approaches.} These methods focus on the alteration of the underlying dataset to make it balanced (e.g. by oversampling or undersampling), thus being classifier-agnostic approaches. They focus on resampling or learning more robust representations. 
\item \textbf{Algorithm-level approaches.} These methods focus on modifying the training approach to make classifiers robust to skewed distributions. They are dedicated to specific learning models, being often more specialized, but less flexible than their data-level counterparts. Algorithm-level modifications focus on identifying mechanisms that suffer from class imbalance, cost-sensitive learning, or one-class classification. 
\end{itemize}

\begin{figure}[b!]
    \centering
    \includegraphics[width=0.7\textwidth]{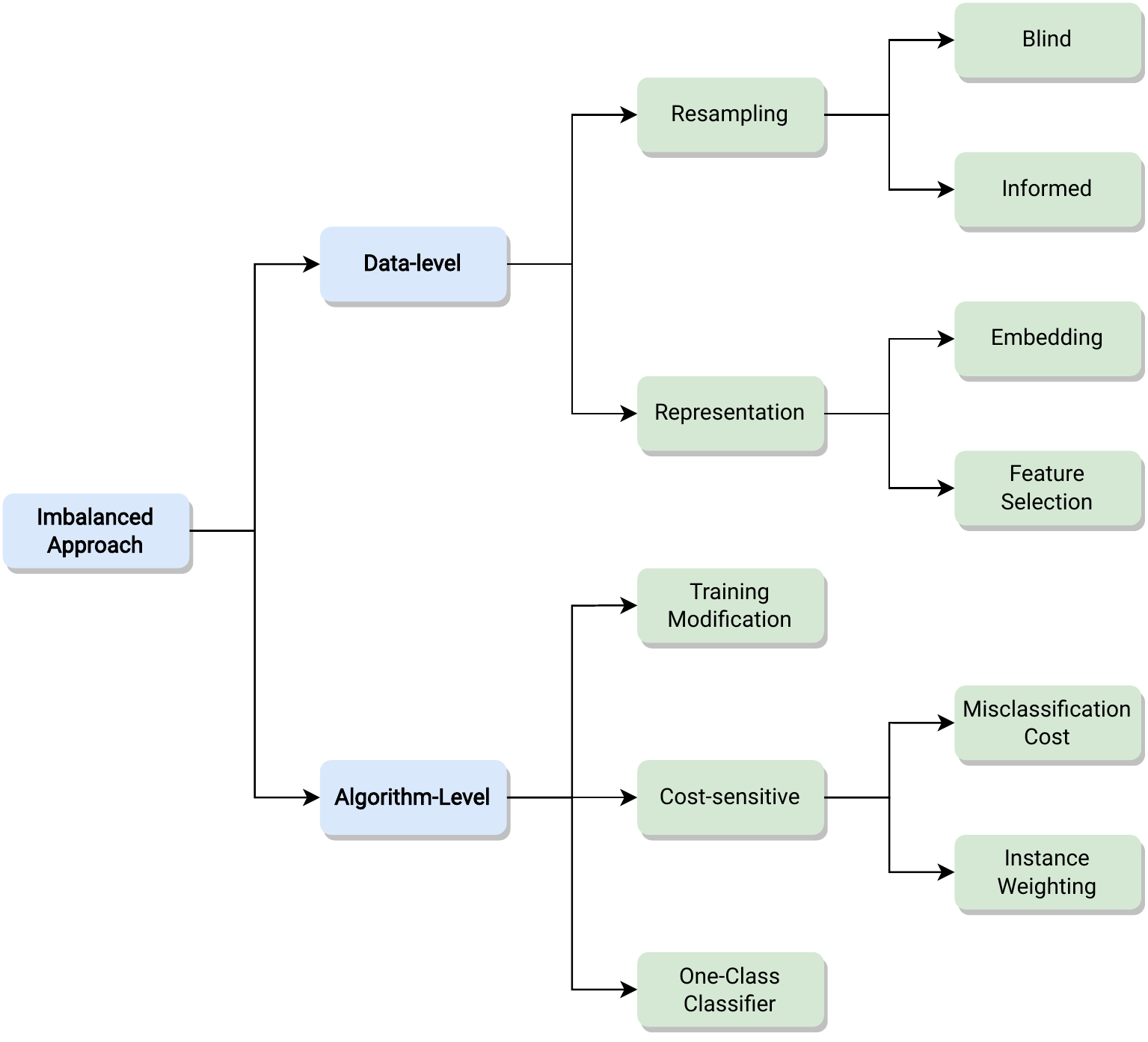}
    \caption{Taxonomy of approaches for imbalanced data streams.}
    \label{fig:imb_taxonomy}
    \vspace*{-0.5cm}
\end{figure}

Figure~\ref{fig:imb_taxonomy} presents a taxonomy \citep{he2009learning,branco2016survey,krawczyk2016learning,fernandez2018learning} of approaches for tackling the class imbalance problem. The specific details are discussed in the following subsections.


\subsection{Data-level approaches}
\vspace{-0.2cm}

While resampling techniques are very popular for static imbalanced problems \citep{fernandez2018learning, aminian2021chebyshev}, they cannot be directly used in the streaming scenario. Concept drift may render resampled data obsolete or even harming to the current state of the stream (e.g. when classes switch roles and resampling starts to empower further the new majority). This calls for dedicated strategies for keeping track of which classes should be resampled at a given moment, as well as for mechanisms capable of dealing with drift by forgetting outdated artificial instances \citep{fernandez2018learning}. 

Resampling algorithms can be categorized as either blind or informed (utilizing information about minority class properties to at least some degree). While blind approaches can be effectively combined with ensembles due to their low computational cost, they do not perform well on their own. Therefore, most resampling methods for data streams are informed and based on a very popular SMOTE (Synthetic Minority Over-sampling Technique) algorithm \citep{ SMOTE2018ann}. Those versions focus on keeping track of changes in the stream by employing either adaptive windows \citep{Korycki2020} or data sketches \citep{Bernardo2021, bernardo2021vfc}. This allows them to generate relevant artificial instances for the current concept and display good reactivity to sudden changes in the stream. It is important to note that the streaming version of SMOTE presented in \citep{Korycki2020} can work with any number of classes, as well as under extremely limited access to class labels. Incremental Oversampling for Data Streams (IOSDS) \citep{anupama2019novel} focuses on replicating instances that are not identified as noisy or overlapping. Clustering of data chunks can be used to identify the most relevant instances to resample \citep{czarnowski2021learning}. Undersampling via Selection-Based Resampling (SRE) \citep{ren2019selection} iteratively removes the safe instances from the majority class without introducing reverse bias towards the minority class. Some works present the usefulness of combining over and under sampling together to obtain a more diverse representation of the minority class \citep{bobowska2019imbalanced}. When handling multi-class imbalanced data streams, resampling can be either conducted using information about all of classes \citep{Korycki2020,sadeghi2021online} or by applying binarization schemes and pairwise resampling \citep{mohammed2020classification}. Active learning techniques such as dynamic budgets \citep{aguiarSAC23} and Racing Algorithms \citep{nguyen2018partial} are also combined with resampling techniques to overcome class imbalance \citep{mohammed2020pwidb}. Disadvantages of data-level methods lie in their high memory use (when oversampling), or the possibility of under-representation of older concepts that are still relevant (when undersampling).

A study by \citet{korycki2021low} discusses an alternative data-level approach to resampling. They propose to create dynamic and low-dimensional embeddings that use information about the class imbalance ratio and separability to find highly discriminative projections. A well-defined low-dimensional embedding may offer better class separability and thus make resampling obsolete, especially when dealing with high-dimensional and difficult imbalanced data streams.

\vspace{-0.3cm}
\subsection{Algorithm-level approaches}
\vspace{-0.2cm}

Among training modifications, the most popular one is the combination of Hoeffding Decision Trees with Hellinger splitting criteria to make skew-insensitive \citep{Lyon2014}. \citet{ksieniewicz2021prior} proposed a method to modify predictions of a base classifier on-the-fly, aiming at modifying prior probabilities based on the frequency of each class. A new loss function was proposed to make neural networks able to handle imbalanced streams in an online setting \citep{ghazikhani2014online}. A combination of online active learning, siamese networks, and multi-queue memory was introduced by \citep{malialis2022nonstationary}. Various modifications of the popular Nearest Neighbors classifier have been adapted to tackling imbalanced data streams by using either dedicated memory formation or skew-insensitive distance metrics \citep{vaquet2020balanced, roseberry2019multi,abolfazli2020drift}. Genetic programming has been successfully used for induction of robust classifiers from the stream \citep{jedrzejowicz2020gep}, as well as increasing skew-insensitive rule interpretability and recovery speed from concept drift \citep{Cano:2019pr}.

Cost-sensitive methods have been applied to streaming decision trees. \citet{krawczyk2017cost} proposed replacing leaves with perceptrons that use cost-sensitive threshold adjustment of class-based outputs. Their cost matrix is adapted in an online fashion to the evolving imbalance ratio, while multiple expositions of difficult instances are used to improve adaptation. Alternatively, Gaussian cost-sensitive decision trees combine cost and accuracy into a hybrid criterion during their training \citep{Guo:2013}. Another approach uses Online Multiple Cost-Sensitive Learning (OMCSL) \citep{yan2017framework} where cost matrices for all classes are adjusted incrementally according to a sliding window. The recent framework proposed two-stage cost-sensitive learning, where a cost matrix is used for both online feature selection and classification \citep{Sun:2020}. Finally, cost-sensitive approaches have been combined with Extreme Learning Machine algorithms via weighting matrices and misclassification costs \citep{li2021online}. 

One-class classification is an interesting solution to class imbalance, where one uses these class-specific models to either describe minority class or all the classes (achieving a one-class decomposition of multi-class problems) \citep{Krawczyk:2018pr}. One-class classifiers can be used for data stream mining scenarios and display good reactivity to concept drift \citep{Krawczyk:2015}. One can use adaptive online one-class Support Vector Machines to track minority classes and their changes over time \citep{klikowski2020employing}. One can combine one-class classification with ensembles, over-sampling, and instance selection \citep{CZARNOWSKI2022101614}. One-class classifiers can be combined with active learning to select the most informative instances from the stream to be used for class modeling \citep{Gao:2015}. Anomaly detection, similar in its assumptions to one-class classifiers can also be used to identify minority and majority instances in the stream \citep{liang2021anomaly}.

\vspace{-0.2cm}
\subsection{Similar domains}
\vspace{-0.2cm}

When talking about learning from imbalanced data streams, it is necessary to mention to similar domains in contemporary machine learning, namely continual learning and long-tailed recognition. 

\smallskip
\noindent \textbf{Similarities to continual learning.} It is important to mention that data stream mining can often be viewed as task-free continual learning \citep{Krawczyk:2021}. While imbalanced problems have not been yet discussed widely in this setup, there are some works noticing the importance of handling skewed class distributions for continual deep learning \citep{Chrysakis:2020,Kim:2020,Arya22,priya2021deep}.

\smallskip
\noindent \textbf{Similarities to long-tailed recognition.} The extreme case of multi-class imbalance is known as long-tailed recognition \citep{Yang:2022}. It deals with situations, where we have hundreds or thousands of classes, with progressively increasing imbalance ratio and smallest classes being extremely imbalanced compared to the majority ones (hence long-tailed class-based distribution of instances). This problem is mainly discussed in the context of deep learning, where various decomposition strategies \citep{Zhu:2022}, loss functions \citep{Zhao:2022}, or cost-sensitive solutions \citep{Peng:2022} are being utilized. Currently, there are but few works that discuss the combined challenge of continual learning from long-tailed distributions \citep{Kim:2020}.

\vspace{-0.2cm}
\section{Ensembles for imbalanced data streams}
\vspace{-0.2cm}
\label{sec:ens_streams}

Combining multiple classifiers into an ensemble is one of the most powerful approaches in modern machine learning, leading to improved predictive performance, generalization capabilities, and robustness. Ensembles have proven themselves to be highly effective for data streams, as they offer unique ways of managing concept drift and class imbalance \citep{krawczyk2017ensemble}. The former can be achieved by adding new classifiers or updating the existing ones, while the latter is achieved by combining classifiers with different skew-insensitive approaches \citep{brzezinski2018ensemble, grzyb2021hellinger, du2021online}. 

Ensembles for data streams can be categorized by the following design choices:
\begin{itemize}
\item \textbf{Classifier pool generation.} There are two major approaches for generating a pool of classifiers for forming an ensemble: heterogeneous and homogeneous \citep{Bian:2007}. Heterogeneous solutions assume that we ensure diversity of the pool by using different classifier models, aiming at exploiting their individual strengths at forming decision boundaries. Homogeneous solutions assume that we select a specific type of classifier (e.g., popular choice are decision trees) and then ensure diversity among them by modification of the training set. This is usually achieved by one of two popular solutions: bagging and boosting. Bagging (bootstrap aggregating) trains multiple independent base learners in parallel and combines their predictions using an aggregation function (e.g. by simple average or simple majority vote). Boosting trains the base learners in a sequential way. Each model in the sequence is fitted giving more importance to observations in the dataset that were poorly handled by the previous models. Predictions are combined using a deterministic strategy (e.g. weighted majority voting). It is worthwhile noting that while the majority of the methods are based on either heterogeneous pool or homogeneous weak learners, there exist alternative approaches, such as generating hybrid pools (using multiple types of models, but also generating multiple learners for each of them) \citep{Luong:2020} and using projections \citep{korycki2021low}.
\item \textbf{Feature space modification.} This defines what feature space input is being used by base classifiers. They can either be trained on full feature space (here their diversity must be ensured in another way), feature subspaces, or completely new feature embeddings (e.g. creating artificial feature spaces).
\item \textbf{Ensemble line-up.} This defines how ensembles are managed during the continual learning from streams. Voting procedures can be used for dynamical adjustment of base learners’ importance. Ensembles can be fixed, meaning that each base learner is continuously updated, but never removed. Alternatively, one can use a dynamic setup, where worst classifiers are pruned and replaced by new ones trained on more recent instances. Finally, all of these mentioned techniques can be combined to create hybrid architectures, capable of better responsiveness to concept drift. 
\end{itemize}

\begin{figure}[b!]
    \centering
    \includegraphics[width=0.9\textwidth]{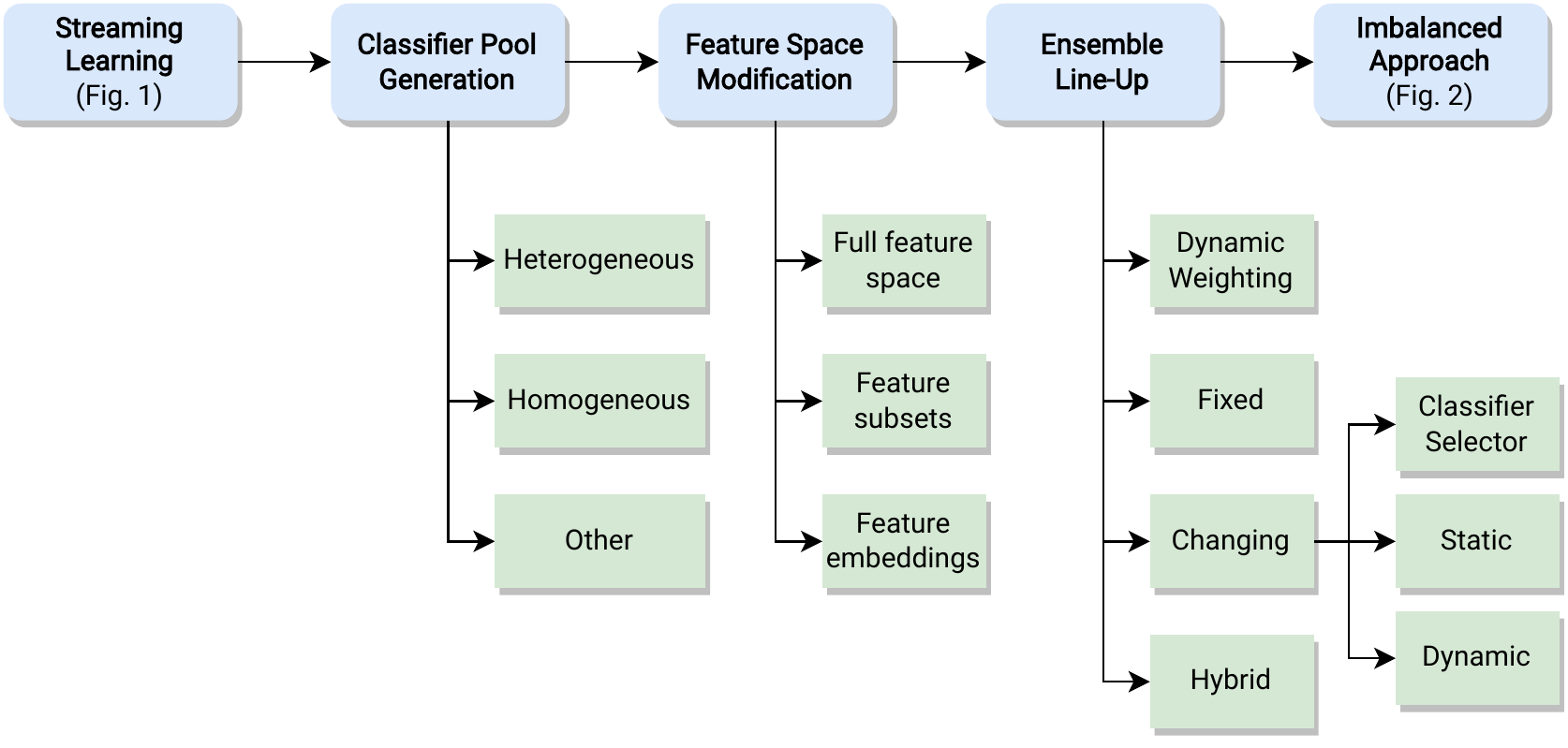}
    \caption{Taxonomy of ensemble definition for imbalanced data streams.}
    \label{fig:ensemble_taxonomy}
\end{figure}

For imbalanced data streams, ensembles are usually combined with techniques mentioned in the previous section. Figure \ref{fig:ensemble_taxonomy} presents a taxonomy  \citep{krawczyk2017ensemble,gomes2017survey} based on how ensembles are built for data streams and how this can be connected with the previously discussed approaches to handle drifting and imbalanced streams. 

The most popular approach lies in combining resampling techniques with Online Bagging \citep{wang2014resampling,wang2016dealing, wang2016online}. Similar strategies can be applied to Adaptive Random Forest \citep{gomes2017adaptive}, Online Boosting \citep{klikowski2019multi, gomes2019streaming}, Dynamic Weighted Majority \citep{lu2017dynamic}, Dynamic Feature Selection \citep{wu2014classifying}, Adaptive Random Forest with resampling \citep{ferreira2019adaptive}, Kappa Updated Ensemble \citep{Cano2020}, Robust Online Self-Adjusting Ensemble \citep{cano2021rose}, Deterministic Sampling Classifier with weighted Bagging \citep{KLIKOWSKI2022108855}, Dynamic Ensemble Selection \citep{jiao2022dynamic,han2023dynamic} or any ensemble that can incrementally update its base learners \citep{ancy2020handling,li2020incremental}. It is interesting to note that preprocessing approaches enhance diversity among base classifiers \citep{zyblewski2019classifier}. Alternatively, cost-sensitive solutions can be used together with ensembles such as Adaptive Random Forest \citep{Loezer2020}.

The effectiveness of ensembles for imbalanced data streams can be further increased by using dedicated combination schemes or adaptive chunk-based learning \citep{Lu2020}. Weights assigned for each base classifier can be continuously updated to reflect their current competencies on minority classes \citep{ren2018gradual}. A reinforcement learning mechanism can be used to increase the weights of the base classifiers that perform better on the minority class \citep{zhang2019resample}. One can use a hybrid approach that combines resampling minority instances with dynamic weighting base classifiers based on their predictive performance on sliding windows of minority samples \citep{yan2021dynamic}. 
Dynamic selection of classifiers and their related preprocessing techniques can be a very effective tool for handling concept drift, as it offers exploitation of diversity among base classifiers \citep{Zyblewski2021, zyblewski2021dynamic}. Alternatively, classifier selection balances subsets of the incoming stream. Cost-sensitive neural networks can be initialized using different random weights and then incrementally improved with new instances \citep{ghazikhani2013ensemble}. OSELM \citep{li2021online} classifiers can be combined using diverse initialization to generate a more robust compound classifier \citep{wang2021ensemble}.

Finally, ensembles found their applications in imbalanced data streams with limited access to class labels. CALMID is a robust framework to deal with limited label access, concept drift, and class imbalance by dynamically inducing new base classifiers with the weighting of the most relevant instances \citep{liu2021comprehensive}. Another approach uses reinforcement learning \citep{zhang2020reinforcement} to select instances for updating the ensemble under labeling constraints. In multi-class imbalance settings, self-training semi-supervised \citep{vafaie2020multi} methods were applied to self-labeling driven by a small subset of labeled instances. It can be realized by an abstaining mechanism temporarily removing uncertain classifiers, with dynamically adjusting the abstaining criterion in favor of minority classes \citep{korycki2019active}.

While the vast majority of mentioned ensembles use Hoeffding Decision Trees (or their variants) as base classifiers, there are several skew-insensitive ensembles dedicated to neural networks. ESOS-ELM \citep{mirza2015ensemble} maintains randomized neural networks that are trained on balanced subsets of the incoming stream. Cost-sensitive neural networks can be initialized using random weights and then incrementally improved with new instances \citep{ghazikhani2013ensemble}. OSELM \citep{li2021online} classifiers can be combined using diverse initialization to generate a more robust compound classifier \citep{wang2021ensemble}.

Finally, ensembles found their applications in imbalanced data streams with limited access to class labels. CALMID is a robust framework to deal with limited label access, concept drift, and class imbalance by dynamically inducing new base classifiers with the weighting of the most relevant instances \citep{liu2021comprehensive}. Another approach uses reinforcement learning \citep{zhang2020reinforcement} to select instances for updating the ensemble under labeling constraints. In multi-class imbalance settings, self-training semi-supervised \citep{vafaie2020multi} methods were applied to self-labeling driven by a small subset of labeled instances. 

\vspace{-0.3cm}
\section{Experimental setup}
\vspace{-0.2cm}
\label{sec:exp_setup}

The experimental study was designed to evaluate the performance of data stream mining algorithms under varied imbalanced scenarios and difficulties. We aim at gaining a better understanding of the data difficulties and how they impact the classifiers. We address the following research questions (RQ):

\vspace{-0.1cm}
\begin{itemize}
\item \textbf{RQ1}: How do different levels of class imbalance ratio affect the algorithms?
\item \textbf{RQ2}: How do static vs dynamic imbalance ratios influence the classifiers?
\item \textbf{RQ3}: How do instance-level difficulties impact the classifiers?
\item \textbf{RQ4}: How do algorithms adapt to simultaneous concept drift and imbalance ratio changes?
\item \textbf{RQ5}: Are there differences on the performance between imbalanced generators and real-world streams?
\item \textbf{RQ6}: Is there trade-off between the accuracy and the computational and memory complexities?
\item \textbf{RQ7}: What are the lessons learned? Which algorithm should I use in my dataset?
\end{itemize}
\vspace{-0.1cm}

\noindent To answer these questions, we formulate a set of benchmark problems building on experiments proposed in previous studies and new ones to assess additional data difficulties in two-class and multi-class imbalanced data streams. 
One of the major issues in this research area is the lack of standardized and agreed-upon procedures and benchmarks on how to evaluate these algorithms holistically. Therefore, we evaluate a comprehensive set of benchmark problems which includes an exhaustive list of data difficulties in imbalanced data streams. The experimental study in Section~\ref{sec:results} is divided into the following experiments whereas Section~\ref{sec:lessons} discusses the lessons learned and recommendations.

\renewcommand{\labelenumiii}{\arabic{enumi}.\arabic{enumii}.\arabic{enumiii}}

\fbox{\parbox{0.9\textwidth}{
\hspace{0.01cm} Index of experiments
\begin{enumerate}
\item[]
\begin{enumerate}
\small
\setcounter{enumi}{7}
    \item \hyperref[sec:bc-experiments]{Binary class experiments}
    \begin{enumerate}
    \item \hyperref[sec:bc-static-imb]{Static imbalance ratio}
    \item \hyperref[sec:bc-dyn-imb]{Dynamic imbalance ratio}
    \item \hyperref[sec:bc-instance-level-diff]{Instance-level difficulties}
    \item \hyperref[sec:bc-cd-static-ratio]{Concept drift and static imbalance ratio}
    \item \hyperref[sec:bc-cd-dynamic-ratio]{Concept drift and dynamic imbalance ratio}
    \item \hyperref[sec:bc-rl-wd]{Real-world binary class imbalanced datasets}
    \end{enumerate}
    \item \hyperref[sec:mc-experiments]{Multi-class experiments}
    \begin{enumerate}
    \item \hyperref[sec:mc-static-IR]{Static imbalance ratio}
    \item \hyperref[sec:mc-dynamic-ir]{Dynamic imbalance ratio}
    \item \hyperref[sec:mc-cd-static-ir]{Concept drift and static imbalance ratio}
    \item \hyperref[sec:mc-cd-dynamic-ir]{Concept drift and dynamic imbalance ratio}
    \item \hyperref[sec:mc-nclasses]{Impact of the number of classes}
    \item \hyperref[sec:mc-datasets]{Real-world multi-class imbalanced datasets}
    \item \hyperref[sec:mc-semisynthetic]{Semi-synthetic multi-class imbalanced datasets}
    \end{enumerate}
    \item \hyperref[sec:overall]{Overall comparison}
    \vspace{-0.2cm}
\end{enumerate}
\end{enumerate}
}}

\vspace{-0.25cm}
\subsection{Algorithms}
\vspace{-0.25cm}

The experiments comprise $24$ state-of-the-art algorithms for data streams, including best-performing general-purpose ensembles and algorithms specifically designed for imbalanced streams. Algorithms are presented in Table~\ref{tab:algorithms} with their characteristics according to the established taxonomies. Specific properties of the ensemble models are presented in Table~\ref{tab:ensembles}. All algorithms are implemented in MOA \citep{bifet2010moa}. The source code of the algorithms and the experiments are publicly available on GitHub to facilitate the transparency and reproducibility of this research\footnote{Source code, experiments, and results are available at \url{https://github.com/canoalberto/imbalanced-streams}}. All results, interactive plots and tables are available on the website\footnote{Interactive plots and tables for all experiments are available at \url{https://people.vcu.edu/~acano/imbalanced-streams}}. Algorithms were run on a cluster with 2,300 AMD EPYC2 cores, 12 TB RAM, and Centos~7. No individual hyperparameter optimization was conducted for any algorithm. All algorithms use the parameter settings recommended by their authors on their respective implementations. All ensembles are evaluated with the same parameter settings of 10 base classifiers using Hoeffding tree as the base learner. We acknowledge that algorithms often depend on parameters that may have a significant impact on the results obtained. Some methods use random generators which require an initial random seed. Different seeds will produce different results and multiple seeds should be run when the number of benchmarks is small due to the central limited theorem.
Other methods have parameters that affect the classifier learning (e.g. the split confidence of the Hoeffding tree) that should be more carefully chosen when fitting a particular dataset.
Due to the large number of benchmarks, experiments, and data size, the results reported on the paper are the median for 5 runs (5 seeds). Complete results to facilitate future comparisons and detailed information about the specific parameter configuration are available on the GitHub repository.

\begin{table}[t!]
\centering
\caption{Data stream algorithms and their taxonomy.\\CS: cost-sensitive, TM: training modification, RI: informed resampling\\ R(O$\mid$U)B: blind (over$\mid$under) resampling, BL: base-learner, H: hybrid.}
\label{tab:algorithms}
\resizebox{\textwidth}{!}{%
\begin{tabular}{@{}lccccc@{}}
\toprule

Algorithm &
  \multicolumn{1}{c}{\begin{tabular}[c]{@{}c@{}}Imbalanced approach\end{tabular}} &
  \multicolumn{1}{c}{\begin{tabular}[c]{@{}c@{}}Learning \\ method\end{tabular}} &
  \multicolumn{1}{c}{\begin{tabular}[c]{@{}c@{}}Concept \\ drift\end{tabular}} &
  \multicolumn{1}{c}{\begin{tabular}[c]{@{}c@{}}Multi\\ class\end{tabular}} &
  \multicolumn{1}{c}{\begin{tabular}[c]{@{}c@{}}BL \\ agnostic\end{tabular}}  \\
\midrule
\acrshort{irl} \citep{bernardo2020incremental} & RI   & Online & Explicit &  \xmark &  \checkmark \\
\acrshort{csmote} \citep{Bernardo2020} & RI  & Online & Explicit &  \xmark  &  \checkmark \\
\acrshort{vfcsmote} \citep{bernardo2021vfc} &  RI   & Online & Explicit &  \xmark &  \checkmark \\

\acrshort{csarf} \citep{Loezer2020} & CS   & Online & Explicit & \checkmark & \xmark\\
\acrshort{ghvfdt} \citep{Lyon2014} & TM  & Online & Implicit & \checkmark & \xmark \\
\acrshort{hdvfdt} \citep{cieslak2008learning} & TM   & Online & Implicit & \checkmark &  \xmark \\

\acrshort{arf} \citep{gomes2017adaptive} &  \xmark  & Online & Explicit & \checkmark &  \xmark \\
\acrshort{kue} \citep{Cano2020} &  H  & Hybrid & Explicit & \checkmark &   \checkmark\\
\acrshort{lb} \citep{bifet2010leveraging} &  \xmark  & Online & Explicit & \checkmark & \checkmark \\ 
\acrshort{oba} \citep{bifet2009new} &  \xmark & Online & Explicit & \checkmark  & \checkmark \\
\acrshort{srp} \citep{gomes2019streaming} &  \xmark  & Online & Explicit &  \checkmark &  \checkmark\\
\acrshort{esoselm} \citep{mirza2015ensemble} &  RB  & Chunk & Explicit & \xmark & \xmark \\
\acrshort{calmid} \citep{liu2021comprehensive} &  H  & Hybrid & Explicit &  \checkmark &  \checkmark \\
\acrshort{micfoal} \citep{liu2021comprehensive} &  H  & Online & Explicit & \checkmark & \checkmark\\
\acrshort{rose} &  H  & Hybrid & Explicit &  \checkmark  &  \checkmark \\
\acrshort{oada} \citep{wang2016online} &  \xmark  & Online & Explicit &  \checkmark &  \checkmark \\
\acrshort{oadaC2} \citep{wang2016online} &  CS  & Online & Explicit & \xmark &  \checkmark \\

\acrshort{arfr} \citep{ferreira2019adaptive} &  RI  & Online & Explicit & \checkmark &  \xmark \\
\acrshort{smoteob} \citep{Bernardo2021} &   RI  & Online & Explicit  &  \xmark & \checkmark \\
\acrshort{osmote} \citep{wang2016online} &   RI   & Online & Explicit &  \xmark &  \checkmark \\
\acrshort{oob} \citep{wang2016dealing} &  ROB  & Online & Implicit &  \checkmark &  \checkmark \\
\acrshort{uob} \citep{wang2016dealing} &  RUB  & Online & Implicit &  \checkmark  & \checkmark \\
\acrshort{orus} \citep{wang2016online} &  RUB  & Online & Explicit &  \xmark & \checkmark \\
\acrshort{ouob} \citep{wang2016online} &  RB  & Online & Explicit & \xmark &   \checkmark \\

\bottomrule
\end{tabular}%
}
\end{table}

\begin{table}[t!]
\centering
\caption{Ensemble algorithms and their taxonomy.}
\label{tab:ensembles}

\begin{tabular}{@{}llll@{}}
\toprule
Ensemble & Meta-algorithm & Feature space & Line-up \\
\midrule

\acrshort{arf} & Bagging & Feature subsets & Fixed \\
\acrshort{kue} & Bagging & Feature subsets & Dynamic weighting \\
\acrshort{lb} & Bagging & Full feature space & Fixed \\
\acrshort{oba} & Boosting & Full feature space & Fixed \\
\acrshort{srp} & Bagging & Feature subsets & Change dynamic \\
\acrshort{esoselm} & Other & Full feature space & Hybrid \\
\acrshort{calmid} & Other & Full feature space & Change dynamic \\
\acrshort{micfoal} & Other & Full feature space & Change dynamic \\
\acrshort{rose} & Bagging & Feature subsets & Dynamic weighting \\
\acrshort{oada} & Boosting & Full feature space & Fixed \\
\acrshort{oadaC2} & Boosting & Full feature space & Fixed \\
\acrshort{arfr} & Bagging & Feature subsets & Fixed \\
\acrshort{smoteob} & Bagging & Feature subsets & Fixed \\
\acrshort{osmote} & Bagging & Feature subsets & Fixed \\
\acrshort{oob} & Bagging & Full feature space & Fixed \\
\acrshort{uob} & Bagging & Full feature space & Fixed \\
\acrshort{orus} & Boosting & Feature subsets & Fixed \\
\acrshort{ouob} & Bagging & Feature subsets & Fixed \\

\bottomrule
\end{tabular}%

\end{table}

\vspace{-0.35cm}
\subsection{Generators}
\vspace{-0.25cm}

To evaluate the classifiers in specific and controlled scenarios, we prepared data streams generators under different imbalanced and drifting settings. Nine generators in MOA \citep{bifet2010moa} plus one generator proposed by Brzezi{\'n}ski \citep{Brzezinski2021} were used. Those generators are presented in Table~\ref{tab:generators}, with their number of attributes, classes, and whether they can generate internal concept drifts. All generators are evaluated on a stream of 200,000 instances. For generators where it is possible to use a configurable number of attributes, the default value on the table was used. The number of classes was adjusted according to the experiment (2 for binary class experiments and 5 for multi-class experiments). 

\begin{table}[h!]
\centering
\caption{Specifications of the data stream generators.}
\label{tab:generators}
\begin{tabular}{lrrl}
\toprule
Generator & Attributes & Classes & Concept drift\\
\midrule
Agrawal              & 10              & 2 & \checkmark \ Functions \\
Asset                & 5               & 2 & \checkmark \ Functions \\
Brzezi{\'n}ski           & N (5)  & 2 &  \checkmark \ Centroids\\
Hyperplane           & 12              & N (2 binary class, 5 multi-class) & \checkmark \ Features\\
Mixed                & 4               & 2 & \checkmark \ Function \\
RandomRBF            & N (10) & N (2 binary class, 5 multi-class) & \checkmark \ Centroids \\
RandomTree           & N (10) & N (2 binary class, 5 multi-class) & \checkmark \ Trees \\
SEA                  & 3               & 2 & \checkmark \ Function \\
Sine                 & 3               & 2 & \checkmark \ Function \\
Text                 & 100             & 2 & \xmark \\
\bottomrule
\end{tabular}
\end{table}

\subsection{Performance evaluation}
\label{sec:metrics}

The algorithms were evaluated using the test-then-train model, where each instance is first used to test then update the classifier in an online manner (instance by instance). We measured seven performance metrics (Accuracy, Kappa, G-Mean, AUC, PMAUC, WMAUC, and EWMAUC). Complete results are available on the website \url{https://people.vcu.edu/~acano/imbalanced-streams}. However, due to the limitations of space in the manuscript, we show results for Kappa, G-Mean, and the Area Under the Curve (AUC). They are calculated over a sliding window of 500 instances. We also acknowledge that there are different schools of thought regarding
the best selection of performance metrics for imbalanced data. Our argument is that in order to have a comprehensive evaluation of the classifier performance on imbalanced datasets, one should not use only one metric, whichever the metric is, since all metrics have biases one way or another, and focus on assessing different aspects. Therefore, in our study, we report pairs of metrics that we have observed they exhibit complementary behaviors.

Kappa is often used to evaluate classifiers in imbalanced settings \citep{japkowicz2013assessment, brzezinski2018visual, brzezinski2019dynamics}. It evaluates the classifier performance by computing the inter-rater agreement between the successful predictions and the statistical distribution of the data classes, correcting agreements that occur by mere statistical chance. Kappa values range from $-100$ (total disagreement) through $0$ (default probabilistic classification) to $100$ (total agreement) as Eq.~\ref{eq:kappa}.

\begin{equation}
\label{eq:kappa}
Kappa = \displaystyle\frac{n\displaystyle\sum_{i=1}^c{x_{ii}}-\displaystyle\sum_{i=1}^c{x_{i.}x_{.i}}}{n^2-\displaystyle\sum_{i=1}^c{x_{i.}x_{.i}}} \cdot 100
\end{equation}

\noindent where $x_{ii}$ is the count of cases in the main diagonal of the confusion matrix, $n$ is the number of examples, $c$ is the number of classes, and $x_{.i}$, $x_{i.}$ are the column and row total counts, respectively. Kappa punishes homogeneous predictions, which is very important to detect in imbalanced scenarios but can be too drastic in penalizing misclassifications on difficult data. Moreover, Kappa provides better insights in detecting changes in the distribution of classes in multi-class imbalanced data. However, some authors recommend to avoid Kappa as Kappa's values vary depending not only on the performance of the model in question, but also on the level of class imbalance in the data, which can make the analyses difficult \citep{luque2019impact}.

To tackle a balance between the performance of classifiers on the majority and minority classes, many researchers consider null-bias metrics such as sensitivity and specificity \citep{brzezinski2018ensemble}. These metrics are based on the confusion matrix: true positive (TP), true negative (TN), false positive (FP), and false negative (FN). Sensitivity, also called recall, is the ratio of correctly classified instances from the minority class (true positive rate) defined in Eq.~\ref{eq:sensitivity}. Specificity is the ratio of instances correctly classified from the majority class (true negative rate) defined in Eq.~\ref{eq:specificity}. The geometric mean (G-Mean) is the product of the two metrics as defined in Eq.~\ref{eq:gmean}. This measure tries to maximize the accuracy of each of the classes while keeping these accuracies balanced. G-Mean is a recommended null-bias metric for class imbalance \citep{luque2019impact}. For multi-class data, the geometric mean is the square root of the product of class-wise sensitivity. However, this introduces the problem that as soon as the recall for one class is 0 the product of the whole geometric mean becomes 0. Therefore, it is much more complicated to use in multi-class experiments with a large number of classes and consequently, AUC would be preferred.

\noindent
\begin{minipage}[h!]{0.5\columnwidth}
\begin{equation}
\label{eq:sensitivity}
Sensitivity = Recall = \displaystyle\frac{TP}{TP + FN}
\end{equation}
\end{minipage}
\begin{minipage}[h!]{0.5\columnwidth}
\begin{equation}
\label{eq:specificity}
Specificity = \displaystyle\frac{TN}{TN + FP}
\end{equation}
\end{minipage}

\begin{equation}
\label{eq:gmean}
G-Mean = \displaystyle\sqrt{Sensitivity \times Specificity}
\end{equation}

The Area Under the Curve (AUC) is invariant to changes in class distribution and provides a statistical interpretation for scoring classifiers. However, to measure the ranking ability of the classifiers the AUC needs to sort the data and iterate through each example. We employ the prequential AUC formulation proposed by~\citet{brzezinski2017prequential} which uses a sorted tree structure with a sliding window. The AUC formulation was extended by \citep{Wang2020} for multi-class problems defining the Prequential Multi-Class (PMAUC) as Eq.~\ref{eq:pmauc}. 

\begin{equation}
\label{eq:pmauc}
PMAUC = \displaystyle\frac{1}{C(C-1)} \cdot \sum_{i\neq j} A (i\mid j)
\end{equation}

\noindent where $A (i\mid j)$ is pairwise AUC when treating class $i$ as the positive class and class $j$ as negative, and $C$ is the number of classes. Extensions of the PMAUC calculation include Weighted Multi-class AUC (WMAUC) and Equal Weighted Multi-class AUC (EWMAUC) \citep{Wang2020}.

Both AUC and G-Mean are blind regarding the level of the class imbalance, while Kappa takes into account the class distribution but makes it more difficult to understand. Therefore, in cases of extreme imbalance ratios, the Kappa metric can be very dissimilar to the G-Mean and AUC, which means a classifier can have a high value of AUC, but a very low Kappa value. This is very useful to understand the behavior of a classifier under high imbalance ratios and how different metrics exhibit complementary facets of the classification performance. Therefore, it is important to evaluate the algorithms using both metrics in other to counterbalance overestimation. Henceforth, in our experiments presented in the manuscript, we evaluated the classifiers with G-Mean and Kappa for binary class scenarios and PMAUC and Kappa for multi-class scenarios. Metrics were calculated prequentially \citep{gama2013evaluating} using a sliding window of 500 examples. Complete results for all metrics (Accuracy, Kappa, G-Mean, AUC, PMAUC, WMAUC, and EWMAUC) are available on the website \url{https://people.vcu.edu/~acano/imbalanced-streams} for analysis and comparison with future works.

\section{Results}
\label{sec:results}

This section presents the experimental results from the set of benchmarks proposed to answer the research questions. Section~\ref{sec:bc-experiments} shows the experiments on binary class imbalanced streams. Section~\ref{sec:mc-experiments} shows the experiments on multi-class imbalanced streams. Finally, Section~\ref{sec:overall} shows overall results and an aggregated comparison of all algorithms.

Due to the very large number of experiments conducted in this work, we present in the manuscript a selection of the most representative results. The experiments are organized to show three levels of detail in the results. First, a more detailed comparison of the top five methods. Second, an aggregated comparison of the top ten methods. Third, a summary of the comparison among all methods. Complete results for all experiments on all algorithms, datasets/generators, and metrics are available on the website\footnote{Complete results for all experiments are available at \url{https://people.vcu.edu/~acano/imbalanced-streams}}.

\vspace*{-0.2cm}
\subsection{Binary class experiments}
\label{sec:bc-experiments}

The first set of experiments focuses on binary class problems with a positive minority class and a negative majority class. These experiments include static imbalance ratio, dynamic imbalance ratio, instance-level difficulties, concept drift and static imbalance ratio, concept drift and dynamic imbalance ratio, and real-world binary class imbalanced datasets.

\subsubsection{Static imbalance ratio}
\label{sec:bc-static-imb}

\begin{figure}[t!]
\centering
\includegraphics[width=0.19\columnwidth]{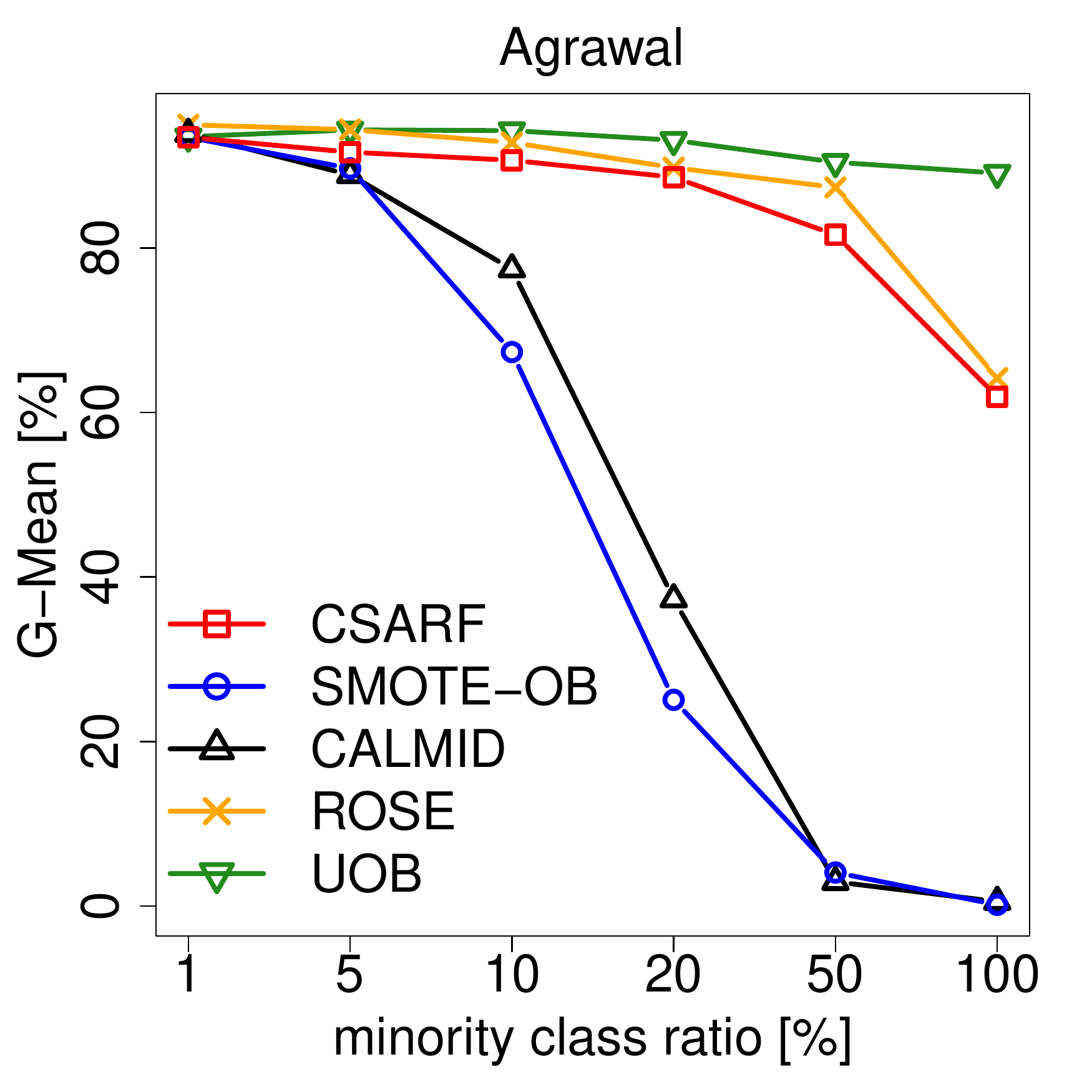}
\includegraphics[width=0.19\columnwidth]{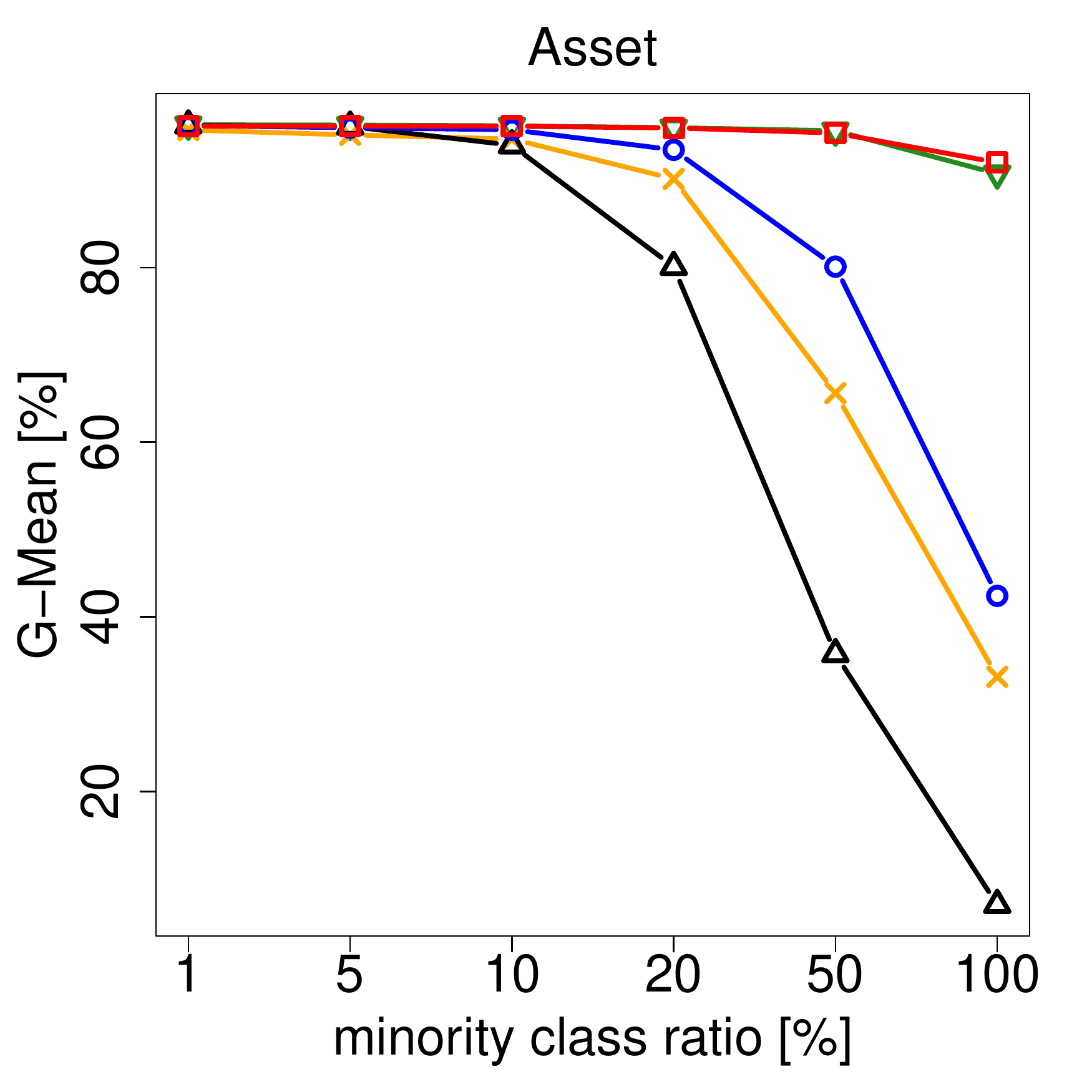}
\includegraphics[width=0.19\columnwidth]{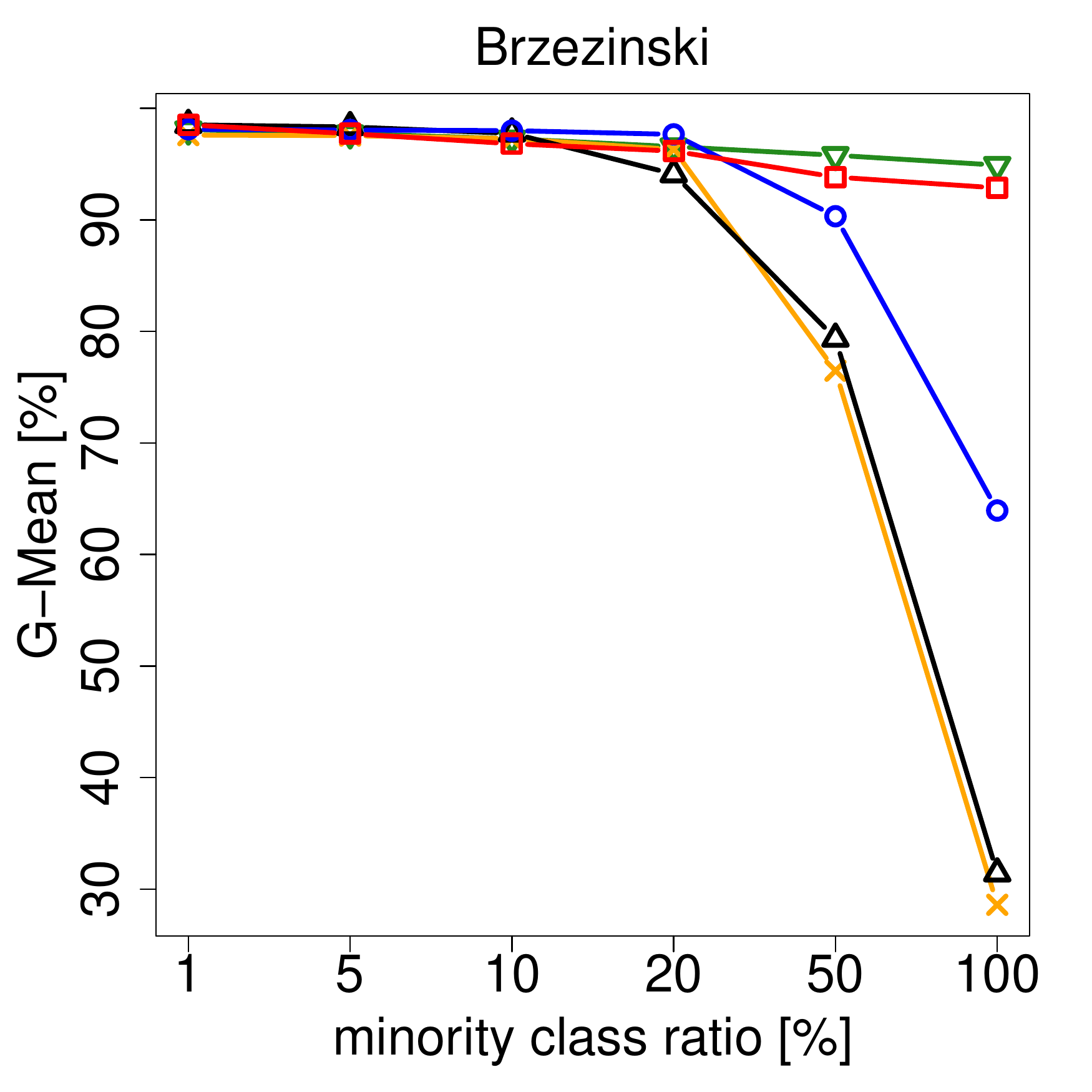}
\includegraphics[width=0.19\columnwidth]{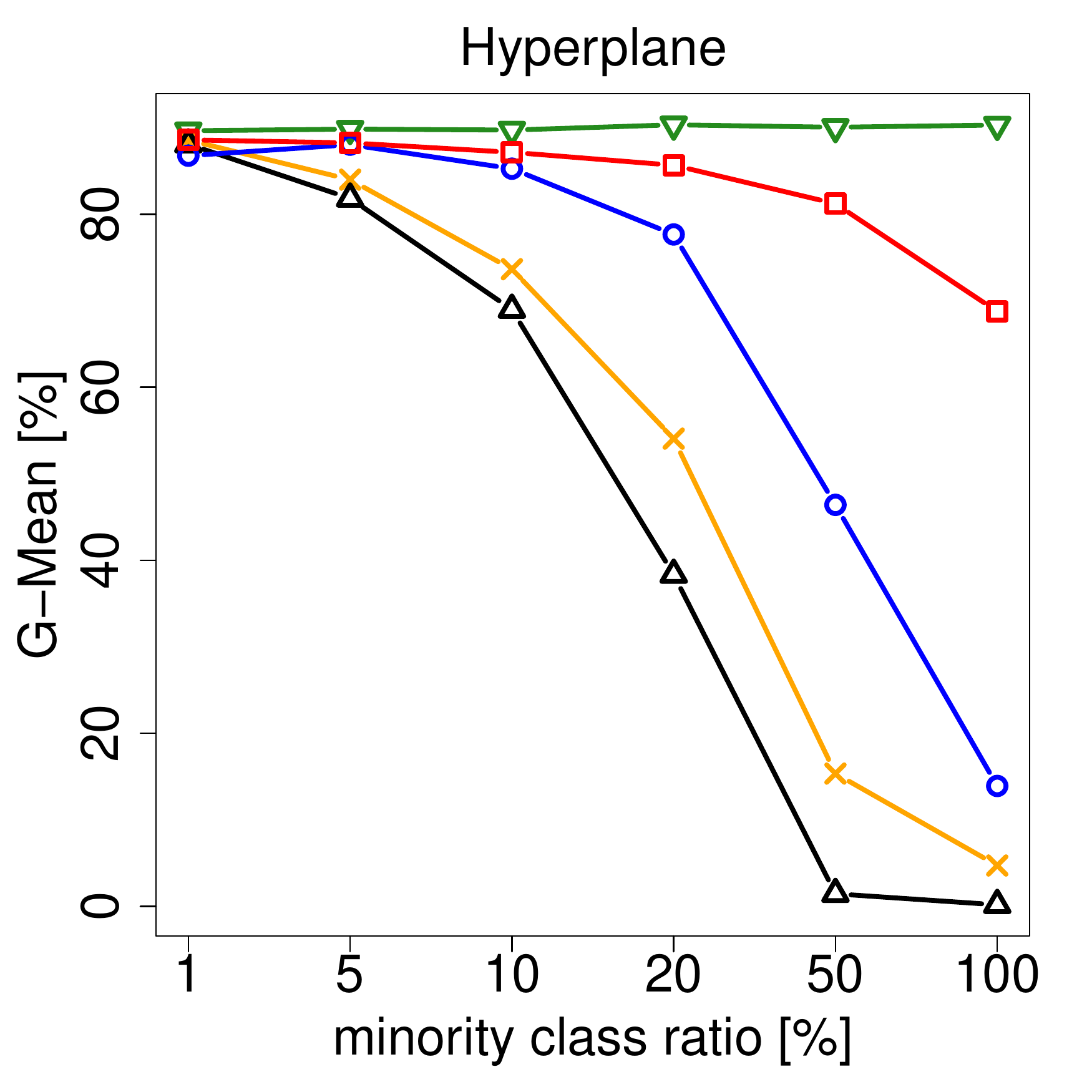}
\includegraphics[width=0.19\columnwidth]{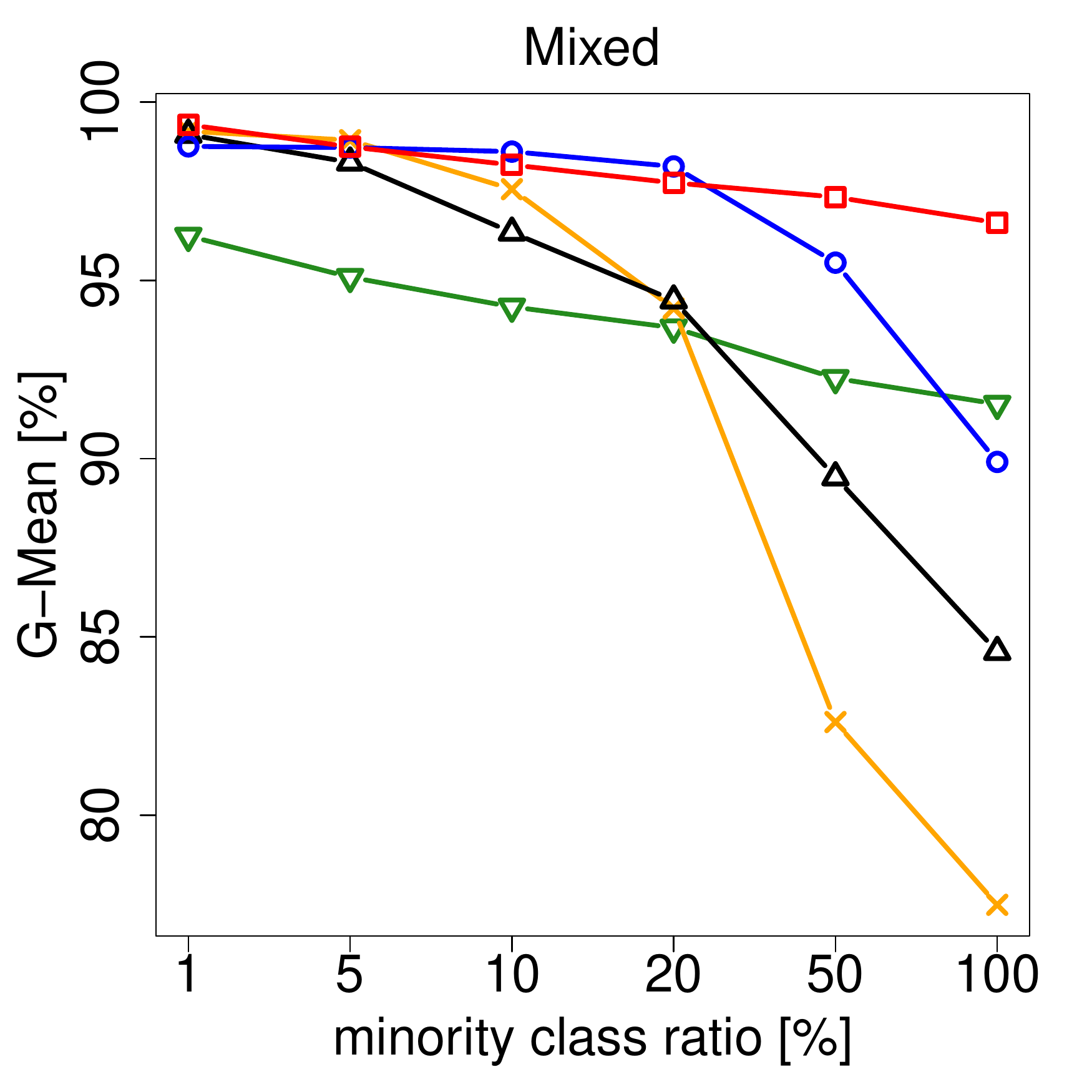}
\includegraphics[width=0.19\columnwidth]{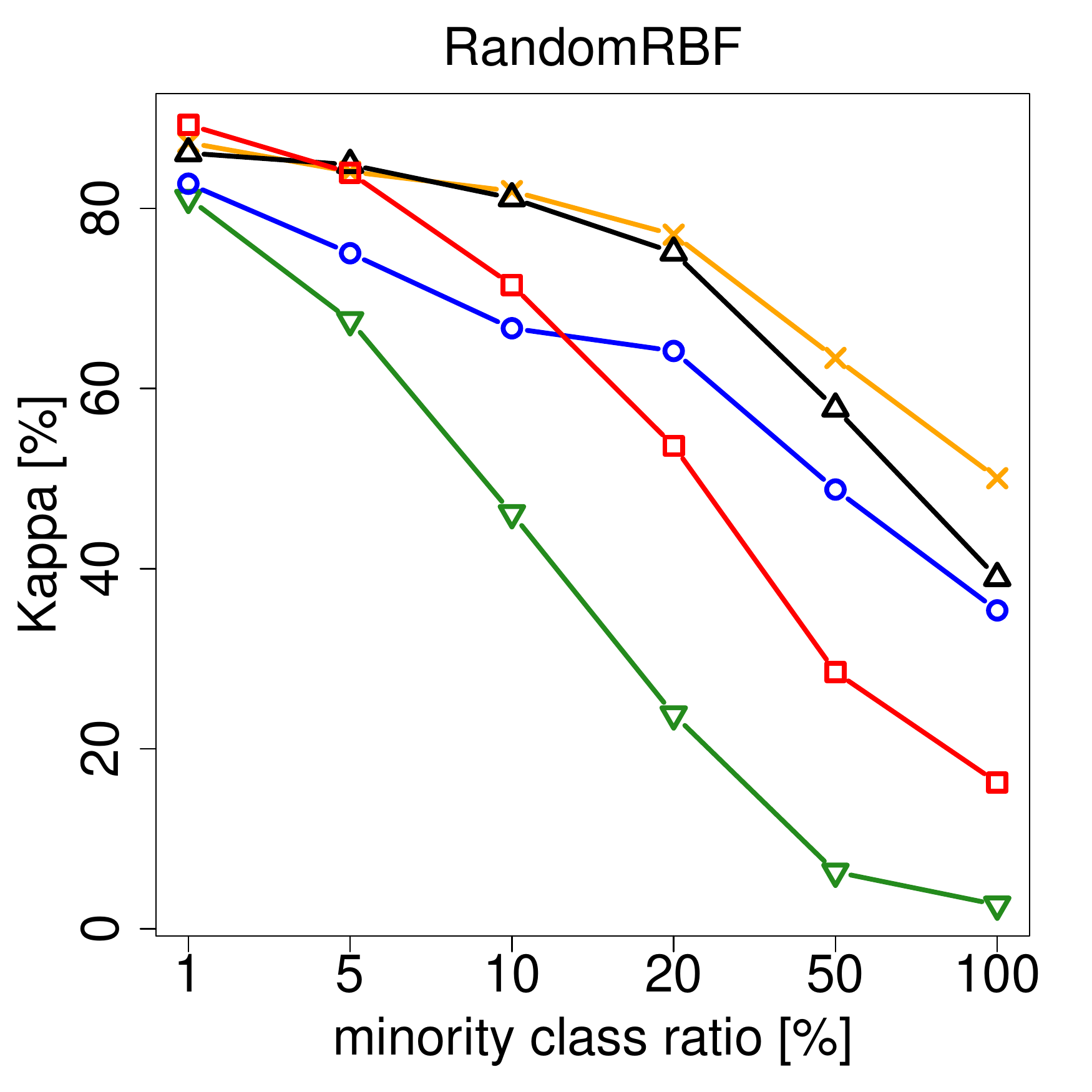}
\includegraphics[width=0.19\columnwidth]{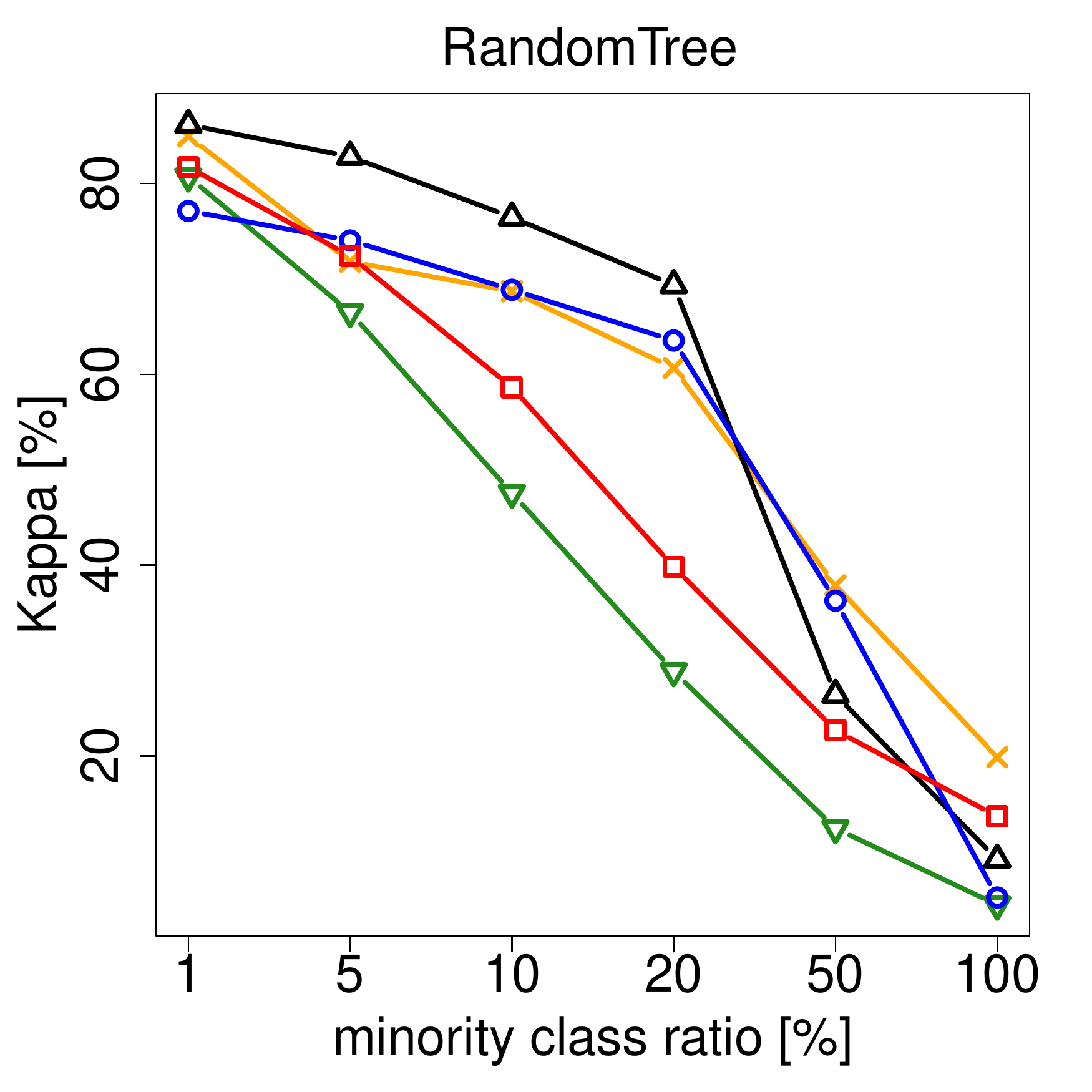}
\includegraphics[width=0.19\columnwidth]{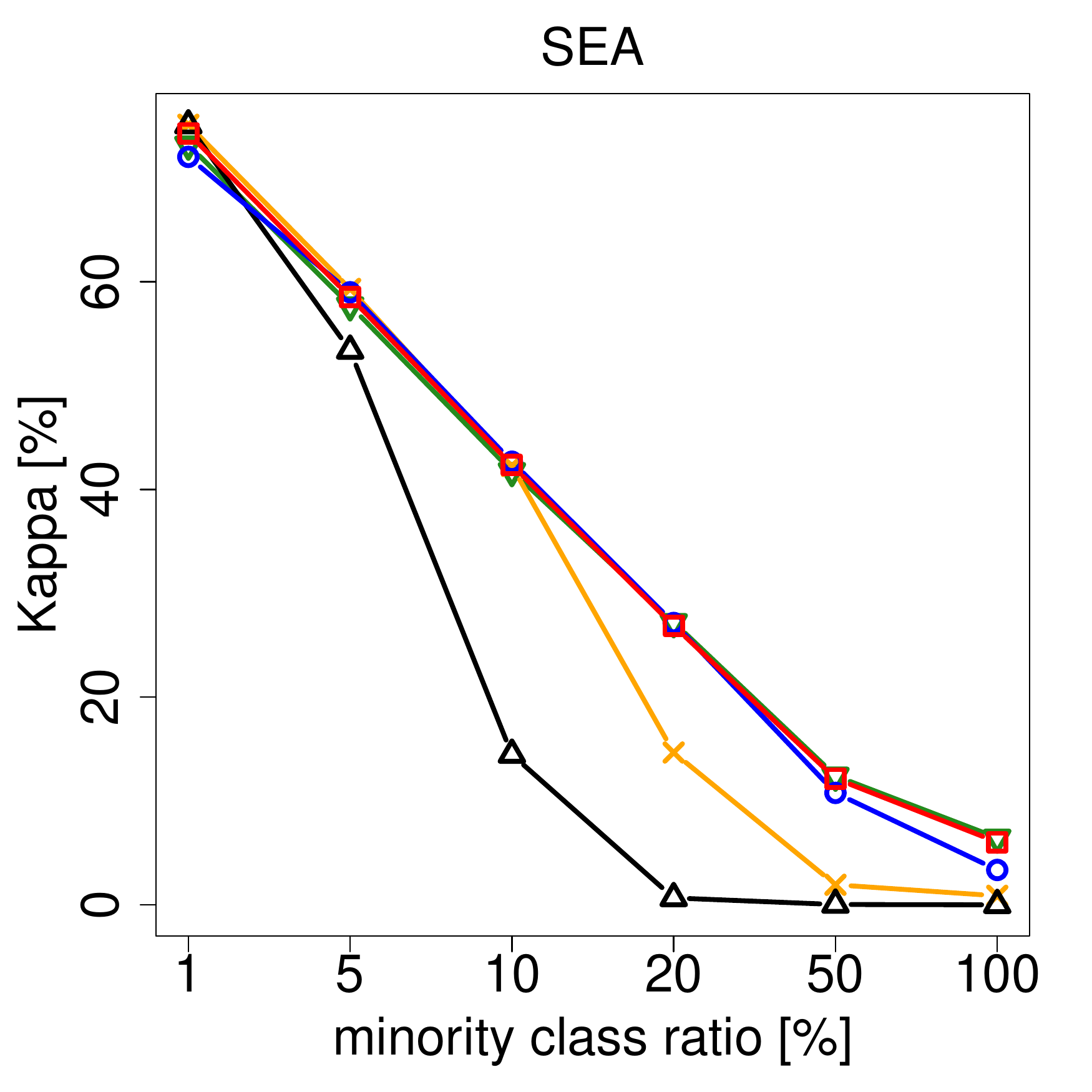}
\includegraphics[width=0.19\columnwidth]{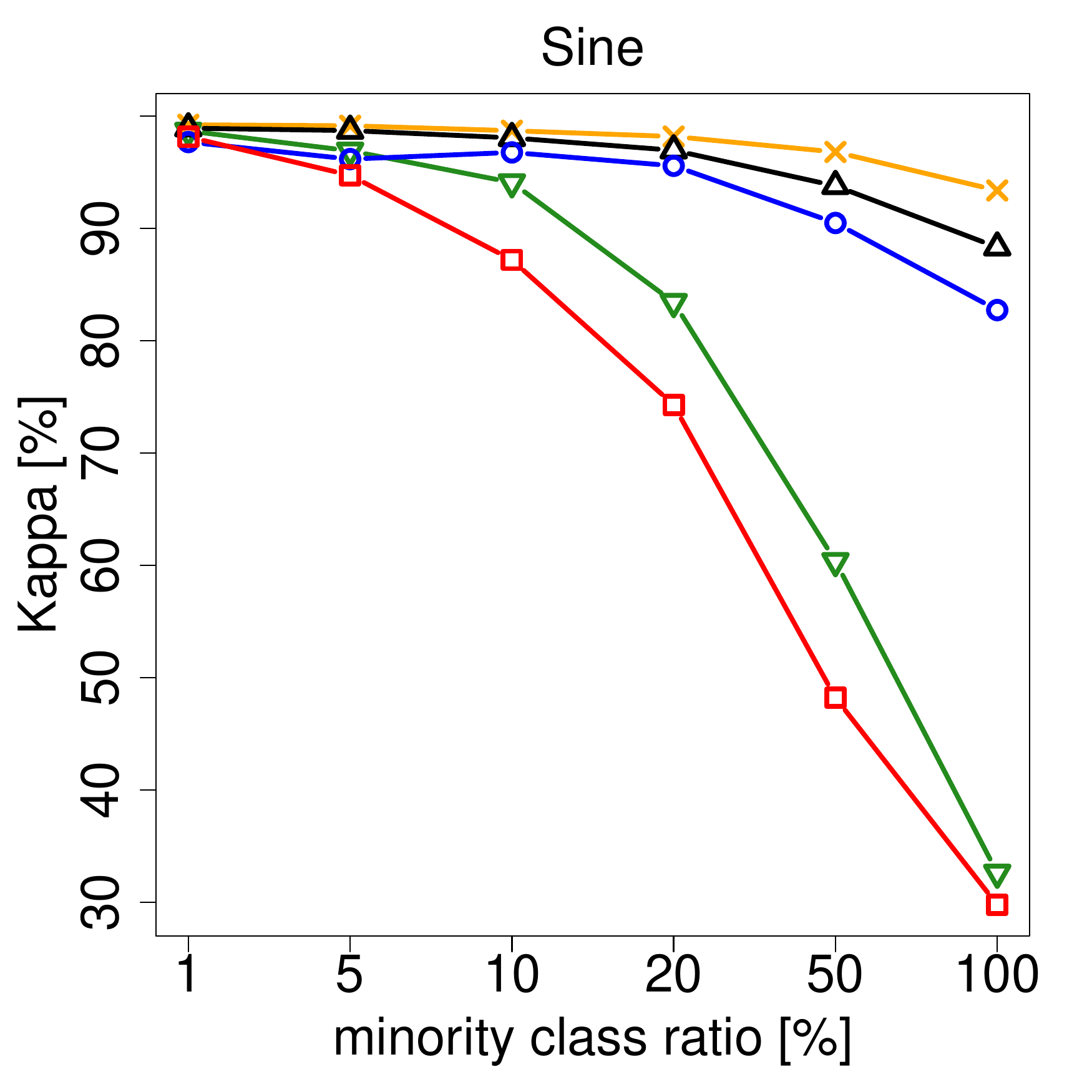}
\includegraphics[width=0.19\columnwidth]{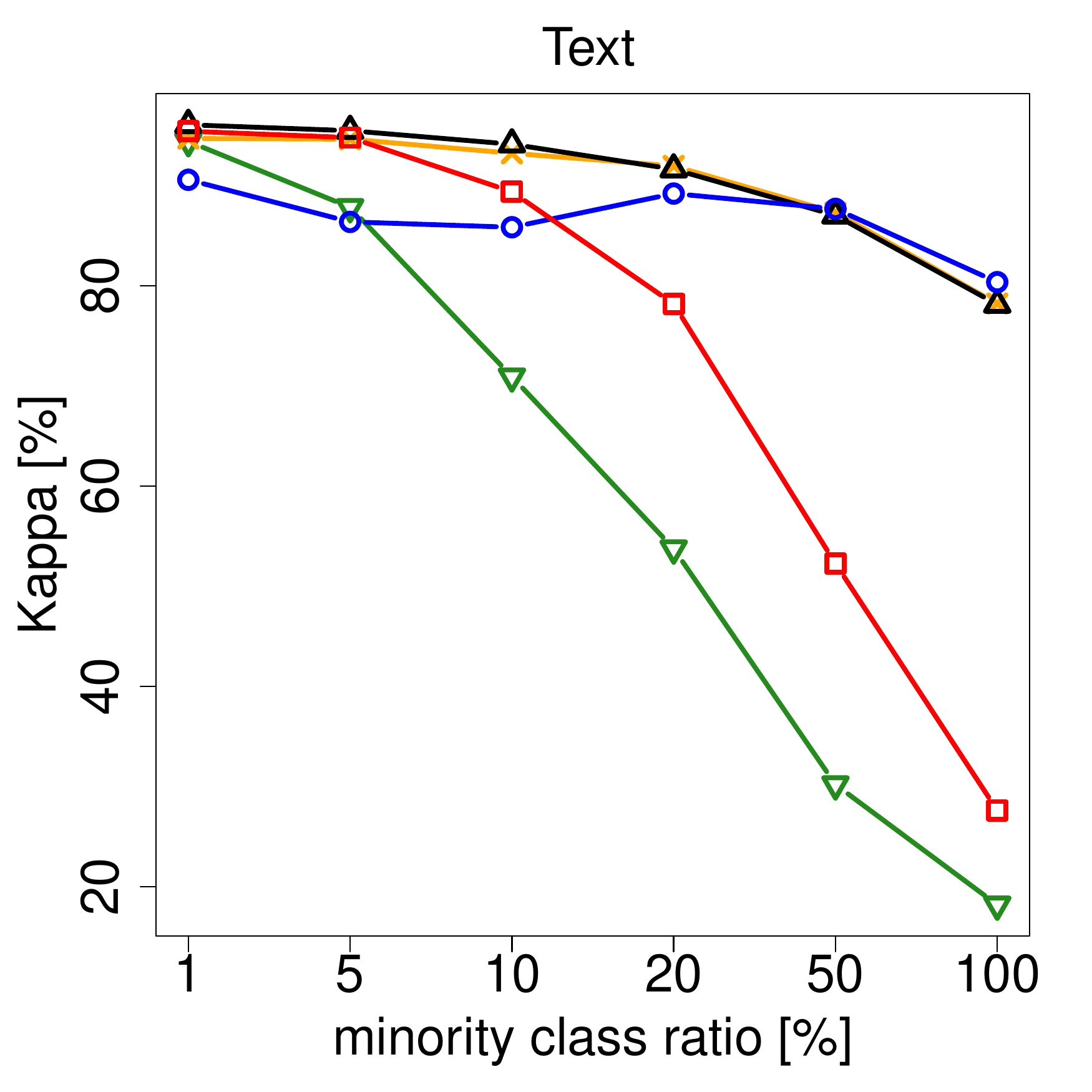}
\caption{Robustness to different levels of static class imbalance ratio (G-Mean and Kappa).}
\label{fig:BC_SIR}
\vspace{0.7cm}
\end{figure}

\begin{figure}[t!]
\centering
\includegraphics[width=\columnwidth]{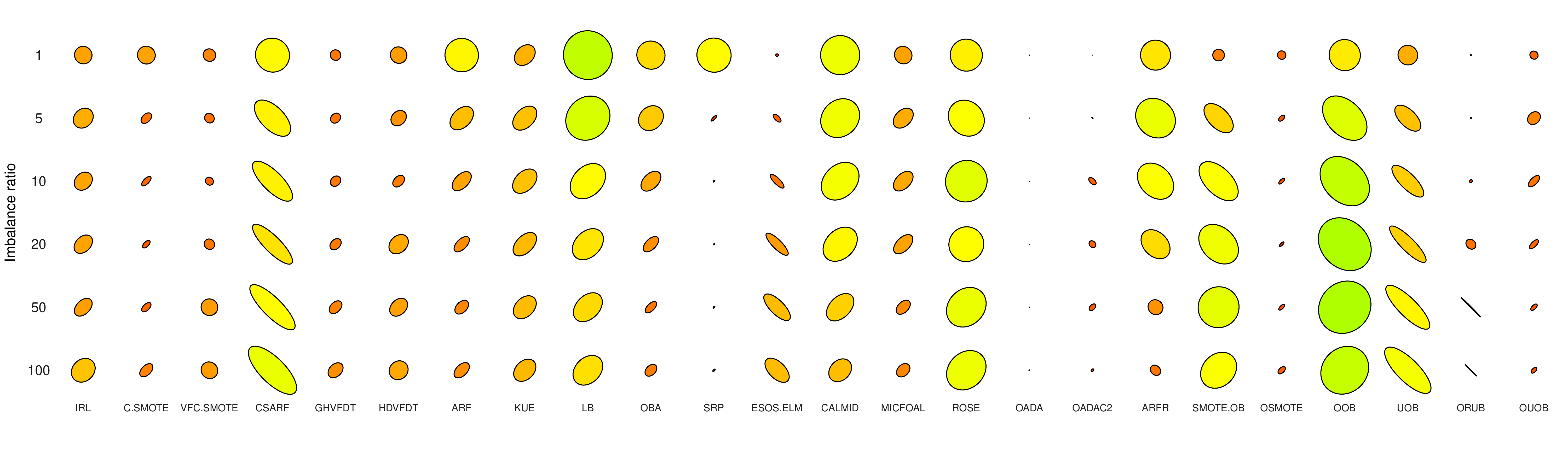}
\caption{Comparison of all 24 algorithms for different levels of static class imbalance ratio. The axes of the ellipse represent G-Mean and Kappa metrics. The bigger the axes the better rank of the algorithm on the metrics. The more rounded the ellipse the more agreement between the metrics. The color gradient represents the product of both metrics' ranks.}
\label{fig:BC_SIR_ellipse}
\vspace{0.7cm}
\end{figure}

\noindent \textbf{Goal of the experiment.} This experiment was designed to address \textbf{RQ1} and evaluate the robustness of the classifiers under different levels of static class imbalance without concept drift. It is expected that classifiers that were designed to tackle class imbalance will present better robustness to different levels of imbalance, i.e., to achieve a stable performance regardless of the imbalance ratio. To evaluate this, we prepared the generators presented in Table~\ref{tab:generators} with static imbalance ratios (ratio of the size of the majority class to the minority class as defined by~\cite{zhu2020adjusting}) of \{5, 10, 20, 50, 100\}.
This allows us to assess how each classifier performs under specific levels of class imbalance. Figure~\ref{fig:BC_SIR} illustrates the performance of five selected algorithms with increasing levels of static imbalance ratio. Table~\ref{tab:BC_SIR} presents the average G-Mean and Kappa for the top 10 classifiers for each of the evaluated imbalance ratios and the overall rank of the algorithms. Figure~\ref{fig:BC_SIR_ellipse} provides a comparison of all algorithms for each level of imbalance ratio. Axes of the ellipse represent G-Mean and Kappa metrics. The bigger the axes the better rank of the algorithm on the metrics. The more rounded the ellipse the more agreement between the metrics. Finally, the color represents a gradient of the product of the two metrics' ranks - red (worse) to green (better).

\noindent \textbf{Discussion}

\noindent \textit{Impact of approach to class imbalance.} First, we will analyze the impact of different skew-insensitive mechanisms used by analyzed ensembles on their robustness to various levels of static imbalance under stationary assumptions. Looking at resampling-based methods we can observe a clear distinction between methods based on blind and informative approaches. Ensembles utilizing blind approaches usually drop their performance with an increasing imbalance ratio. Taking \acrshort{uob} as an example, one can see discrepancies between G-mean and Kappa metrics. For G-mean \acrshort{uob} maintains its predictive performance, to the point that for very high imbalance ratios it outperforms other approaches. However, for the Kappa metric, we can see that performance of \acrshort{uob} deteriorates significantly with each increase in class disproportions. This shows that \acrshort{uob} produces a good true positive ratio but proportionally a larger number of false positives. We can explain that by the limitations of undersampling to extreme class imbalance, as to balance the current distribution one must aggressively discard majority instances. In static problems the higher the disproportion between classes, the higher chance of discarding relevant majority examples. However, in a streaming setting, we analyze the imbalance ratio in an online manner, thus \acrshort{uob} is not able to counter the bias towards the majority class accumulated over time by undersampling incoming instances one by one.  Its counterpart \acrshort{oob} shows the opposite behavior, returning best results for Kappa metric. Additionally, for high imbalance ratios \acrshort{oob} starts displaying balanced performance on both metrics. This shows that blind oversampling in online scenarios are capable of better and faster countering of bias accumulated over time. From informative resampling methods, we can observe that only \acrshort{smoteob} returns satisfactory performance. For the Kappa metric, it can outperform \acrshort{uob} but does not fare well against \acrshort{oob}. All other algorithms that use SMOTE-based resampling perform even worse. This allows us to conclude that blind oversampling performs best from all data-level mechanisms in terms of robustness to static imbalance.

\begin{table*}[t!]
\centering
\footnotesize
\setlength{\tabcolsep}{4pt}
\caption{G-Mean and Kappa averages of all 10 streams on static class imbalance ratio.}
\label{tab:BC_SIR}
\begin{tabular}{ll|C{1cm}C{1cm}C{1cm}C{1cm}C{1cm}C{1cm}C{1cm}C{1cm}C{1cm}C{1cm}}
\toprule
& IR & CSARF & ARF & KUE & LB & CALMID & ROSE & ARFR & SMOTE-OB & OOB & UOB\\
\midrule
\multirow{6}{*}{\rotatebox[origin=c]{90}{G-Mean}}
& 1 & 94.54 & 94.54 & 94.10 & \textbf{95.09} & 94.68 & 94.65 & 94.57 & 93.30 & 94.15 & 93.77\\
& 5 & \textbf{94.12} & 86.79 & 90.28 & 92.10 & 91.87 & 92.76 & 93.68 & 93.44 & 93.92 & 93.07\\
& 10 & \textbf{93.48} & 72.95 & 79.20 & 80.38 & 83.09 & 89.87 & 88.59 & 90.55 & 92.98 & 92.16\\
& 20 & \textbf{92.21} & 57.21 & 69.44 & 70.35 & 69.05 & 78.75 & 79.08 & 83.64 & 90.39 & 91.12\\
& 50 & 88.95 & 39.55 & 49.49 & 47.50 & 49.14 & 63.83 & 47.55 & 68.44 & 73.97 & \textbf{89.22}\\
& 100 & 82.75 & 30.14 & 36.55 & 36.41 & 35.17 & 46.82 & 14.50 & 45.10 & 60.99 & \textbf{86.78}\\
\midrule
\multirow{6}{*}{\rotatebox[origin=c]{90}{Kappa}}
& 1 & 89.13 & 89.12 & 88.76 & \textbf{90.24} & 89.42 & 89.36 & 89.20 & 86.75 & 88.36 & 87.63\\
& 5 & 82.12 & 79.38 & 82.83 & \textbf{84.94} & 84.18 & 83.86 & 84.33 & 80.84 & 83.77 & 78.68\\
& 10 & 70.92 & 65.64 & 72.43 & 74.63 & 74.88 & \textbf{78.00} & 72.95 & 72.72 & 77.54 & 65.44\\
& 20 & 55.19 & 52.42 & 62.79 & 64.10 & 61.53 & 65.95 & 57.38 & 62.52 & \textbf{69.64} & 48.24\\
& 50 & 31.91 & 36.15 & 45.92 & 44.34 & 44.34 & 54.15 & 27.43 & 47.90 & \textbf{57.14} & 26.77\\
& 100 & 17.84 & 28.39 & 34.55 & 34.48 & 32.84 & 41.09 & 5.62 & 35.18 & \textbf{46.93} & 13.92\\
\midrule
\multicolumn{2}{l|}{Avg. G-Mean} & 91.01 & 63.53 & 69.85 & 70.30 & 70.50 & 77.78 & 69.66 & 79.08 & 84.40 & \textbf{91.02}\\
\multicolumn{2}{l|}{Avg. Kappa} & 57.85 & 58.52 & 64.55 & 65.45 & 64.53 & 68.73 & 56.15 & 64.32 & \textbf{70.56} & 53.45\\
\midrule
\multicolumn{2}{l|}{Rank G-Mean} & \textbf{3.08} & 8.08 & 7.87 & 5.69 & 6.32 & 5.15 & 5.93 & 5.25 & 3.51 & 4.12\\
\multicolumn{2}{l|}{Rank Kappa} & 6.40 & 6.49 & 6.00 & 4.11 & 4.77 & 4.33 & 6.32 & 6.00 & \textbf{3.75} & 6.83\\
\bottomrule
\end{tabular}
\vspace{0.7cm}
\end{table*}

Among the algorithm-level solutions, \acrshort{csarf} displays best results for the G-mean metric, outperforming all reference methods. However, it does not hold its performance when evaluated using Kappa. This is another striking example of discrepancies between those metrics and how they highlight different aspects of imbalanced classification. Alternative algorithm-level approaches, such as \acrshort{rose} and \acrshort{calmid}, while performing worse on G-mean, offer a more balanced performance on both metrics at once. Additionally, they display good robustness to increasing imbalance ratios. Therefore, algorithm selection for data streams with static imbalance is far from trivial, as one must choose between methods that perform very well only on one of the metrics, or choose a well-rounded method that, while not exceeding on any single metric, offers more even performance.  

Finally, out of standard ensembles with no skew-insensitive mechanisms, \acrshort{lb} returned the best predictive performance, outperforming several methods dedicated to imbalanced data streams. This did not hold for other methods, such as \acrshort{srp} or \acrshort{arf} that displayed no robustness to increasing imbalance ratios.

\noindent \textit{Impact of ensemble architecture.} When we look at the overall best-performing methods in every scenario, we can see a dominance of ensembles based on bagging or hybrid architectures. Bagging offers an easy and effective way of maintaining instance-based diversity among base learners that benefits both data and algorithm-level approaches and leads to high robustness under various levels of class imbalance. Within bagging methods, only \acrshort{ouob} can be seen as an outlier. We can explain this using our observations from the previous paragraph – that undersampling and oversampling offer contrary performance (one favoring G-mean and the other one Kappa). Therefore, by combining those two approaches we obtain an ensemble that is driven by two conflicting mechanisms. Boosting-based ensembles are usually the worst performing ones. We can explain this by the fact that boosting mechanism focuses on correcting the errors of the previous classifier in a chain. When dealing with high imbalance ratios the errors are driven by a small number of minority instances, leading to too small sample sizes to effectively improve the performance. As usually minority instances are misclassified, assigning high weights to them will lead to high error on the majority class by increasing the number of false positives. In the end, boosting-based ensembles will consist of classifiers biased towards one of the classes. Without proper selection or weighting mechanisms, it is impossible to maintain robustness to high imbalance ratios with such classifiers in the ensemble pool.

\subsubsection{Dynamic imbalance ratio}
\label{sec:bc-dyn-imb}

\begin{figure}[t!]
\centering
\includegraphics[width=0.19\columnwidth]{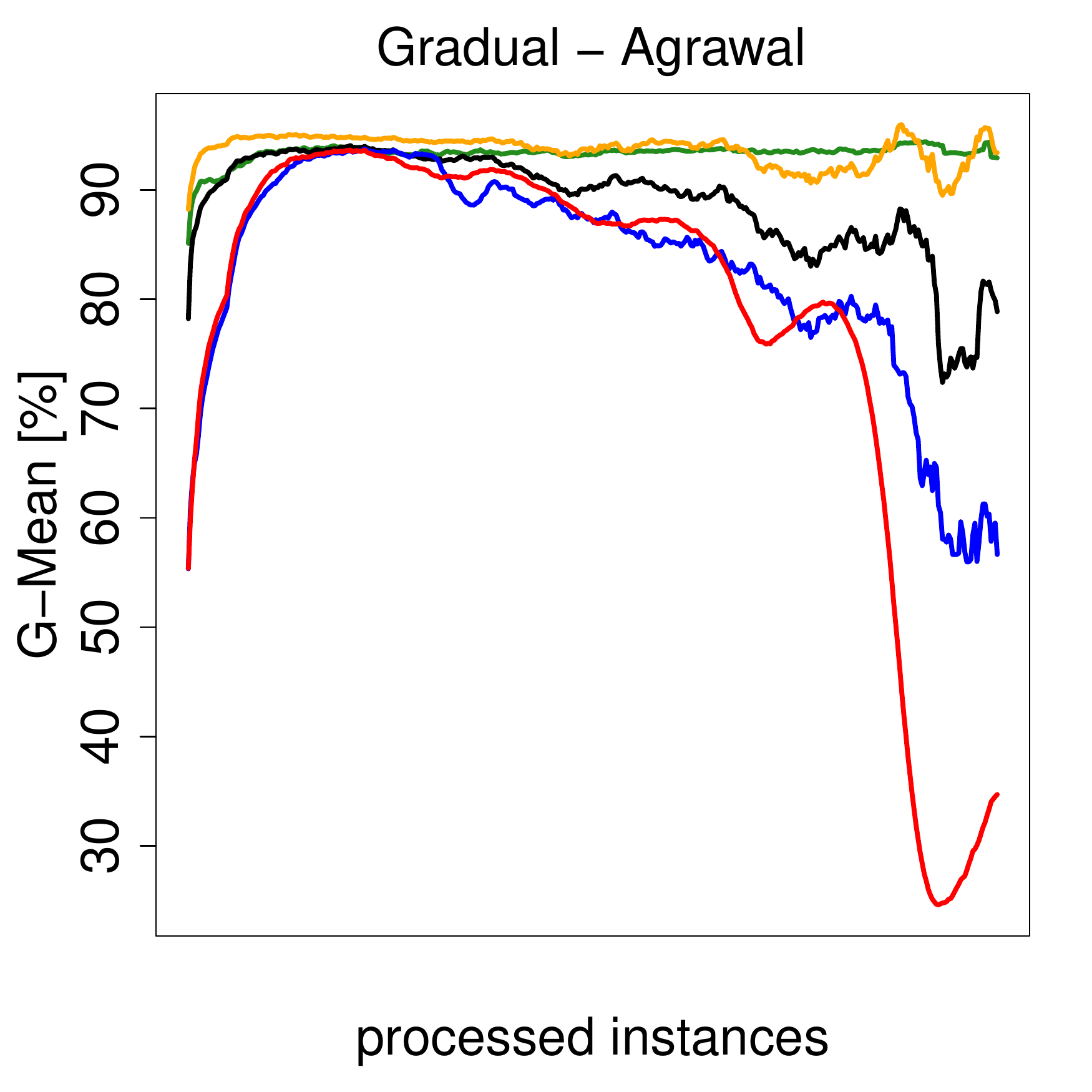}
\includegraphics[width=0.19\columnwidth]{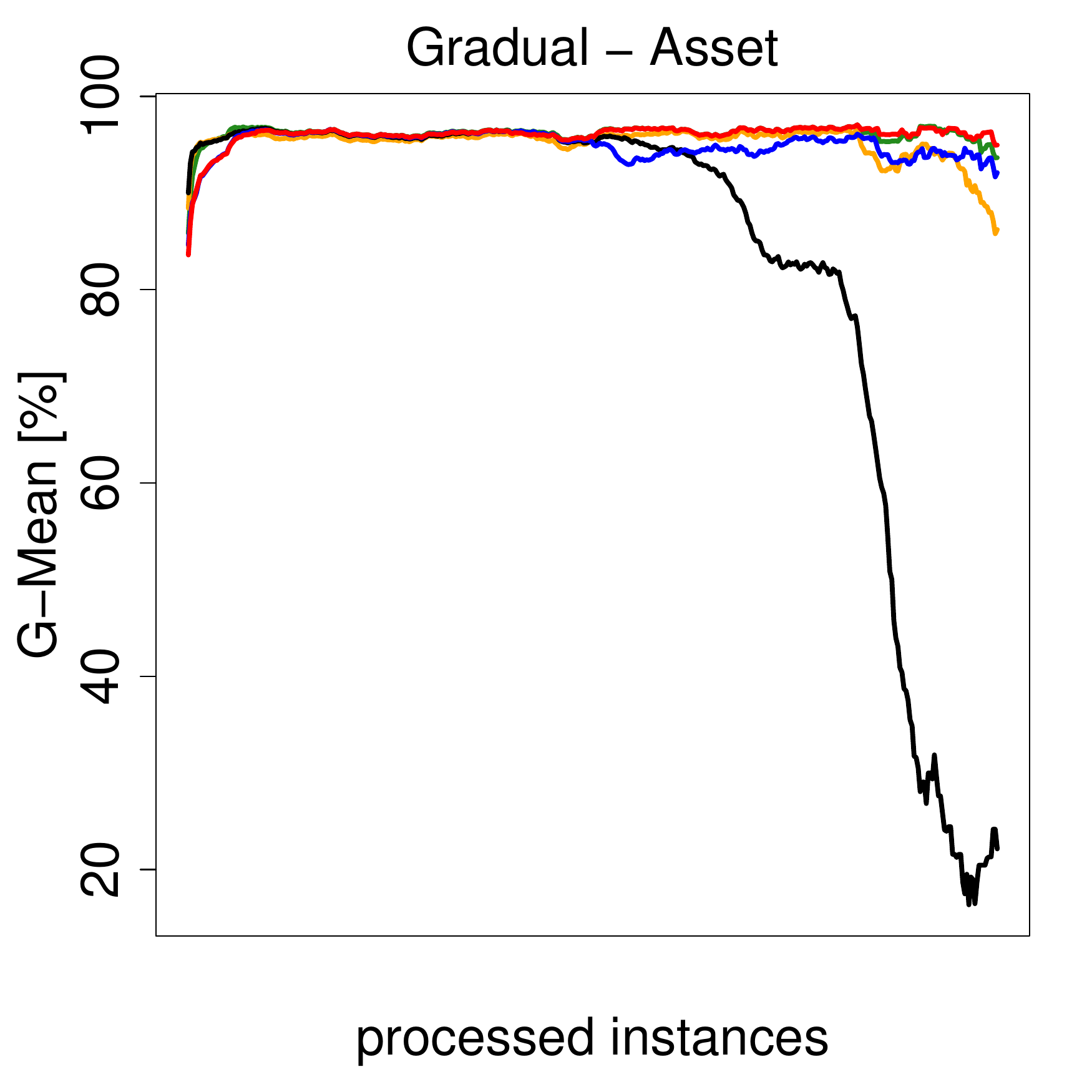}
\includegraphics[width=0.19\columnwidth]{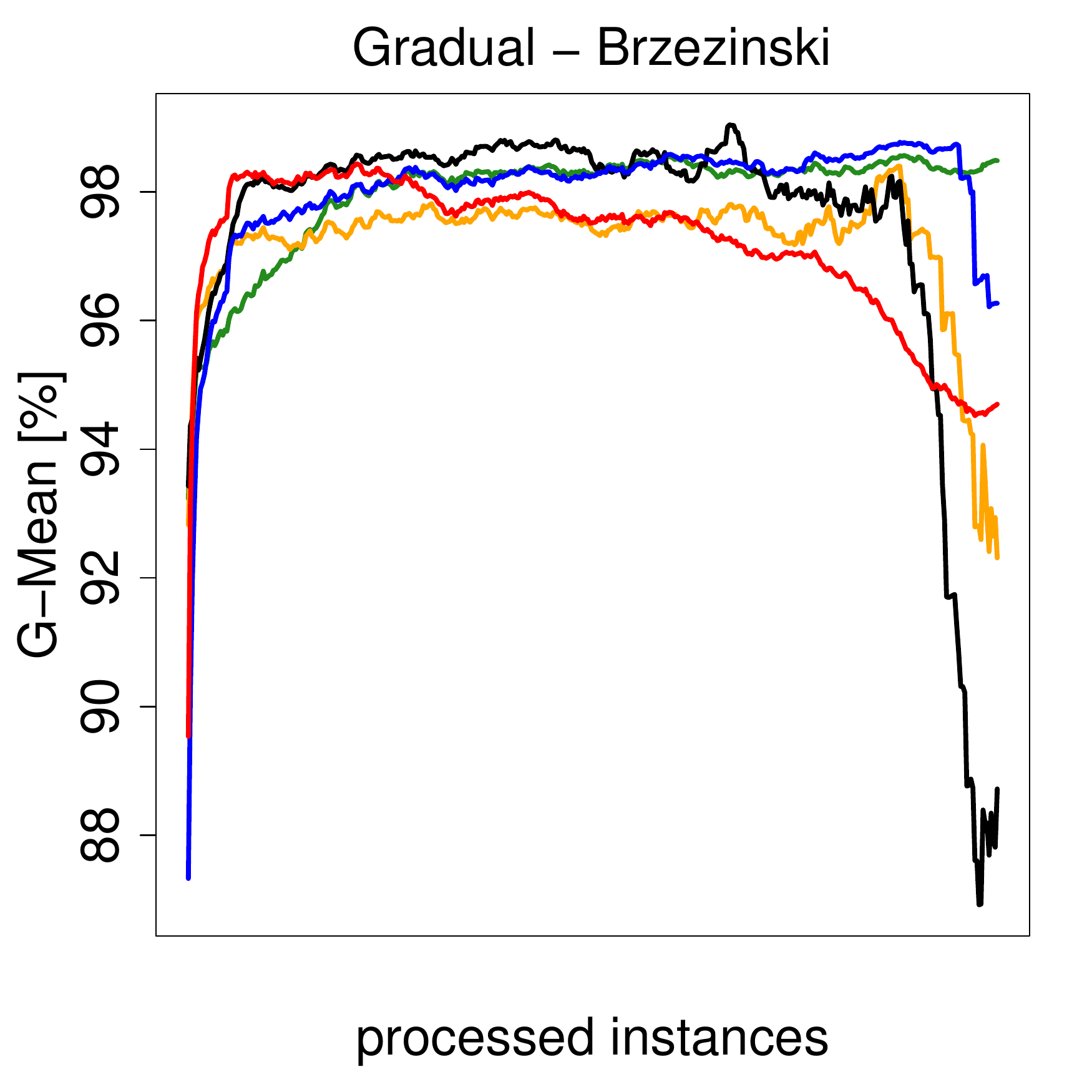}
\includegraphics[width=0.19\columnwidth]{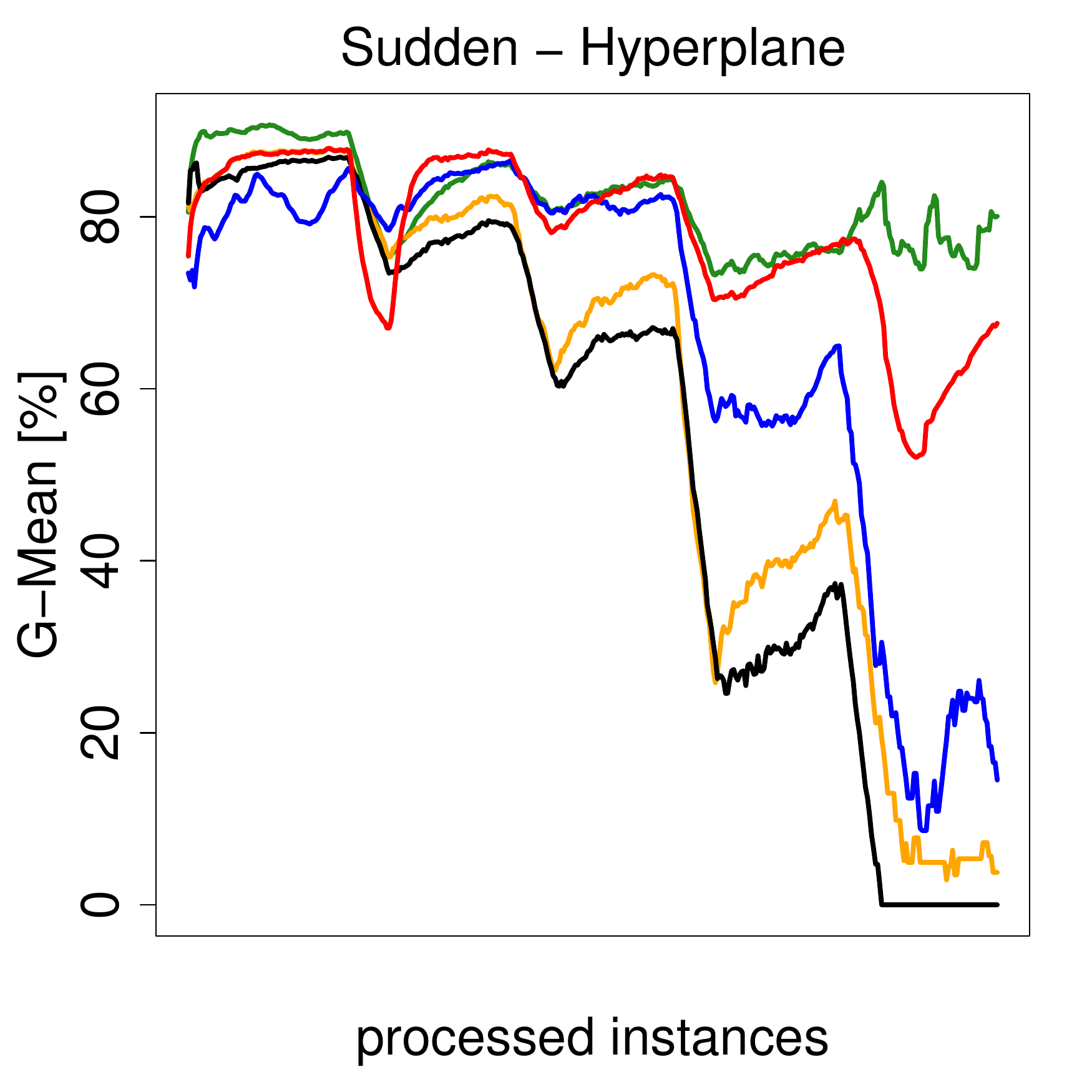}
\includegraphics[width=0.19\columnwidth]{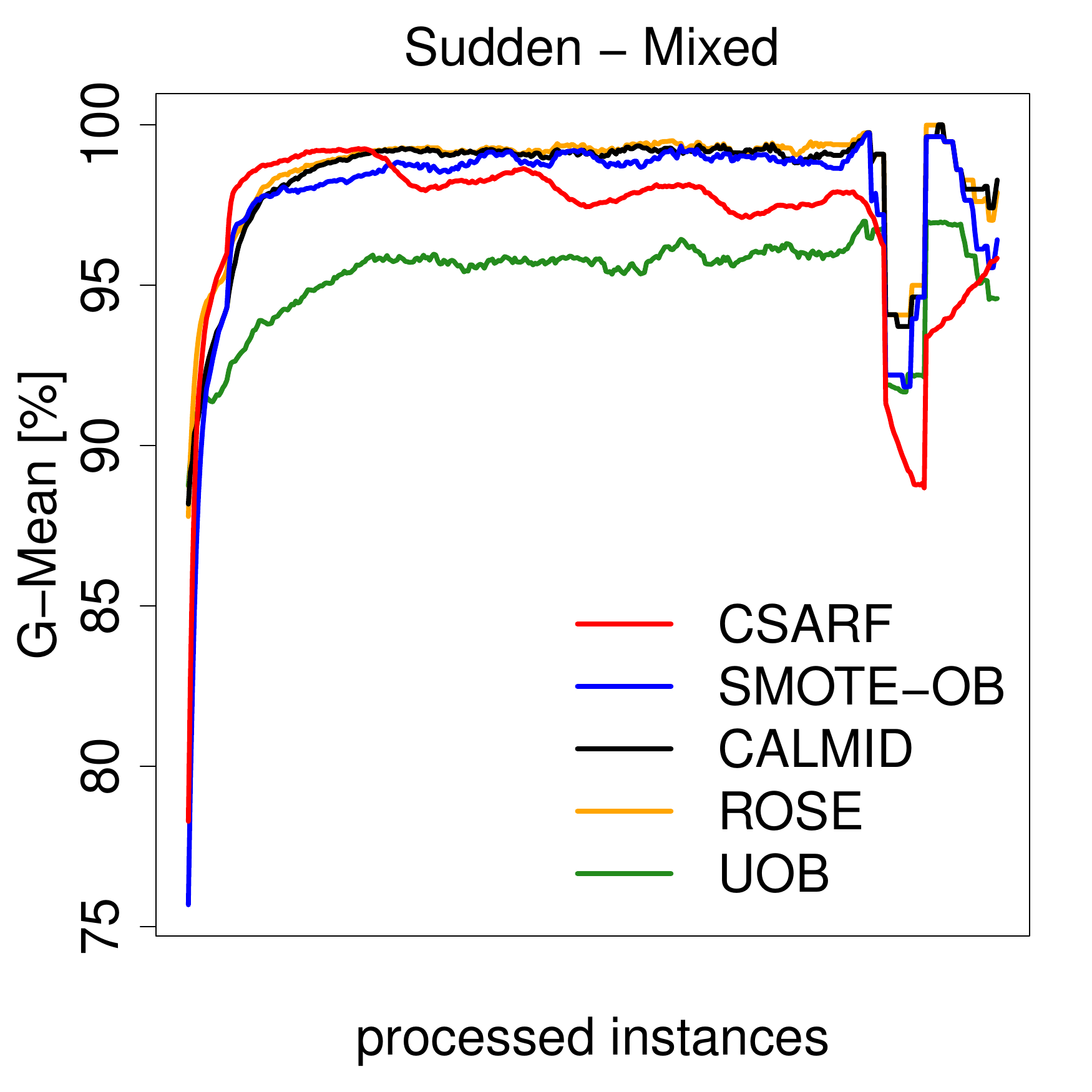}
\includegraphics[width=0.19\columnwidth]{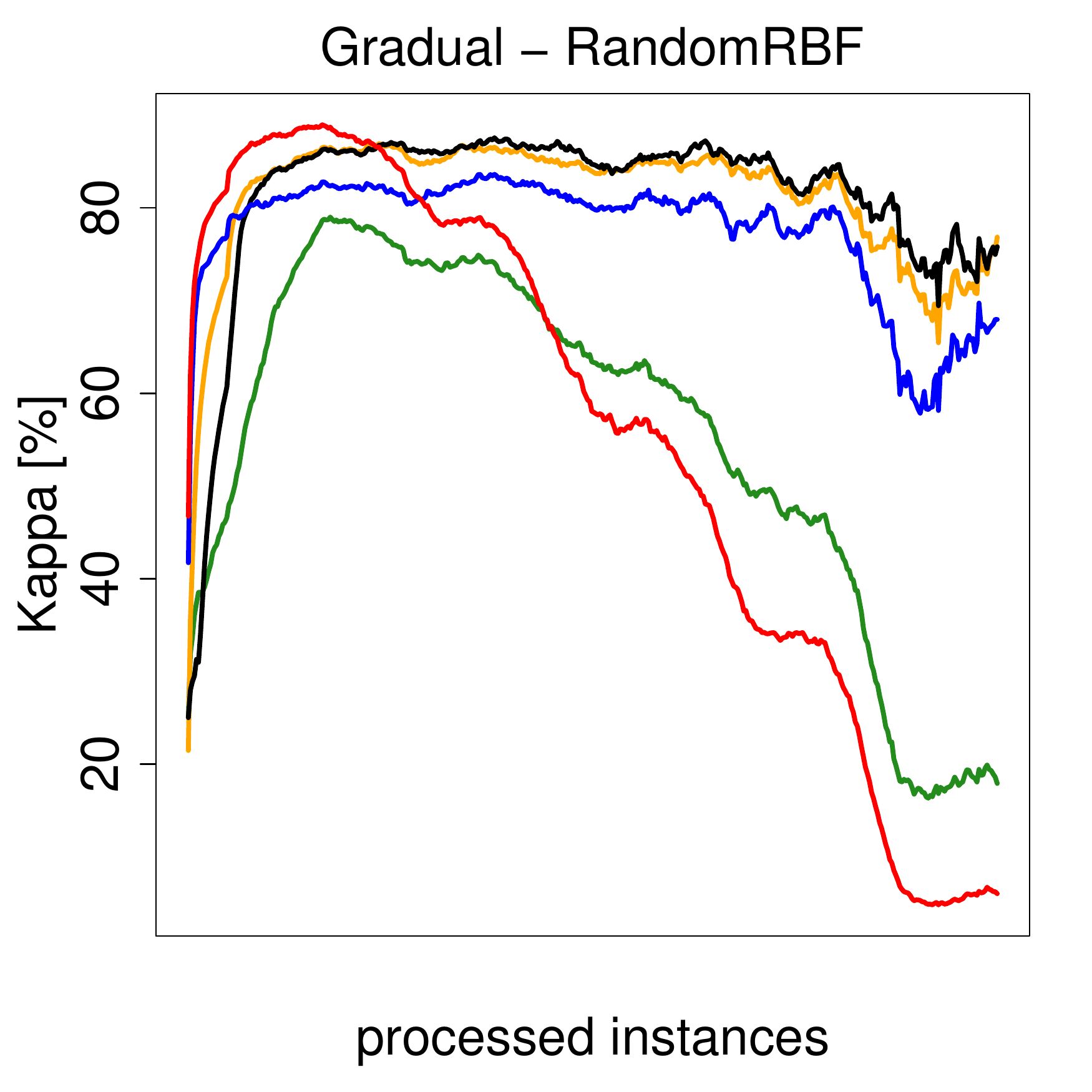}
\includegraphics[width=0.19\columnwidth]{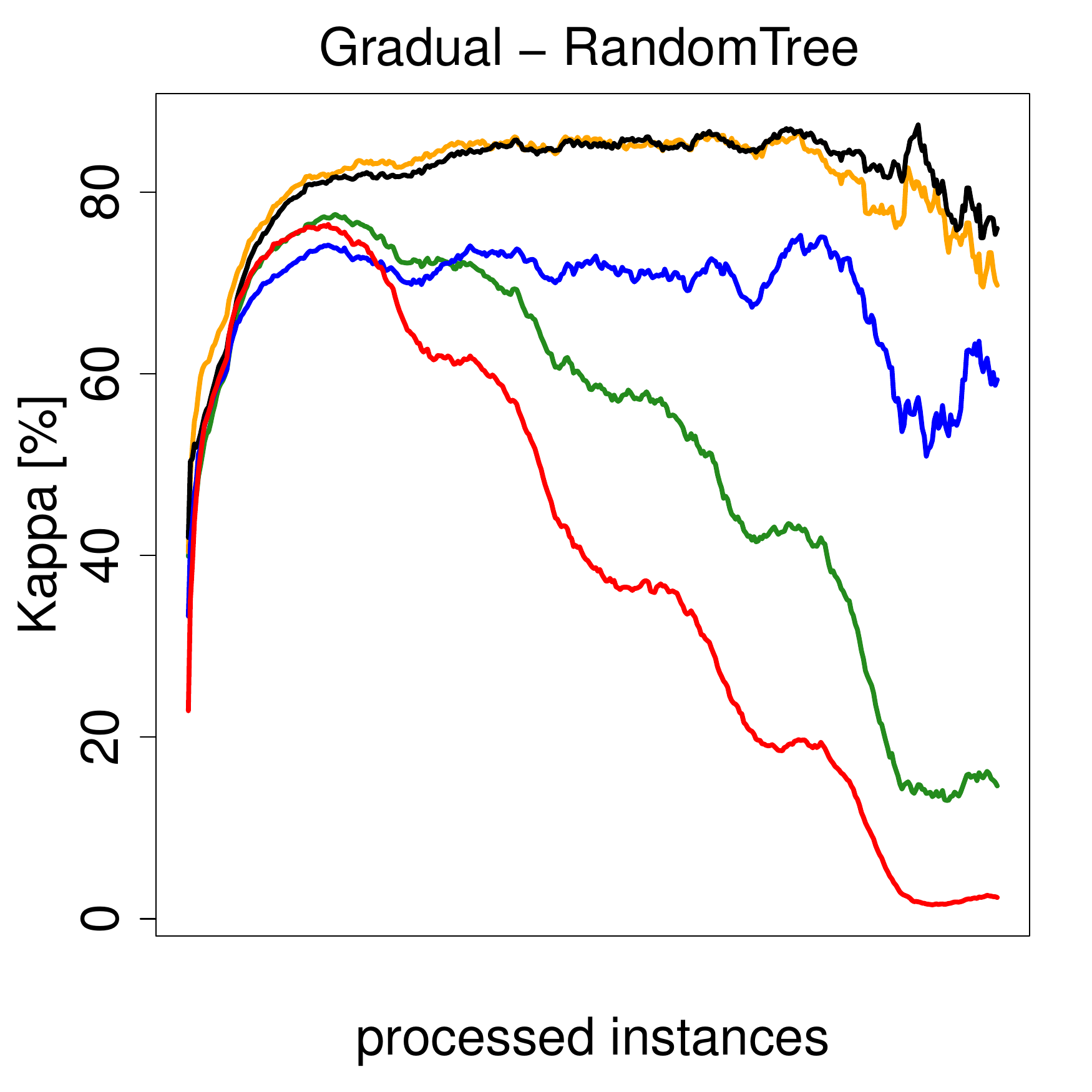}
\includegraphics[width=0.19\columnwidth]{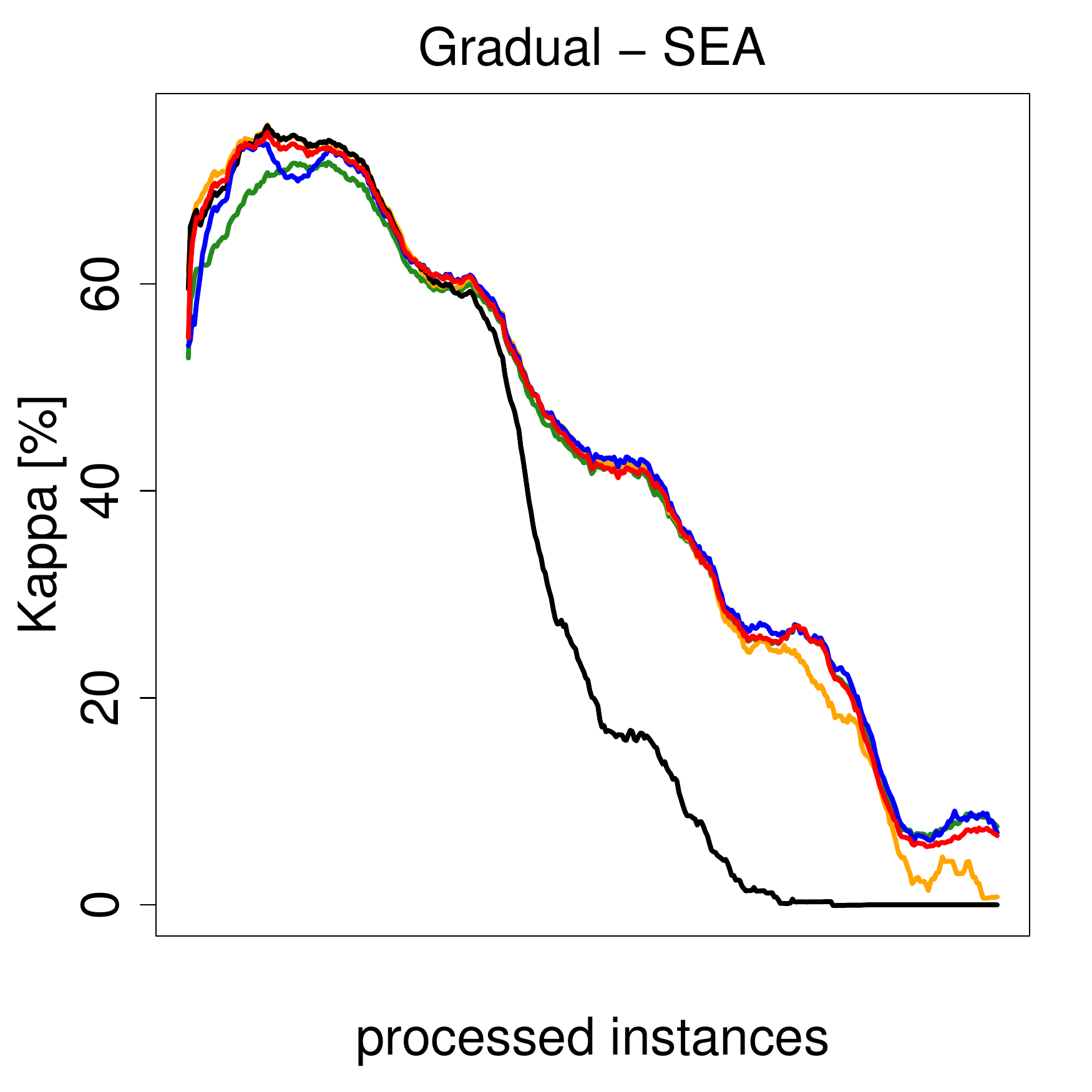}
\includegraphics[width=0.19\columnwidth]{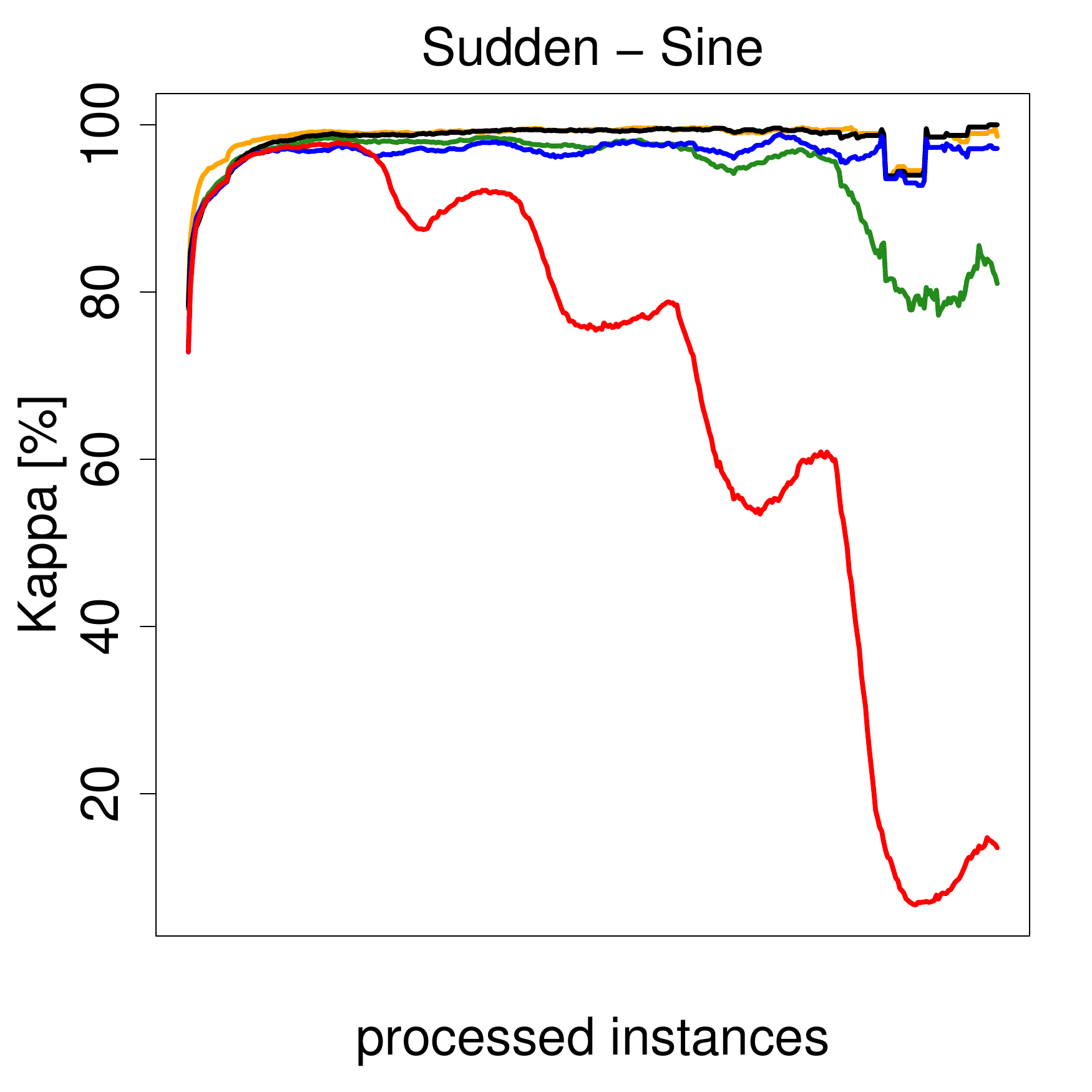}
\includegraphics[width=0.19\columnwidth]{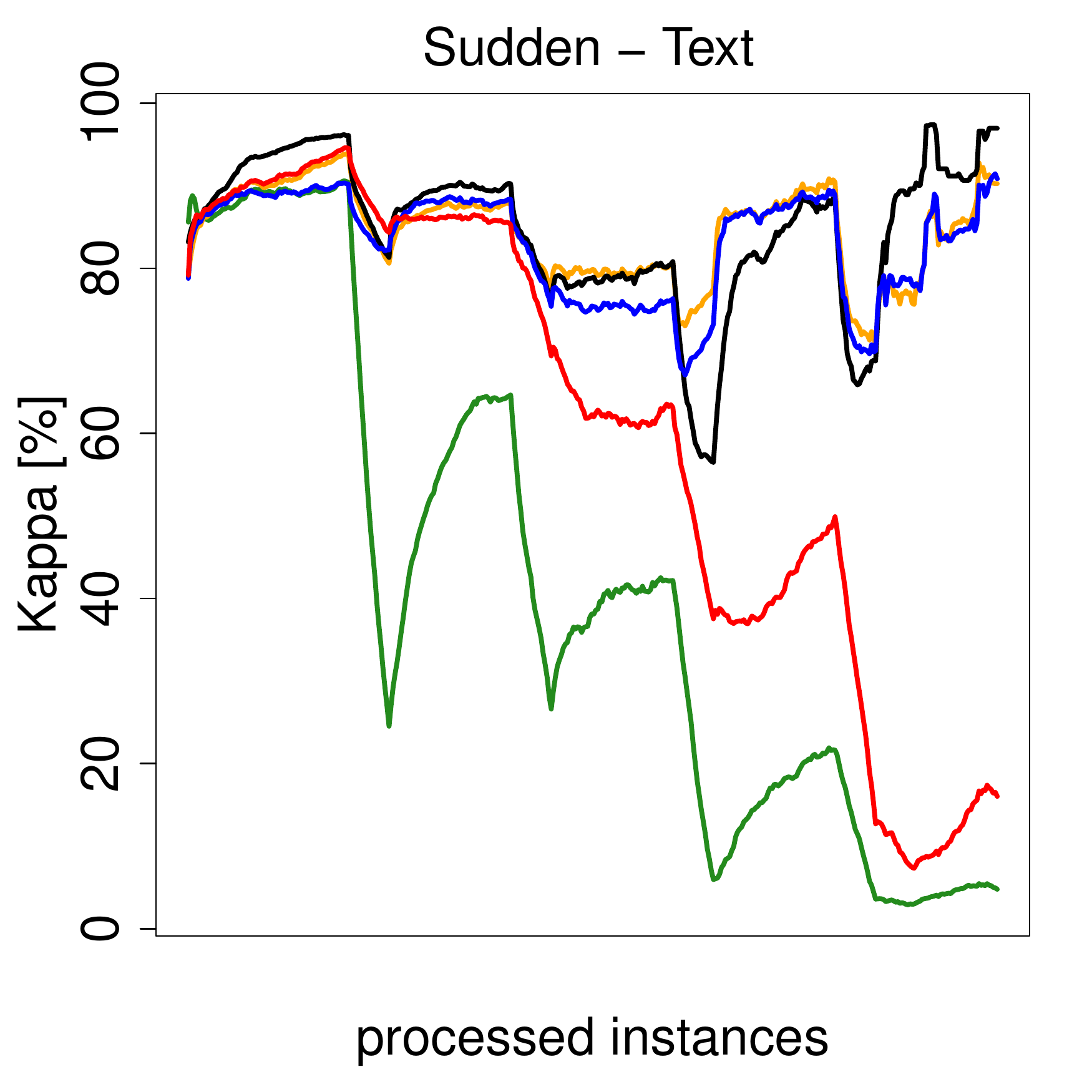}
\caption{G-Mean and Kappa on increasing class imbalance ratio with gradual and sudden drift.}
\label{fig:ir_increasing_study}
\end{figure}

\begin{figure}[t!]
\centering
\includegraphics[width=0.19\columnwidth]{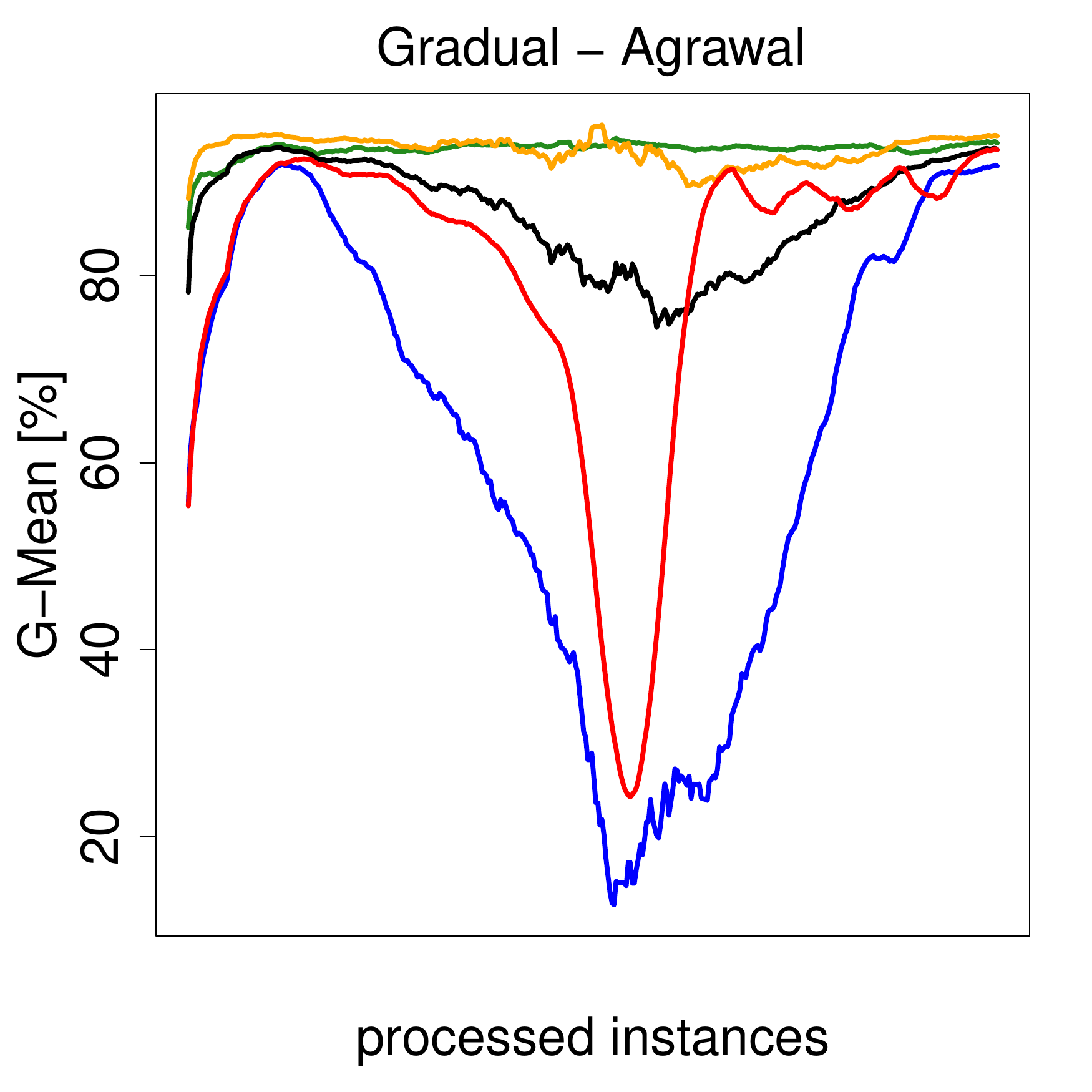}
\includegraphics[width=0.19\columnwidth]{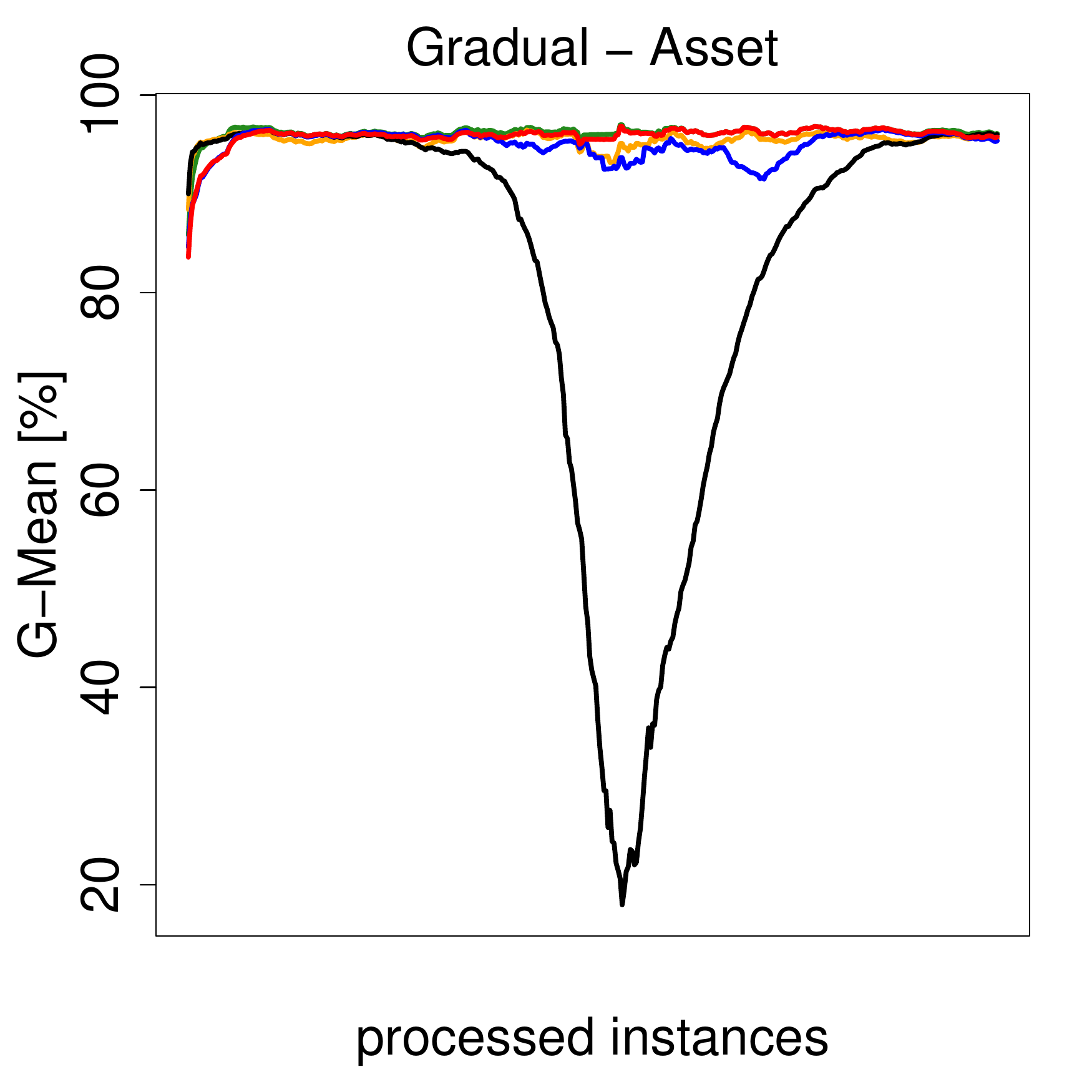}
\includegraphics[width=0.19\columnwidth]{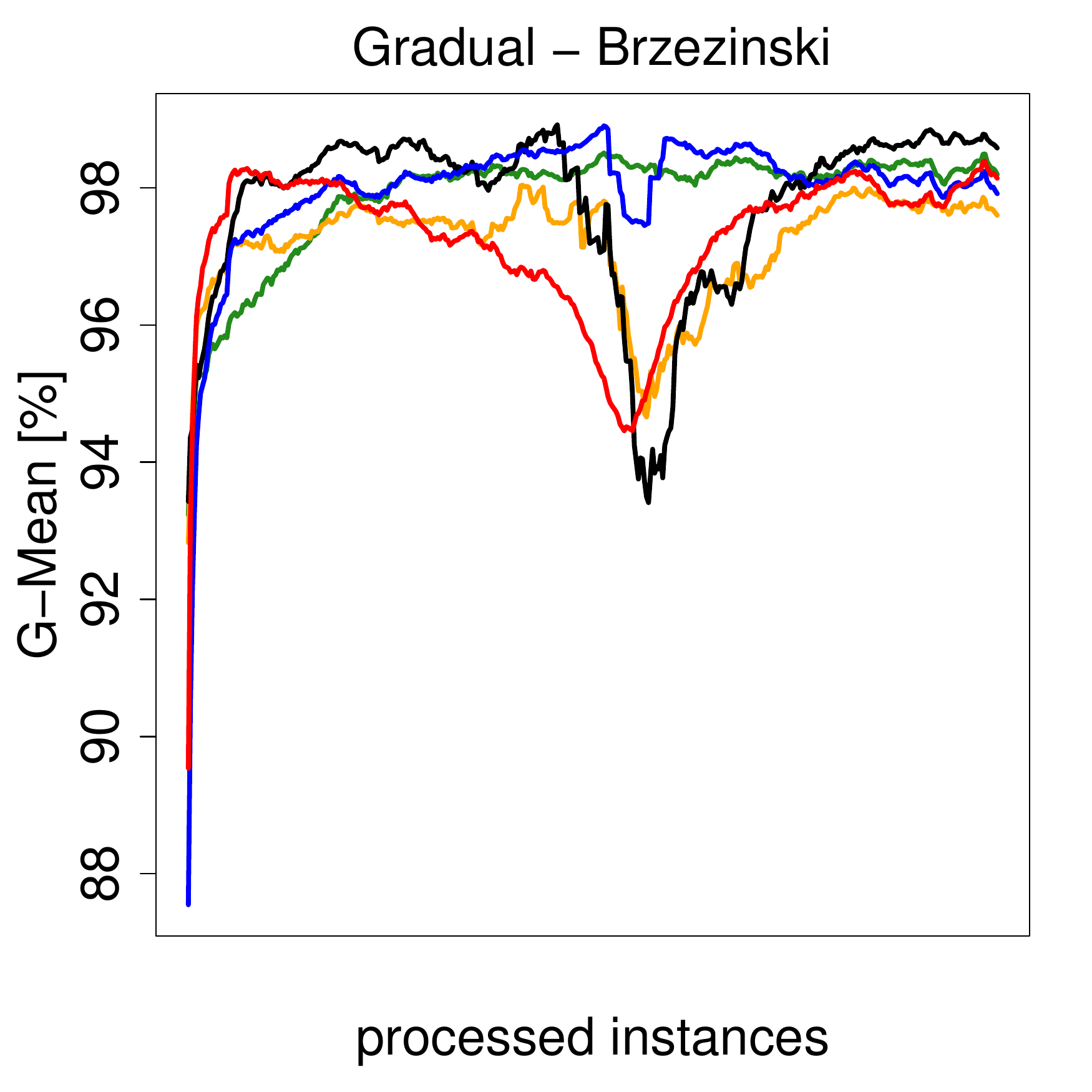}
\includegraphics[width=0.19\columnwidth]{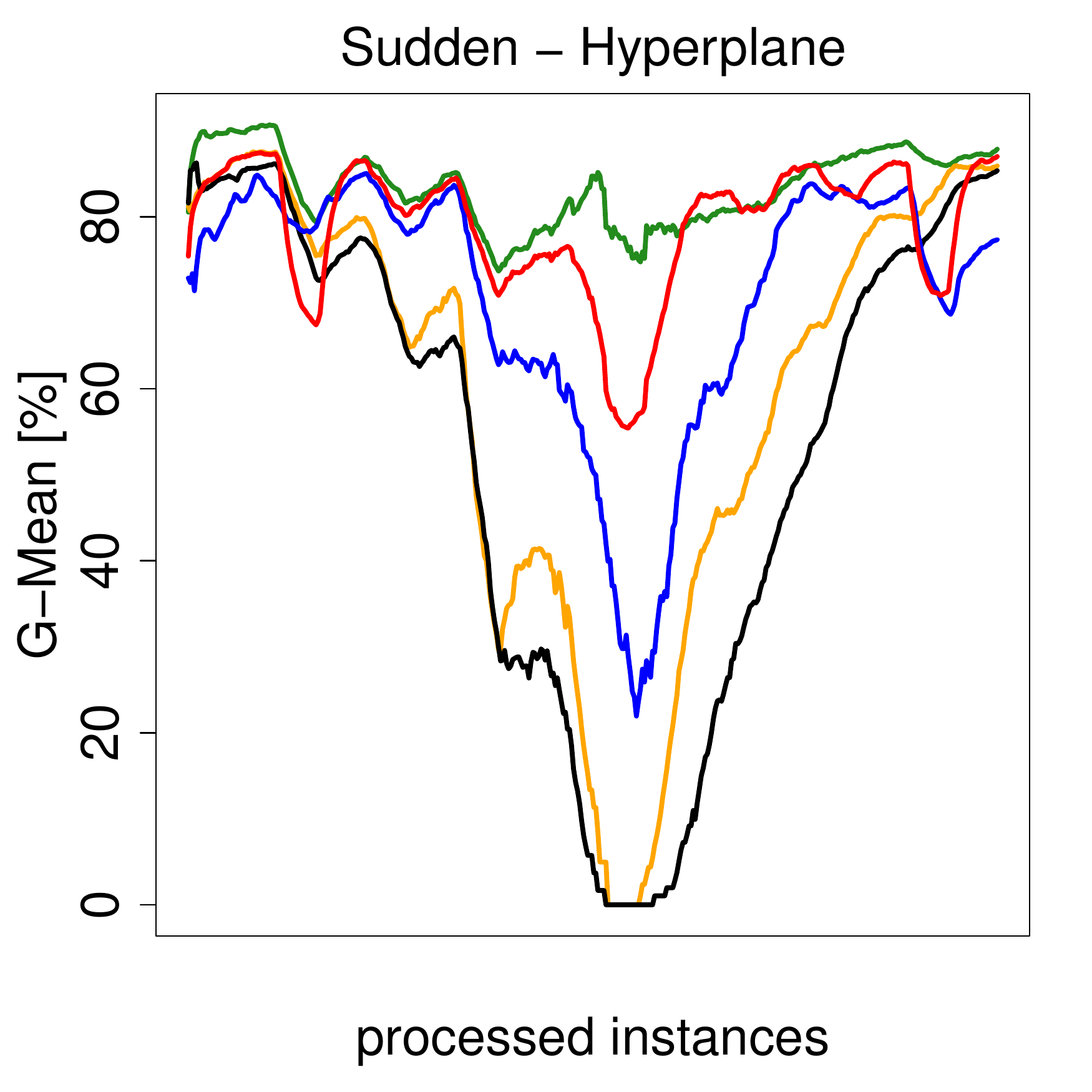}
\includegraphics[width=0.19\columnwidth]{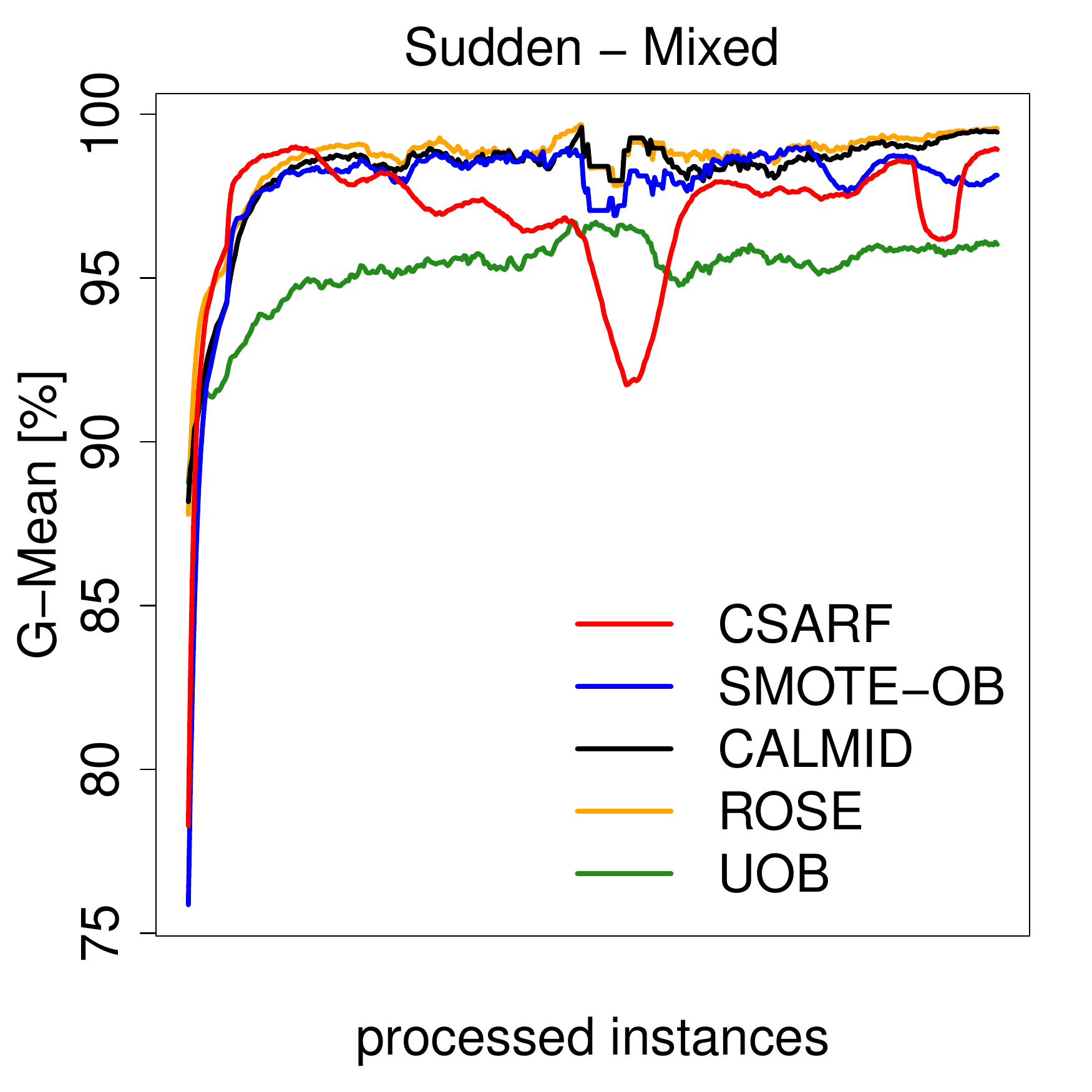}
\includegraphics[width=0.19\columnwidth]{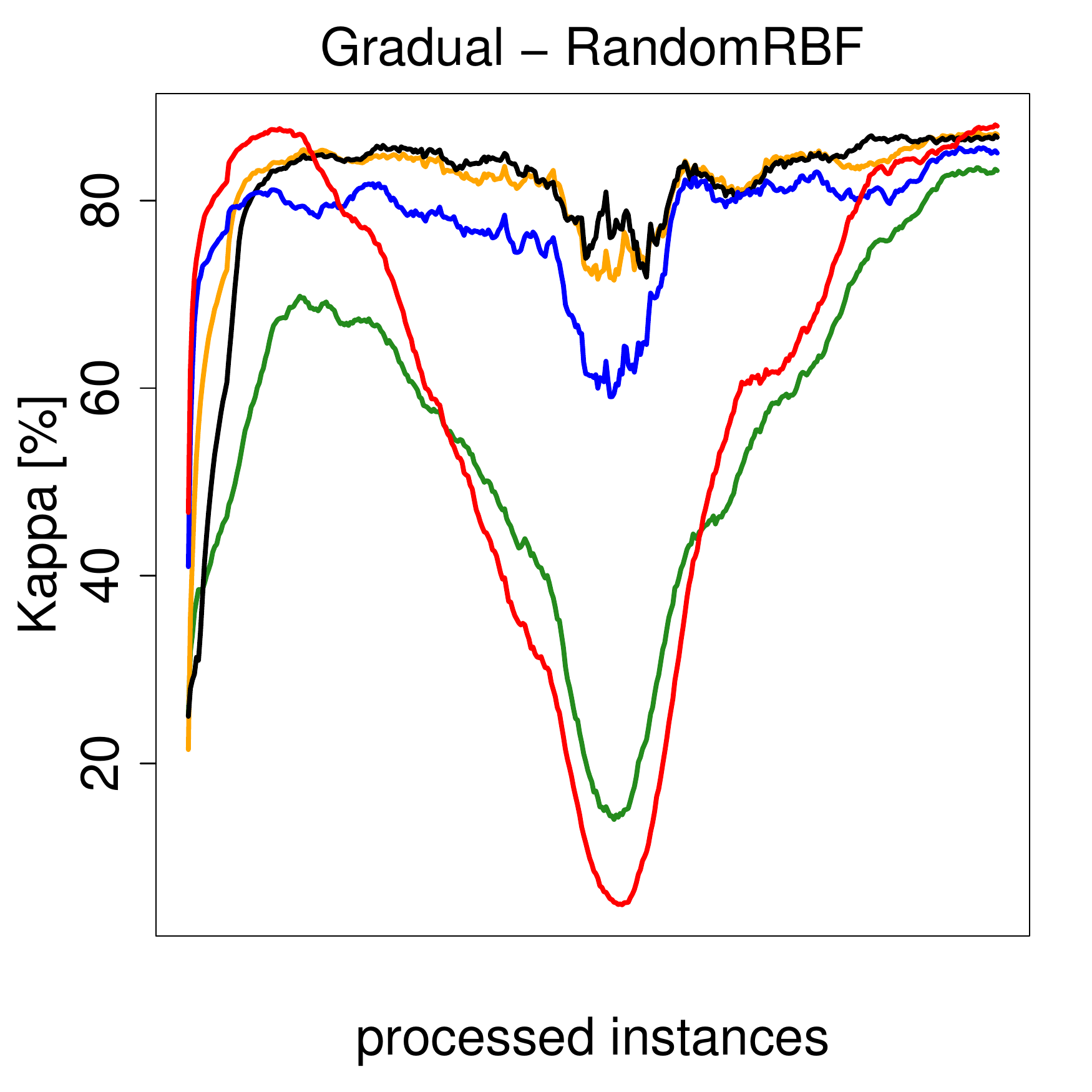}
\includegraphics[width=0.19\columnwidth]{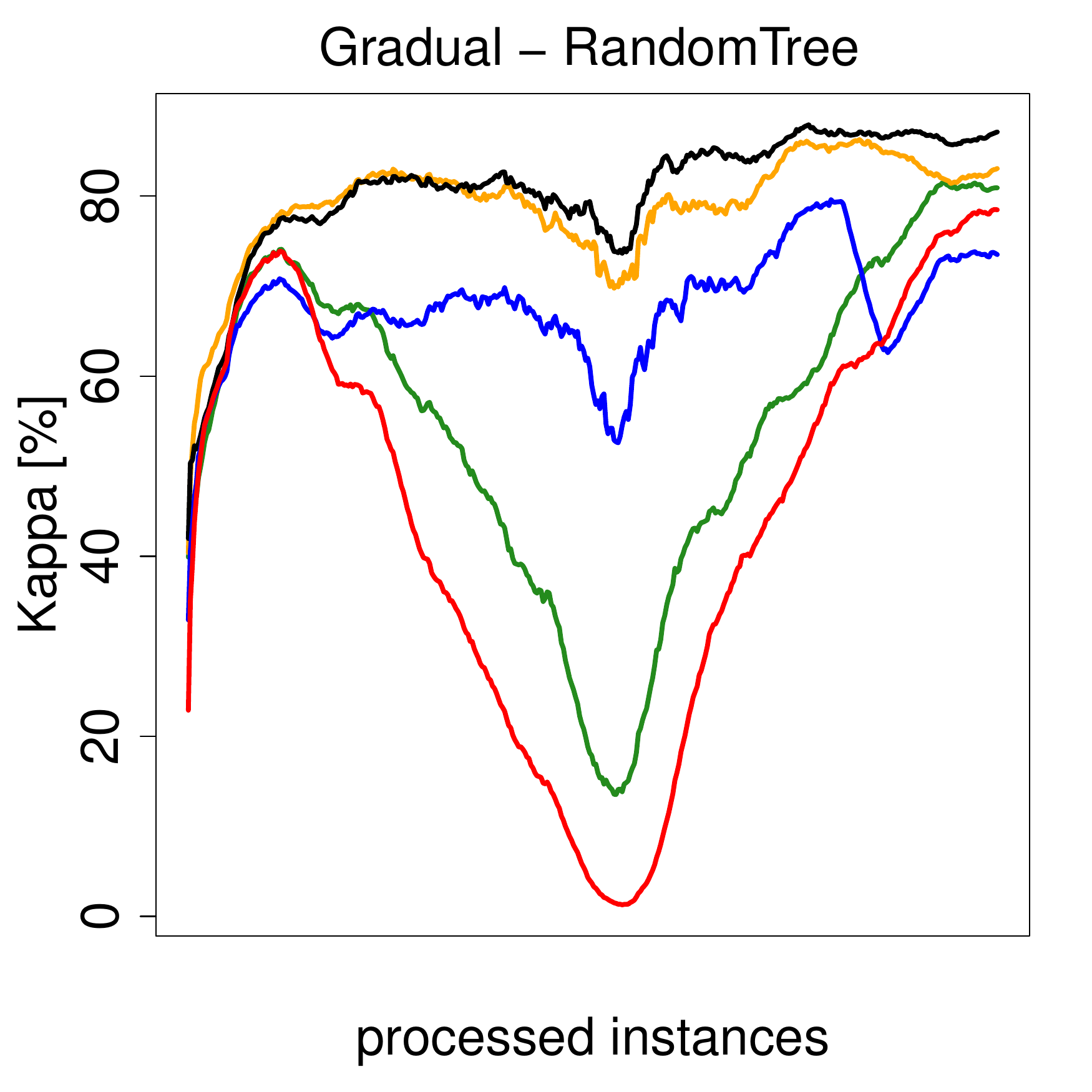}
\includegraphics[width=0.19\columnwidth]{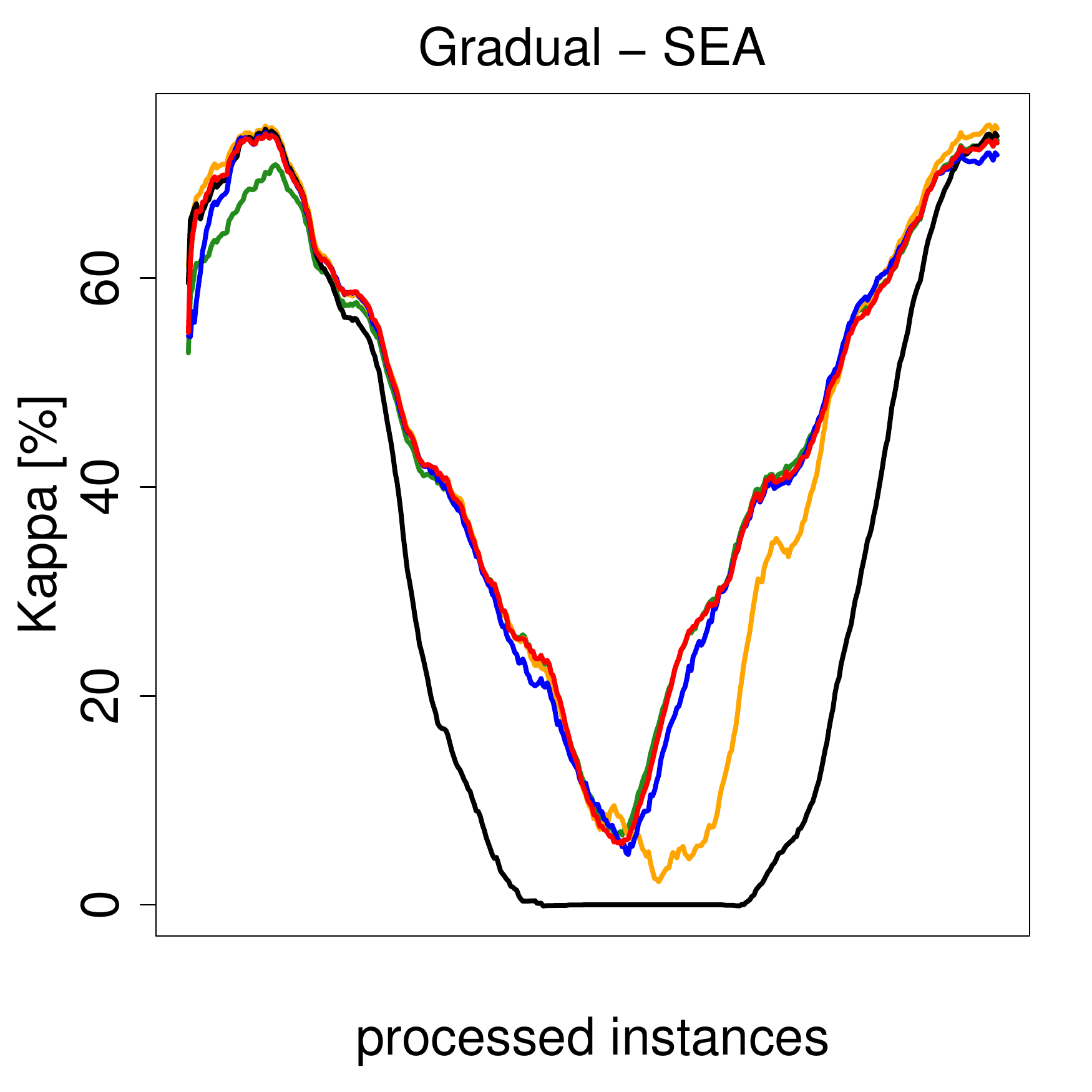}
\includegraphics[width=0.19\columnwidth]{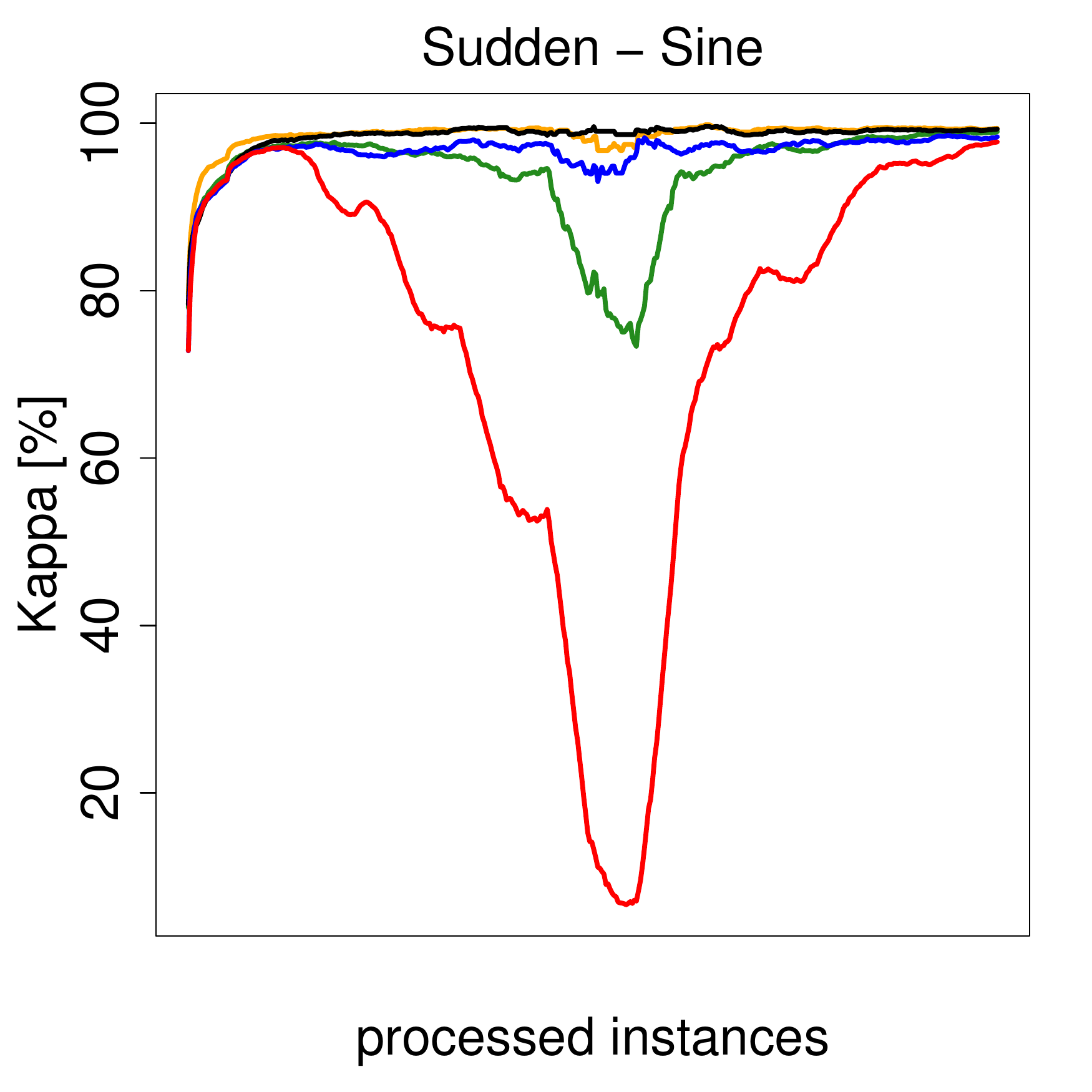}
\includegraphics[width=0.19\columnwidth]{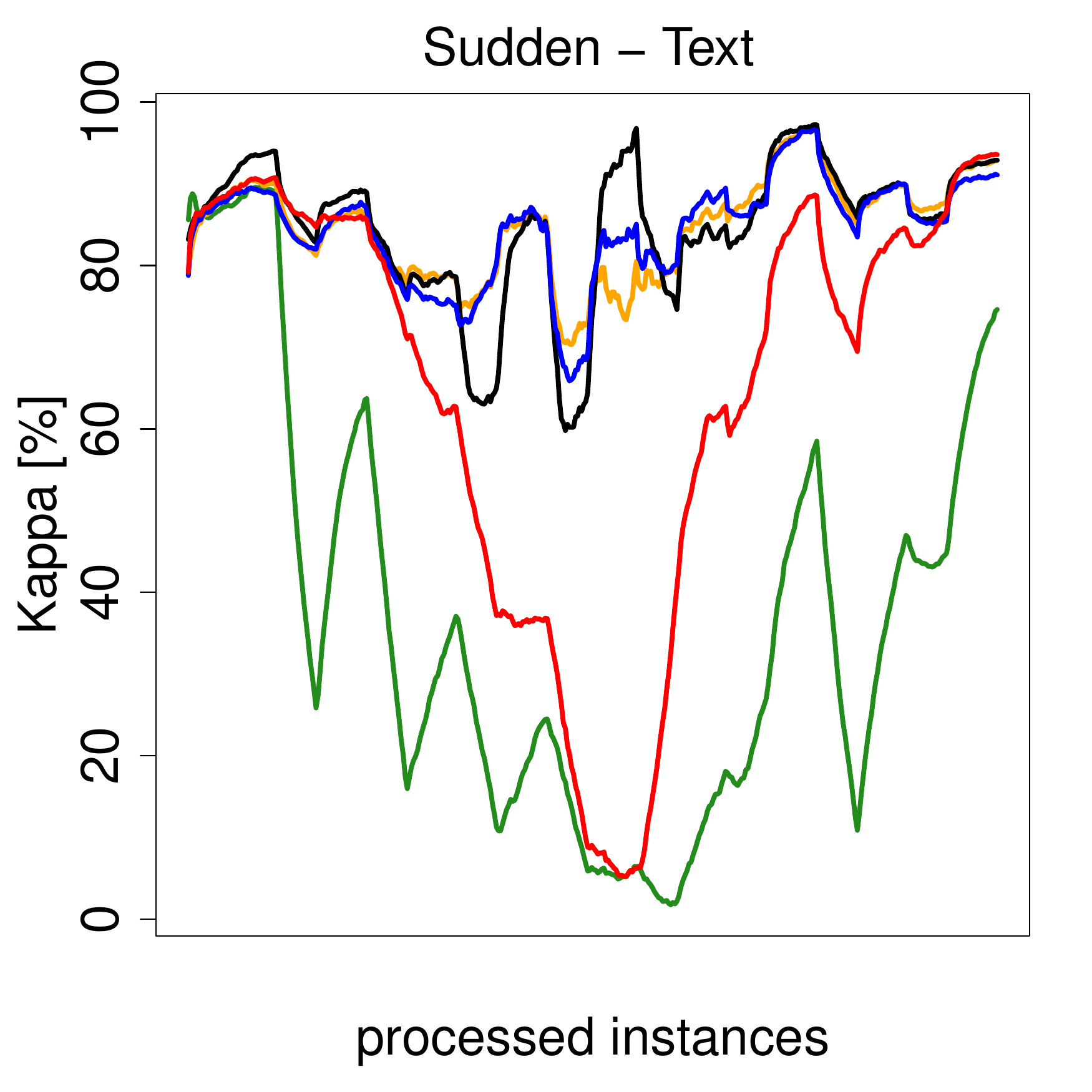}
\caption{G-Mean and Kappa on increasing decreasing class imbalance ratio with gradual and sudden drift.}
\label{fig:ir_increasing_decreasing_study}
\end{figure}

\begin{figure}[t!]
\centering
\includegraphics[width=0.19\columnwidth]{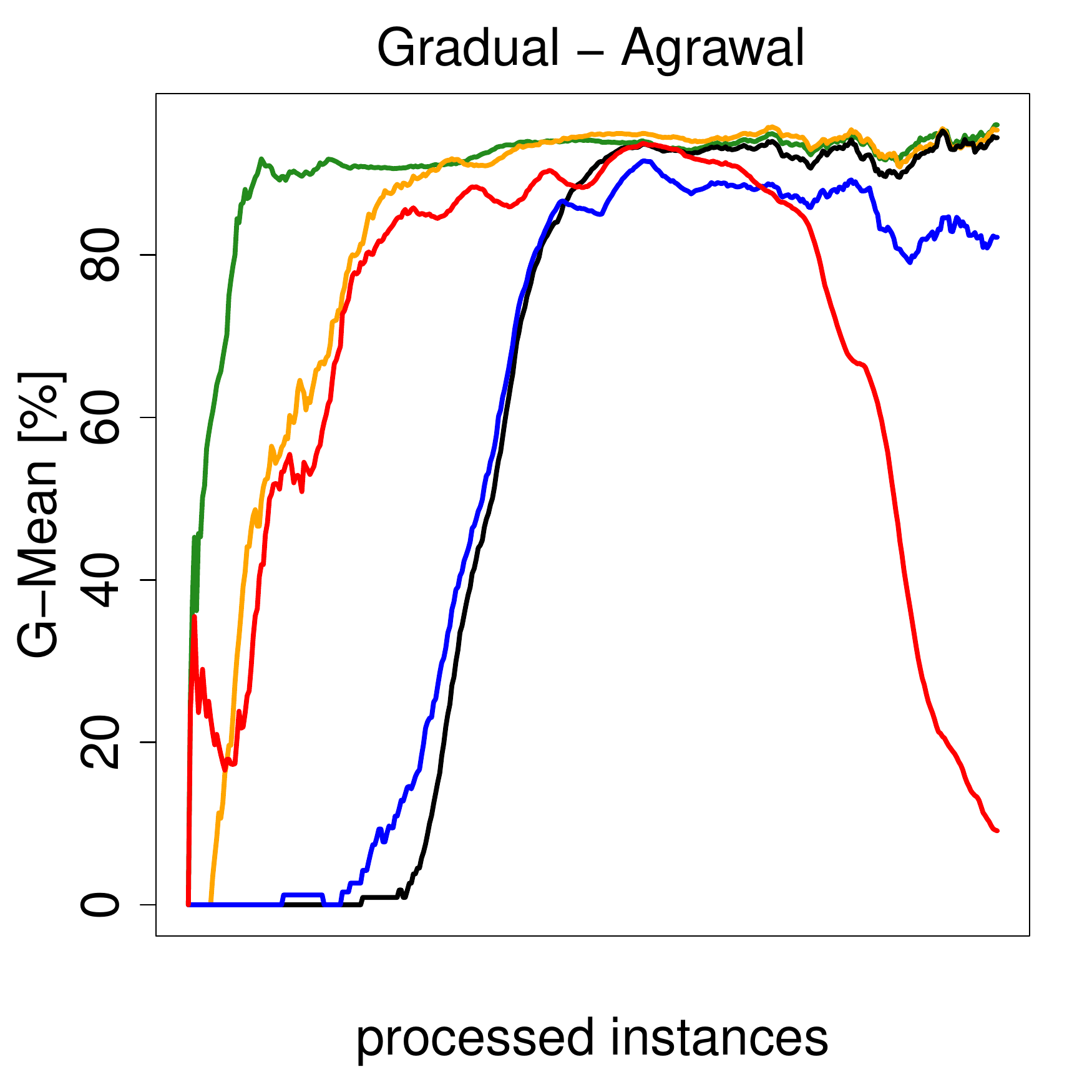}
\includegraphics[width=0.19\columnwidth]{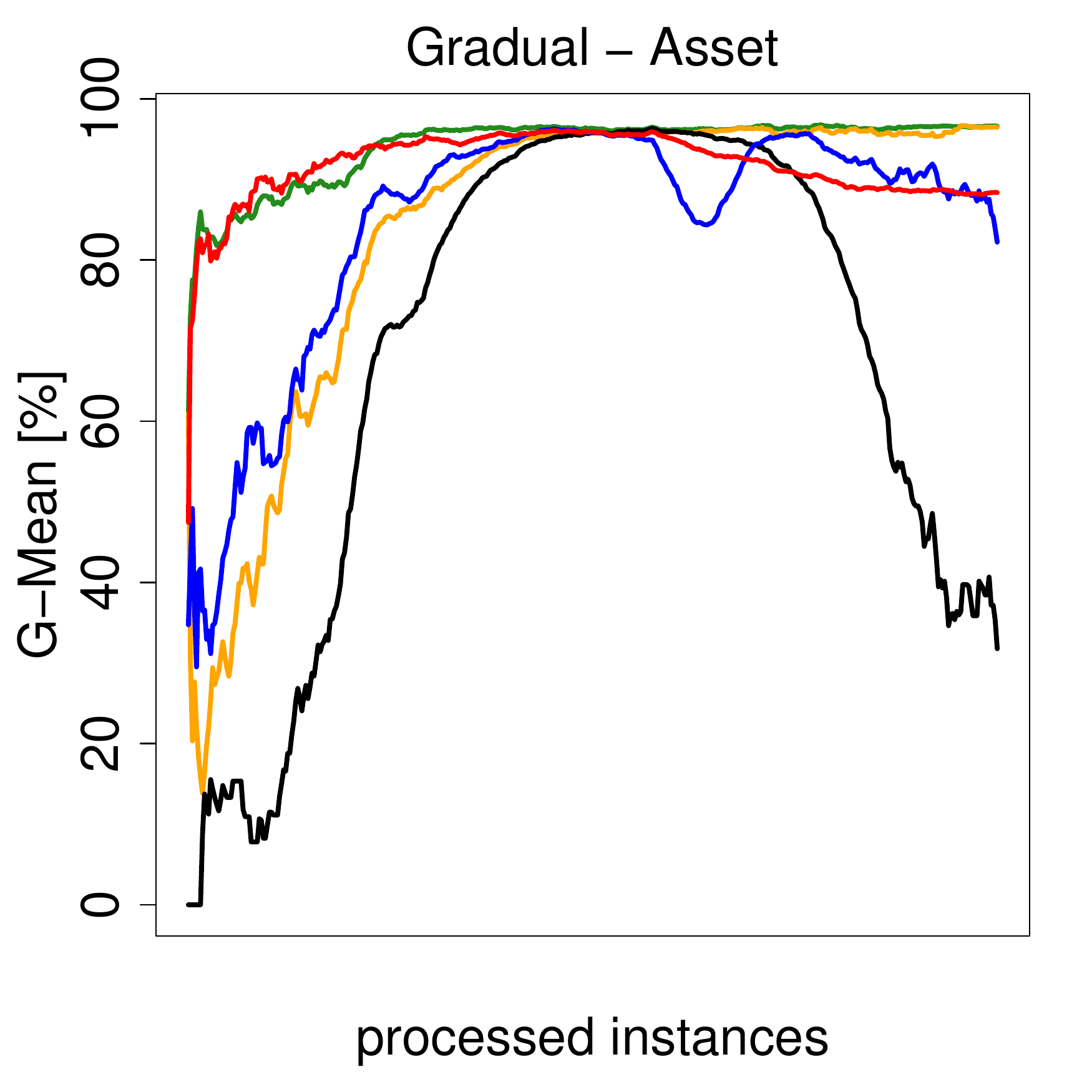}
\includegraphics[width=0.19\columnwidth]{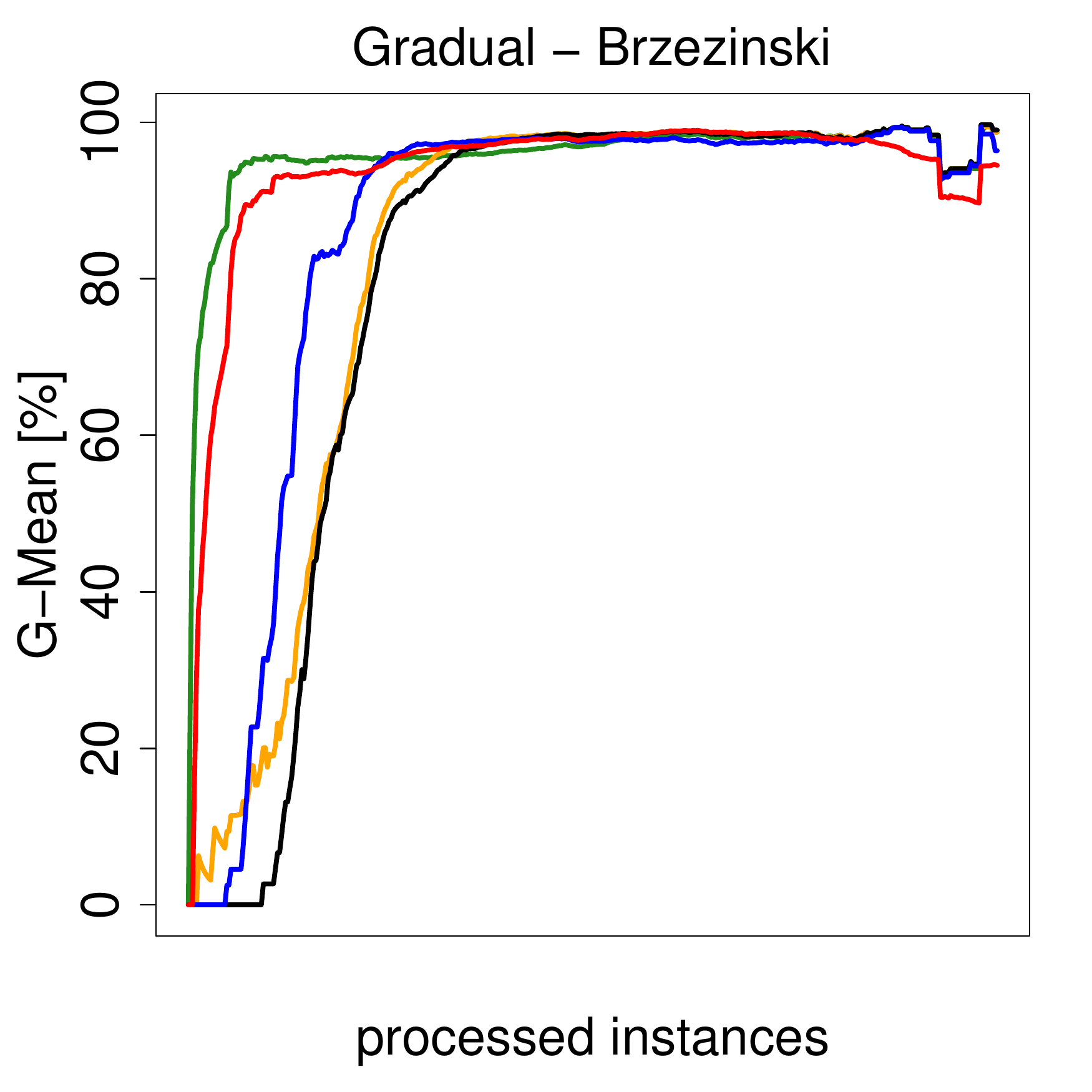}
\includegraphics[width=0.19\columnwidth]{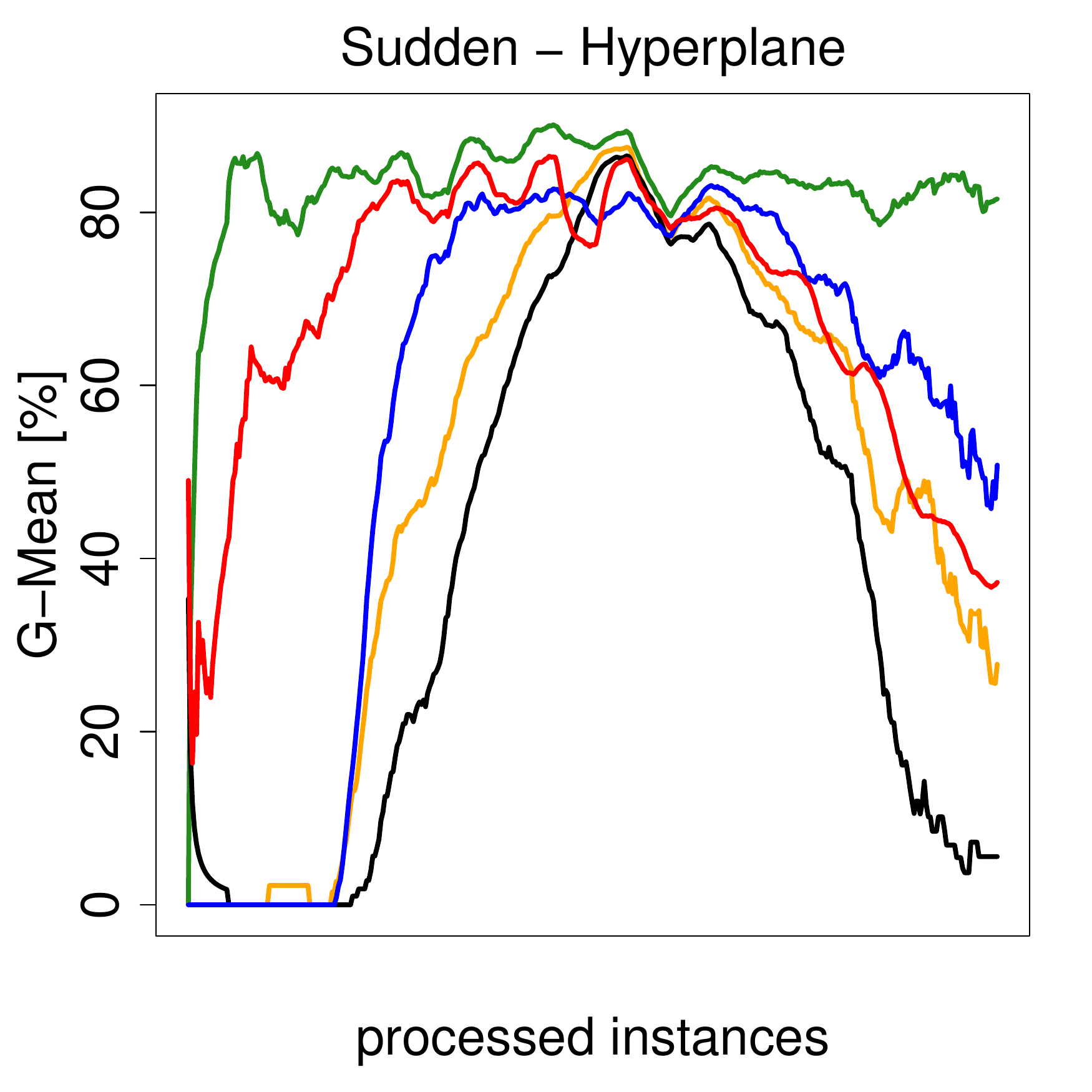}
\includegraphics[width=0.19\columnwidth]{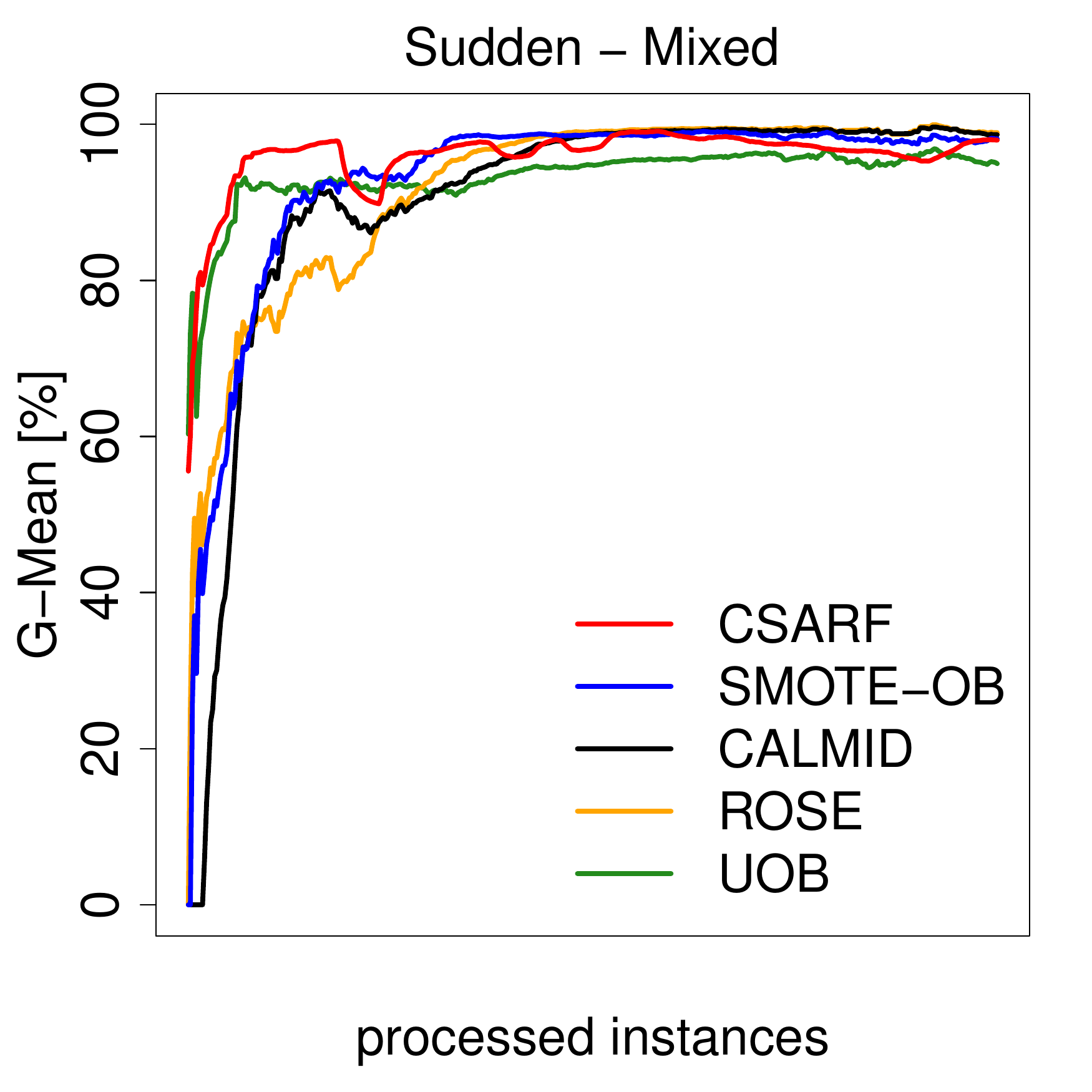}
\includegraphics[width=0.19\columnwidth]{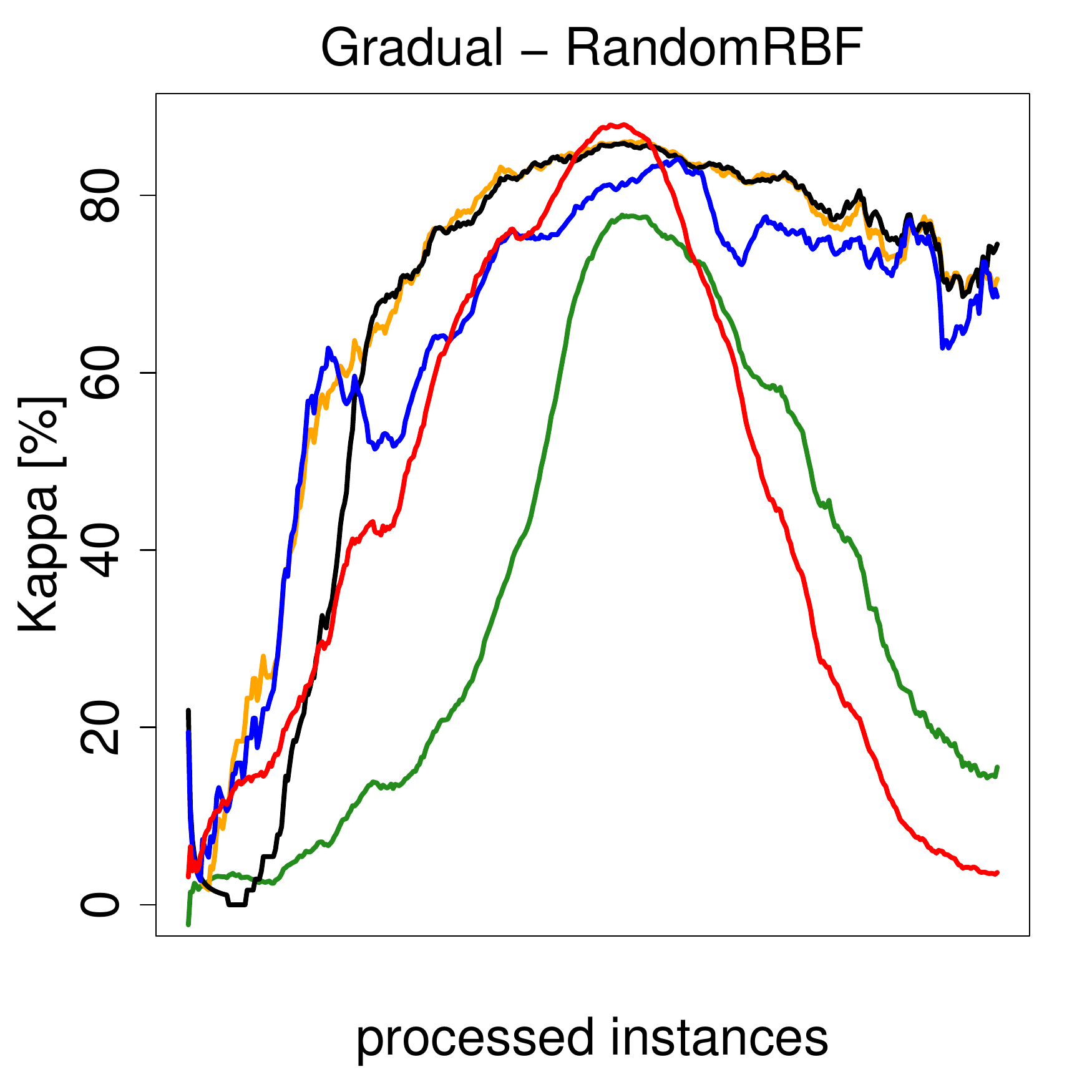}
\includegraphics[width=0.19\columnwidth]{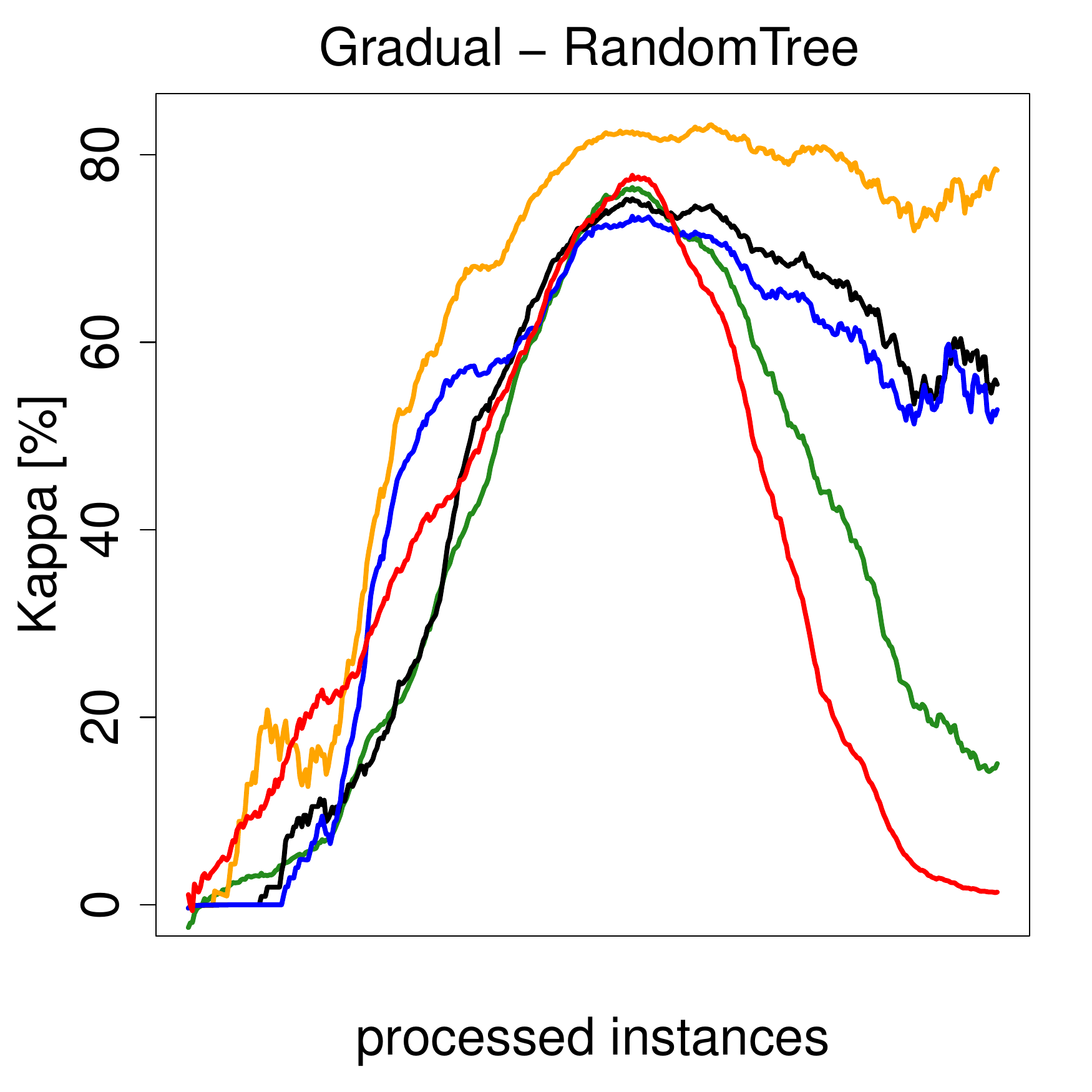}
\includegraphics[width=0.19\columnwidth]{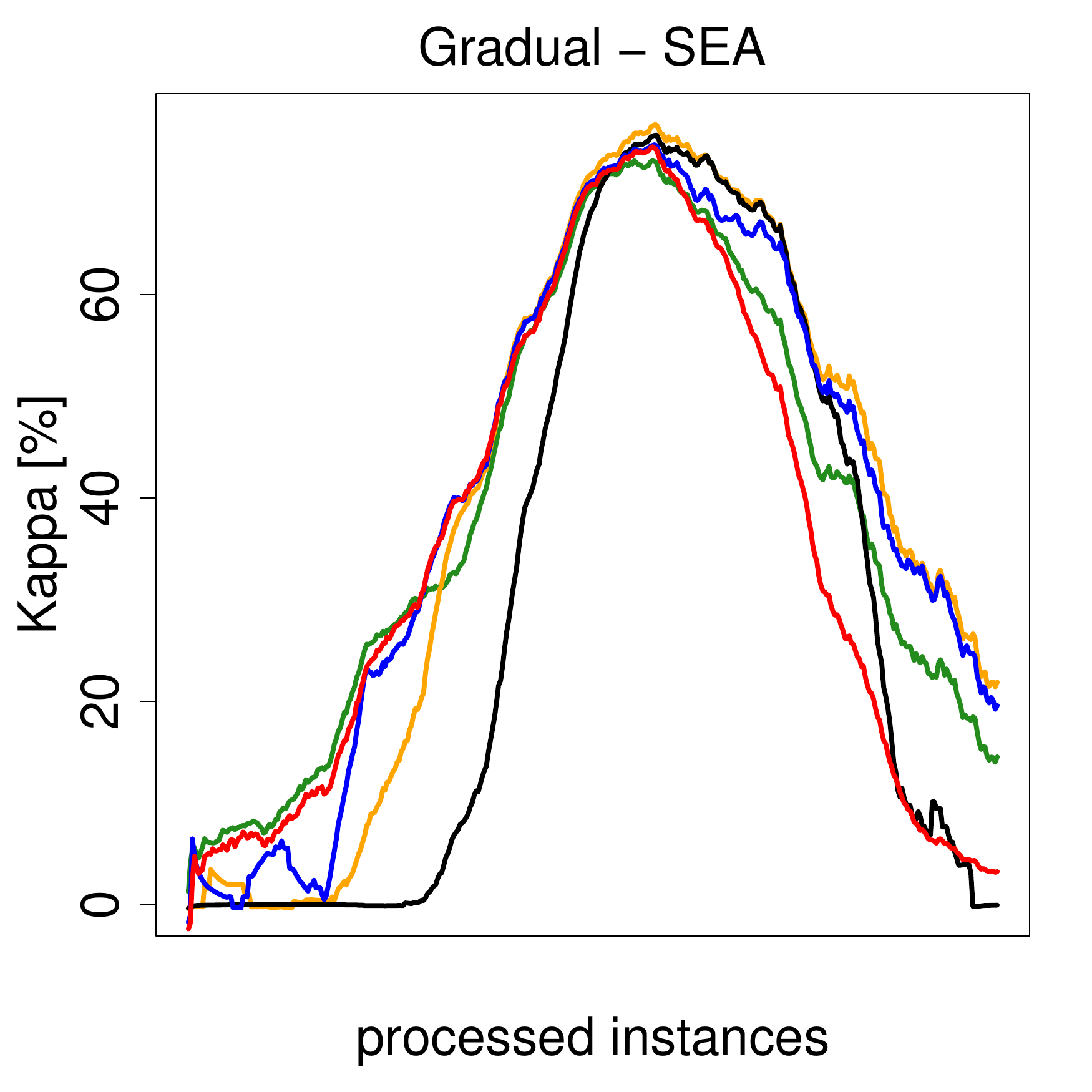}
\includegraphics[width=0.19\columnwidth]{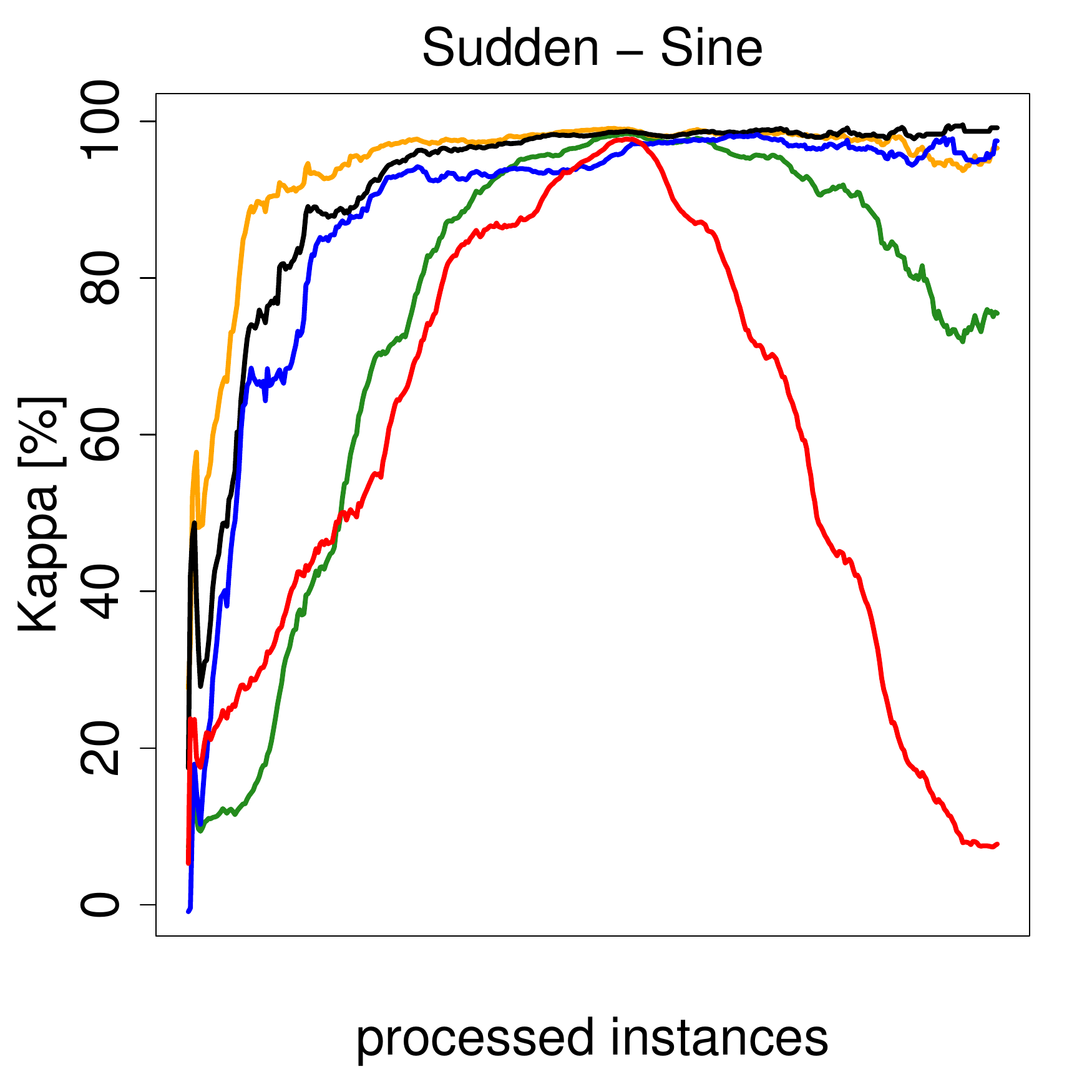}
\includegraphics[width=0.19\columnwidth]{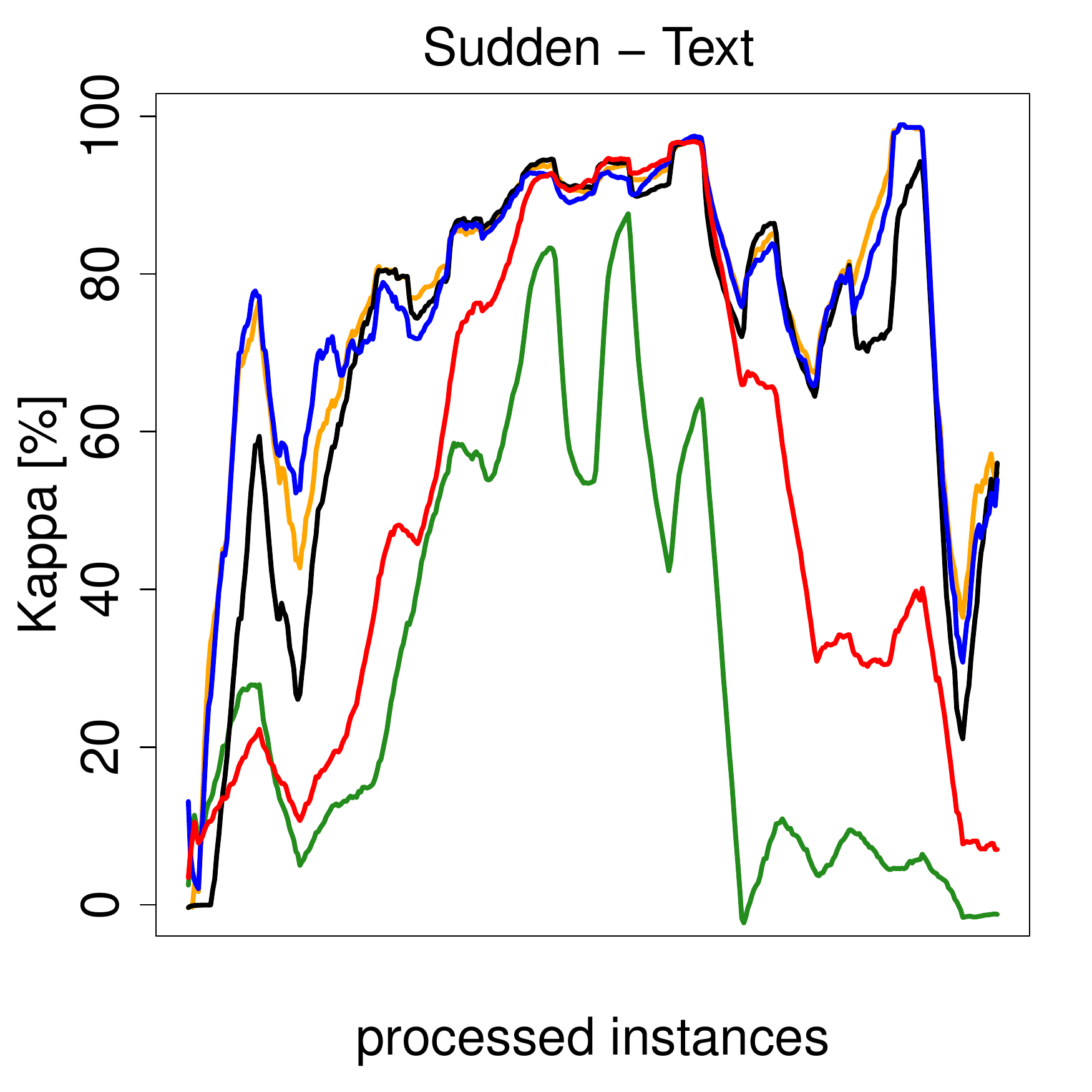}
\caption{G-Mean and Kappa on flipping class imbalance ratio with gradual and sudden drift.}
\label{fig:ir_flipping1_study}
\end{figure}

\begin{figure}[t!]
\centering
\includegraphics[width=0.19\columnwidth]{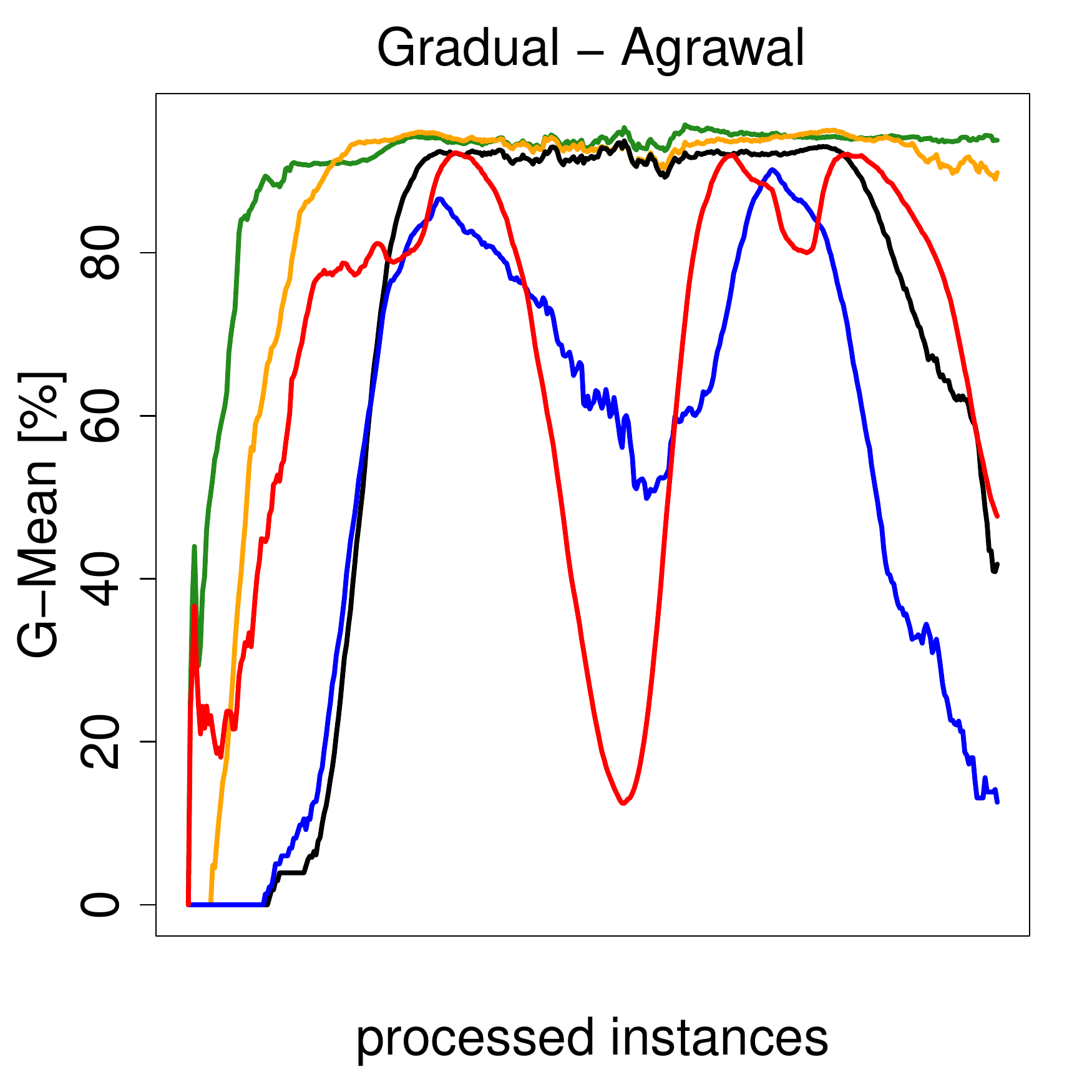}
\includegraphics[width=0.19\columnwidth]{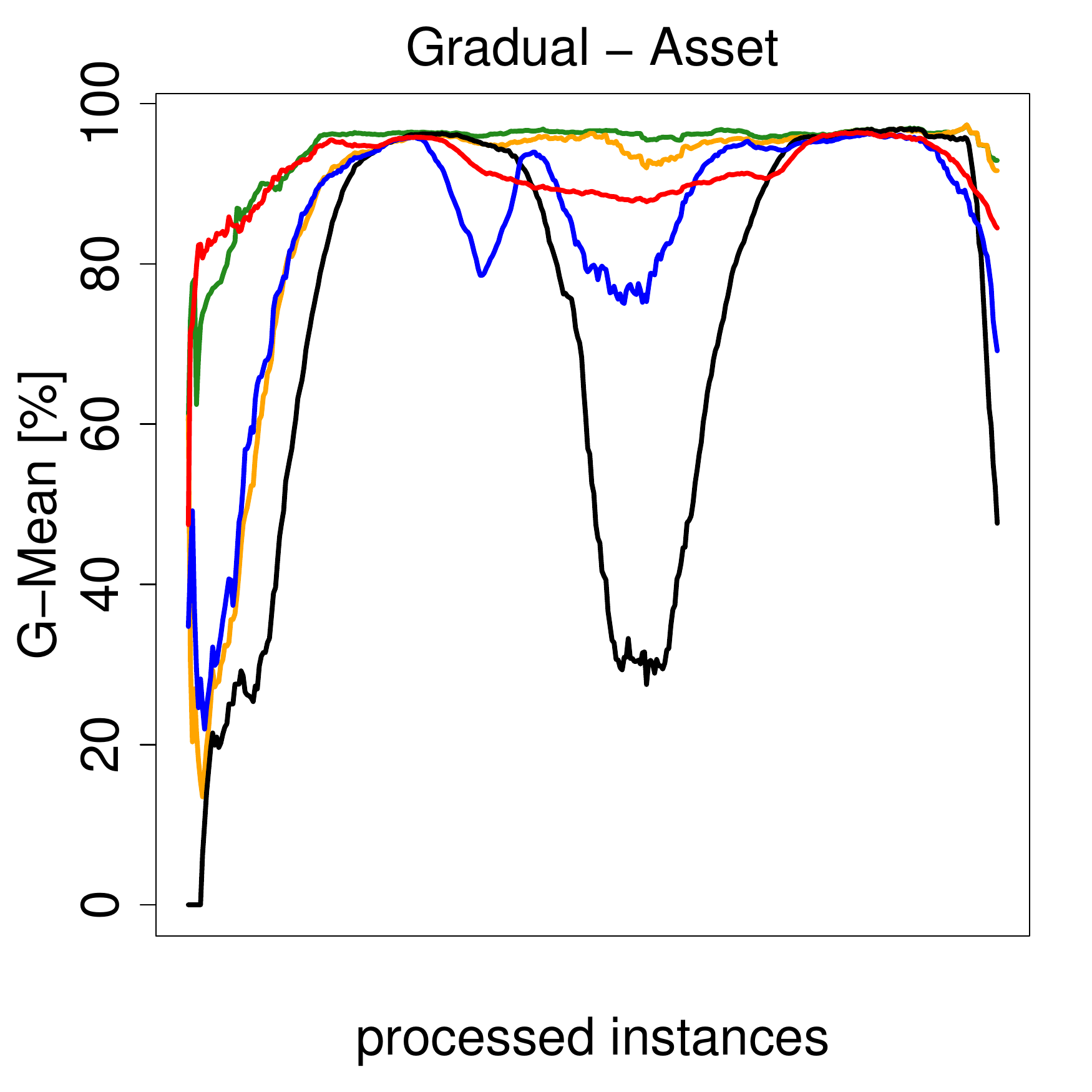}
\includegraphics[width=0.19\columnwidth]{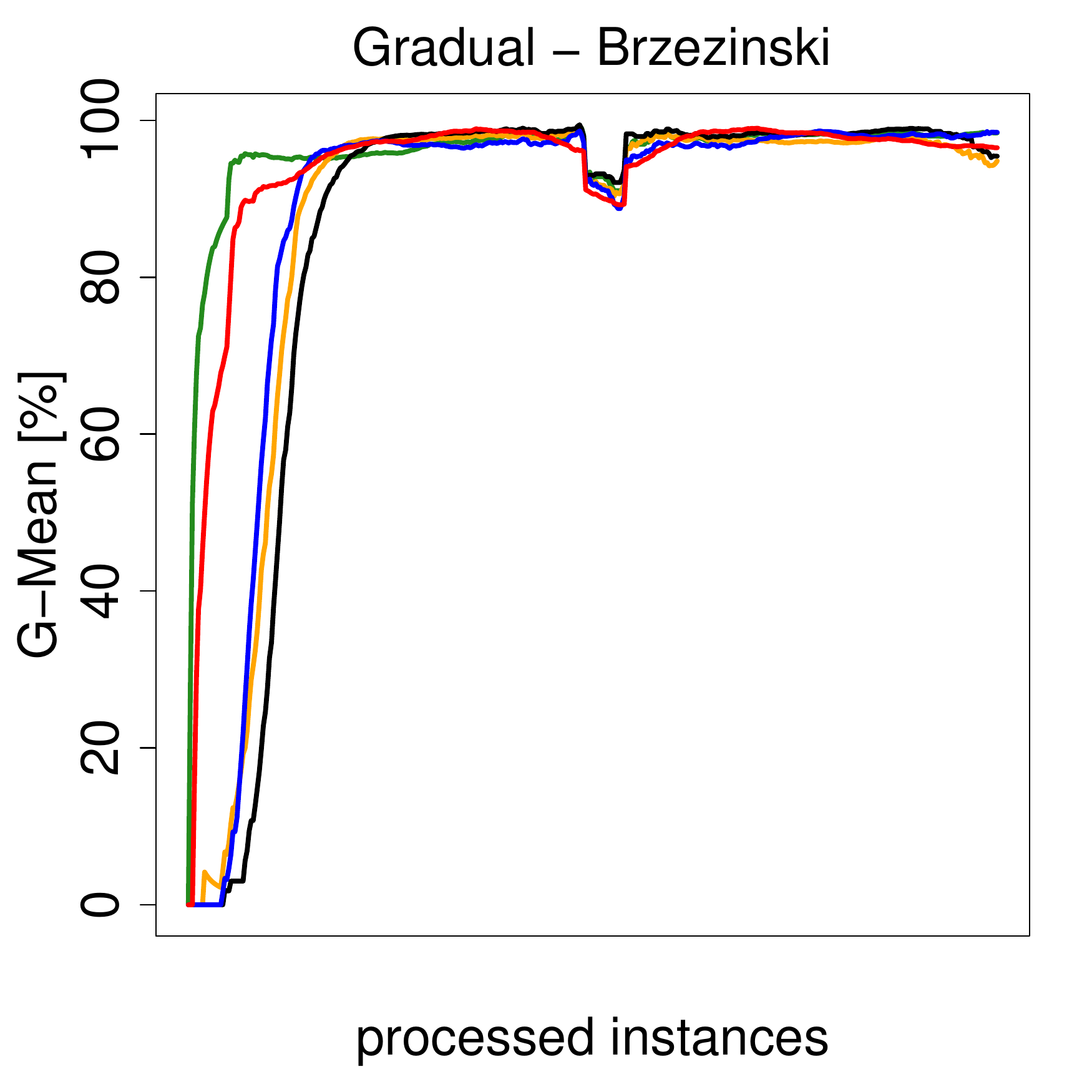}
\includegraphics[width=0.19\columnwidth]{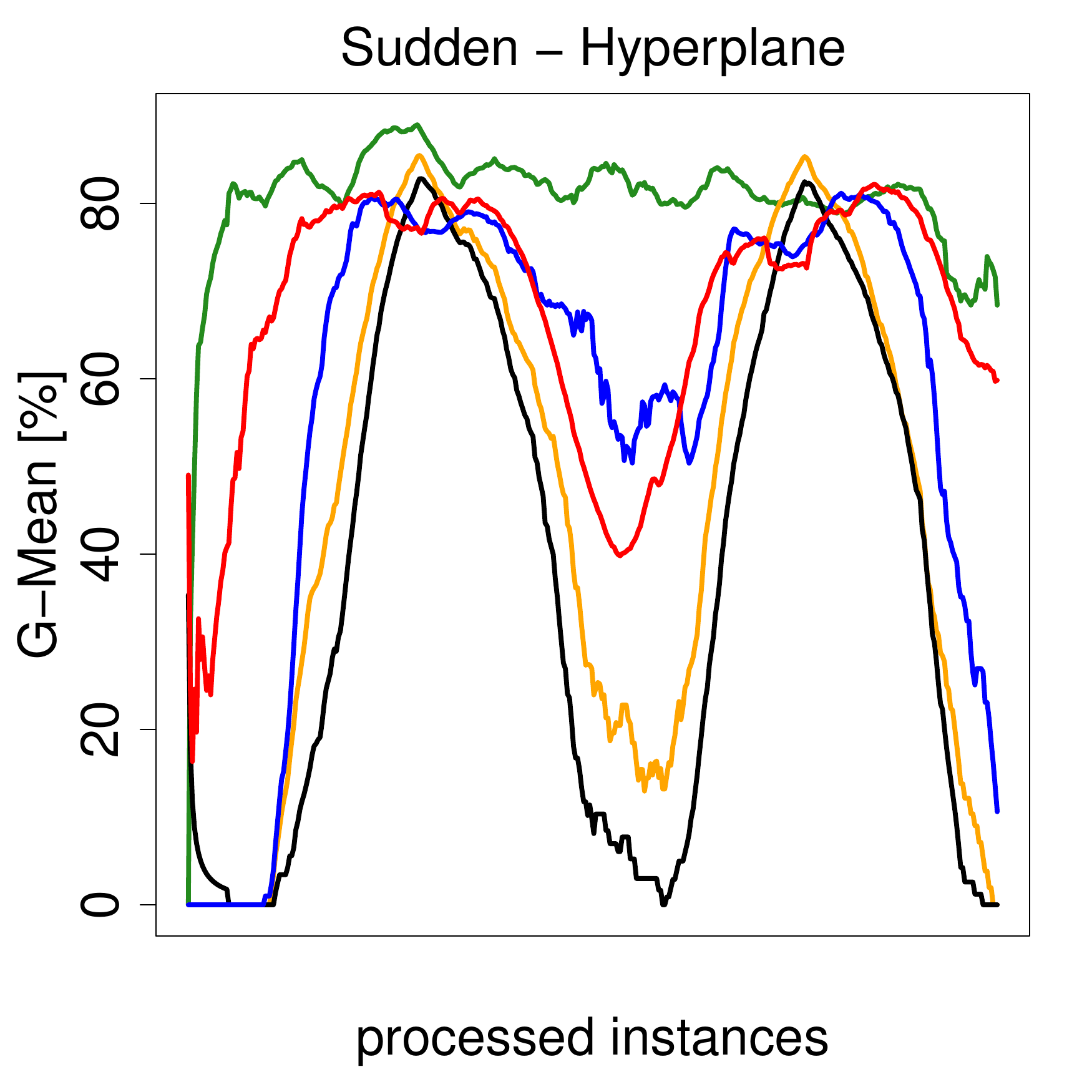}
\includegraphics[width=0.19\columnwidth]{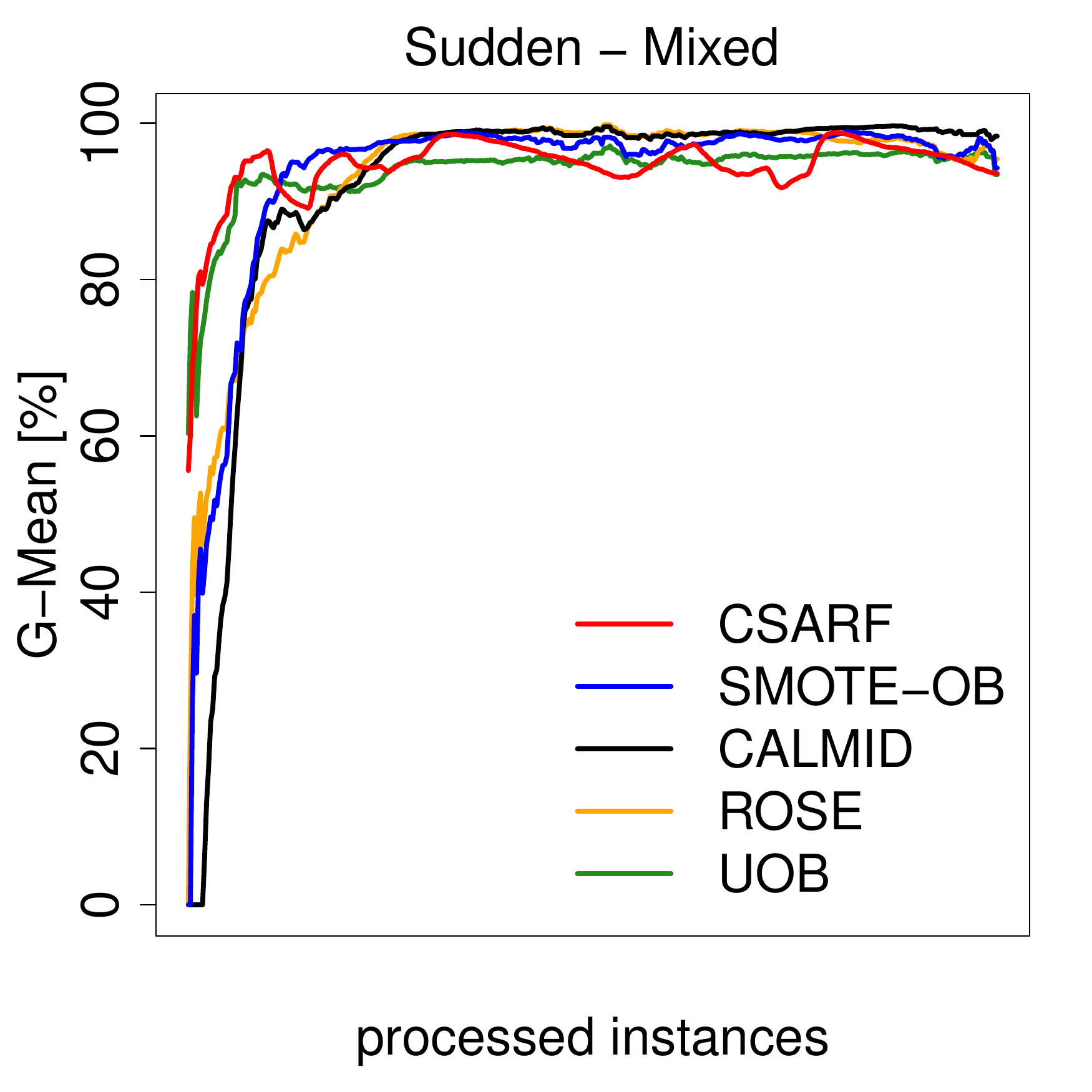}
\includegraphics[width=0.19\columnwidth]{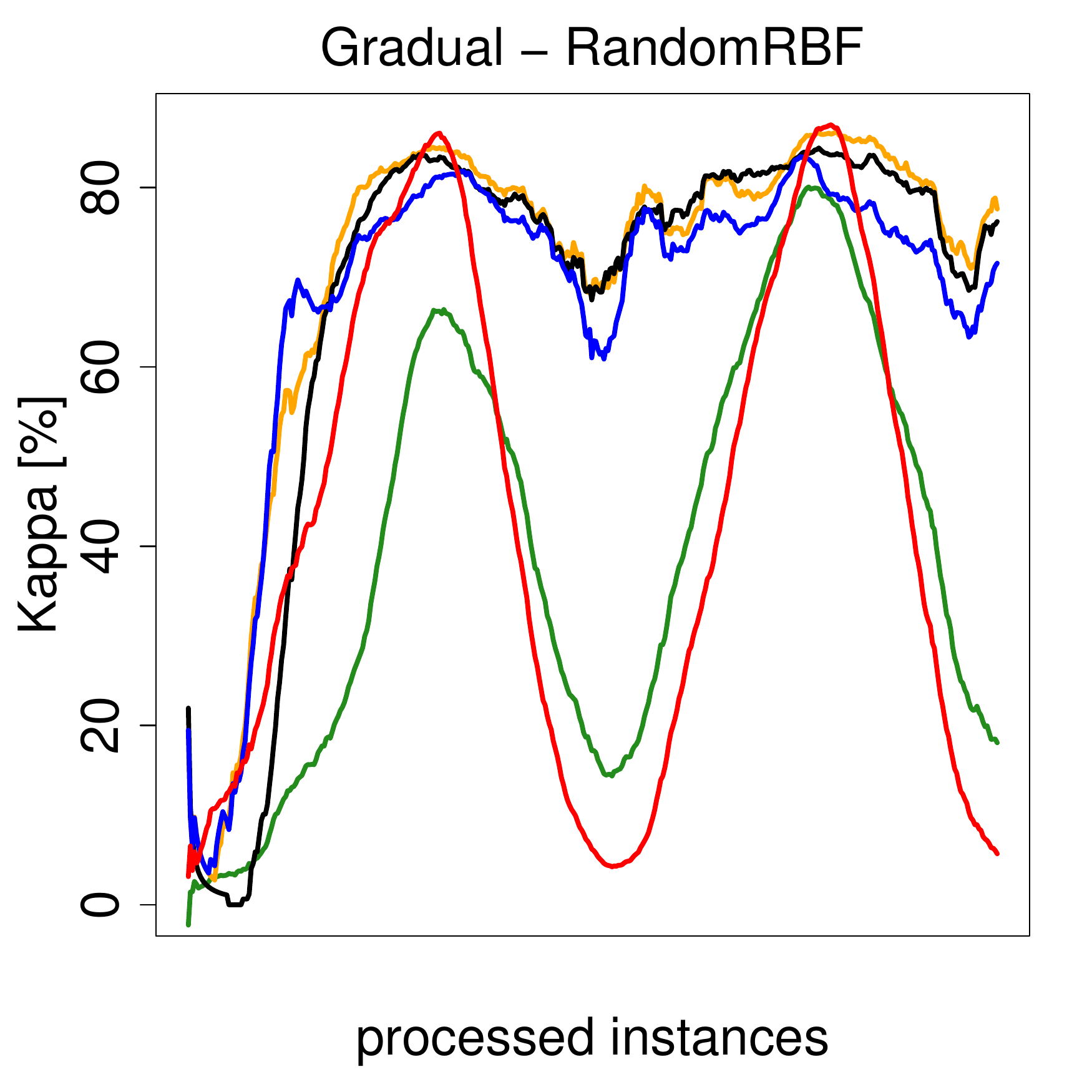}
\includegraphics[width=0.19\columnwidth]{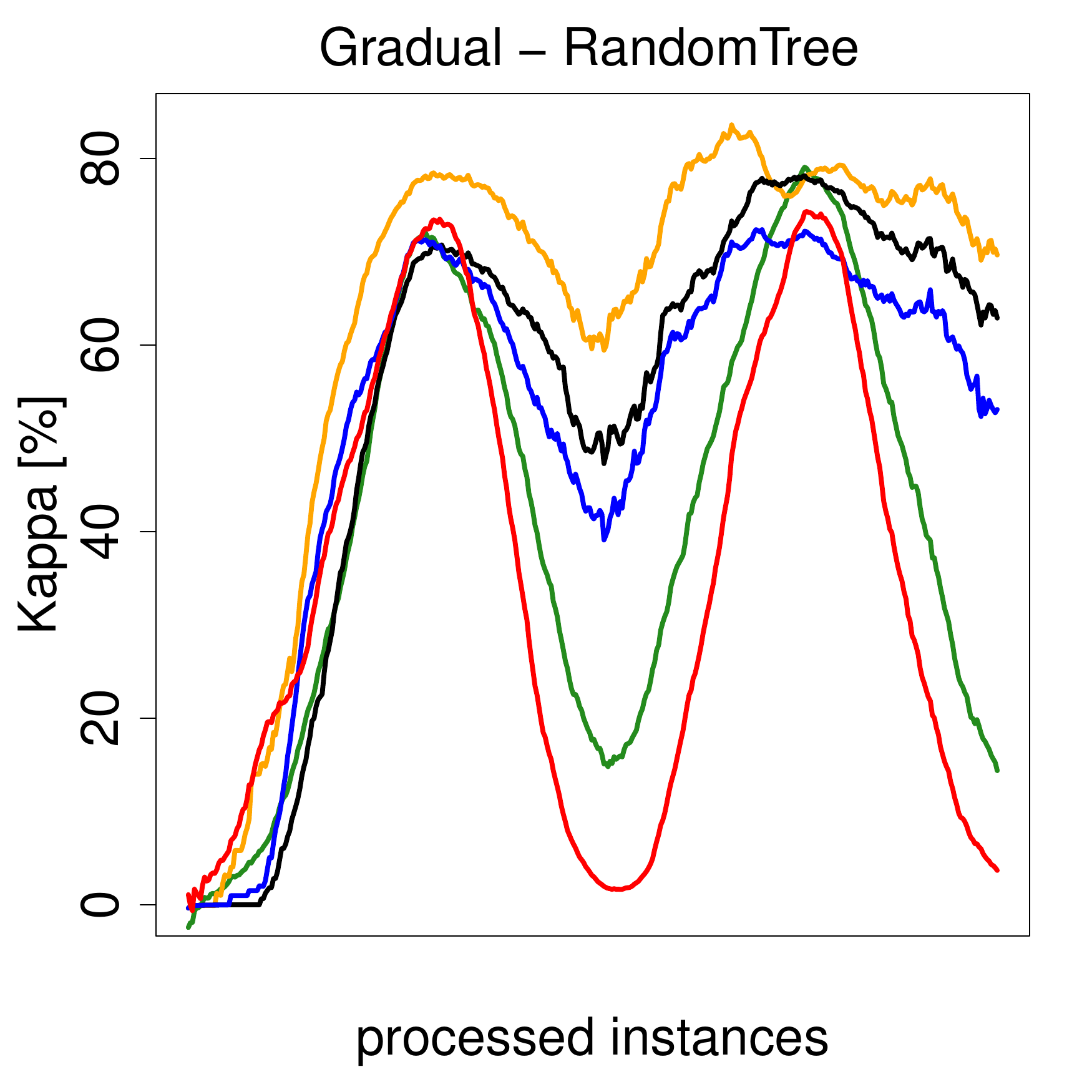}
\includegraphics[width=0.19\columnwidth]{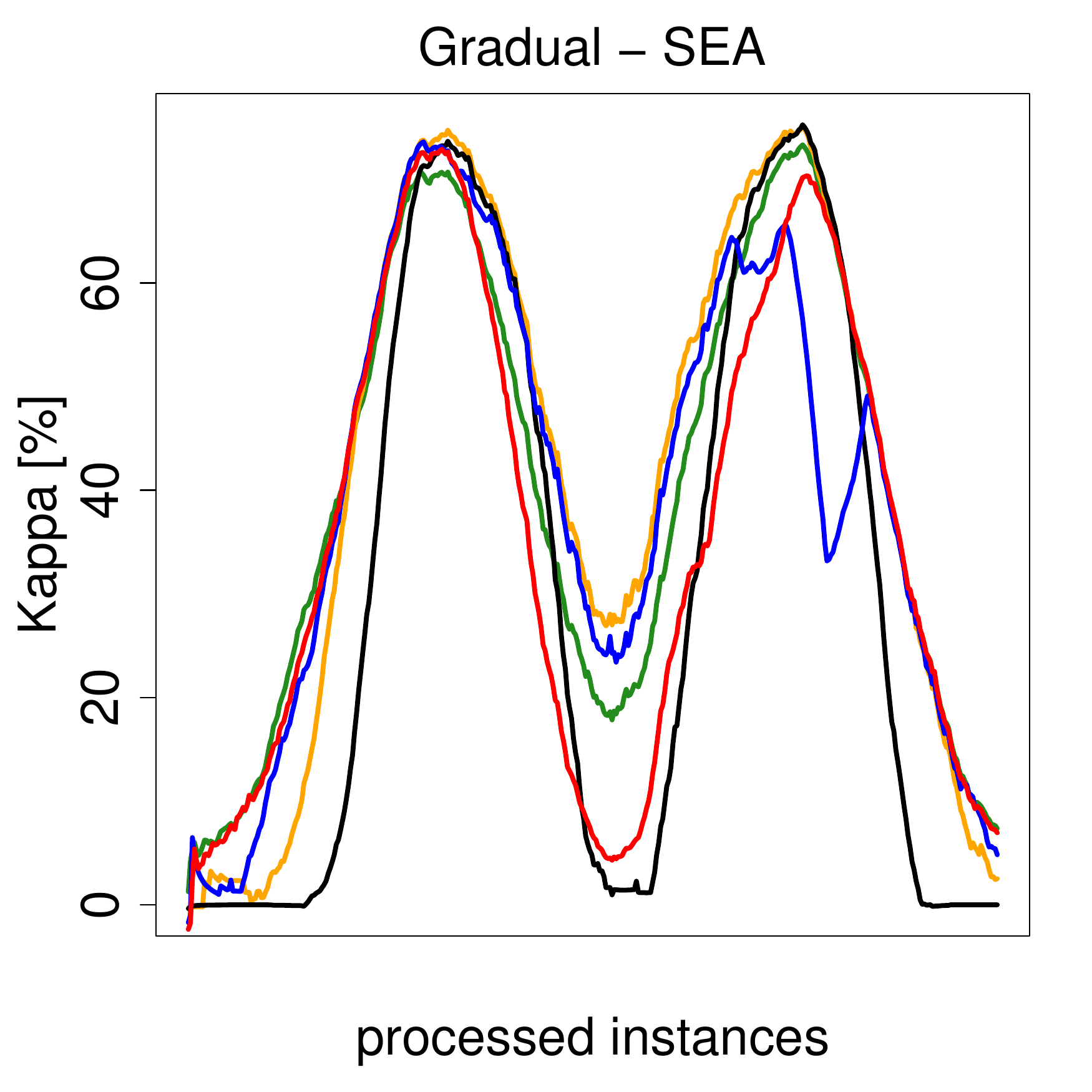}
\includegraphics[width=0.19\columnwidth]{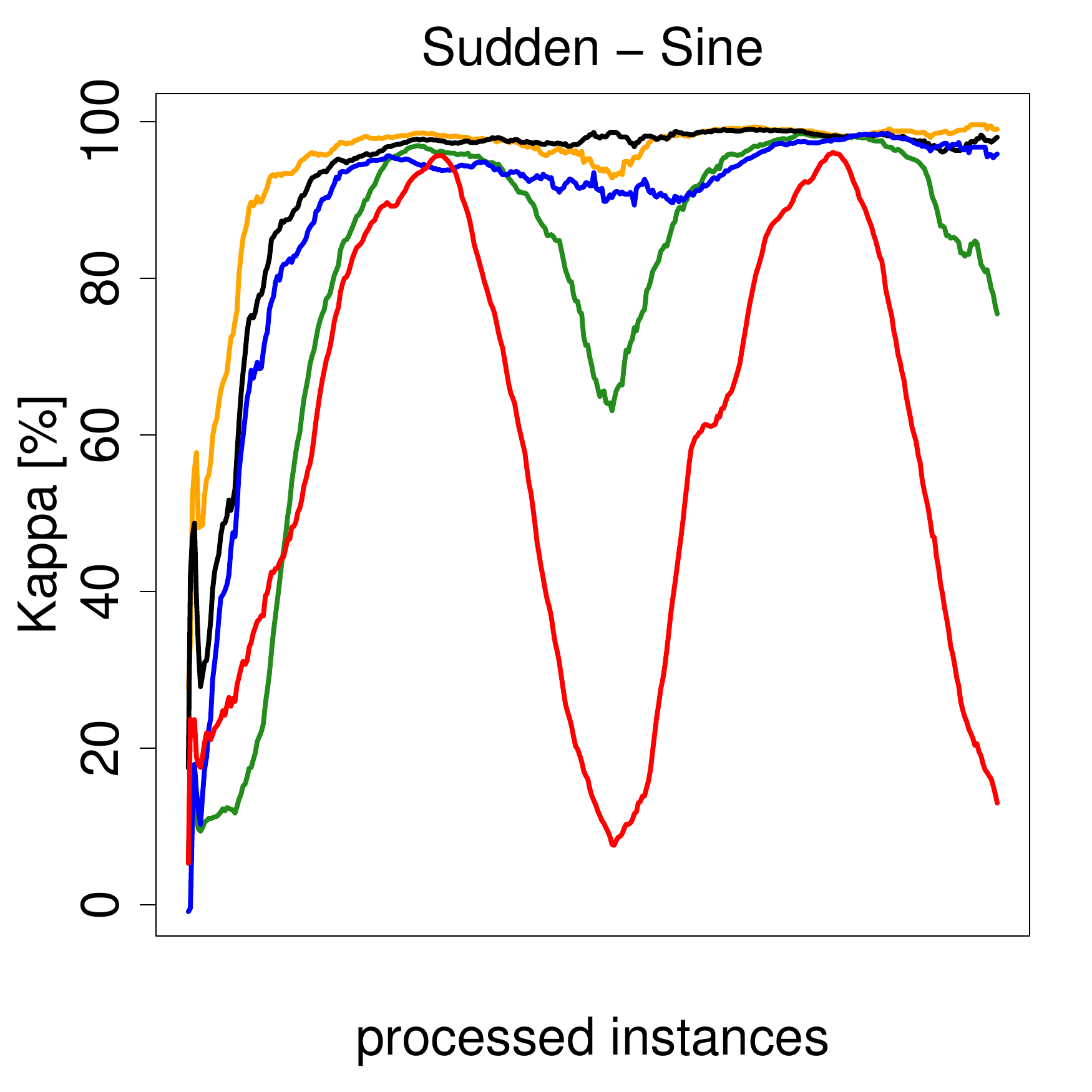}
\includegraphics[width=0.19\columnwidth]{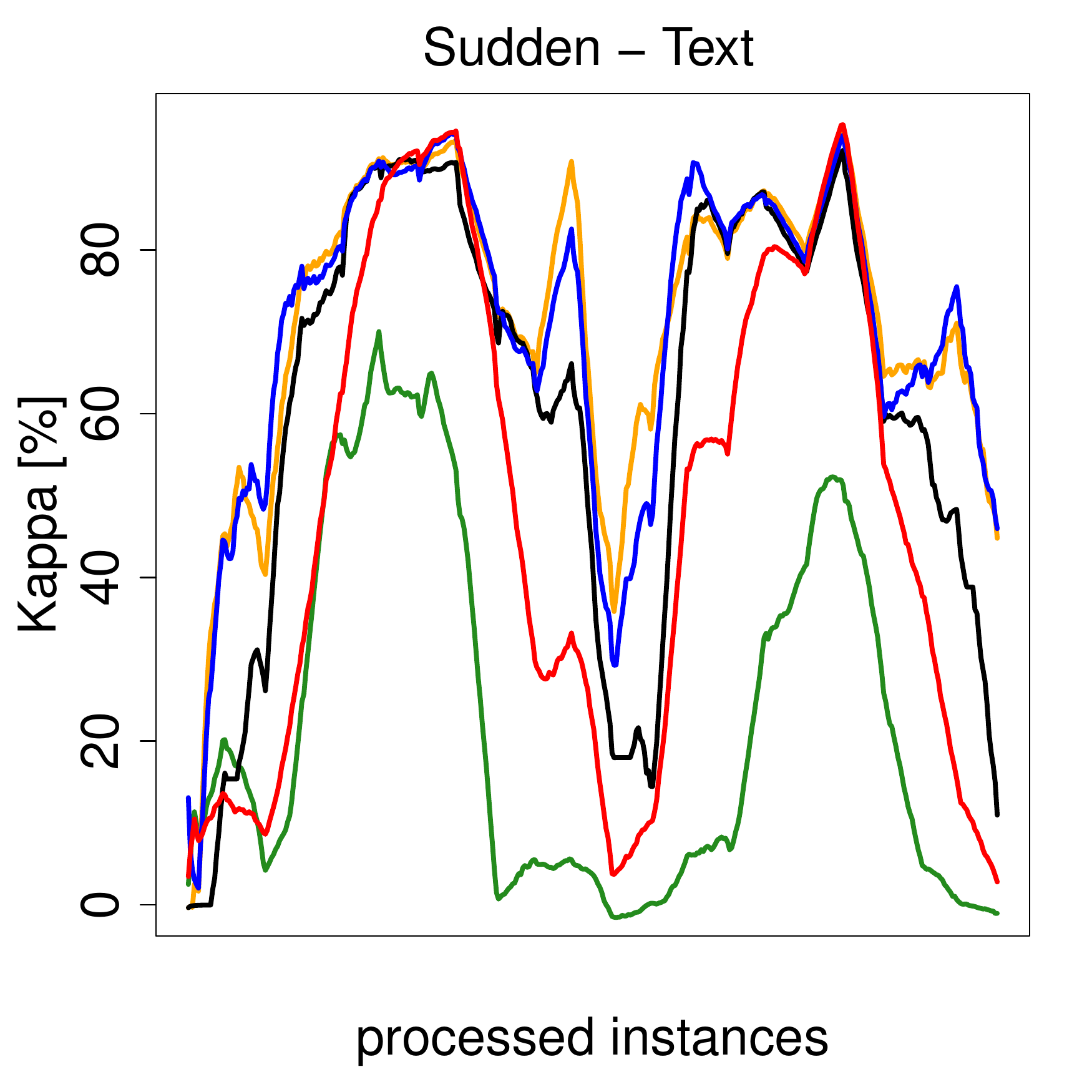}
\caption{G-Mean and Kappa on flipping and reflipping class imbalance ratio with gradual and sudden drift.}
\label{fig:ir_flipping2_study}
\end{figure}

\begin{figure}[t!]
\centering
\includegraphics[width=0.5\columnwidth]{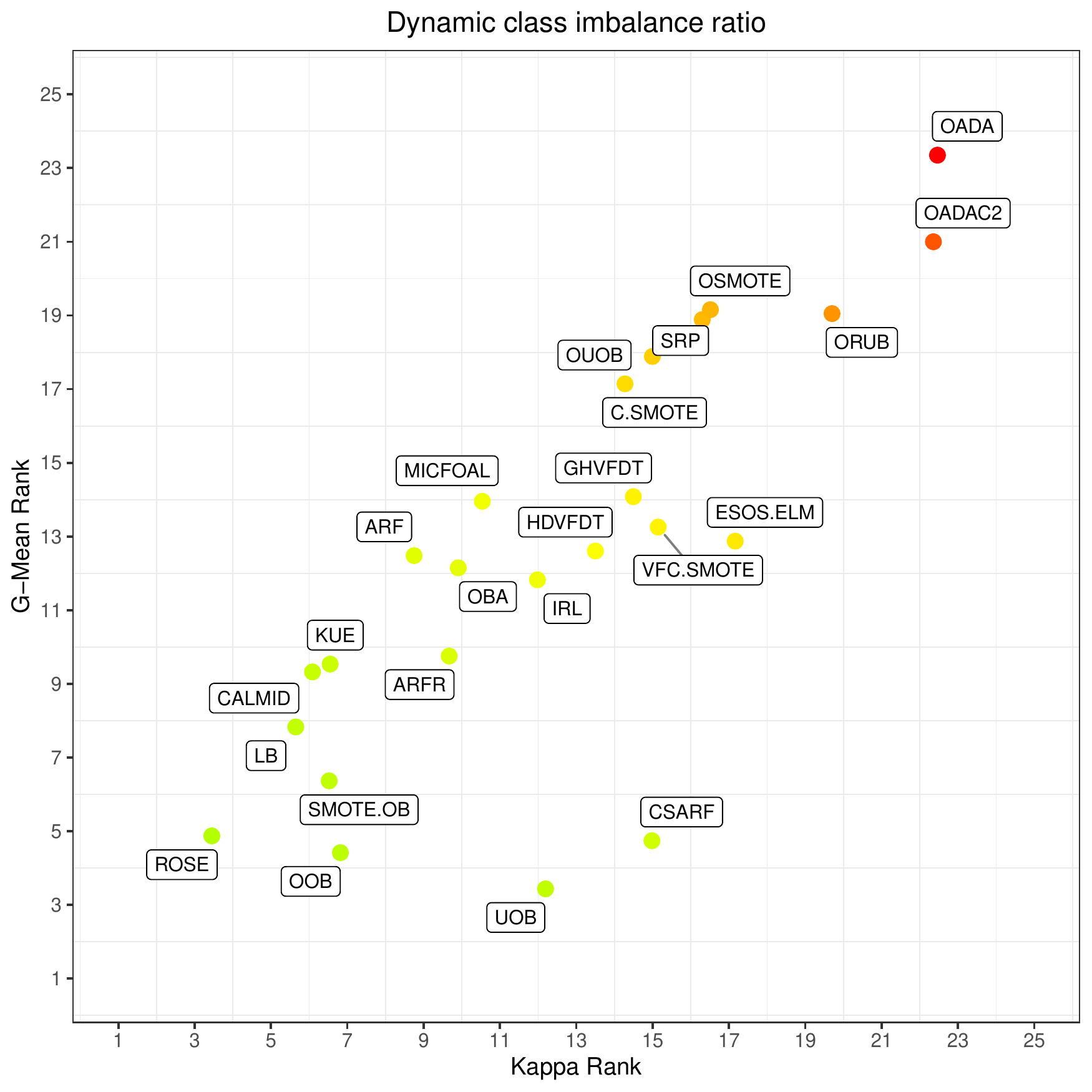}
\caption{Comparison of all 24 algorithms for dynamic class imbalance ratio. Color gradient represents the product of both metrics.}
\label{fig:BC_DIR_scatter}
\end{figure}

\noindent \textbf{Goal of the experiment.} This experiment was designed to address \textbf{RQ2} and to evaluate how classifiers behave under dynamic imbalance ratios. Even though many existing methods were designed to deal with static imbalance ratio, they lack mechanisms that allow adaptation to time-varying changes in the imbalance ratio. To evaluate this, we prepared four scenarios: (i) increasing the imbalance ratio \{1, 5, 10, 20, 50, 100\}, (ii) increasing then decreasing the imbalance ratio \{1, 5, 10, 20, 50, 100, 50, 20, 10, 5, 1\}, (iii) flipping the imbalance ratio, in which the majority becomes the minority class and vice versa \{100, 50, 20, 10, 5, 1, 0.2, 0.1, 0.05, 0.02, 0.01\}, and (iv) flipping then repflipping the imbalance ratio, in which the majority becomes the minority class and then flips back to become the majority and vice versa \{100, 50, 20, 10, 5, 1, 0.2, 0.1, 0.05, 0.02, 0.01, 0.02, 0.05, 0.1, 0.2, 1, 5, 10, 20, 50, 100\}. In this experiment, we also evaluated two types of drift: gradual and sudden. This allows us to analyze how the classifiers can cope with dynamic imbalance ratio changes and how they can adapt when majority and minority change roles. Figures~\ref{fig:ir_increasing_study},\ref{fig:ir_increasing_decreasing_study},\ref{fig:ir_flipping1_study},\ref{fig:ir_flipping2_study} present the G-Mean and Kappa over time for the five selected classifiers for the generators and for both types of drift over (i) increasing imbalance ratio, (ii) increasing then decreasing imbalance ratio, (iii) flipping imbalance ratio, and (iv) flipping then reflipping imbalance ratio. To increase readability, the line plots were smoothed using a moving average of 20 data points.
Table~\ref{tab:BC_DIR} presents the average G-Mean and Kappa for the top 10 classifiers for each of the evaluated dynamic scenarios and the overall rank of the algorithms. Figure~\ref{fig:BC_DIR_scatter} provides an overall comparison among all algorithms.

\noindent \textbf{Discussion}

\noindent \textit{Impact of approach to class imbalance.} In our second experiment, it is interesting to note that best-performing classifiers are similar to the ones in the static scenario with their difference relying on how quickly they can recover from the imbalance drift.
Regarding data-level approaches, \acrshort{uob} and \acrshort{oob} achieve good results, even without having explicit mechanisms for handling concept drift in imbalance ratio. \acrshort{ouob} once again did not display satisfactory results, mainly because of its inability to switch between different resampling approaches that lead to a slower response to changes. \acrshort{smote} based methods had diverging performances. \acrshort{csmote} and \acrshort{osmote} cannot handle increasing imbalance ratio, losing their performance over time and not being able to cope with the increasing disproportion among classes. This can be explained by the fact that increasing imbalance ratio leads to a lower number of minority instances that could be used for the generation of relevant and diverse artificial instances. \acrshort{smoteob} was among the best-performing classifiers. This can be explained by \acrshort{smoteob} undersampling together with oversampling, leading to smaller disproportions between classes and more homogeneous $k$-nearest neighborhoods used for instance generation. 

For algorithms modification methods, \acrshort{csarf} is the best performing one according to G-Mean but suffers under Kappa metric. When dynamic changes are introduced, \acrshort{rose} presented the most balanced results according to both metrics. In the previous experiments, \acrshort{rose} was also one of the best classifiers, but its underlying change adaptation mechanisms and usage of dynamic sliding windows lead to significant improvements for non-stationary imbalance, especially when dealing with high imbalance ratios and flipping role of classes. 

\noindent \textit{Impact of ensemble architecture.} Experiments with the dynamic imbalance ratio confirm our previous observations regarding the most robust architecture choice for ensembles. Boosting-based methods return even worse performance when dealing with an evolving disproportion between classes. We can explain this by the fact that each classifier in boosting change may be built using different class ratios, thus further reinforcing the small sample size problem for minority classes observed for static imbalance. This allows us to conclude that boosting-based ensembles are not best suited for handling difficult imbalanced streams. Bagging-based and hybrid architectures perform significantly better, with bagging being a dominating solution. It is very interesting to see that regardless of the used skew-insensitive mechanism bagging-based ensembles (or hybrid architectures like \acrshort{rose} that utilize bagging) deliver superior performance. This can be explained by the diversity among base classifiers that allow for anticipating different local characteristics of decision boundaries. Therefore, with increasing or decreasing of the imbalance ratio there is a high chance that some of the base classifiers (and thus a subset of instances that they use) offer better generalization and faster adaptation to evolving disproportions between classes.

\noindent \textit{Impact of drift speed in class imbalance.} We can observe that most of the examined algorithms offer similar performance on all types of drifts. Some of the methods do not have explicit mechanisms for change adaptation and this leads to their slower recovery from changes. However, in the long run, there were no significant differences between sudden and gradual drift adaptations for all methods on G-Mean or Kappa. However, the third analyzed scenario with flipping the majority and minority classes has a major impact on analyzed classifiers. \acrshort{arf} significantly suffers on both metrics, showing that its adaptation mechanisms are not suitable for settings where classes can change roles over time. The same observation holds for \acrshort{csarf} that displays an increasing gap between performances on majority and minority classes, as it is not able to effectively adapt its cost matrix to such changes and penalizes wrong class over time. Considering only G-Mean, flipping classes did not impact much \acrshort{uob} demonstrating that undersampling displays potential robustness to switching minority class. However, the same cannot be said for the Kappa metric, leading us to conclude that \acrshort{uob} tends to prioritize one class even when their roles flip. \acrshort{rose} was the most robust and stable classifier for all possible changes in class distribution, being capable of avoiding huge drops of performance for the metrics due to its per-class sliding window adaptation.






\begin{table*}[t!]
\centering
\footnotesize
\setlength{\tabcolsep}{3pt}
\caption{G-Mean and Kappa averages of all 10 streams on dynamic class imbalance ratio.}
\label{tab:BC_DIR}
\begin{tabular}{lll|C{1cm}C{1cm}C{1cm}C{1cm}C{1cm}C{1cm}C{1cm}C{1cm}C{1cm}C{1cm}}
\toprule
& & IR & CSARF & ARF & KUE & LB & CALMID & ROSE & ARFR & SMOTE-OB & OOB & UOB\\
\midrule
\multirow{8}{*}{\rotatebox[origin=c]{90}{G-Mean}} & \multirow{4}{*}{\rotatebox[origin=c]{90}{Gradual}} 
& Increasing & 89.21 & 80.87 & 83.60 & 82.52 & 82.44 & 87.83 & 85.62 & 88.42 & \textbf{90.88} & 90.76\\
& & Inc. then dec. & 89.57 & 76.18 & 80.57 & 82.08 & 81.49 & 87.33 & 83.11 & 85.61 & 89.05 & \textbf{90.17}\\
& & Flipping & 84.59 & 64.34 & 71.08 & 69.73 & 68.01 & 78.32 & 65.67 & 75.83 & 80.33 & \textbf{87.78}\\
& & Reflipling & 84.28 & 64.72 & 75.11 & 72.77 & 70.23 & 80.39 & 64.88 & 75.23 & 82.30 & \textbf{87.35}\\
\cmidrule{2-13}
& \multirow{4}{*}{\rotatebox[origin=c]{90}{Sudden}}
& Increasing & 89.16 & 81.35 & 83.75 & 83.80 & 82.36 & 88.00 & 84.40 & 85.61 & 90.89 & \textbf{90.78}\\
& & Inc. then dec. & 89.88 & 77.16 & 81.82 & 83.10 & 82.32 & 87.96 & 83.41 & 86.62 & 89.30 & \textbf{90.47}\\
& & Flipping & 84.78 & 65.41 & 72.15 & 71.35 & 69.09 & 79.61 & 66.66 & 77.52 & 80.40 & \textbf{87.89}\\
& & Reflipling & 84.69 & 66.12 & 76.10 & 74.09 & 71.74 & 81.37 & 66.23 & 77.26 & 82.22 & \textbf{87.41}\\
\midrule
\multirow{8}{*}{\rotatebox[origin=c]{90}{Kappa}} & \multirow{4}{*}{\rotatebox[origin=c]{90}{Gradual}} 
& Increasing & 54.08 & 71.49 & 72.98 & 72.68 & 73.38 & \textbf{73.96} & 73.61 & 71.47 & 68.44 & 58.68\\
& & Inc. then dec. & 59.21 & 67.50 & 71.08 & 73.14 & 72.95 & \textbf{74.98} & 71.47 & 69.89 & 69.87 & 61.81\\
& & Flipping & 45.14 & 56.97 & 62.00 & 61.40 & 60.20 & \textbf{66.97} & 50.51 & 61.61 & 61.73 & 51.76\\
& & Reflipling & 43.43 & 56.80 & 61.91 & 62.61 & 61.36 & \textbf{67.64} & 52.78 & 60.64 & 60.86 & 52.43\\
\cmidrule{2-13}
& \multirow{4}{*}{\rotatebox[origin=c]{90}{Sudden}} 
& Increasing & 54.48 & 72.50 & 73.53 & 74.45 & 73.98 & \textbf{74.46} & 73.87 & 69.93 & 68.26 & 58.46\\
& & Inc. then dec. & 59.87 & 68.85 & 72.47 & 74.70 & 74.39 & \textbf{75.82} & 72.43 & 71.44 & 69.51 & 61.50\\
& & Flipping & 45.16 & 58.50 & 63.31 & 63.51 & 61.54 & \textbf{68.52} & 51.93 & 63.23 & 62.09 & 51.09\\
& & Reflipling & 43.82 & 58.28 & 62.91 & 64.47 & 62.96 & \textbf{69.19} & 54.16 & 62.61 & 60.66 & 51.81\\
\midrule
\multicolumn{3}{l|}{Avg. G-Mean} & 87.02 & 72.02 & 78.02 & 77.43 & 75.96 & 83.85 & 75.00 & 81.51 & 85.67 & \textbf{89.08}\\
\multicolumn{3}{l|}{Avg. Kappa} & 50.65 & 63.86 & 67.53 & 68.37 & 67.59 & \textbf{71.44} & 62.59 & 66.35 & 65.18 & 55.94\\
\midrule
\multicolumn{3}{l|}{Rank G-Mean} & 3.83 & 8.62 & 7.19 & 5.83 & 6.85 & 4.11 & 6.66 & 5.11 & 3.80 & \textbf{3.01}\\

\multicolumn{3}{l|}{Rank Kappa} & 8.58 & 6.31 & 5.14 & 4.41 & 4.68 & \textbf{2.80} & 6.41 & 4.93 & 4.64 & 7.12\\
\bottomrule
\end{tabular}
\end{table*}

\newpage
\subsubsection{Instance-level difficulties}
\label{sec:bc-instance-level-diff}

\noindent \textbf{Goal of the experiment.}
This experiment addresses \textbf{RQ3} and evaluates the robustness of the classifiers to instance-level difficulties \citep{Brzezinski2021}. We evaluated the Brzezi{\'n}ski generator with borderline or rare instances, and combining both at the same time. The ratio for difficult instances for scenarios where there are only rare or borderline instances are \{0\%, 20\%, 40\%, 60\%, 80\%, 100\%\}. In the combined scenario they represent \{0\%, 20\%, 40\%\} of rare and borderline instances, e.g. 20\% means there are 20\% rare instances and 20\% borderline instances. Difficult instances were created for the minority class to present a challenging scenario for the classifier. We evaluated the influence on classifiers combined with static and dynamic imbalance ratios. Borderline instances pose a challenge to the classifier because they lie in the uncertainty area of the decision space and strongly impact the induction of the classification boundaries. Rare instances are overlapping with the majority class. Also, instances of minority classes are distributed in clusters. Moving, splitting and merging these clusters leads to new challenges for the classifiers since the decision boundary moves accordingly. 

\begin{figure}[t!]
\centering
\includegraphics[width=0.19\columnwidth]{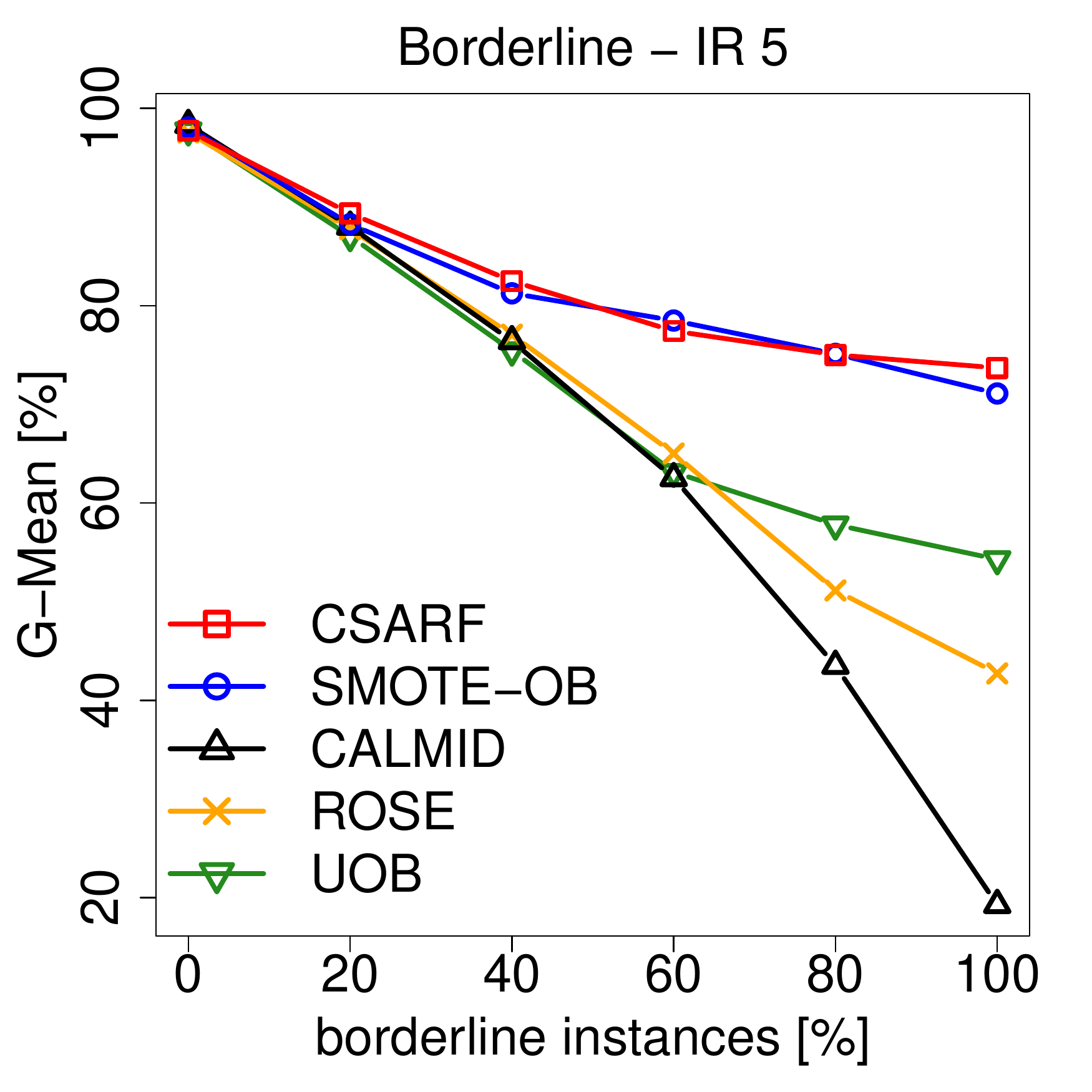}
\includegraphics[width=0.19\columnwidth]{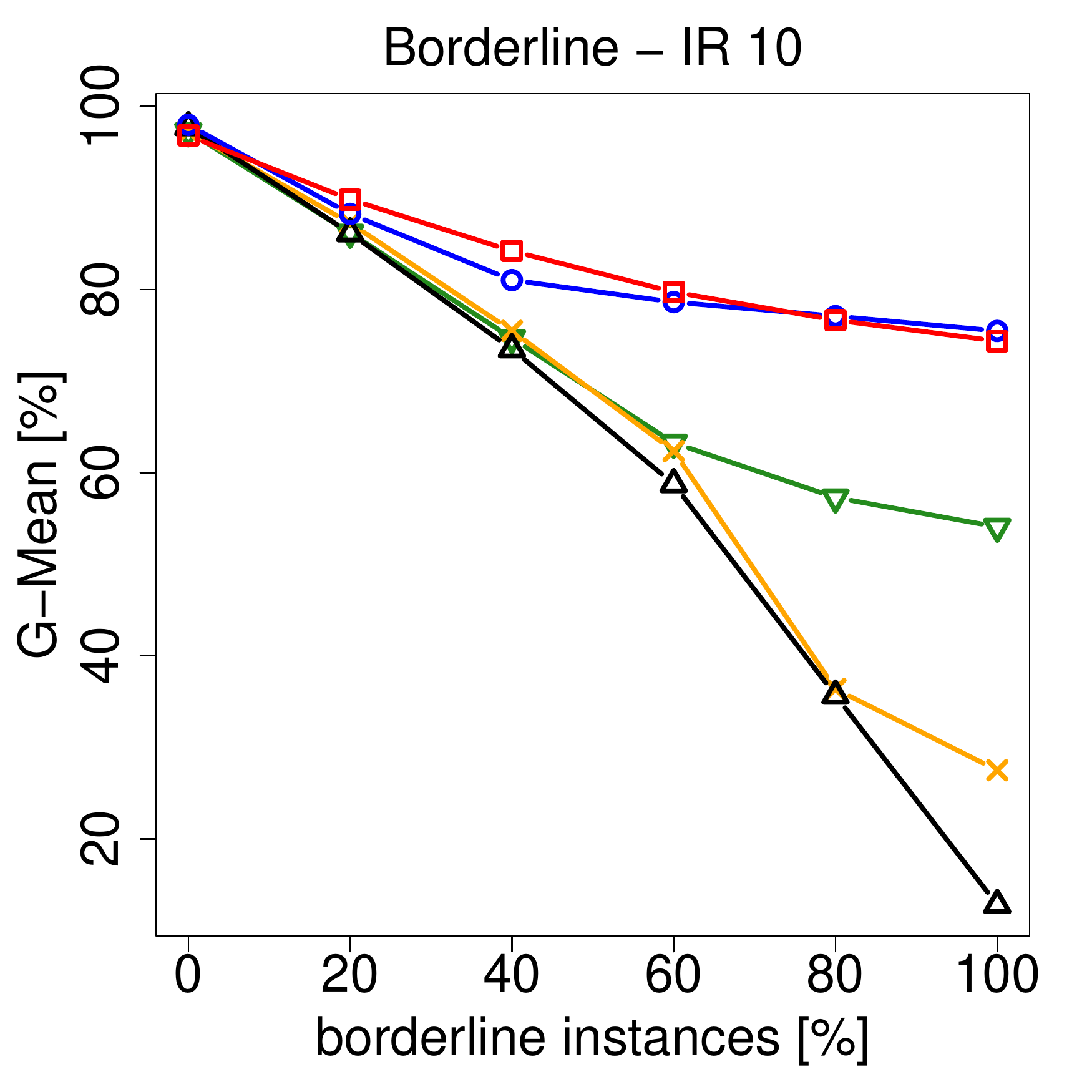}
\includegraphics[width=0.19\columnwidth]{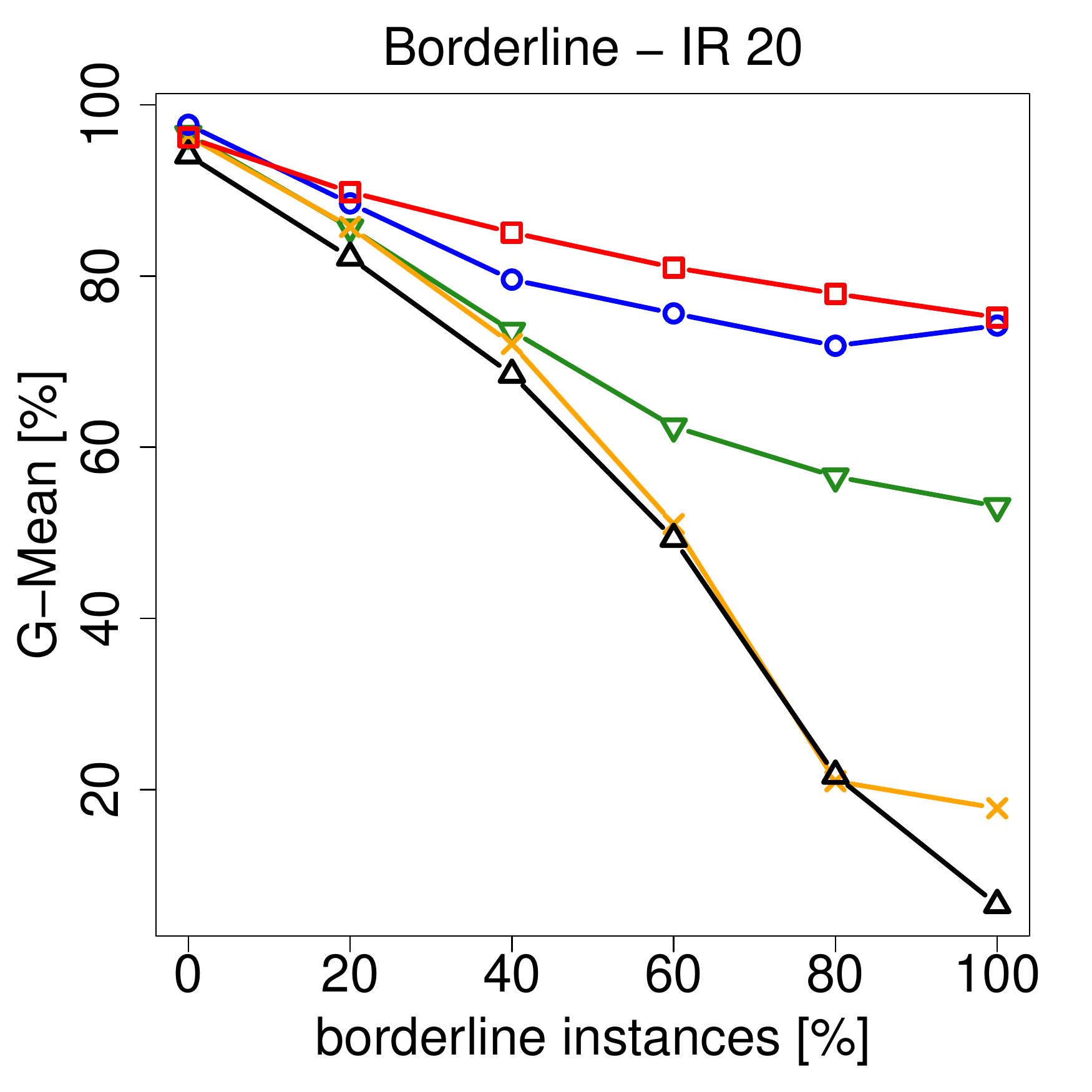}
\includegraphics[width=0.19\columnwidth]{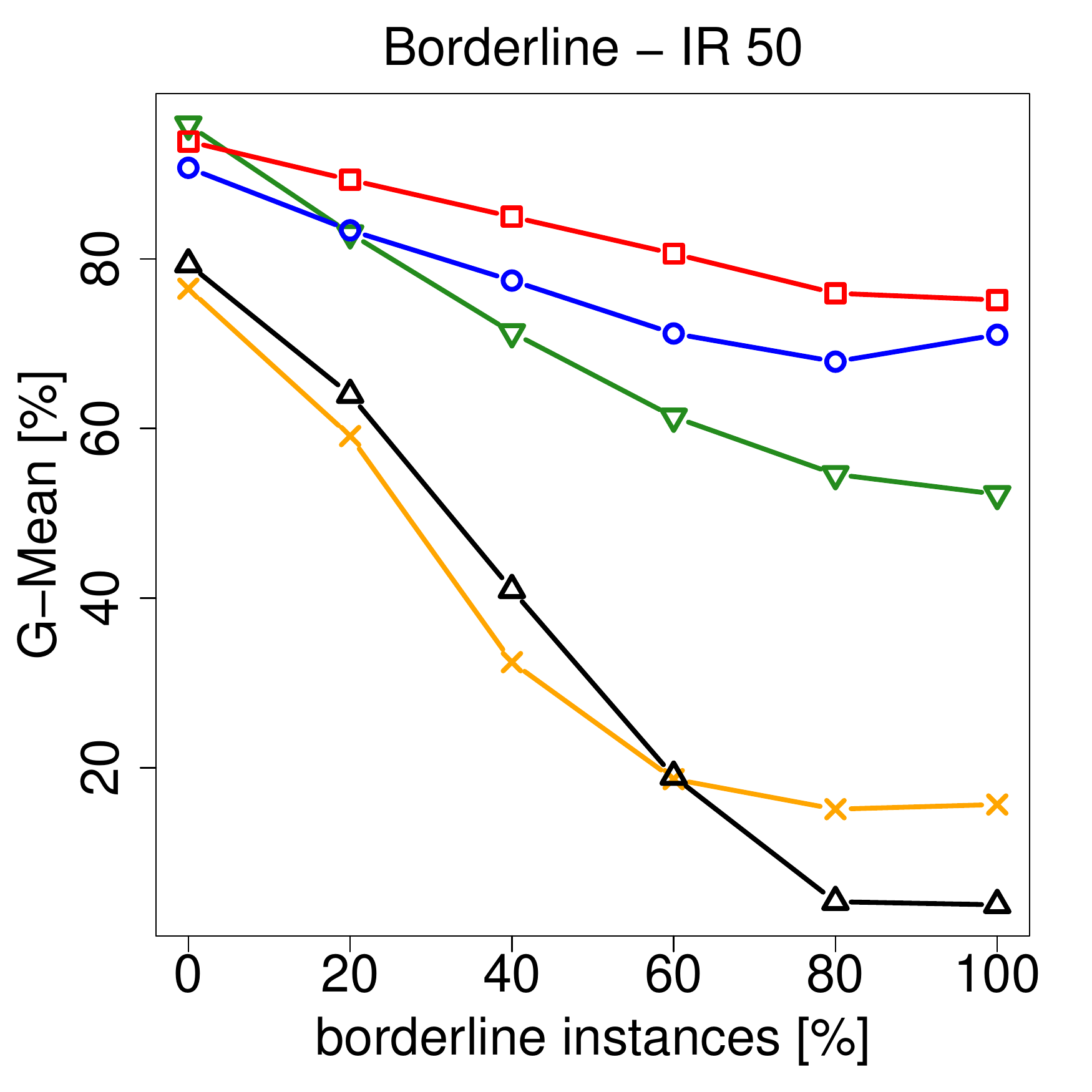}
\includegraphics[width=0.19\columnwidth]{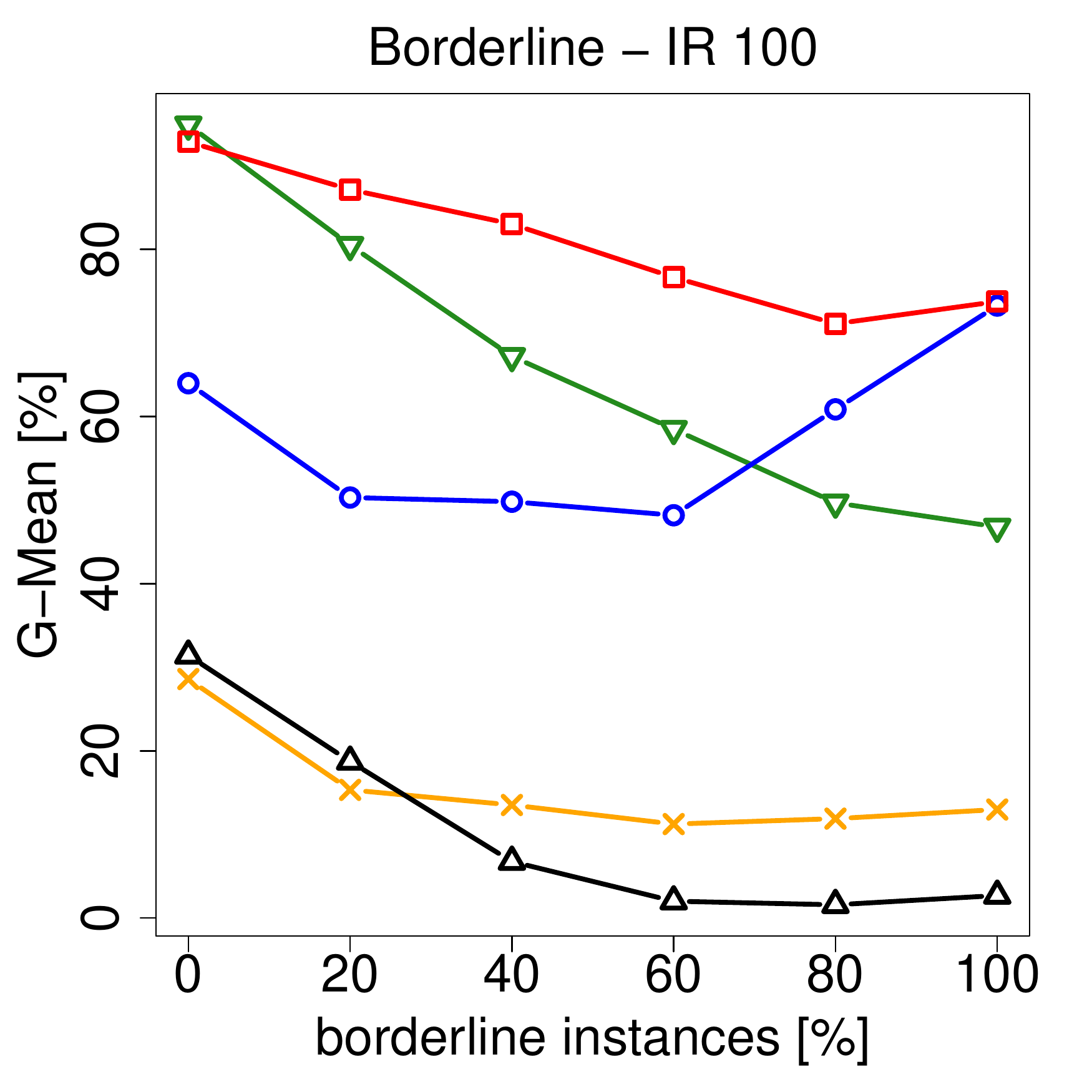}
\includegraphics[width=0.19\columnwidth]{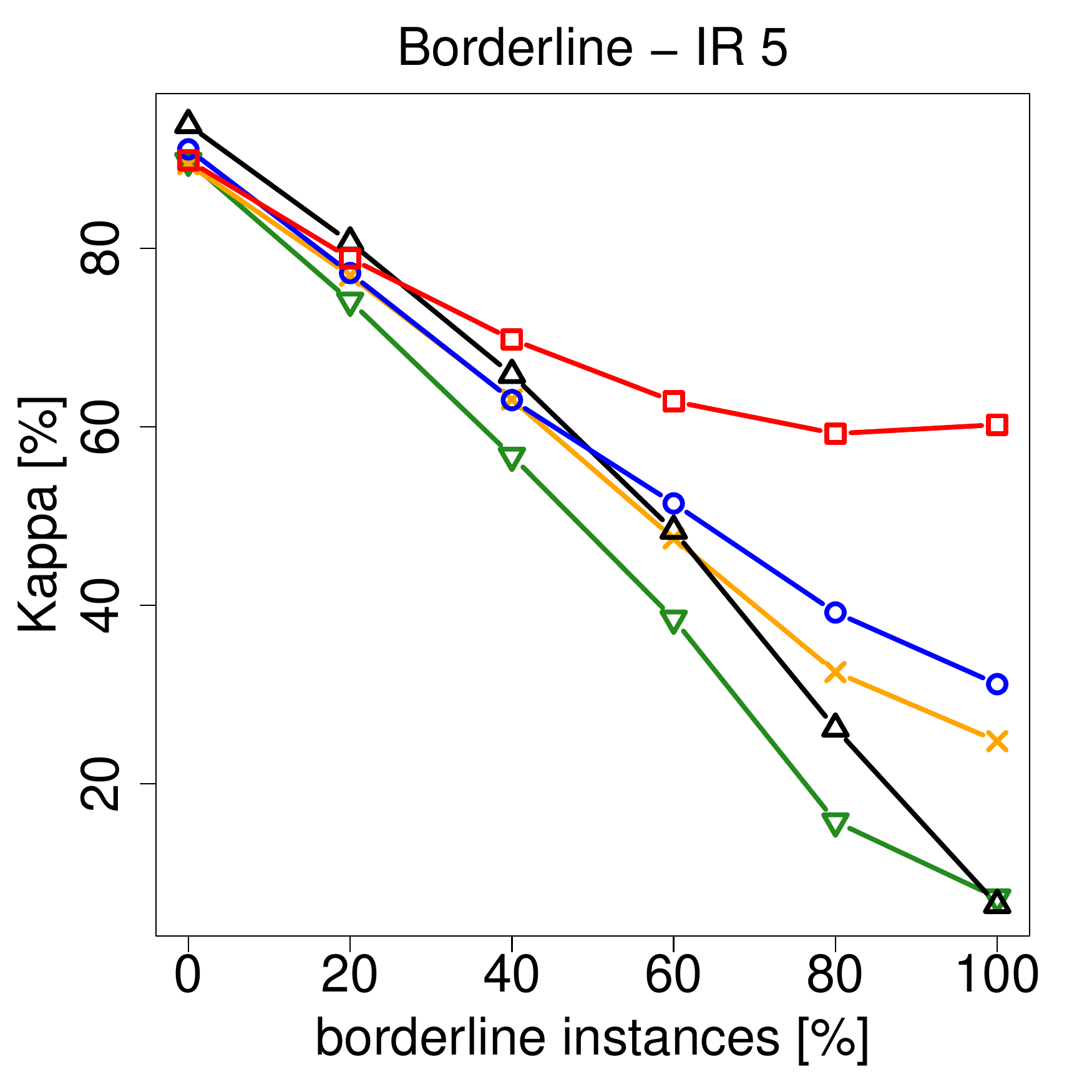}
\includegraphics[width=0.19\columnwidth]{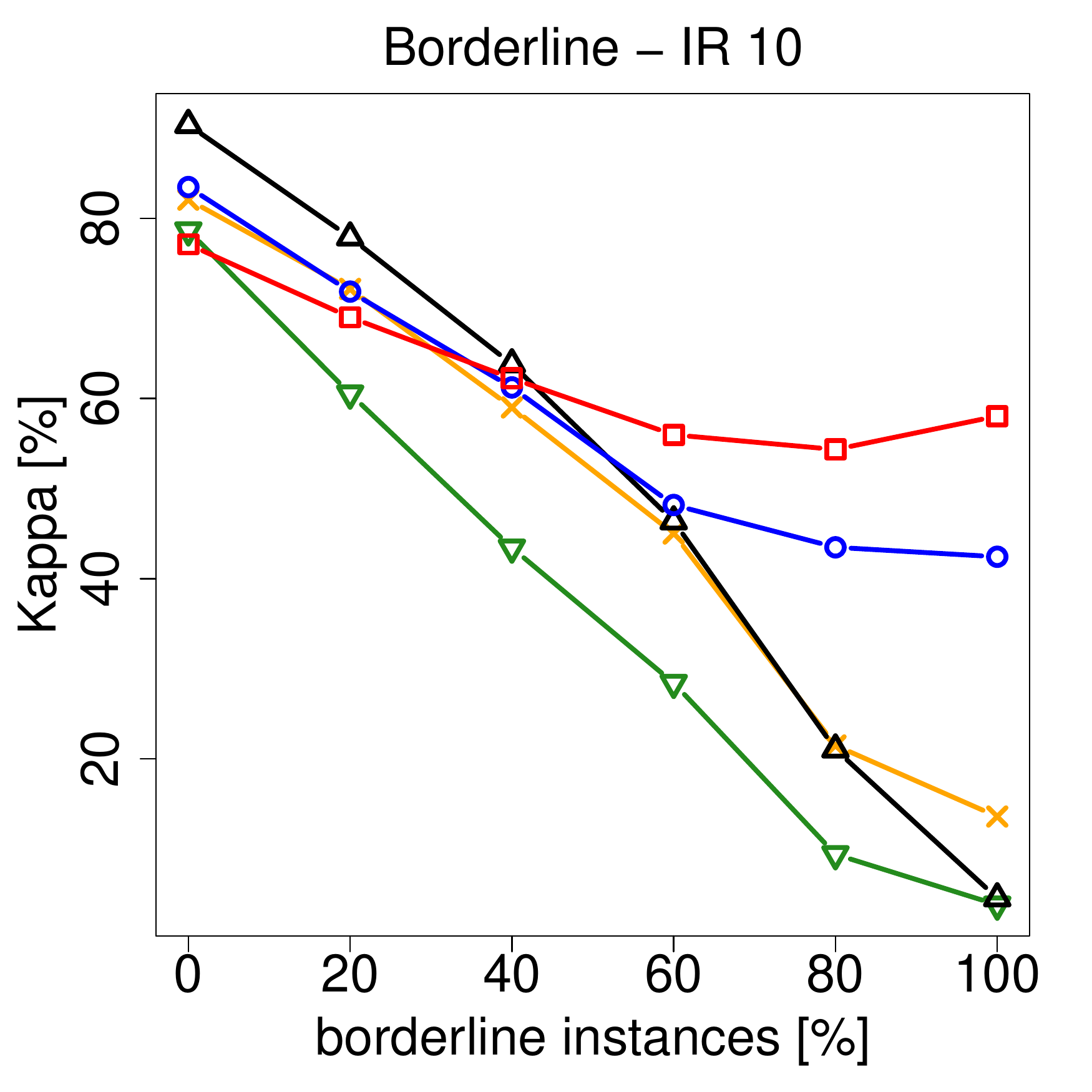}
\includegraphics[width=0.19\columnwidth]{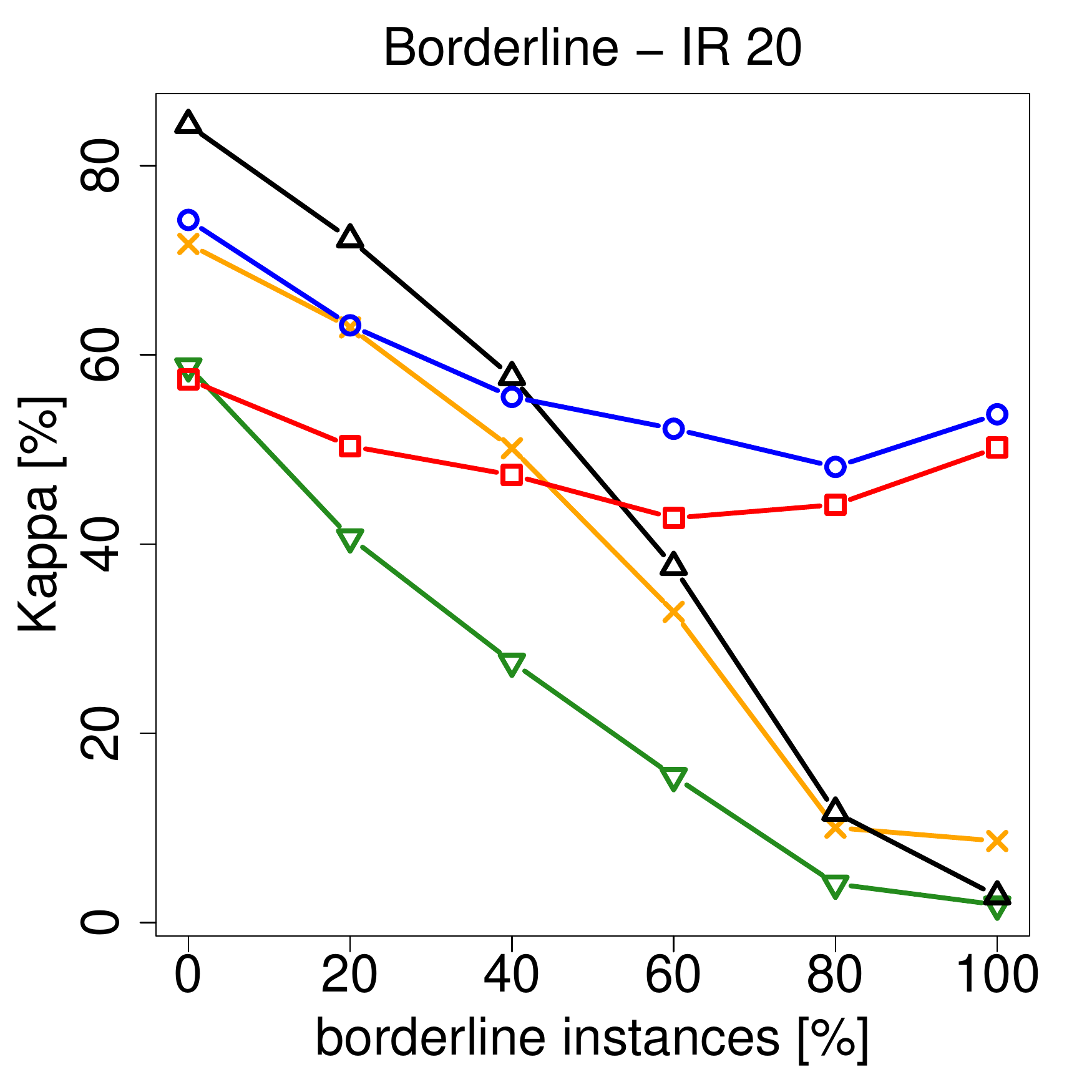}
\includegraphics[width=0.19\columnwidth]{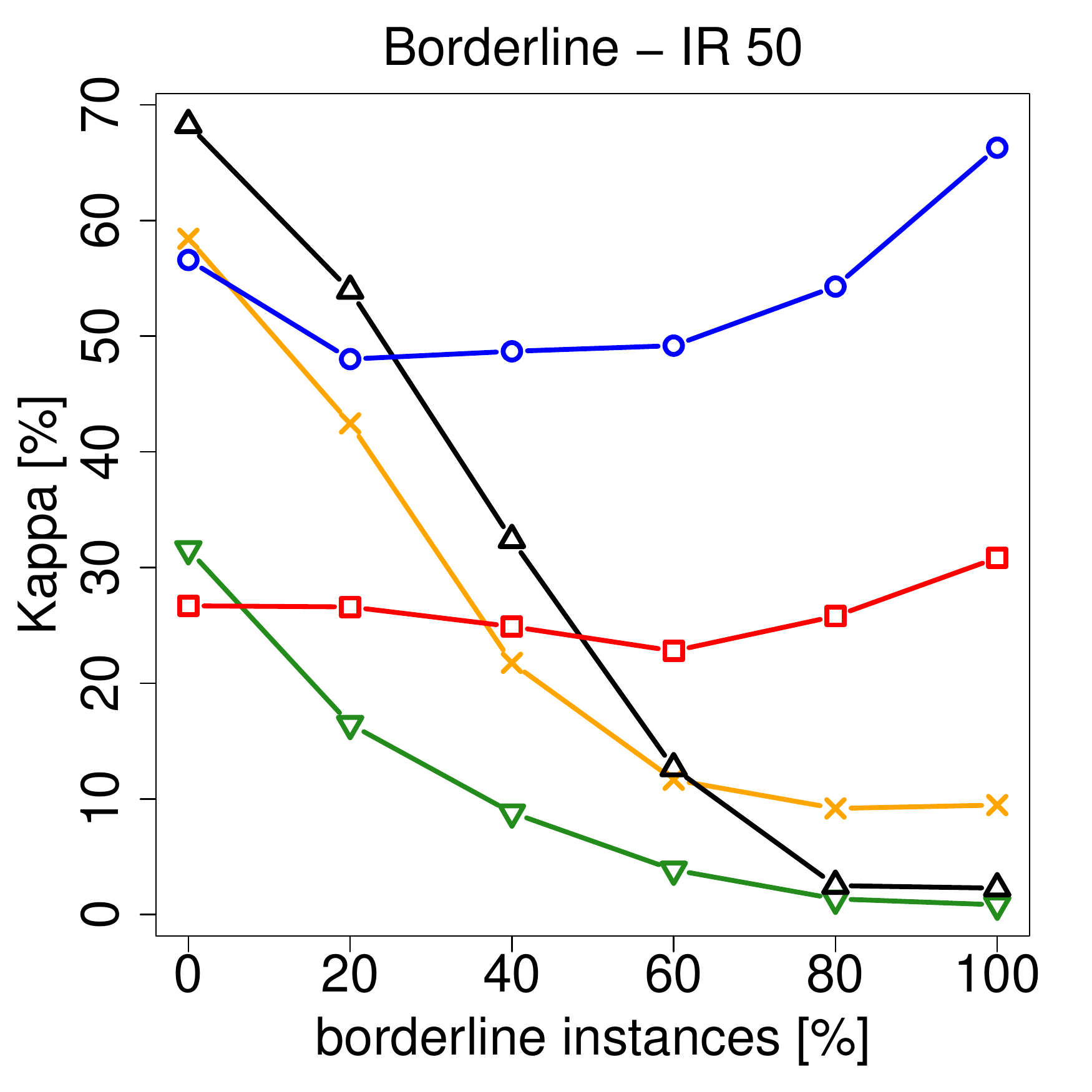}
\includegraphics[width=0.19\columnwidth]{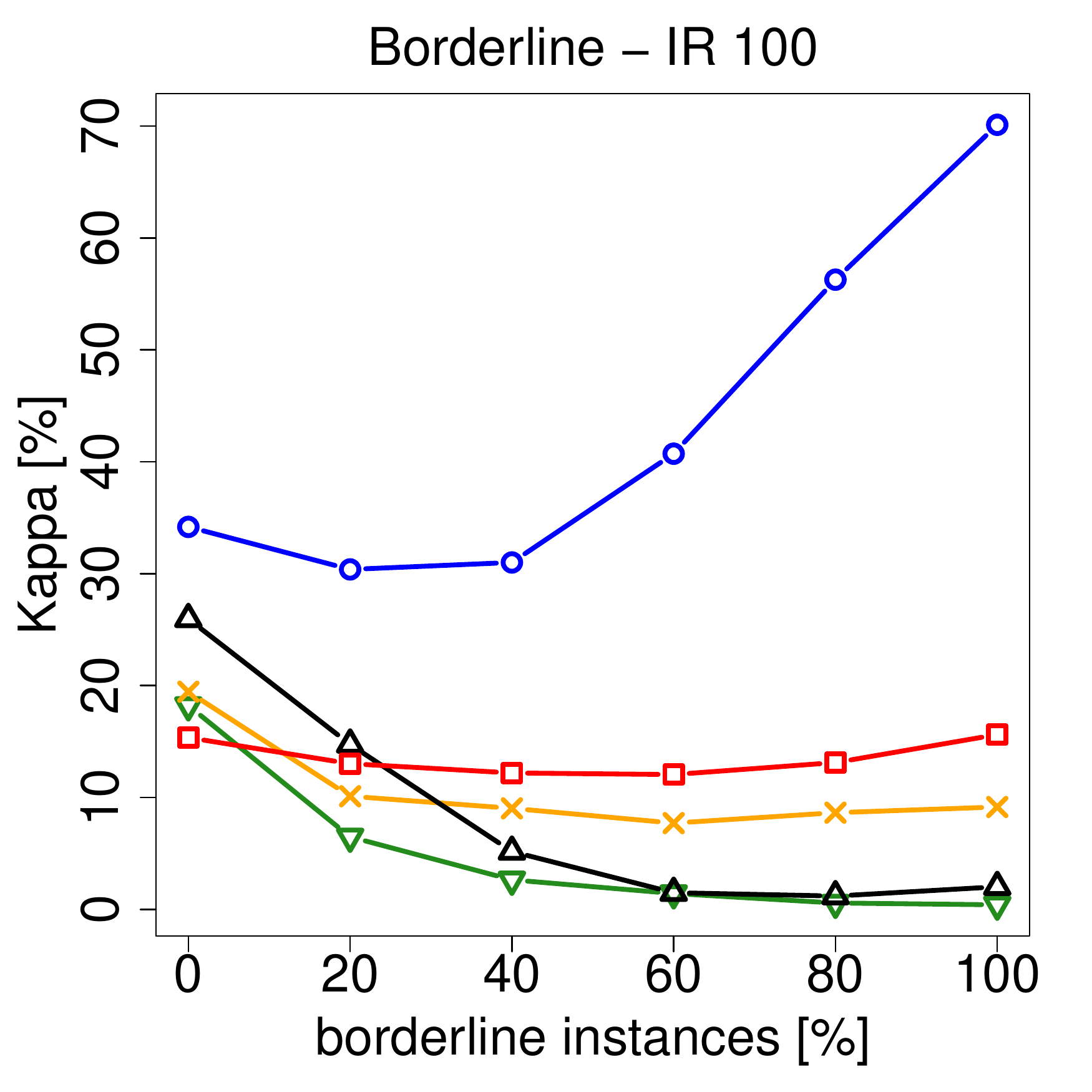}
\caption{Robustness to borderline instances for static imbalance ratio (G-Mean and Kappa).}
\label{fig:ild_static_borderline}
\end{figure}

\begin{figure}[t!]
\centering
\includegraphics[width=0.19\columnwidth]{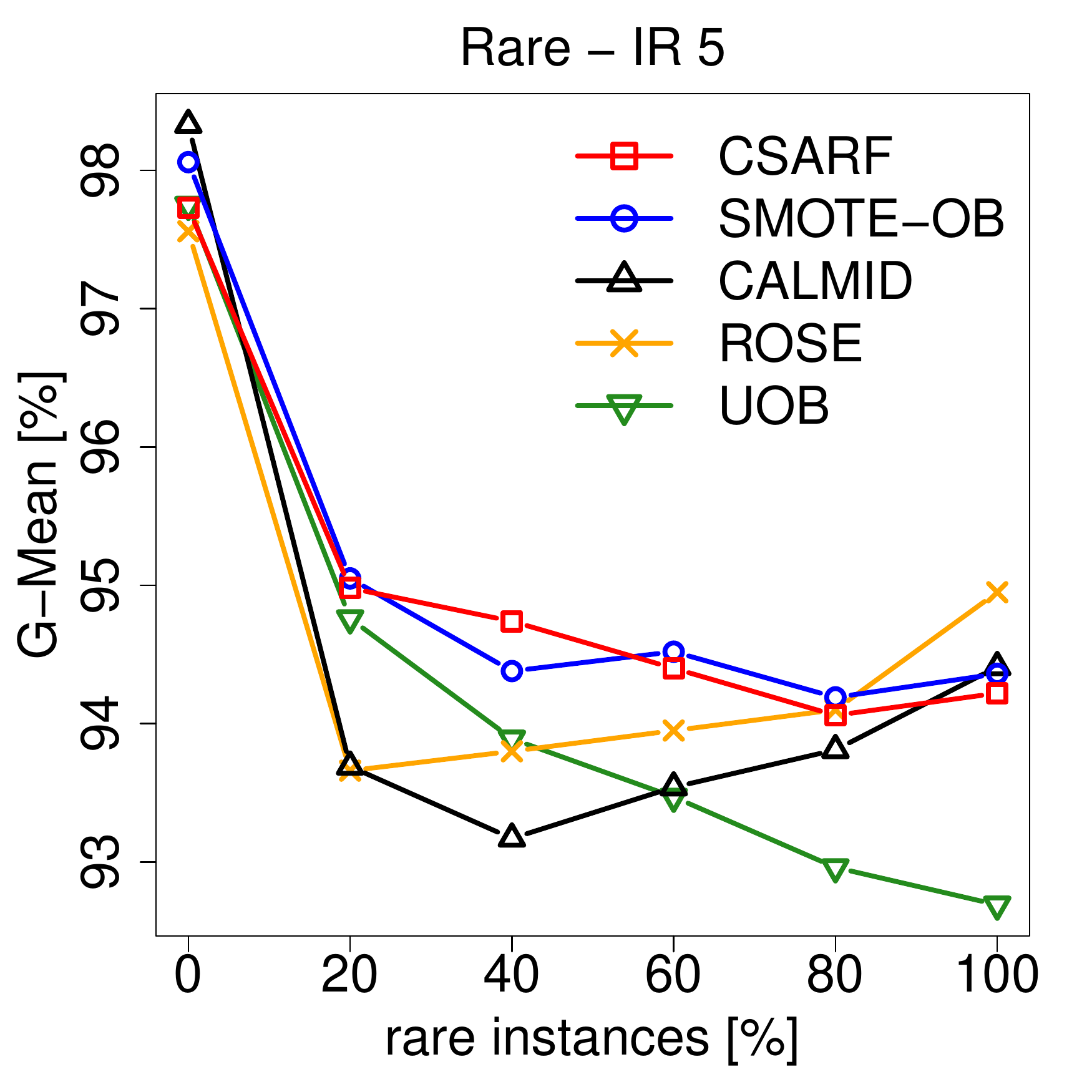}
\includegraphics[width=0.19\columnwidth]{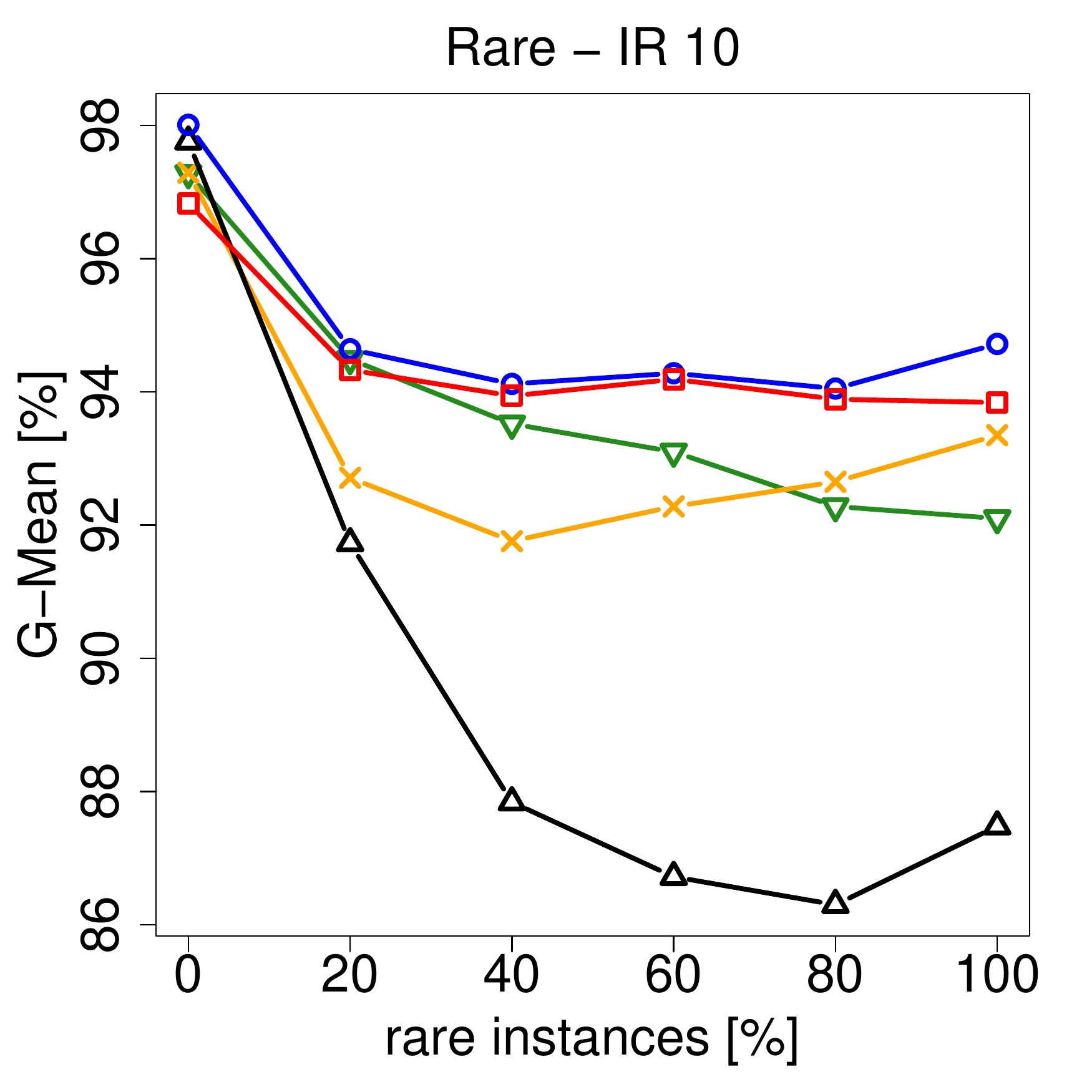}
\includegraphics[width=0.19\columnwidth]{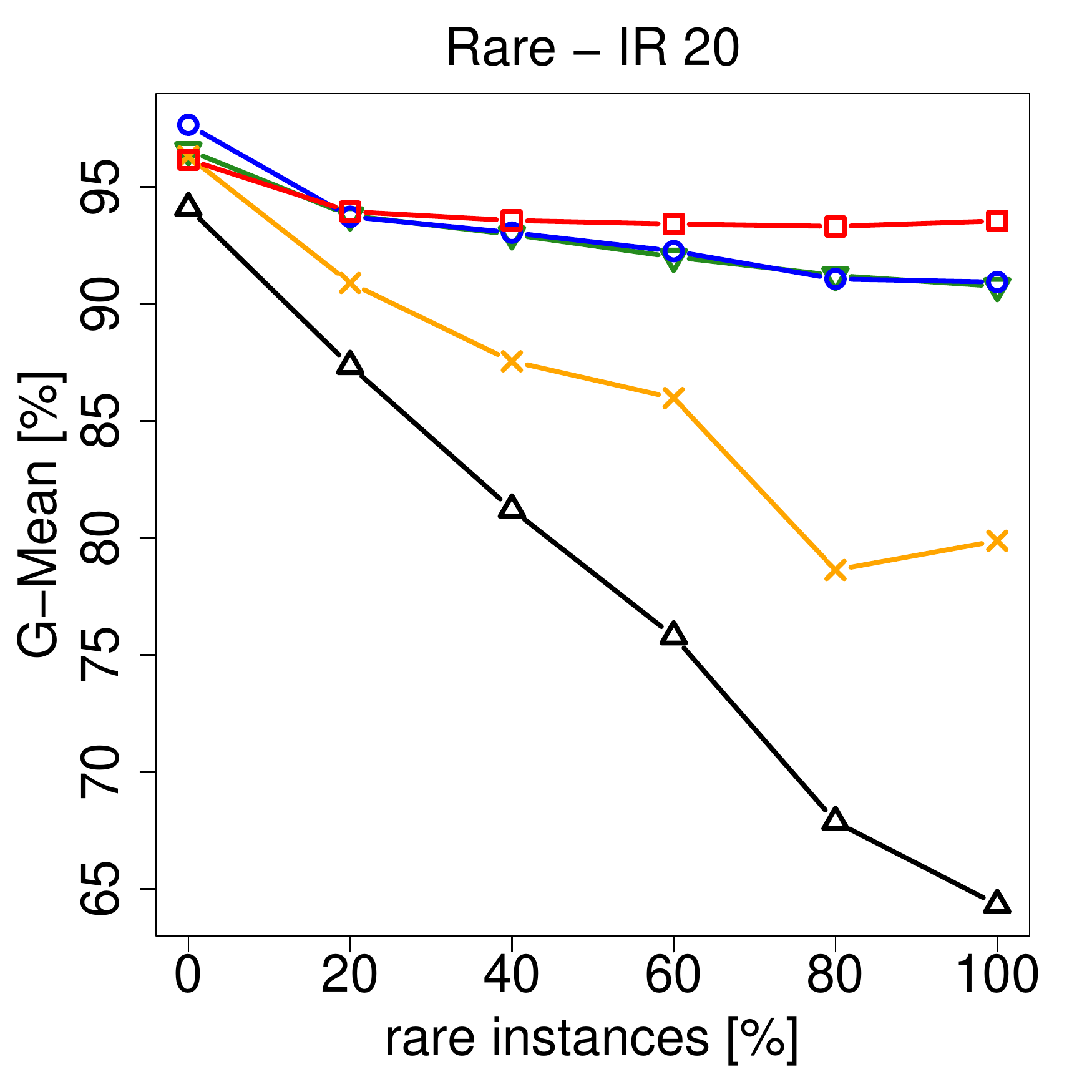}
\includegraphics[width=0.19\columnwidth]{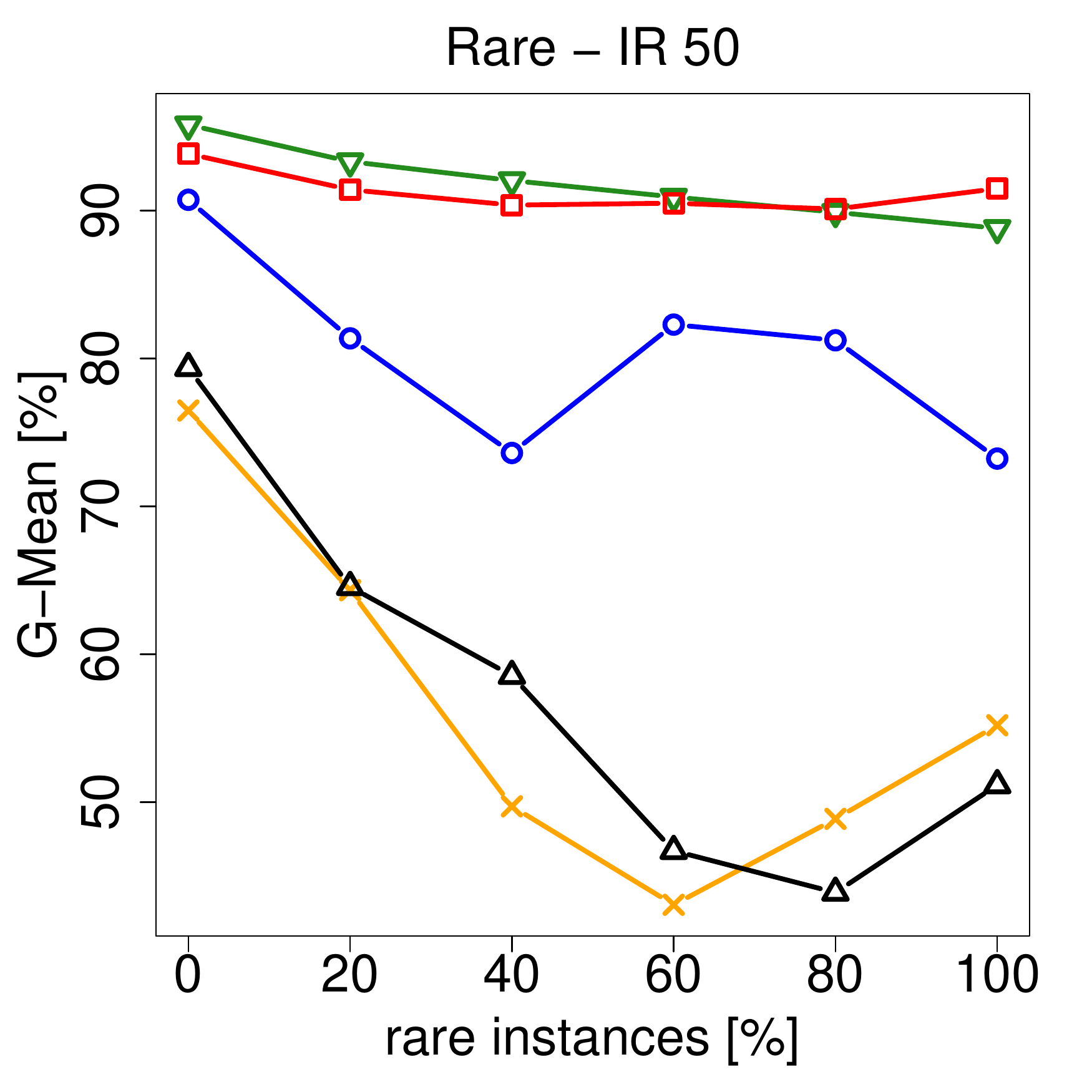}
\includegraphics[width=0.19\columnwidth]{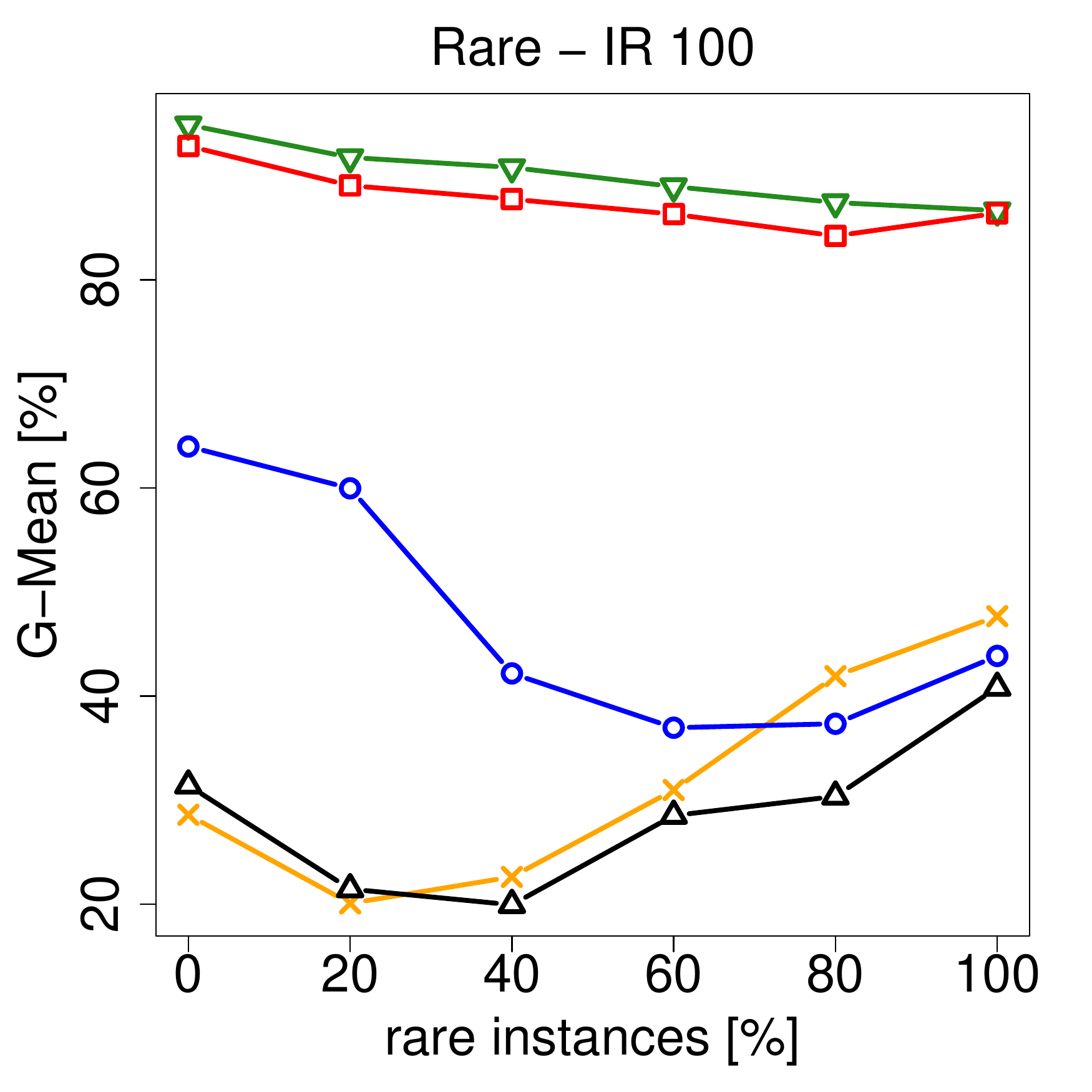}
\includegraphics[width=0.19\columnwidth]{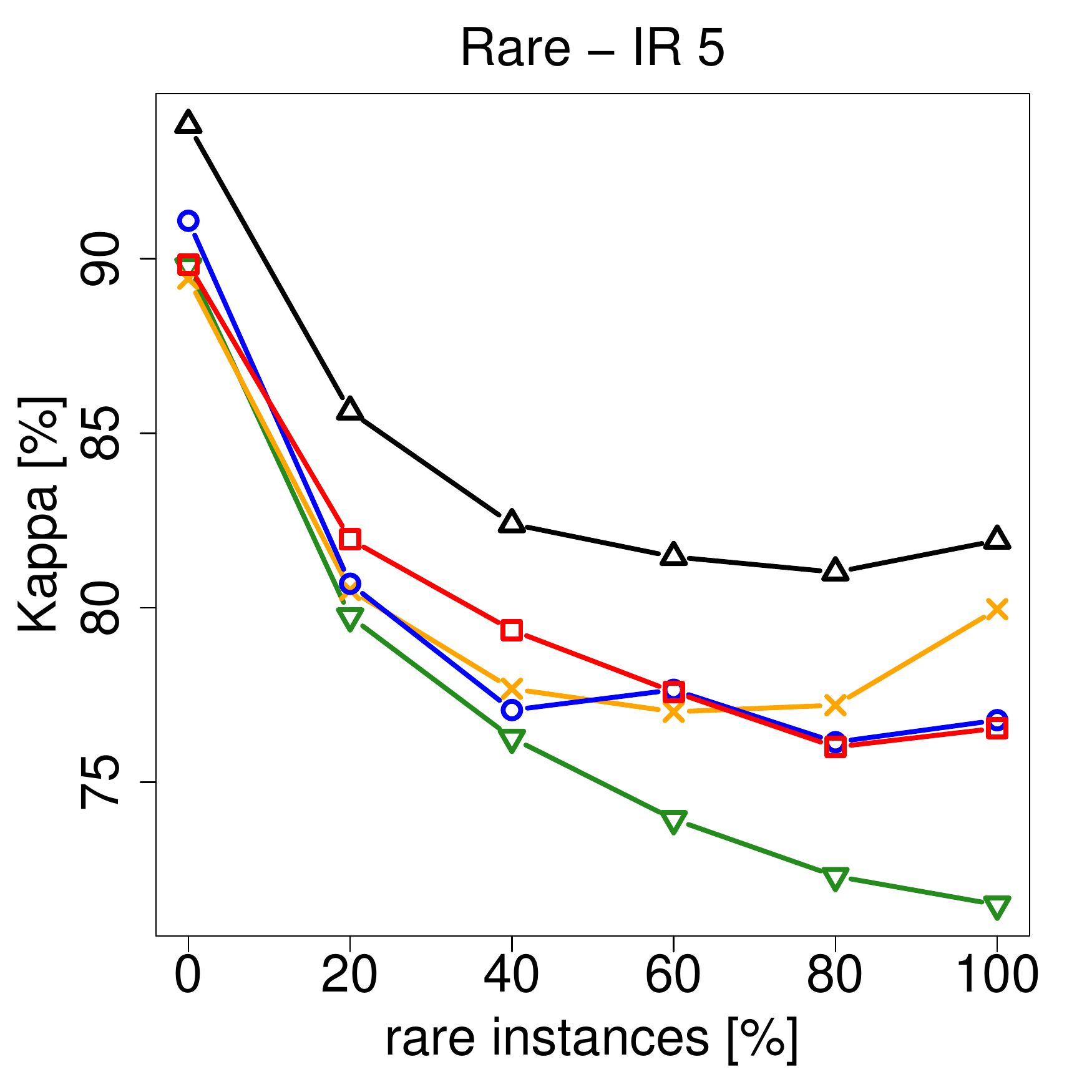}
\includegraphics[width=0.19\columnwidth]{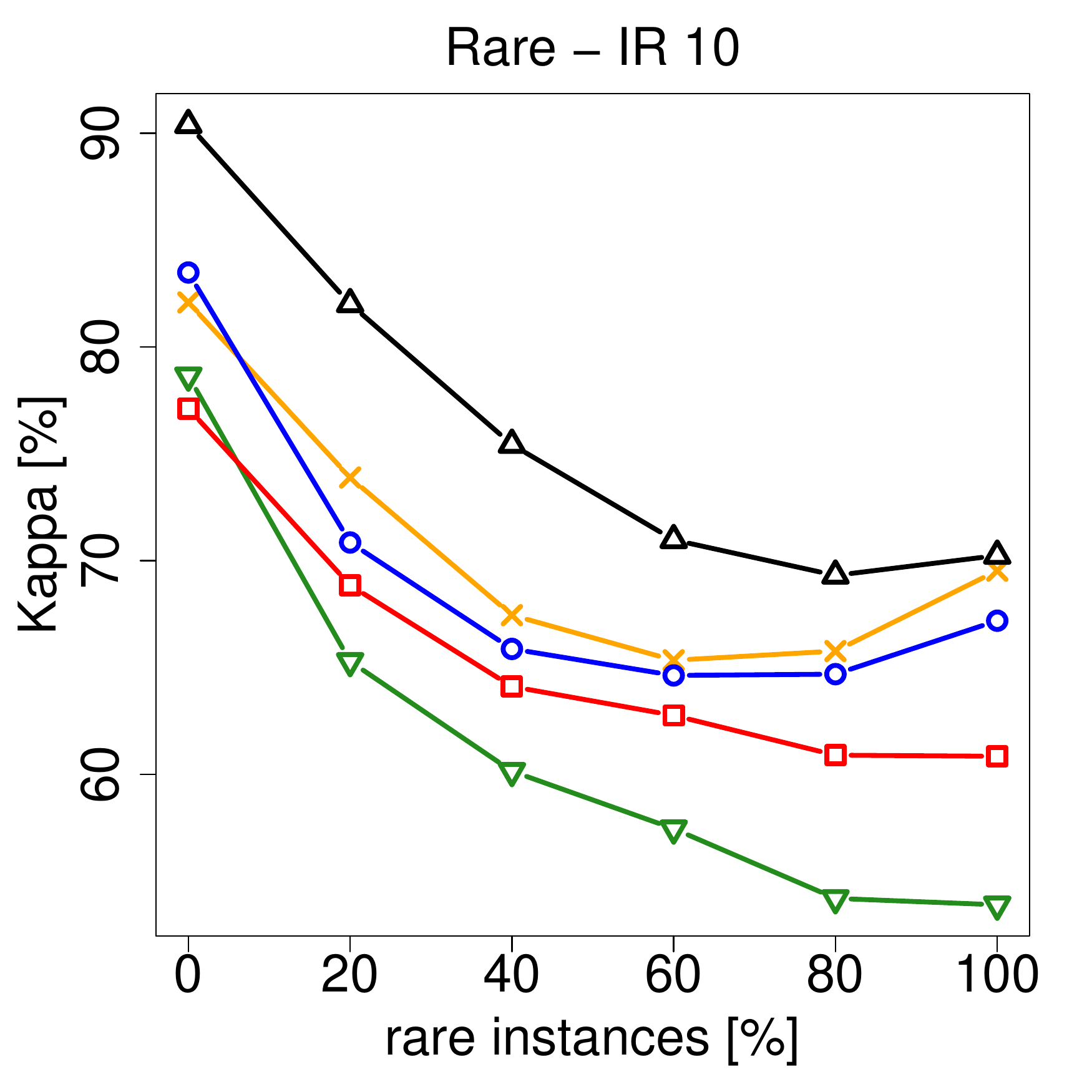}
\includegraphics[width=0.19\columnwidth]{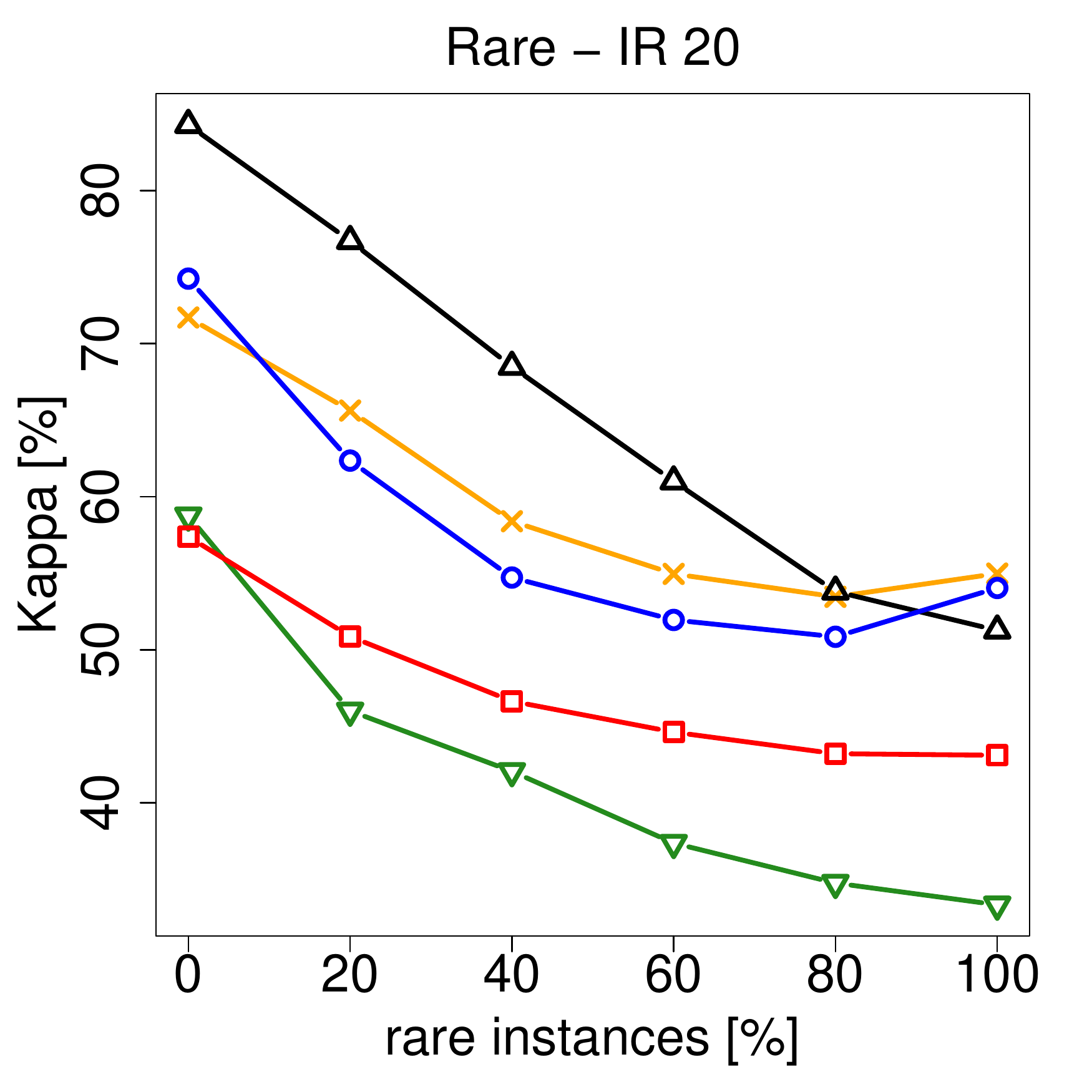}
\includegraphics[width=0.19\columnwidth]{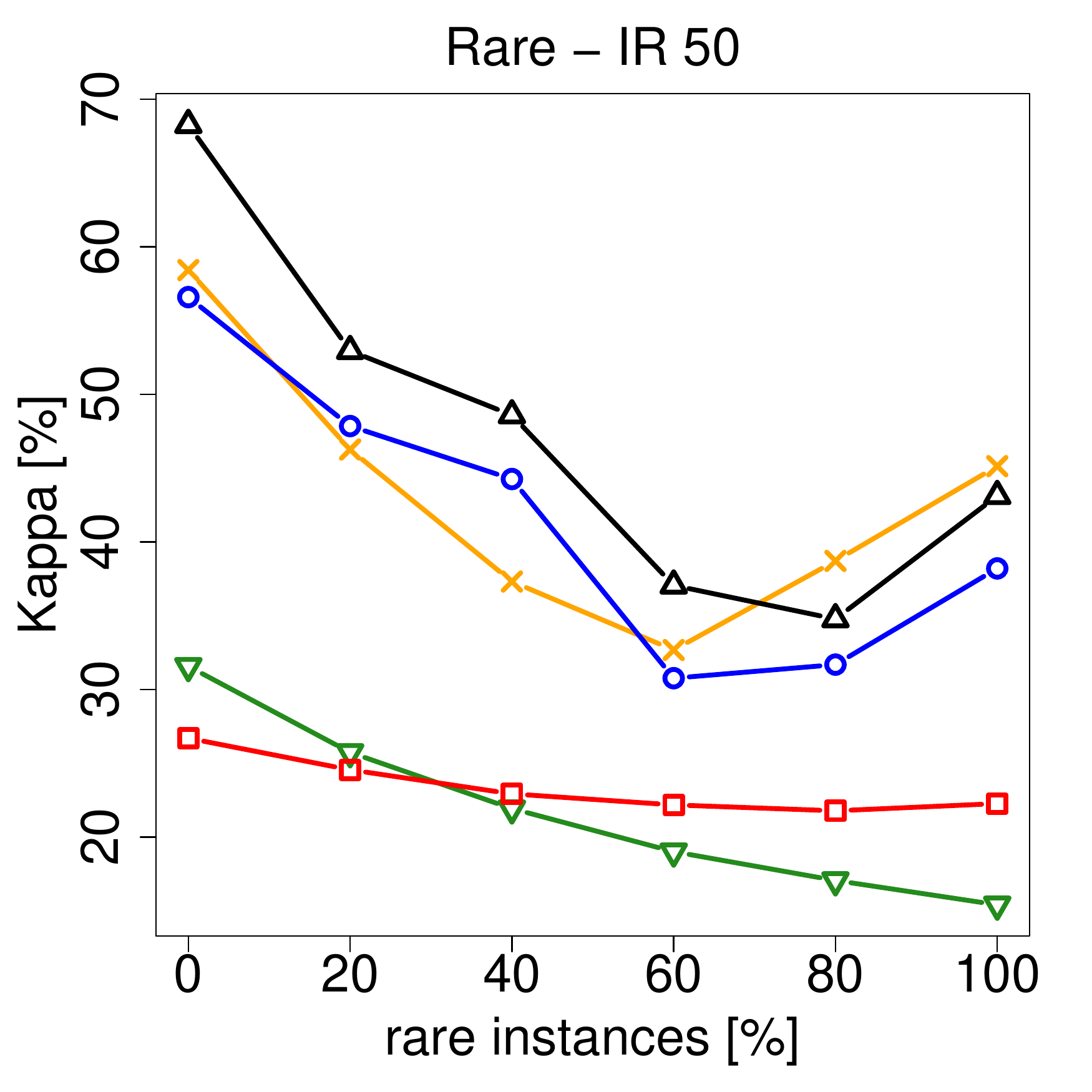}
\includegraphics[width=0.19\columnwidth]{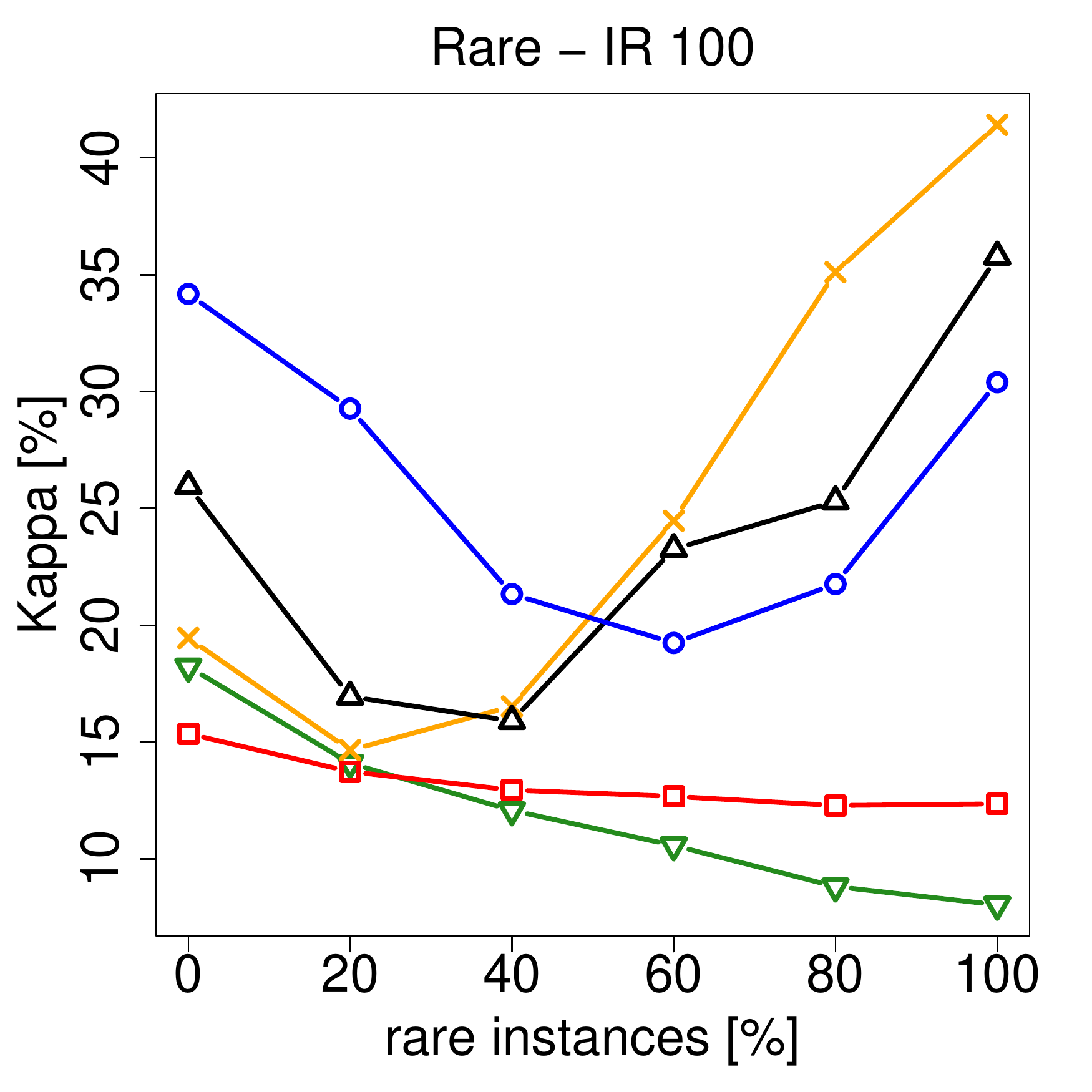}
\caption{Robustness to rare instances for static imbalance ratio (G-Mean and Kappa).}
\label{fig:ild_static_rare}
\end{figure}

\begin{figure}[t!]
\centering
\includegraphics[width=0.19\columnwidth]{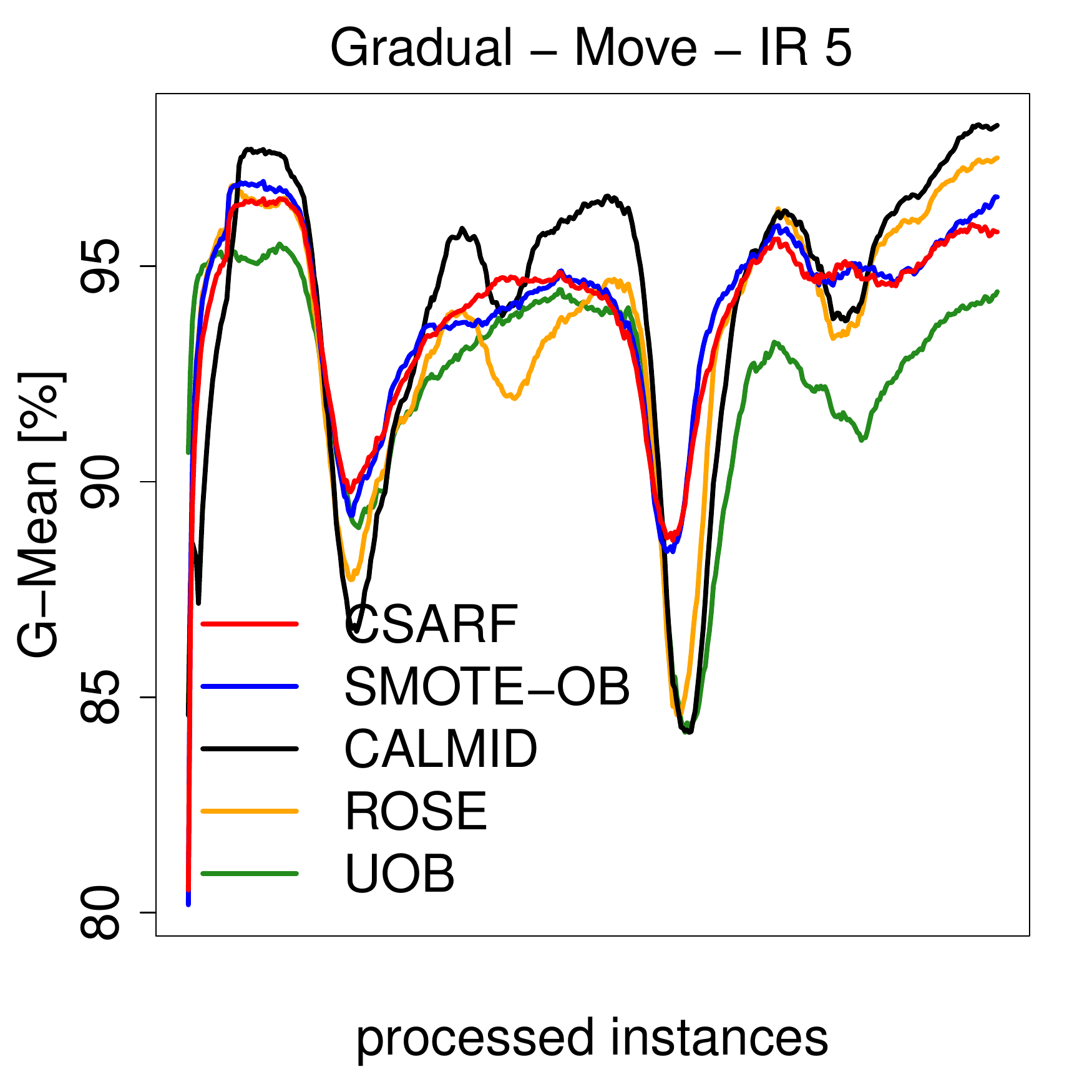}
\includegraphics[width=0.19\columnwidth]{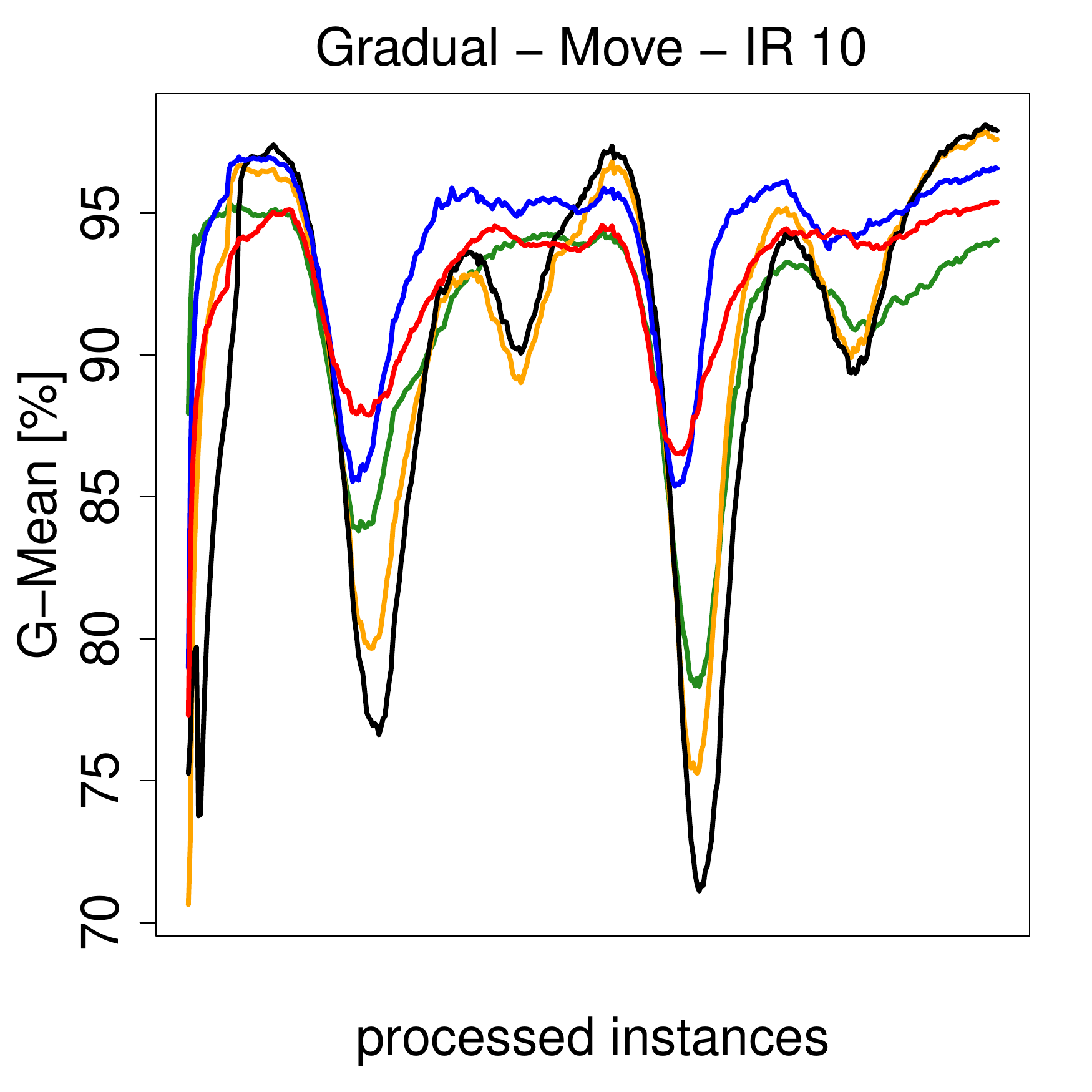}
\includegraphics[width=0.19\columnwidth]{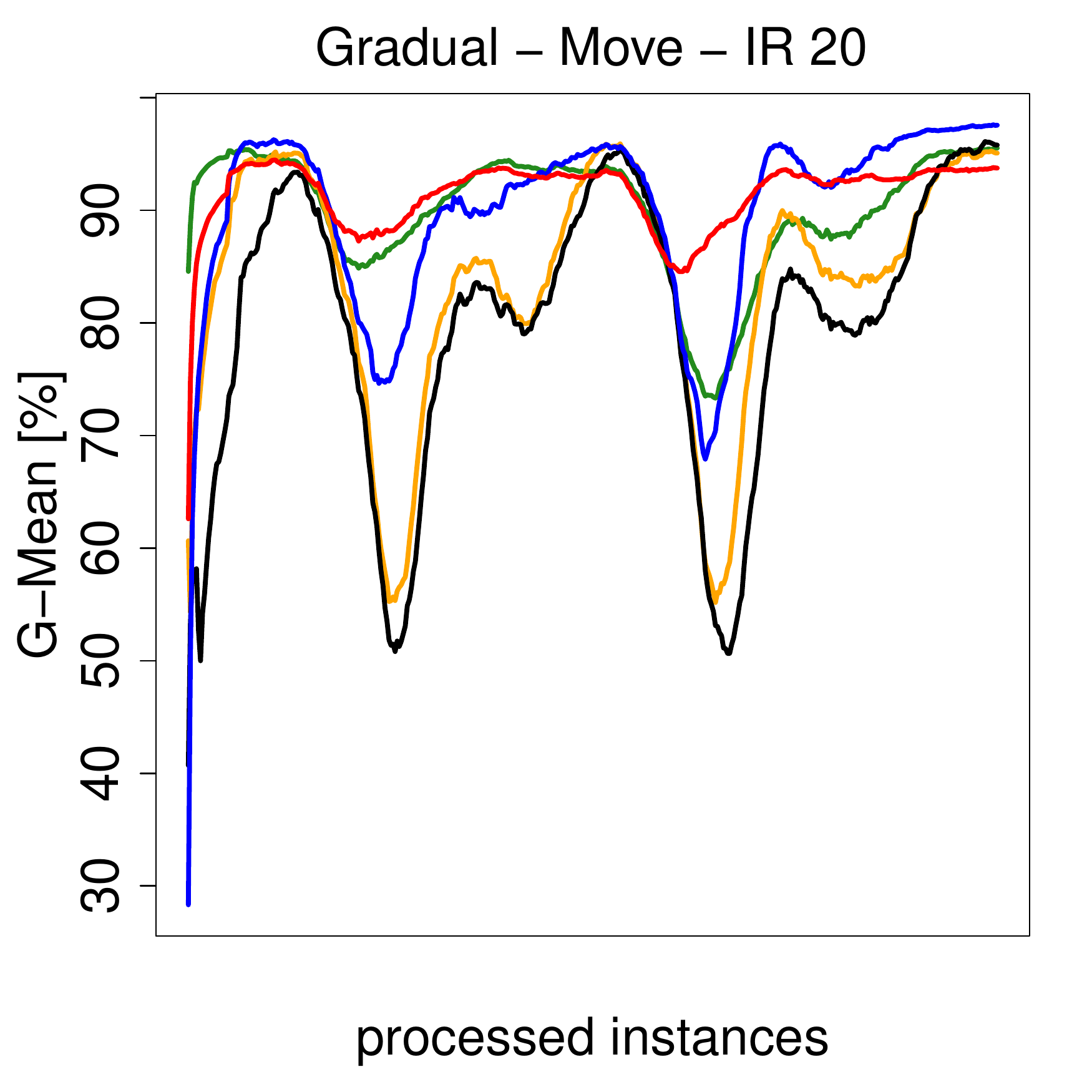}
\includegraphics[width=0.19\columnwidth]{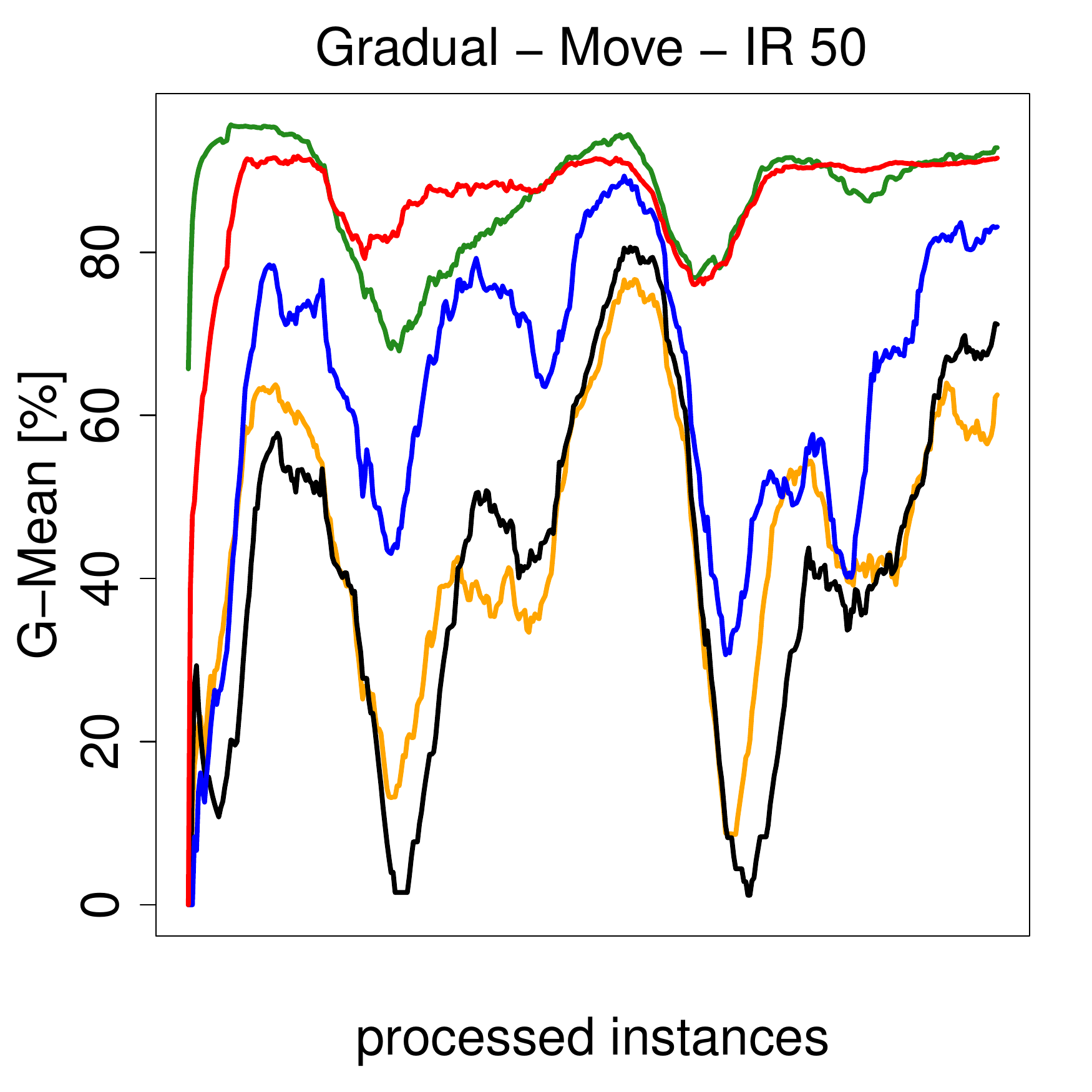}
\includegraphics[width=0.19\columnwidth]{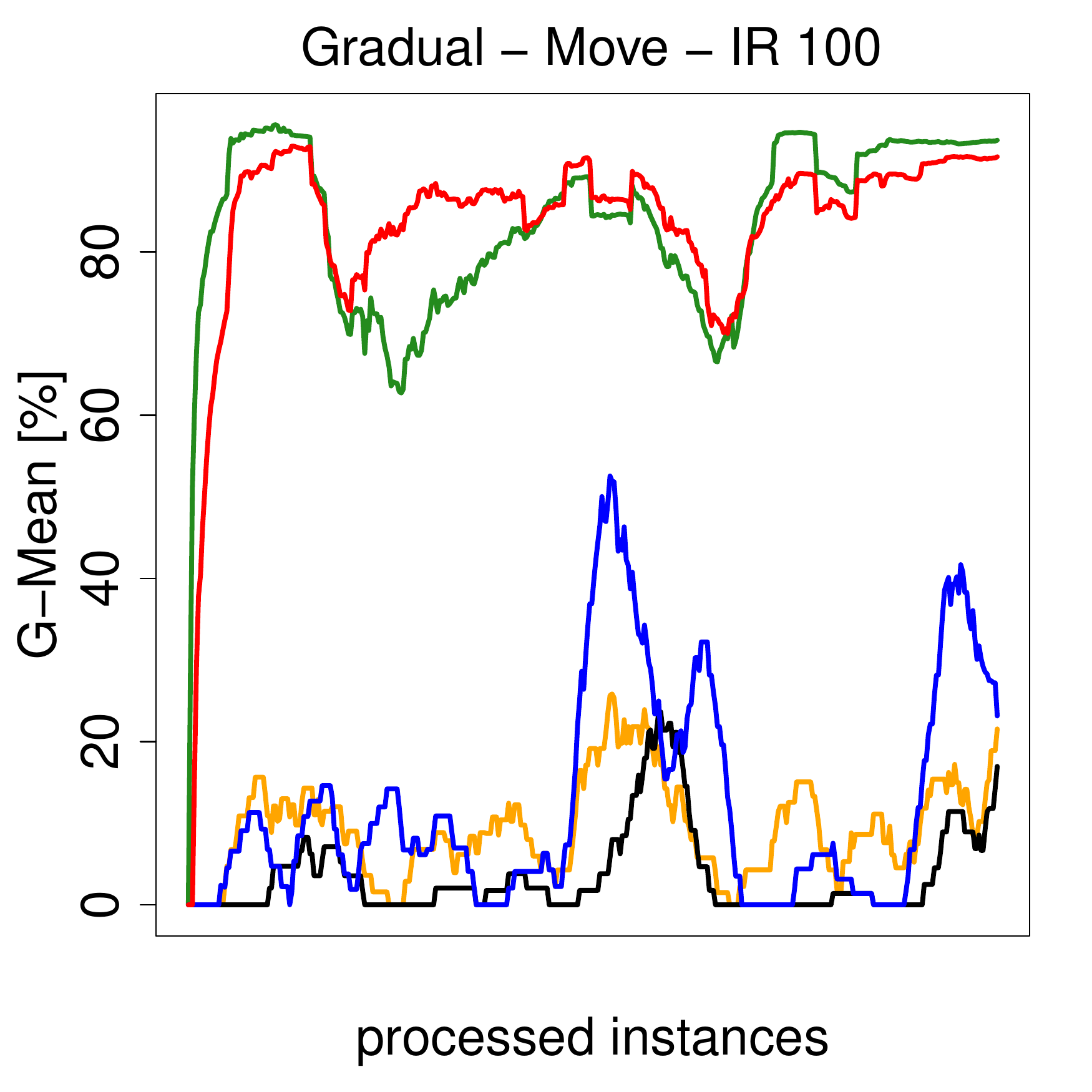}
\includegraphics[width=0.19\columnwidth]{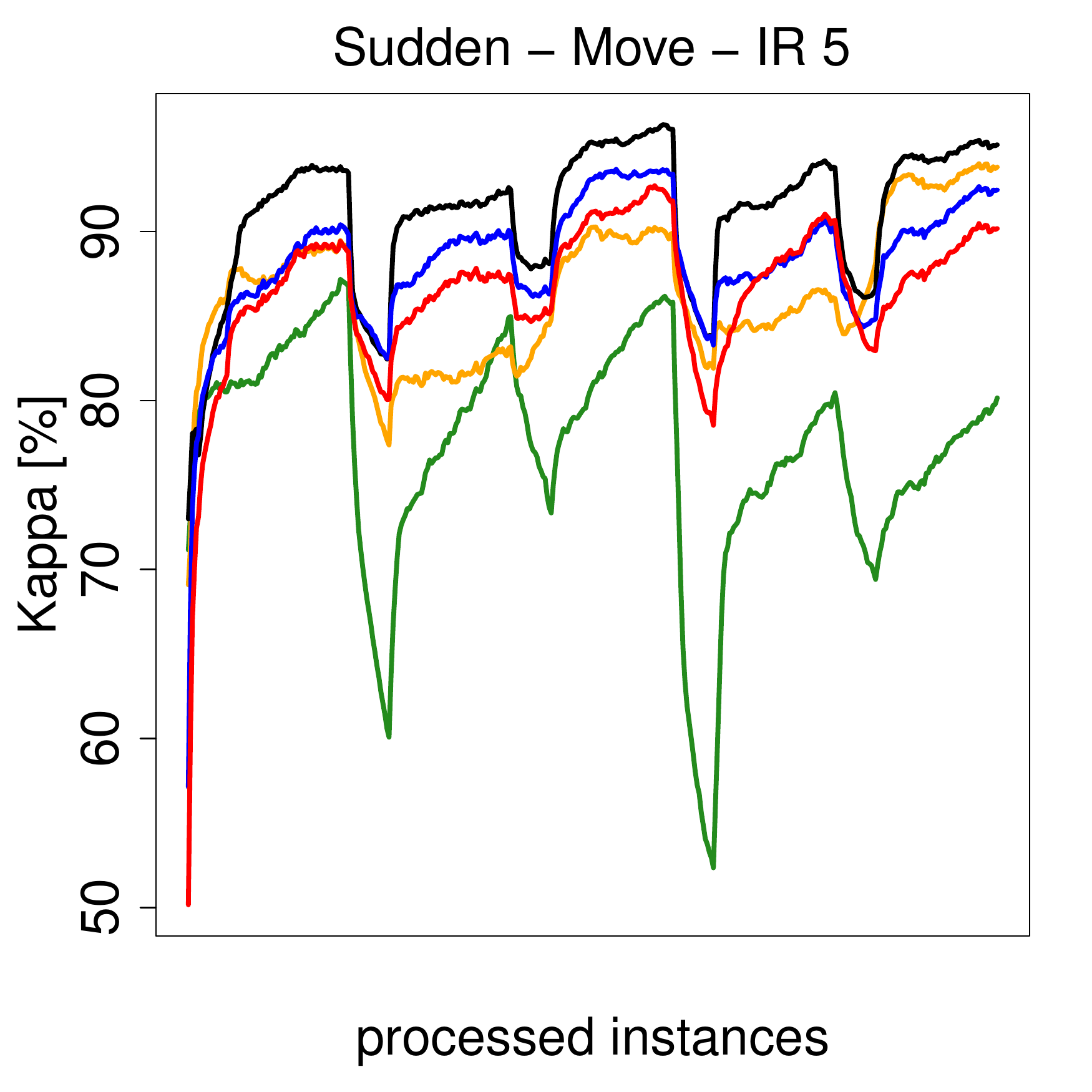}
\includegraphics[width=0.19\columnwidth]{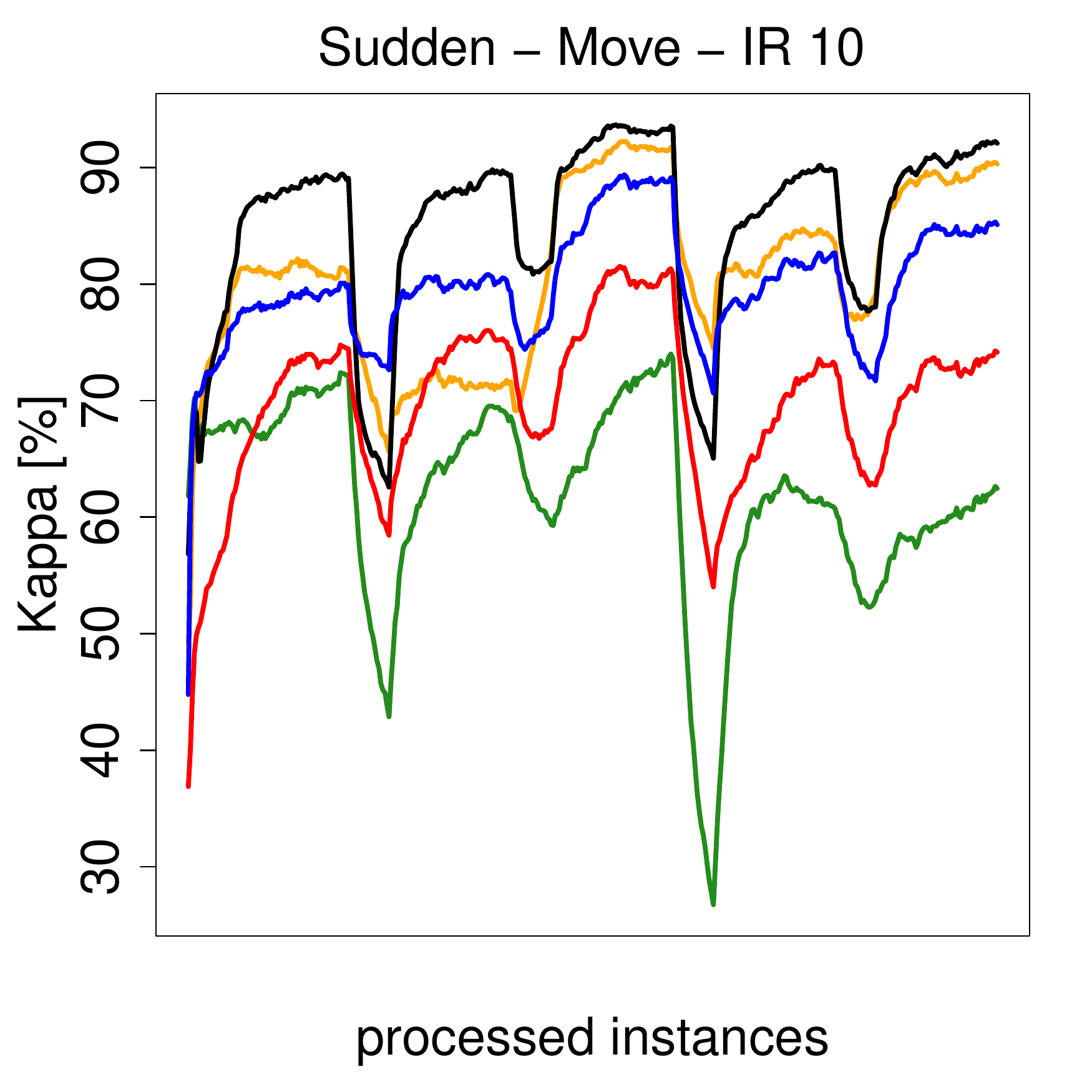}
\includegraphics[width=0.19\columnwidth]{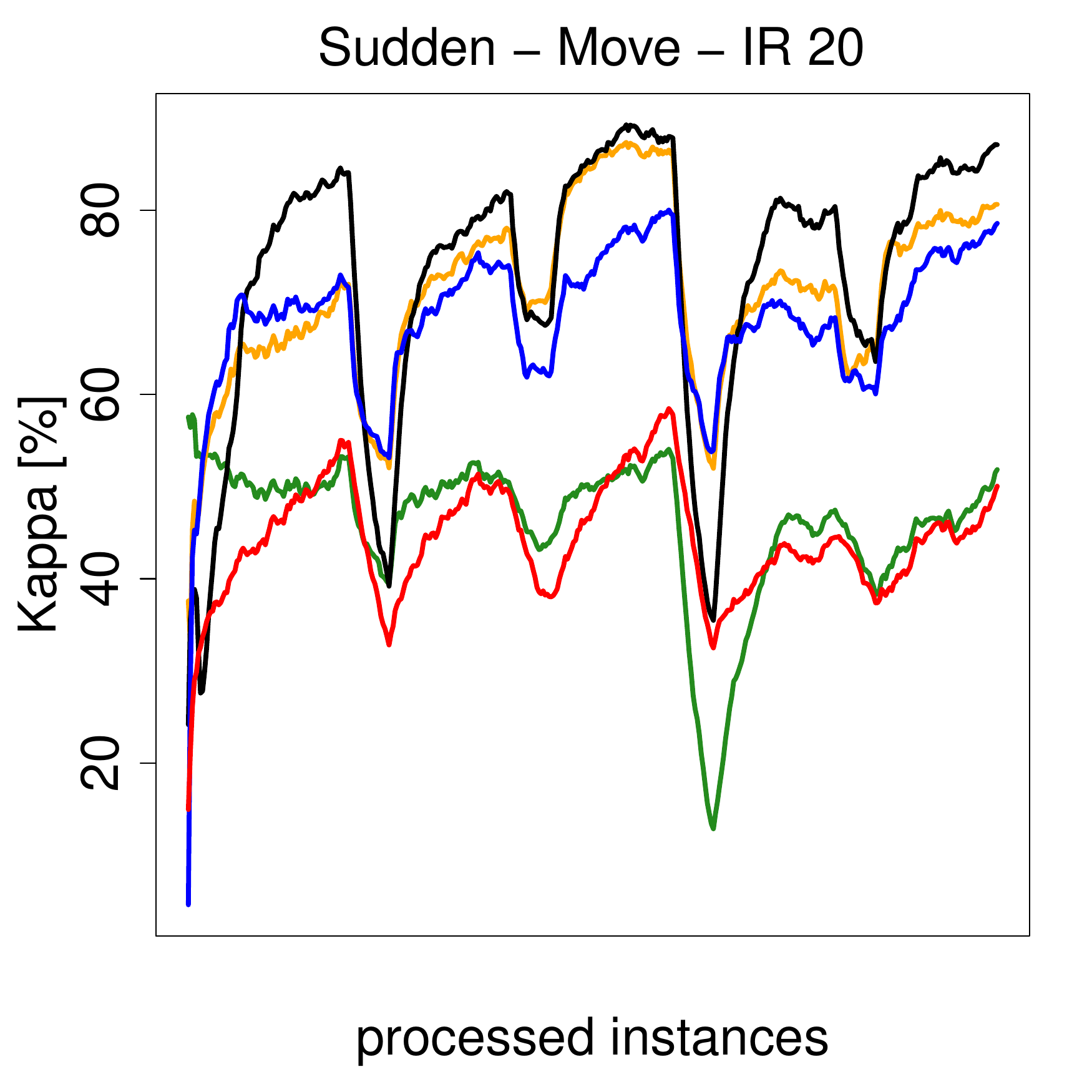}
\includegraphics[width=0.19\columnwidth]{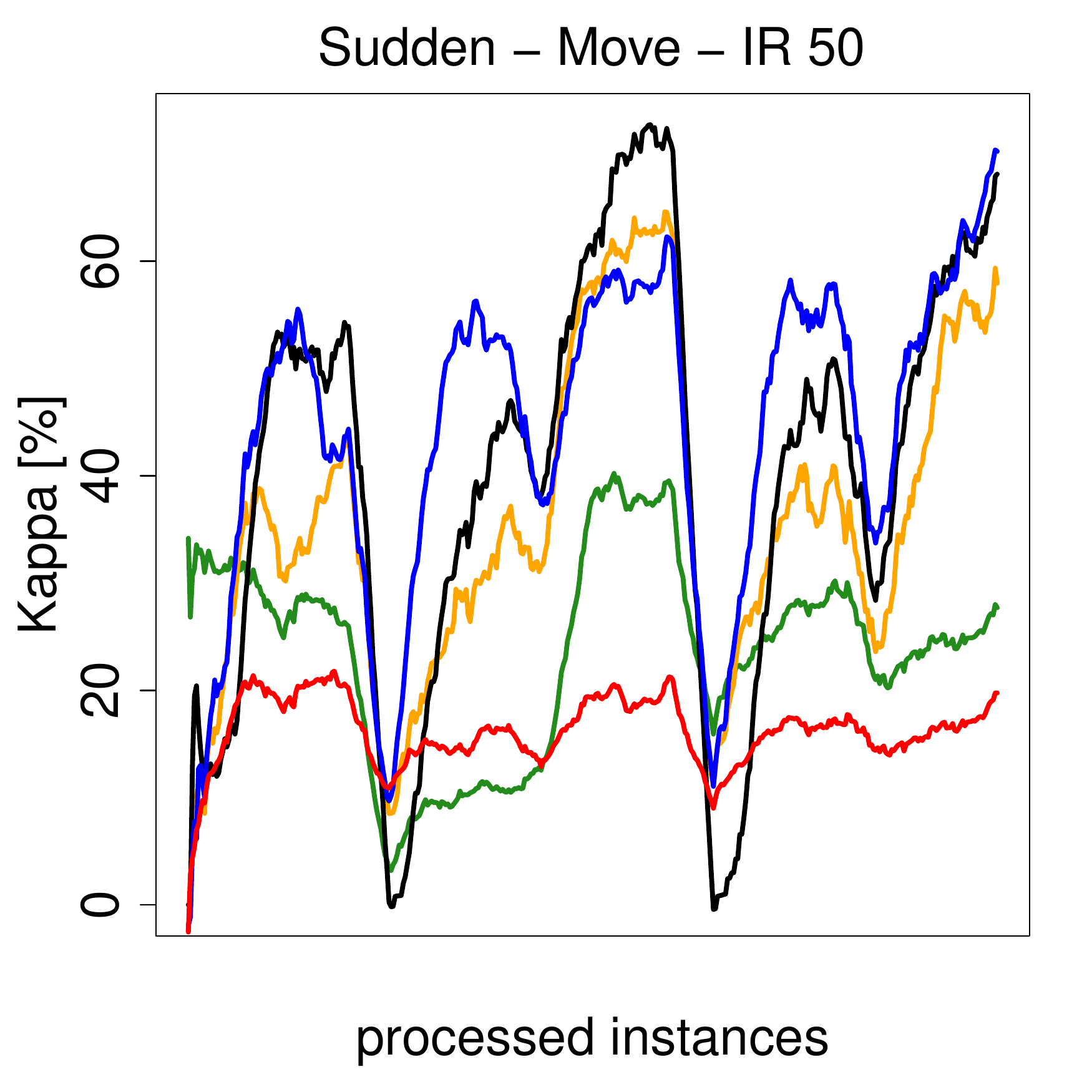}
\includegraphics[width=0.19\columnwidth]{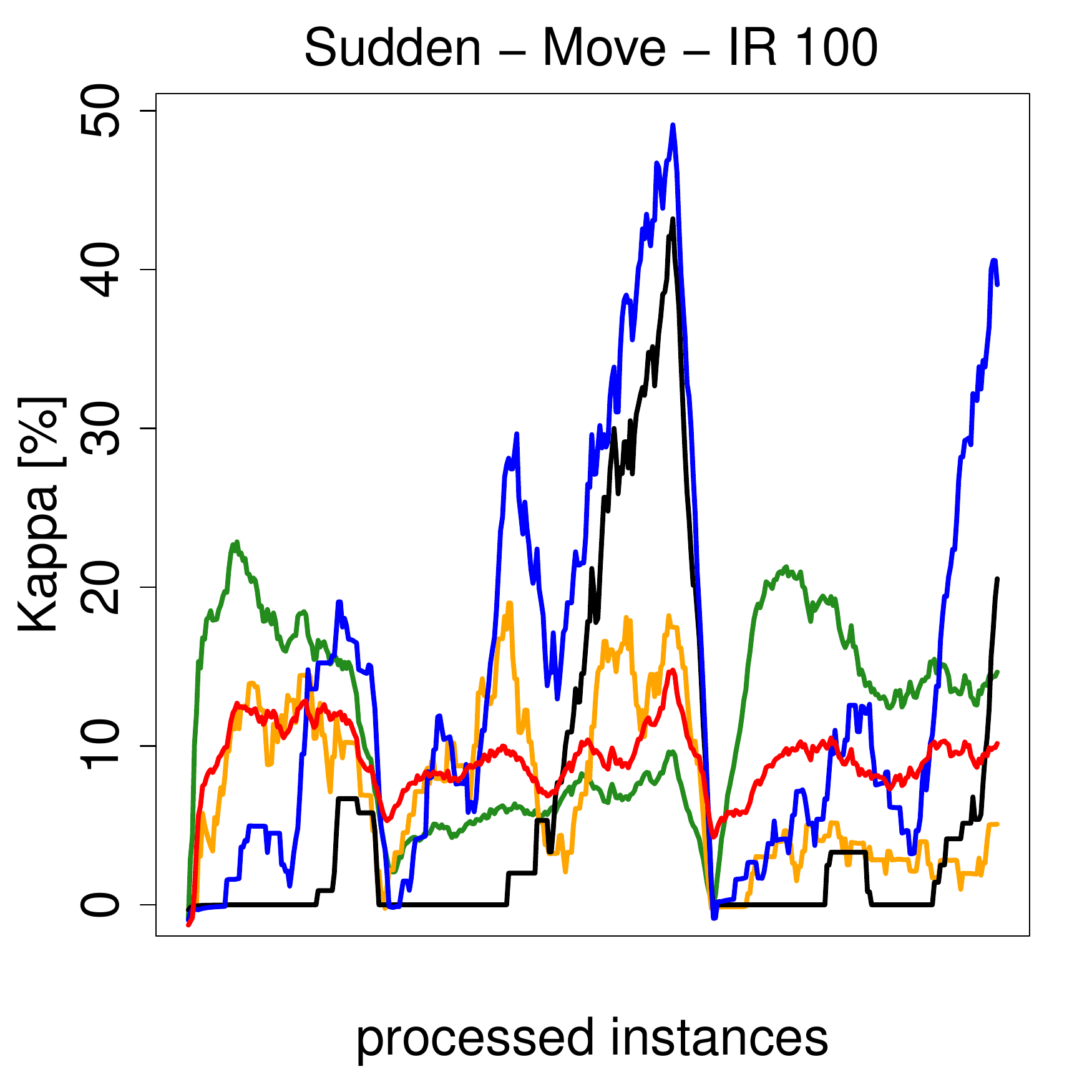}
\caption{G-Mean and Kappa on moving minority clusters for static imbalance ratio.}
\label{fig:ild_static_move}
\vspace*{0.7cm}
\end{figure}

\begin{figure}[t!]
\centering
\includegraphics[width=0.19\columnwidth]{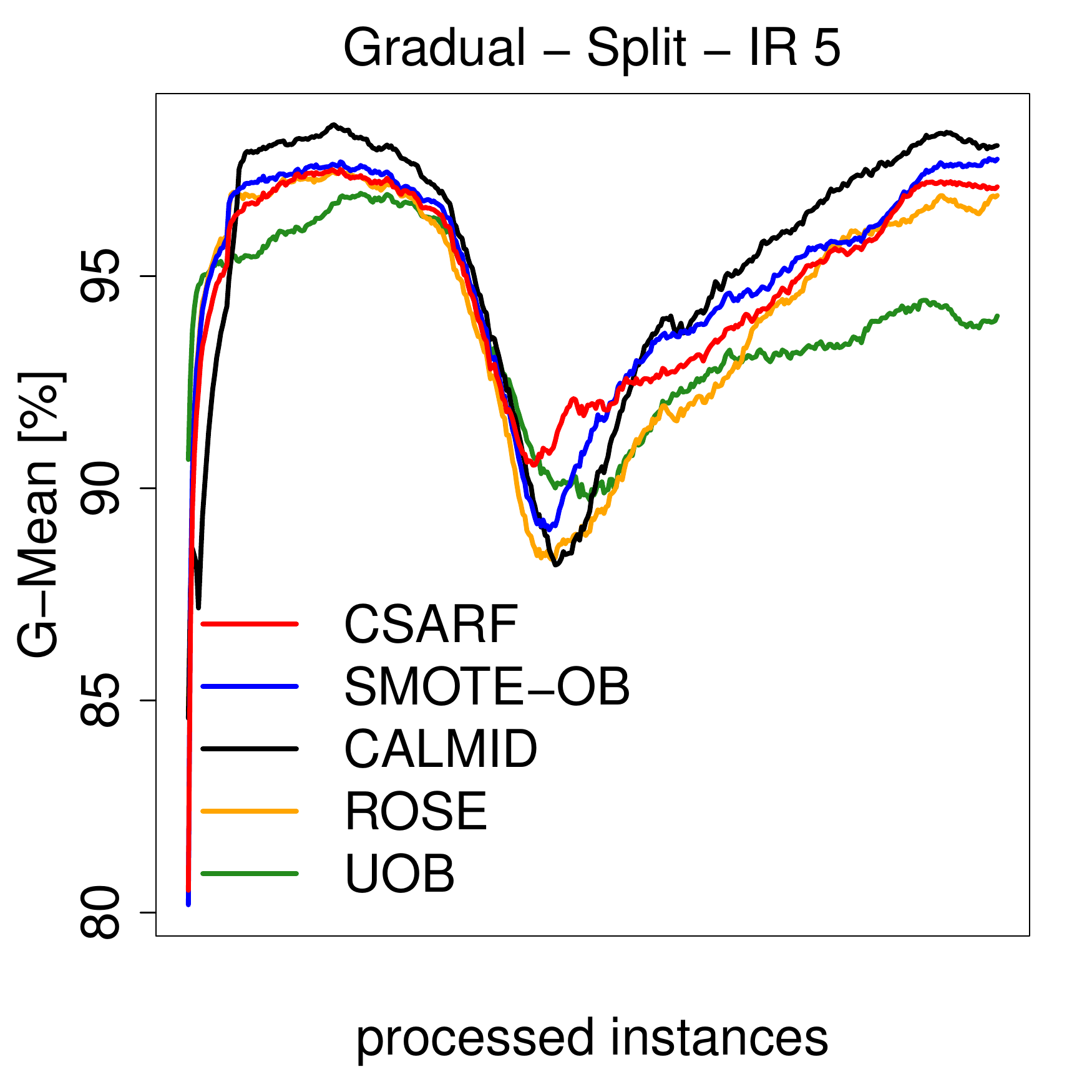}
\includegraphics[width=0.19\columnwidth]{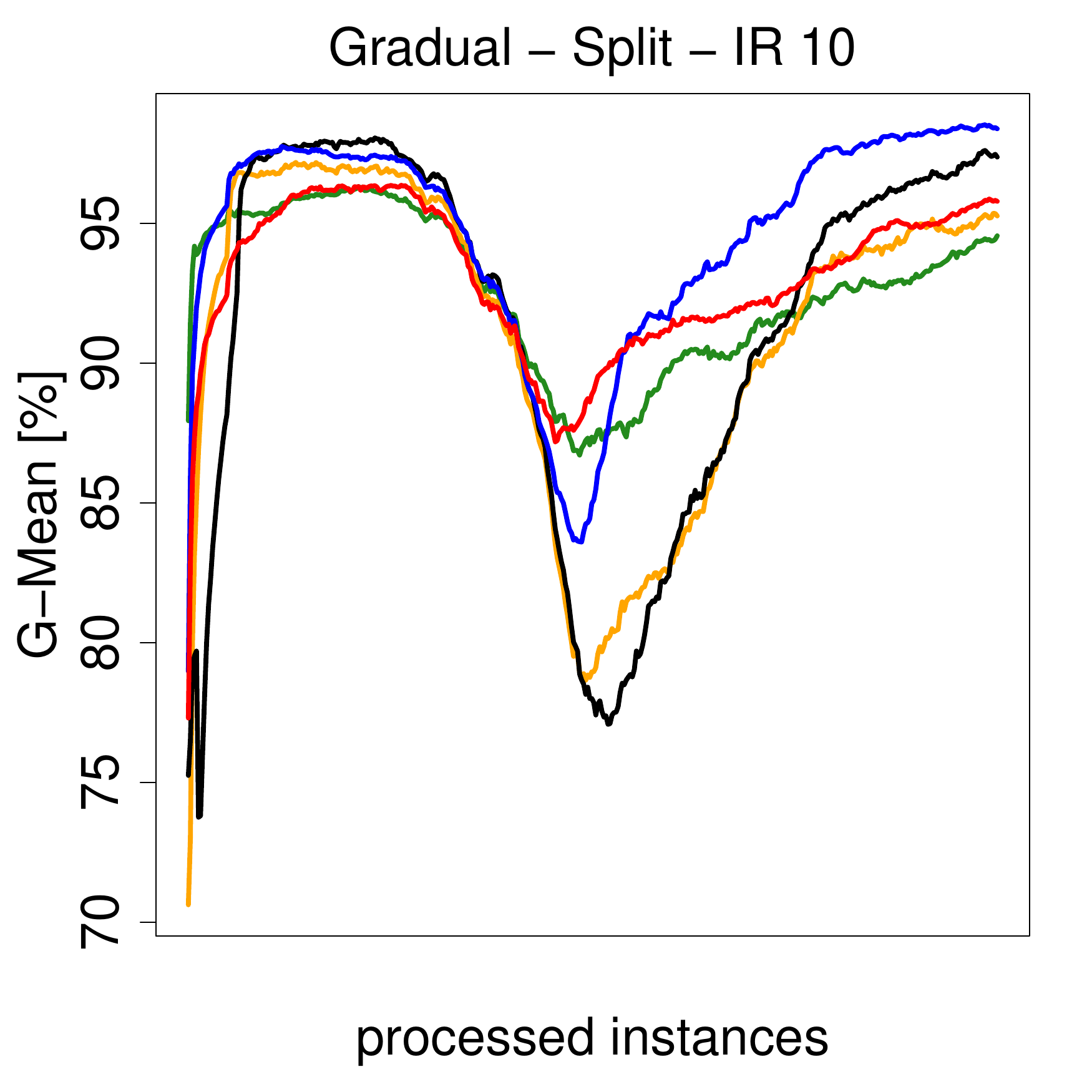}
\includegraphics[width=0.19\columnwidth]{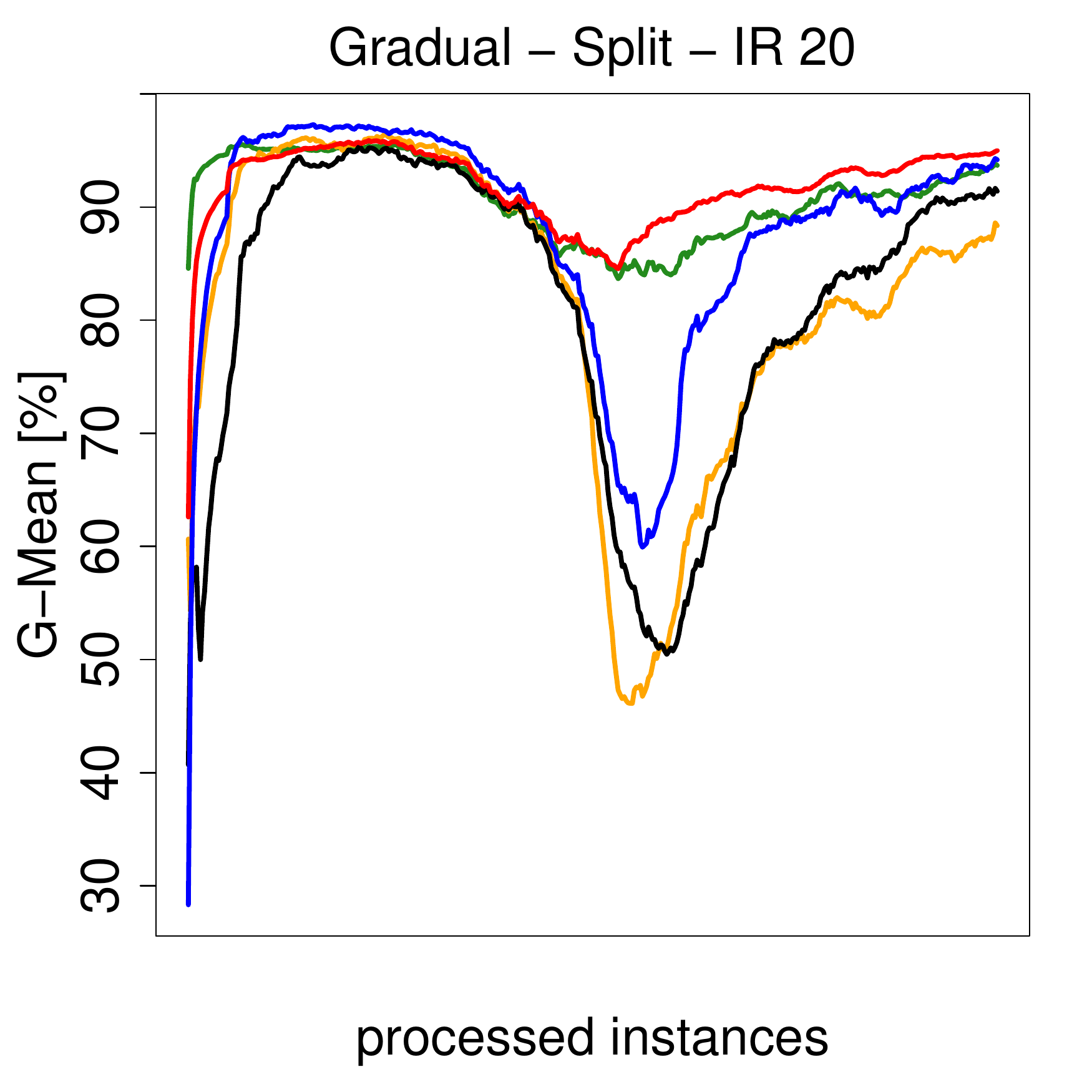}
\includegraphics[width=0.19\columnwidth]{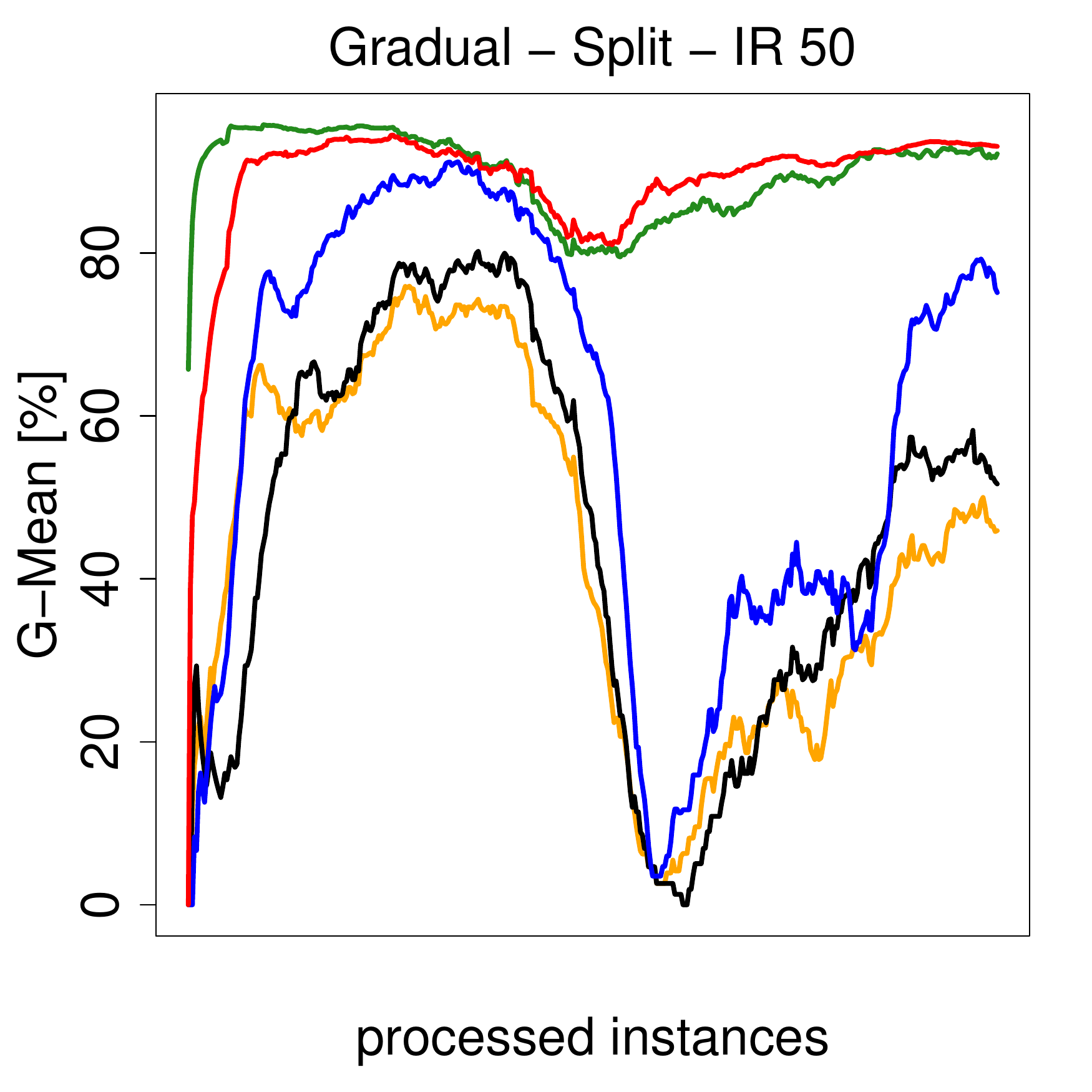}
\includegraphics[width=0.19\columnwidth]{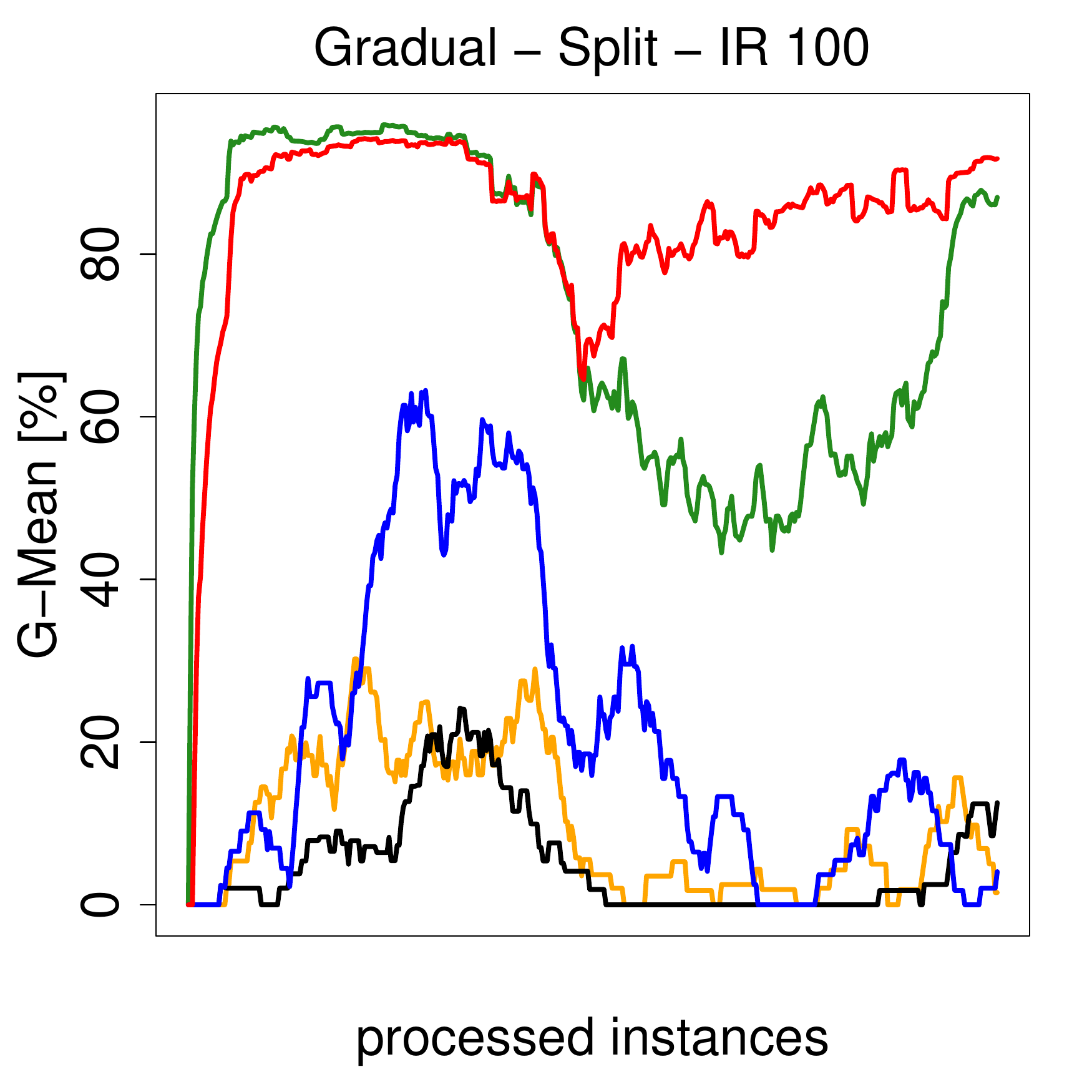}
\includegraphics[width=0.19\columnwidth]{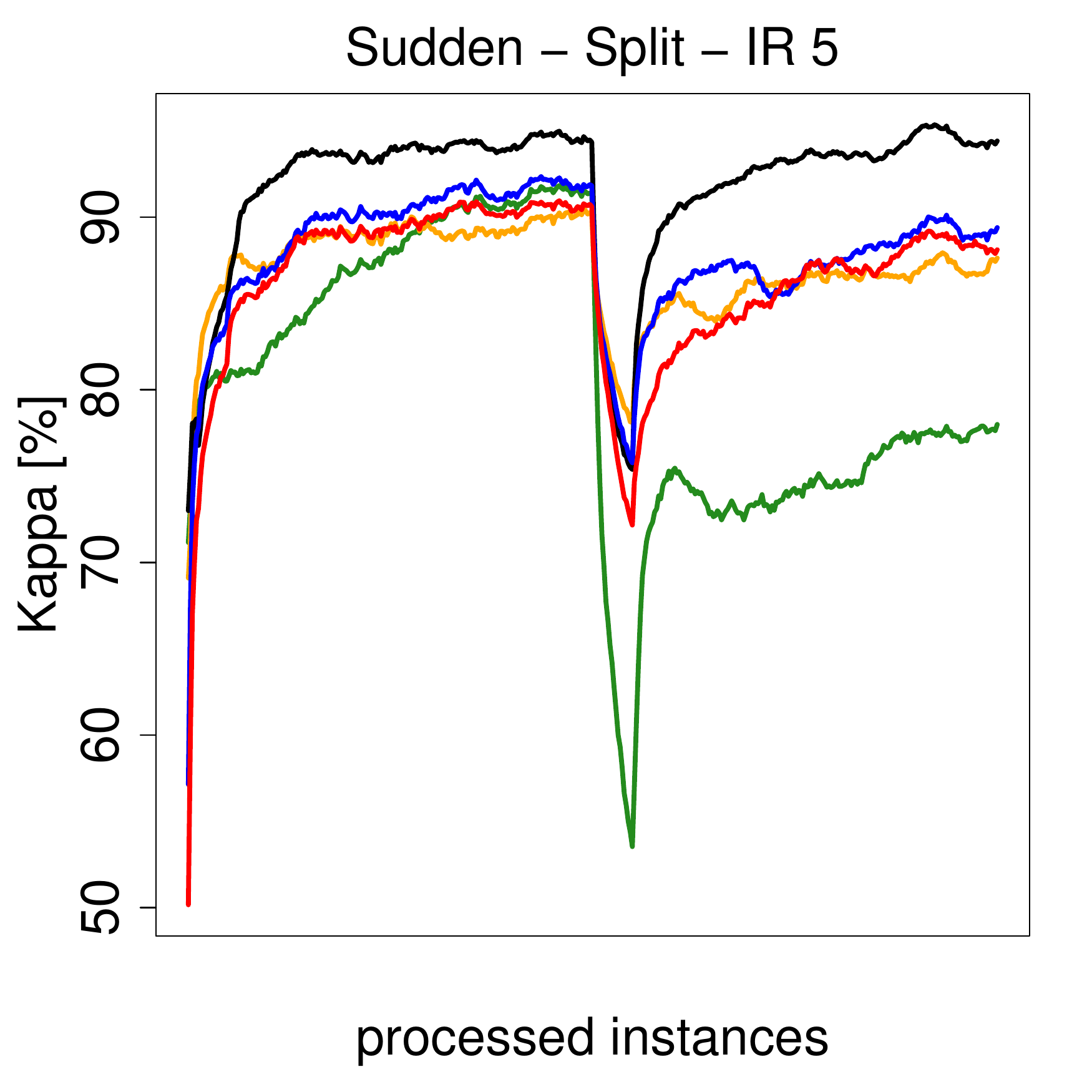}
\includegraphics[width=0.19\columnwidth]{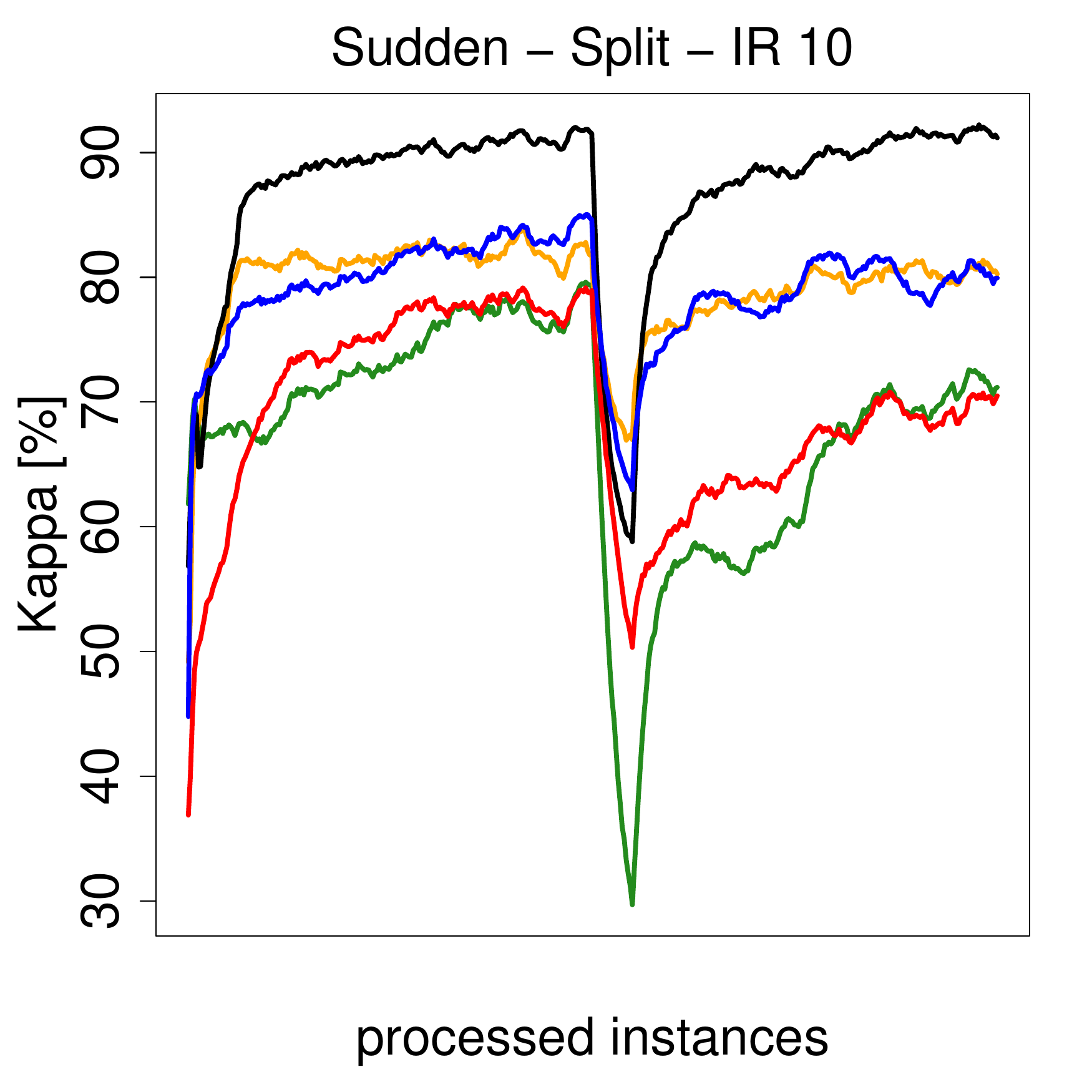}
\includegraphics[width=0.19\columnwidth]{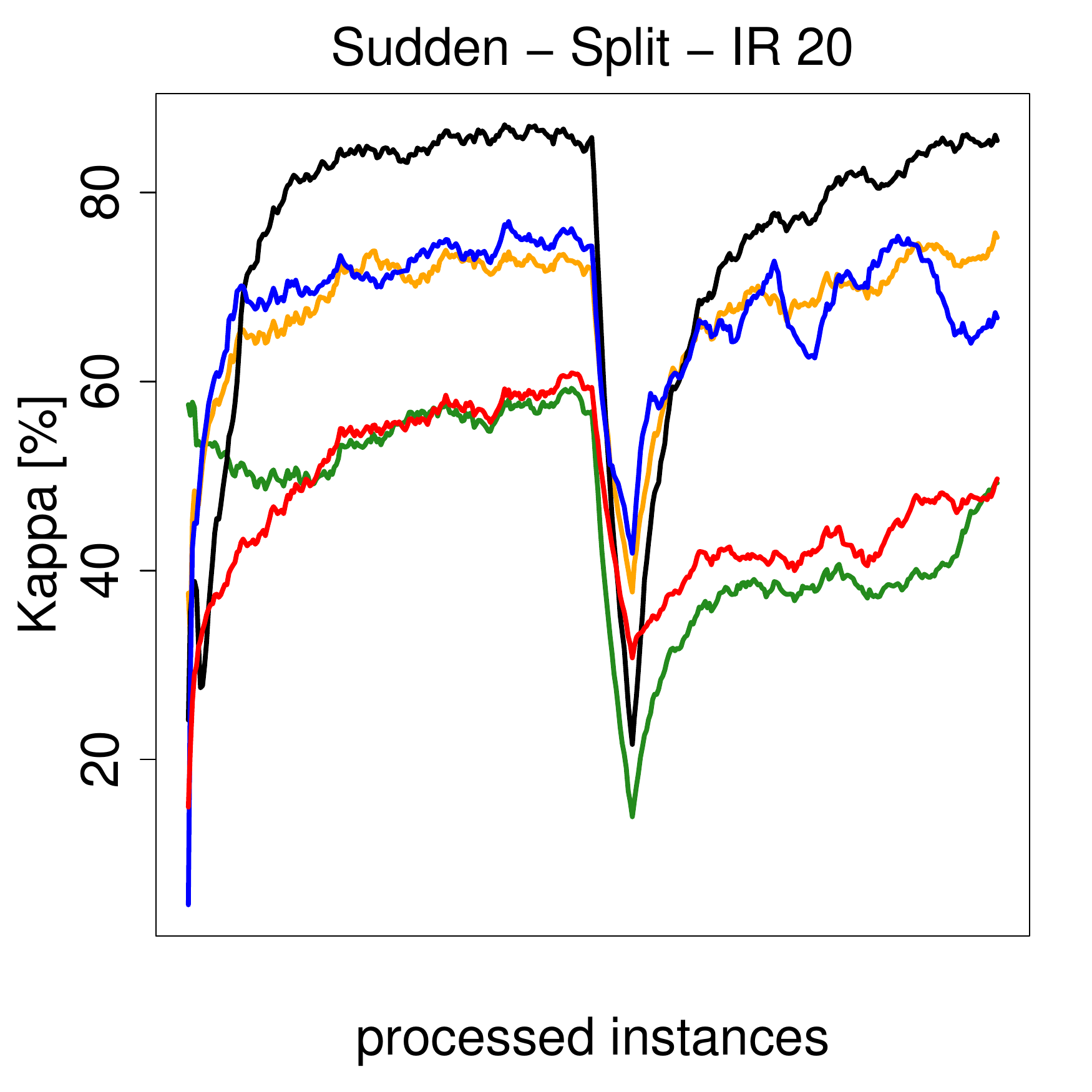}
\includegraphics[width=0.19\columnwidth]{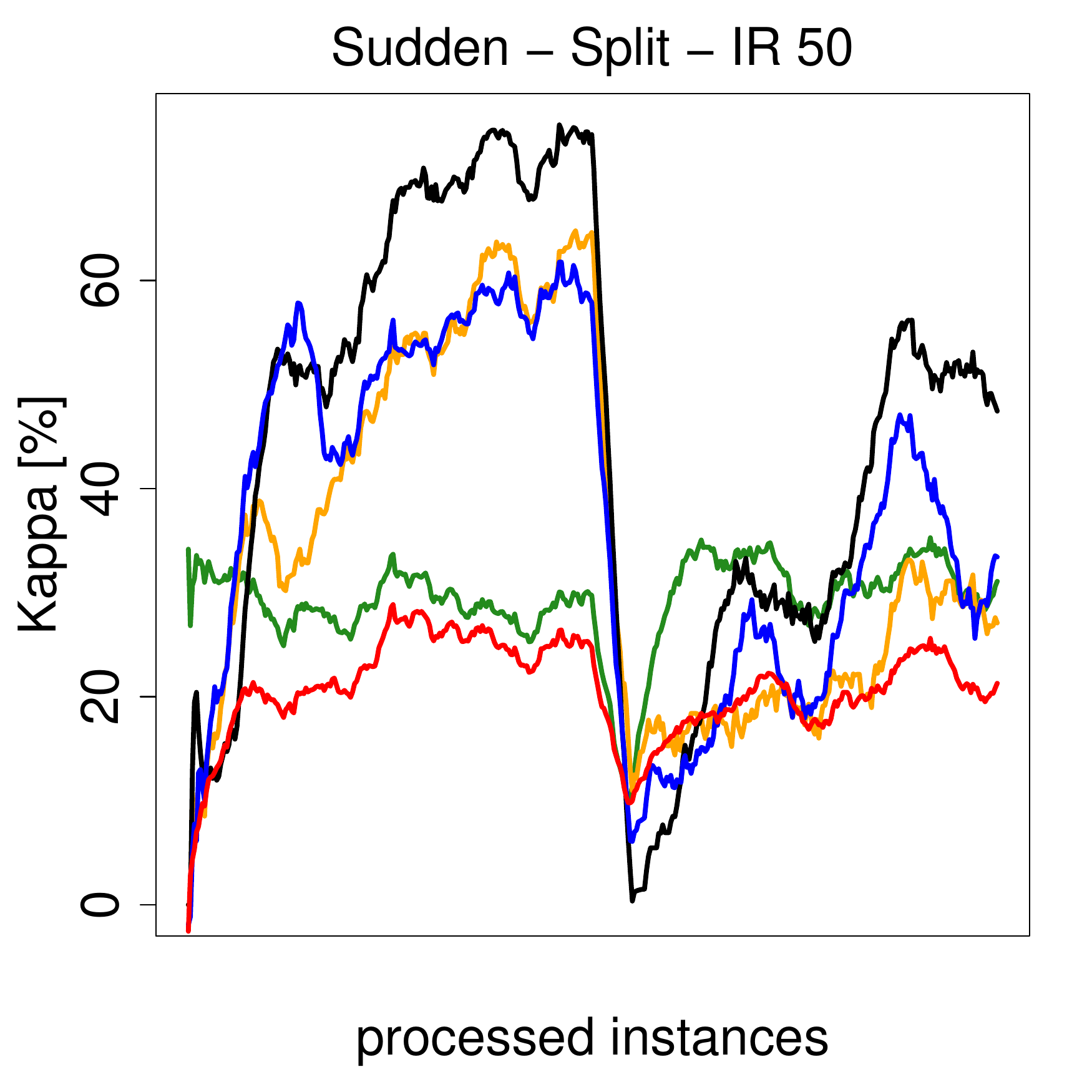}
\includegraphics[width=0.19\columnwidth]{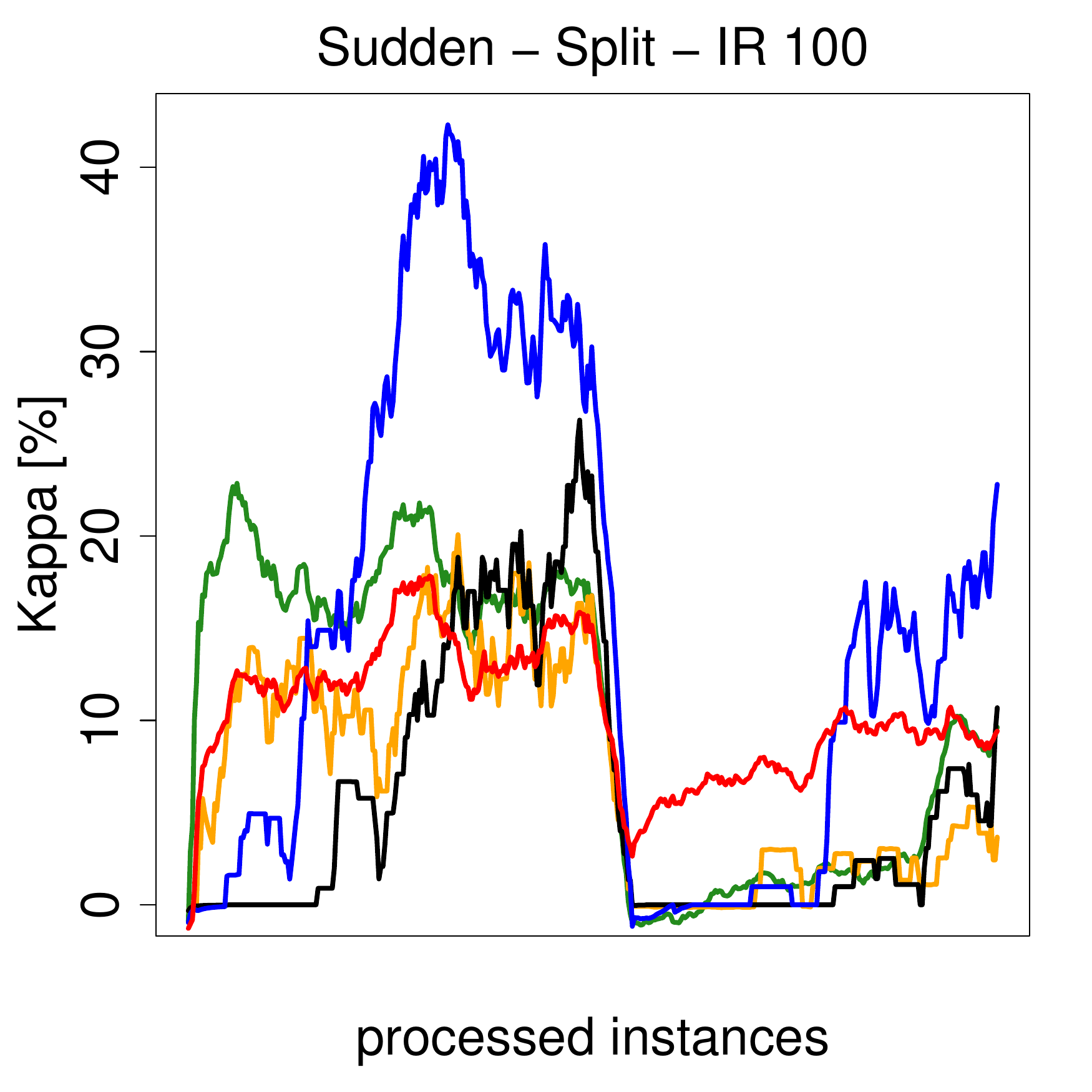}
\caption{G-Mean and Kappa on splitting minority clusters for static imbalance ratio.}
\label{fig:ild_static_split}
\vspace*{0.7cm}
\end{figure}

\begin{figure}[t!]
\centering
\includegraphics[width=0.19\columnwidth]{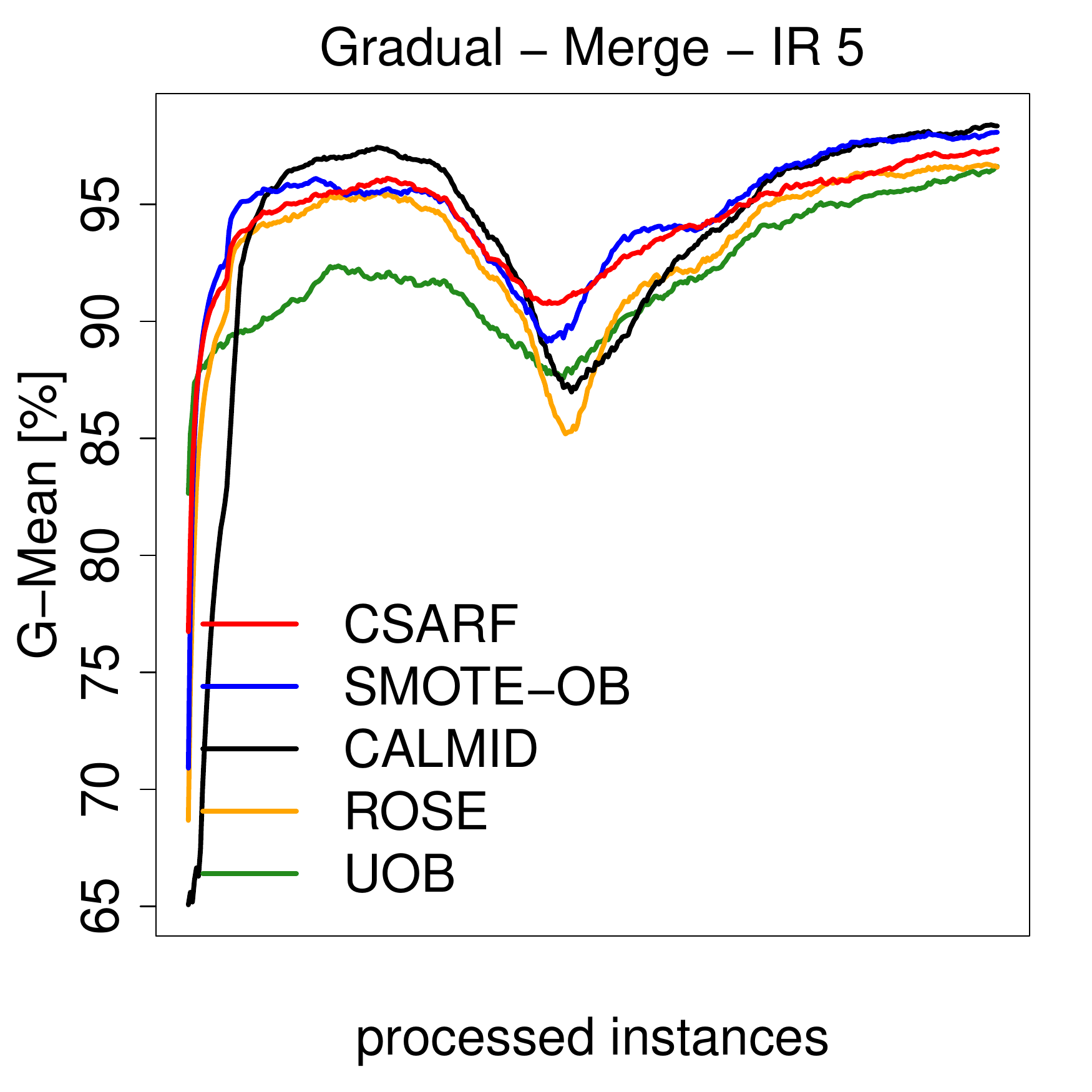}
\includegraphics[width=0.19\columnwidth]{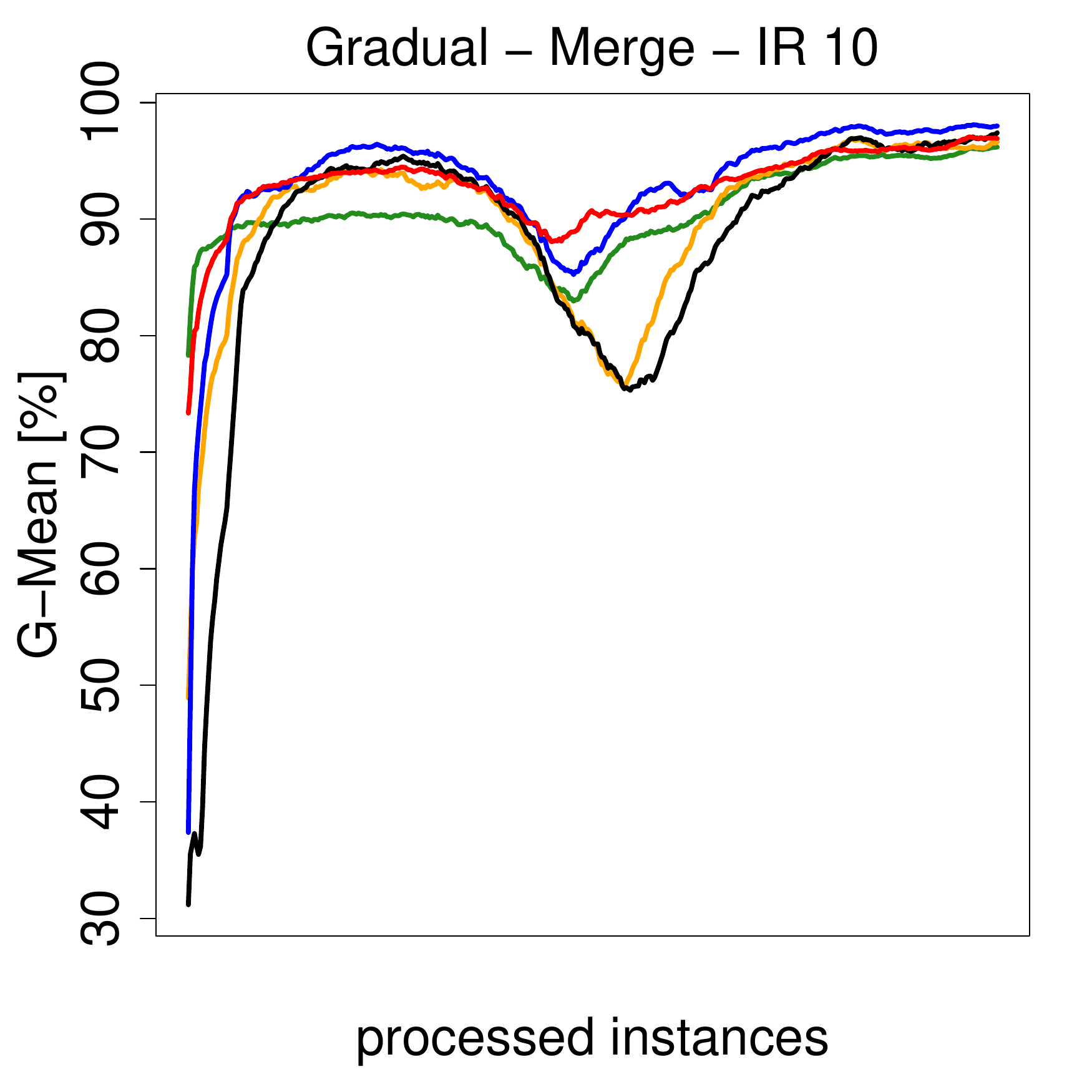}
\includegraphics[width=0.19\columnwidth]{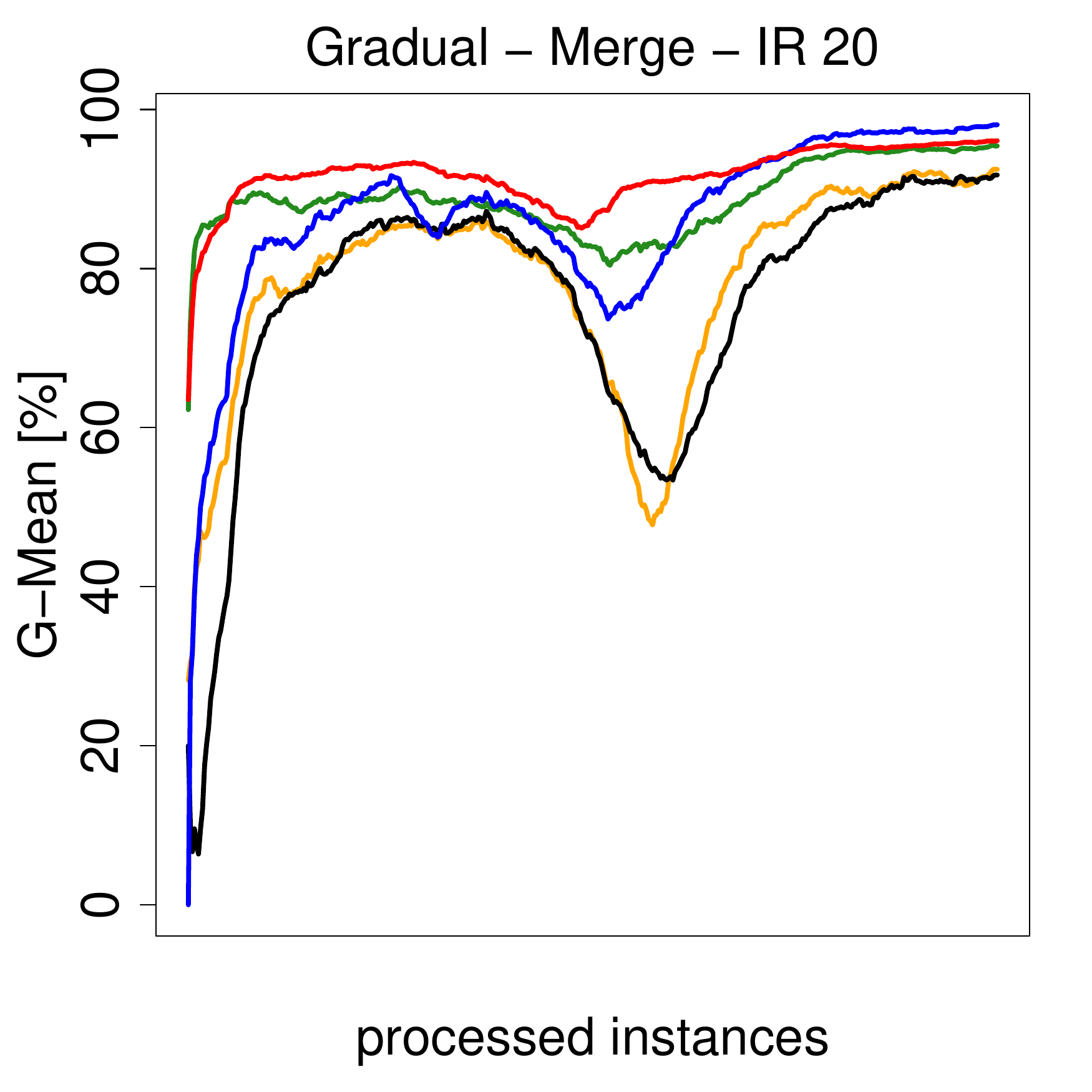}
\includegraphics[width=0.19\columnwidth]{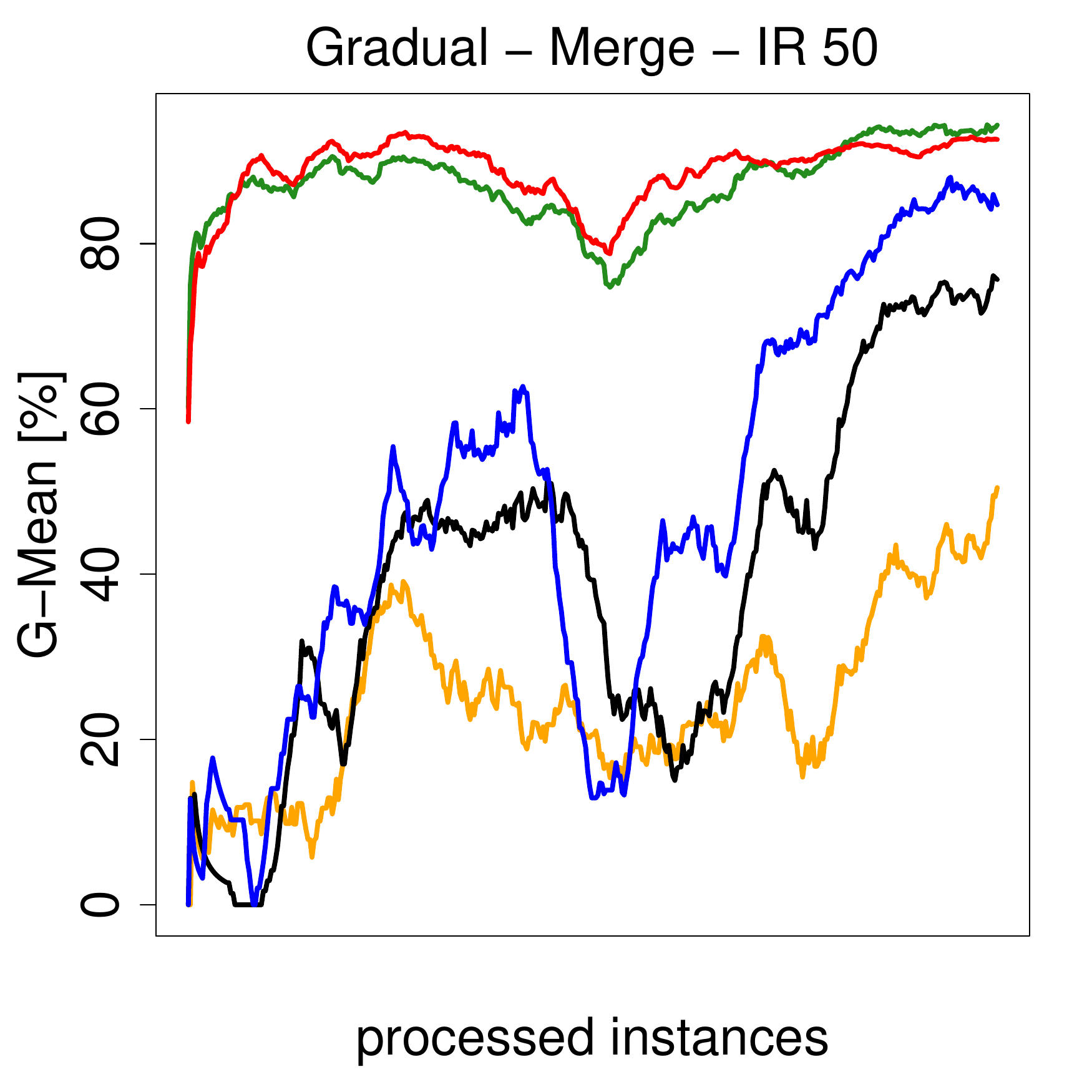}
\includegraphics[width=0.19\columnwidth]{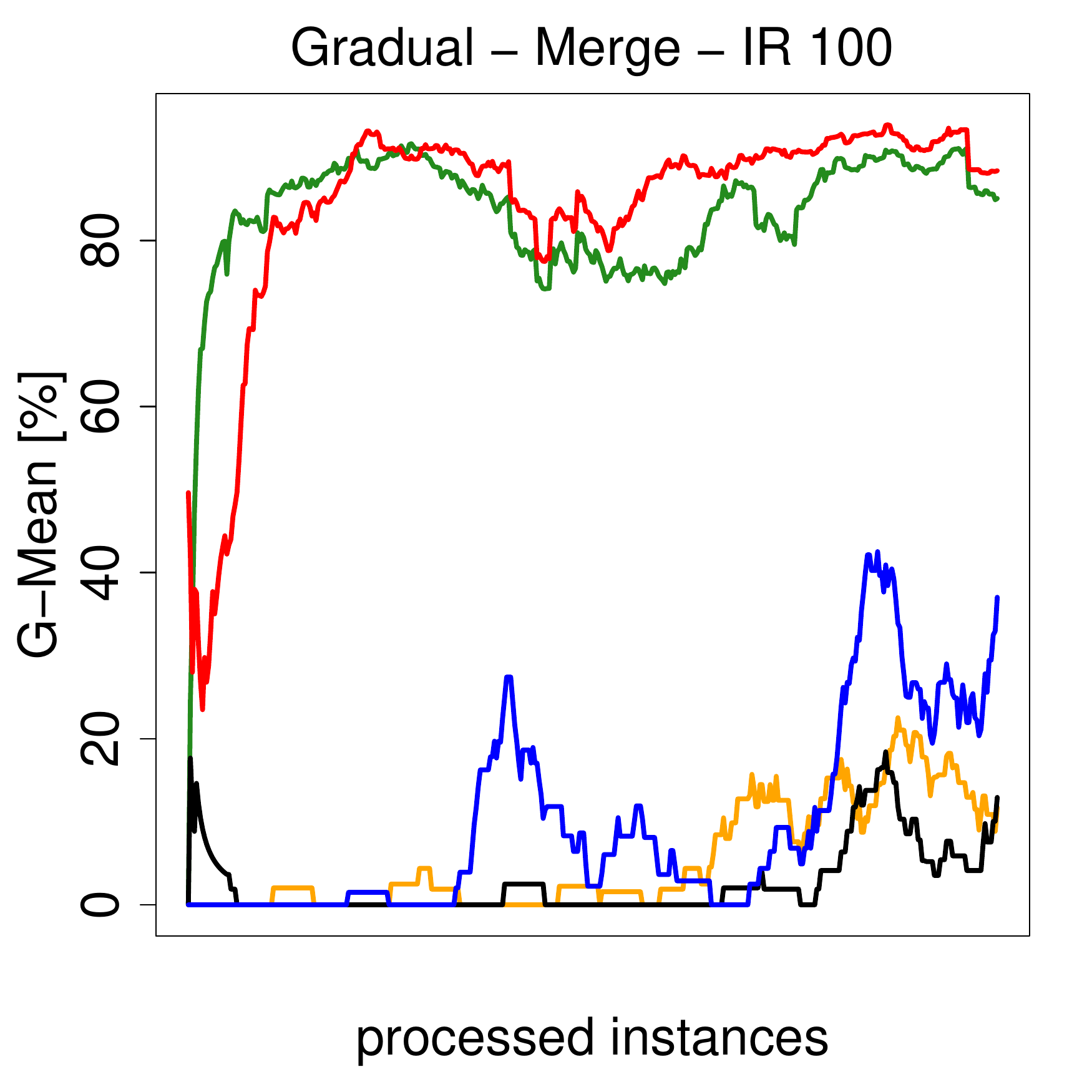}
\includegraphics[width=0.19\columnwidth]{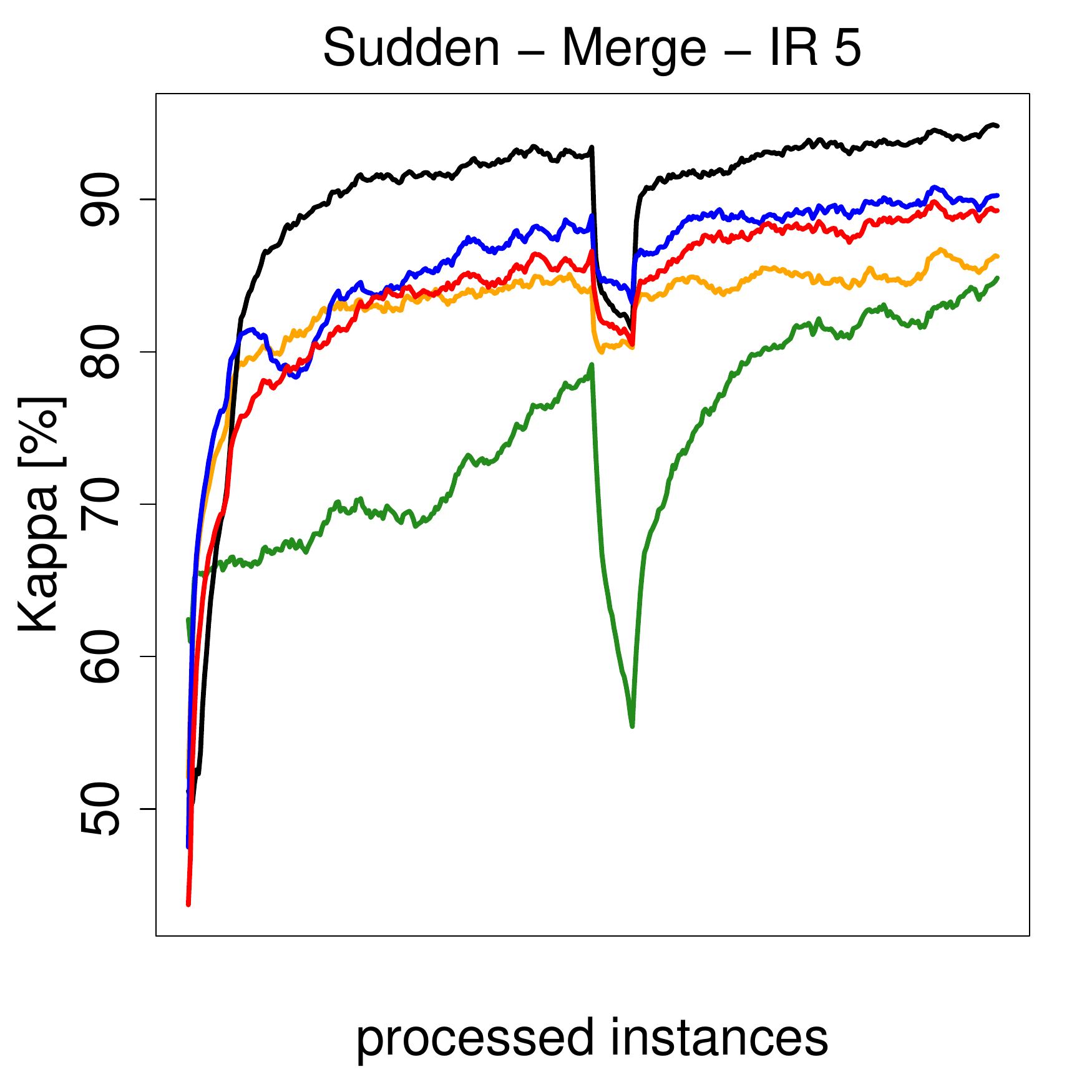}
\includegraphics[width=0.19\columnwidth]{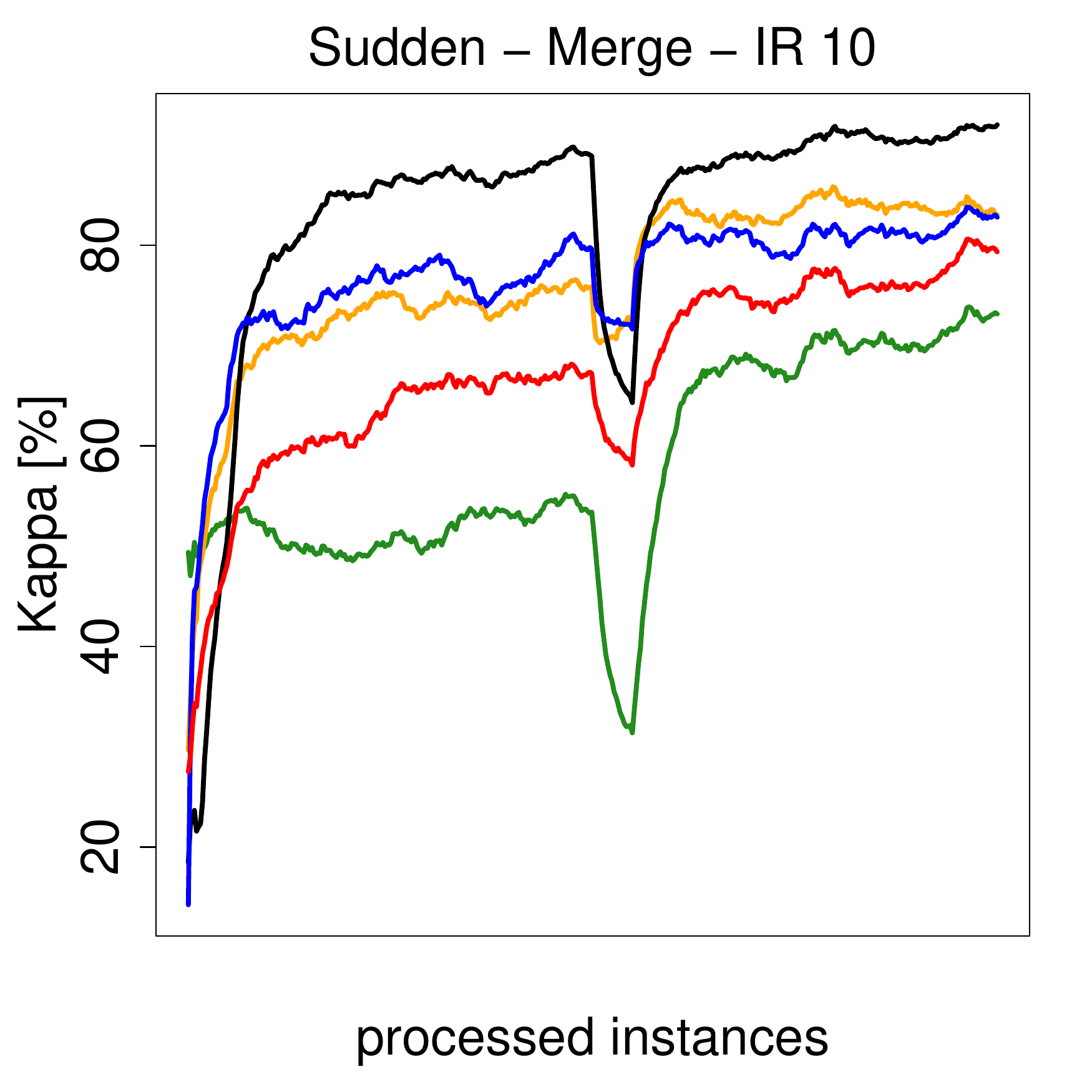}
\includegraphics[width=0.19\columnwidth]{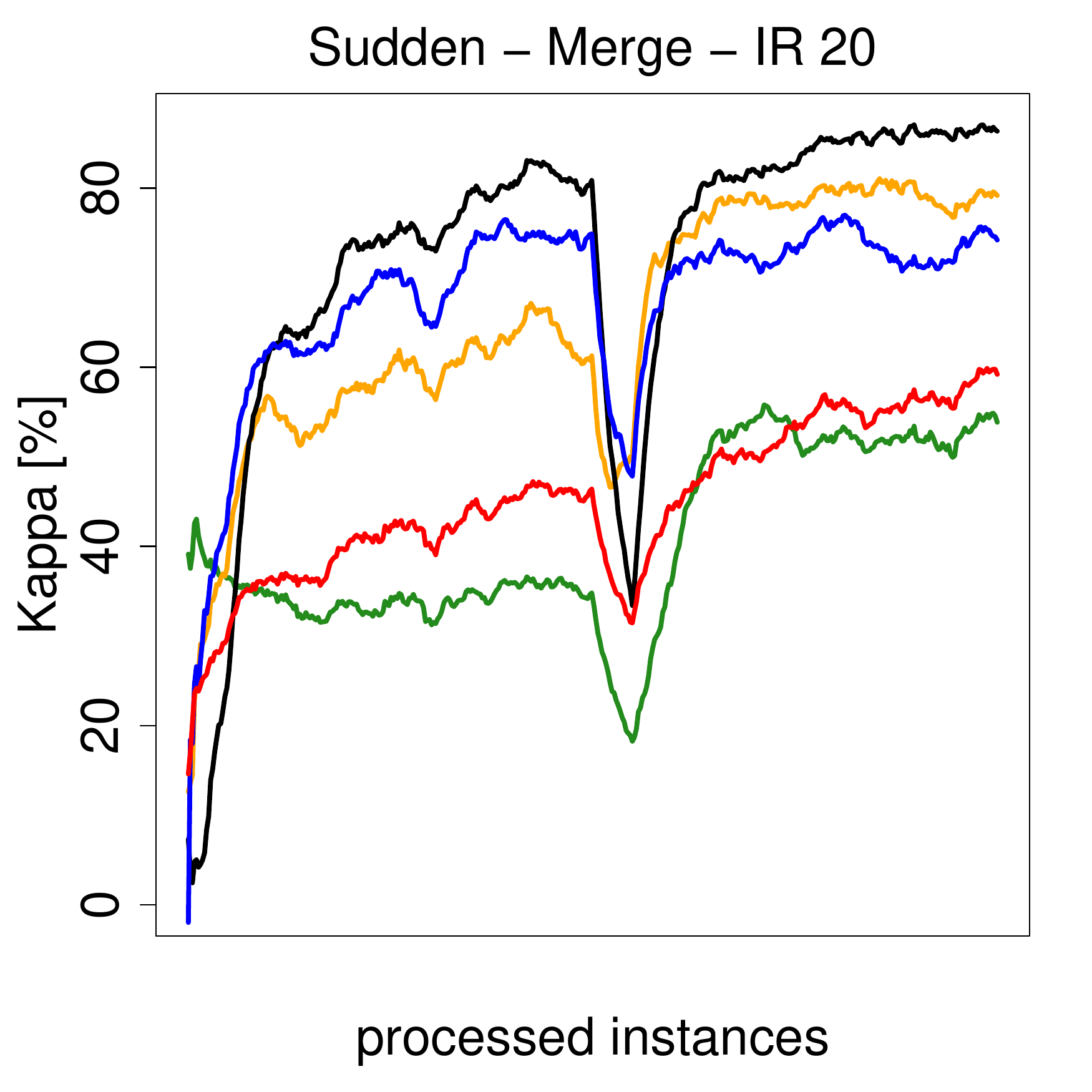}
\includegraphics[width=0.19\columnwidth]{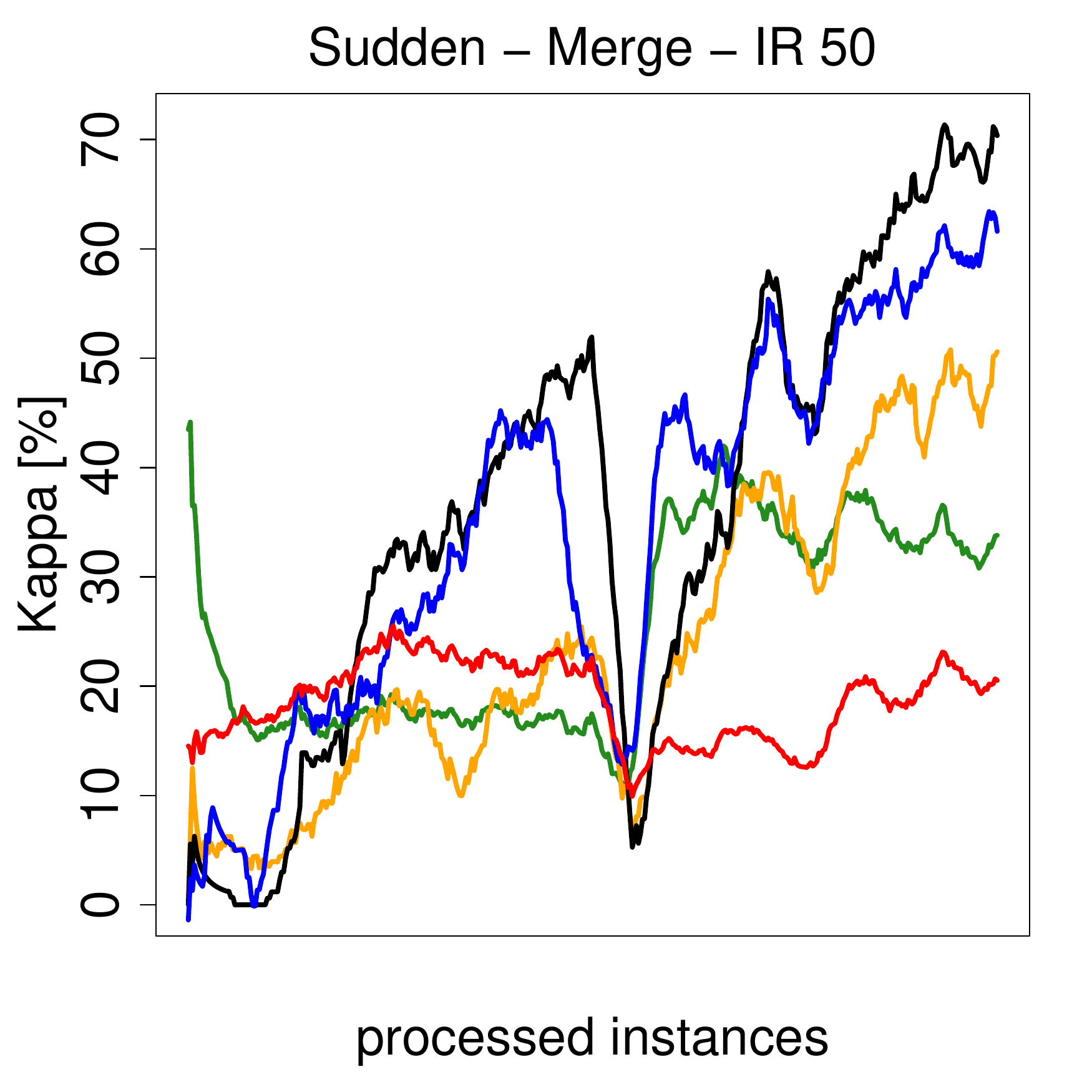}
\includegraphics[width=0.19\columnwidth]{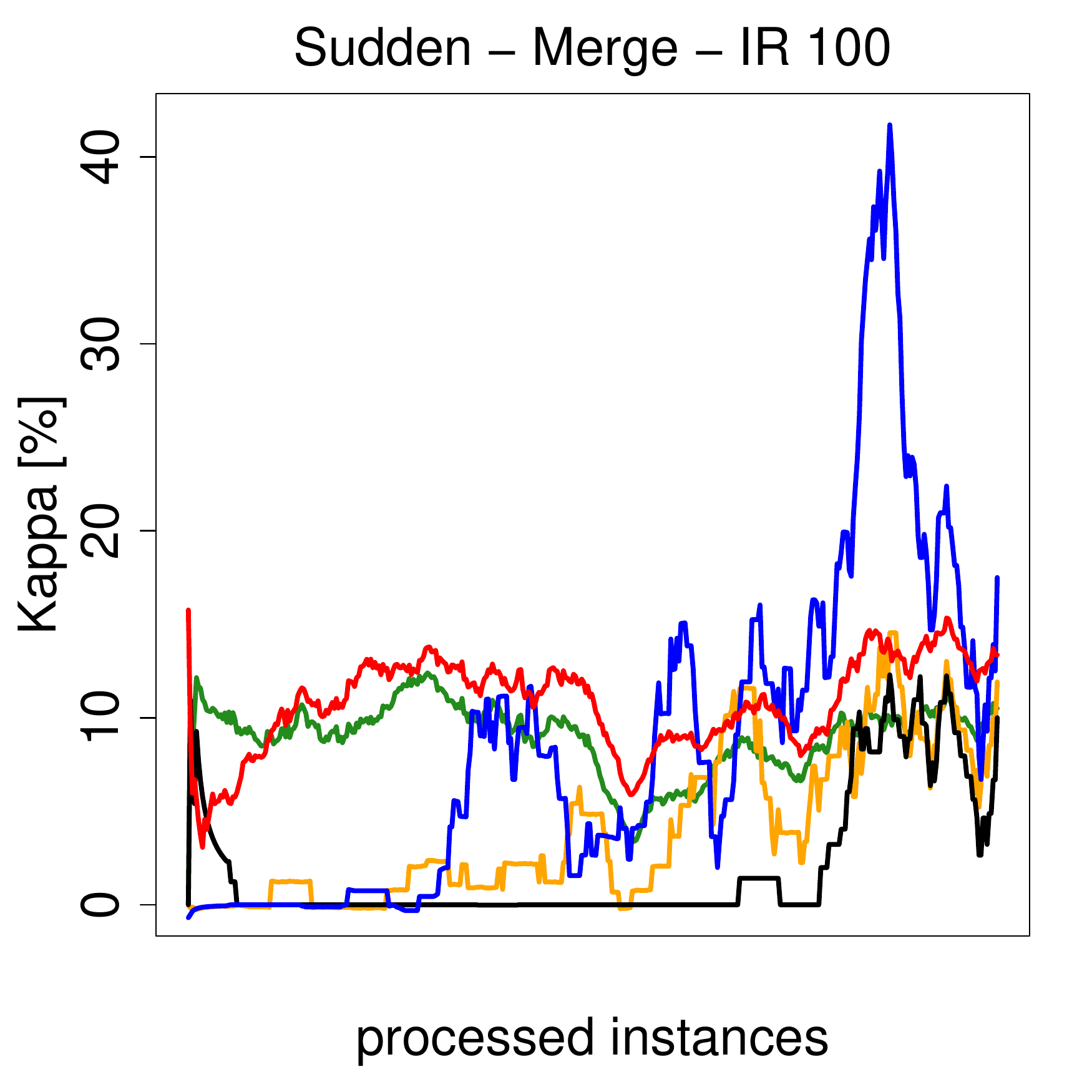}
\caption{G-Mean and Kappa on merging minority clusters for static imbalance ratio.}
\label{fig:ild_static_merge}
\vspace*{0.6cm}
\end{figure}

\begin{figure}[t!]
\centering
\includegraphics[width=\columnwidth]{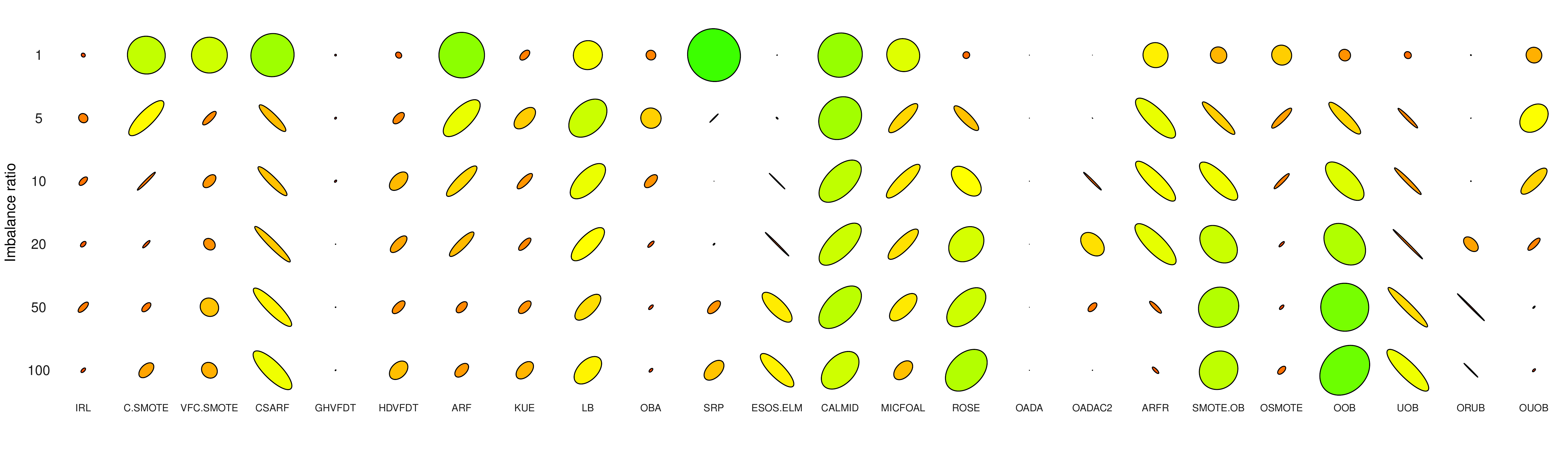}
\caption{Comparison of all 24 algorithms for borderline instances on static class imbalance ratio. Axes of the ellipse represent G-Mean and Kappa metrics. Color gradient represents the product of both metrics.}
\label{fig:BC_ILD_SIR_borderline}
\end{figure}

\begin{figure}[t!]
\centering
\includegraphics[width=\columnwidth]{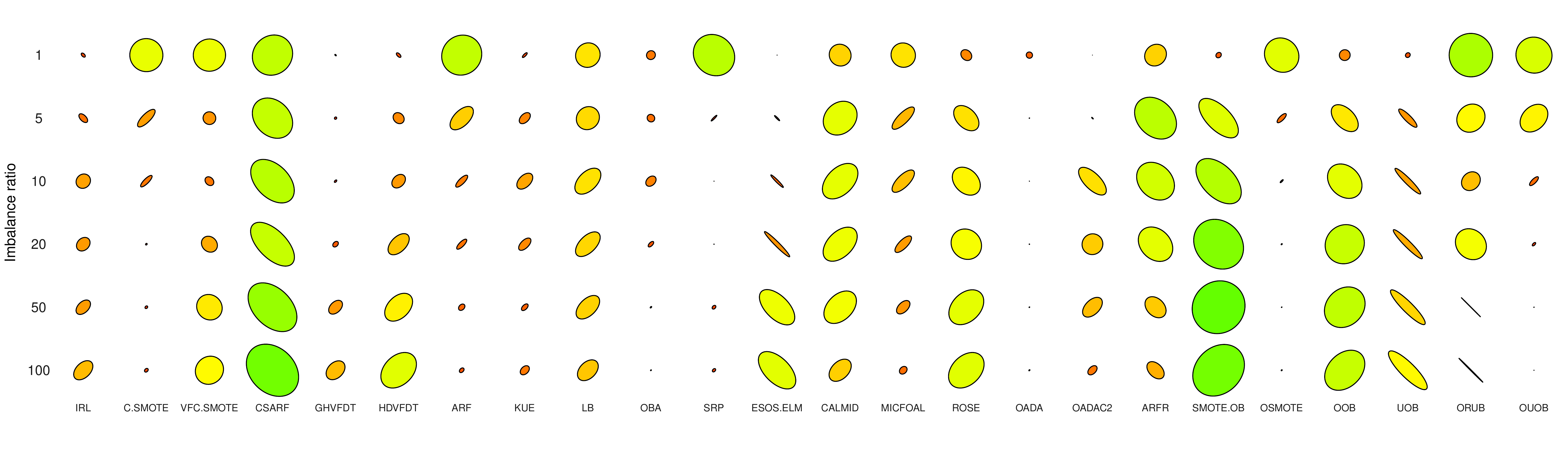}
\caption{Comparison of all 24 algorithms for rare instances on static class imbalance ratio. Axes of the ellipse represent G-Mean and Kappa metrics. Color gradient represents the product of both metrics.}
\label{fig:BC_ILD_SIR_rare}
\end{figure}

\begin{figure}[t!]
\centering
\includegraphics[width=\columnwidth]{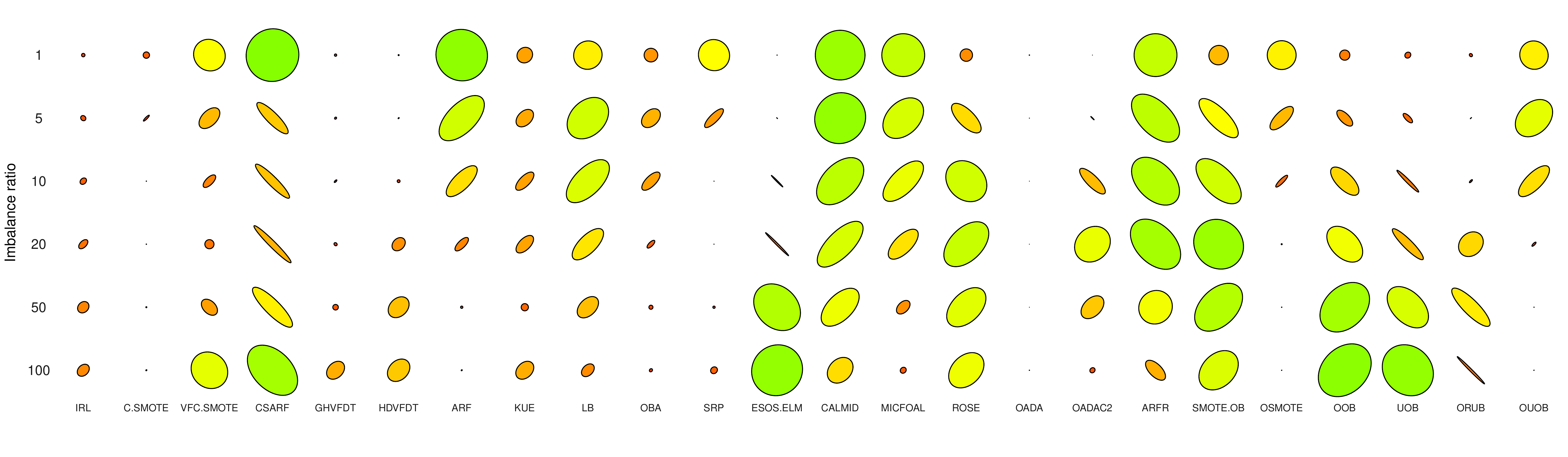}
\caption{Comparison of all 24 algorithms for moving minority clusters on static class imbalance ratio. Axes of the ellipse represent G-Mean and Kappa metrics. Color gradient represents the product of both metrics.}
\label{fig:BC_ILD_SIR_moving}
\end{figure}

\begin{figure}[t!]
\centering
\includegraphics[width=\columnwidth]{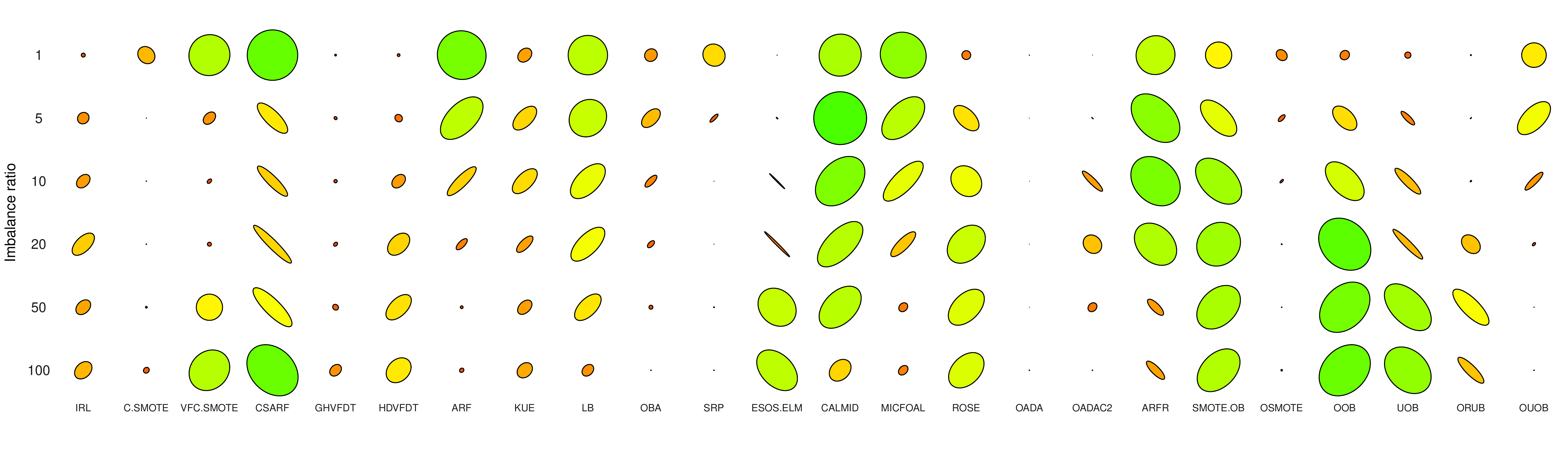}
\caption{Comparison of all 24 algorithms for splitting minority clusters on static class imbalance ratio. Axes of the ellipse represent G-Mean and Kappa metrics. Color gradient represents the product of both metrics.}
\label{fig:BC_ILD_SIR_splitting}
\end{figure}

\begin{figure}[t!]
\centering
\includegraphics[width=\columnwidth]{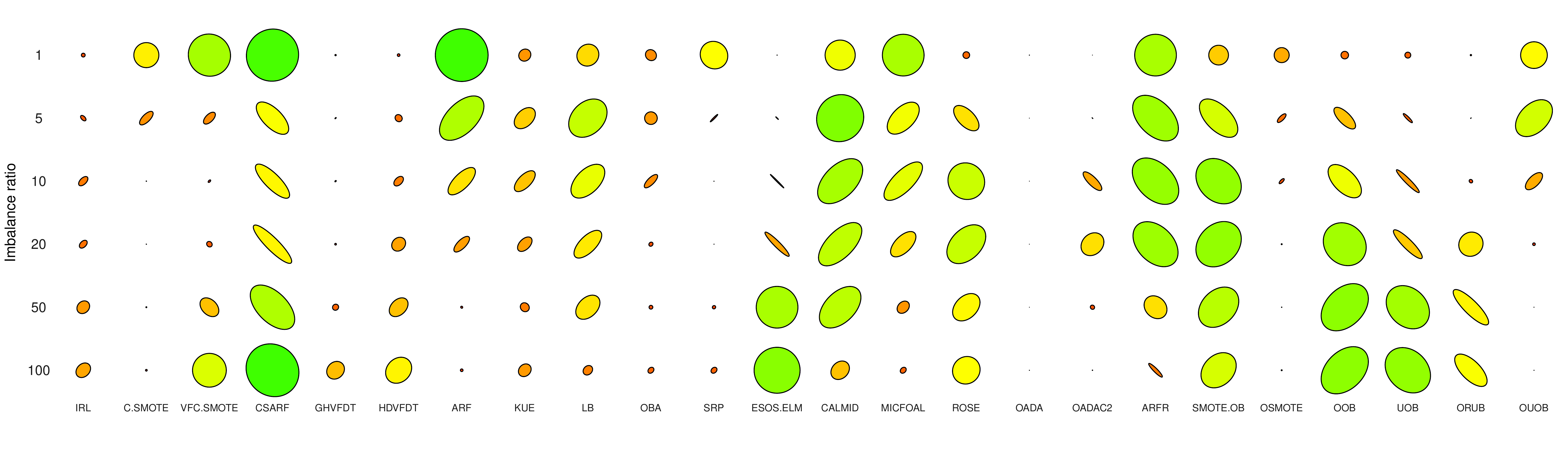}
\caption{Comparison of all 24 algorithms for merging minority clusters on static class imbalance ratio. Axes of the ellipse represent G-Mean and Kappa metrics. Color gradient represents the product of both metrics.}
\label{fig:BC_ILD_SIR_merging}
\end{figure}

Figures~\ref{fig:ild_static_borderline} and \ref{fig:ild_static_rare} present the performance of the five selected classifiers with the increasing presence of borderline and rare instances respectively under static imbalance ratio. Figures~\ref{fig:ild_static_move}, \ref{fig:ild_static_split}, \ref{fig:ild_static_merge} illustrate the performance of the same classifiers with changes in the spatial distribution of minority class instances under static imbalance. Table~\ref{tab:ILD_static} presents the average G-Mean and Kappa for the top 10 classifiers for each imbalance ratio and a given instance-level difficulty, and their average ranking. The overall performance of all classifiers regarding each of the instance difficulties is presented in Figures~\ref{fig:BC_ILD_SIR_borderline},\ref{fig:BC_ILD_SIR_rare},\ref{fig:BC_ILD_SIR_moving},\ref{fig:BC_ILD_SIR_splitting},\ref{fig:BC_ILD_SIR_merging}, in which axes of the ellipse represent G-Mean and Kappa metrics, the more rounded the better, and the color represents the product of both metrics.

\begin{figure}[t!]
\vspace*{1cm}
\centering
\includegraphics[width=0.19\columnwidth]{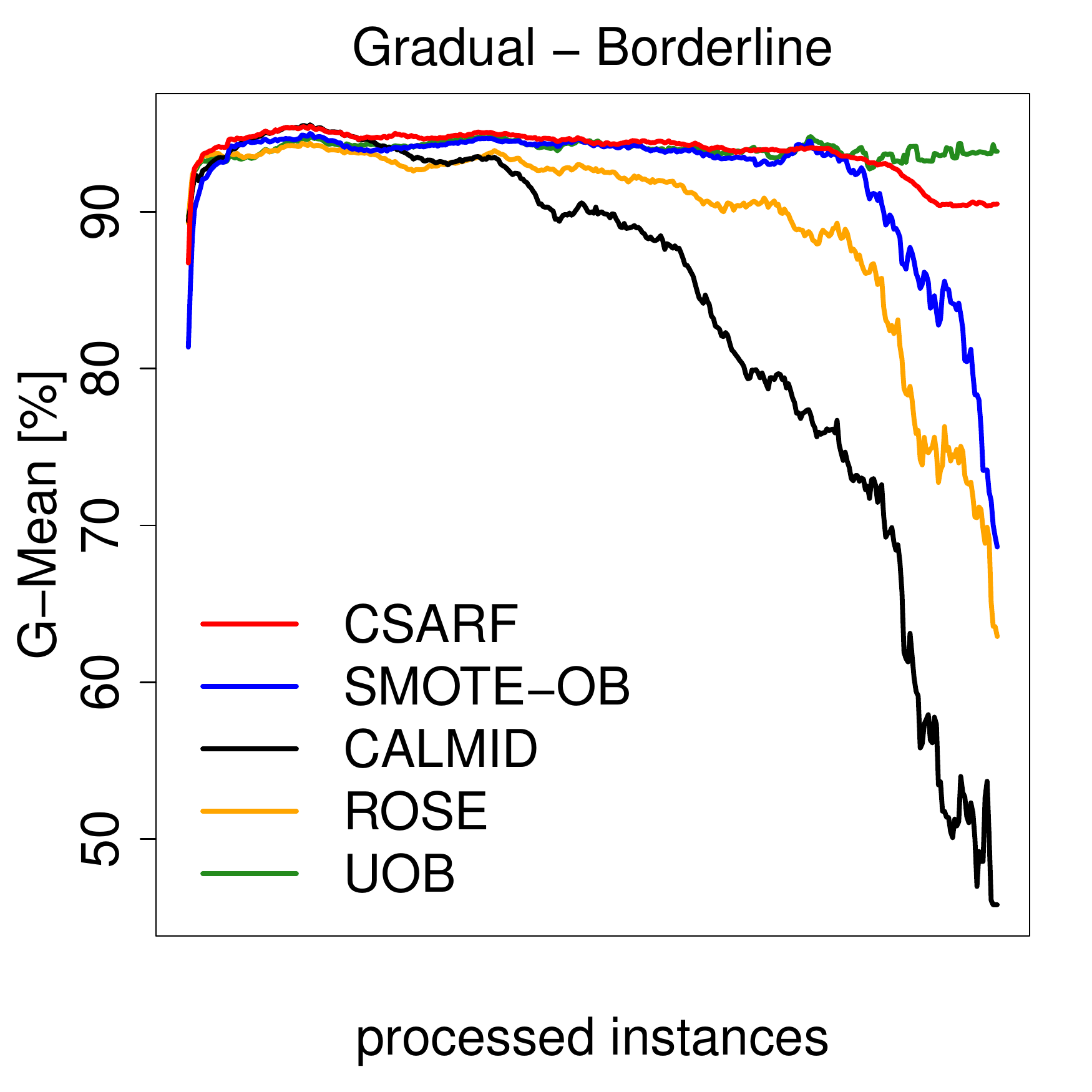}
\includegraphics[width=0.19\columnwidth]{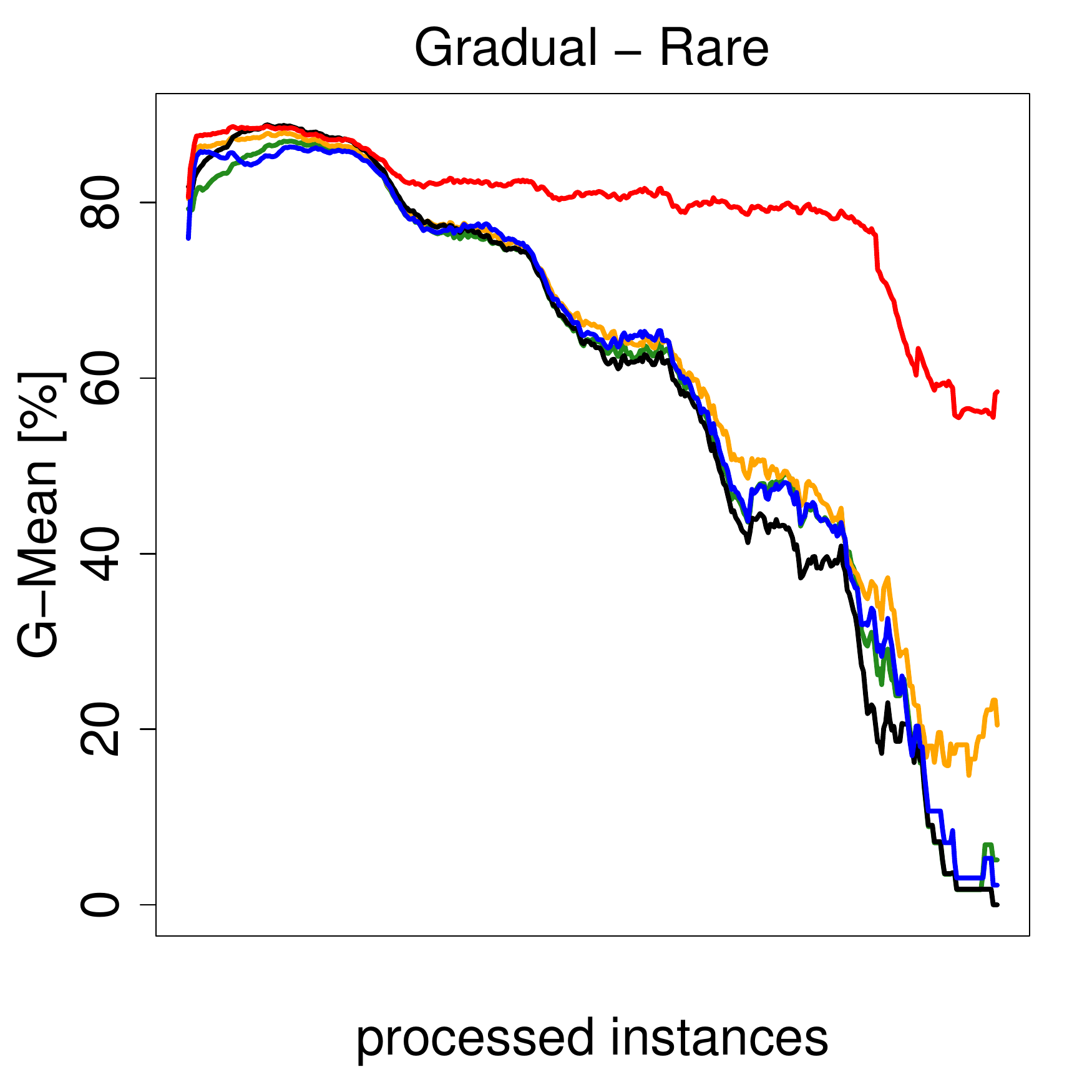}
\includegraphics[width=0.19\columnwidth]{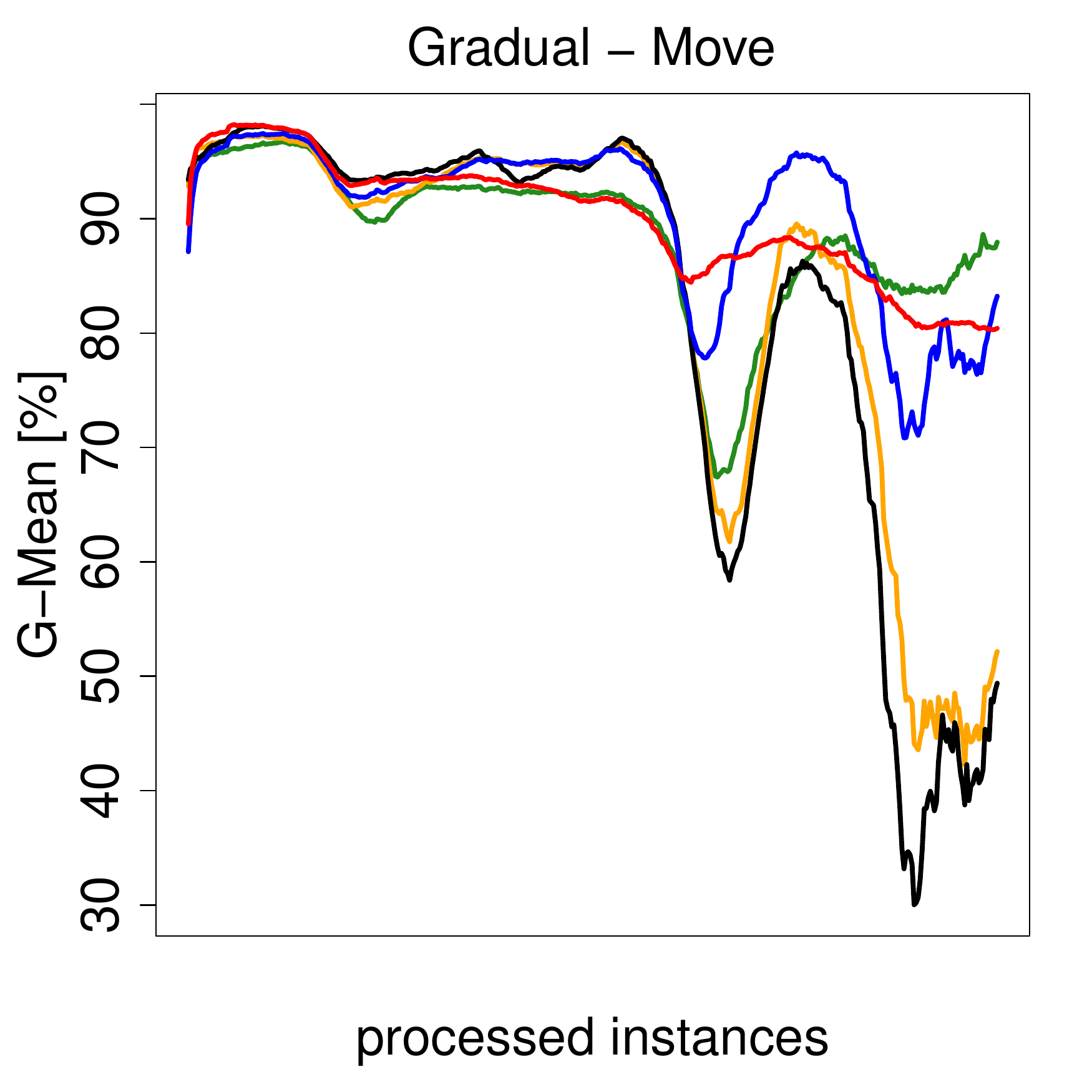}
\includegraphics[width=0.19\columnwidth]{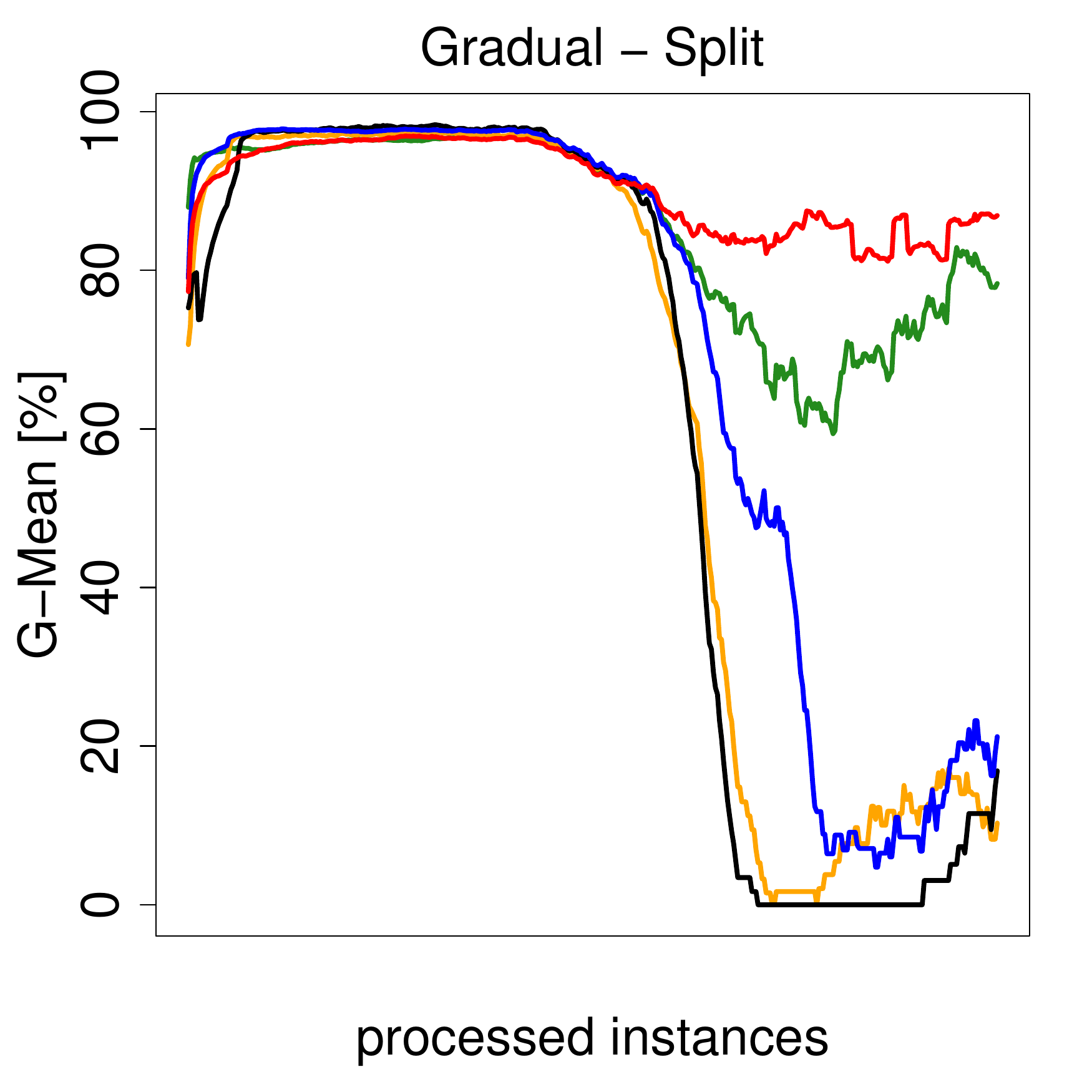}
\includegraphics[width=0.19\columnwidth]{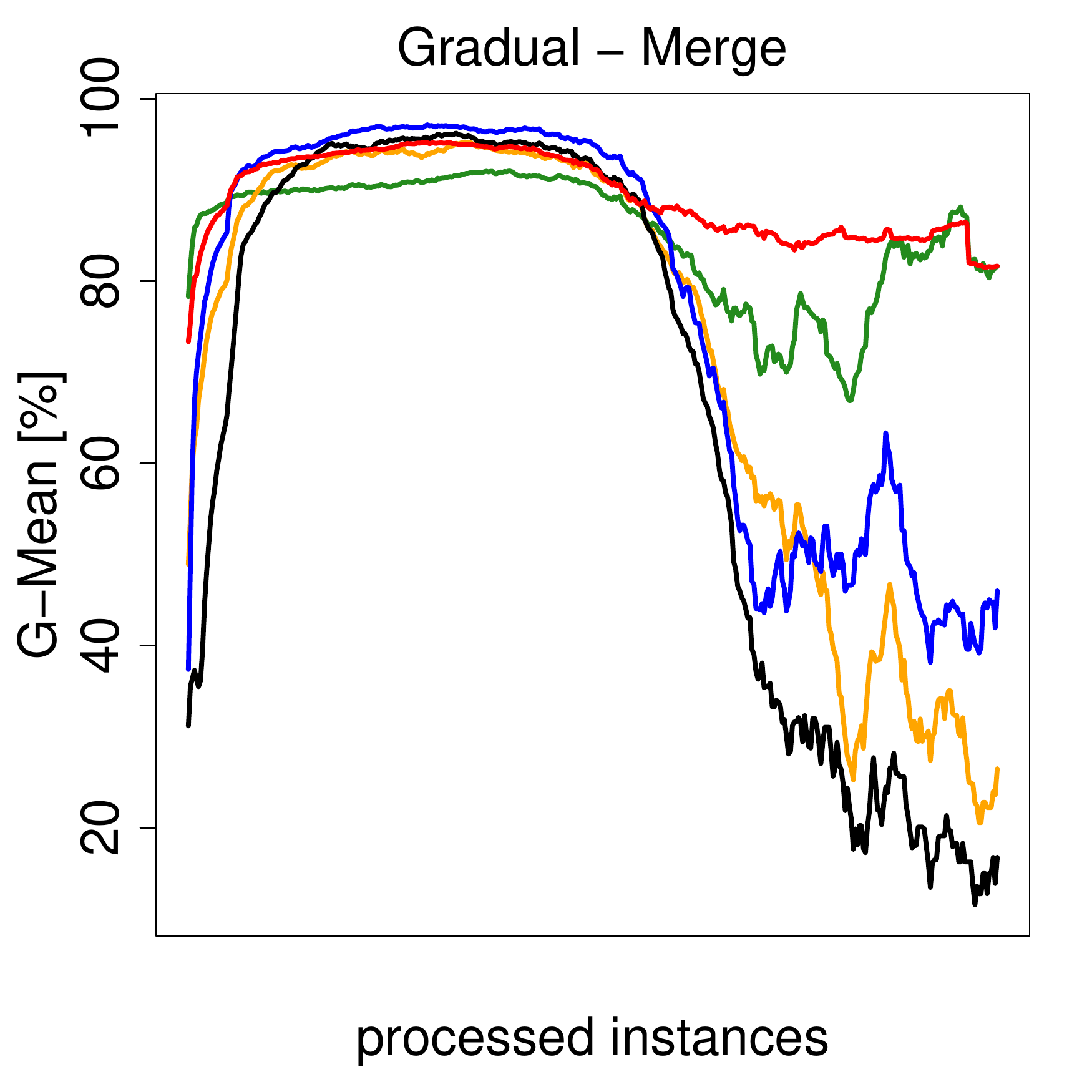}
\includegraphics[width=0.19\columnwidth]{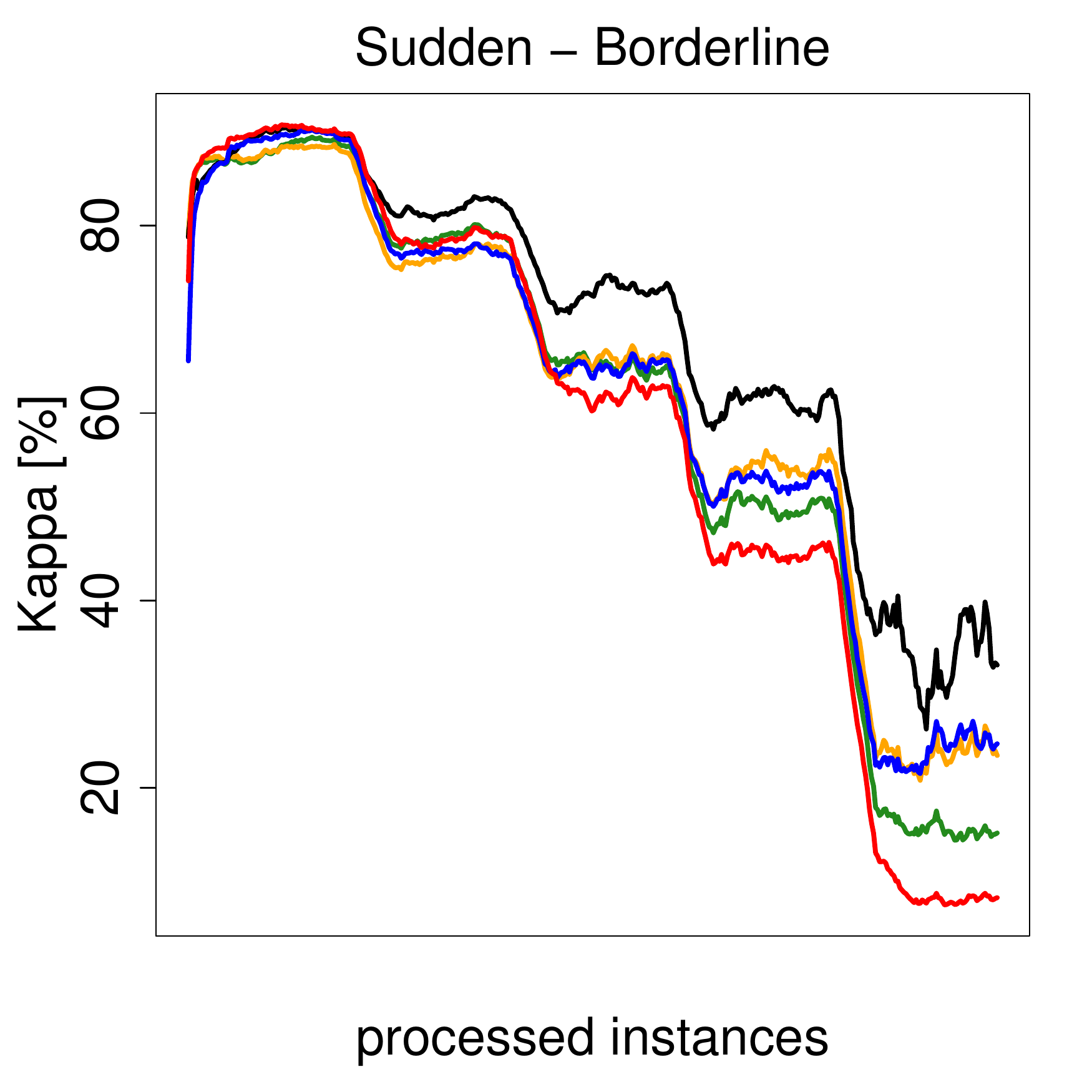}
\includegraphics[width=0.19\columnwidth]{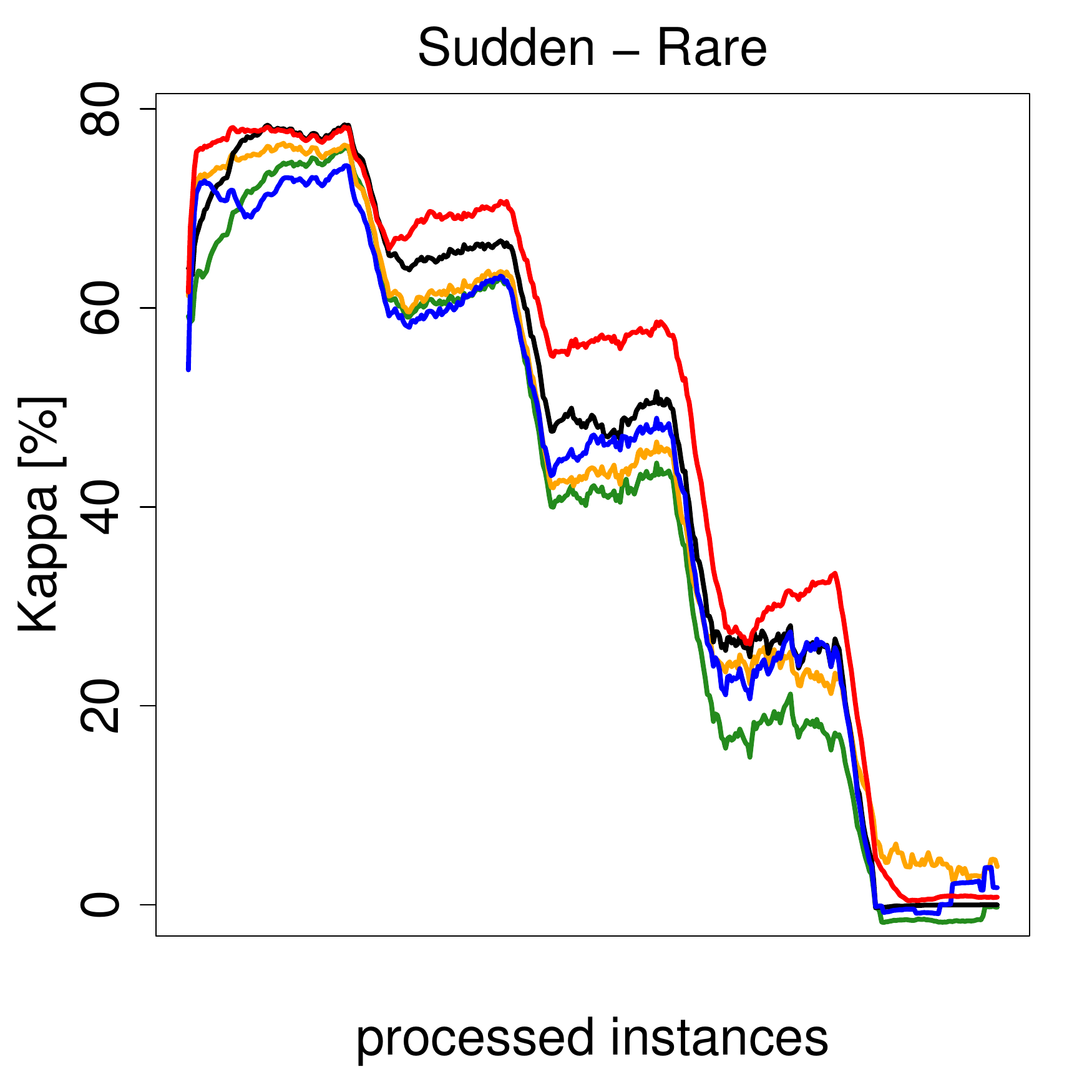}
\includegraphics[width=0.19\columnwidth]{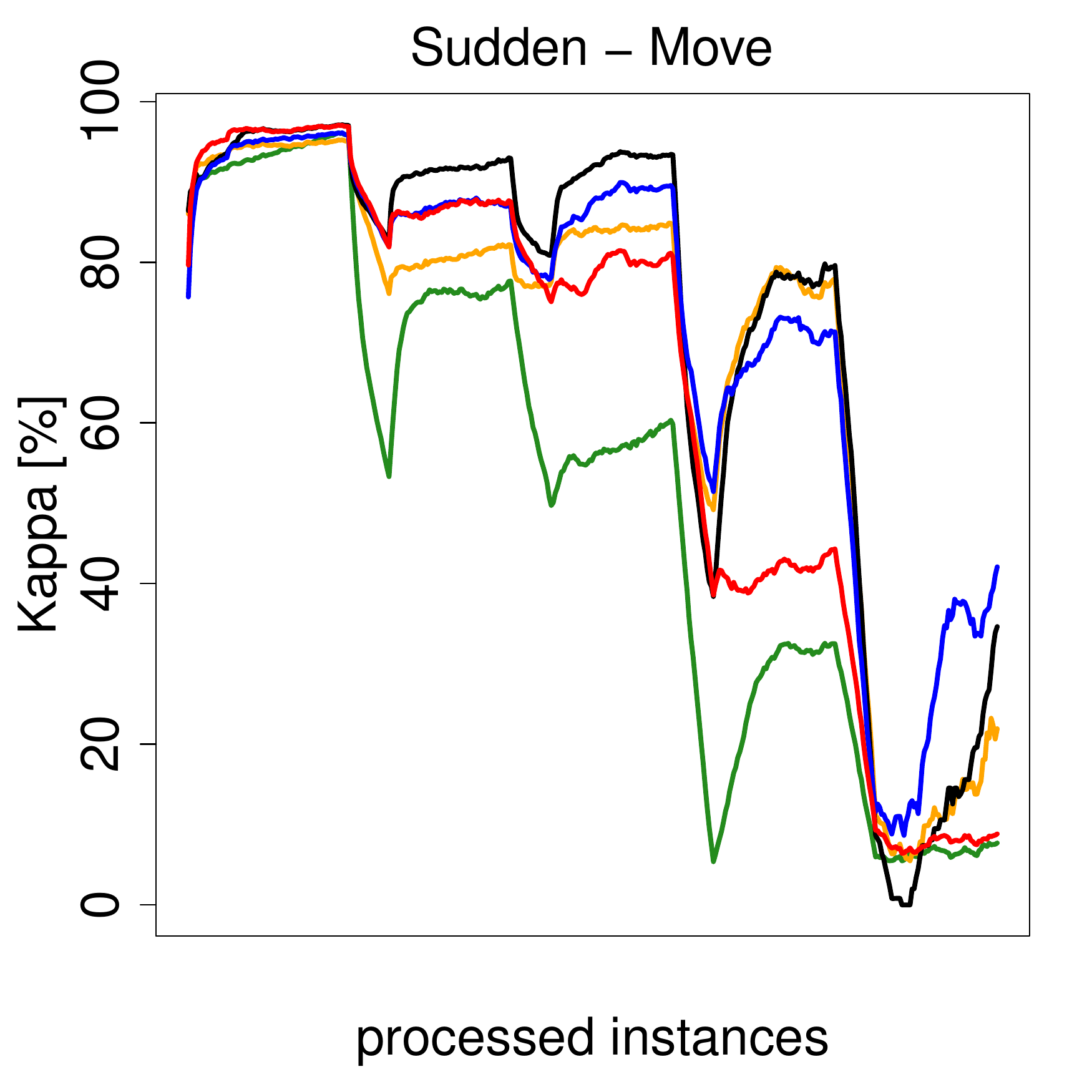}
\includegraphics[width=0.19\columnwidth]{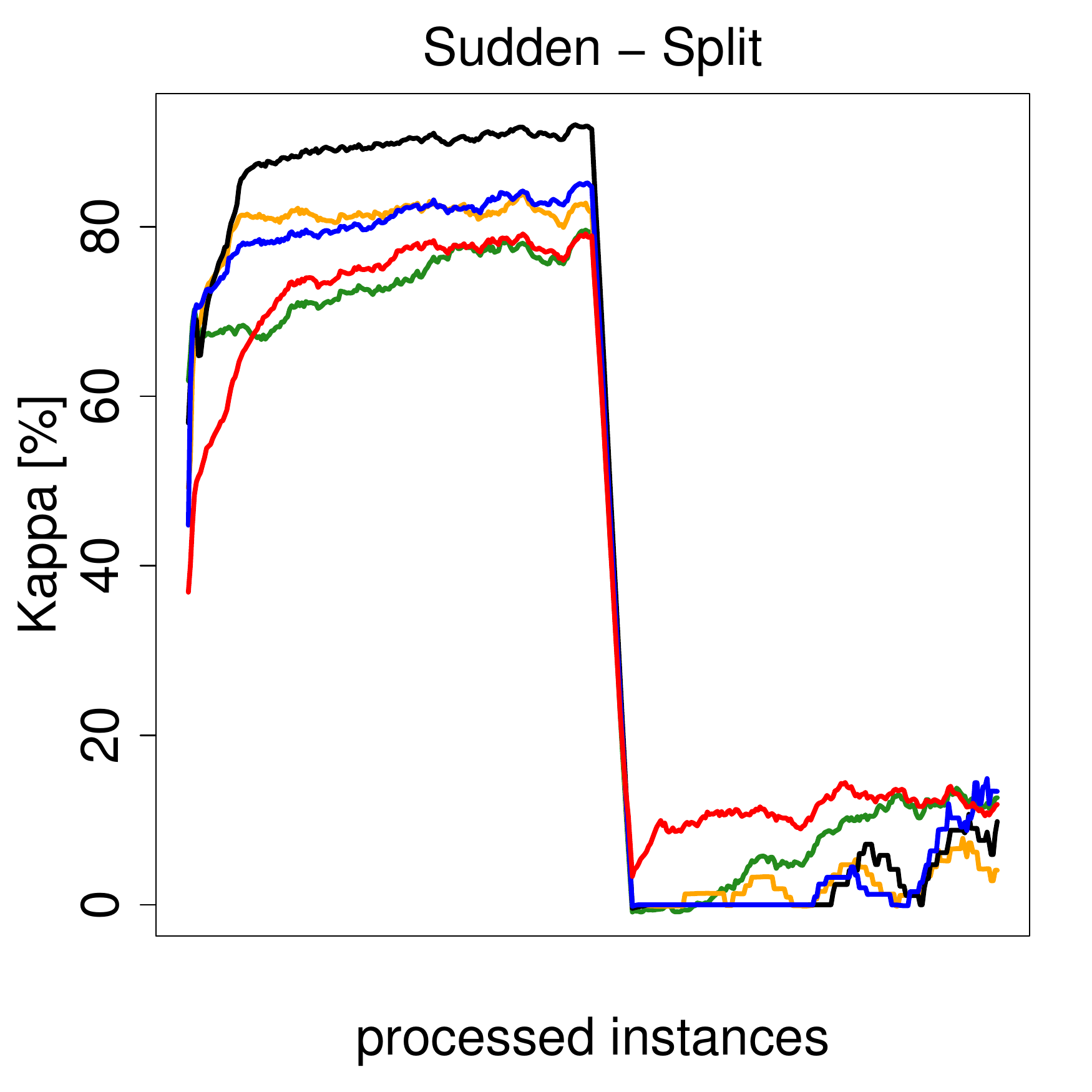}
\includegraphics[width=0.19\columnwidth]{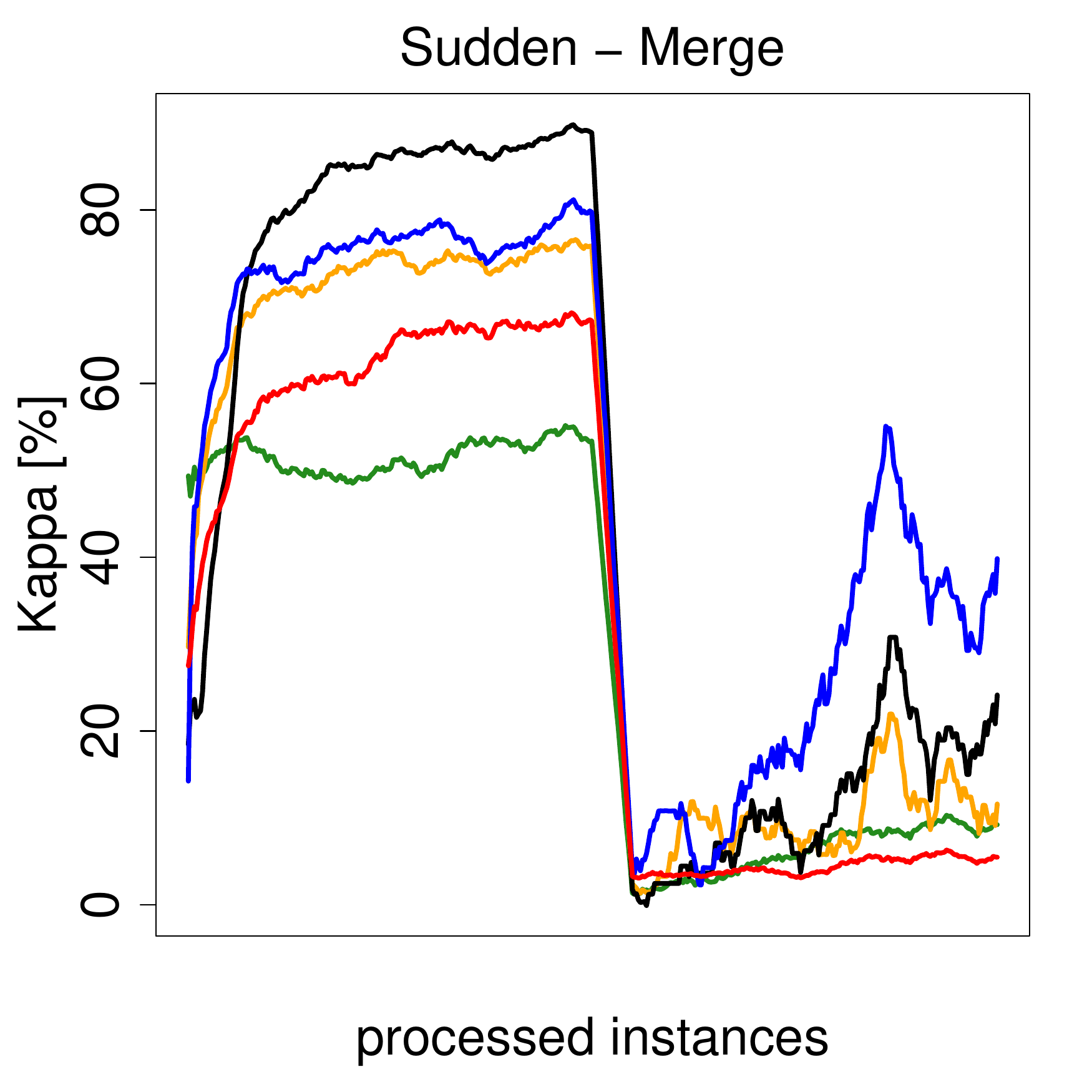}
\caption{G-Mean and Kappa on increasing borderline, rare, moving, splitting, and merging minority clusters and increasing imbalance ratio.}
\label{fig:ild_increasing_borderline_rare_move_split_merge}
\vspace*{1cm}
\end{figure}

\begin{figure}[t!]
\centering
\includegraphics[width=0.32\columnwidth]{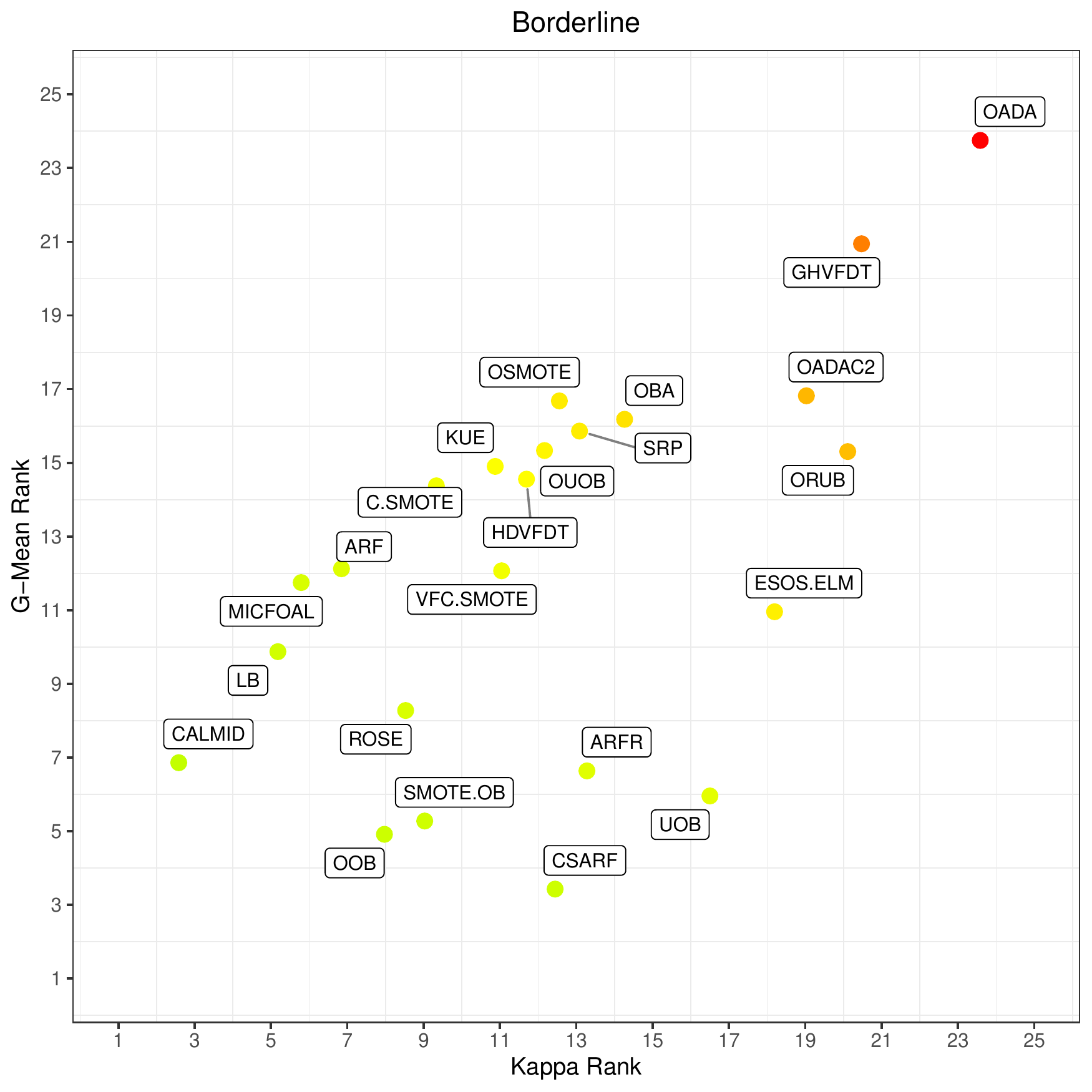}
\includegraphics[width=0.32\columnwidth]{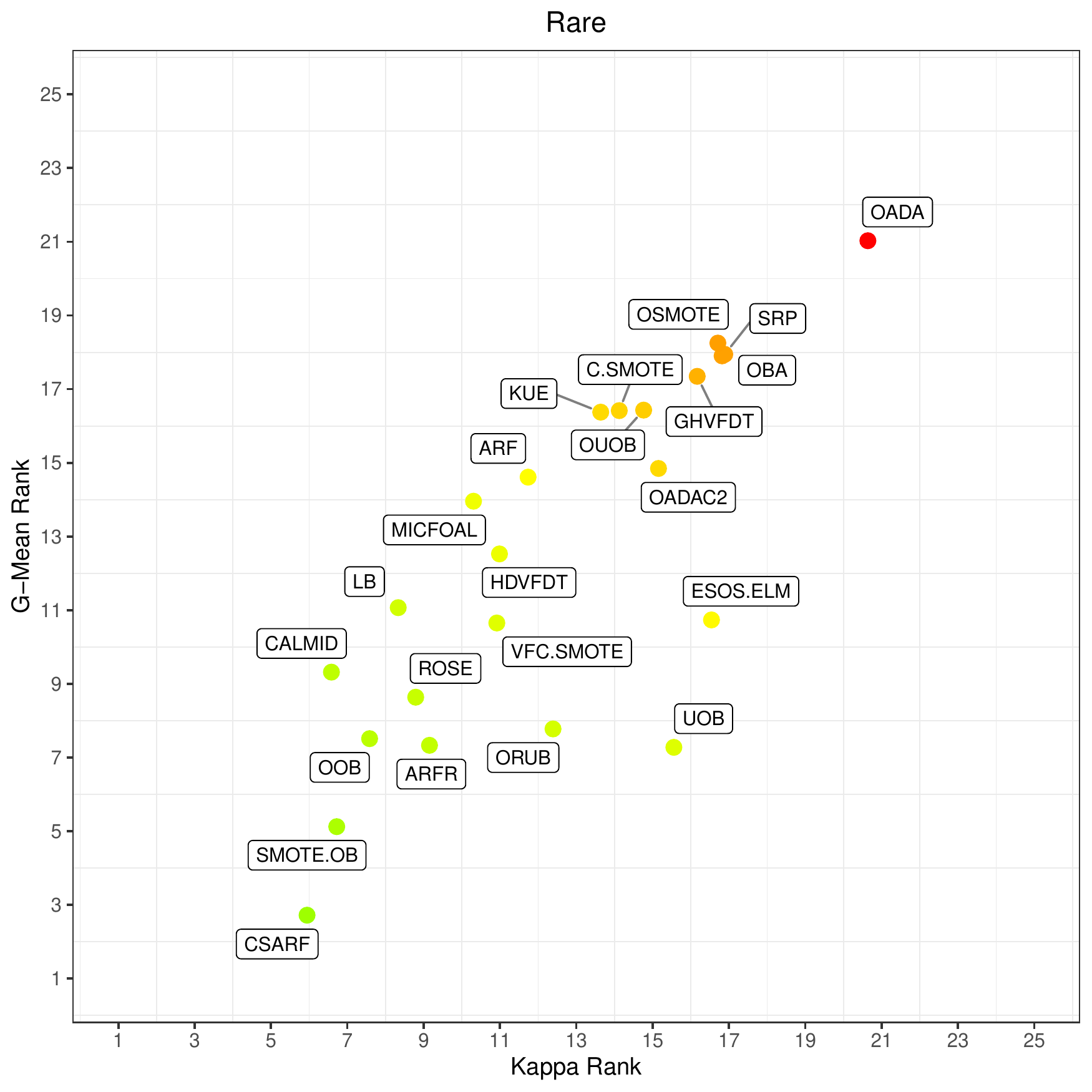}

\includegraphics[width=0.32\columnwidth]{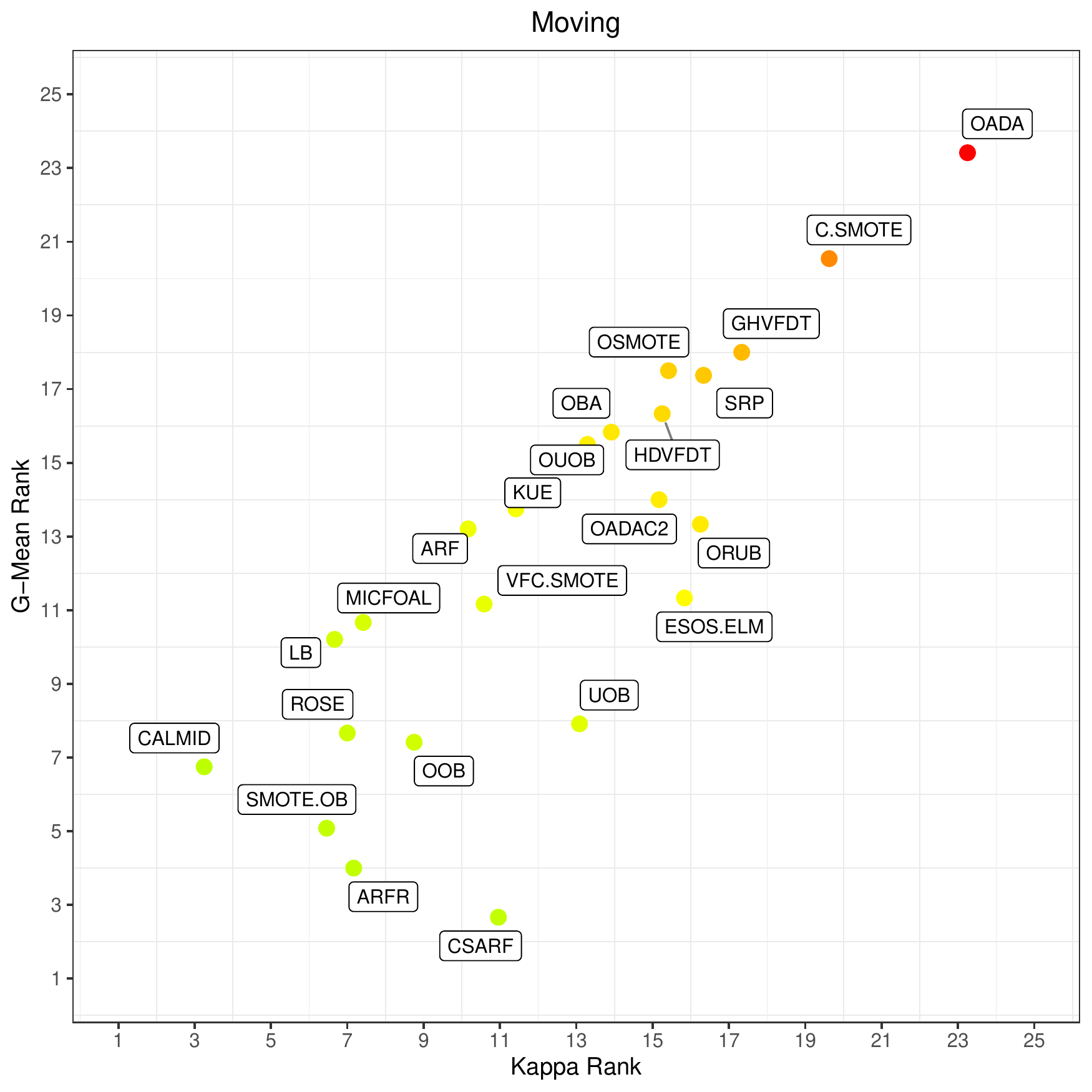}
\includegraphics[width=0.32\columnwidth]{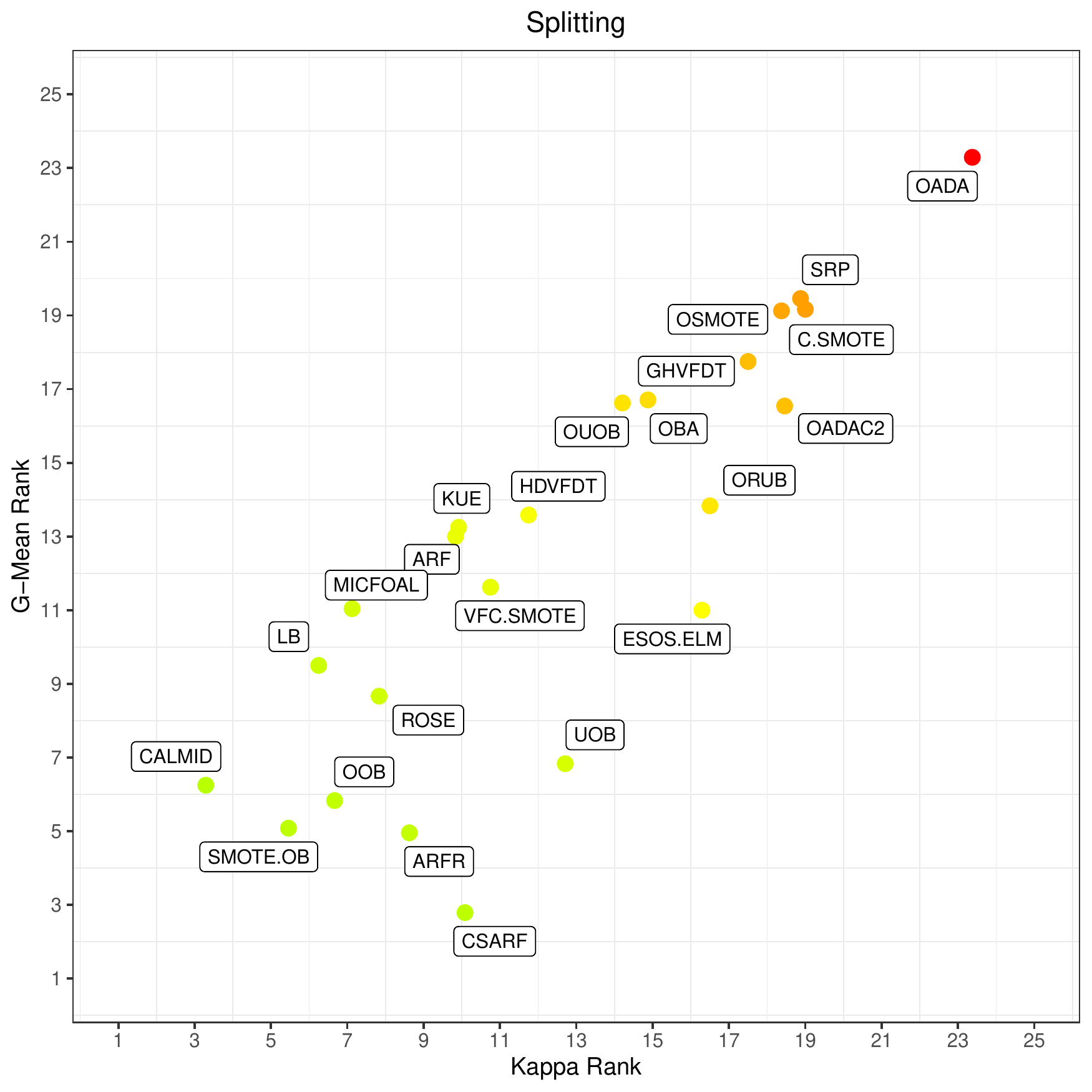}
\includegraphics[width=0.32\columnwidth]{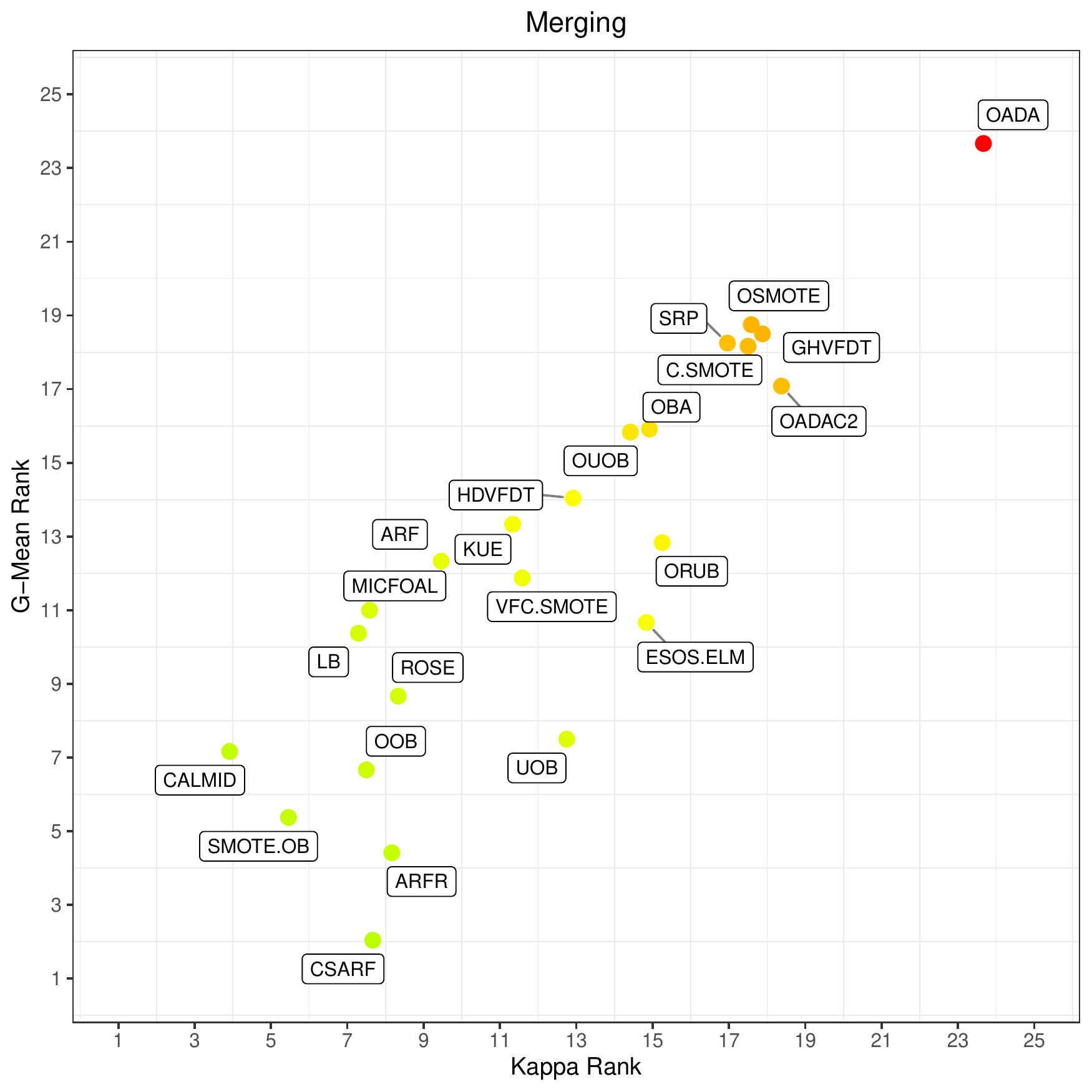}
\caption{Comparison of all 24 algorithms on borderline, rare, moving, splitting, merging minority clusters and increasing imbalance ratio. Color gradient represents the product of both metrics.}
\label{fig:BC_ILD_IIR_scatter}
\end{figure}

Considering the instance-level difficulties combined with dynamic imbalance ratio, Figure~\ref{fig:ild_increasing_borderline_rare_move_split_merge} illustrates the performance of the classifiers with increasing imbalance ratio. Table~\ref{tab:BC_ILD_increasing_IR} presents the average G-Mean and Kappa for the top 10 classifiers for increasing imbalance ratio in the presence of instance-level difficulties, and their overall ranking. To summarize, Figure~\ref{fig:BC_ILD_IIR_scatter} shows the overall performance of all classifiers for each instance-level difficulty.


\noindent \textbf{Discussion}

\noindent \textit{Impact of approach to class imbalance.} First, let us look on how different mechanisms for ensuring robustness to class imbalance tend to perform under diverse data-level difficulties. While guided resampling solutions usually perform better than their blind counterparts, here we can see that most of approaches based on \acrshort{smote} tend to fail. This can be explained by reliance of \acrshort{smote} on the neighborhood. Borderline and rare instances create non-homogeneous neighborhoods that are characterized by high overlapping and classification uncertainty. Oversampling such areas will lead to enhancing these undesirable qualities, instead of simplifying the classification task. This is especially visible for rare instances, where \acrshort{smote}-based methods deliver the worst performance. Rare instances do not form homogeneous neighborhoods and in streaming setup may appear infrequently, leading to lack of both spatial and temporal coherencies. This undermines the basic assumptions of \acrshort{smote}-based algorithms and renders them ineffective. Only \acrshort{smoteob} can handle both types of minority instances, as well as various minority clusters. This can be explained by using bags of instances (subsets) for training, which may offer better separation of instances and impose partial coherence on artificially generated instances. Blind oversampling employed by \acrshort{oob} also performs well under data-level difficulties, as it does not rely on the neighborhood analysis. However, when dealing with rare instances, \acrshort{oob} tends to significantly fail, as it amplifies these often-scattered instances, leading to overfitted decision boundaries. Where \acrshort{oob} and \acrshort{uob} excel is for minority class clusters, as due to their online nature they can swiftly adapt to changes in cluster structures and resample even small drifts to make them viable for their base learners. Algorithms based on training modifications, such as \acrshort{hdvfdt}, \acrshort{rose}, or \acrshort{calmid} can handle well all types of difficulties. Their robustness lies in their data manipulation and usage of modified mechanisms for training, which can display all-around, yet implicit, robustness to such challenges. Especially for Kappa metric, \acrshort{rose} and \acrshort{calmid} can be seen as good choices for these scenarios. Their offshoot, a cost-sensitive approach of \acrshort{csarf}, displays best performance on G-Mean metric, yet suffers under Kappa evaluation. This shows that \acrshort{csarf} strongly focuses on the minority class, but this hits back in the form of an increased number of false positives. So, while it is capable of handling minority difficulties and clusters by cost penalty-lead training, it increases the false positives in these overlapping or uncertain regions.

\begin{table*}[t!]
\centering
\footnotesize
\setlength{\tabcolsep}{3pt}
\caption{G-Mean and Kappa averages on borderline, rare, moving, splitting, merging minority clusters for static imbalance ratio.}
\label{tab:ILD_static}
\begin{tabular}{lll|C{1cm}C{1cm}C{1cm}C{1cm}C{1cm}C{1cm}C{1cm}C{1cm}C{1cm}C{1cm}}
\toprule
\multicolumn{3}{c}{Instance-level diff.} & CSARF & ARF & KUE & LB & CALMID & ROSE & ARFR & SMOTE-OB & OOB & UOB\\
\midrule
\multirow{10}{*}{\rotatebox[origin=c]{90}{G-Mean}} &
\multirow{5}{*}{\rotatebox[origin=c]{90}{IR 5}} &
Borderline & 95.02 & 93.05 & 92.93 & 93.95 & 94.49 & 94.67 & \textbf{95.27} & 95.09 & 95.04 & 94.25\\
&&Rare & \textbf{82.62} & 61.15 & 59.84 & 61.64 & 64.69 & 70.21 & 74.15 & 82.05 & 67.86 & 72.57\\
&&Moving & 95.45 & 94.81 & 92.75 & 94.99 & 95.70 & 95.24 & \textbf{96.32} & 95.70 & 93.51 & 92.69\\
&&Splitting & 95.93 & 95.06 & 94.34 & 95.25 & 96.45 & 95.51 & \textbf{96.86} & 96.20 & 95.37 & 94.13\\
&&Merging & 95.47 & 94.06 & 92.57 & 94.55 & 95.57 & 94.44 & \textbf{96.39} & 95.86 & 94.28 & 92.55\\
\cmidrule{2-13}
& \multirow{5}{*}{\rotatebox[origin=c]{90}{IR 100}} &
Borderline & 87.78 & 16.83 & 18.57 & 24.24 & 28.75 & 31.98 & 14.27 & 47.38 & 64.69 & \textbf{90.09}\\
&&Rare & \textbf{80.76} & 2.55 & 3.03 & 7.74 & 10.53 & 15.59 & 10.10 & 57.74 & 39.45 & 66.26\\
&&Moving & \textbf{85.32} & 0.06 & 5.84 & 3.81 & 9.53 & 14.20 & 10.50 & 18.86 & 53.82 & 84.30\\
&&Splitting & \textbf{86.32} & 1.05 & 4.27 & 3.49 & 9.06 & 15.84 & 11.19 & 24.81 & 61.12 & 73.18\\
&&Merging & \textbf{86.14} & 0.88 & 4.26 & 3.64 & 5.35 & 9.62 & 6.18 & 11.47 & 51.49 & 84.44\\
\midrule
\multirow{10}{*}{\rotatebox[origin=c]{90}{Kappa}} &
\multirow{5}{*}{\rotatebox[origin=c]{90}{IR 5}} &
Borderline & 80.22 & 84.16 & 82.71 & 84.19 & \textbf{84.38} & 80.30 & 81.52 & 79.91 & 80.70 & 77.26\\
&&Rare & \textbf{70.14} & 51.37 & 50.18 & 51.64 & 53.56 & 55.70 & 61.70 & 58.85 & 52.83 & 47.03\\
&&Moving & 82.56 & \textbf{89.00} & 84.87 & 88.27 & 88.70 & 84.55 & 86.94 & 83.77 & 79.74 & 76.18\\
&&Splitting & 84.29 & \textbf{90.40} & 87.49 & 88.97 & 90.34 & 85.03 & 88.67 & 85.76 & 83.99 & 79.91\\
&&Merging & 82.18 & \textbf{88.99} & 84.75 & 88.03 & 88.59 & 81.57 & 86.74 & 84.65 & 80.74 & 73.56\\
\cmidrule{2-13}
& \multirow{5}{*}{\rotatebox[origin=c]{90}{IR 100}} &
Borderline & 13.22 & 13.52 & 15.08 & 20.03 & 23.86 & 25.27 & 2.51 & 26.03 & \textbf{38.09} & 11.97\\
&&Rare & 13.56 & 1.96 & 2.16 & 5.94 & 8.43 & 10.68 & 3.00 & \textbf{43.78} & 24.51 & 4.96\\
&&Moving & 9.19 & 0.05 & 4.14 & 2.80 & 7.37 & 9.31 & 1.79 & 11.19 & \textbf{31.14} & 10.98\\
&&Splitting & 10.44 & 0.83 & 3.09 & 2.62 & 6.98 & 9.98 & 1.92 & 13.20 & \textbf{36.78} & 9.94\\
&&Merging & 10.92 & 0.65 & 3.27 & 2.66 & 4.05 & 6.36 & 1.33 & 8.15 & \textbf{34.58} & 8.94\\
\midrule
\multirow{2}{*}{All IRs} & \multicolumn{2}{l|}{Avg. G-Mean} & \textbf{88.40} & 52.60 & 53.32 & 57.81 & 62.98 & 65.06 & 64.94 & 77.21 & 75.87 & 83.10\\
& \multicolumn{2}{l|}{Avg. Kappa} & 50.01 & 46.06 & 45.16 & 49.79 & 53.68 & 51.26 & 47.05 & \textbf{56.75} & 54.67 & 41.29\\
\midrule
\multirow{2}{*}{All IRs} & \multicolumn{2}{l|}{Rank G-Mean} & \textbf{2.06} & 8.13 & 9.17 & 7.10 & 5.69 & 5.78 & 4.42 & 3.61 & 4.34 & 4.71\\
& \multicolumn{2}{l|}{Rank Kappa} & 5.65 & 5.63 & 7.06 & 4.49 & \textbf{2.85} & 5.29 & 5.99 & 4.52 & 4.87 & 8.64\\
\bottomrule
\end{tabular}
\end{table*}

\noindent \textit{Impact of ensemble architecture.} We can observe that all best-performing methods for various types of data-level difficulties are based either on bagging or hybrid architectures. All boosting methods are among the worst performing ones. This can be explained by the nature of boosting, as it focuses on correcting the mistakes of the previous classifier in the ensemble. Rare, borderline, or clustered minority instances will always introduce a high uncertainty into the training procedure. This may significantly destabilize boosting, as by focusing on correcting errors on those uncertain instances it will be continuously introducing other errors, locking itself in a cycle of never reducing the overall error. Bagging methods offer natural partitioning of instances, allowing to break difficult neighborhood or clusters and introduce more instance-level diversity into base classifiers. This aids the used mechanisms for handling class imbalance, making bagging methods more robust to scenarios where learning difficulties lie in spatial characteristics of data. 

\noindent \textit{Comparison with standard ensembles.} Interestingly, general-purpose ensembles display better robustness to various instance-level difficulties than over half of the classifiers dedicated to imbalanced data streams. \acrshort{lb}, \acrshort{arf}, and \acrshort{kue} can relatively effectively handle both types of difficult instances, as well as various types of evolving clusters within the minority class. They always significantly outperform all methods based on boosting and most of approaches using informative oversampling (except for \acrshort{smoteob}). Of course, we can observe a drop in their performance with the increase of imbalance ratios, yet even for IR = 100 they can perform better than several dedicated approaches. Their robustness to instance types can be explained by the fact that all three mentioned ensembles use instance subsets for training their base classifiers. Therefore, such subsampling may implicitly lead to more sparse neighborhoods (reducing overlapping and uncertainty) and thus to reduction of difficulty levels for certain instances. When analyzing the robustness to evolving clusters, one can explain this by concept drift adaptation mechanisms employed by \acrshort{lb}, \acrshort{arf}, and \acrshort{kue}. Changes in minority class clusters can be picked up by their drift detectors, leading to adaptation to the current state of the minority class. Therefore, any splitting or merging of clusters will be picked up as changes in data distributions and managed by simple online adaptation to the most recent instances. Finding that such general-purpose ensembles display significant robustness to data-level difficulties stands as a testament to how well designed those methods are. However, to excel when dealing with such challenging learning scenario, highly specialized ensemble models enhance with skew-insensitive mechanisms can deliver much better performance. 

\noindent \textit{Relationships between instance-level difficulties and imbalance ratios.} When analyzing algorithms for their robustness to data-level difficulties, we must understand the relationship between them and the class imbalance ratio. Ideally, we are looking for a method that will be insensitive to changing imbalance ratios and will display stable robustness to data-level difficulties. Most of the existing algorithms do not possess this quality, displaying either drops in the performance with increasing imbalance ratio (\textit{e.g.} \acrshort{micfoal}), or lack of any stability (\textit{e.g.} \acrshort{vfcsmote}). The most reliable methods are \acrshort{csarf}, \acrshort{smoteob}, \acrshort{oob}, and \acrshort{rose} that offer stable, or improving, robustness with increasing imbalance. It is important to note that \acrshort{csarf} performance is skewed towards G-Mean, while the remaining methods tend to perform well on both metrics.

\noindent \textit{What difficulties are the most challenging.} While analyzing the performance of the methods, we can see significant drops in the performance in two scenarios: when dealing with rare instances and splitting/merging clusters. Rare instances are one of the biggest challenges for any imbalanced algorithms, as they combine small sample size, class overlapping, and potential presence of noise. With increasing ratio of rare instances, the minority class in the stream starts losing any coherent structure, converging towards a collection of sparsely distributed and spatially uncorrelated instances, more akin to a cloud of points than any structure. This makes the formulation of decision boundaries especially difficult and requires dedicated mechanisms that can either learn under small sample size or can create more coherent representations (either via resampling like \acrshort{oob}, or via instance buffers like \acrshort{rose}). Minority clusters pose even bigger challenge, as they force classifiers to track sub-concepts in minority classes (each cluster should be treated as a sub-concept). Both cases require fast adaptation and are strongly aided by a presence of underlying drift detector. With splitting clusters, previously learned decision boundaries become to general and are not able to capture the emergence of sub-concepts in minority class. With merging clusters, we are left with too complex decision boundaries that are not able to generalize well over the current state of the stream. 

\begin{table*}[t!]
\vspace*{1cm}
\centering
\footnotesize
\setlength{\tabcolsep}{4pt}
\caption{G-Mean and Kappa averages on borderline, rare, moving, splitting, merging minority clusters and increasing imbalance ratio.}
\label{tab:BC_ILD_increasing_IR}
\begin{tabular}{ll|C{1cm}C{1cm}C{1cm}C{1cm}C{1cm}C{1cm}C{1cm}C{1cm}C{1cm}C{1cm}}
\toprule
\multicolumn{2}{c}{Instance-level diff.} & CSARF & ARF & KUE & LB & CALMID & ROSE & ARFR & SMOTE-OB & OOB & UOB\\
\midrule
\multirow{5}{*}{\rotatebox[origin=c]{90}{G-Mean}}
&Borderline & 93.73 & 81.55 & 80.95 & 82.08 & 82.22 & 88.58 & 89.19 & 91.89 & 93.81 & \textbf{93.97}\\
&Rare & \textbf{77.98} & 55.37 & 54.81 & 55.94 & 55.42 & 60.65 & 55.85 & 56.78 & 56.58 & 56.34\\
&Moving & \textbf{92.00} & 66.93 & 69.21 & 71.70 & 80.45 & 81.54 & 79.93 & 87.91 & 89.35 & 87.61\\
&Splitting & \textbf{90.01} & 51.00 & 53.83 & 51.63 & 55.09 & 57.65 & 56.19 & 59.40 & 80.39 & 78.81\\
&Merging & \textbf{88.10} & 52.07 & 50.19 & 60.06 & 60.64 & 63.47 & 57.99 & 69.83 & 81.59 & 82.82\\
\midrule
\multirow{5}{*}{\rotatebox[origin=c]{90}{Kappa}}
&Borderline & 57.38 & 68.43 & 67.06 & 67.23 & \textbf{68.56} & 61.26 & 65.30 & 61.88 & 61.92 & 59.73\\
&Rare & \textbf{47.66} & 44.43 & 43.16 & 44.07 & 44.87 & 42.83 & 43.34 & 41.70 & 41.94 & 39.34\\
&Moving & 56.13 & 60.61 & 60.35 & 64.81 & \textbf{71.25} & 68.12 & 70.10 & 69.19 & 62.31 & 50.86\\
&Splitting & 42.51 & 46.48 & 46.84 & 46.36 & 48.25 & 45.38 & 45.78 & 45.53 & \textbf{50.90} & 42.41\\
&Merging & 34.26 & 44.70 & 42.53 & 50.28 & 51.19 & 46.32 & 43.12 & \textbf{51.75} & 50.50 & 30.08\\
\midrule
\multicolumn{2}{l|}{Avg. G-Mean} & \textbf{87.76} & 62.28 & 62.61 & 65.06 & 66.88 & 70.73 & 68.54 & 73.43 & 79.71 & 79.55\\
\multicolumn{2}{l|}{Avg. Kappa} & 47.83 & 53.15 & 52.28 & 54.55 & \textbf{56.63} & 52.30 & 53.46 & 53.60 & 53.06 & 45.40\\
\midrule
\multicolumn{2}{l|}{Rank G-Mean} & \textbf{1.42} & 9.17 & 9.58 & 7.42 & 7.25 & 4.75 & 6.25 & 3.75 & 2.75 & 2.67\\
\multicolumn{2}{l|}{Rank Kappa} & 8.00 & 4.17 & 5.67 & 3.75 & \textbf{1.71} & 6.50 & 5.08 & 5.63 & 5.17 & 9.33\\
\bottomrule
\end{tabular}
\vspace*{1cm}
\end{table*}

\clearpage

\subsubsection{Concept drift and static imbalance ratio}
\label{sec:bc-cd-static-ratio}

\begin{figure}[t!]
\centering
\includegraphics[width=0.19\columnwidth]{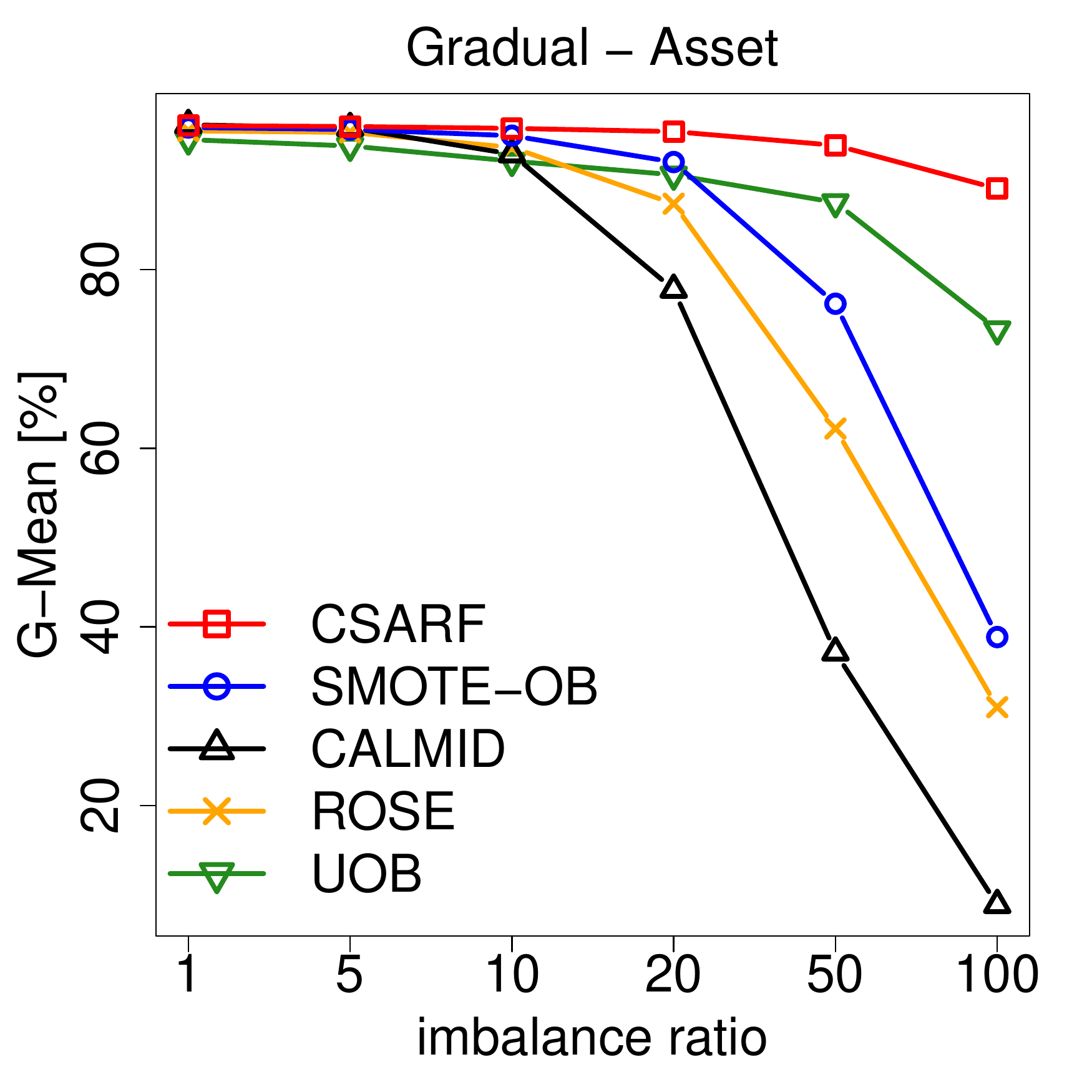}
\includegraphics[width=0.19\columnwidth]{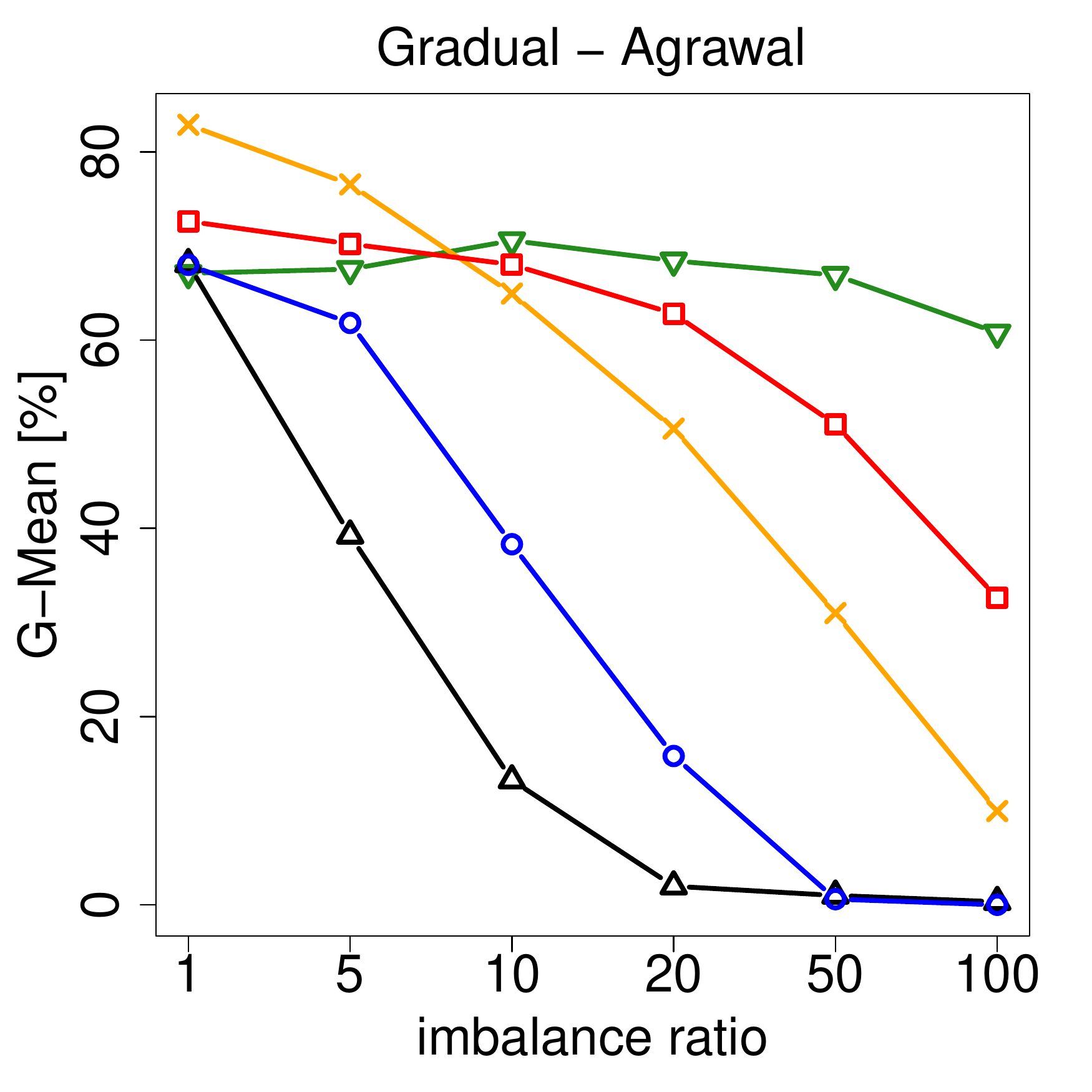}
\includegraphics[width=0.19\columnwidth]{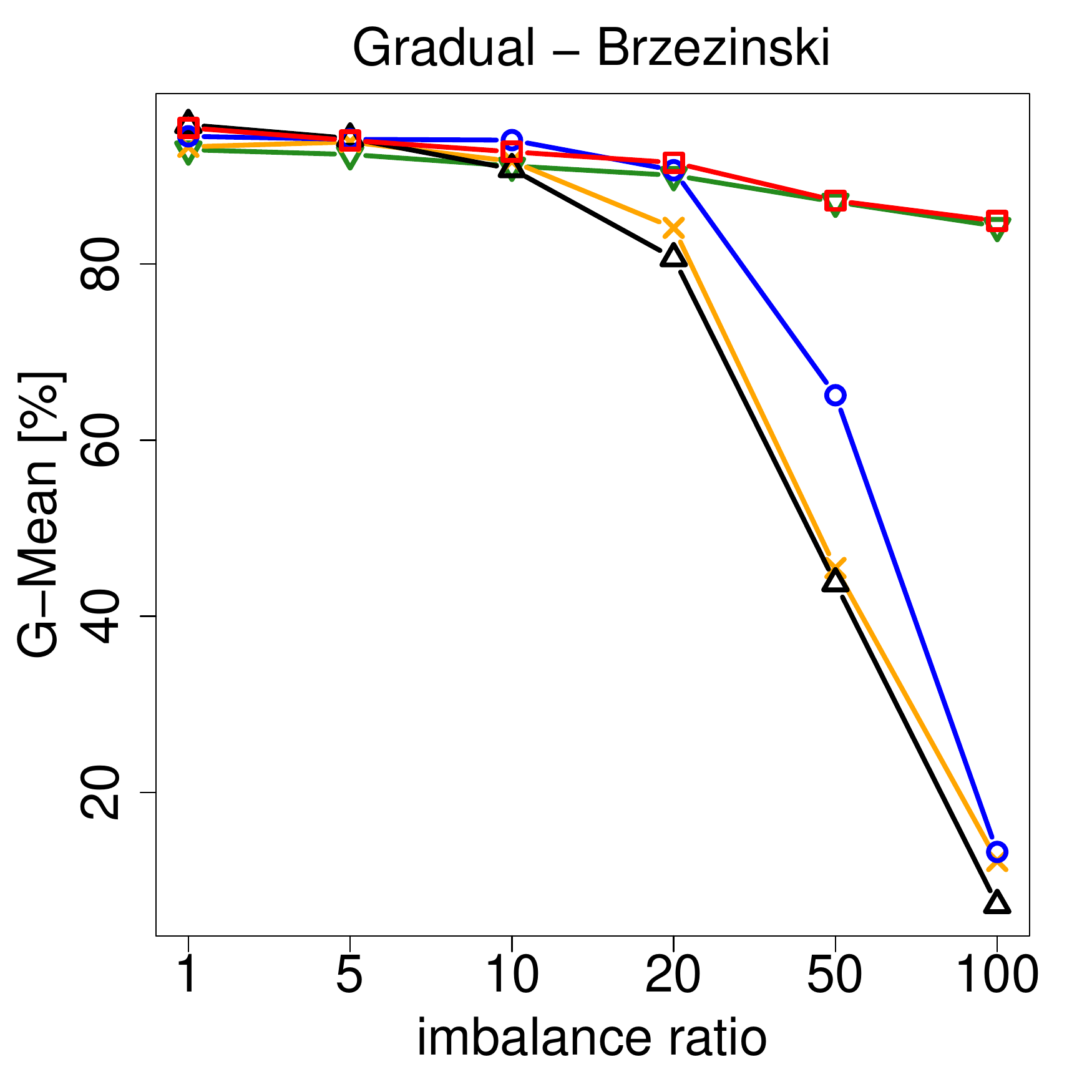}
\includegraphics[width=0.19\columnwidth]{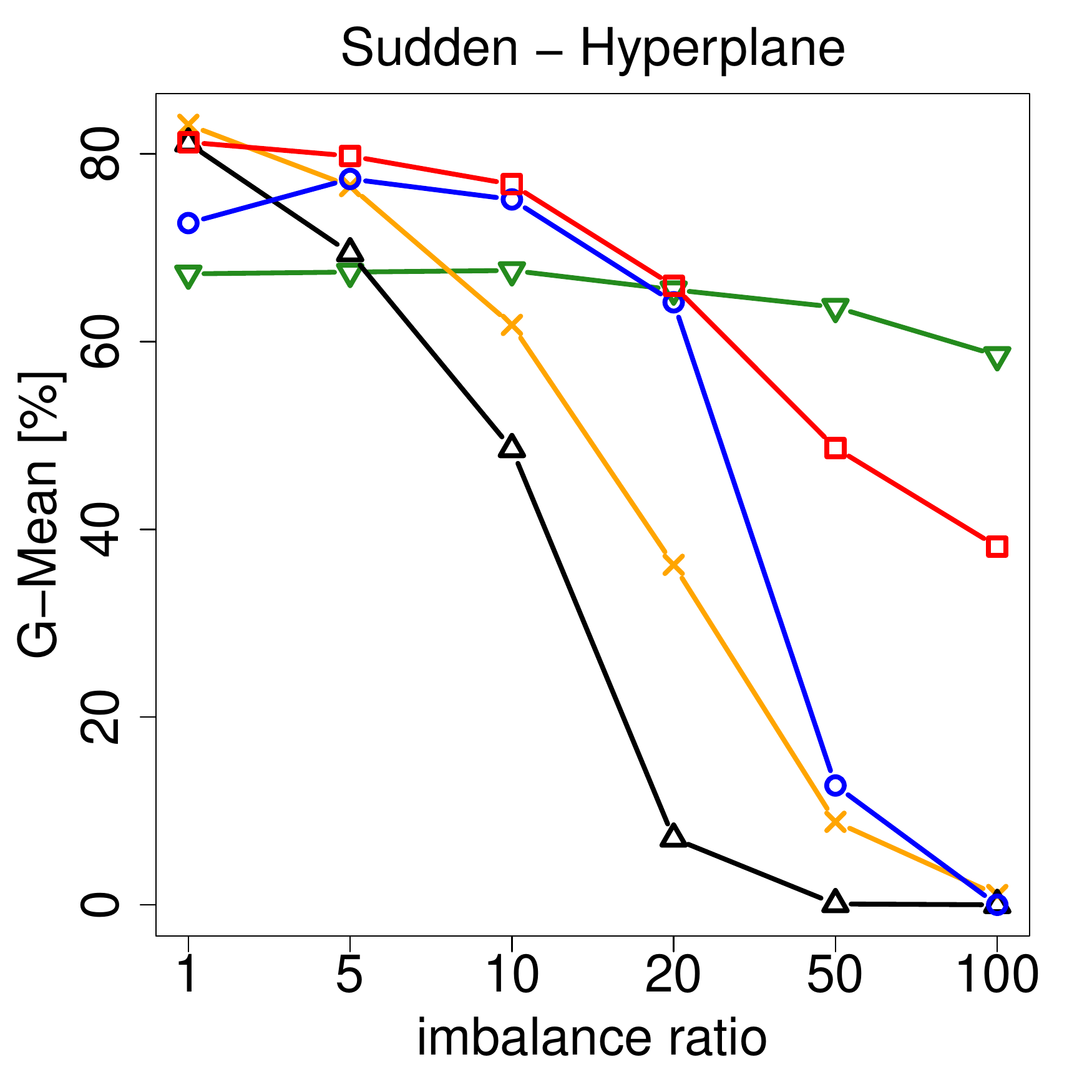}
\includegraphics[width=0.19\columnwidth]{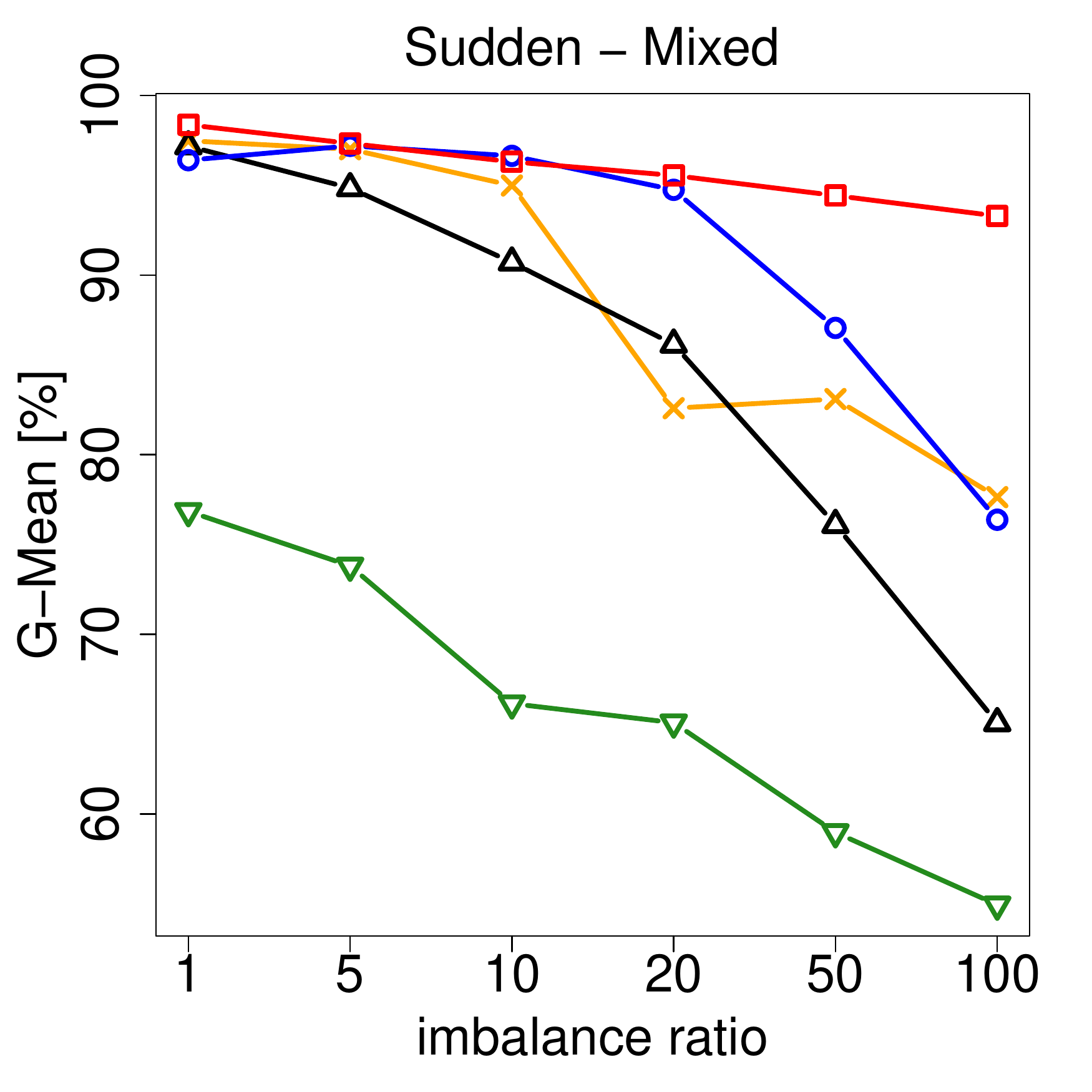}
\includegraphics[width=0.19\columnwidth]{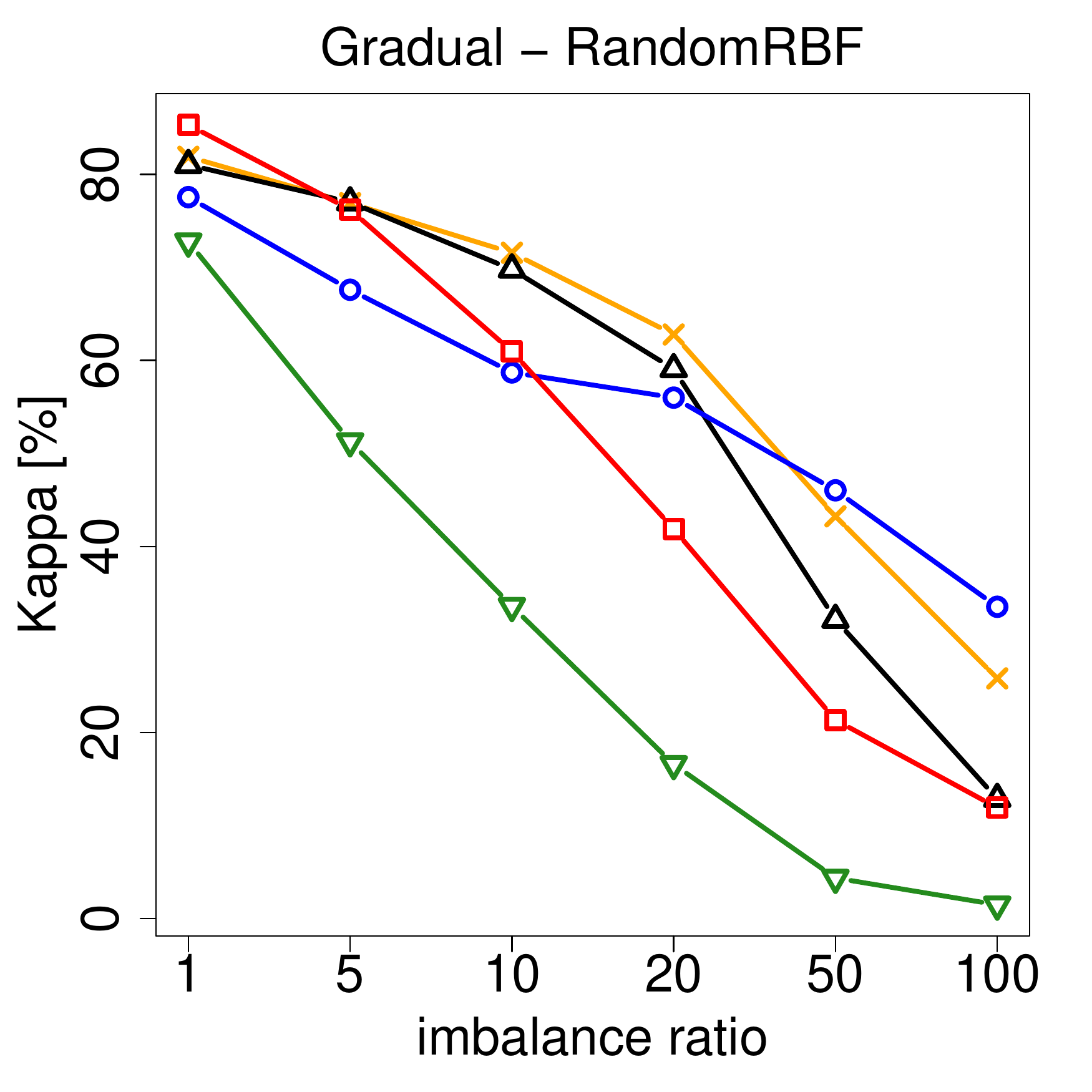}
\includegraphics[width=0.19\columnwidth]{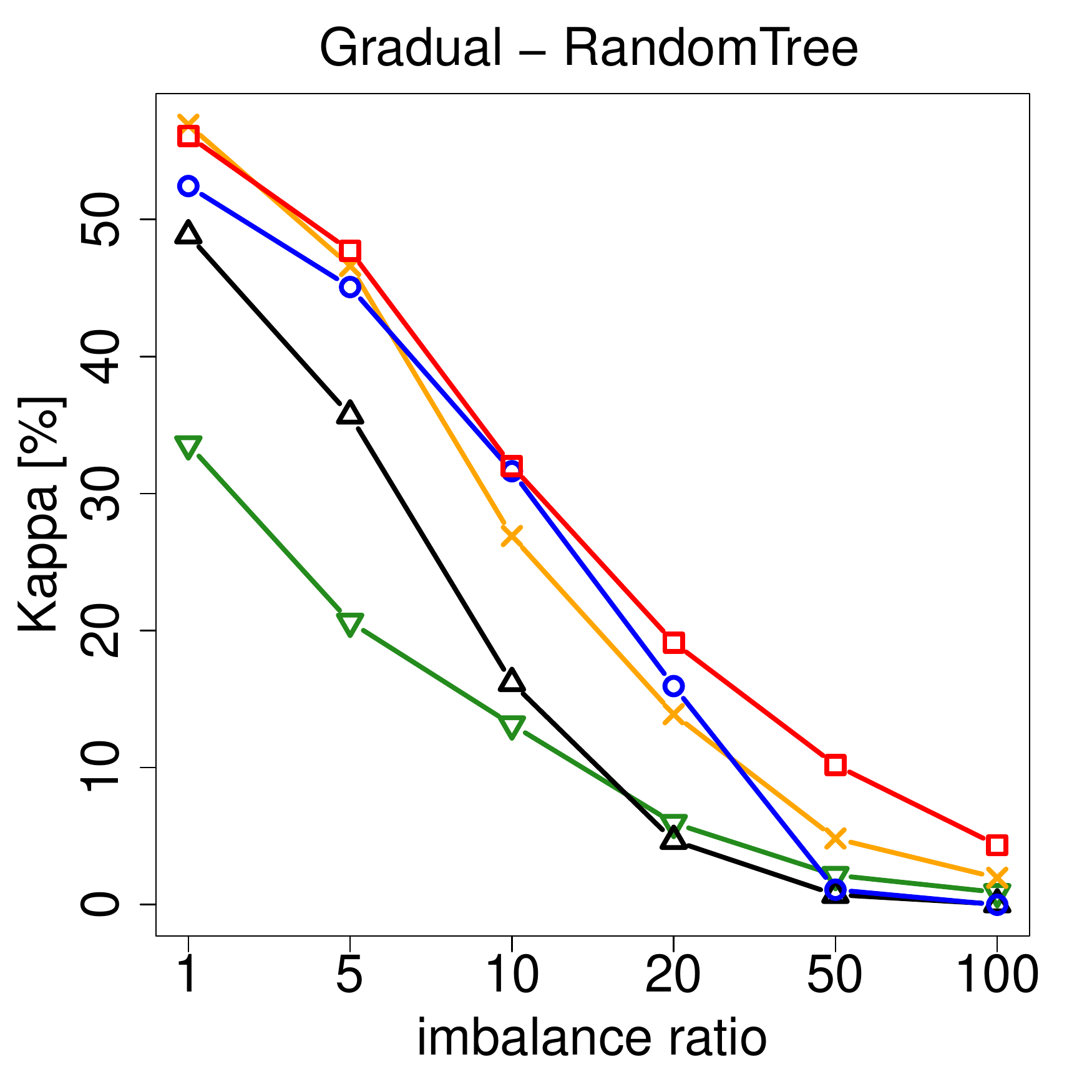}
\includegraphics[width=0.19\columnwidth]{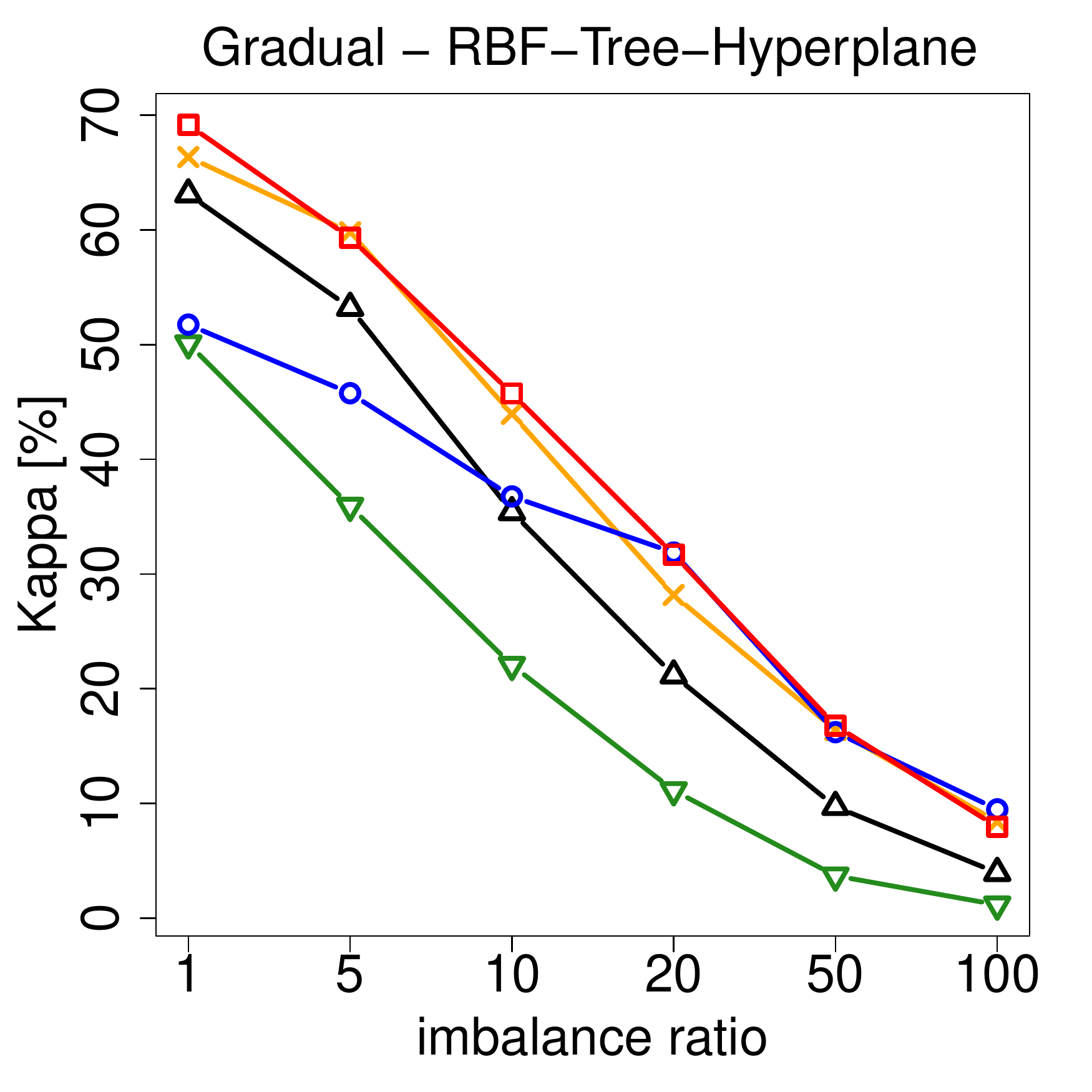}
\includegraphics[width=0.19\columnwidth]{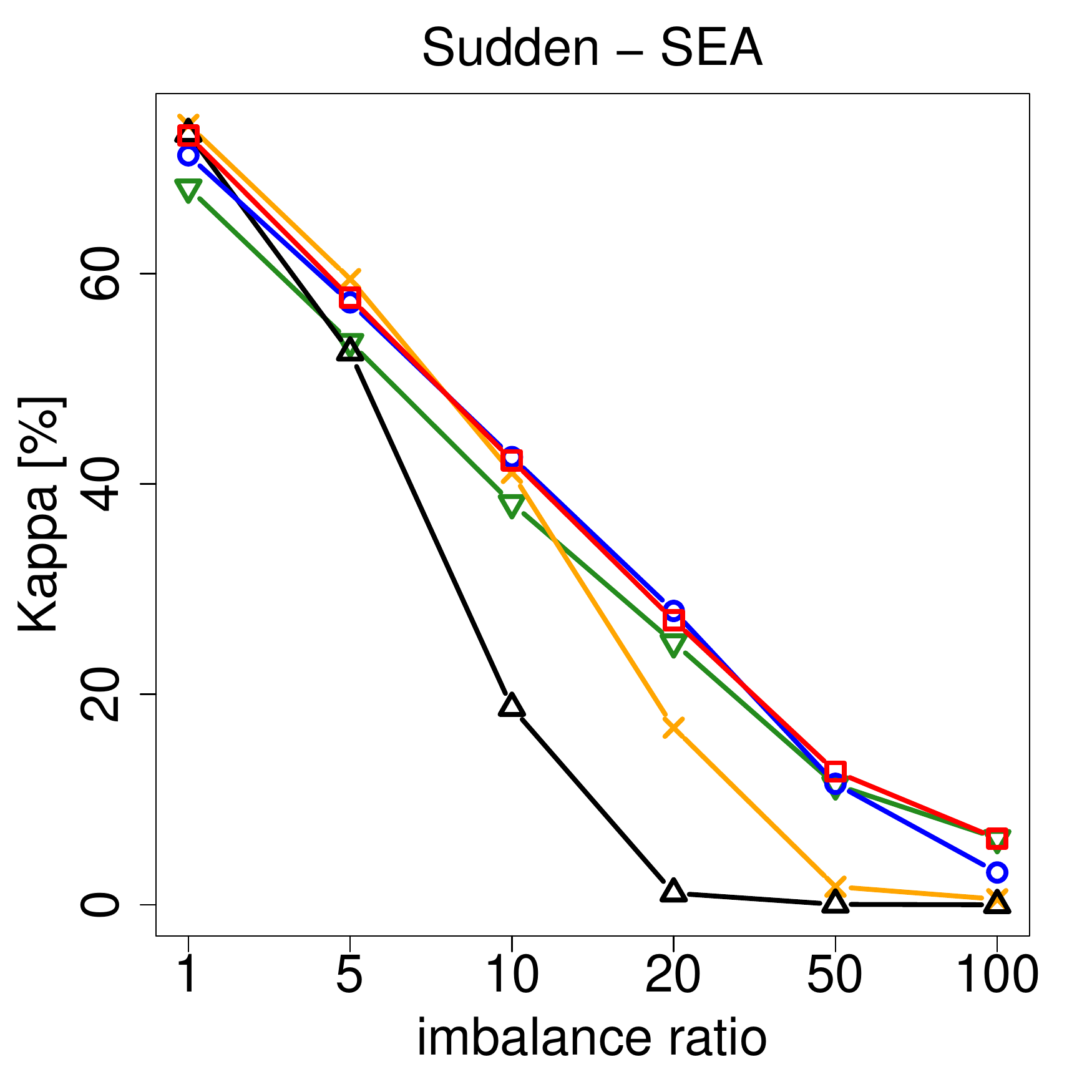}
\includegraphics[width=0.19\columnwidth]{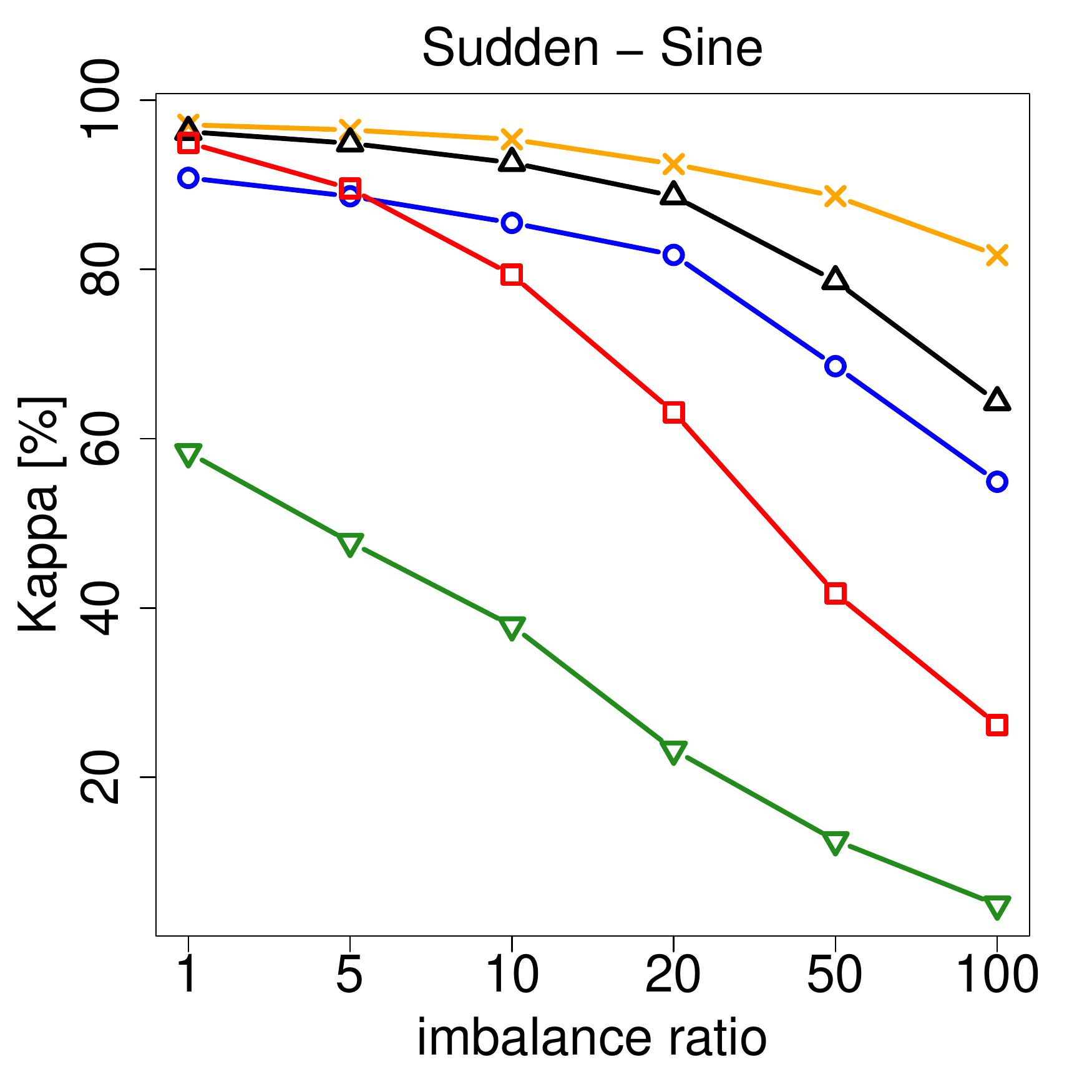}
\caption{Robustness to concept drift with static class imbalance ratio (G-Mean and Kappa).}
\label{fig:cd_static_ir_study}
\end{figure}

\begin{figure}[t!]
\centering
\includegraphics[width=\columnwidth]{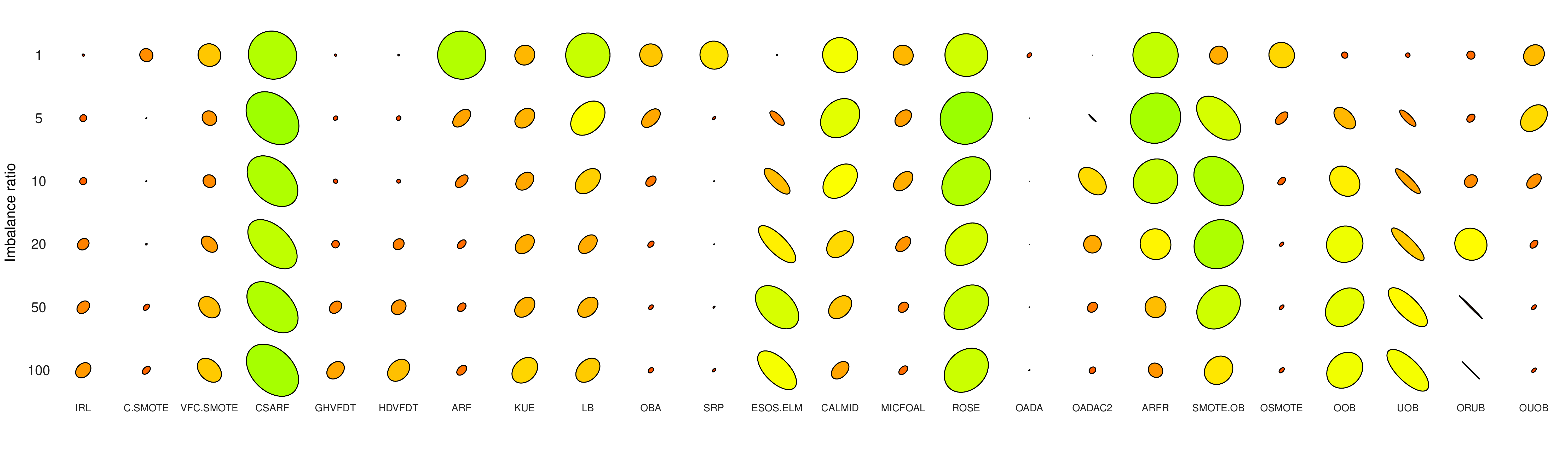}
\caption{Comparison of all 24 algorithms for concept drift with static class imbalance ratio. Axes of the ellipse represent G-Mean and Kappa metrics. Color gradient represents the product of both metrics.}
\label{fig:BC_CD_SIR_ellipse}
\end{figure}

\noindent \textbf{Goal of the experiment.}
This experiment aims to address \textbf{RQ4} and to evaluate the robustness of the data stream classifiers to the static imbalance in the presence of concept drift. Even though the classifiers are designed to deal with imbalance ratios, they also have mechanisms to deal with concept changes. Concept drift affects decision boundaries, thus leading to a more challenging skewed learning scenario with a higher degree of overlap between classes. To evaluate this, we prepared the same generators used in experiment~\ref{sec:bc-static-imb} with two types of concept drift: gradual and sudden. They were combined with the static imbalance ratio examined in experiment~\ref{sec:bc-static-imb}. Figure~\ref{fig:cd_static_ir_study} illustrates the G-Mean and Kappa over time for the five selected classifiers with static imbalance ratio under the presence of concept drift. Table~\ref{tab:BC_CD_SIR} presents the G-Mean and Kappa for the top 10 classifiers and each imbalanced ratio, and their overall ranking as well. Figure~\ref{fig:BC_CD_SIR_ellipse} summarizes the overall performance of all classifiers in this scenario.

\noindent \textbf{Discussion}

\noindent \textit{Impact of approach to class imbalance.} In this experiment, we extend the problem of analyzing the robustness of classifiers to various imbalance ratios by adding concept drift affecting the decision boundaries. It is important to notice that the drift did not influence the disproportion between classes. \acrshort{oob} and \acrshort{uob}, two methods that offered excellent performance for stationary and imbalanced streams suffer from a significant drop in performance when handling non-stationary problems. \acrshort{oob} had the biggest performance drop under concept drift for higher imbalance ratios. This is expected since both methods do not have mechanisms to deal with changes in feature distribution. While the changes in the imbalance ratio could be tackled by resampling approaches, they do not allow for any efficient adaptation to evolving decision boundaries. Classifiers based on informed resampling, such as \acrshort{csmote}, \acrshort{osmote} and \acrshort{vfcsmote} offered only a slightly better performance than the mentioned blind resampling ensembles. This shows that under the presence of concept drift, adaptation mechanisms play a more important role than the solutions used to tackle class imbalance. 
For algorithm-level methods, \acrshort{csarf} demonstrated the best results, thanks to its underlying implicit mechanisms for handling non-stationary data. While, similarly to previous experiments \acrshort{csarf} suffered under Kappa evaluation, this time it was the second-best regarding this metric. \acrshort{rose} remained as the most balanced classifier displaying robustness to changes since it can adapt both to concept drift and imbalance ratio. \acrshort{arf}, \acrshort{arfr}, \acrshort{lb} and \acrshort{srp} achieved decent results in a scenario with concept drift, however, their performance drops as the imbalance ratio increases. 

\begin{table*}[t!]
\centering
\footnotesize
\setlength{\tabcolsep}{4pt}
\caption{G-Mean and Kappa averages of all 10 streams for concept drift with static class imbalance ratio.}
\label{tab:BC_CD_SIR}
\begin{tabular}{ll|C{1cm}C{1cm}C{1cm}C{1cm}C{1cm}C{1cm}C{1cm}C{1cm}C{1cm}C{1cm}}
\toprule
& IR & CSARF & ARF & KUE & LB & CALMID & ROSE & ARFR & SMOTE-OB & OOB & UOB\\
\midrule
\multirow{6}{*}{\rotatebox[origin=c]{90}{G-Mean}}
& 1 & 87.83 & 87.82 & 85.44 & 87.99 & 86.31 & \textbf{88.70} & 87.71 & 83.74 & 78.80 & 77.90\\
& 5 & \textbf{86.62} & 67.56 & 71.74 & 75.41 & 77.08 & 84.74 & 83.96 & 84.60 & 76.62 & 76.35\\
& 10 & \textbf{84.64} & 46.23 & 52.48 & 54.02 & 61.48 & 75.57 & 75.44 & 80.31 & 71.15 & 74.96\\
& 20 & \textbf{81.06} & 31.49 & 38.67 & 38.94 & 41.65 & 59.10 & 57.25 & 70.20 & 60.33 & 72.97\\
& 50 & \textbf{73.97} & 15.43 & 19.40 & 20.63 & 23.71 & 37.90 & 28.59 & 42.82 & 36.64 & 68.99\\
& 100 & \textbf{64.86} & 9.93 & 12.11 & 13.16 & 11.71 & 21.80 & 9.53 & 21.54 & 24.81 & 63.78\\
\midrule
\multirow{6}{*}{\rotatebox[origin=c]{90}{Kappa}} 
& 1 & 75.76 & 75.76 & 71.67 & 76.10 & 72.73 & \textbf{77.52} & 75.59 & 69.71 & 57.93 & 56.23\\
& 5 & 66.58 & 56.51 & 58.24 & 64.02 & 64.12 & \textbf{70.99} & 68.15 & 62.24 & 52.71 & 45.30\\
& 10 & 53.18 & 39.00 & 42.57 & 45.61 & 50.06 & \textbf{60.09} & 54.30 & 53.17 & 45.58 & 33.72\\
& 20 & 37.11 & 26.41 & 30.96 & 32.65 & 34.34 & \textbf{45.03} & 33.83 & 42.97 & 36.70 & 22.06\\
& 50 & 19.26 & 12.69 & 15.89 & 17.31 & 19.04 & \textbf{28.36} & 13.39 & 26.54 & 22.33 & 10.14\\
& 100 & 10.54 & 8.65 & 10.42 & 11.70 & 10.15 & \textbf{17.25} & 2.82 & 15.11 & 15.79 & 4.52\\
\midrule
\multicolumn{2}{l|}{Avg. G-Mean} & \textbf{79.83} & 43.08 & 46.64 & 48.36 & 50.32 & 61.30 & 57.08 & 63.87 & 58.06 & 72.49\\
\multicolumn{2}{l|}{Avg. Kappa} & 43.74 & 36.50 & 38.29 & 41.23 & 41.74 & \textbf{49.87} & 41.35 & 44.96 & 38.51 & 28.66\\
\midrule
\multicolumn{2}{l|}{Rank G-Mean} & \textbf{1.83} & 7.95 & 7.54 & 6.69 & 6.50 & 4.21 & 4.93 & 4.27 & 5.99 & 5.08\\
\multicolumn{2}{l|}{Rank Kappa} & 3.95 & 7.05 & 6.68 & 5.50 & 5.40 & \textbf{2.92} & 4.88 & 4.72 & 5.93 & 7.97\\
\bottomrule
\end{tabular}
\end{table*}

\noindent \textit{Impact of ensembles architecture.} As observed in the previous experiments, boosting-based methods deliver the worst performance among all ensemble architectures. This can be explained by drift destabilizing boosting classifier chains, as errors made by previous classifiers may no longer be meaningful for the updating of their follow-ups. There is a need to improve drift adaptation procedures for boosting-based ensembles so they can become competitive with their bagging peers. While bagging-based architectures are still the core of the best-performing methods, we can see the increasing dominance of hybrid architectures for concept drift scenarios. While all of them use bagging, they combine it with the dynamic weighting of the base classifier and dynamic line-up, demonstrating that a combination of several mechanisms is necessary to tackle class imbalance and concept drift at the same time.

\noindent \textit{Relationship between concept drift and imbalance ratios.} In the context of this experiment, it is crucial to analyze and understand the interplay between the concept drift impacting the class boundaries and static imbalance ratios affecting the disproportion between them. While focusing on how the classifiers try to tackle concept drift, we do not see significant differences between the ones utilizing implicit or explicit drift detection. This shows that there is no obvious choice for adaptation mechanisms and that the classifier performance for drifting and imbalance streams is a product of their learning architecture, drift adaptation mechanism, and approach to tackling class imbalance. We can see that popular classifier for drifting data streams, such as \acrshort{arf}, \acrshort{lb}, or \acrshort{srp} cannot handle increasing imbalance ratios. At the same time, solutions dedicated to online learning from imbalanced data streams, such as \acrshort{uob} or \acrshort{oob} cannot deal with the non-stationary nature of data streams. Best performing methods, such as \acrshort{rose}, \acrshort{csarf} and \acrshort{smoteob} combine adaptation and skew-insensitive mechanisms for all-round robustness. 

\subsubsection{Concept drift and dynamic imbalance ratio}
\label{sec:bc-cd-dynamic-ratio}

\begin{figure}[t!]
\centering
\includegraphics[width=0.19\columnwidth]{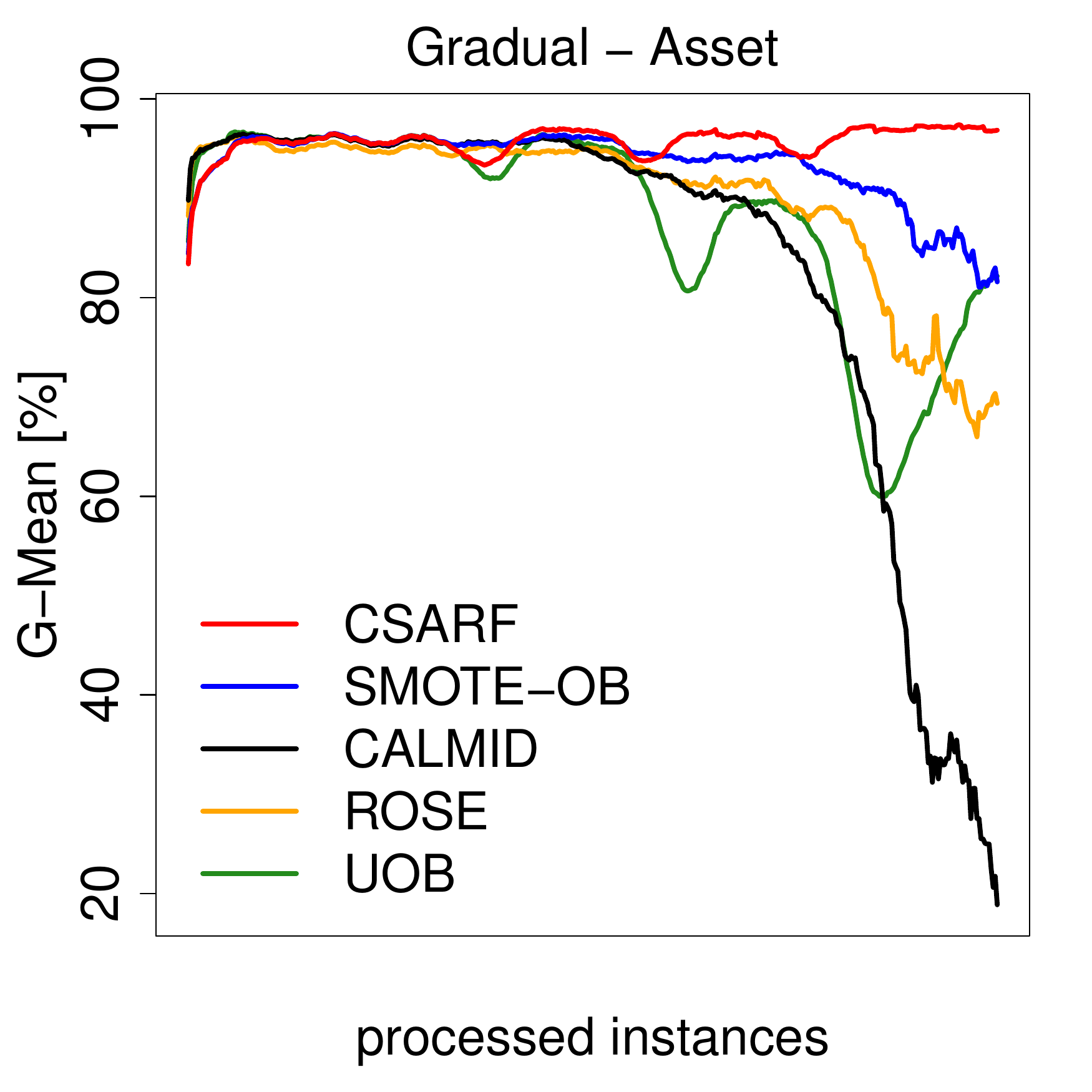}
\includegraphics[width=0.19\columnwidth]{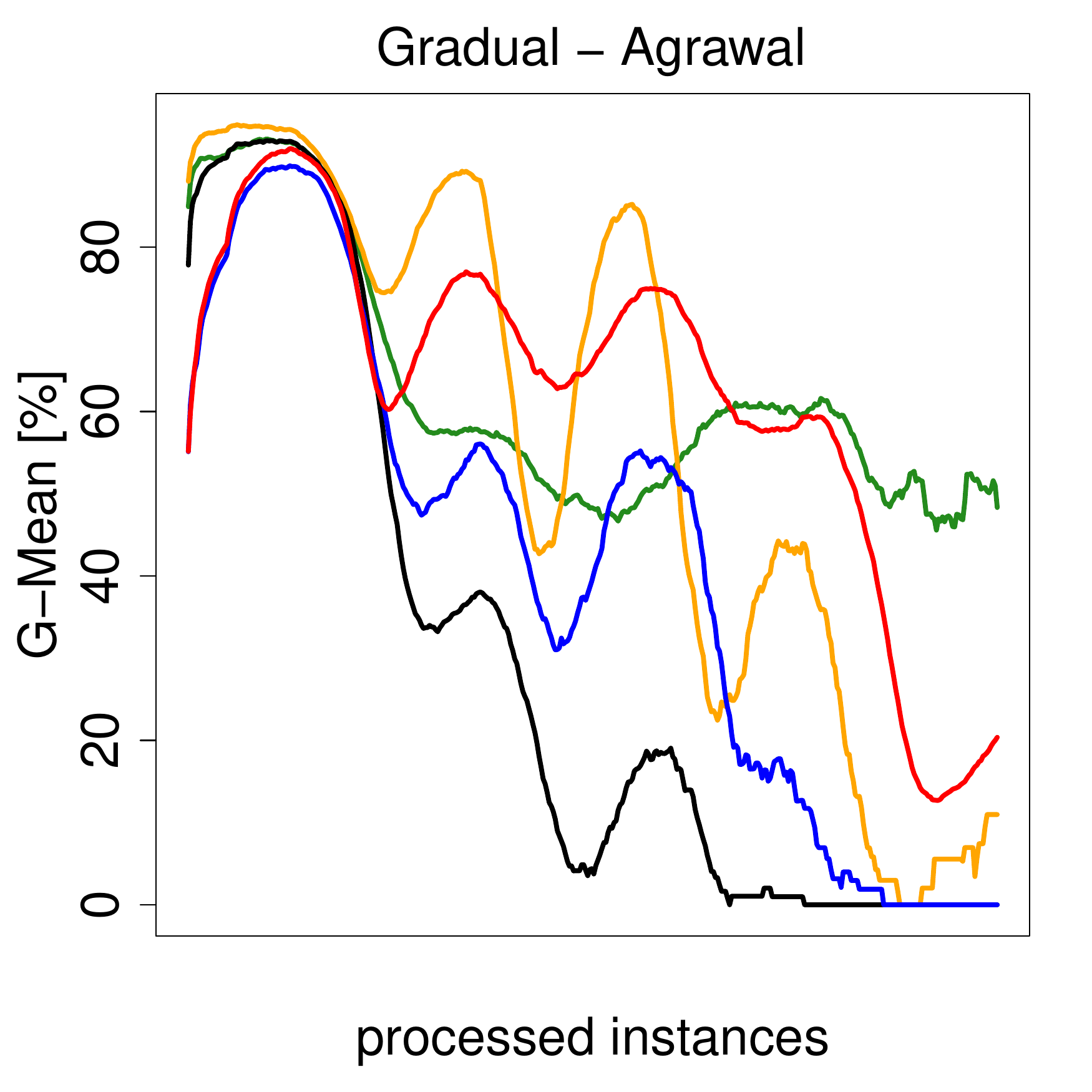}
\includegraphics[width=0.19\columnwidth]{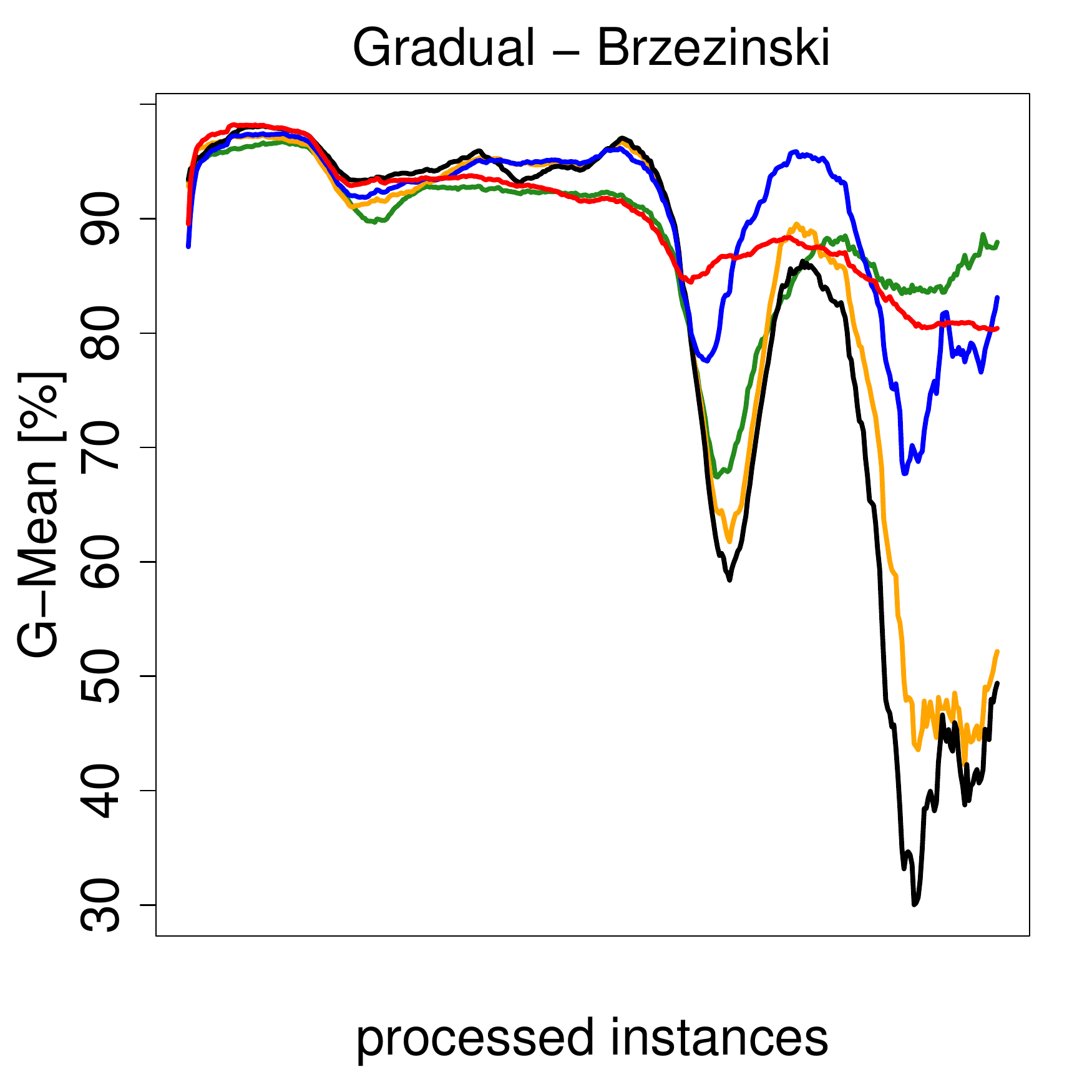}
\includegraphics[width=0.19\columnwidth]{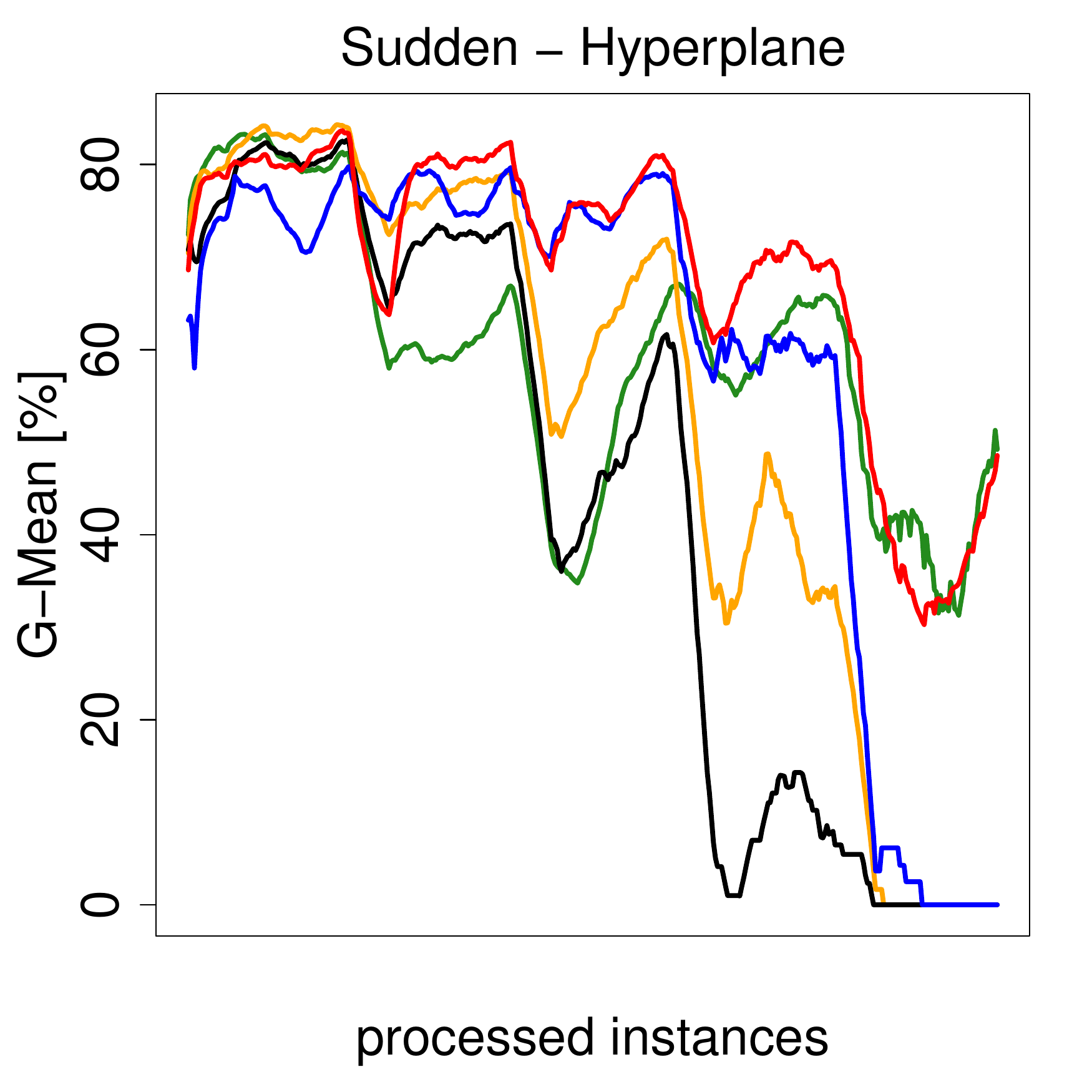}
\includegraphics[width=0.19\columnwidth]{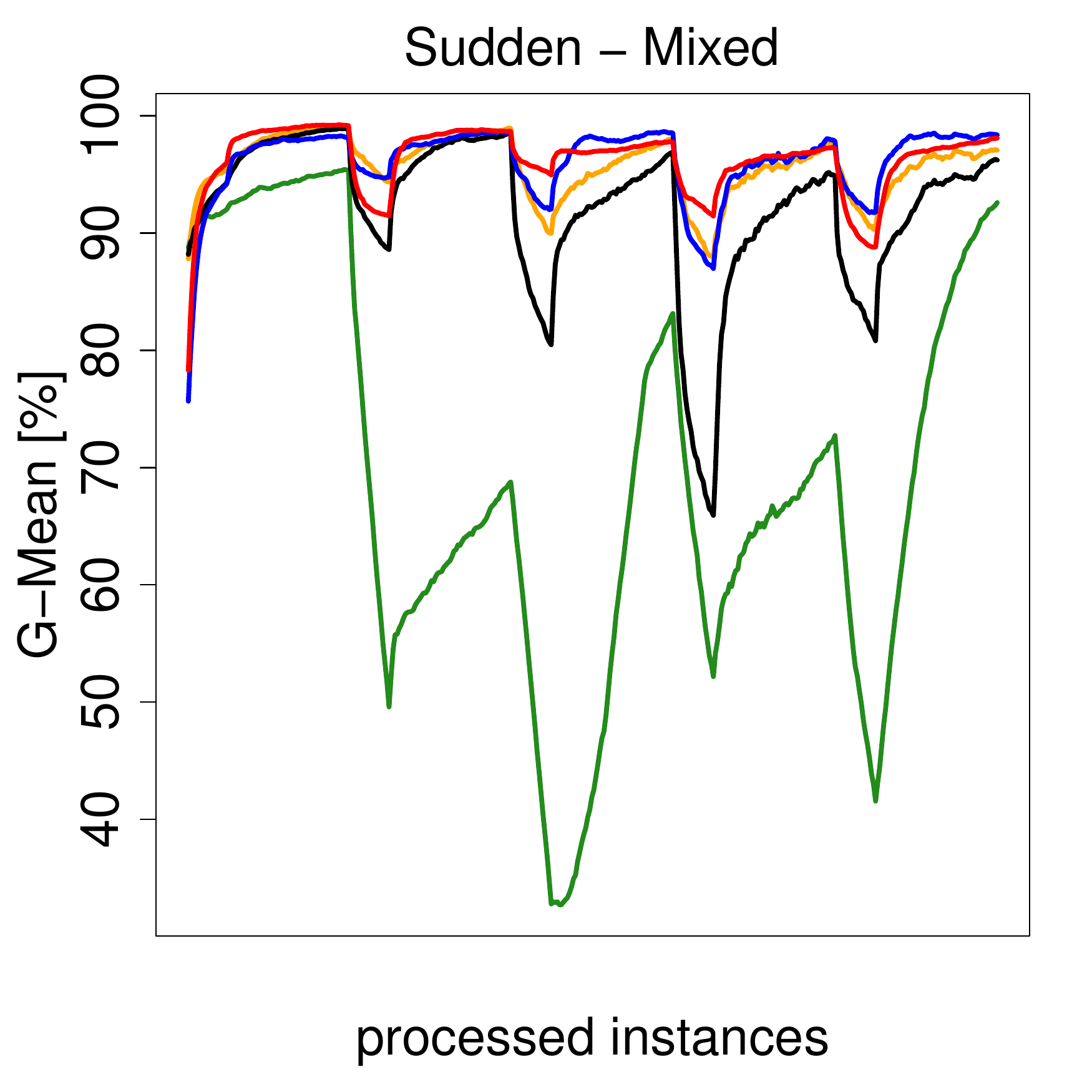}
\includegraphics[width=0.19\columnwidth]{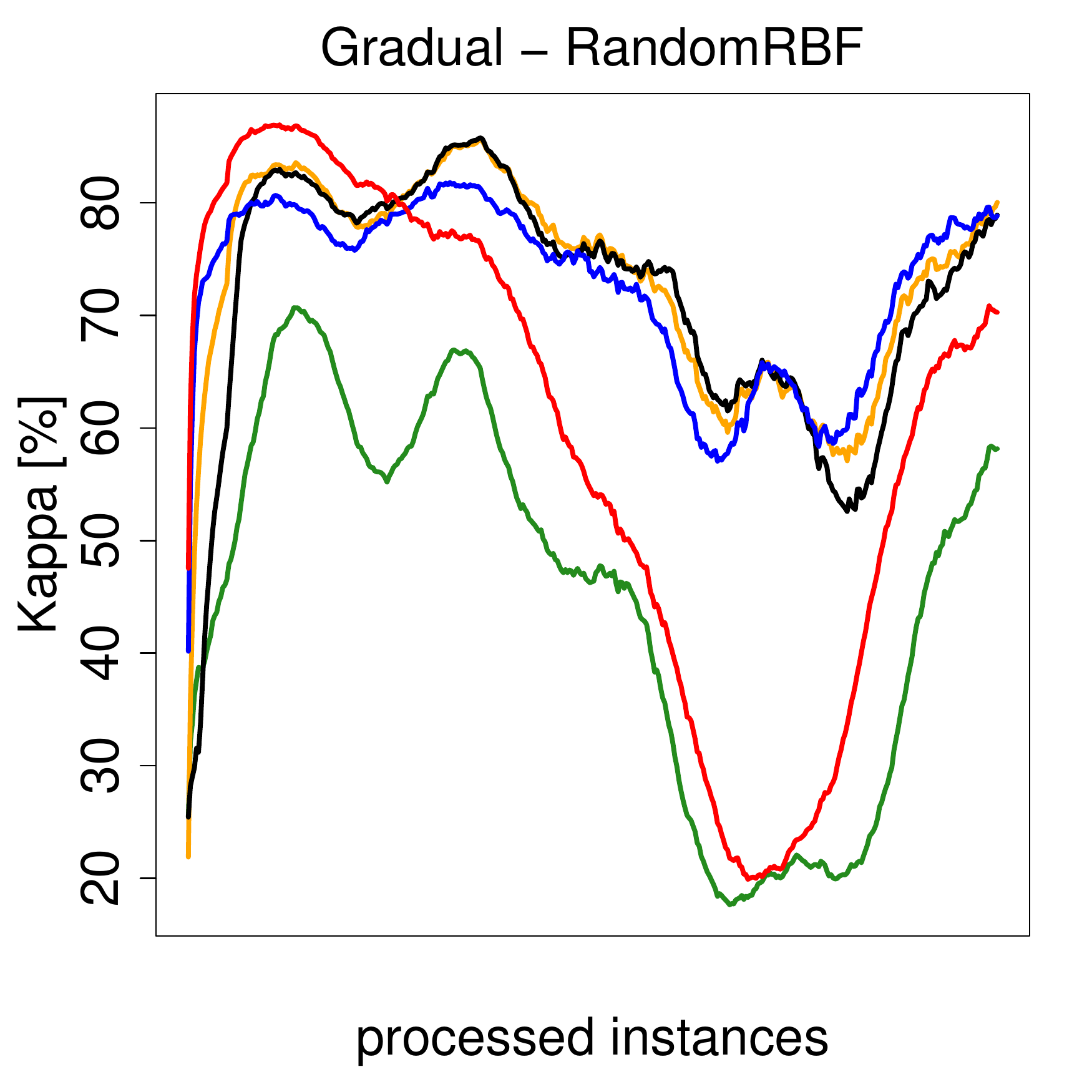}
\includegraphics[width=0.19\columnwidth]{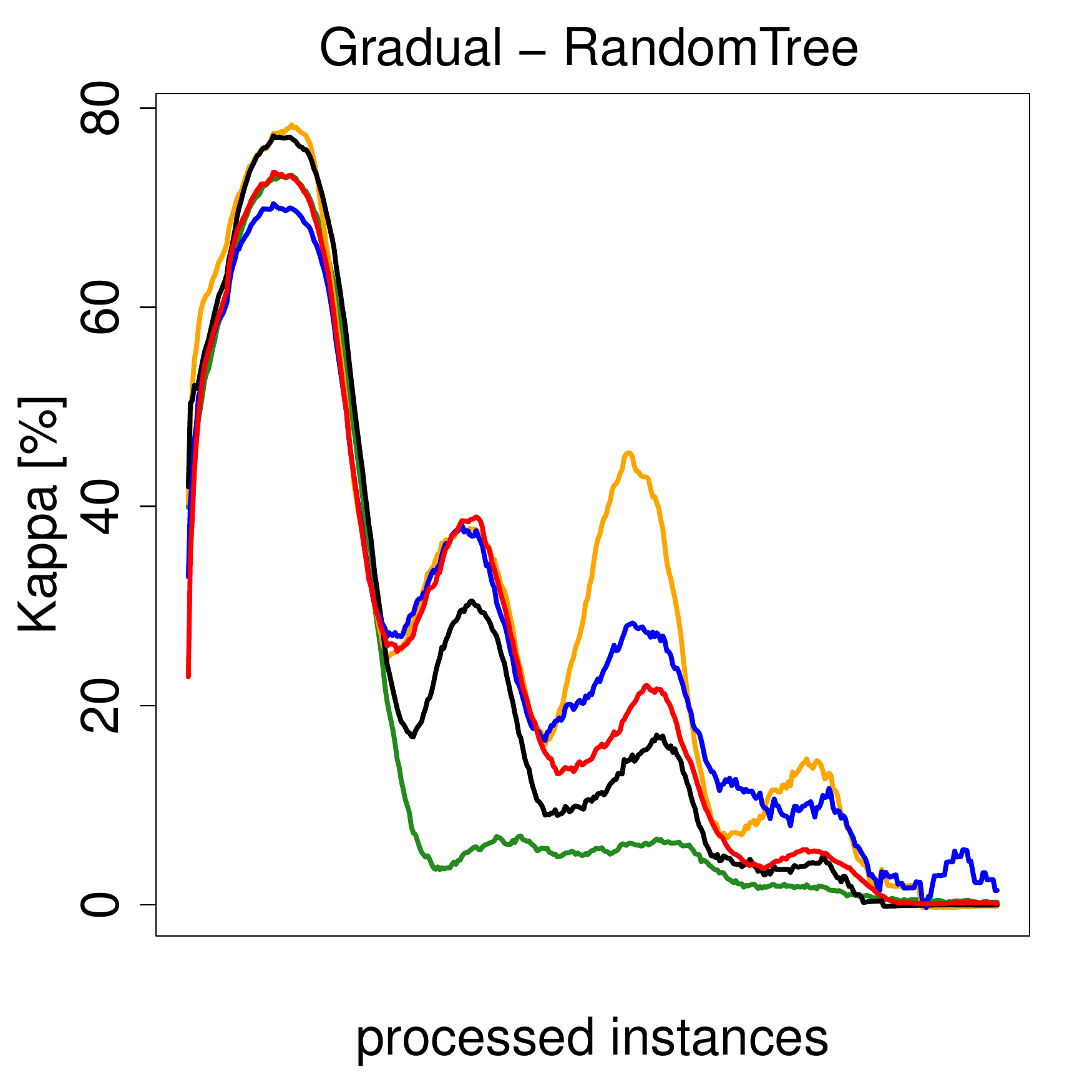}
\includegraphics[width=0.19\columnwidth]{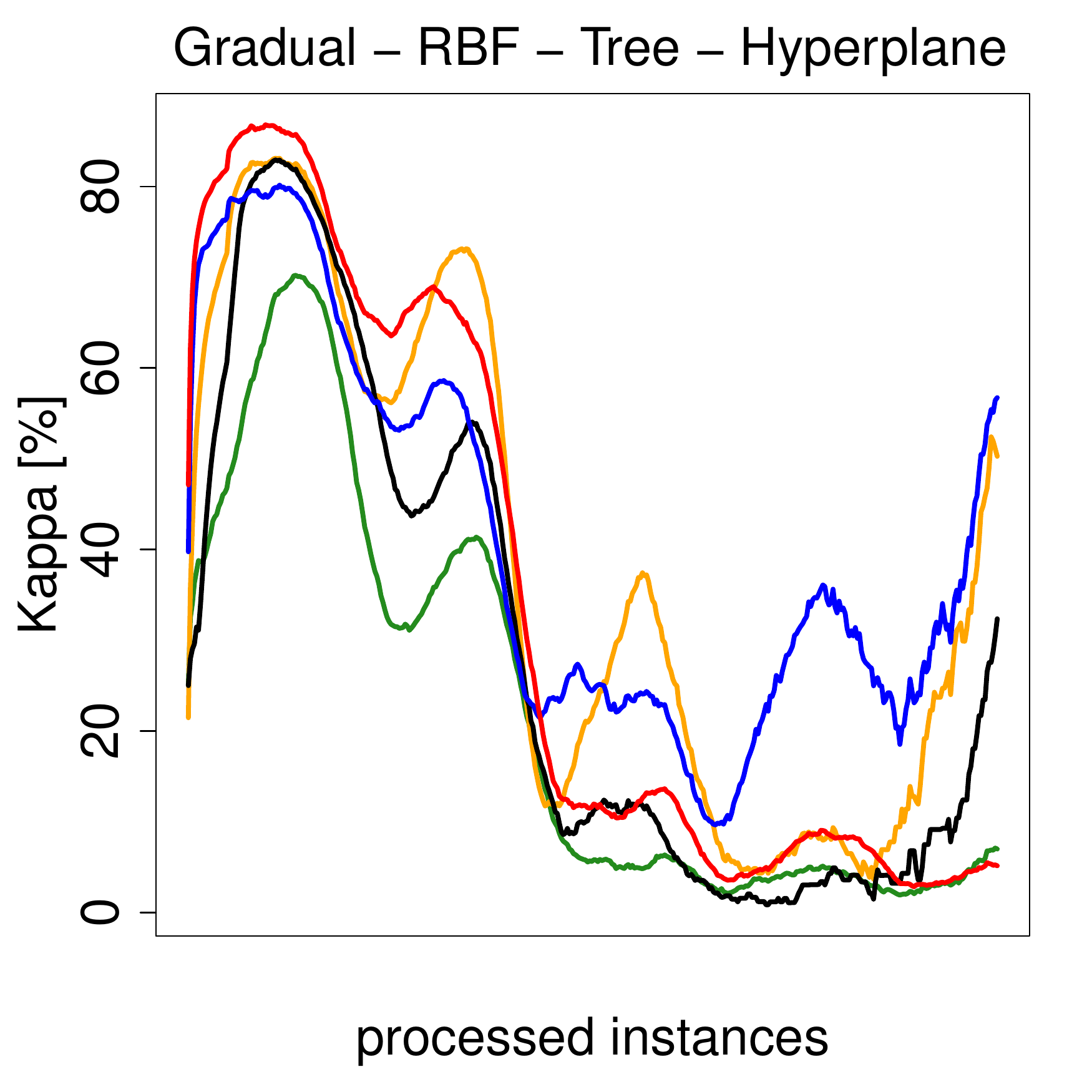}
\includegraphics[width=0.19\columnwidth]{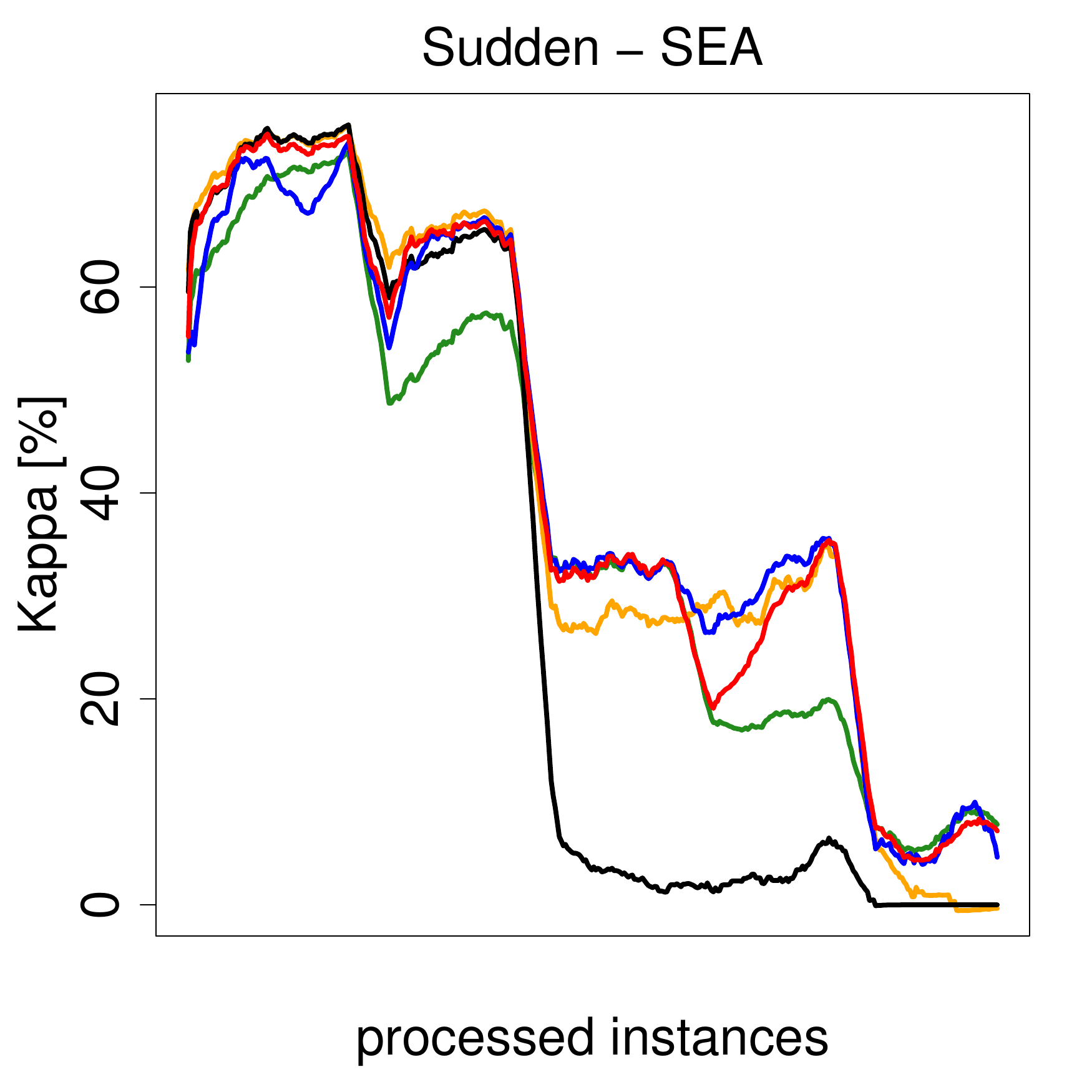}
\includegraphics[width=0.19\columnwidth]{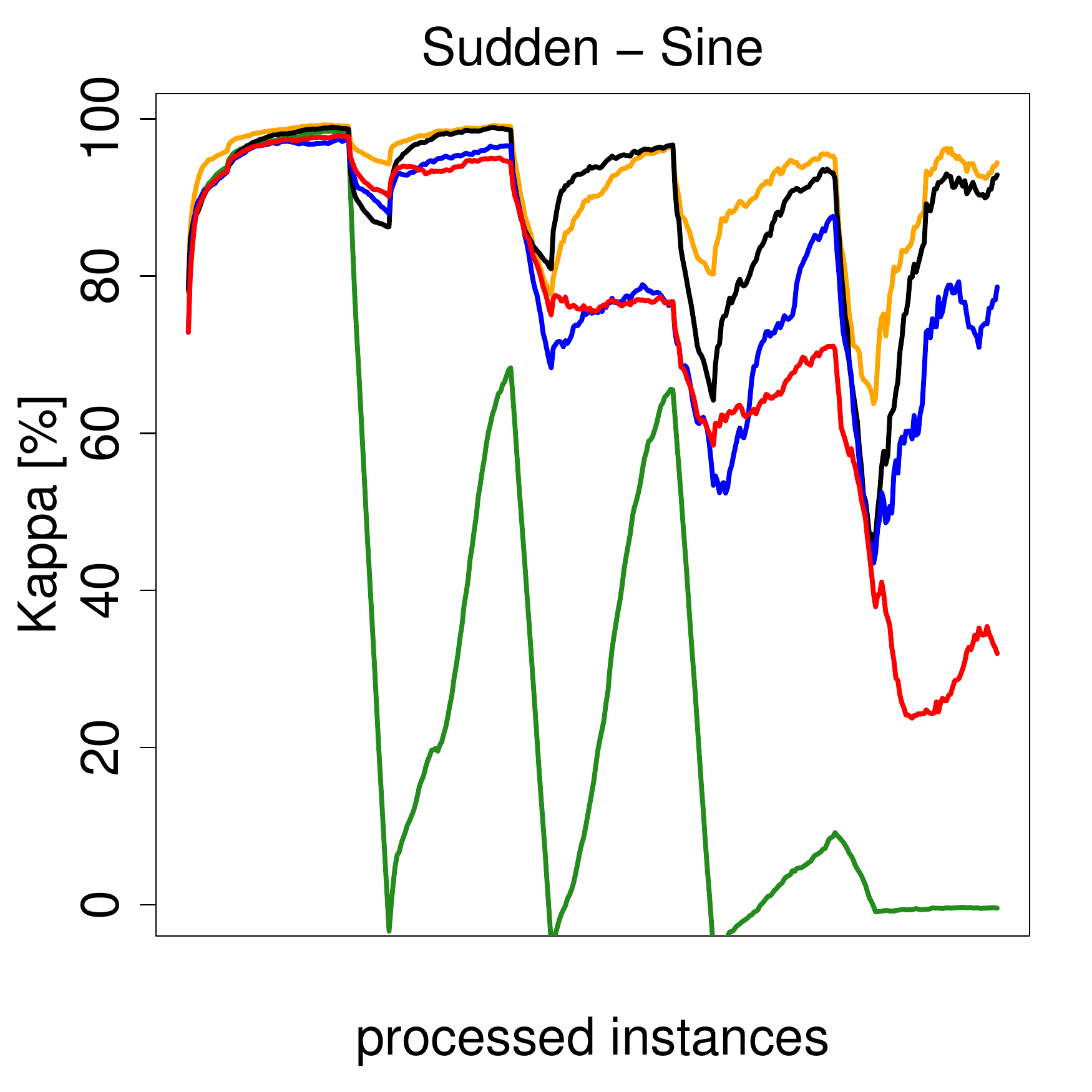}
\caption{G-Mean and Kappa on concept drift with increasing imbalance ratio.}
\label{fig:cd_increasing_ir_study}
\end{figure}

\begin{figure}[t!]
\centering
\includegraphics[width=0.5\columnwidth]{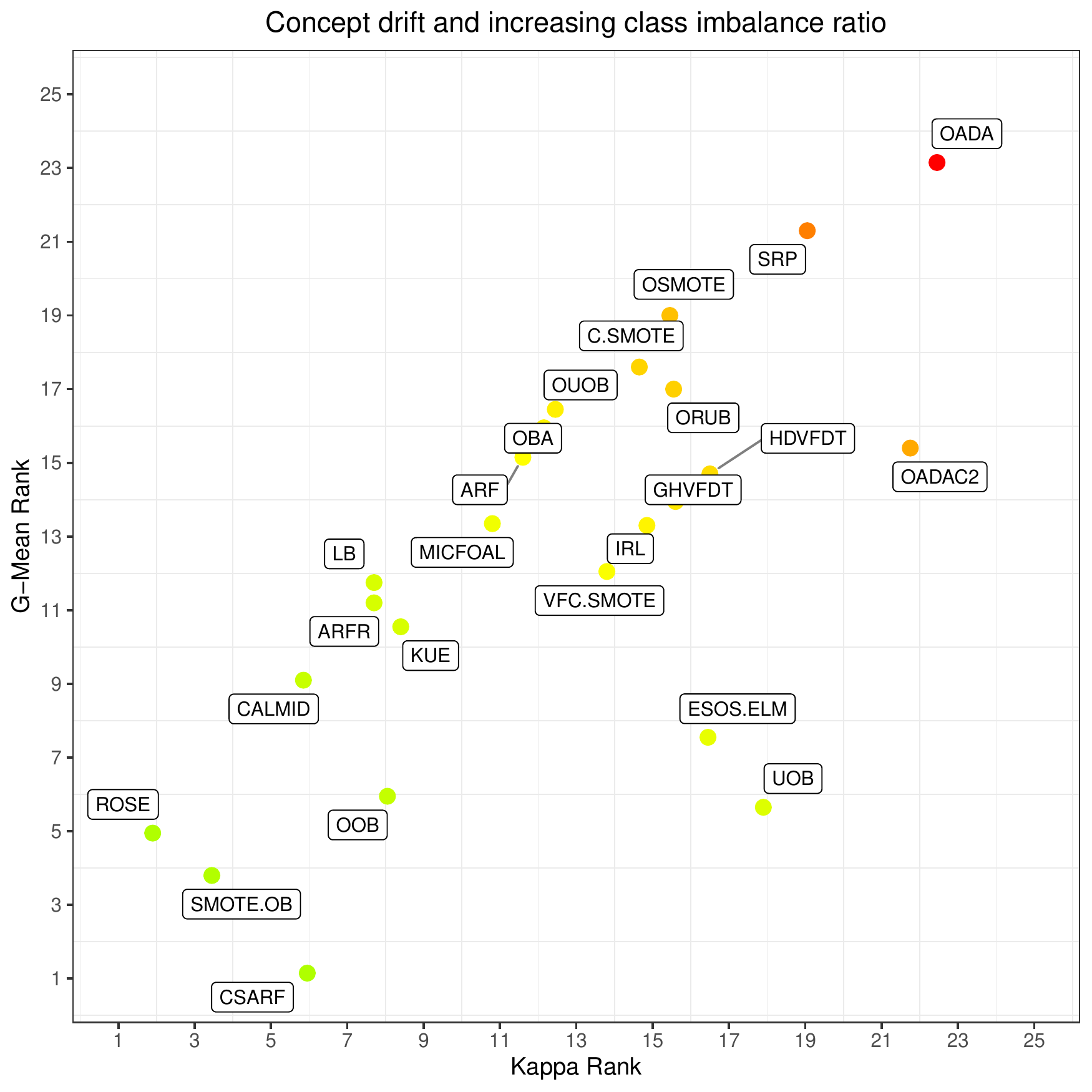}
\caption{Comparison of all 24 algorithms for concept drift with increasing class imbalance ratio. Color gradient represents the product of both metrics.}
\label{fig:BC_CD_IIR_scatter}
\end{figure}

\noindent \textbf{Goal of the experiment.} This experiment was designed to complement the previous experiment, and completely address \textbf{RQ2} and \textbf{RQ4}, examining the classifiers in the presence of concept drift combined with dynamic imbalance ratio. Combining concept drift at the same time with changes in the class imbalance poses a complex challenge to classifiers. To evaluate this, we prepared the same generators in experiment \ref{sec:bc-cd-static-ratio} with gradual and sudden concept drift, and combined them with the dynamic increasing imbalance ratio proposed in experiment~\ref{sec:bc-dyn-imb}. Figure~\ref{fig:cd_increasing_ir_study} illustrates the performance of the selected classifiers with dynamic increasing imbalance ratio under the presence of concept drift. Table~\ref{tab:BC_CD_IIR} presents the G-Mean and Kappa for the top 10 classifiers for each type of concept drift and the average ranking for each evaluated metric. Figure~\ref{fig:BC_CD_IIR_scatter} provides an overall comparison of all classifiers in the proposed scenario.

\noindent \textbf{Discussion}

\noindent \textit{Impact of approach to class imbalance.} Let us focus on changes in the behavior of classifier as compared with the previous case of evolving class imbalance without explicit concept drift. All methods based on blind resampling display drops in performance, as usually they lack explicit mechanisms for handling concept drift, leading to their deterioration over time. \acrshort{smote} based methods followed the behavior experienced in previous experiments, mainly because concept drift and increasing imbalance ratio may lead to temporal incoherence which can enhance problems of oversampling. Only \acrshort{smoteob} displayed satisfactory robustness to simultaneously evolving imbalance ratio and concept drift, while additionally achieving good balance between Kappa and G-Mean metrics.

Classifiers based on training modifications, such as \acrshort{rose} and \acrshort{calmid} displayed robustness to concept drift and dynamic imbalance ratio, especially for the G-mean metric. Their training procedures provide reliability in a scenario where multiple changes happen simultaneously. The cost-sensitive approach of \acrshort{csarf} presents outstanding results regarding the G-Mean metric, with almost $1$ as the average rank. Nevertheless, when analyzing the Kappa metric, we can see shortcomings of \acrshort{csarf}, where it ranks the third. This shows that the \acrshort{csarf} adaptation to evolving data characteristics is not balanced over both classes. \acrshort{rose} displays balanced performance on both metrics, which can be explained by the combination of concept drift detector with balanced buffers for each of classes, allowing for equalized performance on the majority and minority classes.

\noindent \textit{Impact of ensemble architecture.} This highly difficult scenario further shows that bagging-based and hybrid architectures are the only ones capable of handling drifting and evolving class imbalance. Their superiority over boosting methods becomes even more evident in these experiments. However, another interesting observation is the increasing gap between dynamic and static ensemble line-ups. Here we can see that most of the best performing methods use dynamic replacement of the ensemble members. This can be explained by the fact that when concept drift is combined with evolving class imbalance, especially under rapid changes, it is more efficient to train a new classifier from scratch and replace the weakest one, instead of trying to adapt the existing members to a vastly different new concept. 

\begin{table*}[t!]
\centering
\footnotesize
\setlength{\tabcolsep}{4pt}
\caption{G-Mean and Kappa averages of all 10 streams for concept drift with increasing class imbalance ratio.}
\label{tab:BC_CD_IIR}
\begin{tabular}{ll|C{1cm}C{1cm}C{1cm}C{1cm}C{1cm}C{1cm}C{1cm}C{1cm}C{1cm}C{1cm}}
\toprule
& Drift & CSARF & ARF & KUE & LB & CALMID & ROSE & ARFR & SMOTE-OB & OOB & UOB\\
\midrule
\multirow{2}{*}{\rotatebox[origin=c]{90}{\scalebox{.65}{G-Mean}}}
& Sudden & \textbf{85.96} & 55.56 & 62.45 & 62.61 & 64.19 & 74.05 & 60.64 & 75.79 & 72.91 & 71.00\\
& Gradual & \textbf{77.90} & 45.73 & 53.47 & 48.12 & 53.78 & 64.60 & 50.70 & 70.29 & 66.23 & 70.32\\
\midrule
\multirow{2}{*}{\rotatebox[origin=c]{90}{\scalebox{.65}{Kappa}}}
& Sudden & 55.13 & 48.88 & 52.29 & 55.39 & 55.50 & \textbf{63.23} & 52.70 & 57.48 & 48.48 & 33.09\\
& Gradual & 41.57 & 36.44 & 40.59 & 39.22 & 42.33 & \textbf{48.45} & 39.77 & 45.67 & 39.12 & 29.69\\
\midrule
\multicolumn{2}{l|}{Avg. G-Mean} & \textbf{81.93} & 50.64 & 57.96 & 55.36 & 58.98 & 69.33 & 55.67 & 73.04 & 69.57 & 70.66\\
\multicolumn{2}{l|}{Avg. Kappa} & 48.35 & 42.66 & 46.44 & 47.31 & 48.91 & \textbf{55.84} & 46.23 & 51.57 & 43.80 & 31.39\\
\midrule
\multicolumn{2}{l|}{Rank G-Mean} & \textbf{1.10} & 9.30 & 7.20 & 7.70 & 6.45 & 4.10 & 7.35 & 3.30 & 4.35 & 4.15\\
\multicolumn{2}{l|}{Rank Kappa} & 4.75 & 8.00 & 6.35 & 5.90 & 4.70 & \textbf{1.75} & 5.65 & 3.10 & 5.60 & 9.20\\
\bottomrule
\end{tabular}
\end{table*}

\noindent \textit{Impact of concept drift speed.} As mentioned in the previous observation, the speed of changes (velocity of concept drift) significantly impacts the classifiers. We observed that all of the classifiers tend to react worse to gradual drift, while displaying better robustness on sudden drift. While this observation can be surprising, we can explain it by taking a deeper look on how the adaptation mechanisms work in these ensembles. Under sudden concept drift, we cn observe a rapid deterioration of the ensemble performance. However, new instances coming from a stable concept are readily available, allowing for a recovery and adaptation with sufficient sample size. When dealing with gradual drift, classifiers do not see the new, fully formed concept so quickly. Therefore, the adaptation process becomes more tedious, as the sample size from the new concept may not be big enough. This may mislead some pruning or weighting mechanisms, forcing costly false adaptations. While in case of gradual drift we do not observe one single drop of performance, the negative impact of change is prolonged over time and thus may sum up to a bigger challenge for the classifier in the long run. 

\noindent \textit{Relationship between concept drift and increasing class imbalance.} In this scenario each classifier must be able to simultaneously handle concept drift (impacting the decision boundaries) and evolving class imbalance ratio (impacting both skew-insensitive mechanisms and decision boundaries). This creates a trade-off, with classifiers displaying different behavior patterns. Some methods, like \acrshort{lb} or \acrshort{kue} display high adaptability to concept drift. Others, like \acrshort{oob} focus on robustness to evolving class imbalance. The most balanced method, offering best trade-off between those two factors is \acrshort{rose}, followed by \acrshort{smoteob} and \acrshort{csarf}.

\vspace*{-0.25cm}
\subsubsection{Real-world binary class imbalanced datasets}
\vspace*{-0.2cm}
\label{sec:bc-rl-wd}

\begin{figure}[t!]
\centering
\includegraphics[width=0.19\columnwidth]{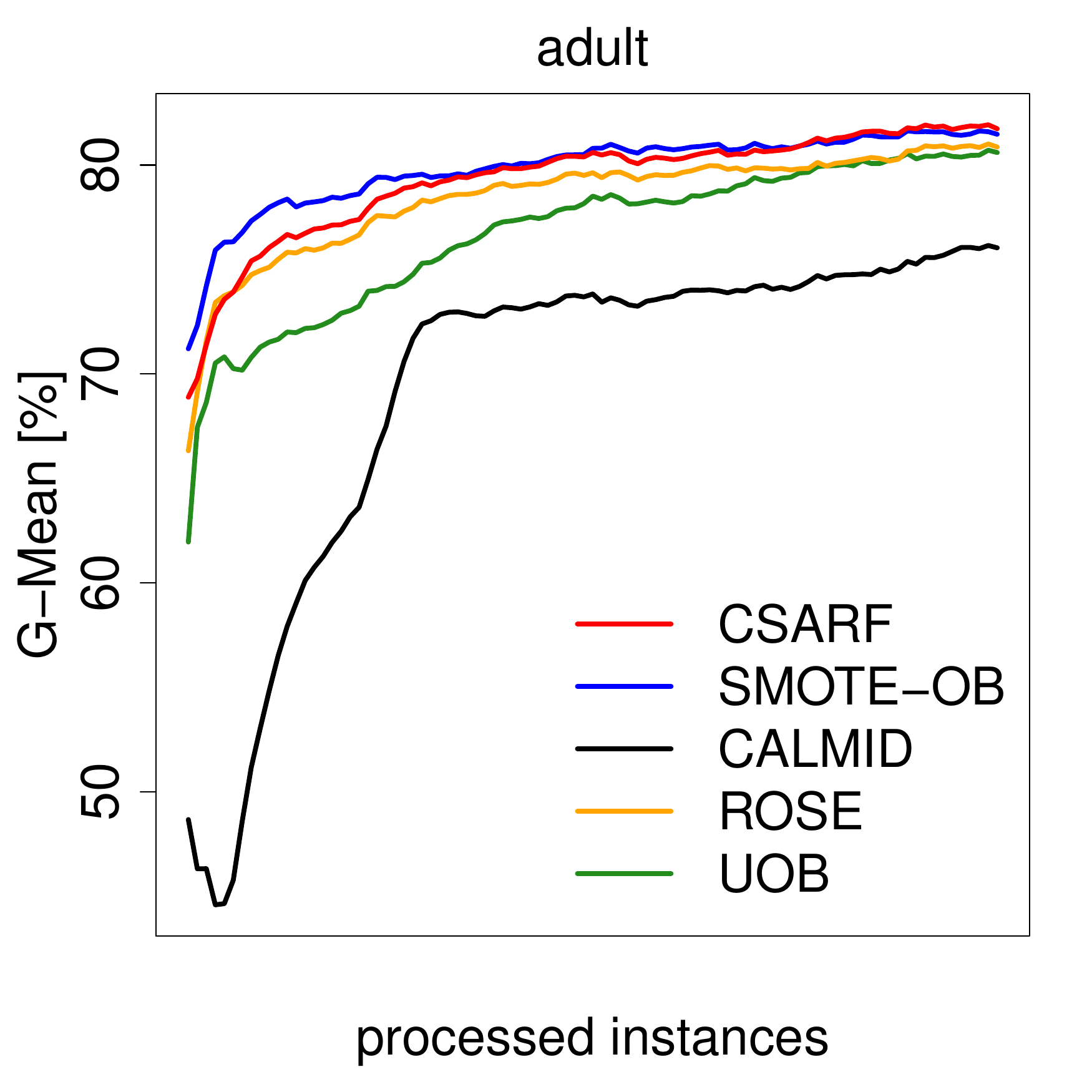}
\includegraphics[width=0.19\columnwidth]{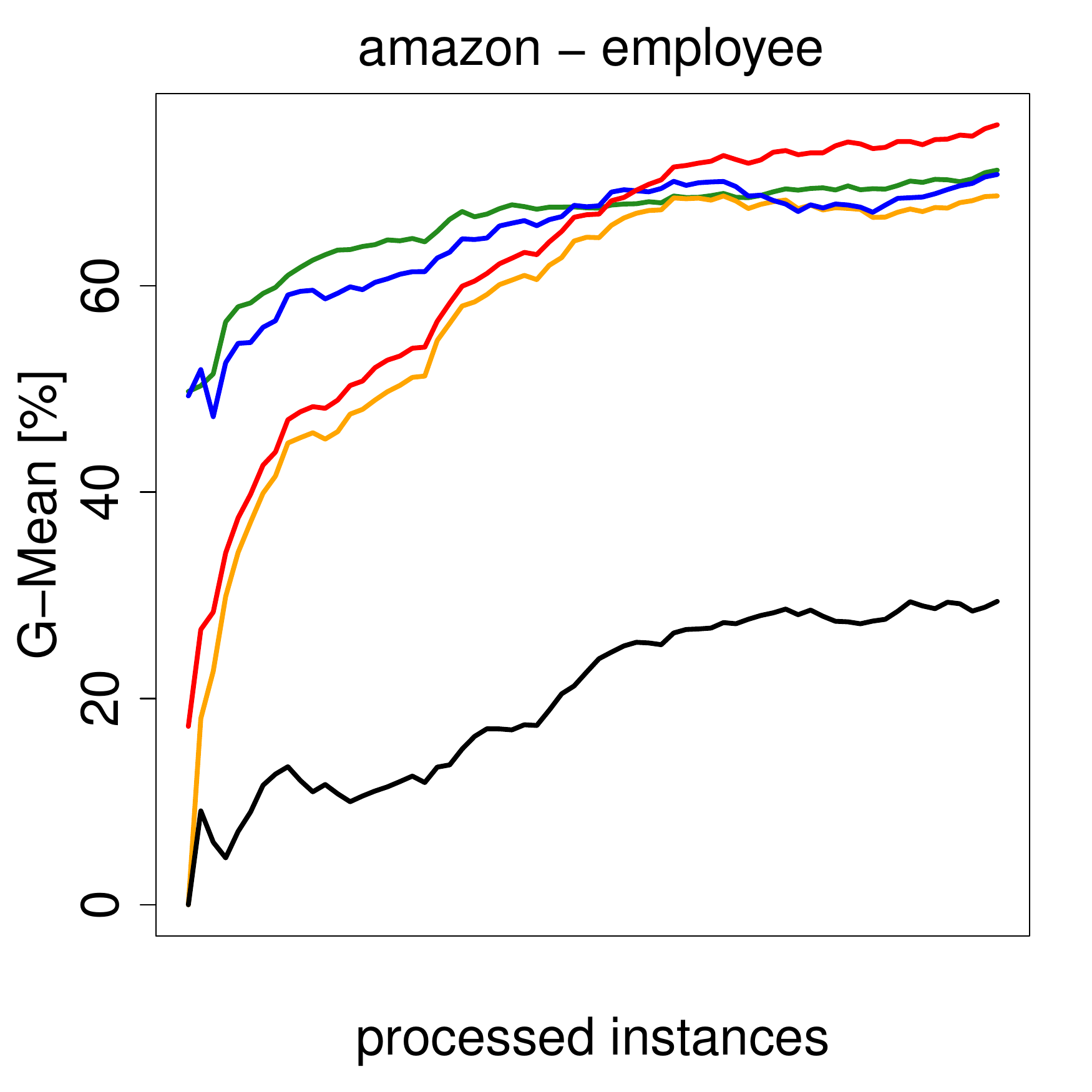}
\includegraphics[width=0.19\columnwidth]{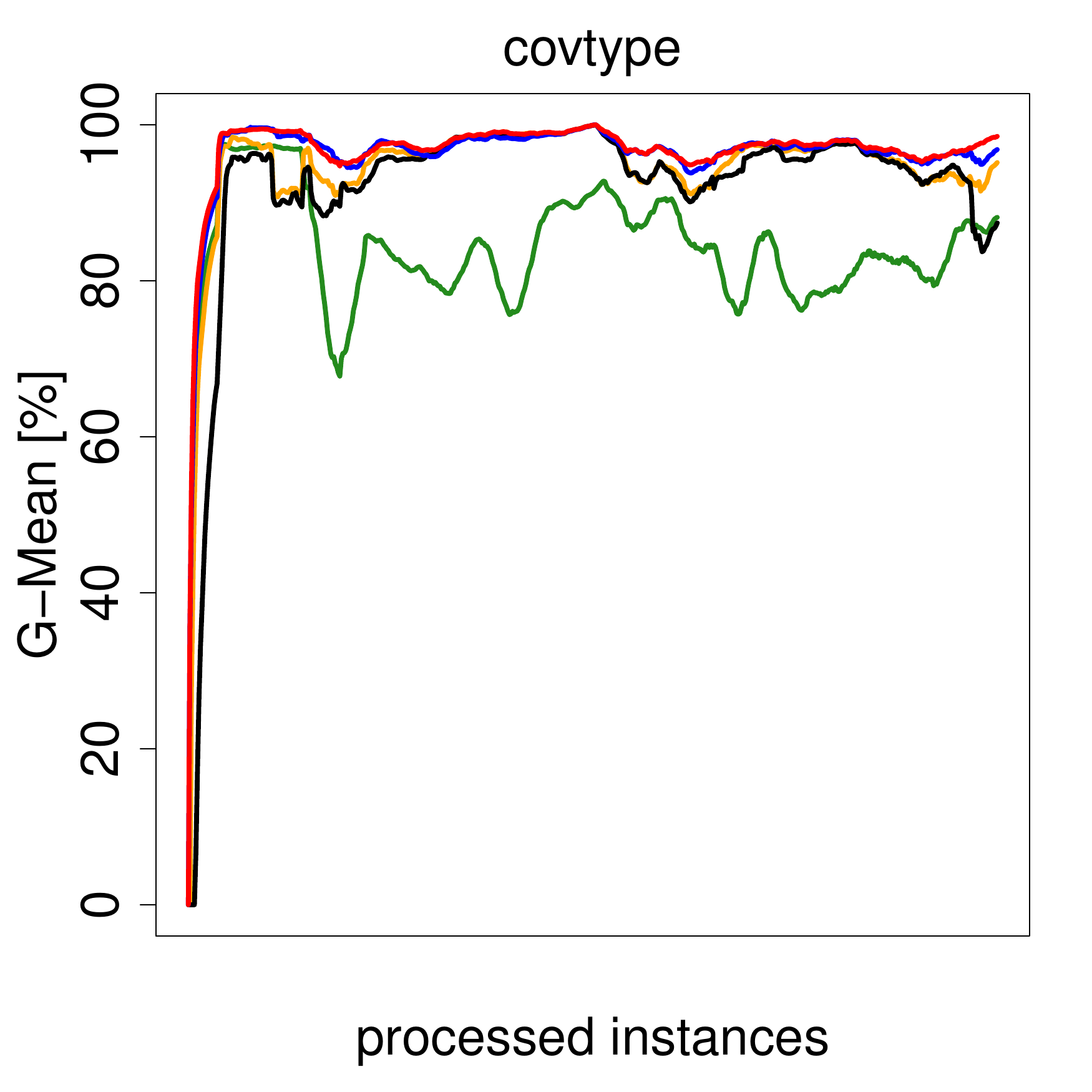}
\includegraphics[width=0.19\columnwidth]{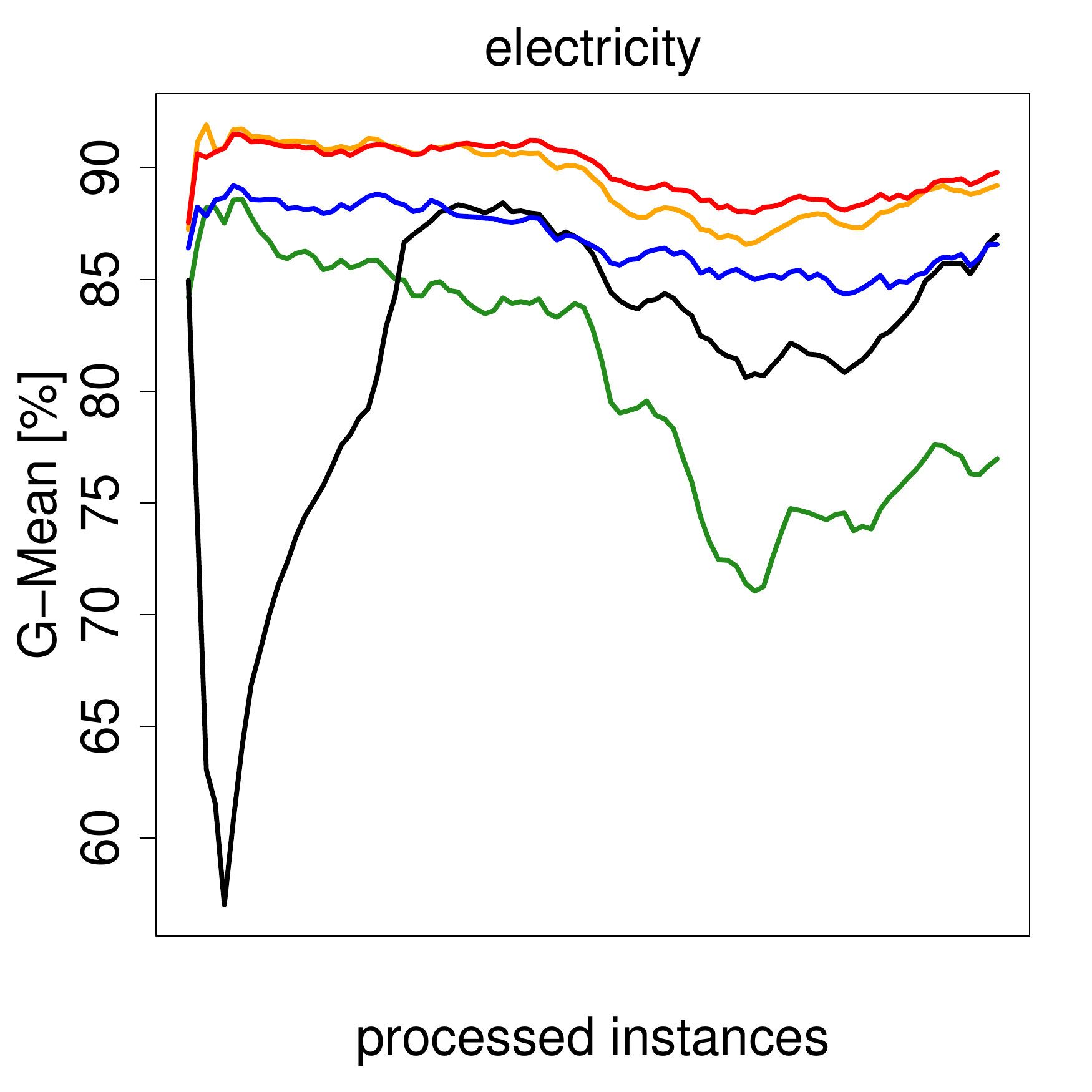}
\includegraphics[width=0.19\columnwidth]{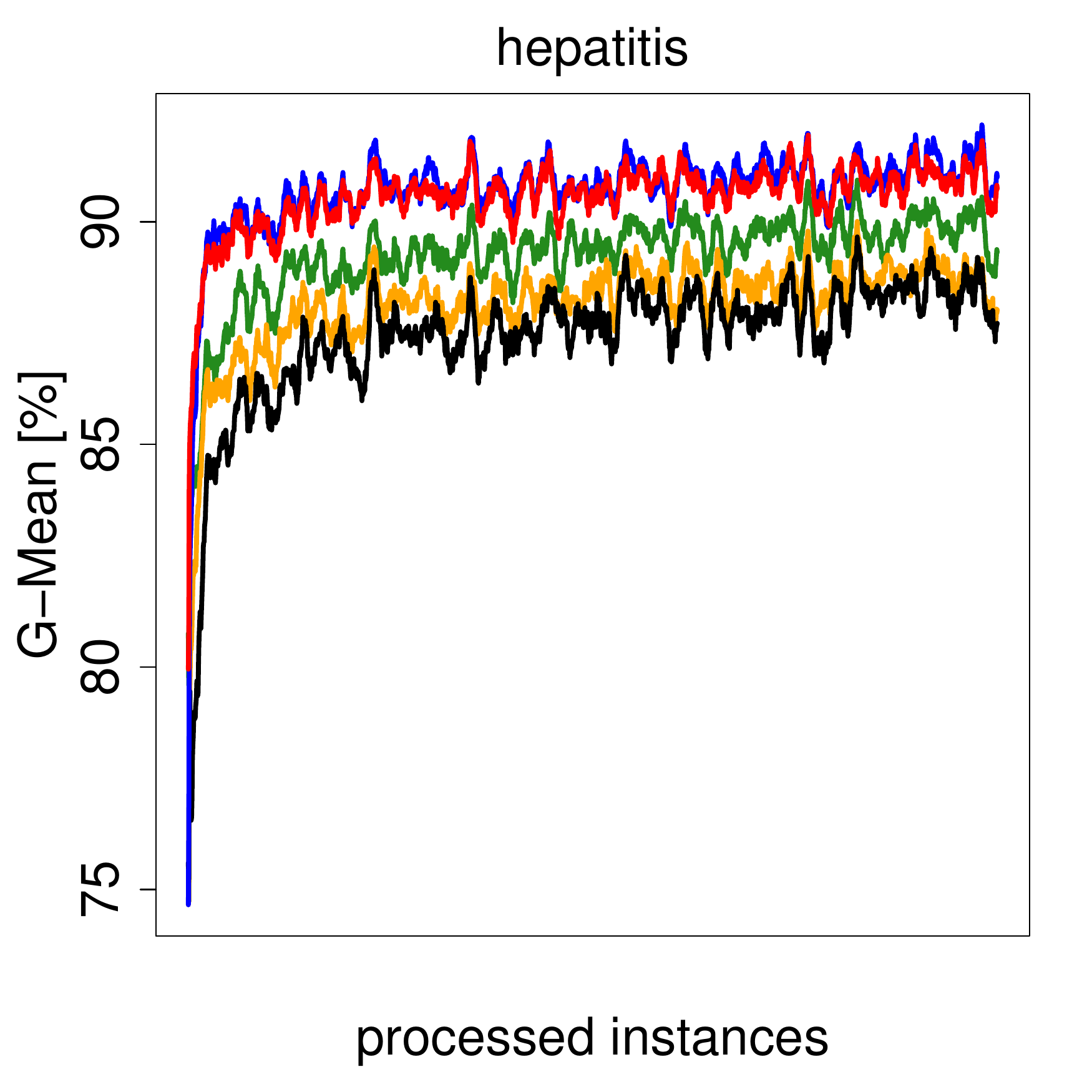}
\includegraphics[width=0.19\columnwidth]{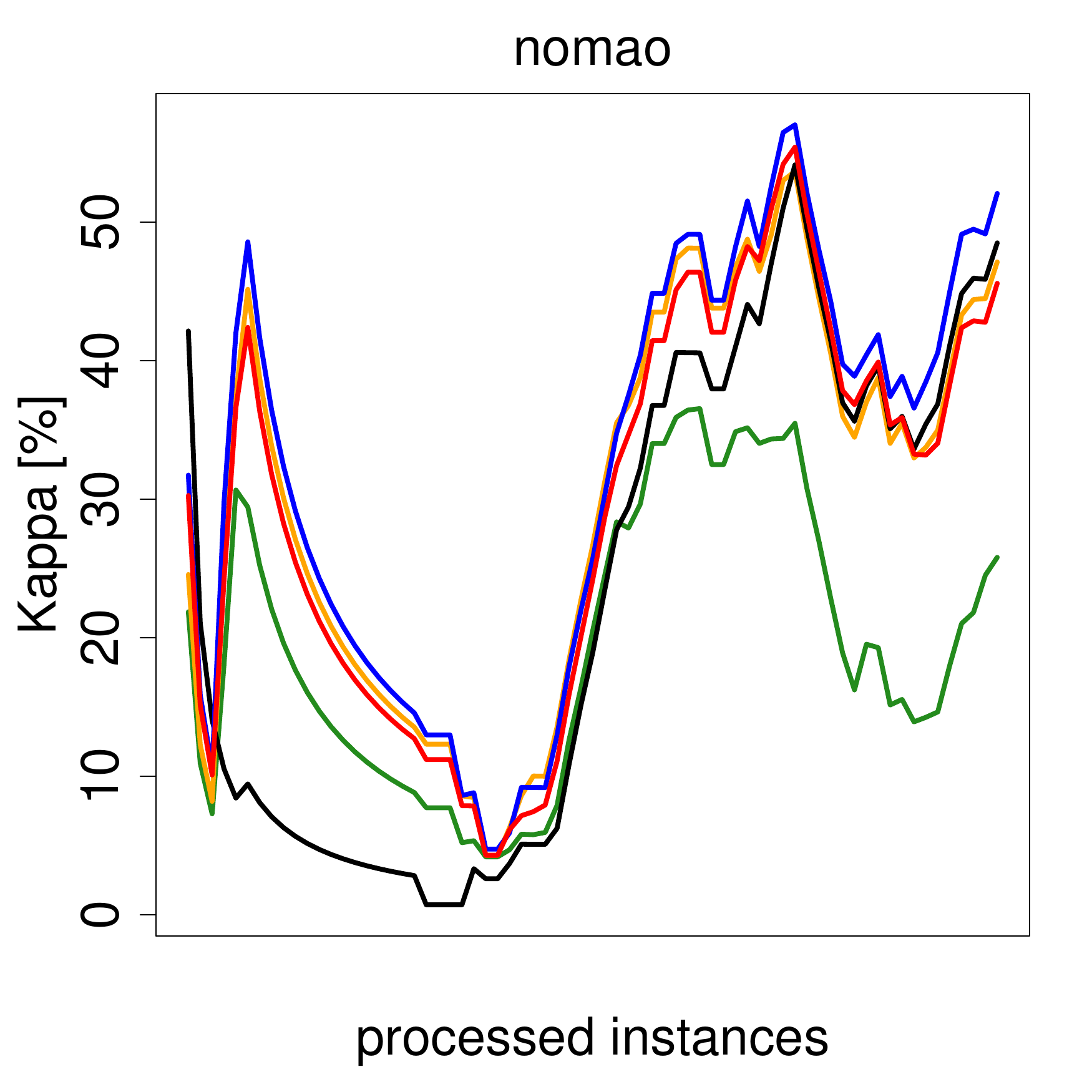}
\includegraphics[width=0.19\columnwidth]{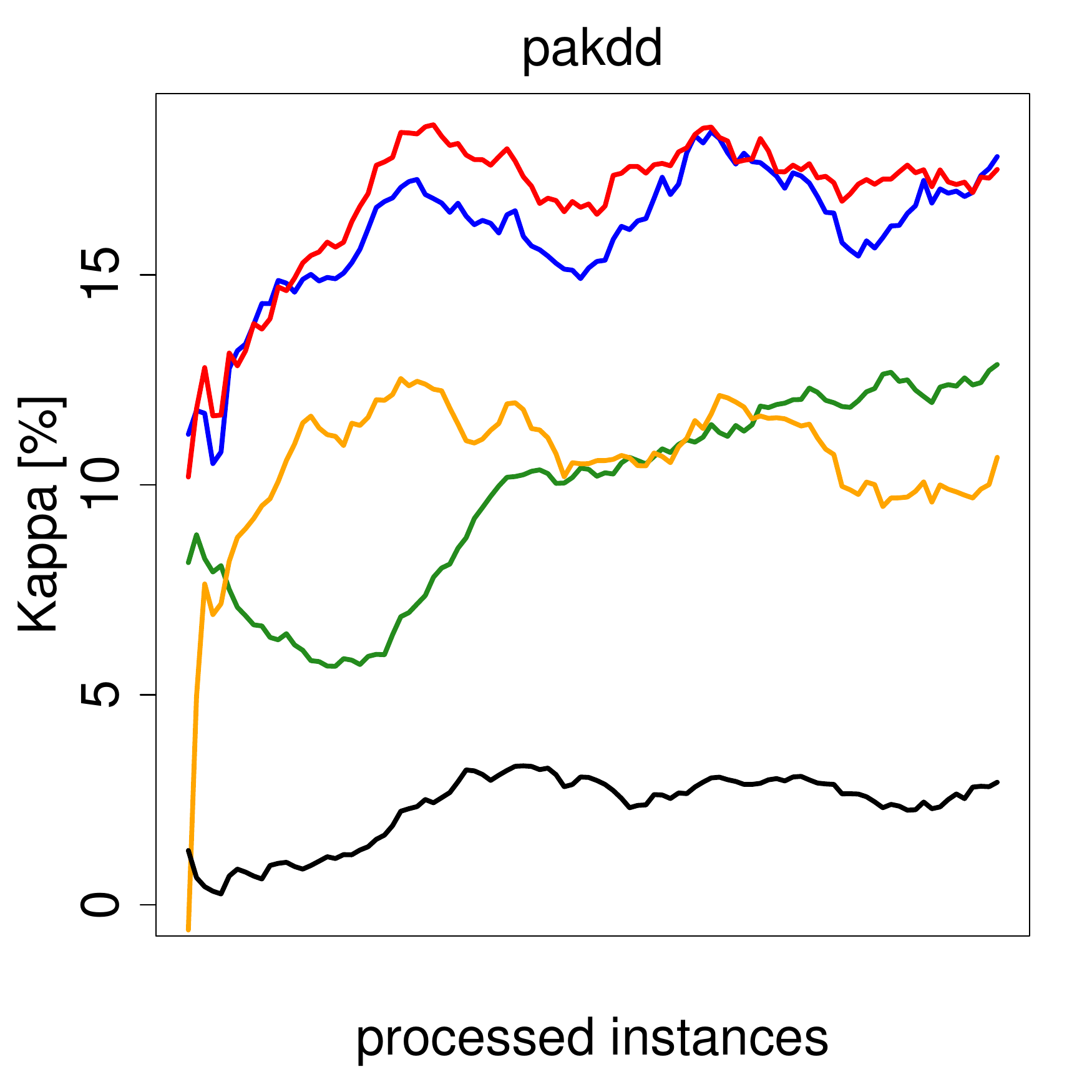}
\includegraphics[width=0.19\columnwidth]{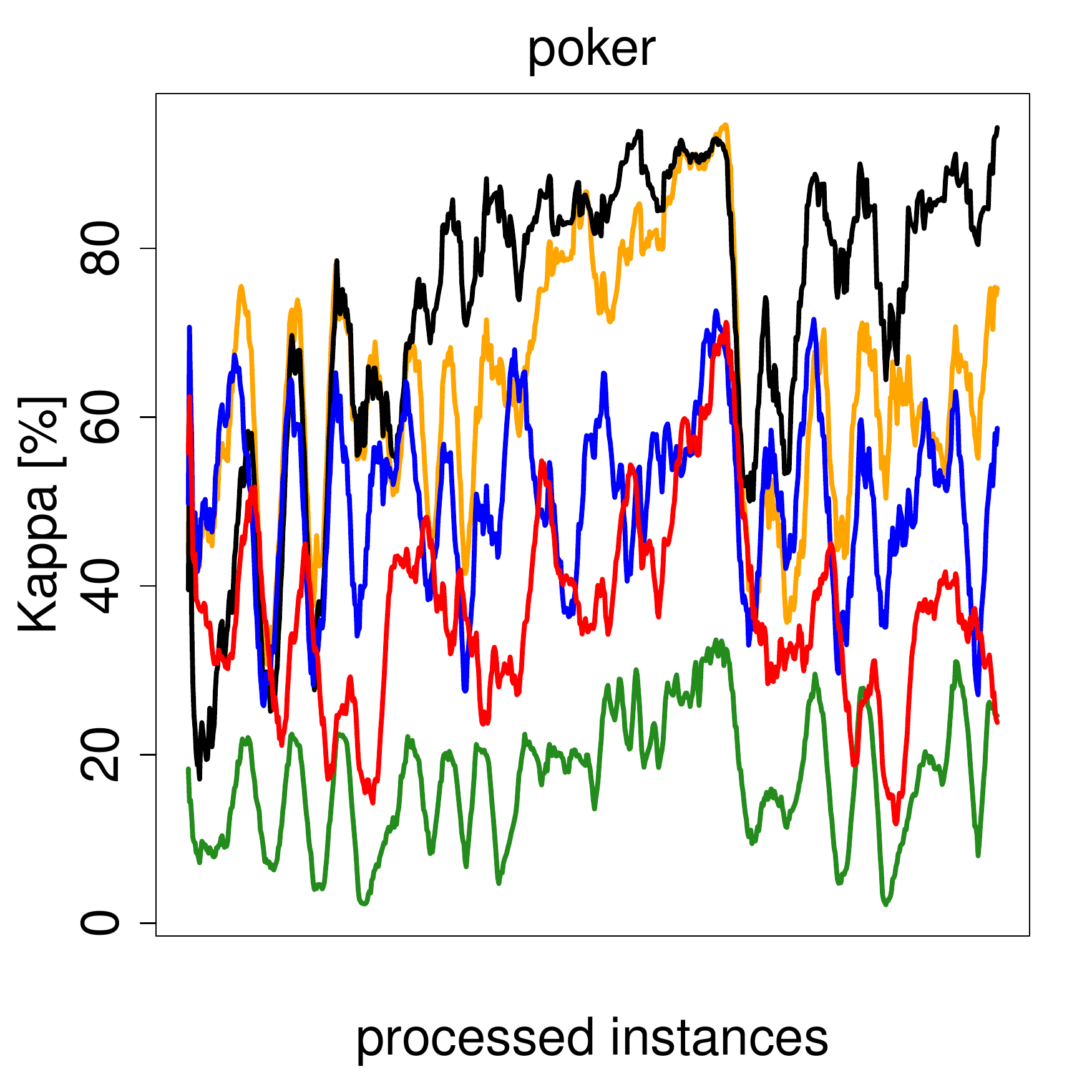}
\includegraphics[width=0.19\columnwidth]{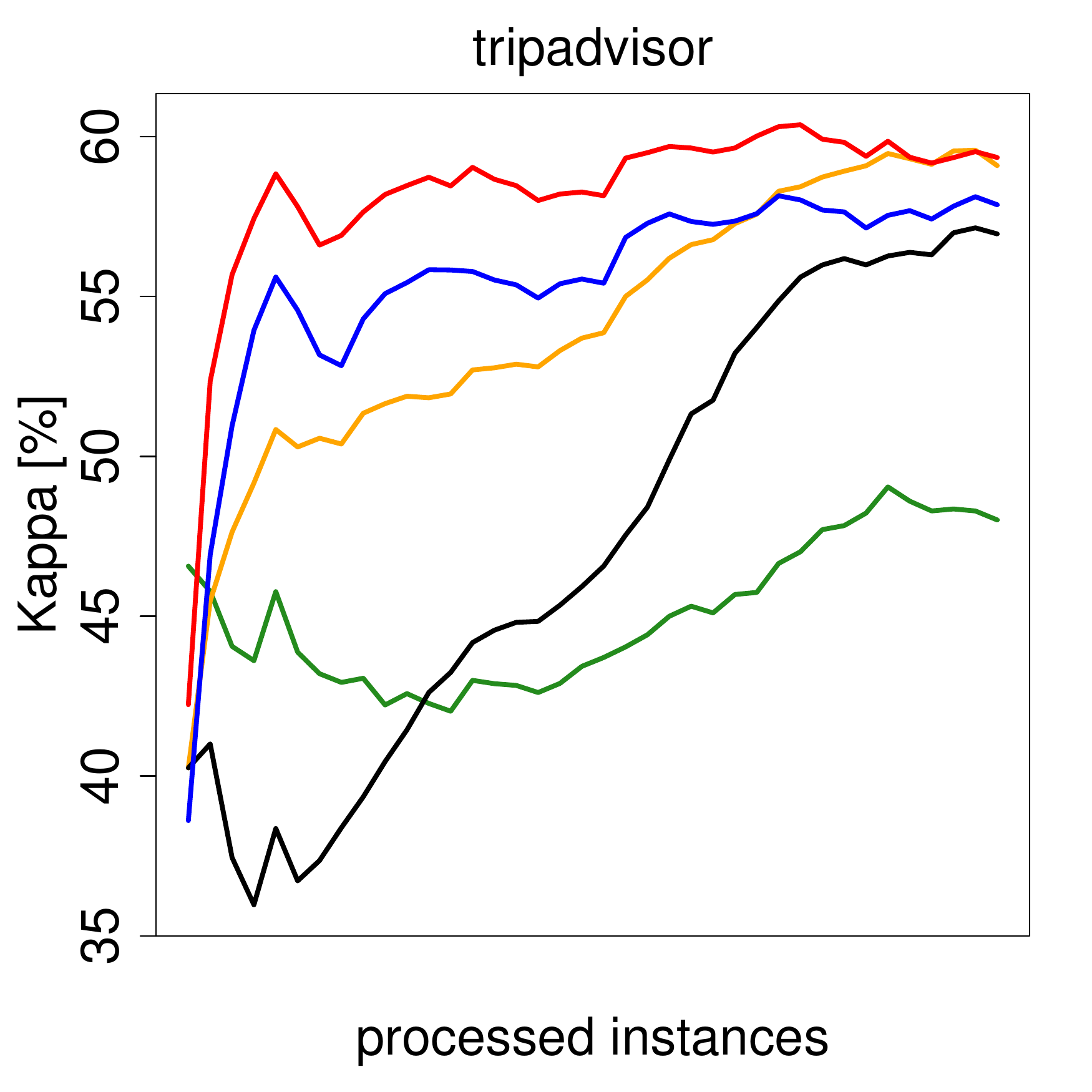}
\includegraphics[width=0.19\columnwidth]{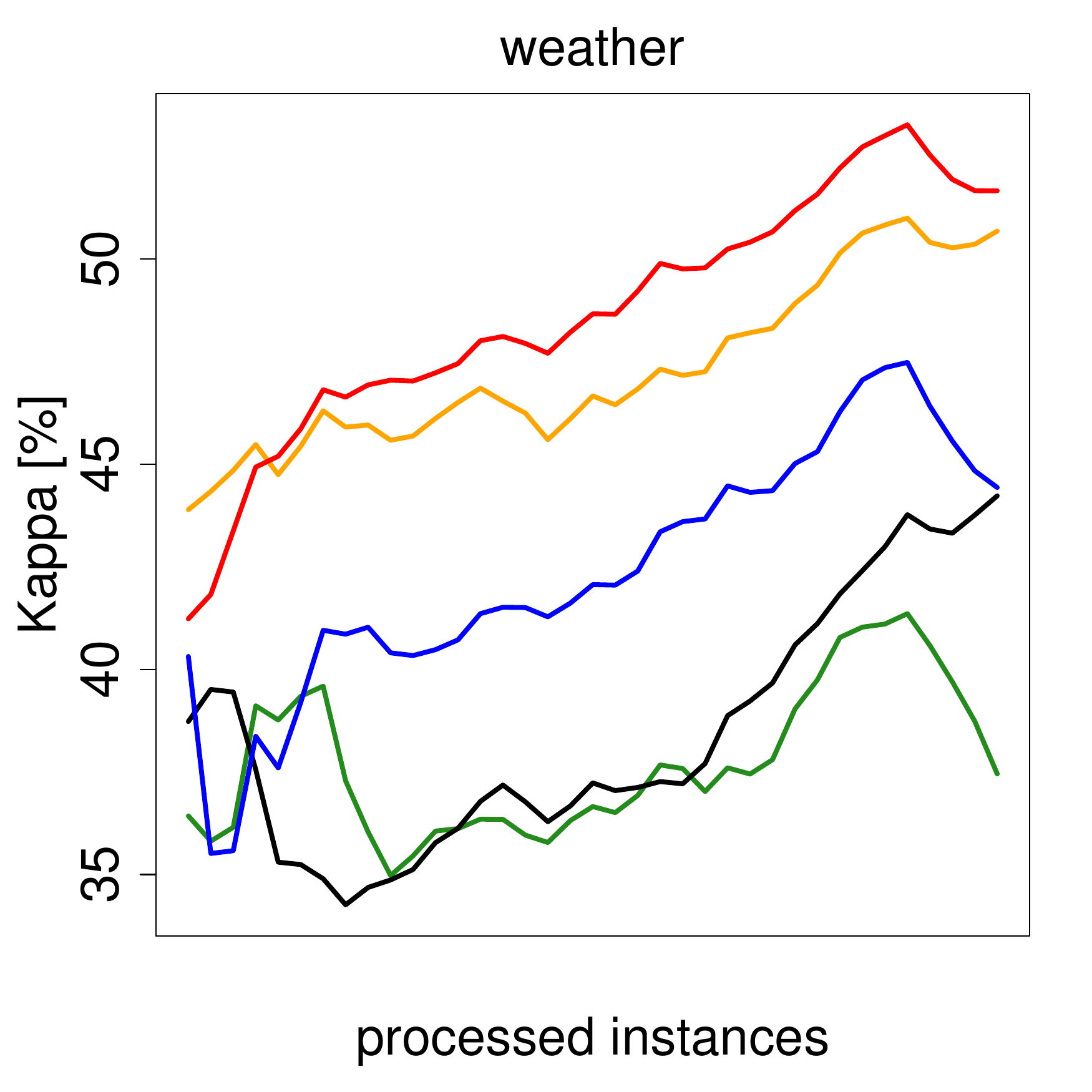}
\caption{G-Mean and Kappa on binary class imbalanced datasets.}
\label{fig:bc_datasets}
\vspace{1cm}
\end{figure}

\begin{figure}[t!]
\centering
\includegraphics[width=0.5\columnwidth]{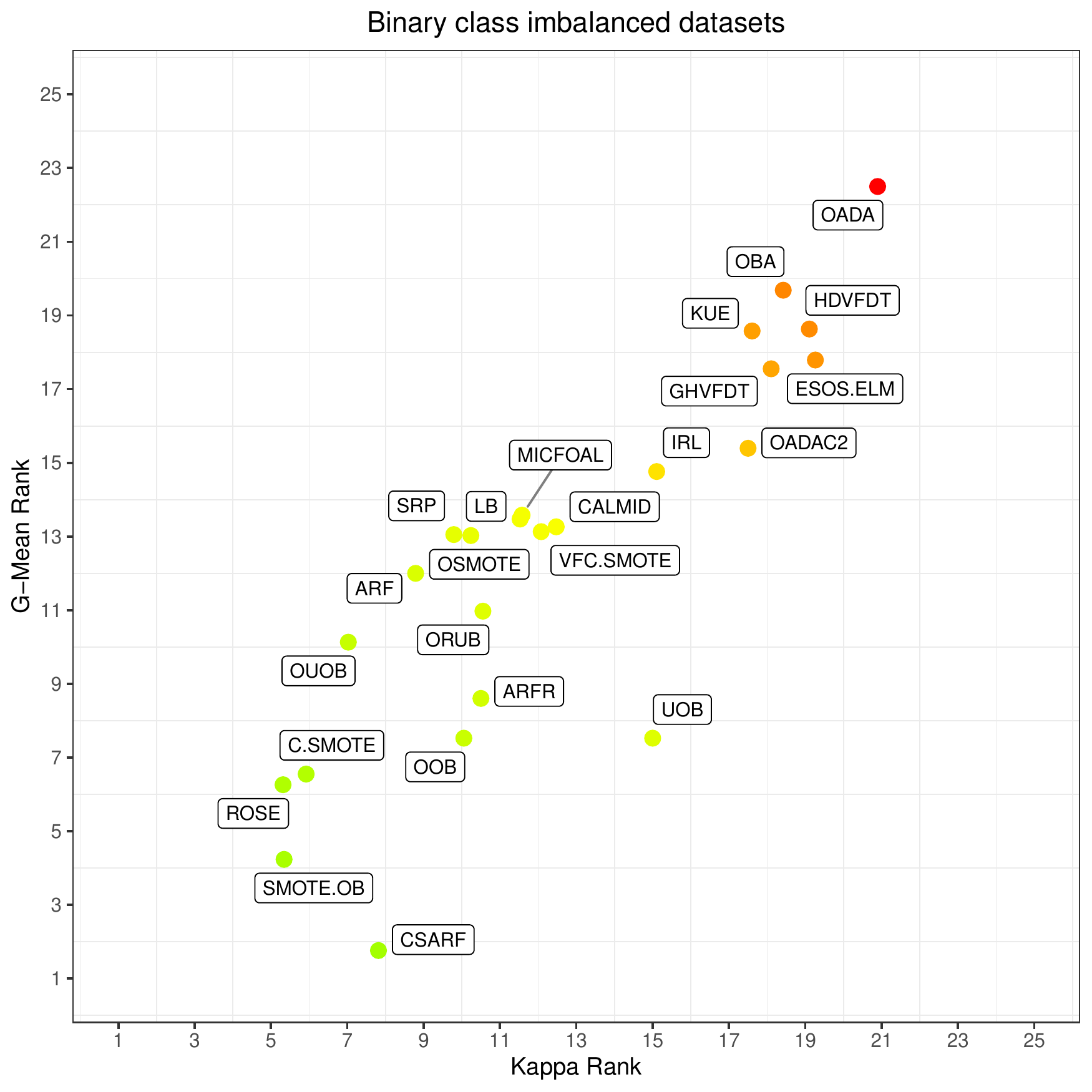}
\caption{Comparison of all 24 algorithms for binary class imbalanced datasets. Color gradient represents the product of both metrics.}
\label{fig:BC_datasets_ellipse}
\end{figure}

\noindent \textbf{Goal of the experiment.}
This experiment was designed to address \textbf{RQ5} and to evaluate the performance of the classifiers on $19$ real-world imbalanced data streams. The previous experiments focused on analyzing how the classifiers cope with various learning difficulties present in imbalanced data streams using synthetic generators, allowing us to inspect how the classifiers behave in specific and controlled scenarios. Meanwhile, real-world datasets pose specific challenges to classifiers, as they are not generated in a controlled environment. They are characterized by a combination of various learning difficulties that appear with varying intensity or frequency. Their imbalance ratio changes over time, while concept drift may oscillate among different types with varying speed. Therefore, assessing the performance of all classifiers on real-world data is a major step towards evaluation. The real-world data streams employed in the experiments are popular benchmarks for imbalanced data streams classifiers, and their specifications are presented at Table~\ref{tab:bc_datasets_spec}. Figure~\ref{fig:bc_datasets} illustrates the performance of the five selected classifiers in the real-world datasets. Table~\ref{tab:BC_datasets} presents the performance for the top 10 classifier on each dataset. Figure~\ref{fig:BC_datasets_ellipse} summarizes the overall performance of all classifiers in the real-world datasets scenario.

\begin{table}[b!]
\centering
\caption{Real-world binary datasets specifications.}
\label{tab:bc_datasets_spec}
\begin{tabular}{@{}lrr@{}}
\toprule
Dataset & Instances & Features\\
\midrule
adult & 45,222 & 14\\
amazon & 8,000 & 30\\
amazon-emp & 32,769 & 9\\
census & 299,284 & 41\\
coil2000 & 9,822 & 85\\
covtype & 267,001 & 54\\
creditcard & 284,807 & 30\\
electricity & 45,312 & 8\\
gmsc & 150,000 & 10\\
hepatitis & 1,000,000 & 19\\
internet-ads & 3,279 & 1,558\\
kddcup & 494,021 & 41\\
nomao & 34,465 & 118\\
pakdd & 50,000 & 27\\
poker & 359,999 & 10\\
spam & 9,324 & 499\\
tripadvisor & 18,569 & 30\\
twitter & 9,090 & 30\\
weather & 18,159 & 8\\
\bottomrule
\end{tabular}
\end{table}

\noindent \textbf{Characteristics of real-world imbalanced data streams.} Before analyzing the classifiers' performance in real-world datasets, it is important to point up the difference between artificial and real-world imbalanced data streams. Generators are probabilistic and base the generation of instances on prior probability taken from the parametric imbalance ratio. Their appearance in the stream is dictated strictly by these priors, leading to bounded windows in which minority and majority instances appear. In real-world datasets, this does not happen, since they were collected to model specific phenomenon observations and does not respect such clear probabilistic mechanisms. All of this poses unique challenges to classifiers, such as the latency with which instances from a specific class arrive, or long periods when instances from only one class appear. This configuration of data streams presents many more challenges for streaming classifiers. Such benchmarks allow us to gain insights about the classifiers examining them under unique and challenging conditions.

\begin{table*}[t!]
\centering
\footnotesize
\setlength{\tabcolsep}{4pt}
\caption{G-Mean and Kappa on binary class imbalanced datasets.}
\label{tab:BC_datasets}
\begin{tabular}{ll|C{1cm}C{1cm}C{1cm}C{1cm}C{1cm}C{1cm}C{1cm}C{1cm}C{1cm}C{1cm}}
\toprule
& Dataset & CSARF & ARF & KUE & LB & CALMID & ROSE & ARFR & SMOTE-OB & OOB & UOB\\
\midrule
\multirow{19}{*}{\rotatebox[origin=c]{90}{G-Mean}} 
& adult & 80.09 & 72.58 & 69.56 & 72.65 & 72.15 & 79.20 & \textbf{80.61} & 80.40 & 79.28 & 77.81\\
 & amazon & 67.19 & 52.05 & 32.57 & 25.17 & 24.35 & 62.38 & 50.54 & 67.18 & 58.15 & \textbf{68.11}\\
 & amazon-emp & \textbf{23.15} & 6.63 & 0.00 & 4.08 & 5.40 & 20.77 & 18.42 & 23.04 & 11.00 & 18.78\\
 & census & \textbf{48.67} & 27.79 & 22.97 & 29.11 & 32.45 & 39.29 & 41.84 & 44.01 & 36.57 & 45.59\\
 & coil2000 & 56.59 & 4.74 & 4.51 & 6.77 & 2.86 & 18.58 & 17.94 & 50.89 & 19.60 & \textbf{61.83}\\
 & covtype & \textbf{97.22} & 94.35 & 90.39 & 93.98 & 93.73 & 95.39 & 93.40 & 96.99 & 94.08 & 84.57\\
 & creditcard & \textbf{32.47} & 27.37 & 16.02 & 25.07 & 26.46 & 28.34 & 27.54 & 28.25 & 29.99 & 32.17\\
 & electricity & \textbf{89.79} & 89.42 & 70.49 & 88.71 & 83.11 & 89.14 & 89.45 & 86.67 & 81.67 & 79.67\\
 & gmsc & \textbf{76.58} & 30.55 & 32.02 & 30.31 & 43.89 & 58.33 & 65.77 & 71.65 & 71.35 & 70.65\\
 & hepatitis & 90.56 & 85.66 & 85.37 & 86.36 & 87.45 & 88.33 & 90.39 & \textbf{90.75} & 89.71 & 89.25\\
 & internet-ads & \textbf{13.90} & 13.54 & 0.00 & 12.98 & 12.98 & 13.36 & 0.00 & \textbf{13.90} & 13.36 & 13.36\\
 & kddcup & 7.71 & 7.26 & 4.19 & 7.44 & 7.61 & 7.83 & 7.27 & 7.52 & 7.65 & \textbf{7.86}\\
 & nomao & 42.91 & 37.38 & 30.35 & 34.89 & 34.24 & 40.09 & 35.68 & 42.13 & 40.04 & \textbf{43.73}\\
 & pakdd & \textbf{59.48} & 2.09 & 12.82 & 4.37 & 15.66 & 39.71 & 57.06 & 54.60 & 54.66 & 55.06\\
 & poker & \textbf{79.61} & 45.27 & 53.38 & 57.52 & 76.98 & 72.81 & 49.88 & 62.31 & 73.39 & 72.45\\
 & spam & \textbf{80.63} & 74.24 & 67.13 & 73.13 & 72.15 & 77.44 & 79.37 & 80.31 & 75.80 & 72.86\\
 & tripadvisor & 64.11 & 18.66 & 14.21 & 22.54 & 21.83 & 45.74 & 57.10 & 64.06 & 60.92 & \textbf{64.55}\\
 & twitter & \textbf{79.61} & 78.53 & 58.03 & 76.34 & 55.57 & 77.14 & 77.77 & 78.95 & 71.39 & 70.39\\
 & weather & \textbf{76.45} & 68.02 & 59.75 & 67.38 & 67.86 & 74.28 & 74.19 & 73.53 & 69.68 & 67.19\\
\midrule
\multirow{19}{*}{\rotatebox[origin=c]{90}{Kappa}} 
 & adult & 53.17 & 53.80 & 52.31 & 54.05 & 52.80 & \textbf{55.68} & 53.60 & 53.68 & 52.72 & 47.99\\
 & amazon & 29.30 & 31.99 & 12.99 & 7.78 & 5.38 & \textbf{33.07} & 17.84 & 26.25 & 25.85 & 13.00\\
 & amazon-emp & 1.91 & 0.35 & 0.00 & 0.13 & 0.19 & \textbf{2.42} & 0.88 & 1.21 & 0.59 & 1.15\\
 & census & 20.87 & 19.72 & 14.66 & 20.47 & 22.13 & 25.95 & 23.46 & \textbf{26.78} & 24.20 & 22.75\\
 & coil2000 & 12.53 & 1.24 & 0.74 & 2.23 & 0.41 & 6.22 & 2.65 & \textbf{13.20} & 5.82 & 9.49\\
 & covtype & 85.26 & \textbf{91.22} & 83.43 & 89.94 & 88.60 & 90.13 & 83.64 & 89.51 & 83.21 & 49.74\\
 & creditcard & 17.96 & 26.43 & 15.45 & 24.30 & 25.56 & 27.30 & 22.61 & 27.33 & \textbf{28.44} & 6.71\\
 & electricity & 79.89 & \textbf{79.95} & 50.14 & 78.44 & 68.81 & 78.64 & 79.70 & 73.91 & 65.57 & 62.13\\
 & gmsc & 29.09 & 15.98 & 17.02 & 15.74 & 26.26 & 35.23 & 31.53 & \textbf{35.75} & 35.22 & 23.95\\
 & hepatitis & 72.92 & 76.83 & 75.32 & 76.62 & \textbf{77.30} & 74.26 & 74.73 & 75.33 & 74.17 & 69.47\\
 & internet-ads & \textbf{13.53} & 13.12 & 0.00 & 12.48 & 12.48 & 12.91 & 0.00 & \textbf{13.53} & 12.91 & 12.91\\
 & kddcup & 7.04 & 7.14 & 3.56 & 7.31 & 7.45 & \textbf{7.49} & 6.99 & 7.25 & 7.27 & 6.26\\
 & nomao & 32.56 & 32.70 & 18.71 & 29.71 & 28.11 & 33.24 & 29.11 & \textbf{36.02} & 26.73 & 21.56\\
 & pakdd & \textbf{17.32} & 0.29 & 1.29 & 0.52 & 2.77 & 11.06 & 15.88 & 16.52 & 15.08 & 10.38\\
 & poker & 36.84 & 37.95 & 43.85 & 50.82 & \textbf{72.67} & 63.49 & 30.14 & 50.72 & 54.39 & 17.05\\
 & spam & \textbf{58.74} & 58.33 & 49.72 & 54.56 & 51.84 & 56.45 & 58.50 & 56.68 & 49.81 & 45.43\\
 & tripadvisor & \textbf{24.97} & 6.15 & 2.55 & 7.51 & 6.51 & 18.12 & 23.67 & 24.12 & 23.53 & 22.25\\
 & twitter & 73.73 & 75.21 & 50.07 & 70.89 & 50.43 & 72.39 & 71.53 & \textbf{75.24} & 60.97 & 58.81\\
 & weather & \textbf{49.81} & 46.77 & 34.79 & 44.00 & 41.19 & 48.32 & 45.29 & 43.08 & 40.32 & 36.19\\
\midrule
\multicolumn{2}{l|}{Avg. G-Mean} & \textbf{61.41} & 44.01 & 38.09 & 43.10 & 44.04 & 54.11 & 53.38 & 58.80 & 54.65 & 57.68\\
\multicolumn{2}{l|}{Avg. Kappa} &  37.76 & 35.53 & 27.72 & 34.08 & 33.73 & \textbf{39.60} & 35.36 & 39.27 & 36.15 & 28.27\\
\midrule
\multicolumn{2}{l|}{Rank G-Mean} & \textbf{1.50} & 6.95 & 9.39 & 7.61 & 7.61 & 4.37 & 5.13 & 3.24 & 4.74 & 4.47\\
\multicolumn{2}{l|}{Rank Kappa} & 4.08 & 4.84 & 8.87 & 6.03 & 6.34 & 3.11 & 5.39 & \textbf{3.03} & 5.63 & 7.68\\
\bottomrule
\end{tabular}
\end{table*}

\noindent \textbf{Discussion.}

\noindent \textit{Impact of approach to class imbalance.} First, it is interesting to note than on average all examined methods displayed much better Kappa than G-mean performance. We can observe that ensembles utilizing blind resampling, such as \acrshort{oob} and \acrshort{uob}, returned poor performance over real-world data streams. We can explain this by their purely online nature paired up with catastrophic forgetting, as these ensembles adapt their resampling strategy to the newest arriving instances, and thus are not being able to retain any memory of previously seen concepts. As in real-world scenarios instances do not arrive in stratified windows, one of classes may disappear for a while. This confuses such ensembles and leads to high skewness towards one class that is very difficult to overcome via online blind resampling. The methods based on informed resampling, such as \acrshort{csmote} and \acrshort{smoteob} displayed satisfactory results, showing that their learning mechanisms are robust to various characteristics of real-world streams. Also, \acrshort{smoteob} achieved balanced results regarding both metrics, demonstrating a high stability and reliability in real-world cases. Interestingly, \acrshort{csmote} was underperforming in previous synthetic cases, showing discrepancies between artificial and real-world domains. This allows us to conclude that there is a need for further research in real-world imbalanced streams and capturing more realistic benchmarks that reflect various learning difficulties.

When analyzing algorithm-level modification classifiers, \acrshort{rose} displayed the best results, especially for the Kappa metric. While for synthetic datasets \acrshort{rose} was consistently among the best methods, for real-world cases we can see that its robustness to a variety of learning difficulties allowed it to demonstrate its potential. \acrshort{rose} stores buffers for each class independently contributing to scenarios with high latency of instances. \acrshort{csarf} remained as one of the best-performing classifier, displaying the best results on G-Mean, and being among the best regarding Kappa.

The worst-performing algorithms in the real-world scenario differ from what we saw in previous scenarios (with exception of \acrshort{oada} still being the weakest classifier). Algorithms that achieved average to good performance in other experiments such as \acrshort{kue}, \acrshort{hdvfdt} and \acrshort{oba} did not maintain their performance over real-world datasets. 

It is interesting to see that \acrshort{arf} which was not among the best-performing classifiers in the experiment with synthetic data, was the third-best classifier regarding Kappa. This shows that in the used real-world datasets the impact of concept drift was much more significant than the impact of class imbalance, allowing for a method focusing purely on adaptation to changes to rank so high.

\noindent \textit{Impact of ensemble architecture.} Real-world datasets allow us to evaluate how each type of ensemble architecture deals with streams under multiple difficulties appearing at the same time. We can see that all ensemble-based methods display much better performance on average than in the previous experiments. This is especially true of boosting-based methods that reduced the gap in their performance when compared to top-performing algorithms. However, bagging-based and hybrid ensembles still are the superior choices. This shows how these architectures offer better robustness in scenarios where data does not follow uniform characteristics over extended periods.

\subsection{Multi-class experiments}
\label{sec:mc-experiments}

The second set of experiments focuses on multi-class problems where the relationships among the many classes may vary over time \citep{LANGO2022116962}. Multi-class imbalanced data is more difficult and less frequently studied than its binary counterpart. There are relative imbalance ratios among classes and overlapping of the minority and majority classes becomes a greater issue \citep{SANTOS2023228,stefanowski2021classification,lipska2022influence}. These experiments include static imbalance ratio, dynamic imbalance ratio, concept drift and static imbalance ratio, concept drift and dynamic imbalance ratio, analysis on the impact of the number of classes, real-world multi-class datasets, and semi-synthetic multi-class imbalanced datasets. The number of examined algorithms in this set of experiments is reduced to $15$ following their multi-class capabilities shown in Table~\ref{tab:algorithms}.

\subsubsection{Static imbalance ratio}
\label{sec:mc-static-IR}

\begin{figure}[t!]
\centering
\includegraphics[width=0.16\columnwidth]{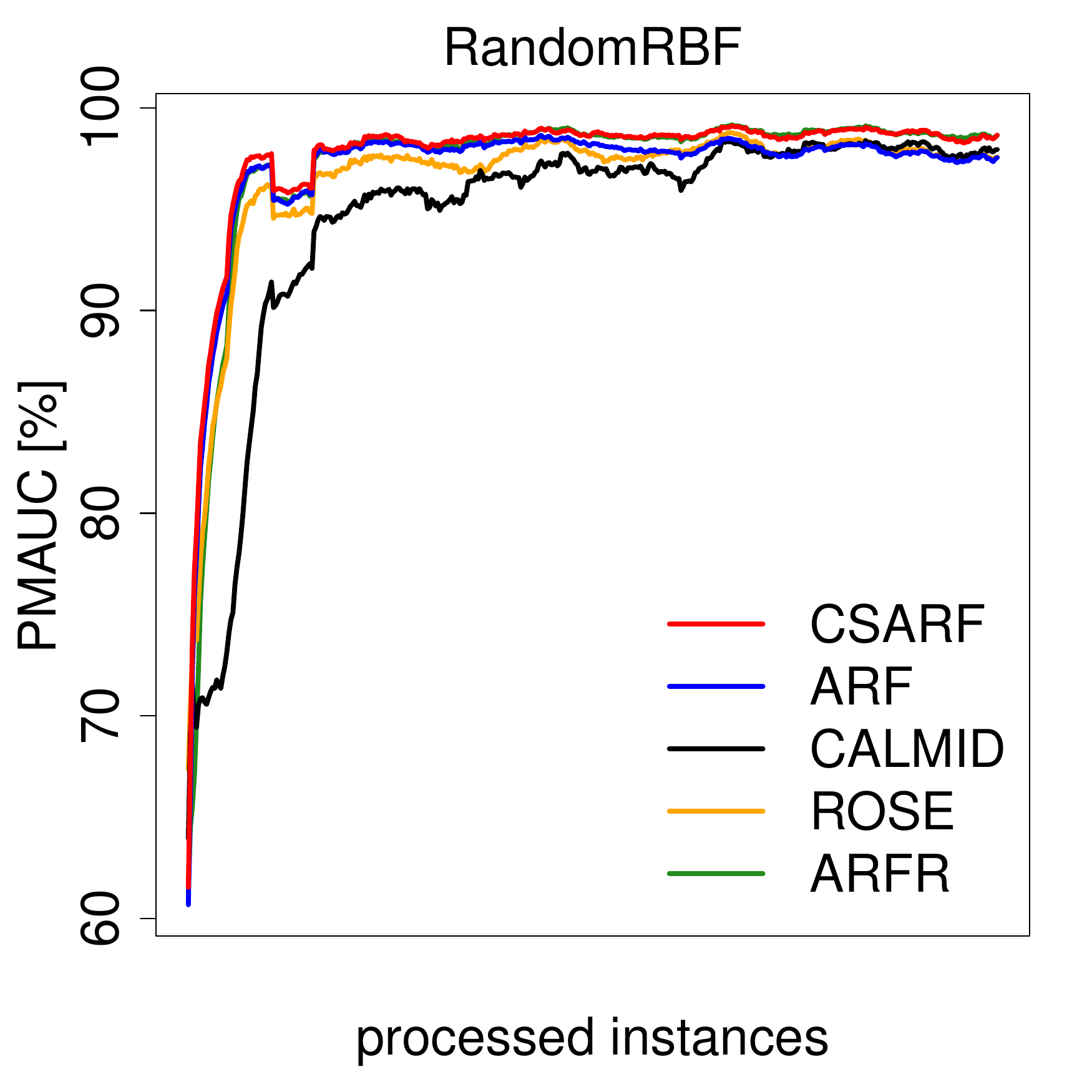}
\includegraphics[width=0.16\columnwidth]{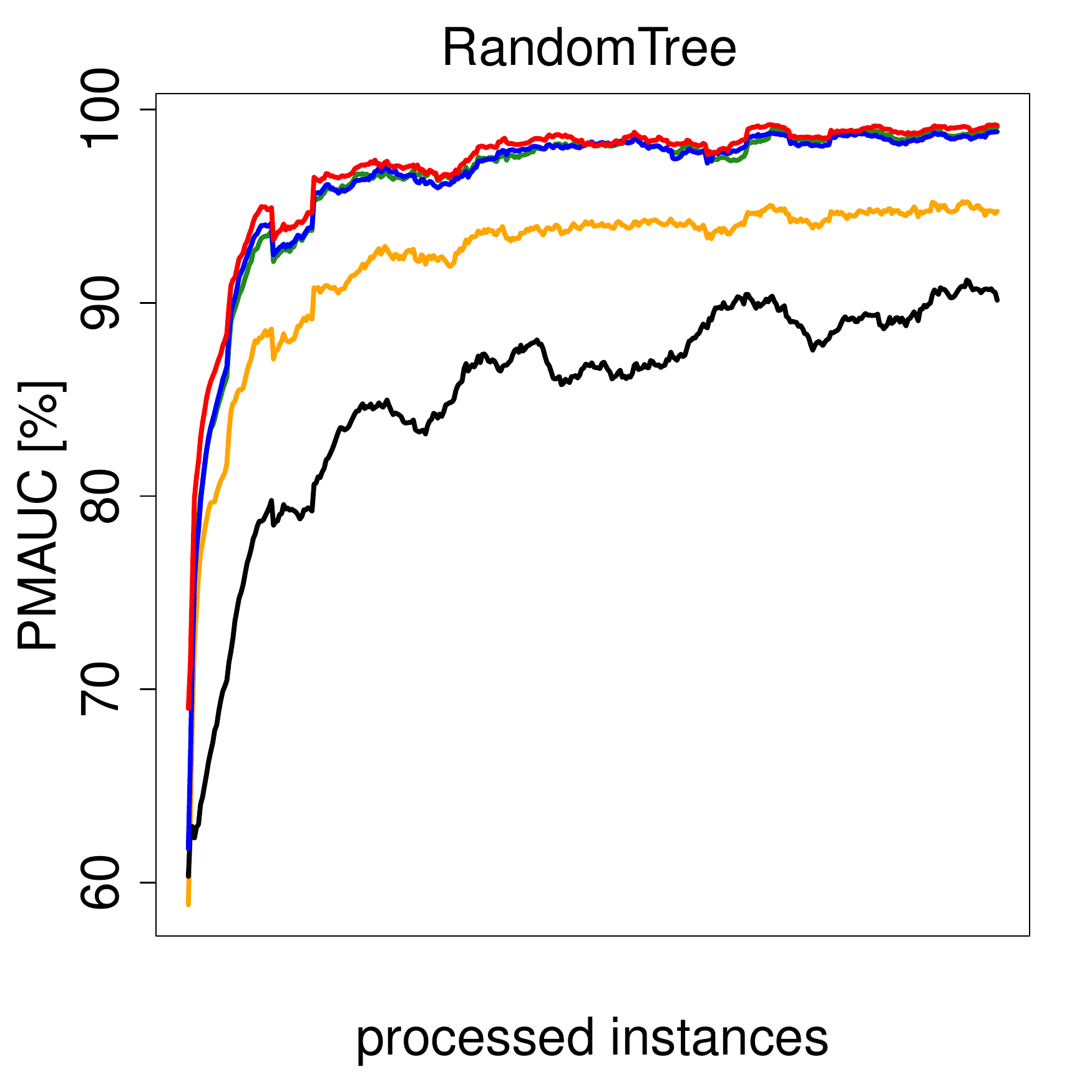}
\includegraphics[width=0.16\columnwidth]{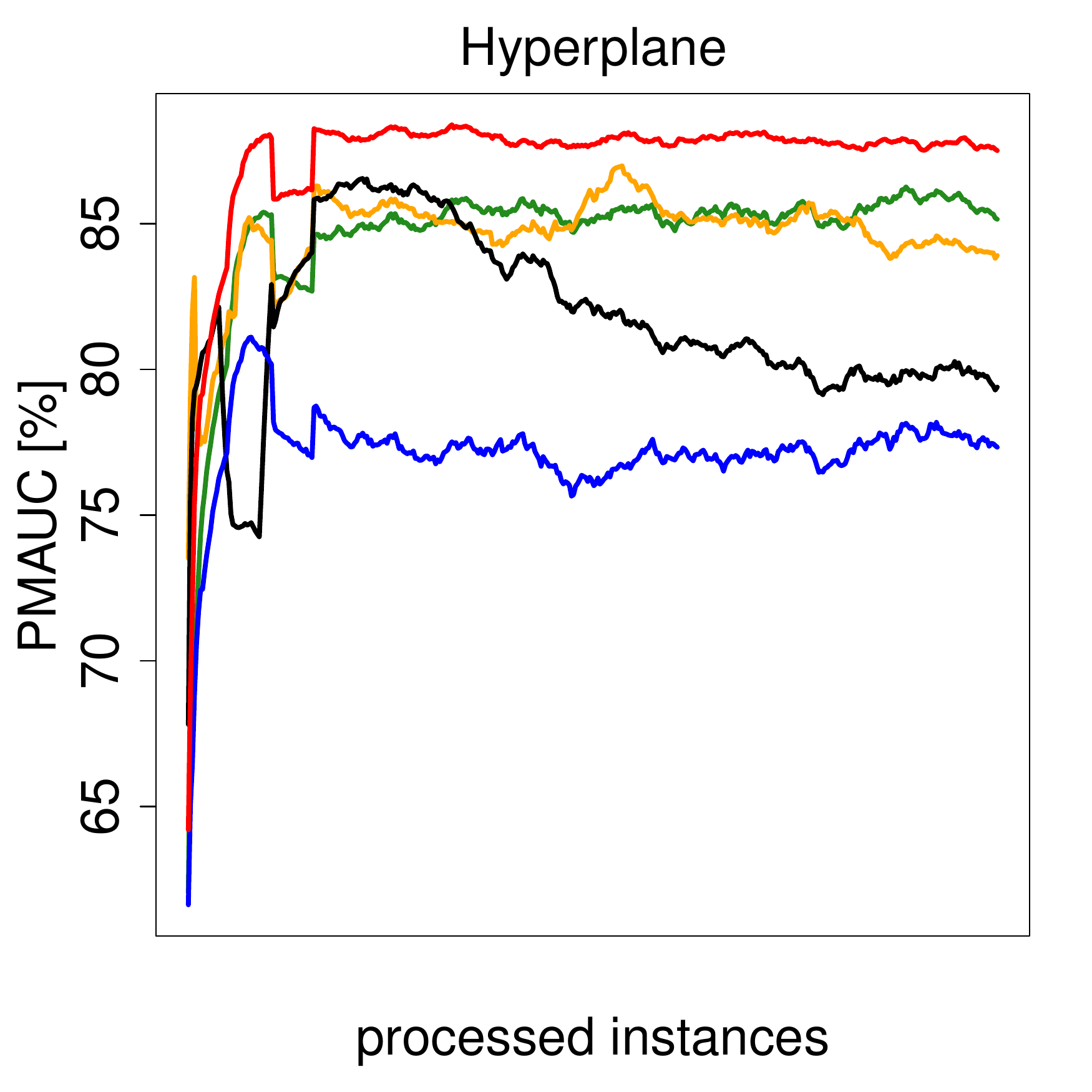}
\includegraphics[width=0.16\columnwidth]{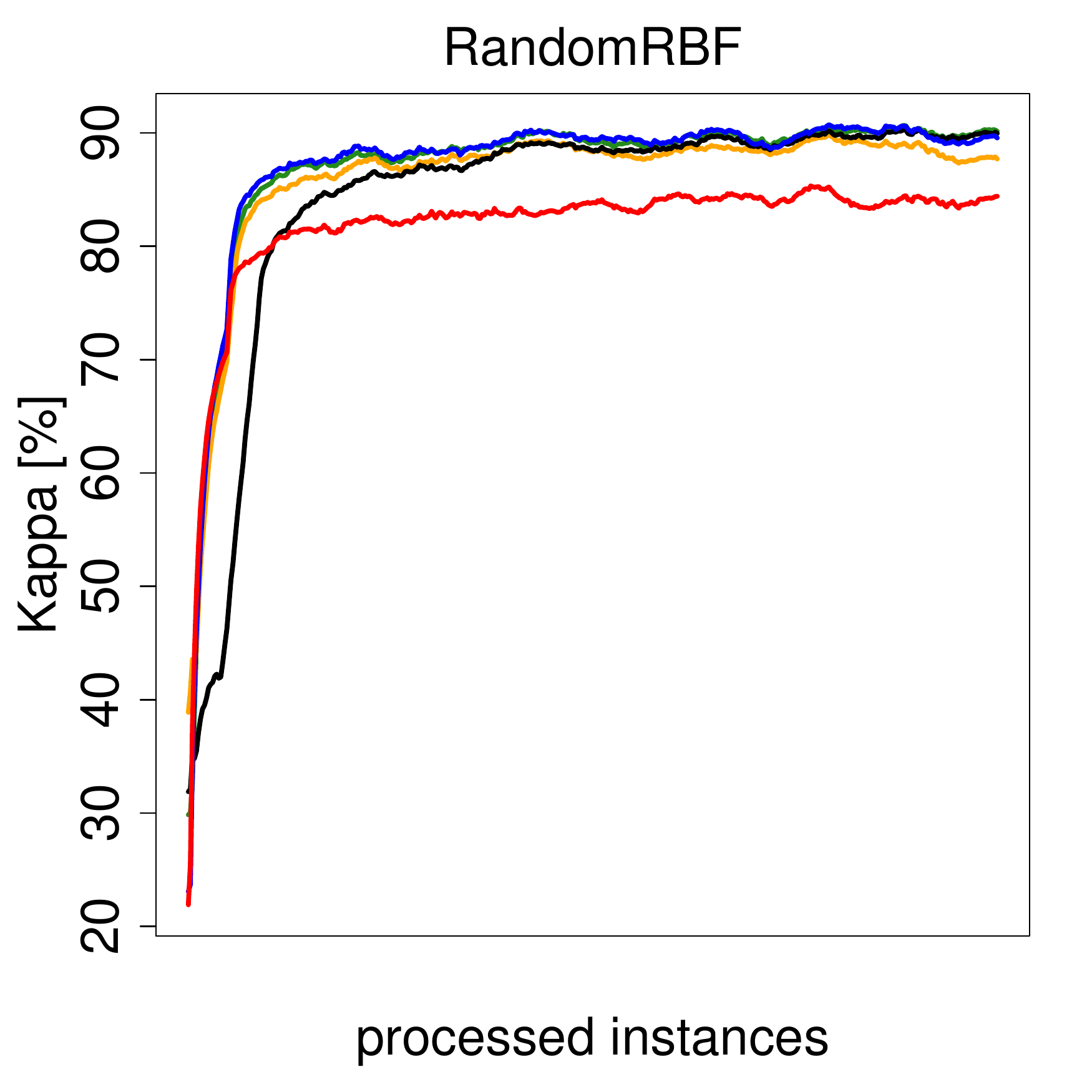}
\includegraphics[width=0.16\columnwidth]{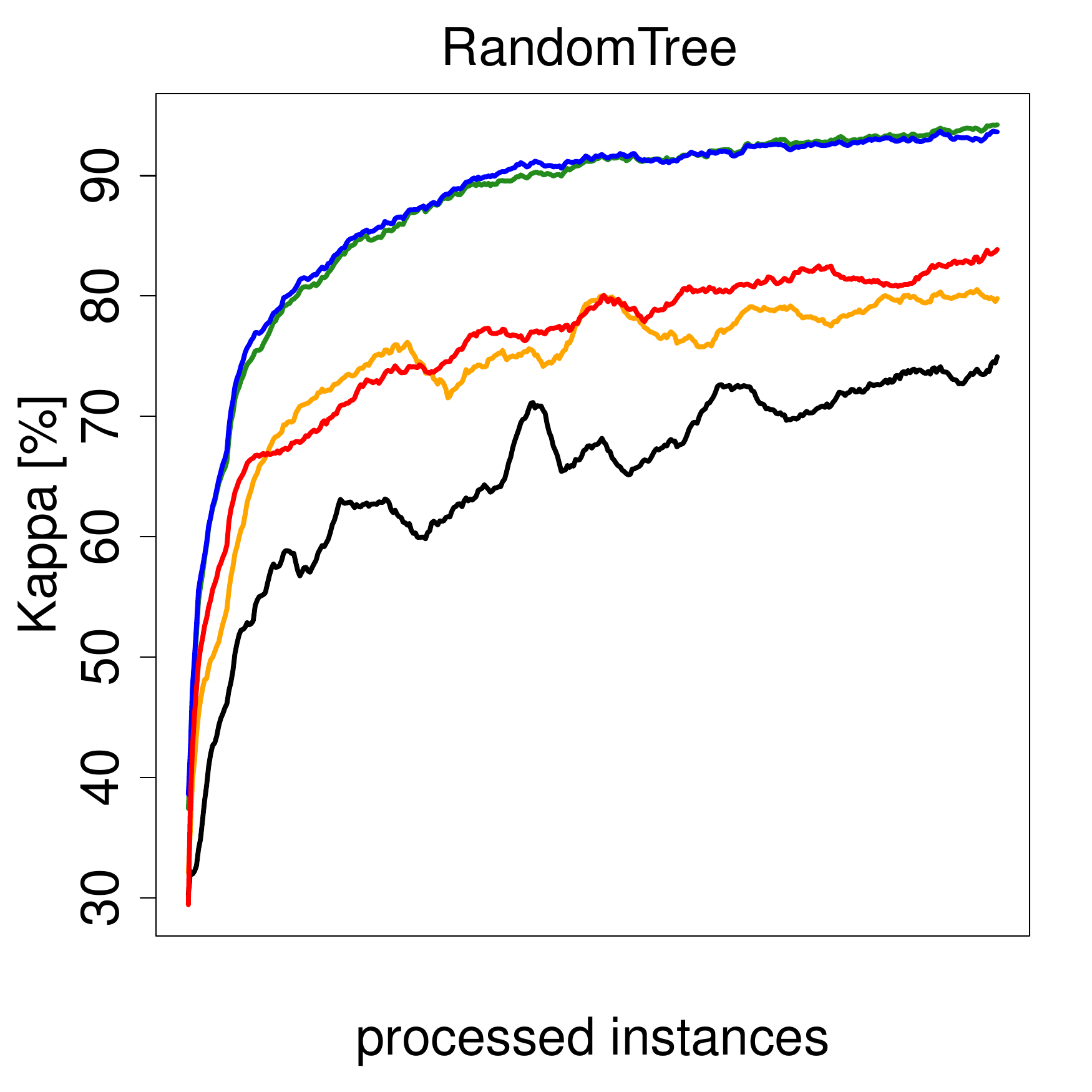}
\includegraphics[width=0.16\columnwidth]{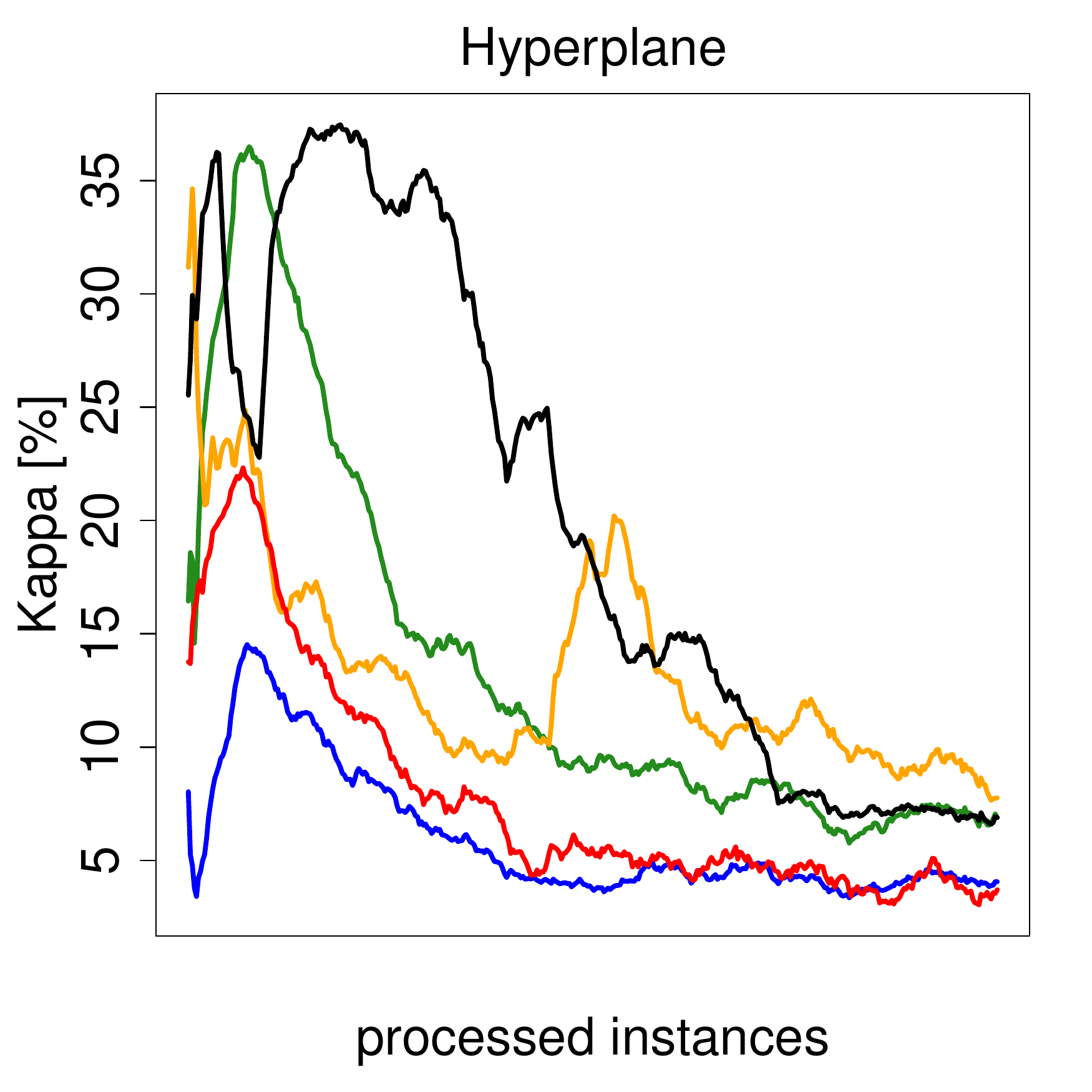}
\caption{PMAUC and Kappa on multi-class static imbalance ratio.}
\label{fig:mc_static_imbalance_ratio}
\end{figure}

\begin{figure}[t!]
\vspace*{0.5cm}
\centering
\includegraphics[width=0.5\columnwidth]{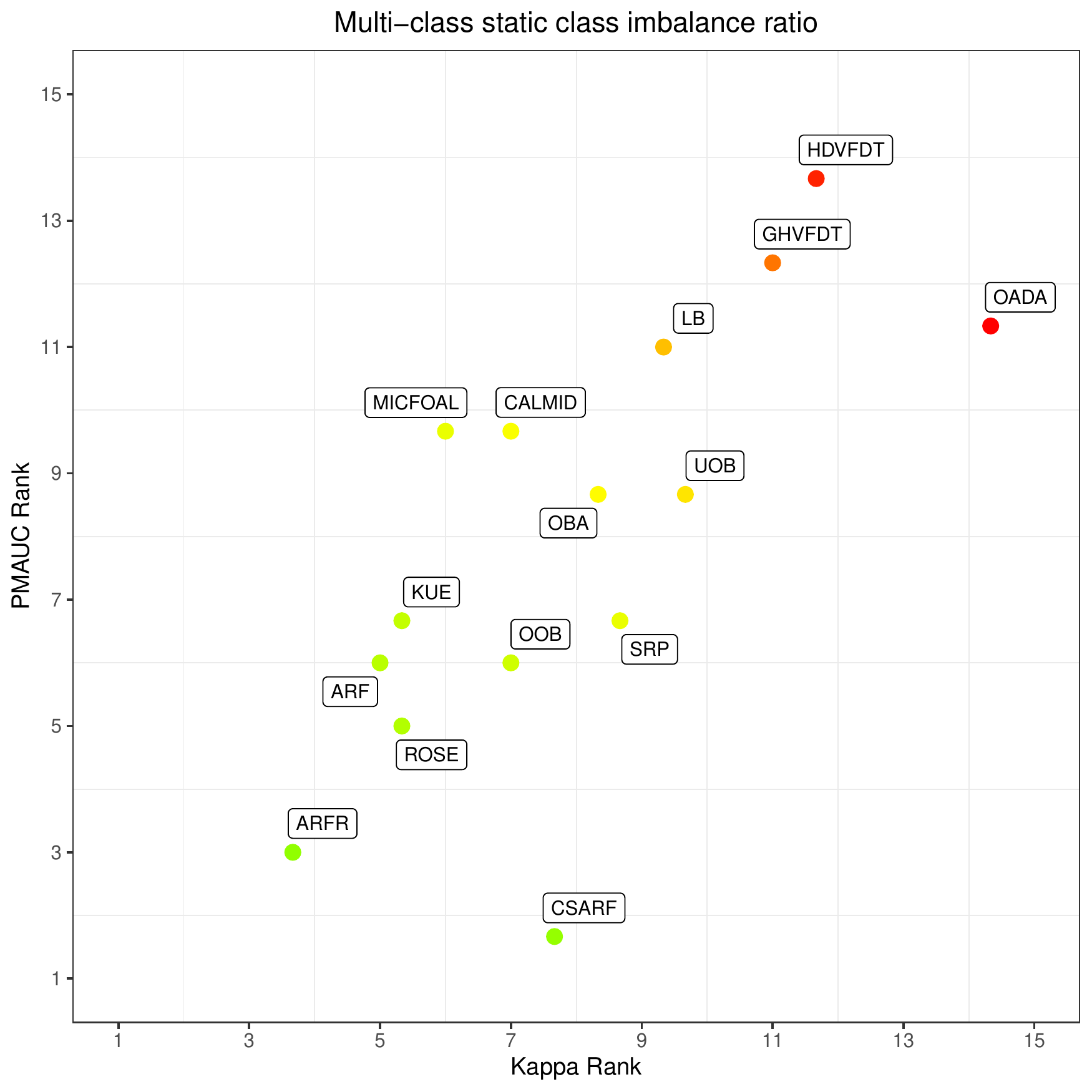}
\caption{Comparison of all 15 algorithms for multi-class static class imbalance ratio. Color gradient represents the product of both metrics.}
\label{fig:MC_SIR_scatter}
\end{figure}

\noindent \textbf{Goal of the experiment.}
This experiment was designed to address \textbf{RQ1} and to evaluate the performance and robustness of the classifiers to the static class imbalance in a scenario with multiple classes. In multi-class settings, the class imbalance can be even more challenging than in binary settings, since now multiple classes can be underrepresented. Also, relations among the classes are no longer obvious, since one class may be a majority when compared to some other classes, but a minority for the rest of them. This allows us to analyze how each classifier behaves under specific class distributions. To evaluate this, we prepared three multi-class generators \{Hyperplane, RandomRBF, and RandomTree\}, all of them with $5$ classes using the class distribution \{50, 20, 10, 5, 1\}.
Figure~\ref{fig:mc_static_imbalance_ratio} illustrates the performance of the five selected algorithms classifiers for each multi-class stream. Table~\ref{tab:MC_SIR} summarizes the performance of the top 10 classifiers for each generator and their average ranking regarding each metric. For overall comparison, Figure~\ref{fig:MC_SIR_scatter} presents the overall aggregated performance of all classifiers. Axes of the ellipse represent PMAUC and Kappa metrics, the more rounded the better, and the color represents the product of both metrics.

\noindent \textbf{Discussion}

\noindent \textit{Impact of class imbalance approach.} First, we need to observe that the performance of the algorithms in multi-class problems significantly differs from the binary problems. This shows that multi-class imbalance data streams pose a series of unique challenges and thus this requires developing specific mechanisms dedicated to tackling more than two classes. Simple adaptation of binary mechanisms tends to fail and underperform, especially when dealing with a large number of classes.

For blind resampling methods, we can see a drop in their performance. \acrshort{oob} returns mediocre results, much below the ranks observed in binary scenarios. \acrshort{uob} becomes completely unusable in multi-class problems, failing to achieve any acceptable predictive power. This shows that when dealing with multiple distributions, blind resampling methods cannot capture complex relationships among classes and tend to further increase the difficulty factors (such as class overlapping or noise). This happens because blind resampling approaches consider only a single class, thus discarding valuable information about other classes. There is a need to develop novel resampling algorithms dedicated specifically to multi-class data streams. 

\acrshort{arfr} was the best algorithm regarding the Kappa metric and the second-best regarding PMAUC. Its weighing mechanism led to good robustness to multi-class imbalance ratio, as it assigns importance to every tree in the ensemble based on the class distribution (independently of the number of classes. The cost-sensitive \acrshort{csarf} displays the best performance on the G-Mean metric, yet suffers under Kappa evaluation. This shows that \acrshort{csarf} focuses on the minority classes but at the cost of suffering a larger number of false positives. The best three classifiers are based on \acrshort{arf}, showing that it is very reliable in multi-class scenarios. Among classifiers based on training modifications, only \acrshort{rose} achieved good results, demonstrating that keeping buffers for each class is a good choice for this scenario. On the other hand, \acrshort{hdvfdt} and \acrshort{ghvfdt} were among the worst. It is worth mentioning that \acrshort{calmid} and \acrshort{micfoal} were not able to outperform the mentioned classifiers, despite being specifically designed for multi-class imbalanced data streams. 

\noindent \textit{Impact of ensemble architecture.} Ensembles once again are predominant among the best performing methods for multi-class imbalanced streams. Within bagging-based methods, only \acrshort{lb} underperformed. However, \acrshort{lb} is a general-purpose ensemble, therefore it was expected not to display robustness on pair with dedicated skew-insensitive solutions. \acrshort{kue} and \acrshort{srp} could satisfactory handle static multi-class imbalance. Also, it is interesting to note that most bagging methods displayed balanced performance considering Kappa and PMAUC, demonstrating that their natural partitioning of instances contributes to a balanced performance among all classes. We have much less information on boosting-based ensembles, as only one of the examined classifiers were suitable for multi-class problems. However, this single case performed poorly, allowing us to assume that the performance of boosting-based methods will follow trends from binary scenarios. 

\begin{table*}[t!]
\vspace*{0.5cm}
\centering
\footnotesize
\setlength{\tabcolsep}{4pt}
\caption{PMAUC and Kappa on multi-class static imbalance ratio.}
\label{tab:MC_SIR}
\begin{tabular}{ll|C{1cm}C{1cm}C{1cm}C{1cm}C{1cm}C{1cm}C{1cm}C{1cm}C{1cm}C{1cm}}
\toprule
& Generator & CSARF & ARF & KUE & LB & SRP & CALMID & MICFOAL & ROSE & ARFR & OOB\\
\midrule
\multirow{3}{*}{\rotatebox[origin=c]{90}{\scalebox{.8}{PMAUC}}}
& Hyperplane & \textbf{87.48} & 77.30 & 82.75 & 76.81 & 76.24 & 81.53 & 75.59 & 85.21 & 84.91 & 87.83\\
& RandomRBF & \textbf{97.97} & 97.32 & 95.48 & 96.36 & 97.43 & 95.17 & 96.59 & 96.96 & 97.81 & 95.28\\
& RandomTree & \textbf{97.11} & 96.57 & 93.31 & 84.52 & 94.99 & 85.69 & 91.07 & 92.57 & 96.65 & 92.14\\
\midrule
\multirow{3}{*}{\rotatebox[origin=c]{90}{\scalebox{.8}{Kappa}}}
& Hyperplane & 7.63 &  6.13 &  \textbf{21.63} & 7.10 &  2.06 &  20.15 & 6.37 &  15.69 & 13.86 & 21.39\\
& RandomRBF & 82.59 & \textbf{88.27} & 82.82 & 86.04 & 87.63 & 85.57 & 87.73 & 86.97 & 88.11 & 79.89\\
& RandomTree & 76.98 & \textbf{88.55} & 74.99 & 57.06 & 71.89 & 65.27 & 80.32 & 75.26 & 88.43 & 68.63\\
\midrule
\multicolumn{2}{l|}{Avg. PMAUC} & \textbf{94.19} & 90.40 & 90.51 & 85.89 & 89.55 & 87.46 & 87.75 & 91.58 & 93.13 & 91.75\\
\multicolumn{2}{l|}{Avg. Kappa} & 55.73 & 60.98 & 59.81 & 50.06 & 53.86 & 57.00 & 58.14 & 59.30 & \textbf{63.47} & 56.64\\
\midrule
\multicolumn{2}{l|}{Rank PMAUC} & \textbf{1.33} &  4.67 &  6.00 &  8.33 &  5.33 &  8.33 &  8.00 &  4.67 &  2.67 &  5.67\\
\multicolumn{2}{l|}{Rank Kappa} & 6.33 &  3.67 &  5.00 &  7.67 &  7.00 &  6.33 &  4.67 &  4.67 & \textbf{3.00} &  6.67\\
\bottomrule
\end{tabular}
\end{table*}

\subsubsection{Dynamic imbalance ratio}
\label{sec:mc-dynamic-ir}

\begin{figure}[t!]
\vspace*{0.5cm}
\centering
\includegraphics[width=0.16\columnwidth]{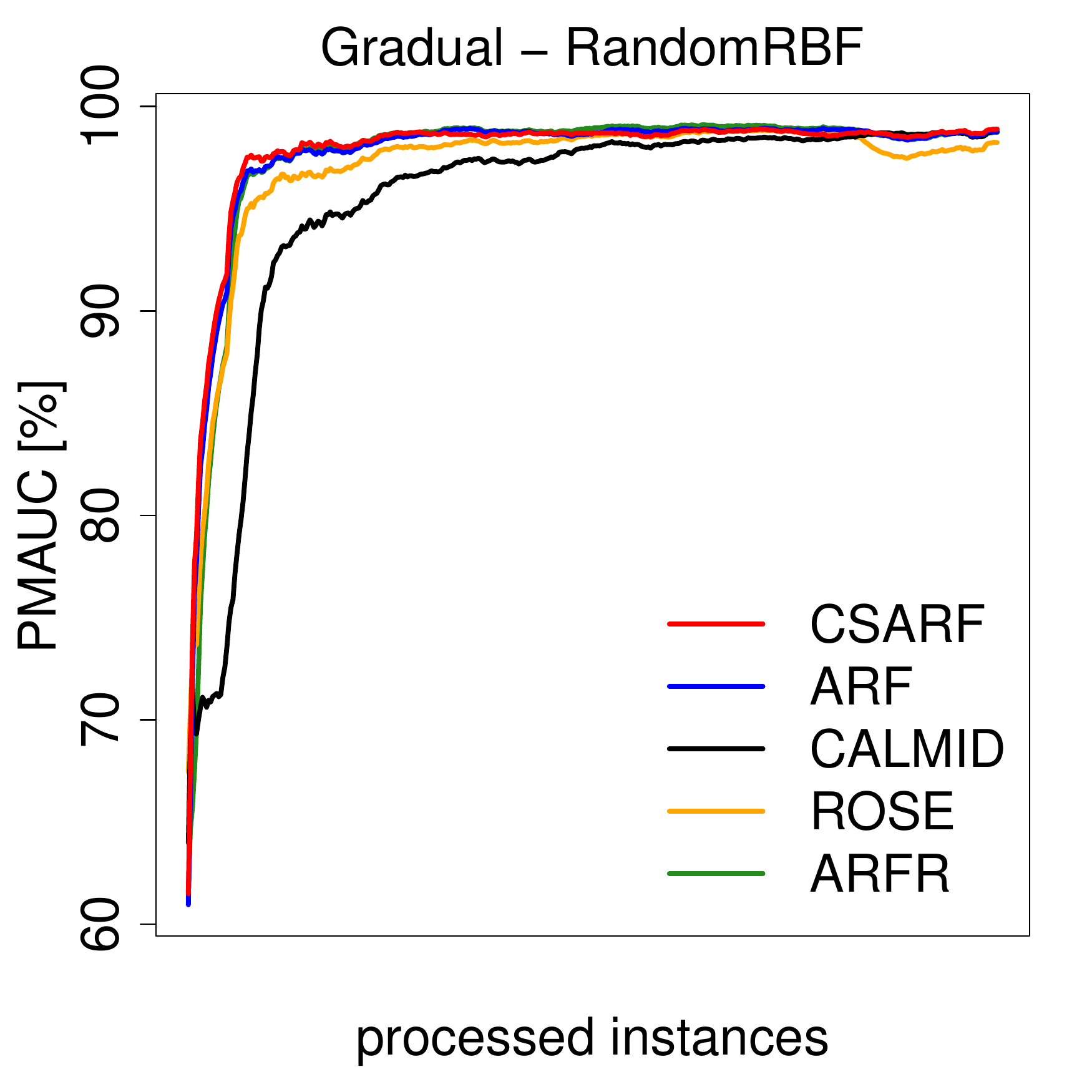}
\includegraphics[width=0.16\columnwidth]{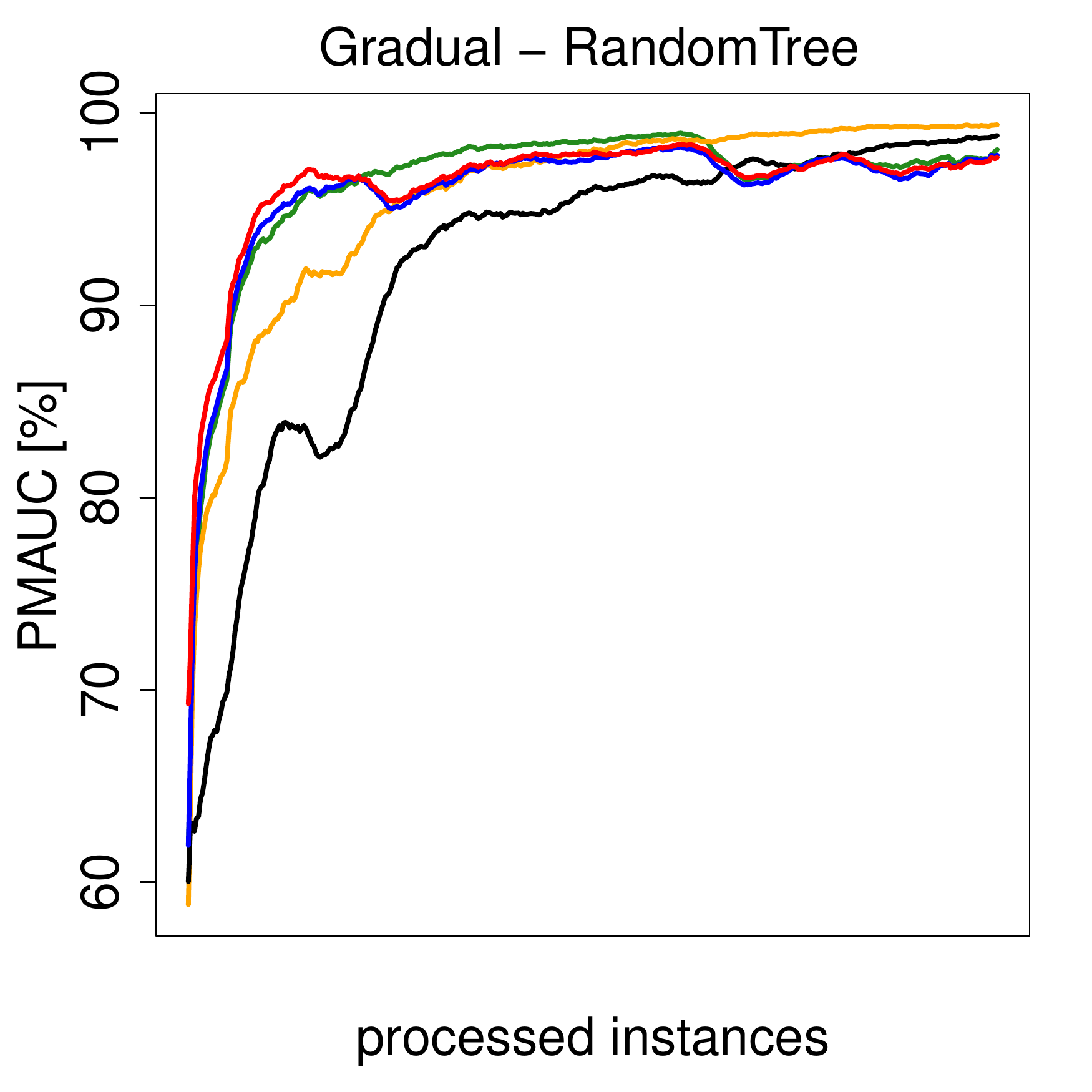}
\includegraphics[width=0.16\columnwidth]{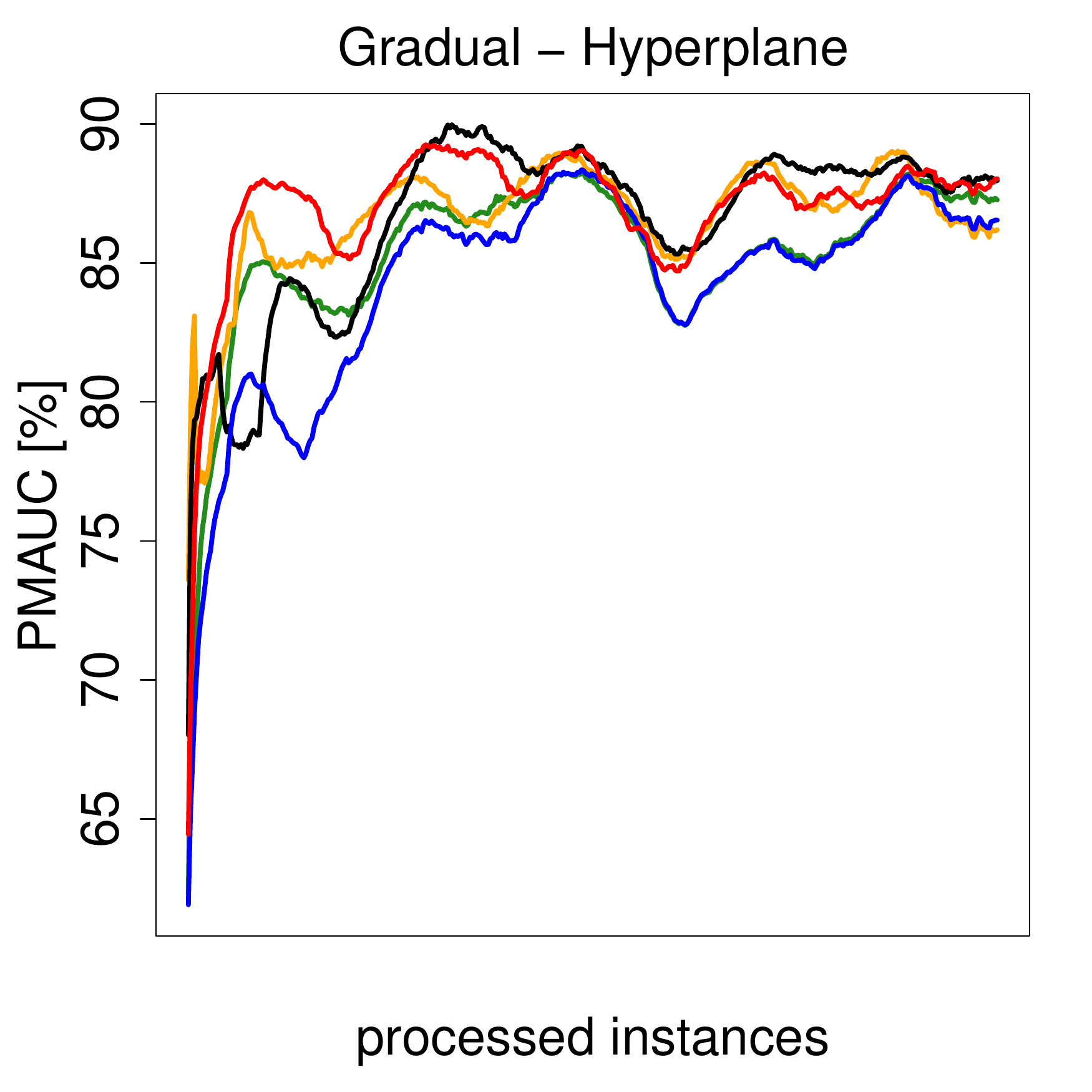}
\includegraphics[width=0.16\columnwidth]{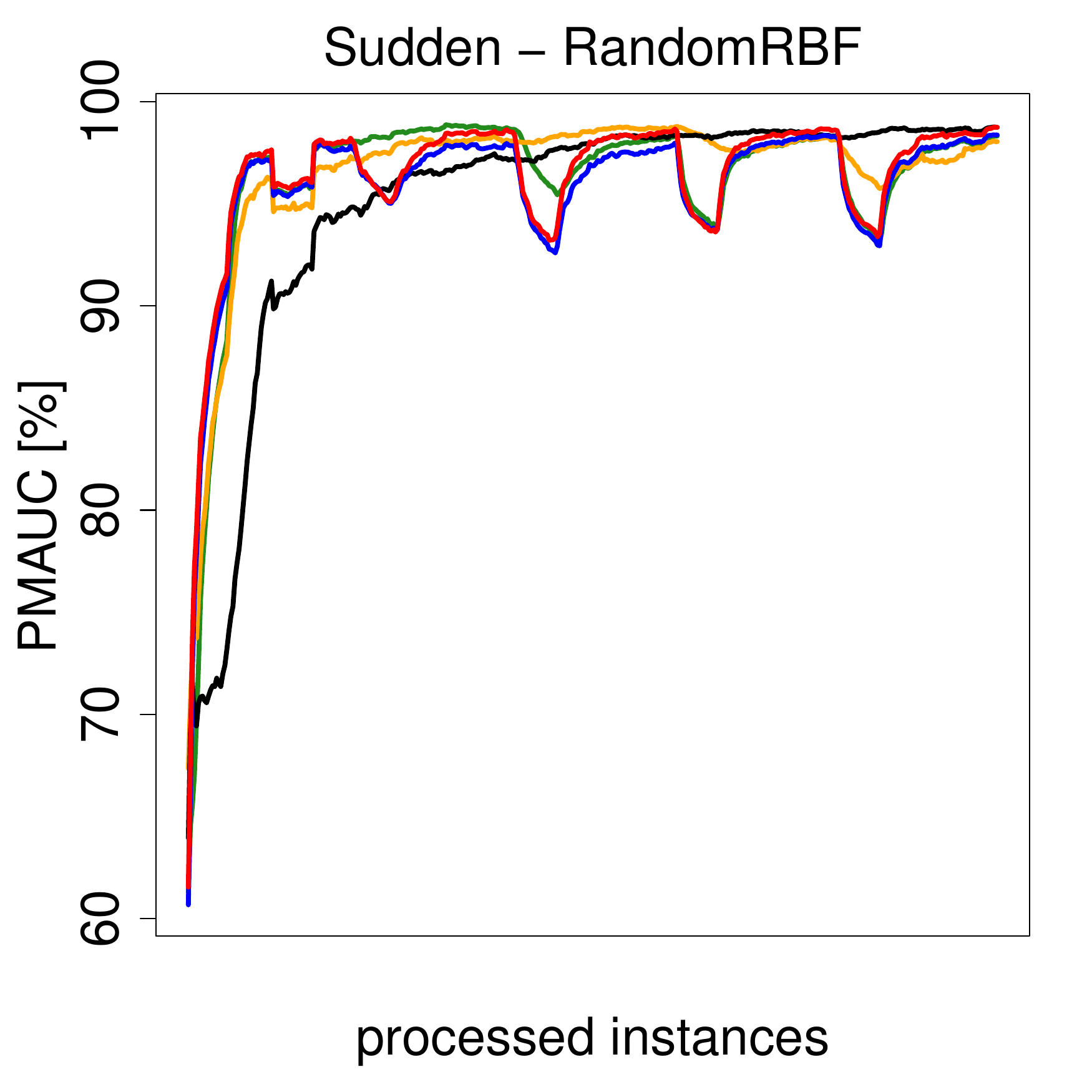}
\includegraphics[width=0.16\columnwidth]{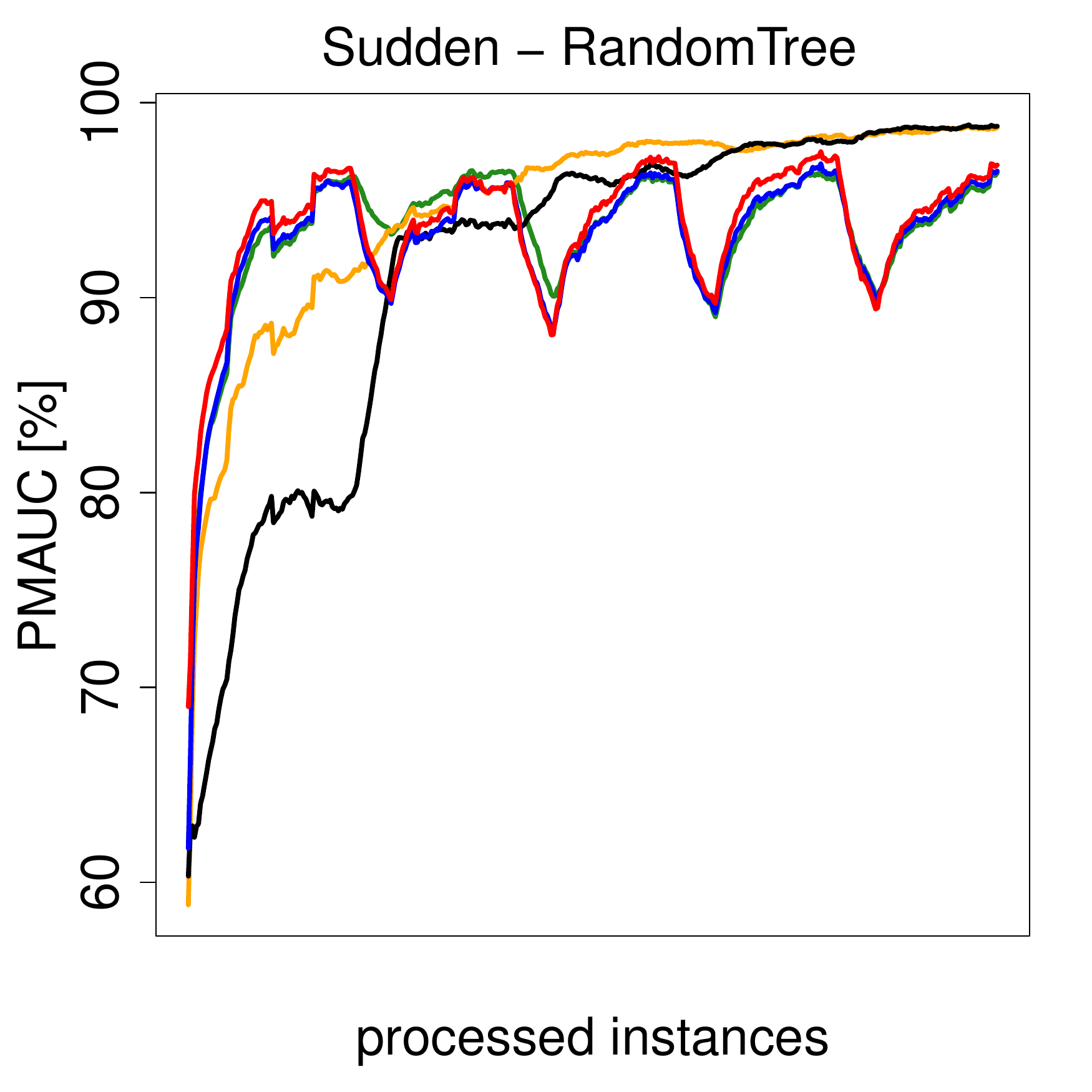}
\includegraphics[width=0.16\columnwidth]{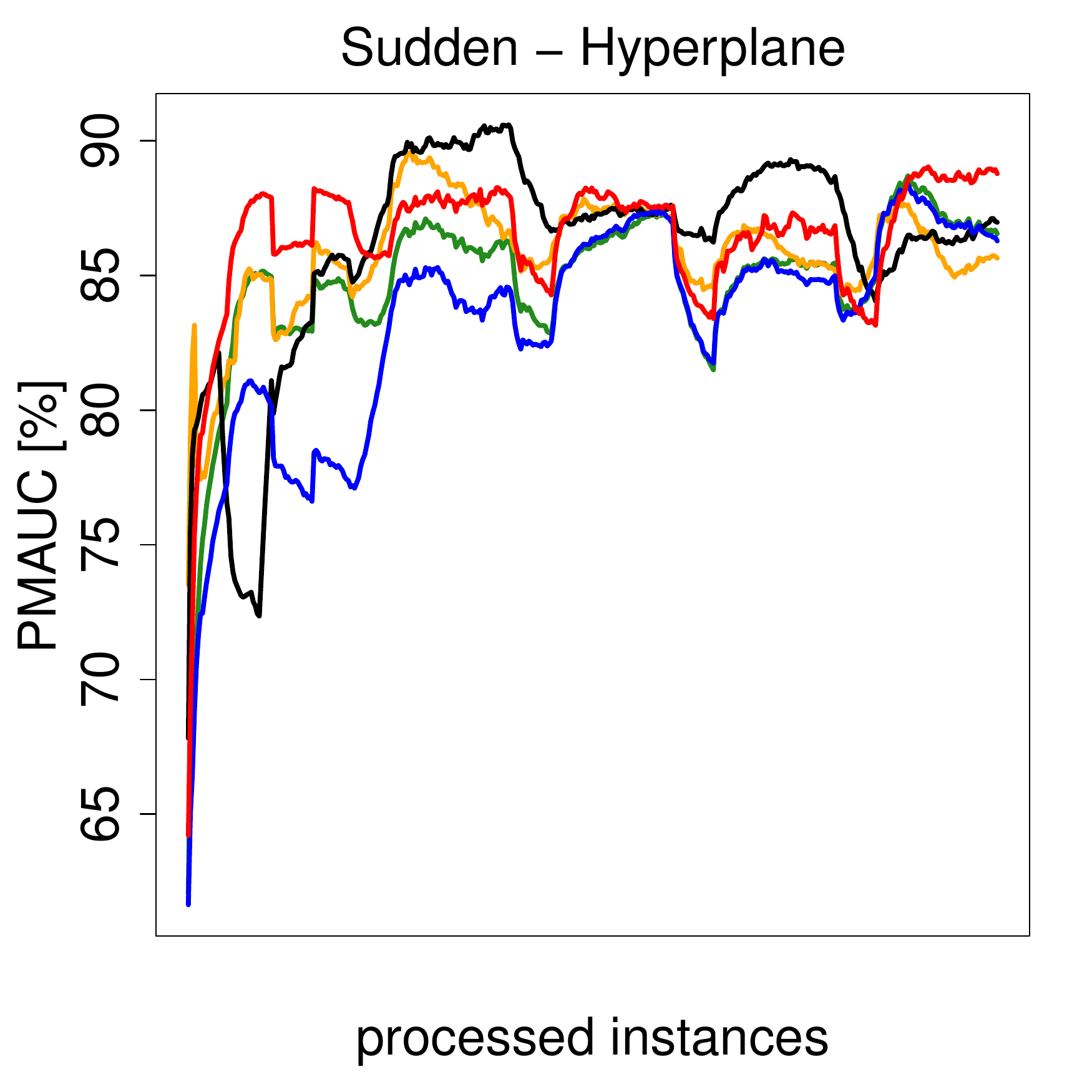}
\includegraphics[width=0.16\columnwidth]{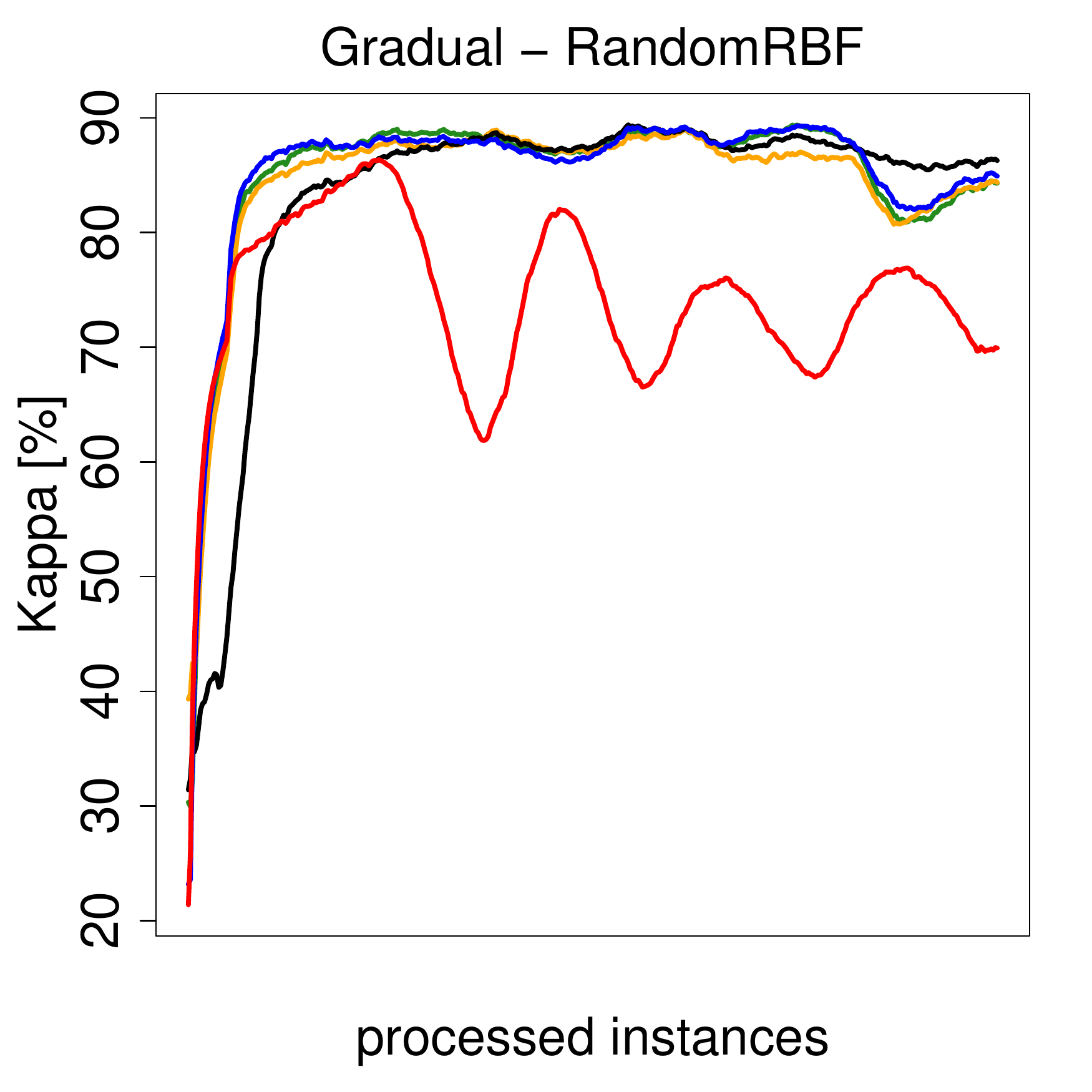}
\includegraphics[width=0.16\columnwidth]{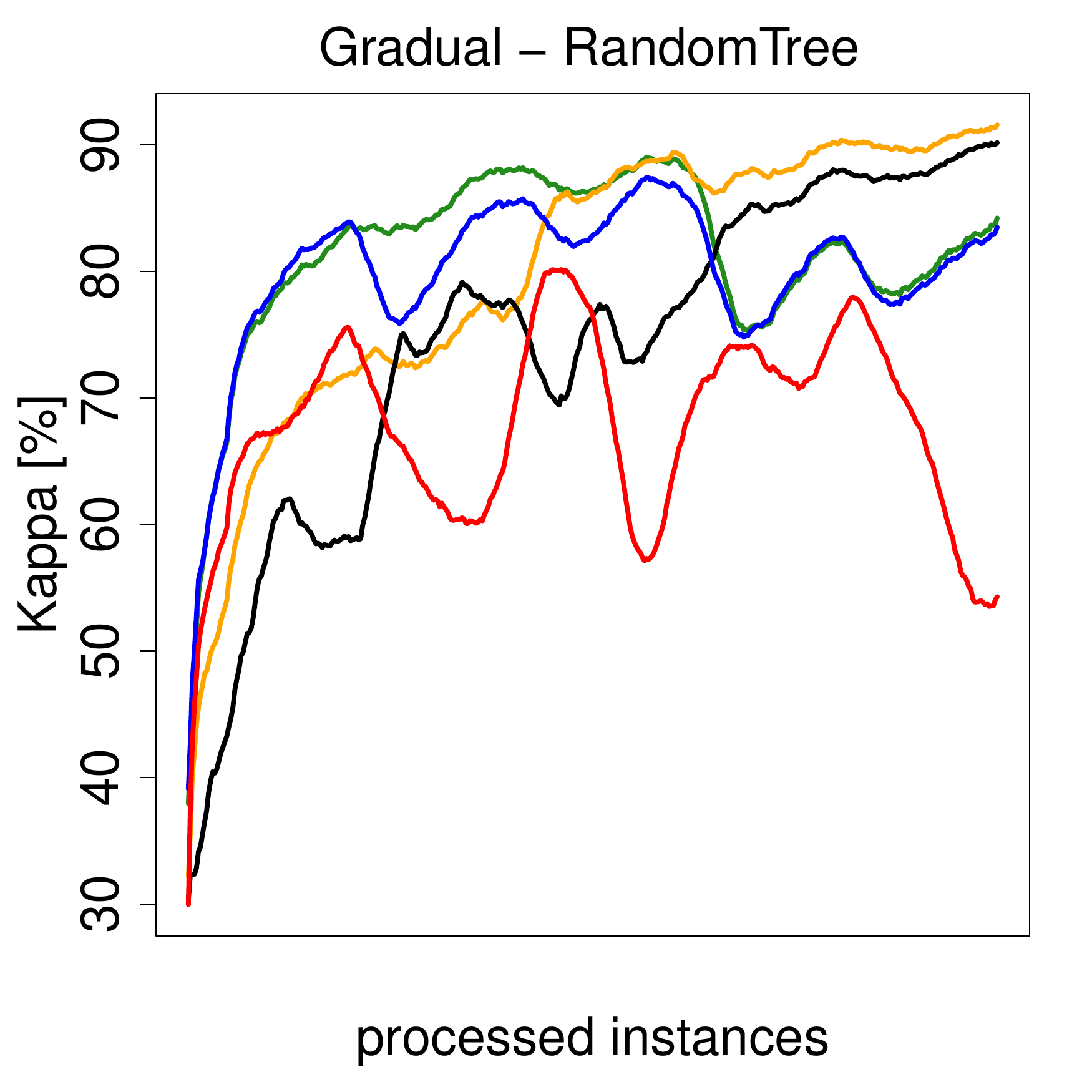}
\includegraphics[width=0.16\columnwidth]{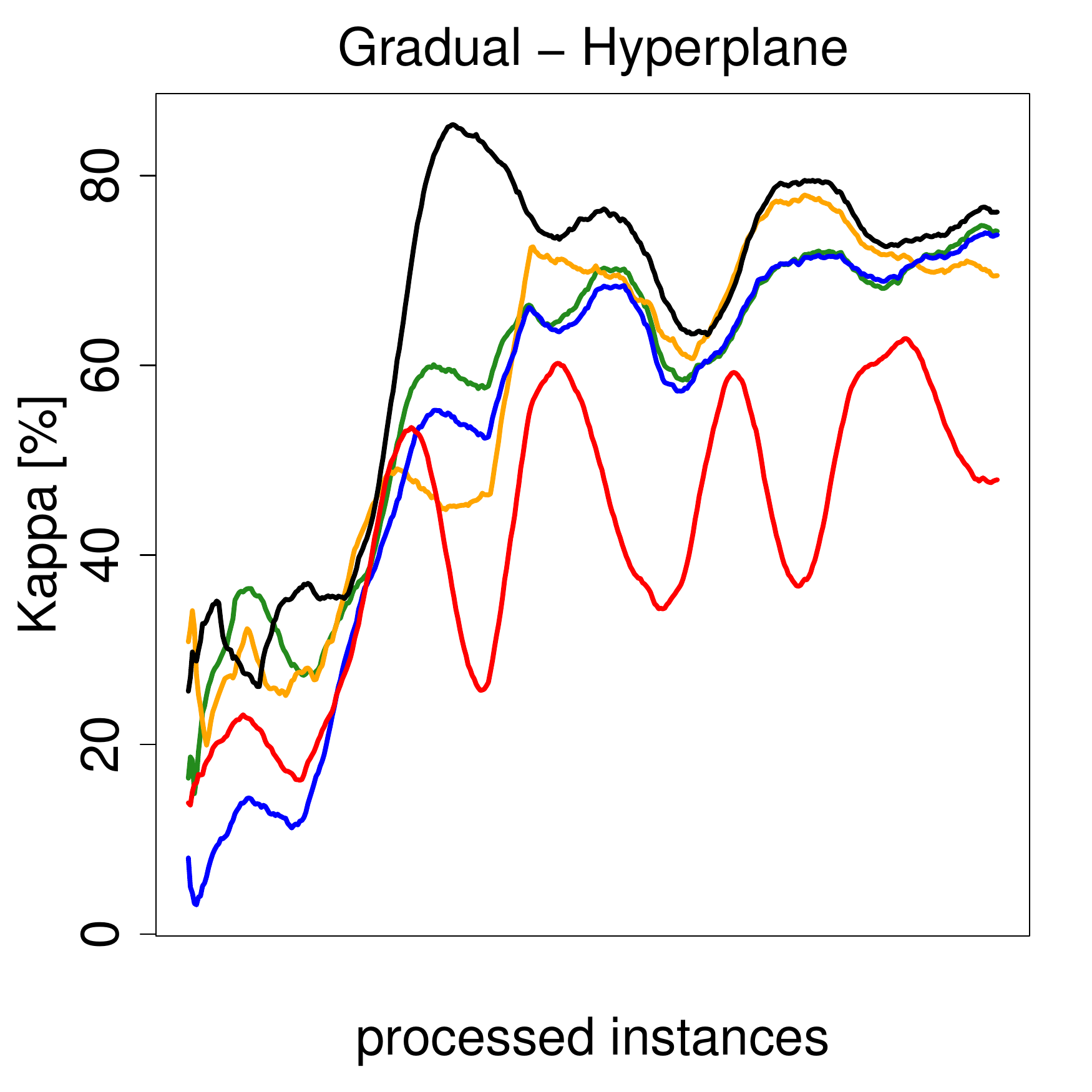}
\includegraphics[width=0.16\columnwidth]{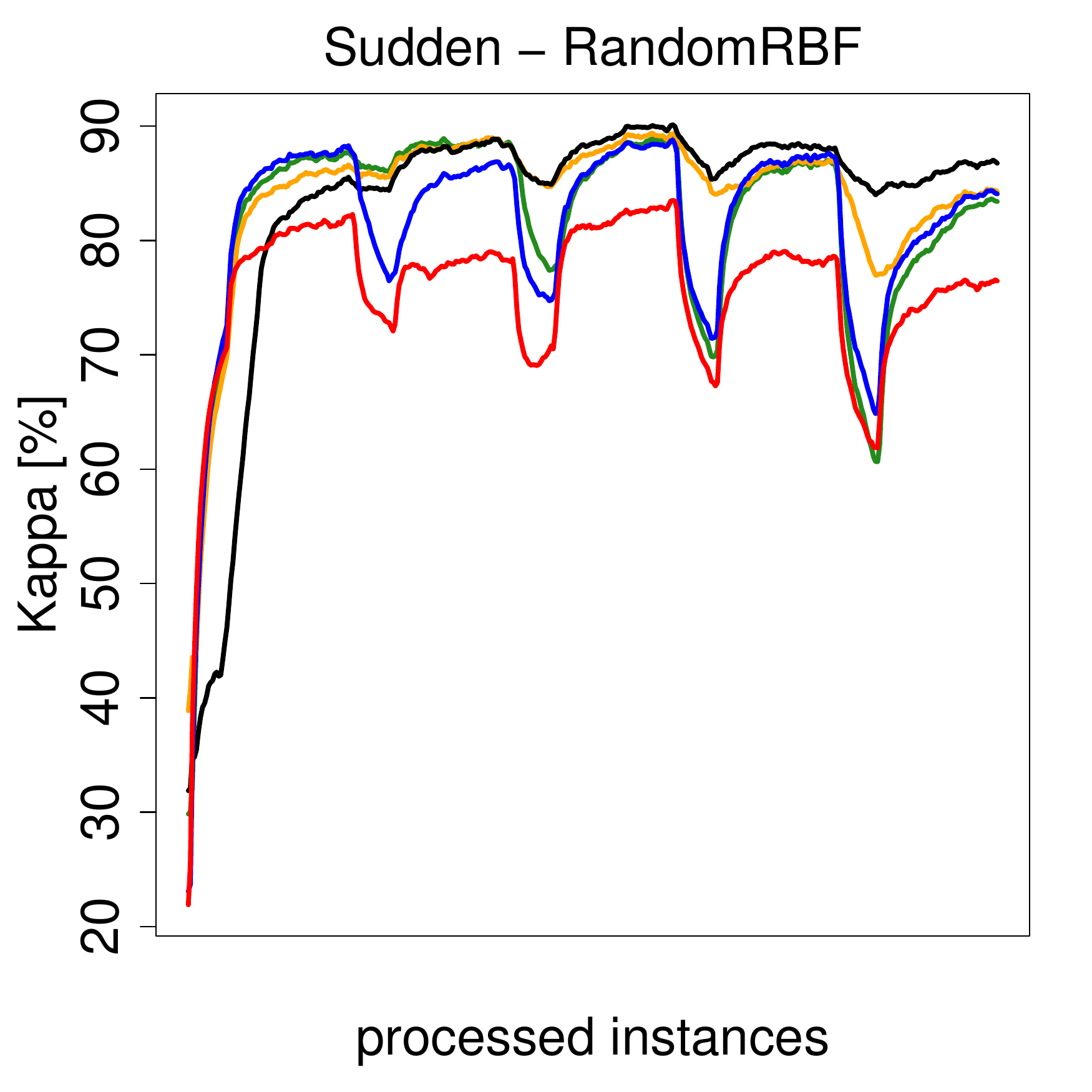}
\includegraphics[width=0.16\columnwidth]{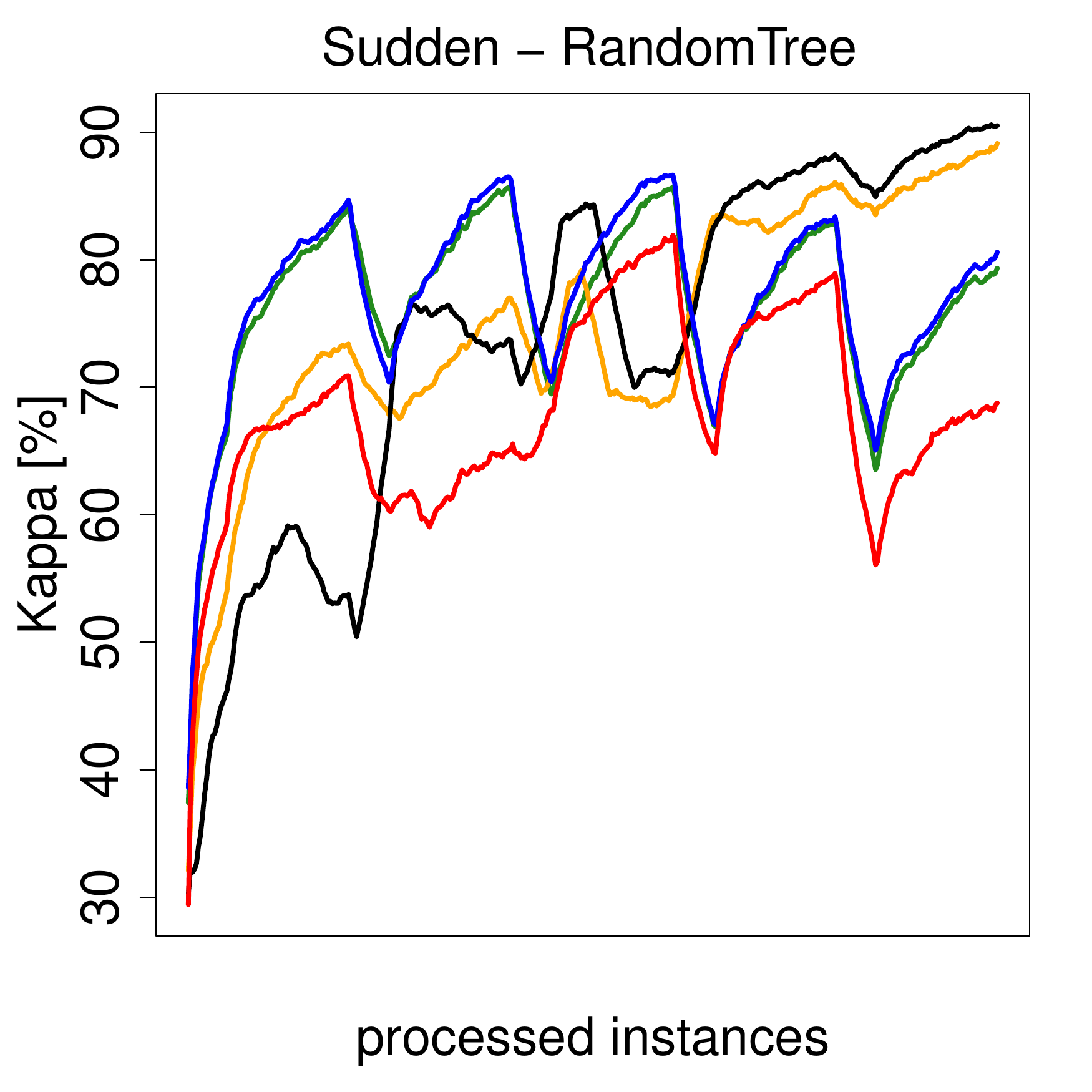}
\includegraphics[width=0.16\columnwidth]{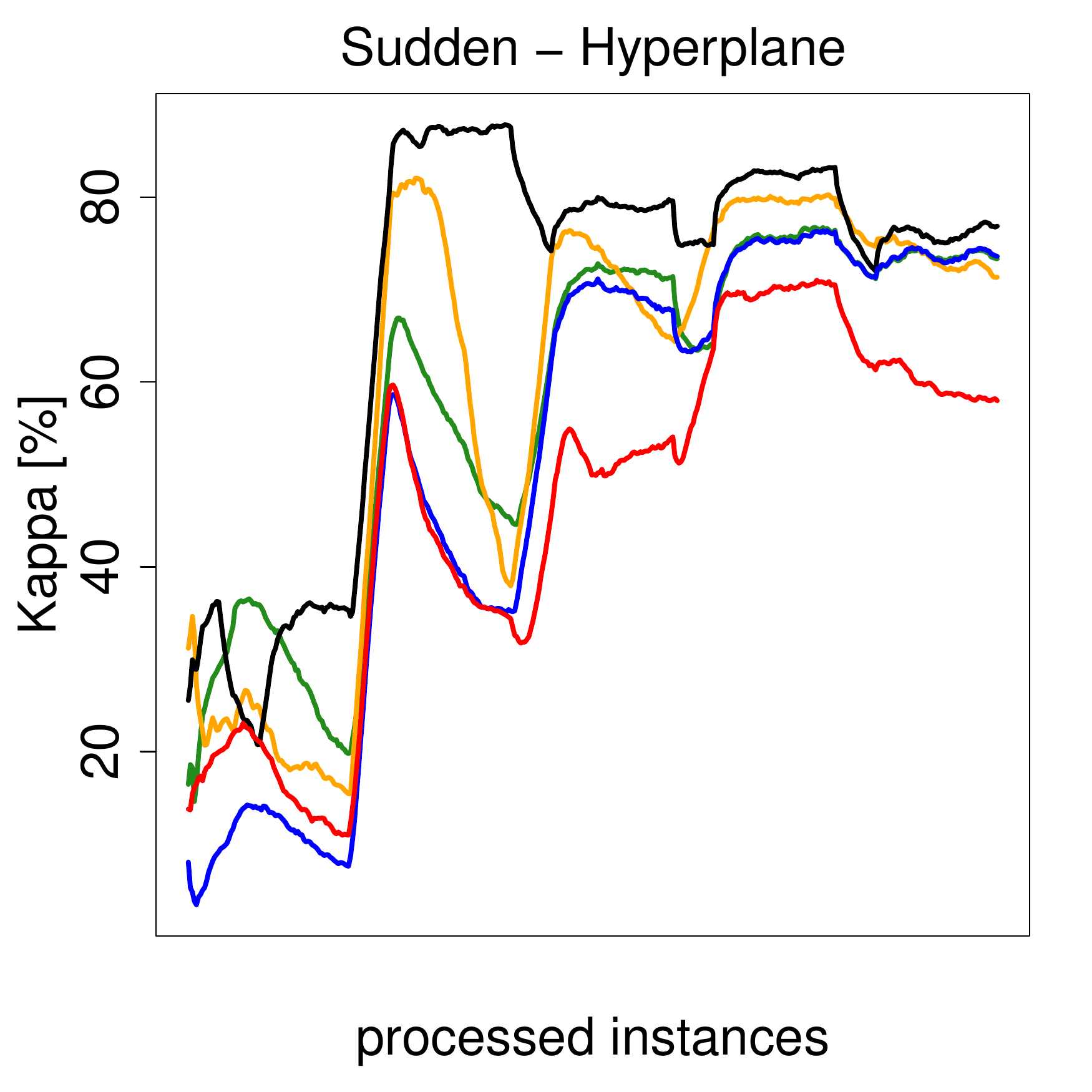}
\caption{PMAUC and Kappa on multi-class shifting imbalance ratio.}
\label{fig:mc_shifting_imbalance_ratio}
\vspace*{0.8cm}
\end{figure}

\begin{figure}[t!]
\vspace*{0.8cm}
\centering
\includegraphics[width=0.5\columnwidth]{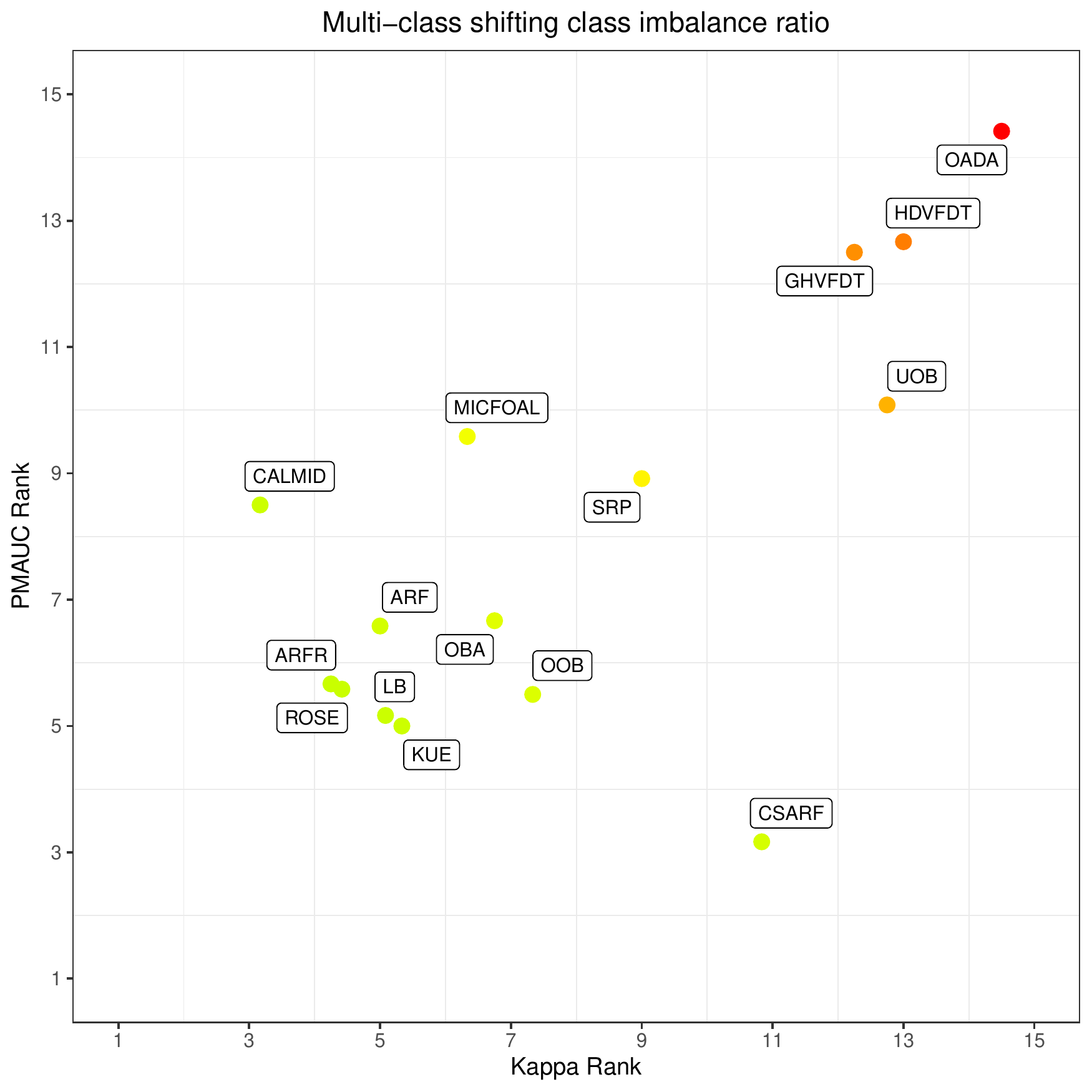}
\caption{Comparison of all 15 algorithms for multi-class shifting class imbalance ratio. Color gradient represents the product of both metrics.}
\label{fig:MC_SHF_IR_scatter}
\end{figure}

\noindent \textbf{Goal of the experiment.} This experiment was designed to complement the previous experiment and address \textbf{RQ2} to evaluate the robustness of the classifiers to dynamic changes in imbalance ratio with multiple classes. To evaluate this, we prepared three multi-class generators \{Hyperplane, RandomRBF and RandomTree\}, all of them with $5$ classes shifting imbalance ratio through the following distributions: \{\{50, 20, 10, 5, 1\}, \{20, 10, 5, 1, 50\}, \{10, 5, 1, 50, 20\}, \{5, 1, 50, 20, 10\}, \{1, 50, 20, 10, 5\}\}. The speed of the changes was evaluated both sudden and gradual. This allows us to analyze how classifiers are able to cope with dynamic imbalance ratio changes and how they are able to adapt. Figure~\ref{fig:mc_shifting_imbalance_ratio} illustrates the prequential PMAUC and Kappa for each generator over time for the selected classifiers. Table~\ref{tab:MC_SHF_IR} presents the performance for the top 10 classifiers, and their average ranking. To summarize, Figure~\ref{fig:MC_SHF_IR_scatter} shows the overall performance of all classifiers

\noindent \textbf{Discussion}

\noindent \textit{Impact of class imbalance approach.} Interestingly, the average performance of all evaluated classifiers is higher under the dynamic imbalance than under the static skewness ratio. We can explain that by the fact that evolving imbalance and class roles lead to each of the classes being the majority class for a given period of time, thus allowing for a better exposure of it to the classier, as well as countering the small sample size problem (which is a big challenge for multi-class imbalance data). 

Blind resampling methods repeated the trends observed in the previous experiment, with \acrshort{oob} returning acceptable performance and \acrshort{uob} failing to deliver predictive power. Undersampling, by reducing the size of majority class, can be enhancing the small sample size difficulty, instead of temporarily alleviating it. This prevents \acrshort{uob} from capitalizing on stats of the stream when a minority class transforms to majority one. 

Ensembles based on \acrshort{arf} maintain their very good performance and robustness to evolving imbalance ratio. \acrshort{arfr} displayed one of the best performances, showing that adding a level of resampling really enhanced the robustness to drifting class imbalance. \acrshort{csarf} exhibited great performance on the PMAUC metric, but again failed to return satisfactory Kappa. Algorithms based on training modifications like \acrshort{rose} and \acrshort{calmid} showed better robustness and ability to handle drifting imbalance ratios. \acrshort{calmid} exceeds on Kappa metric, but drops several ranks under PMAUC. \acrshort{rose} presented the most balanced results in this scenario regarding both metrics, therefore it can be seen as the most reliable and trustworthy choice for a multi-class imbalanced data stream.

\noindent \textit{Impact of ensemble architecture.} As we saw in previous experiments, bagging methods are among the best-performing, and it can be seen easily on the overall figure (Figure~\ref{fig:MC_SHF_IR_scatter}), where $7$ classifiers form a cluster in the bottom-left side of the distribution, all of them being bagging-based ensembles. Hybrid architectures presented by \acrshort{calmid} and \acrshort{micfoal} that were designed specifically for multi-class imbalanced streams significantly improve their performance when dealing with evolving imbalance ratios. 

\noindent \textit{Impact of drift speed in class imbalance.} Considering the speed of changes in class imbalance ratios, we can notice that the impact of speed is marginal on most of the classifiers. Some methods, mainly the ones without any drift handling mechanisms, have slower responses, but it did not translate to significant changes over predictive performance. PMAUC metric seems to be more sensitive to differentiation between gradual and sudden changes, while Kappa values are similar for both speeds. We can explain that by the fact that PMAUC does not consider the class imbalance ratios, thus responding differently to varying speed of changes. Kappa offers a more stable monitoring of the stream changes, not affected by the velocity of imbalance ratio evolution. 

\begin{table*}[t!]
\centering
\footnotesize
\setlength{\tabcolsep}{4pt}
\caption{PMAUC and Kappa on multi-class shifting imbalance ratio.}
\label{tab:MC_SHF_IR}
\begin{tabular}{ll|C{1cm}C{1cm}C{1cm}C{1cm}C{1cm}C{1cm}C{1cm}C{1cm}C{1cm}C{1cm}}
\toprule
& Drift & CSARF & ARF & KUE & LB & SRP & CALMID & MICFOAL & ROSE & ARFR & OOB\\
\midrule
\multirow{2}{*}{\rotatebox[origin=c]{90}{\scalebox{.6}{PMAUC}}}
& Sudden & 92.32 & 90.95 & \textbf{92.58} & 92.29 & 90.44 & 91.38 & 89.93 & 91.65 & 91.35 & 92.49\\
& Gradual & \textbf{93.85} & 92.87 & 93.04 & 92.79 & 91.88 & 91.86 & 90.94 & 93.56 & 93.26 & 93.04\\
\midrule
\multirow{2}{*}{\rotatebox[origin=c]{90}{\scalebox{.65}{Kappa}}}
& Sudden & 63.49 & 70.99 & 72.60 & 73.04 & 65.73 & \textbf{76.87} & 70.13 & 72.24 & 71.91 & 68.33\\
& Gradual & 63.81 & 73.42 & 72.42 & 71.55 & 68.26 & 74.36 & 70.61 & \textbf{75.11} & 74.97 & 70.47\\
\midrule
\multicolumn{2}{l|}{Avg. PMAUC} & \textbf{93.09} & 91.91 & 92.81 & 92.54 & 91.16 & 91.62 & 90.43 & 92.61 & 92.31 & 92.76\\
\multicolumn{2}{l|}{Avg. Kappa} & 63.65 & 72.20 & 72.51 & 72.29 & 66.99 & \textbf{75.62} & 70.37 & 73.68 & 73.44 & 69.40\\
\midrule
\multicolumn{2}{l|}{Rank PMAUC} & \textbf{2.58} &  5.58 &  4.42 &  4.50 &  7.42 &  7.50 &  7.92 &  4.83 &  4.92 &  5.33\\
\multicolumn{2}{l|}{Rank Kappa} & 9.50 &  4.50 &  5.00 &  4.67 &  8.00 &  \textbf{3.08} &  5.58 &  4.00 &  3.92 &  6.75\\
\bottomrule
\end{tabular}
\end{table*}

\newpage
\subsubsection{Concept drift and static imbalance ratio}
\label{sec:mc-cd-static-ir}

\begin{figure}[t!]
\vspace*{0.5cm}
\centering
\includegraphics[width=0.19\columnwidth]{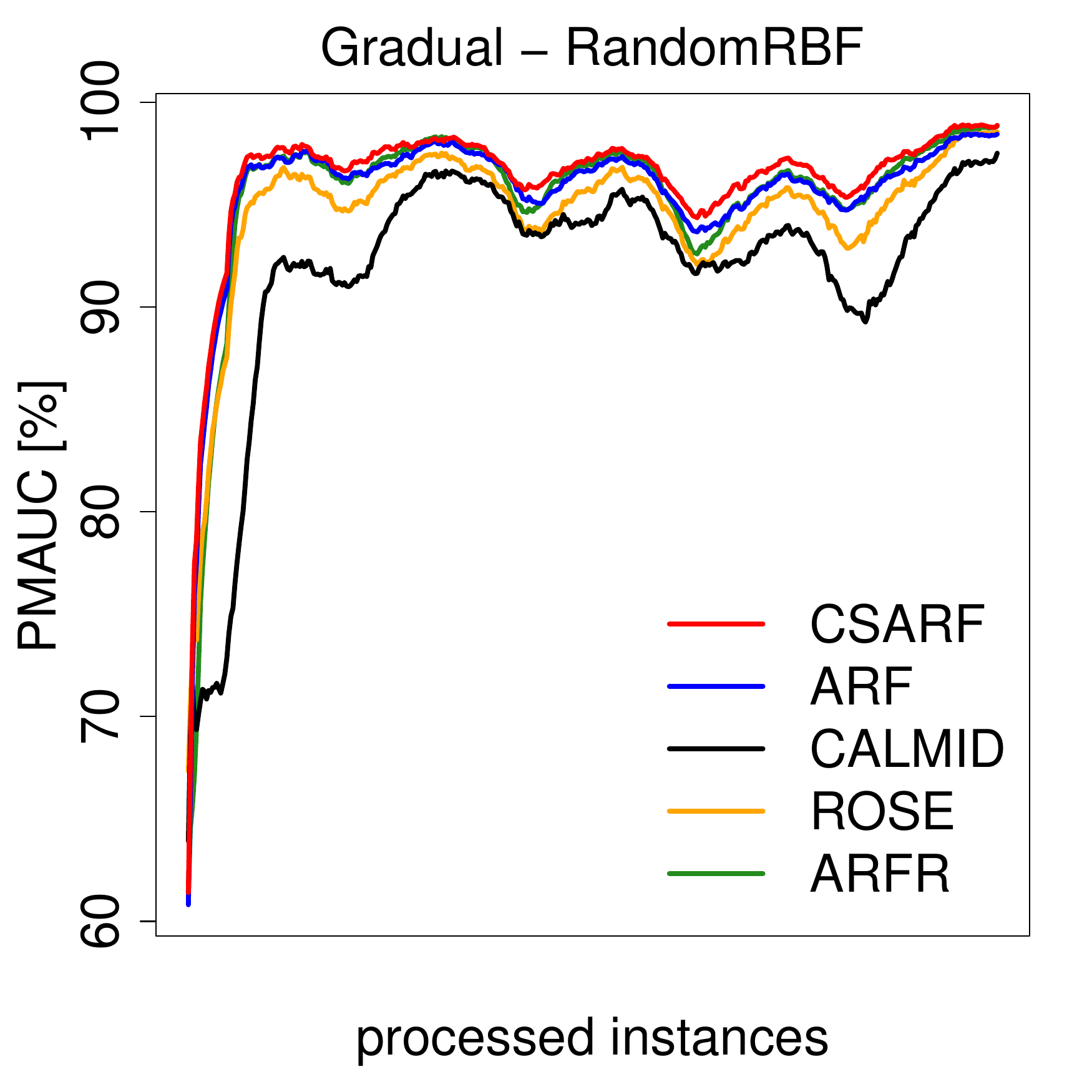}
\includegraphics[width=0.19\columnwidth]{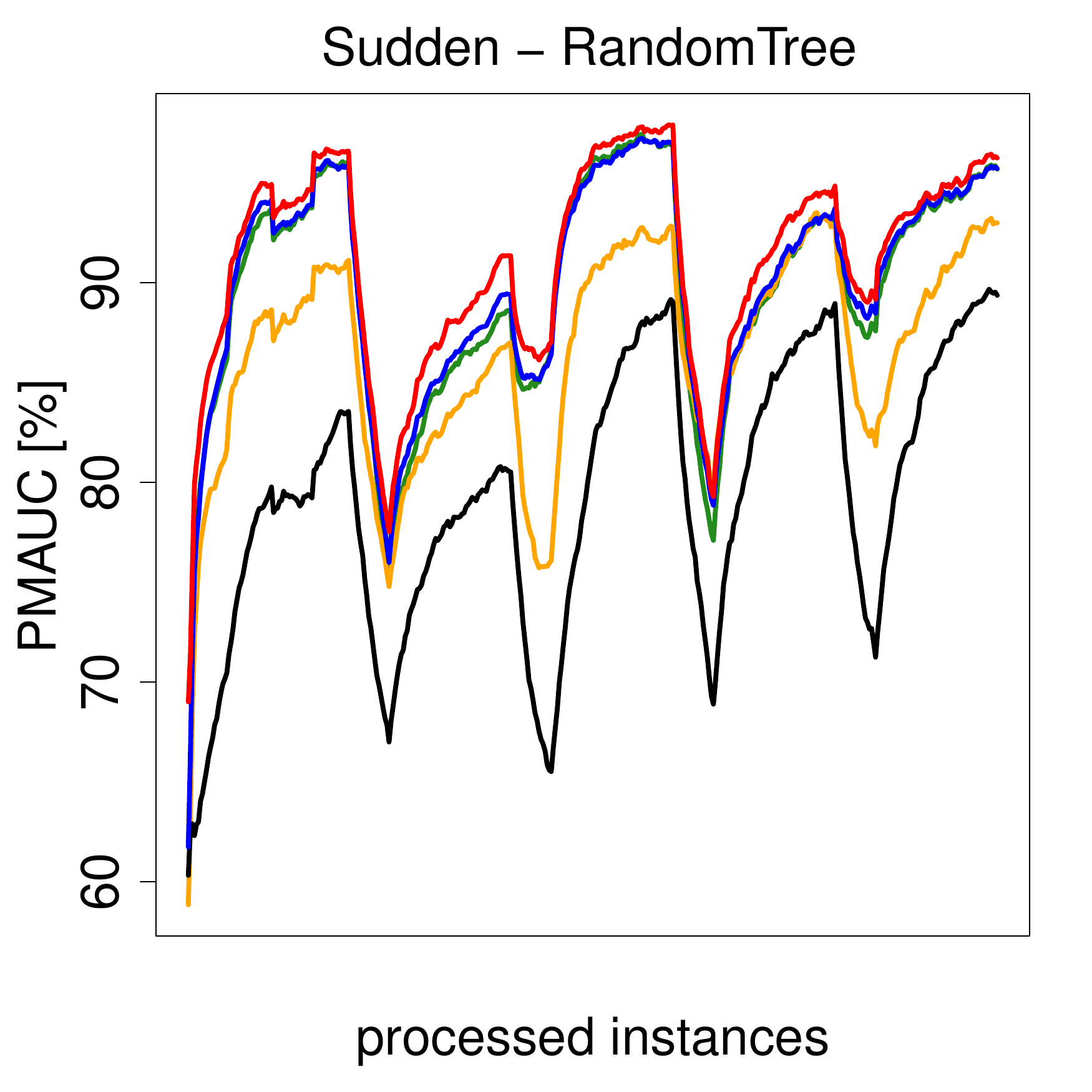}
\includegraphics[width=0.19\columnwidth]{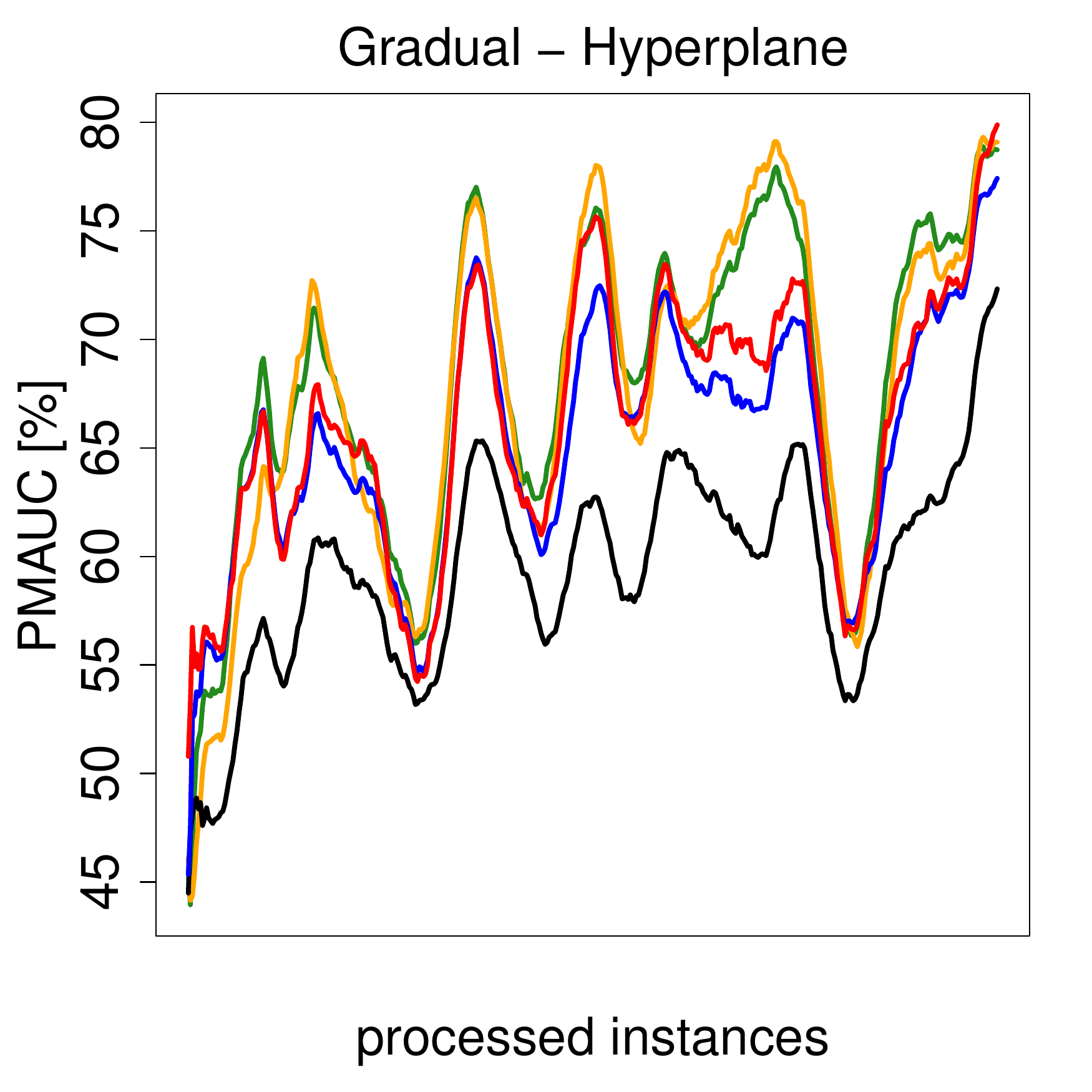}
\includegraphics[width=0.19\columnwidth]{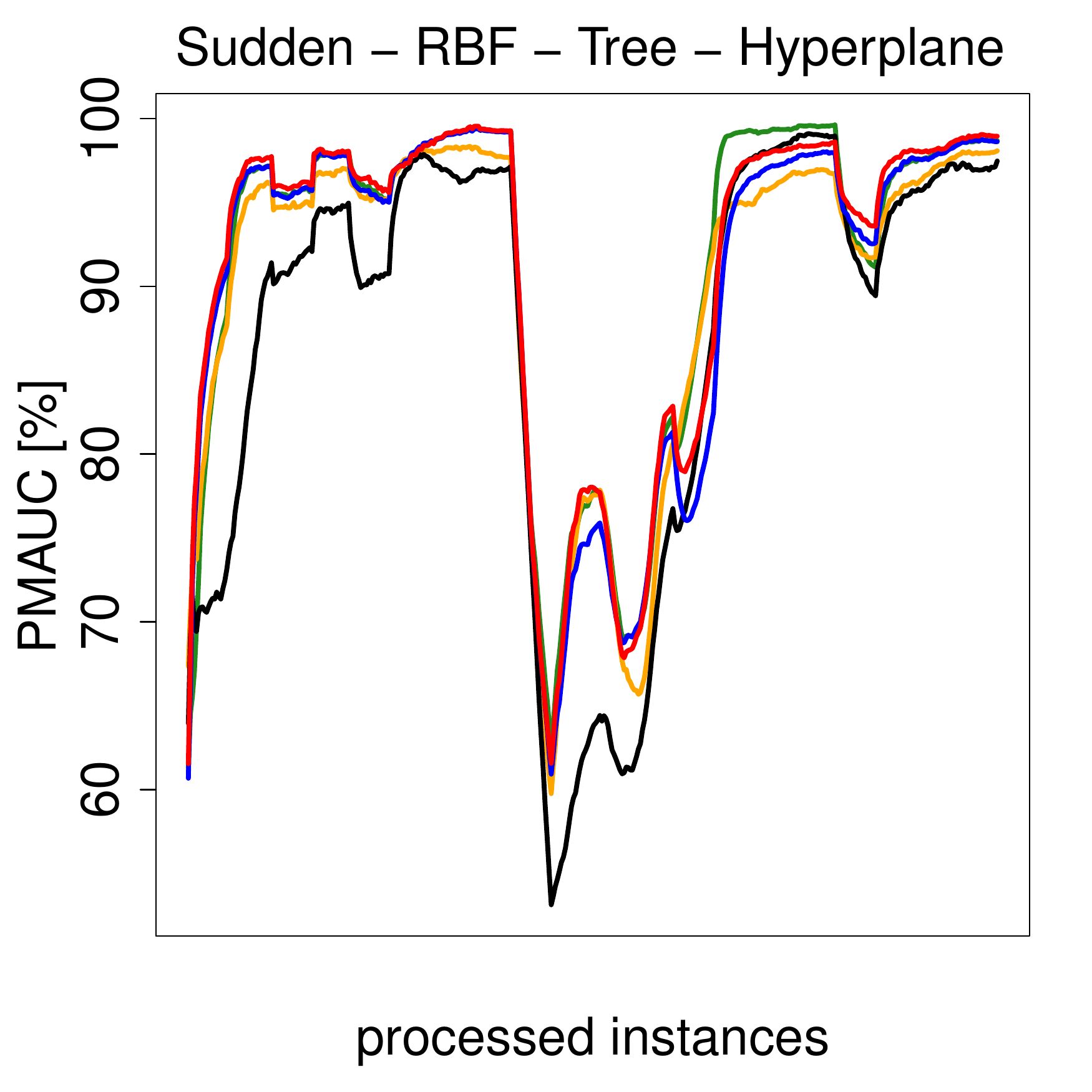}
\\
\includegraphics[width=0.19\columnwidth]{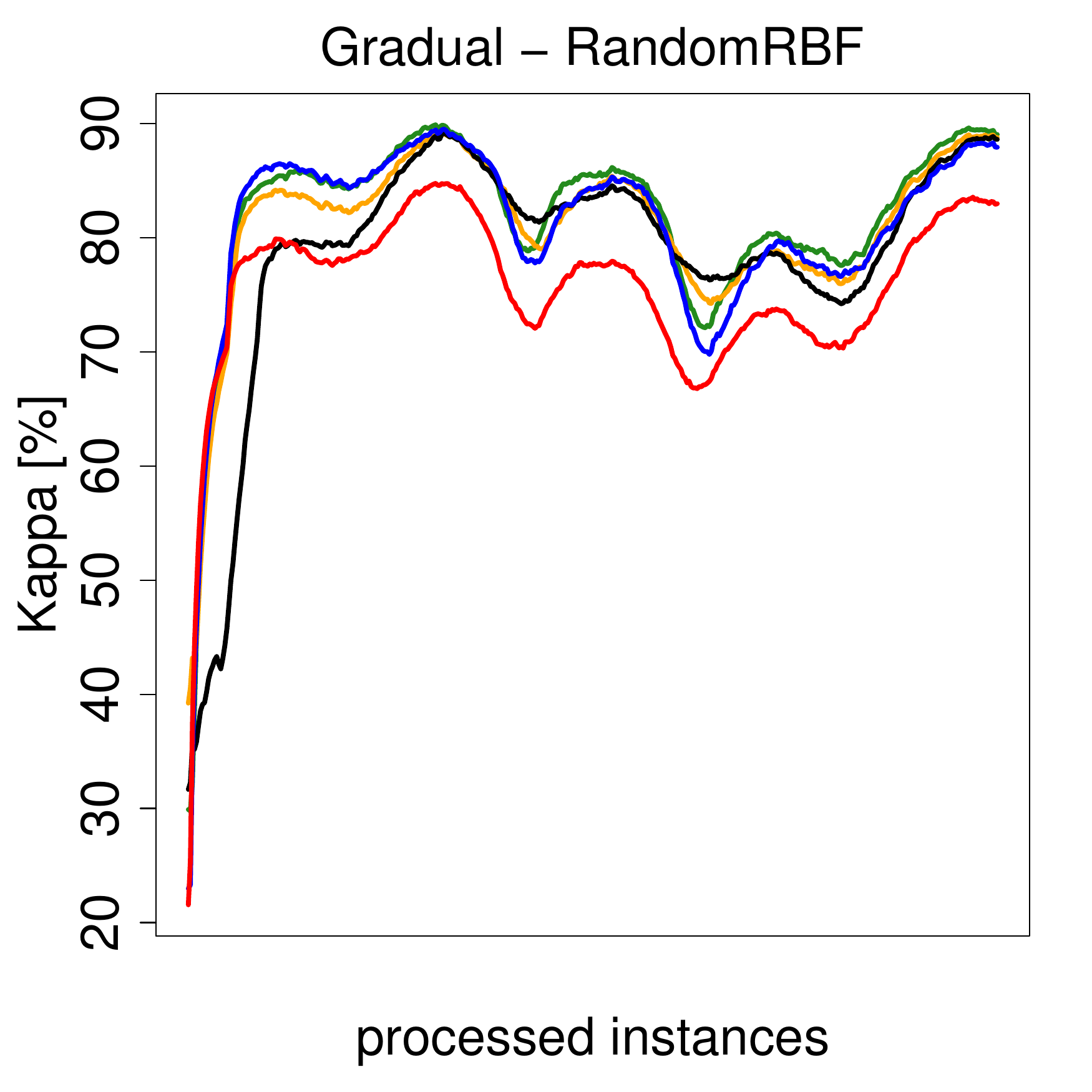}
\includegraphics[width=0.19\columnwidth]{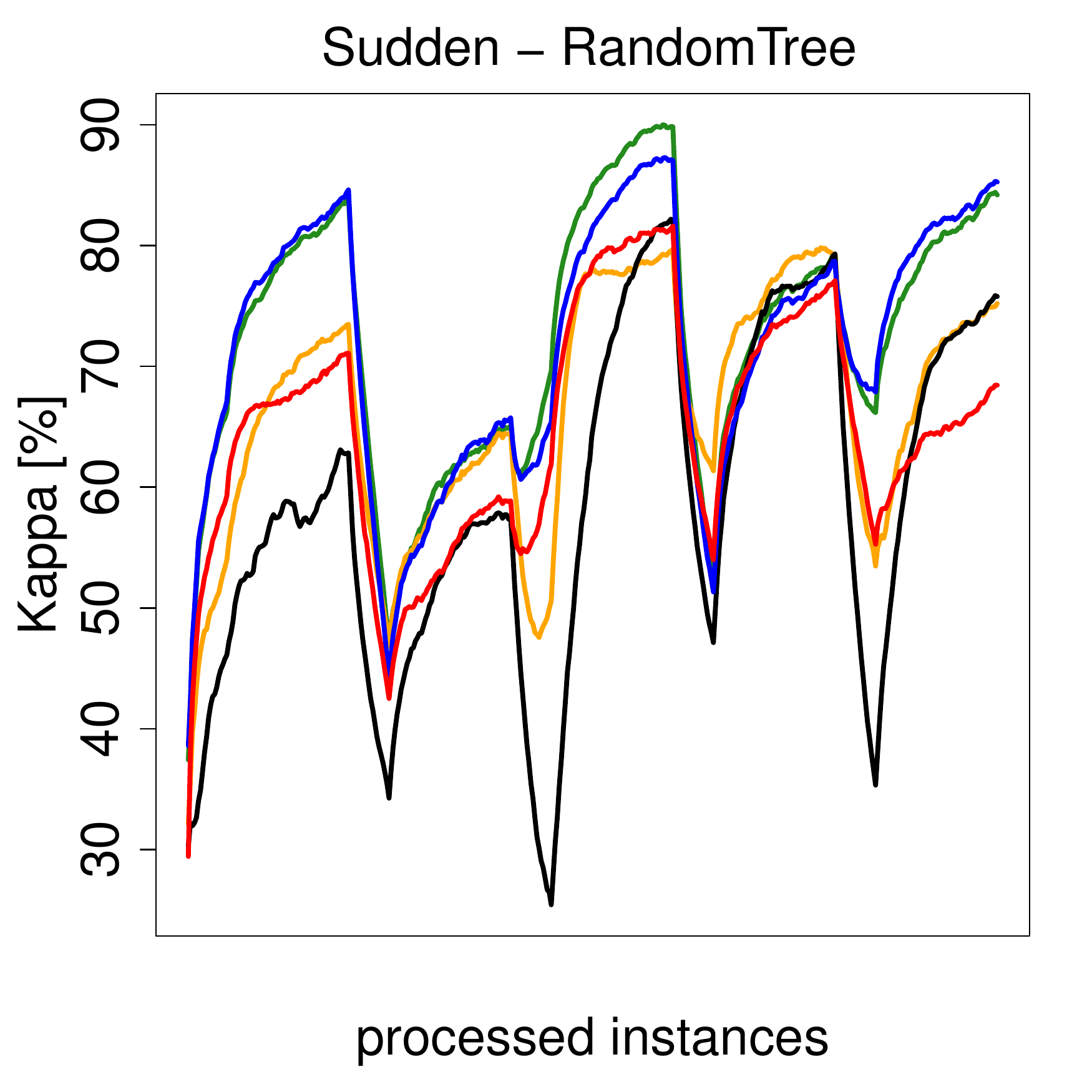}
\includegraphics[width=0.19\columnwidth]{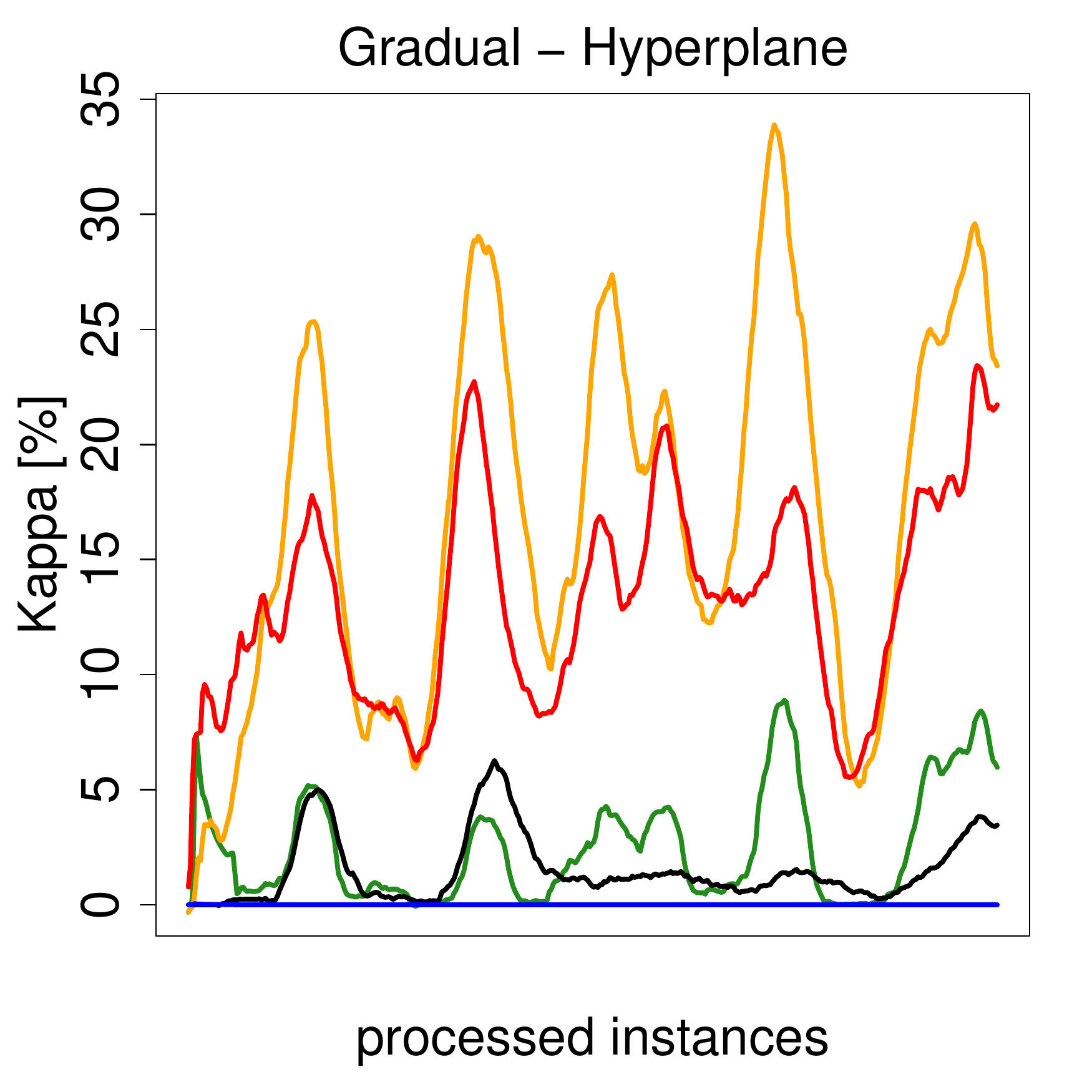}
\includegraphics[width=0.19\columnwidth]{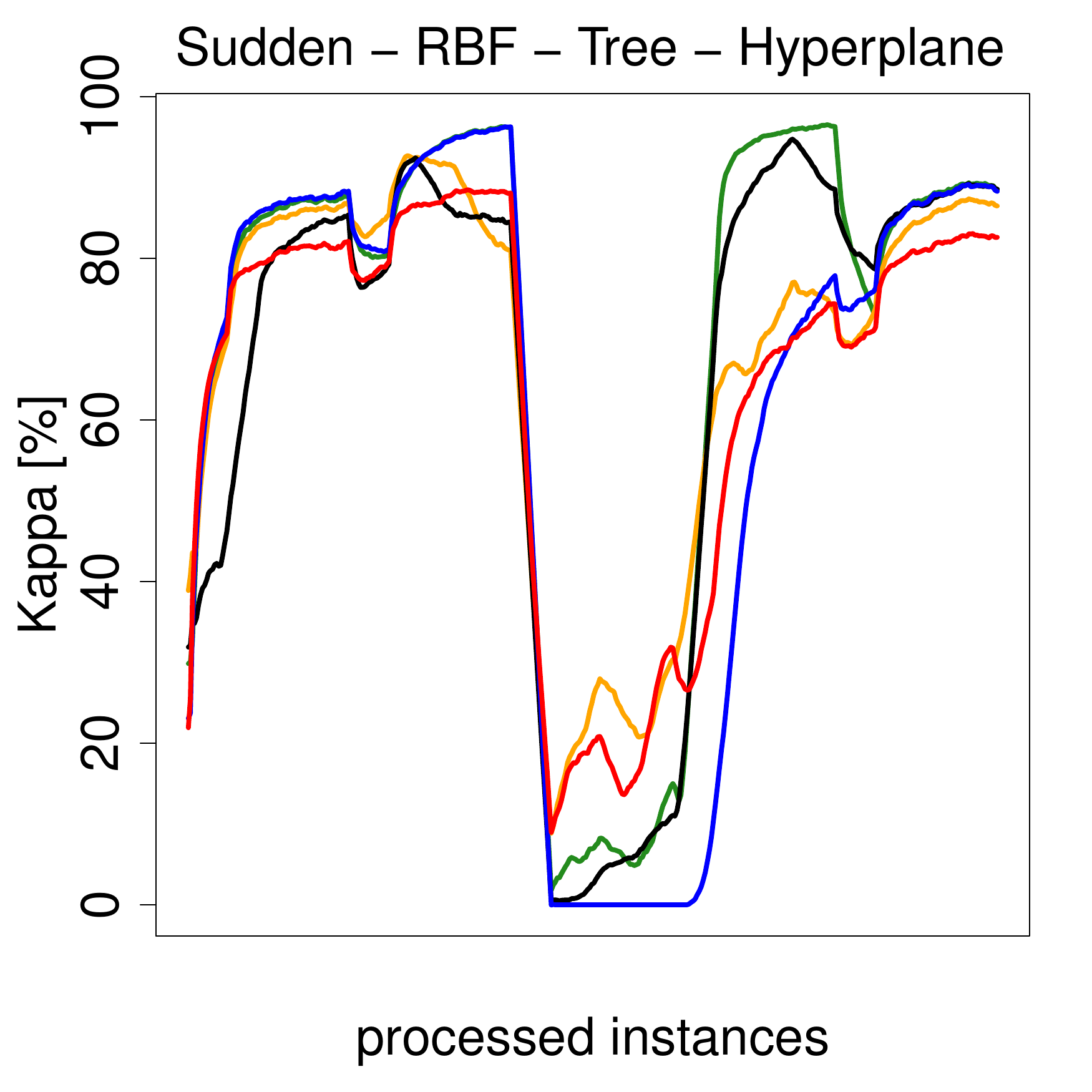}
\caption{PMAUC and Kappa on concept drift and multi-class static imbalance ratio.}
\label{fig:mc_cd_static_imbalance_ratio}
\end{figure}

\begin{figure}[t!]
\vspace*{0.8cm}
\centering
\includegraphics[width=0.5\columnwidth]{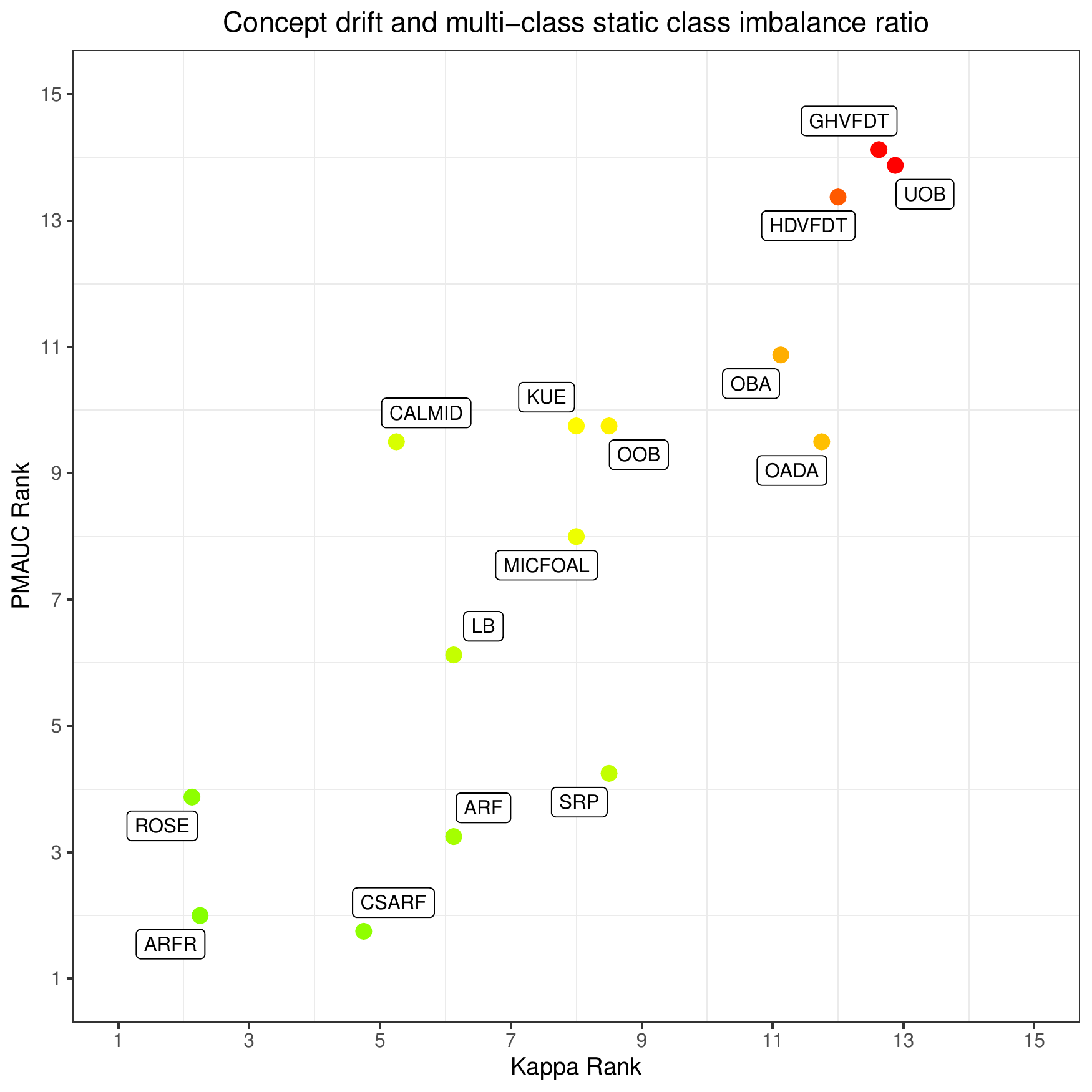}
\caption{Comparison of all 15 algorithms on concept drift and multi-class static class imbalance ratio. Color gradient represents the product of both metrics.}
\label{fig:MC_CD_SIR_scatter}
\end{figure}

\noindent \textbf{Goal of the experiment.} This experiment was designed to complement previous experiments and address \textbf{RQ2} and \textbf{RQ4} and to evaluate the behavior of the classifiers in a scenario with multiple classes in the presence of concept drift and static imbalance ratio. Concept drift leads to changes in decision boundaries, creating a challenge for classifiers to cope with and react to change. To evaluate this, we prepared three streams generators similarly to experiment \ref{sec:mc-static-IR}, plus a concatenation of all three streams, and introduced concept drifts along the stream gradually or suddenly. Figure~\ref{fig:mc_cd_static_imbalance_ratio} presents the performance of the five selected classifiers for each evaluated drifting stream. Table~\ref{tab:MC_CD_SIR} provides the PMAUC and Kappa for the top 10 classifiers for both types of drift and their average value and ranking. Figure~\ref{fig:MC_CD_SIR_scatter} illustrates the overall performance for all classifiers. 

\noindent \textbf{Discussion}

\noindent \textit{Impact of class imbalance approach.} Concept drift poses an increased difficulty in multi-class scenarios, as it changes complex relationships among classes. Multi-class problems tend to have much more complex decision boundaries than their binary counterparts and thus adaptation to drift requires more training instances or increased amount of time. 

When analyzing resampling-based approaches, we can observe a significant drop in predictive power for both \acrshort{oob} and \acrshort{uob}. We already established that \acrshort{uob} is incapable of handling multi-class problems, but the additional presence of concept drift positions it among the worst performing classifiers. This follows our observations from the binary experiments, where we showed that lack of explicit or implicit drift adaptation mechanisms in \acrshort{oob} and \acrshort{uob} inhibits their learning capabilities from non-stationary data. \acrshort{arfr} returned best results among resampling-based algorithms, being at the same time competitive with other top performing classifiers.

The basic version of \acrshort{arf} displayed a loss of performance, showing that this algorithm cannot handle well changes appearing in multiple classes at once, especially when these classes are skewed. Its cost-sensitive modification maintained the very good performance observed in previous experiments, additionally improving under Kappa metric. This shows that \acrshort{csarf} is capable of an efficient adaptation to concept drift. \acrshort{calmid} and \acrshort{micfoal} displayed good results, being methods natively designed for multi-class scenarios. \acrshort{rose} was among the best performing algorithms, without relying on resampling or cost-sensitive modifications. This shows that \acrshort{rose} mechanisms, mainly effective classifier replacement and class-based buffers, allow for an improved robustness in drifting and imbalanced multi-class scenarios.

\noindent \textit{Impact of ensemble architecture.} Once again bagging-based and hybrid architectures tend to dominate the experimental study. Even methods such as \acrshort{lb} and \acrshort{srp} returned decent results, despite their lack of skew-insensitive mechanisms. This shows that well-designed drift adaptation goes a long way in every streaming scenario and that bagging-based architectures can utilize their diversity to better anticipate the drift occurrence. Two exceptions to this rule are \acrshort{kue}, which as we observed in binary class cannot perform well with concept drift and imbalance, and \acrshort{uob} that does not adapt well to multi-class imbalance.

\noindent \textit{Impact of concept drift speed.} We can see that the speed of concept drift does not significantly affect the results of individual classifiers. However, we can see different behavior of the metrics as compared to the previous experiment. Here Kappa reacts differently to gradual and sudden drifts, showing that the speed of evolution of class boundaries can be picked up by Kappa analysis. This allows us to conclude that when concept drift is combined with imbalance, both PMAUC and Kappa become sensitive to speed of changes.

\begin{table*}[t!]
\vspace*{0.8cm}
\centering
\footnotesize
\setlength{\tabcolsep}{4pt}
\caption{PMAUC and Kappa on concept drift and multi-class static imbalance ratio.}
\label{tab:MC_CD_SIR}
\begin{tabular}{ll|C{1cm}C{1cm}C{1cm}C{1cm}C{1cm}C{1cm}C{1cm}C{1cm}C{1cm}C{1cm}}
\toprule
& Drift & CSARF & ARF & KUE & LB & SRP & CALMID & MICFOAL & ROSE & ARFR & OOB\\
\midrule
\multirow{2}{*}{\rotatebox[origin=c]{90}{\scalebox{.6}{PMAUC}}}
& Sudden & \textbf{87.02} & 86.26 & 77.49 & 83.35 & 84.82 & 80.19 & 81.02 & 85.37 & 86.94 & 77.42\\
& Gradual & 84.02 & 83.29 & 76.84 & 81.12 & 82.42 & 78.06 & 79.00 & 83.41 & \textbf{84.22} & 77.01\\
\midrule
\multirow{2}{*}{\rotatebox[origin=c]{90}{\scalebox{.65}{Kappa}}}
& Sudden & 56.12 & 54.82 & 46.27 & 52.28 & 49.92 & 52.30 & 49.49 & \textbf{59.32} & 58.12 & 41.93\\
& Gradual & 49.05 & 46.12 & 42.37 & 47.00 & 42.97 & 46.65 & 43.16 & \textbf{55.05} & 50.32 & 39.94\\
\midrule
\multicolumn{2}{l|}{Avg. PMAUC} & 85.52 & 84.77 & 77.16 & 82.24 & 83.62 & 79.12 & 80.01 & 84.39 & \textbf{85.58} & 77.22\\
\multicolumn{2}{l|}{Avg. Kappa} & 52.58 & 50.47 & 44.32 & 49.64 & 46.45 & 49.48 & 46.32 & \textbf{57.18} & 54.22 & 40.93\\
\midrule
\multicolumn{2}{l|}{Rank PMAUC} & \textbf{1.75} &  3.25 &  8.75 &  6.00 &  4.25 &  8.75 &  7.50 &  3.88 &  2.00 &  8.88\\
\multicolumn{2}{l|}{Rank Kappa} & 4.75 &  5.00 &  7.75 &  5.38 &  7.25 &  5.25 &  7.00 &  \textbf{2.13} &  2.25 &  8.25\\
\bottomrule
\end{tabular}
\end{table*}

\newpage
\subsubsection{Concept drift and dynamic imbalance ratio}
\label{sec:mc-cd-dynamic-ir}

\begin{figure}[t!]
\vspace*{0.5cm}
\centering
\includegraphics[width=0.19\columnwidth]{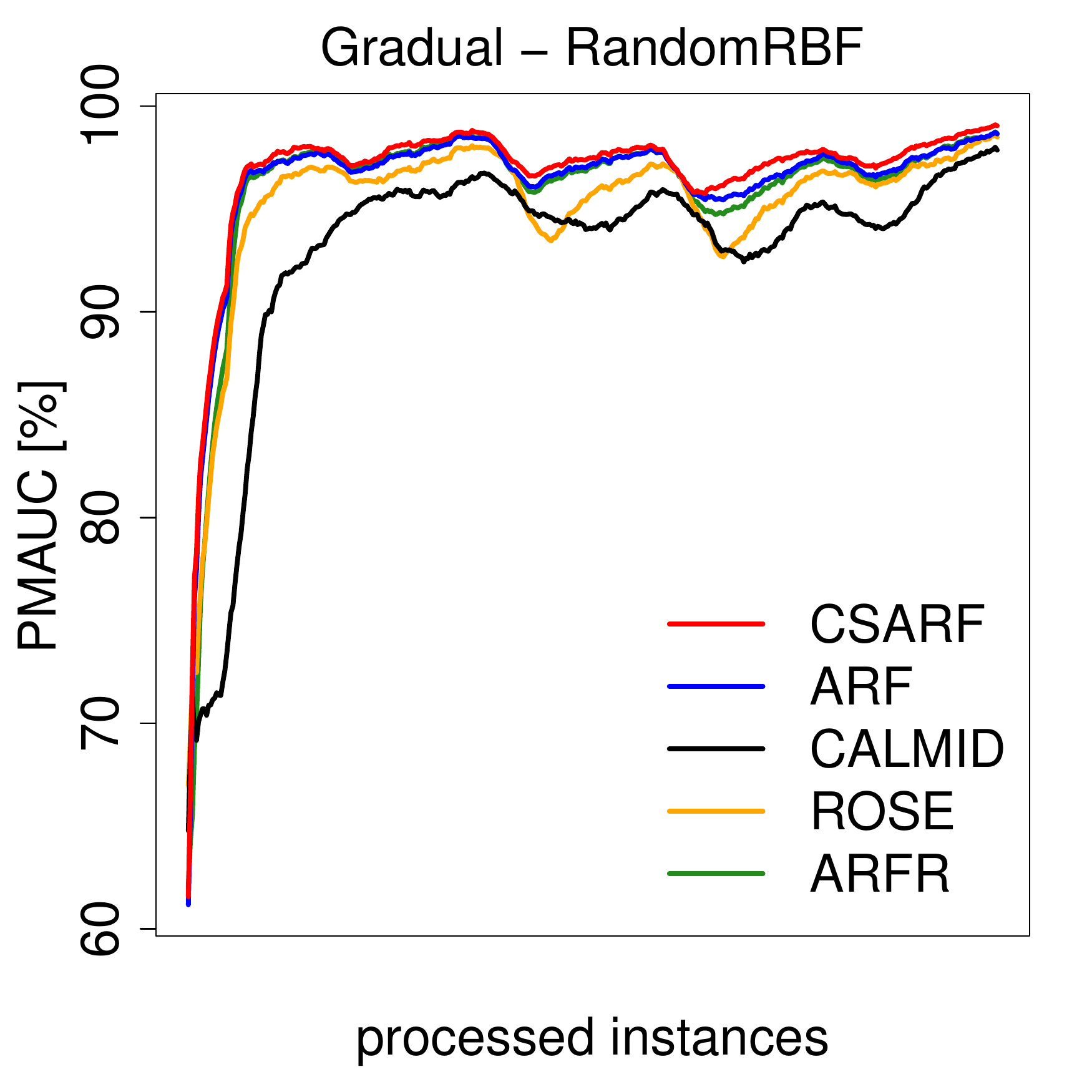}
\includegraphics[width=0.19\columnwidth]{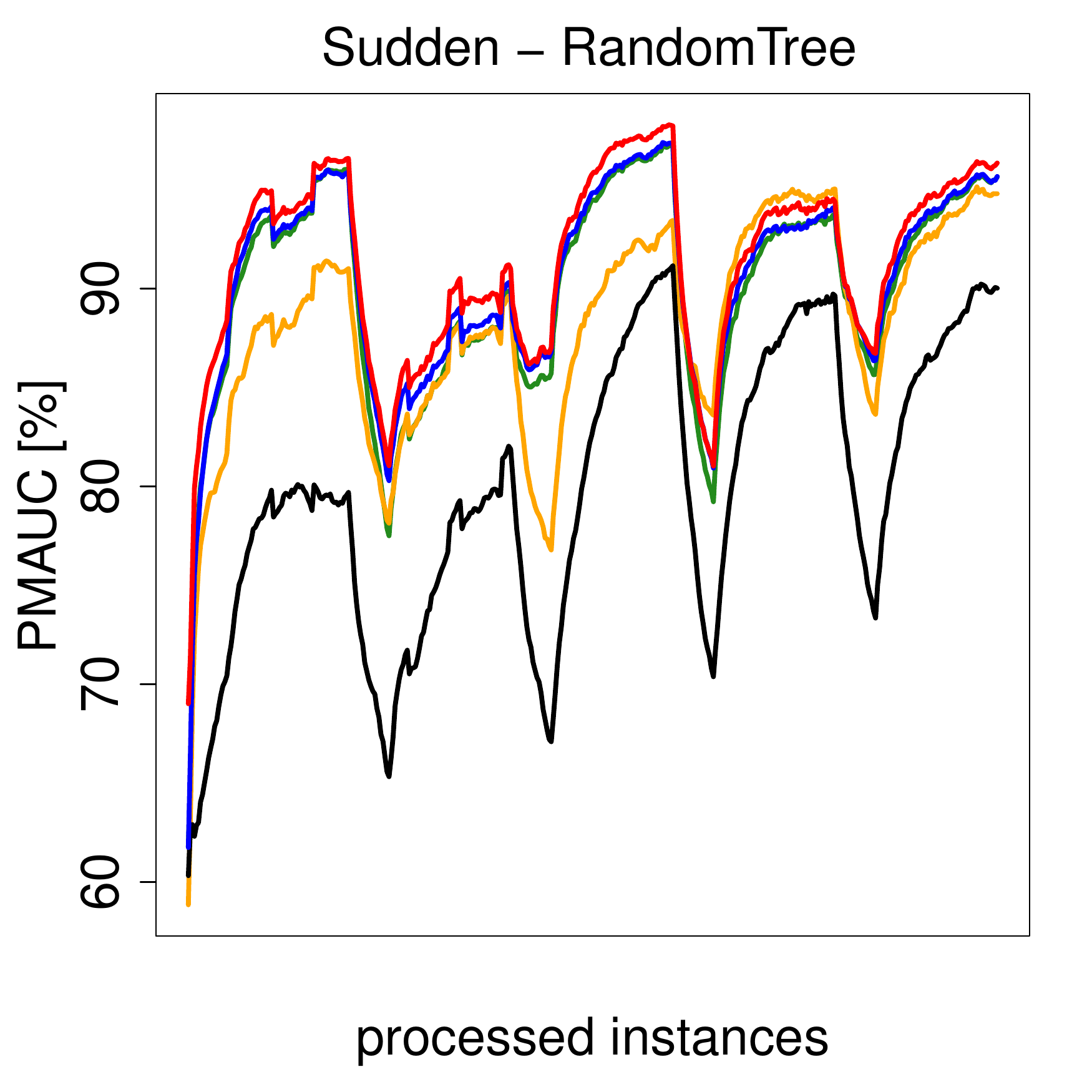}
\includegraphics[width=0.19\columnwidth]{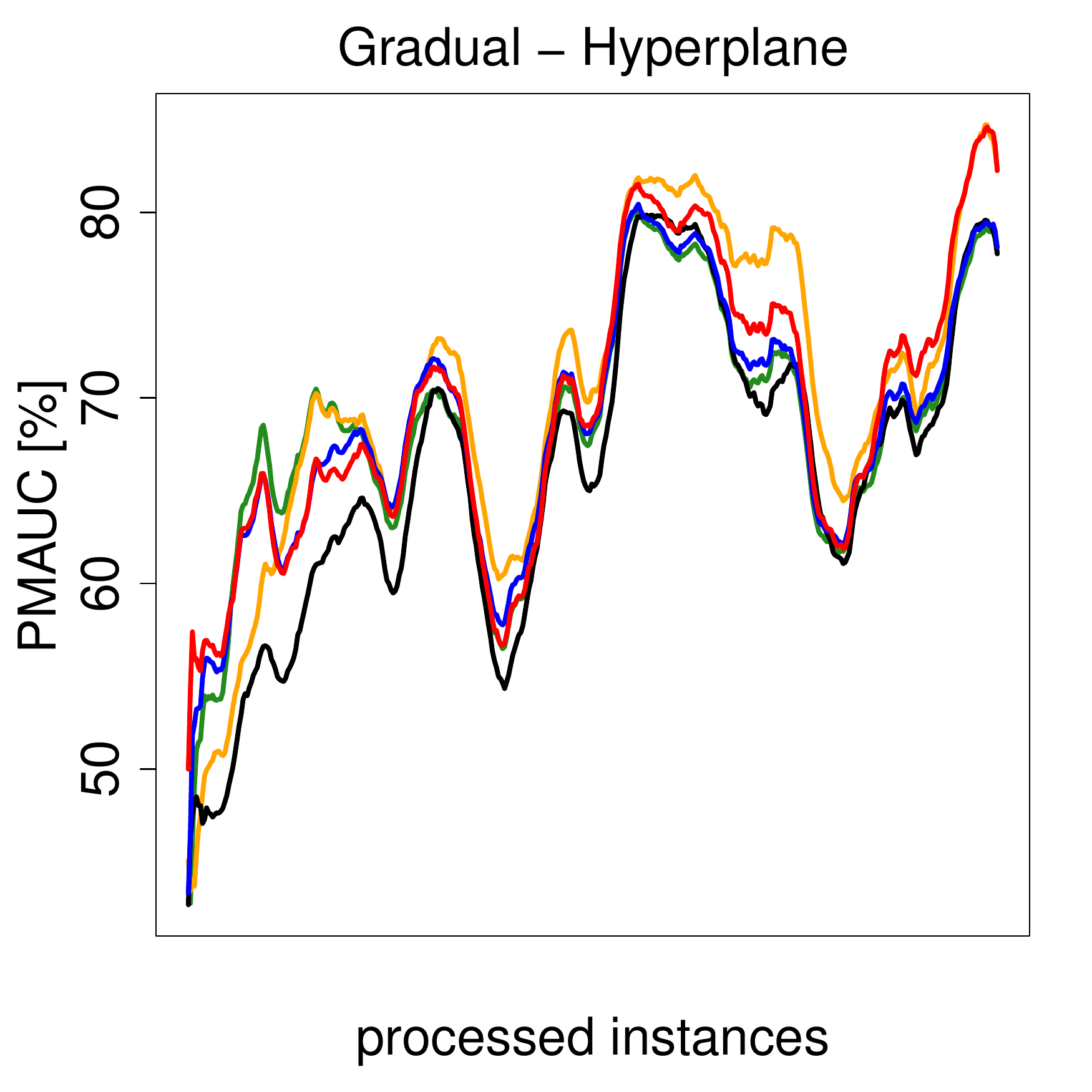}
\includegraphics[width=0.19\columnwidth]{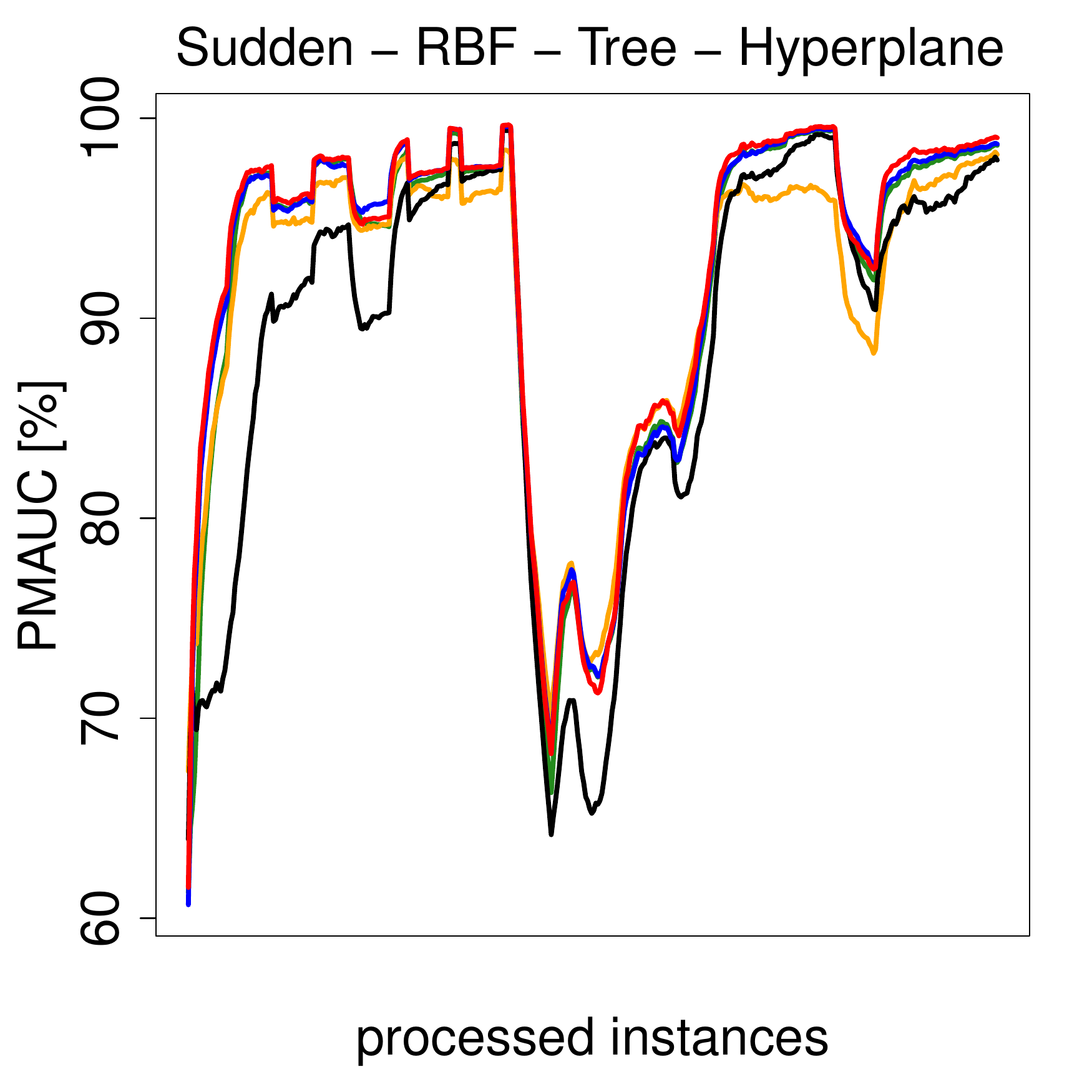}
\\
\includegraphics[width=0.19\columnwidth]{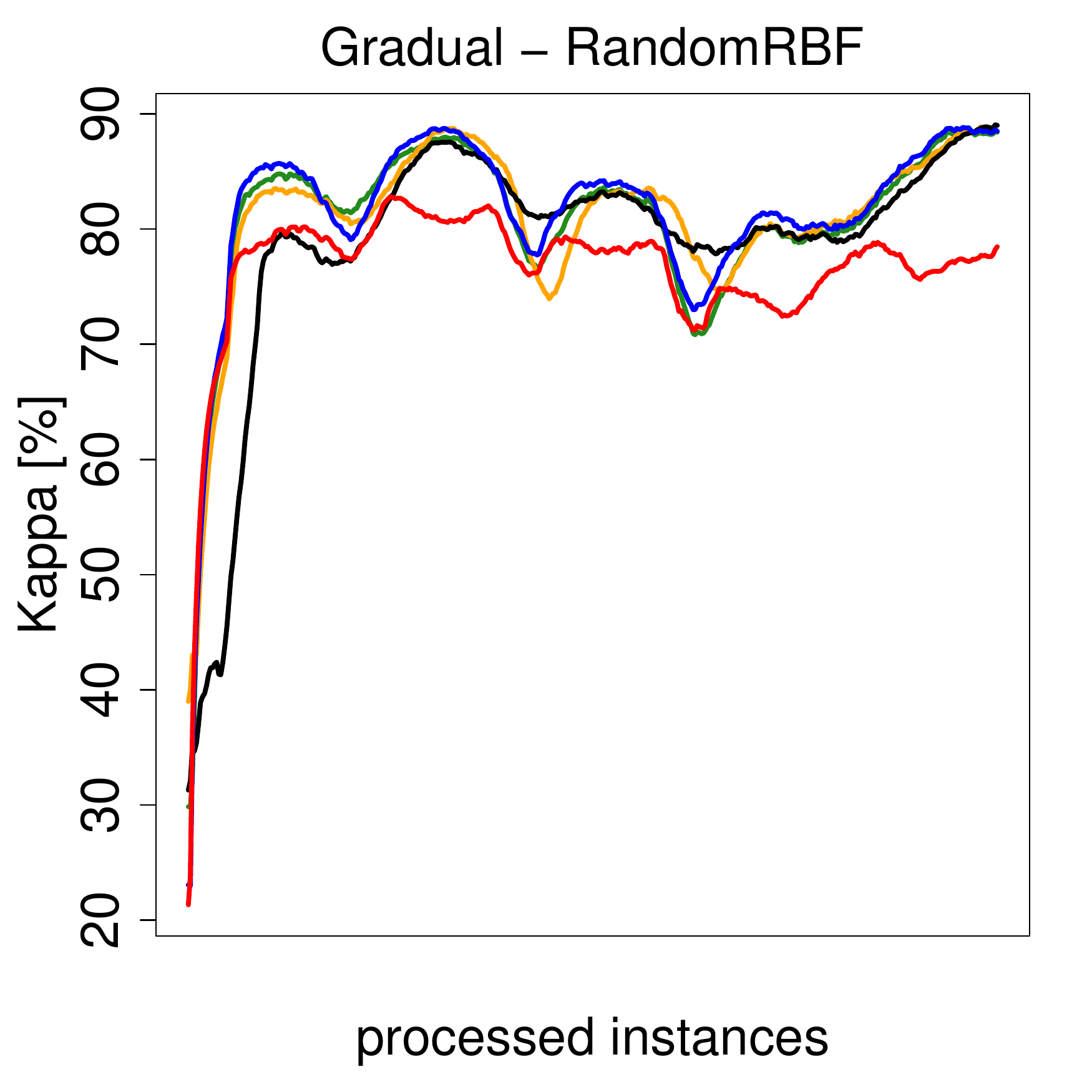}
\includegraphics[width=0.19\columnwidth]{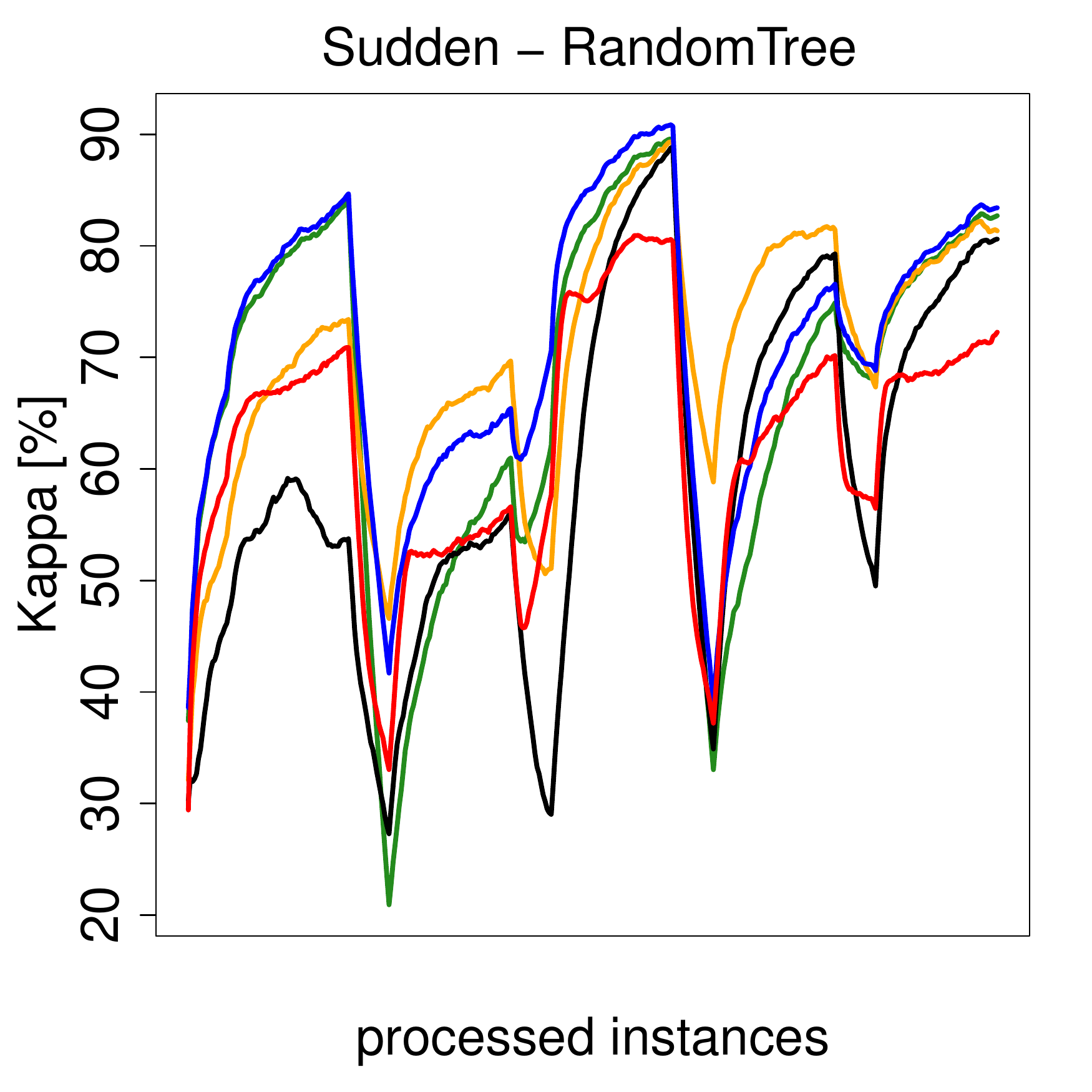}
\includegraphics[width=0.19\columnwidth]{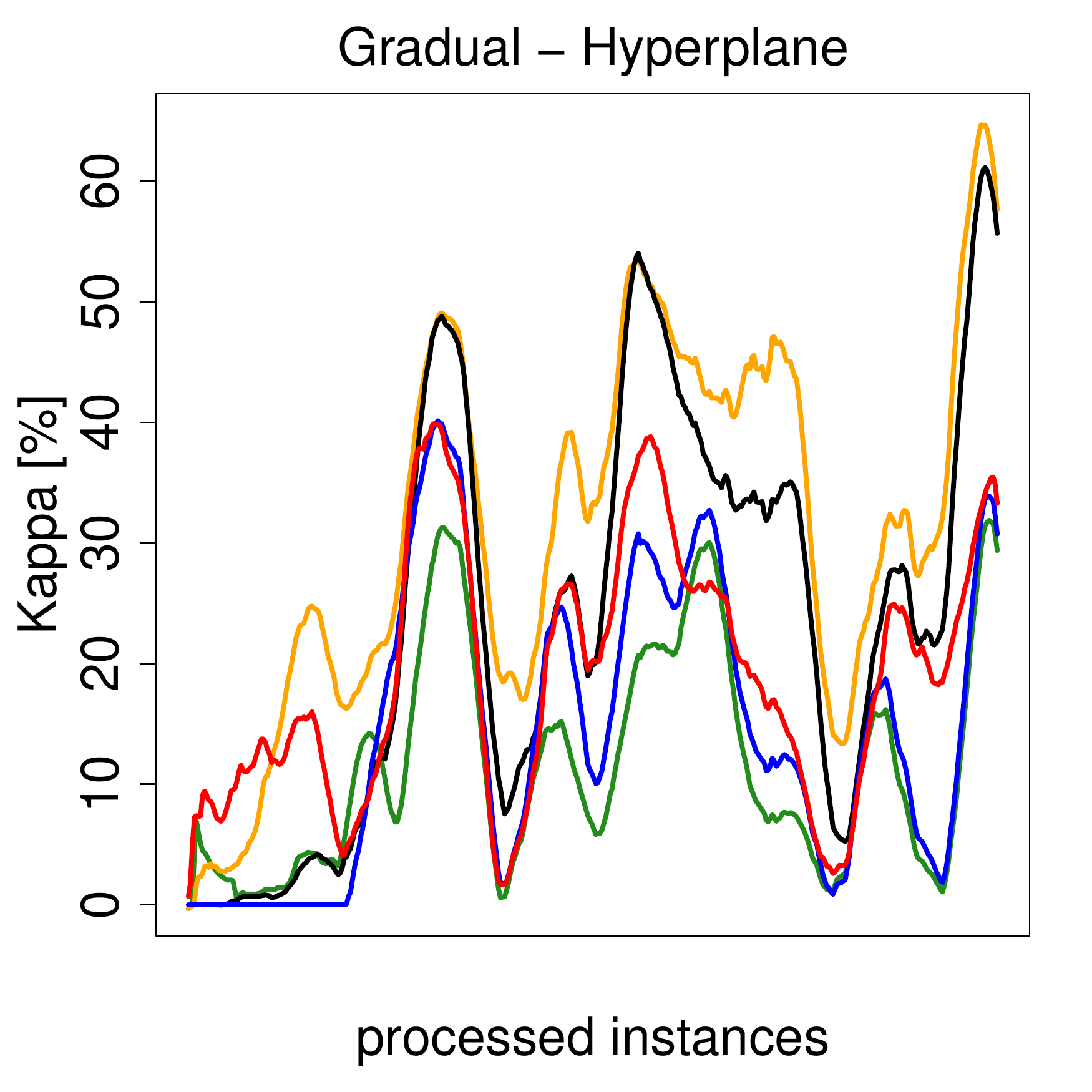}
\includegraphics[width=0.19\columnwidth]{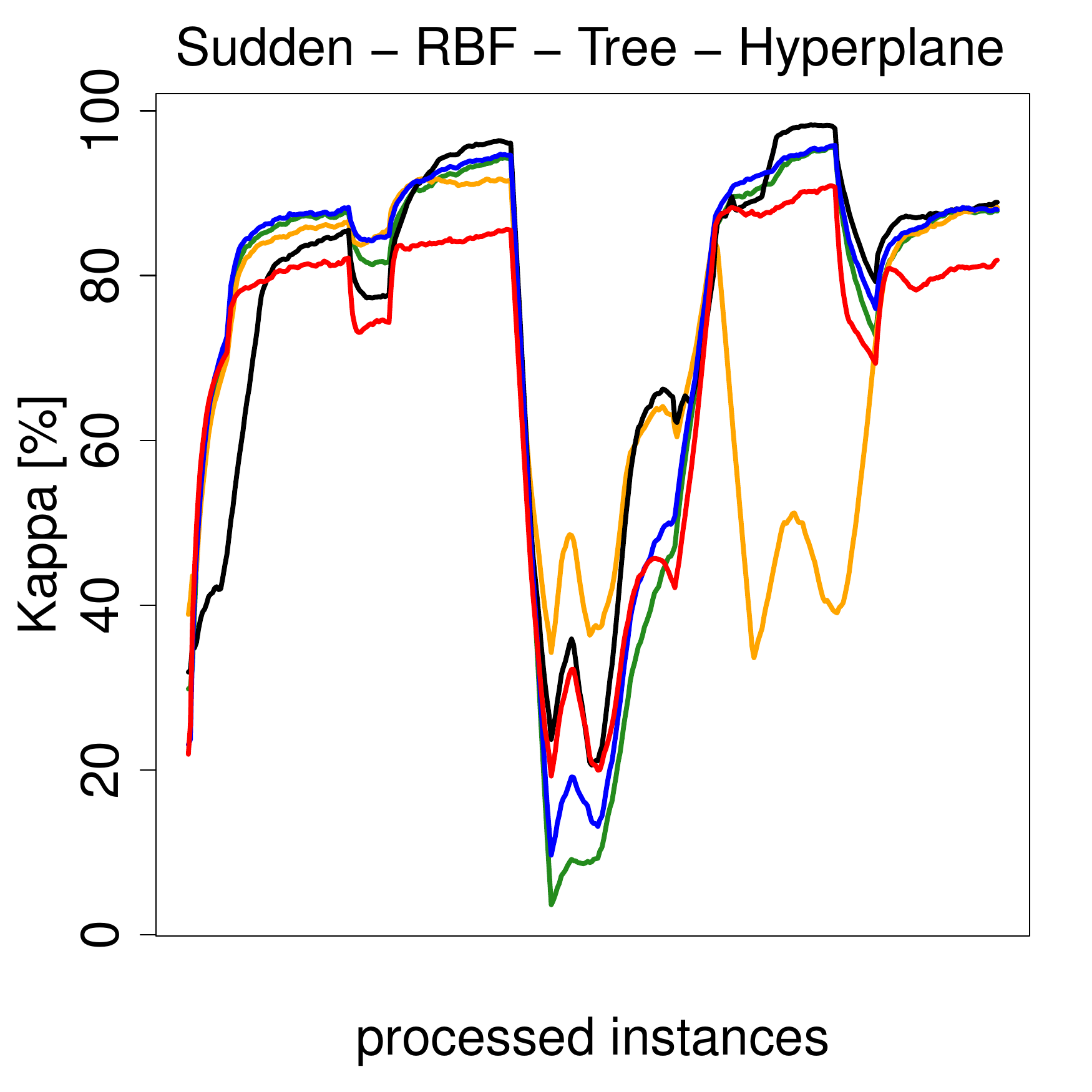}
\caption{PMAUC and Kappa on concept drift and multi-class shifting imbalance ratio.}
\label{fig:mc_cd_shifting_imbalance_ratio}
\end{figure}

\begin{figure}[t!]
\vspace*{0.5cm}
\centering
\includegraphics[width=0.5\columnwidth]{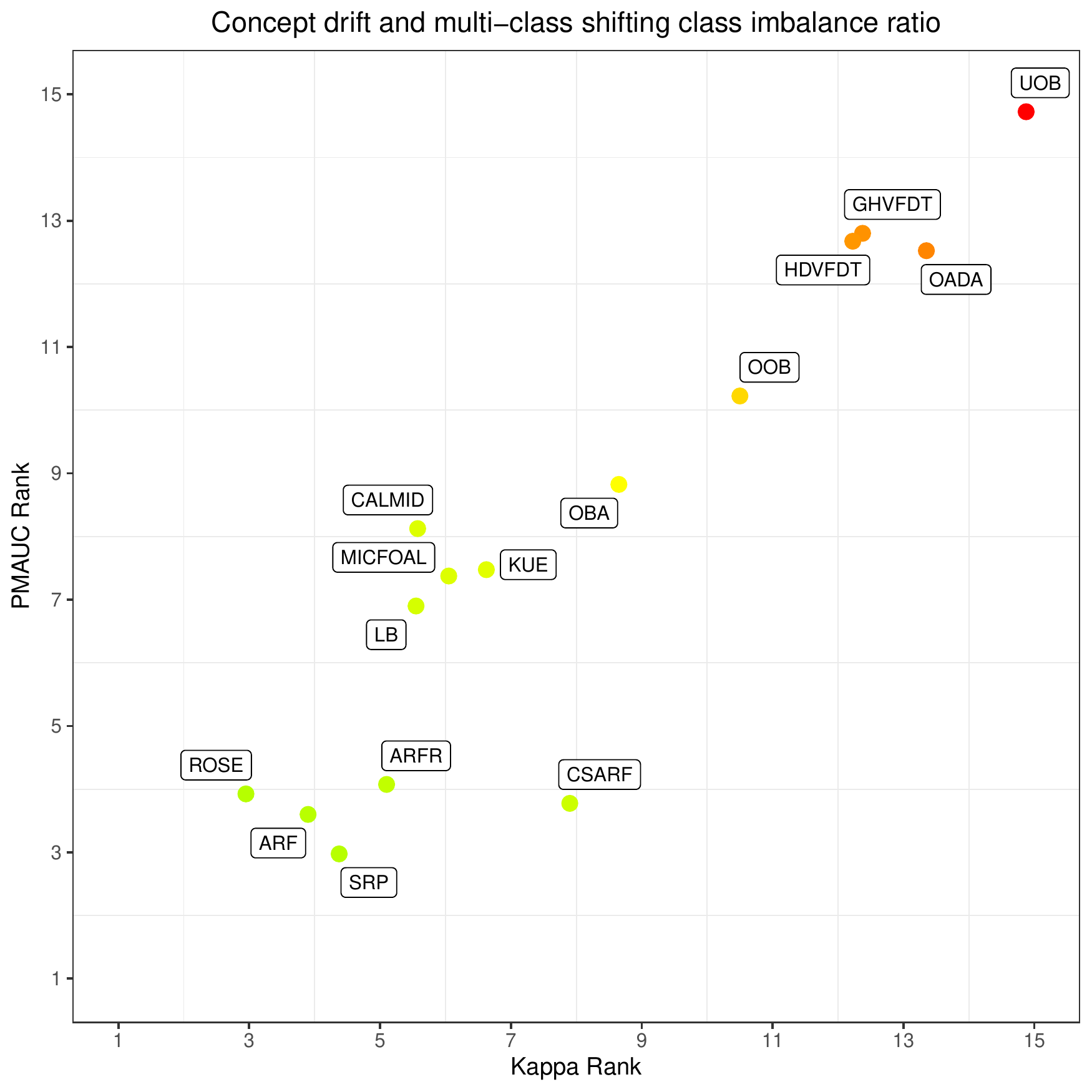}
\caption{Comparison of all 15 algorithms on concept drift and multi-class shifting class imbalance ratio. Color gradient represents the product of both metrics.}
\label{fig:MC_CD_SHF_IR_scatter}
\end{figure}

\noindent \textbf{Goal of the experiment.} This experiment was designed to complement previous experiments and address \textbf{RQ4} to evaluate the behavior of the classifiers in a scenario with multiple classes in the presence of concept drift and dynamic imbalance ratio. Besides concept drift, changes in the imbalance ratio poses obstacles for classifiers that have to deal with multiple changes in data distribution. To evaluate this, we prepared three streams generators similarly to experiment~\ref{sec:mc-dynamic-ir}, but introducing concept drifts along the stream gradually and suddenly. Figure~\ref{fig:mc_cd_shifting_imbalance_ratio} illustrates the PMAUC and Kappa metrics of the selected classifiers for each evaluated drifting stream. Table~\ref{tab:MC_CD_SHF_IR} presents the PMAUC and Kappa for the top 10 classifiers for both types of drift and their average value and ranking. Figure~\ref{fig:MC_CD_SHF_IR_scatter} provides the overall performance for all classifiers. 

\noindent \textbf{Discussion}

\noindent \textit{Impact of class imbalance approach.} Regarding blind resampling methods, we can see that the combination of concept drift and evolving imbalance ratios led to significant deterioration of \acrshort{oob} results, showing that the blind oversampling cannot adapt well to changes happening in both feature space and class characteristics. \acrshort{uob} was impacted even more significantly, making it the worst classifier in this scenario. \acrshort{arfr} is still among the best performing methods, however we can see small drop in the performance compared to the previous experiment. This shows that informed resampling techniques still require more work regarding the adaptation to both drifting and evolving imbalance ratios, as especially class role switching became challenging for \acrshort{arfr}.

When analyzing algorithm-level solutions we can see that \acrshort{csarf}, while still performing well on PMAUC, displayed reduced performance on Kappa. This shows that it cannot handle evolving imbalance ratios and class roles well, having high bias towards the initial role of classes. \acrshort{calmid} and \acrshort{micfoal} improved their relative ranking regarding previous experiments, showing that they are resilient enough to handle both challenges at the same time. 

\acrshort{rose} is a clear winner in this scenario, showing the best robustness to multiple types of changes affecting the data stream. Its adaptation and skew-insensitive mechanisms allow it to efficiently handle the combination of concept drift and dynamic class imbalance, easily adapting to the new incoming concepts, even with changed class roles. 

\noindent \textit{Impact of ensemble architecture.} Once again we can see a clear dominance of bagging-based and hybrid architectures. However, this difficult learning scenario gives us a very unexpected insight. We can see that \acrshort{srp} and \acrshort{lb} are able to outperform \acrshort{calmid} and \acrshort{micfoal}. This is highly surprising, as the former methods are general-purpose classifiers, while the latter ones were specifically designed to handle imbalanced multi-class streams. Additionally, \acrshort{kue} achieved similar performance to dedicated skew-insensitive ensembles. This allows us to conclude that combination of bagging-based or hybrid architecture with an effective drift adaptation mechanism is a leading factor in the performance of ensemble classifiers for drifting and dynamically imbalanced streams. Therefore, it is crucial for future researchers not to focus solely on how to handle class imbalance, but firstly how to handle non-stationary characteristics, and then make this adaptation mechanism skew-insensitive. 

\noindent \textit{Impact of concept drift speed.} Once again, we are unable to see a clear relationship between the speed of concept drift and classifier performance. Even under sudden drifts, most of the examined methods were able to quickly recover and return to their performance before the change. Therefore, end results are similar for any speed of change. The differences can be observed very locally during the drift occurrence, but they did not have a long-lasting effect on any classifier. 

\noindent \textit{Relationship between concept drift and shifting imbalance ratio.} This scenario combines two types of changes, creating a more realistic and challenging scenario. Therefore, we need to understand the impact of each of these types of changes on the underlying classifier. Analyzing the results, we can see that most of existing algorithms are characterized by a trade-off: either focusing on adaptation to changes or on robustness to class imbalance. Only \acrshort{rose} and \acrshort{srp} displayed a balanced performance on both tasks. This supports our previous conclusion that there is a need to design novel methods where both adaptation and skew-insensitiveness will be solved as a joint problem.

\begin{table*}[t!]
\vspace*{0.5cm}
\centering
\footnotesize
\setlength{\tabcolsep}{4pt}
\caption{PMAUC and Kappa on concept drift and multi-class shifting imbalance ratio.}
\label{tab:MC_CD_SHF_IR}
\begin{tabular}{ll|C{1cm}C{1cm}C{1cm}C{1cm}C{1cm}C{1cm}C{1cm}C{1cm}C{1cm}C{1cm}}
\toprule
& Drift & CSARF & ARF & KUE & LB & SRP & CALMID & MICFOAL & ROSE & ARFR & OOB\\
\midrule
\multirow{2}{*}{\rotatebox[origin=c]{90}{\scalebox{.6}{PMAUC}}}
& Sudden & 82.02 & 81.97 & 77.82 & 78.66 & \textbf{82.40} & 77.60 & 78.75 & 81.80 & 81.80 & 74.14\\
& Gradual & 80.83 & 81.54 & 80.04 & 79.14 & \textbf{82.05} & 78.25 & 79.37 & 81.93 & 81.43 & 76.56\\
\midrule
\multirow{2}{*}{\rotatebox[origin=c]{90}{\scalebox{.65}{Kappa}}}
& Sudden & 48.85 & 54.31 & 48.93 & 50.05 & 51.95 & 50.46 & 51.54 & \textbf{55.57} & 52.81 & 37.27\\
& Gradual & 43.35 & 49.18 & 46.40 & 46.91 & 49.01 & 45.83 & 47.19 & \textbf{51.68} & 48.26 & 37.07\\
\midrule
\multicolumn{2}{l|}{Avg. PMAUC} & 81.43 & 81.75 & 78.93 & 78.90 & \textbf{82.23} & 77.92 & 79.06 & 81.87 & 81.61 & 75.35\\
\multicolumn{2}{l|}{Avg. Kappa} & 46.10 & 51.75 & 47.66 & 48.48 & 50.48 & 48.15 & 49.37 & \textbf{53.62} & 50.54 & 37.17\\
\midrule
\multicolumn{2}{l|}{Rank PMAUC} & 3.55 &  3.40 &  7.13 &  6.28 &  \textbf{2.93} &  7.75 &  6.93 &  3.90 &  3.88 &  9.28\\
\multicolumn{2}{l|}{Rank Kappa} & 7.30 &  3.70 &  6.45 &  5.18 &  4.15 &  5.30 &  5.70 &  \textbf{2.93} &  4.88 &  9.43\\
\bottomrule
\end{tabular}
\end{table*}

\subsubsection{Impact of the number of classes}
\label{sec:mc-nclasses}

\begin{figure}[t!]
\vspace*{0.5cm}
\centering
\includegraphics[width=0.19\columnwidth]{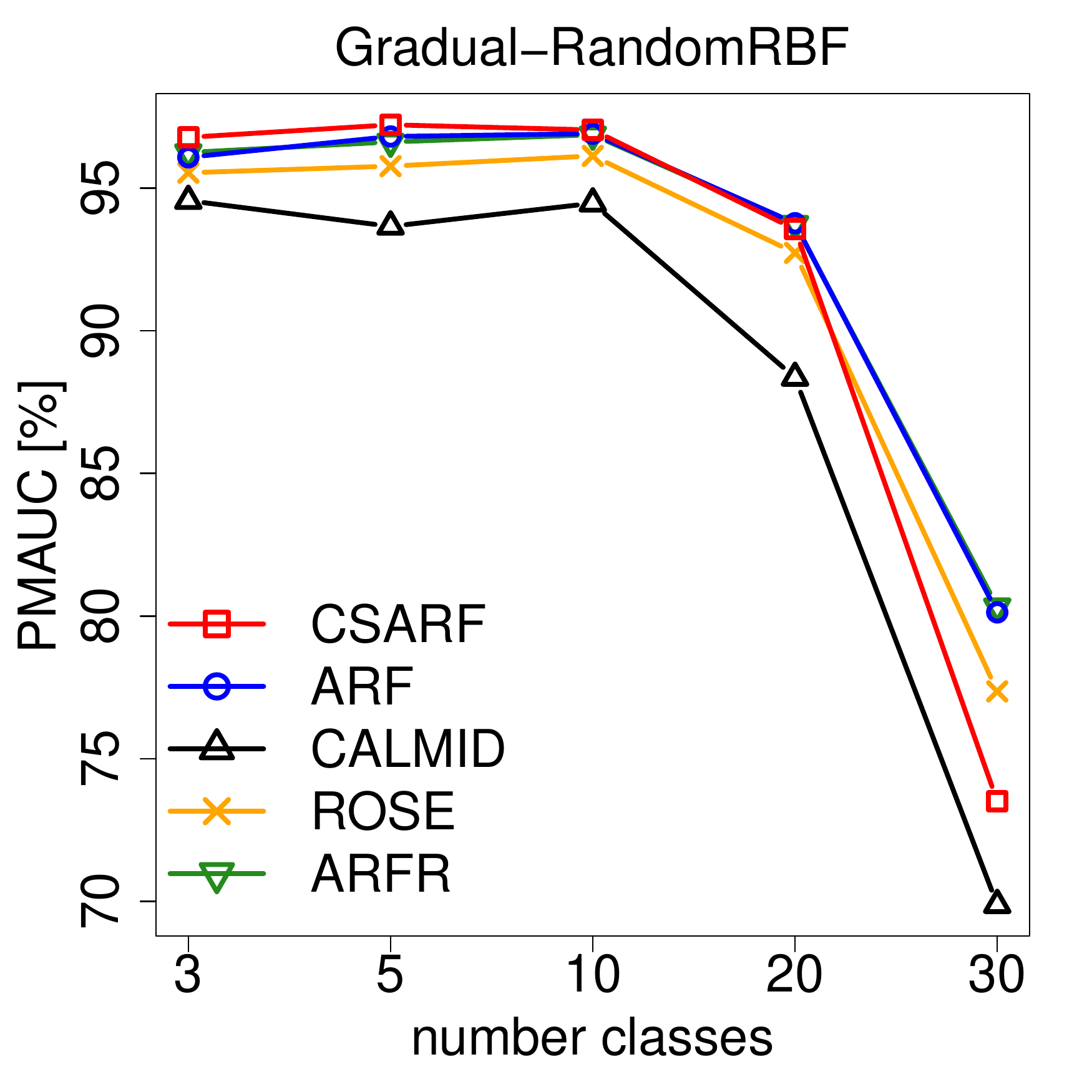}
\includegraphics[width=0.19\columnwidth]{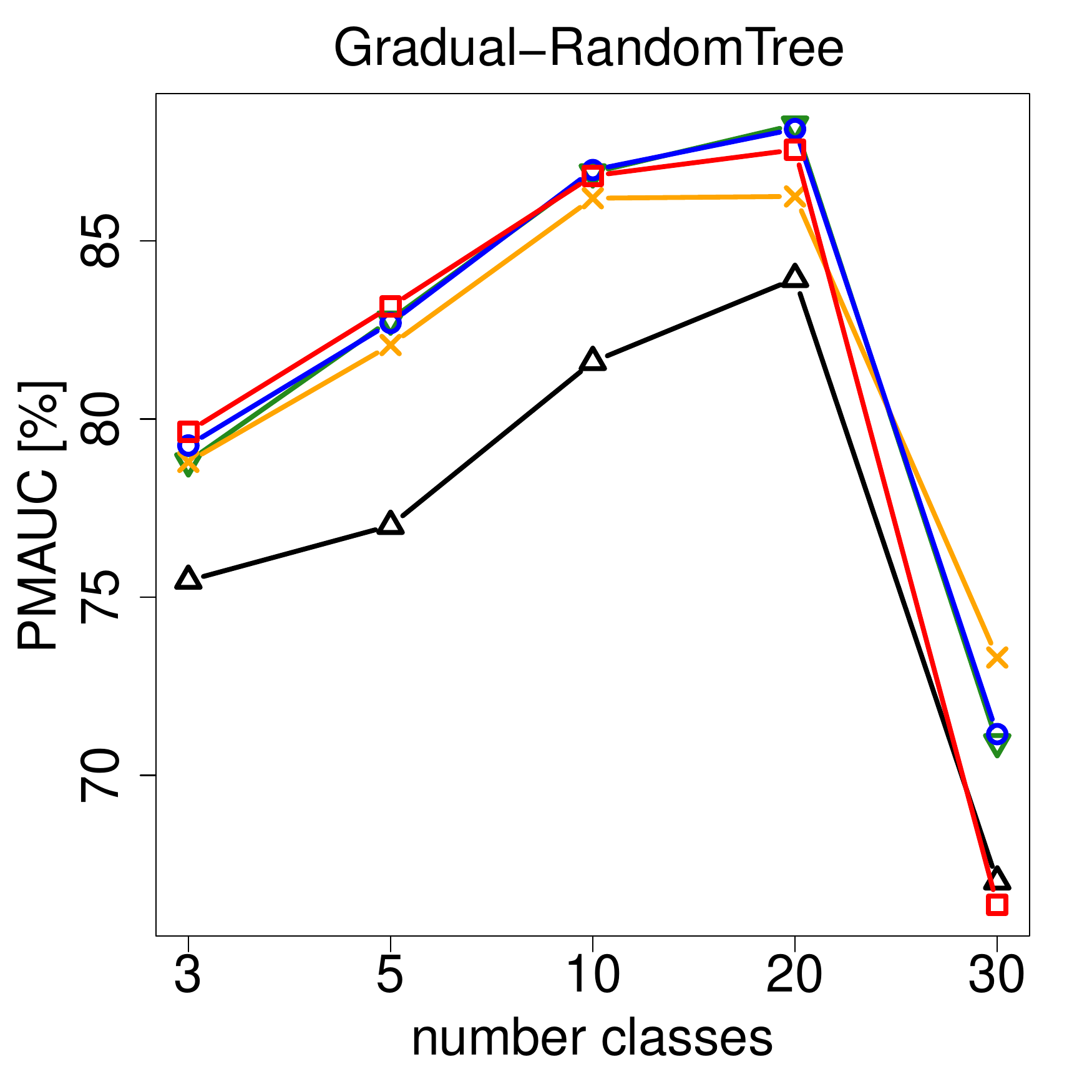}
\includegraphics[width=0.19\columnwidth]{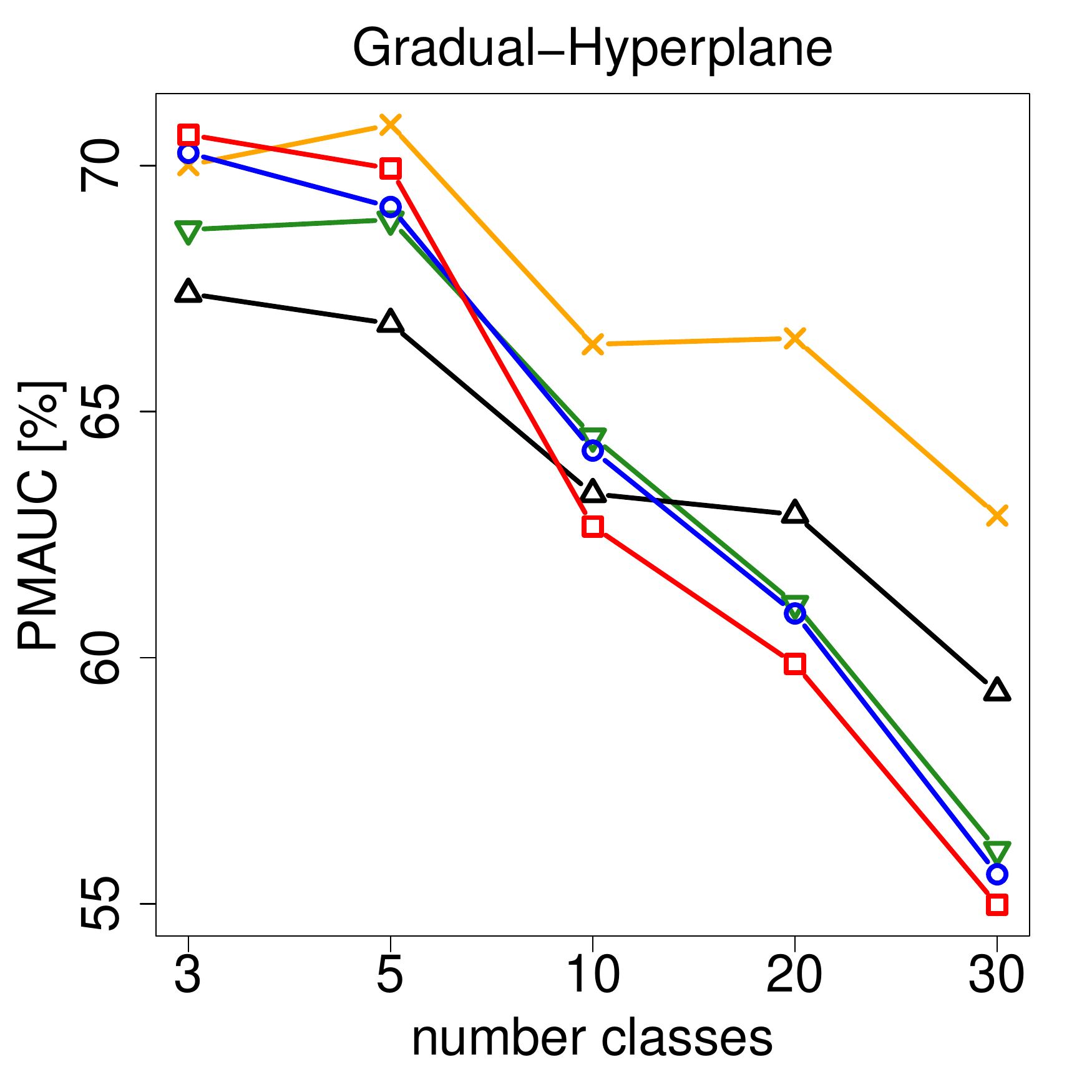}
\includegraphics[width=0.19\columnwidth]{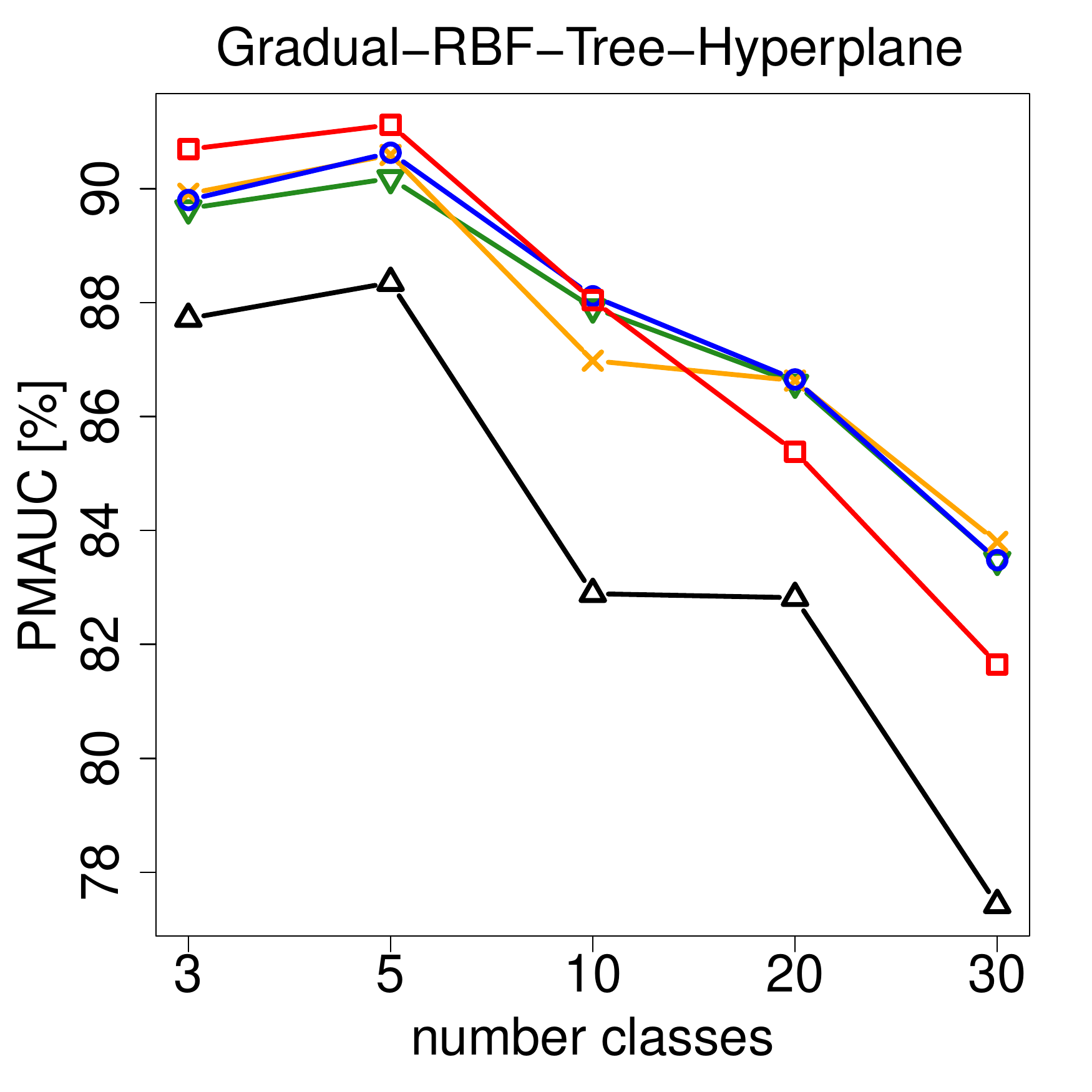}
\\
\includegraphics[width=0.19\columnwidth]{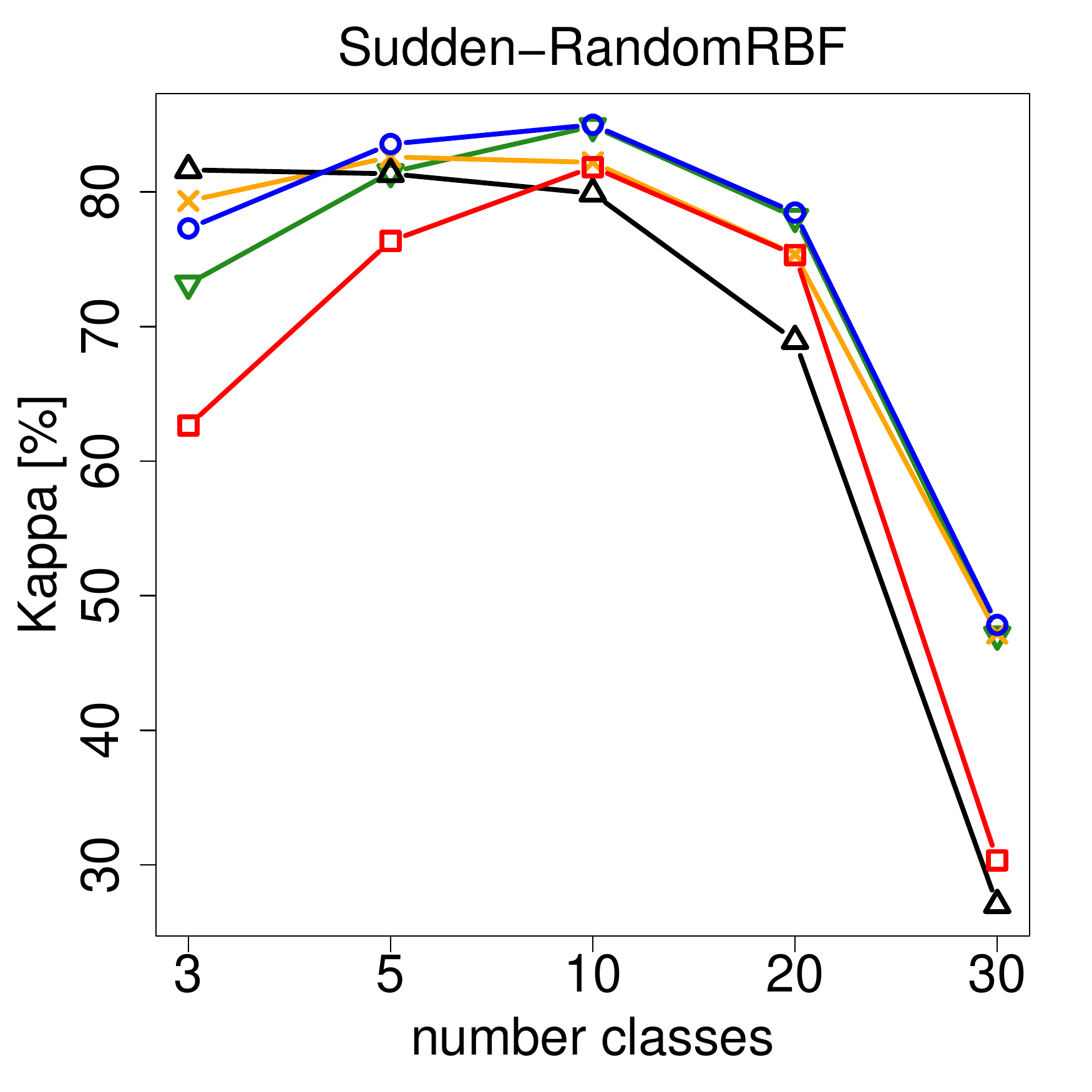}
\includegraphics[width=0.19\columnwidth]{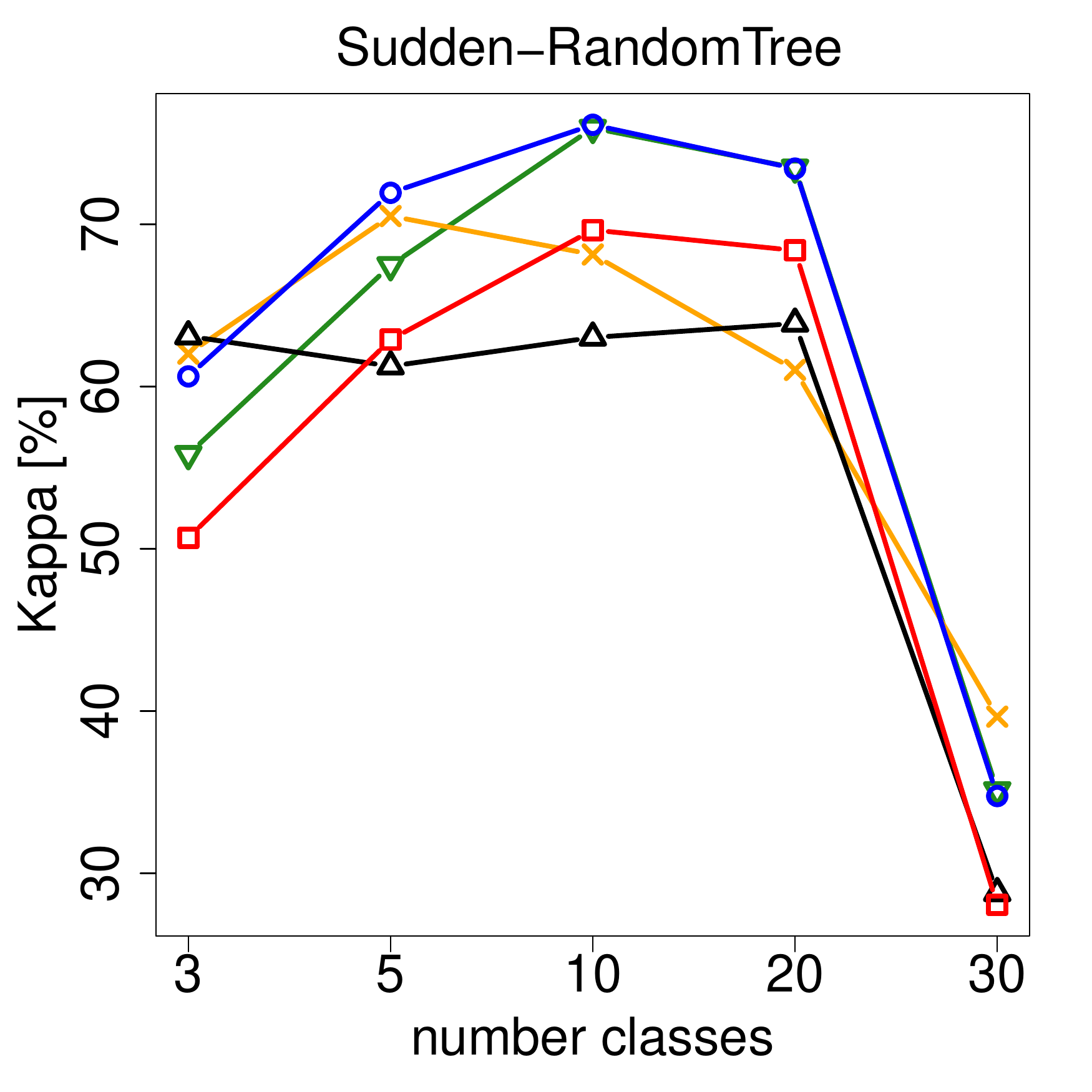}
\includegraphics[width=0.19\columnwidth]{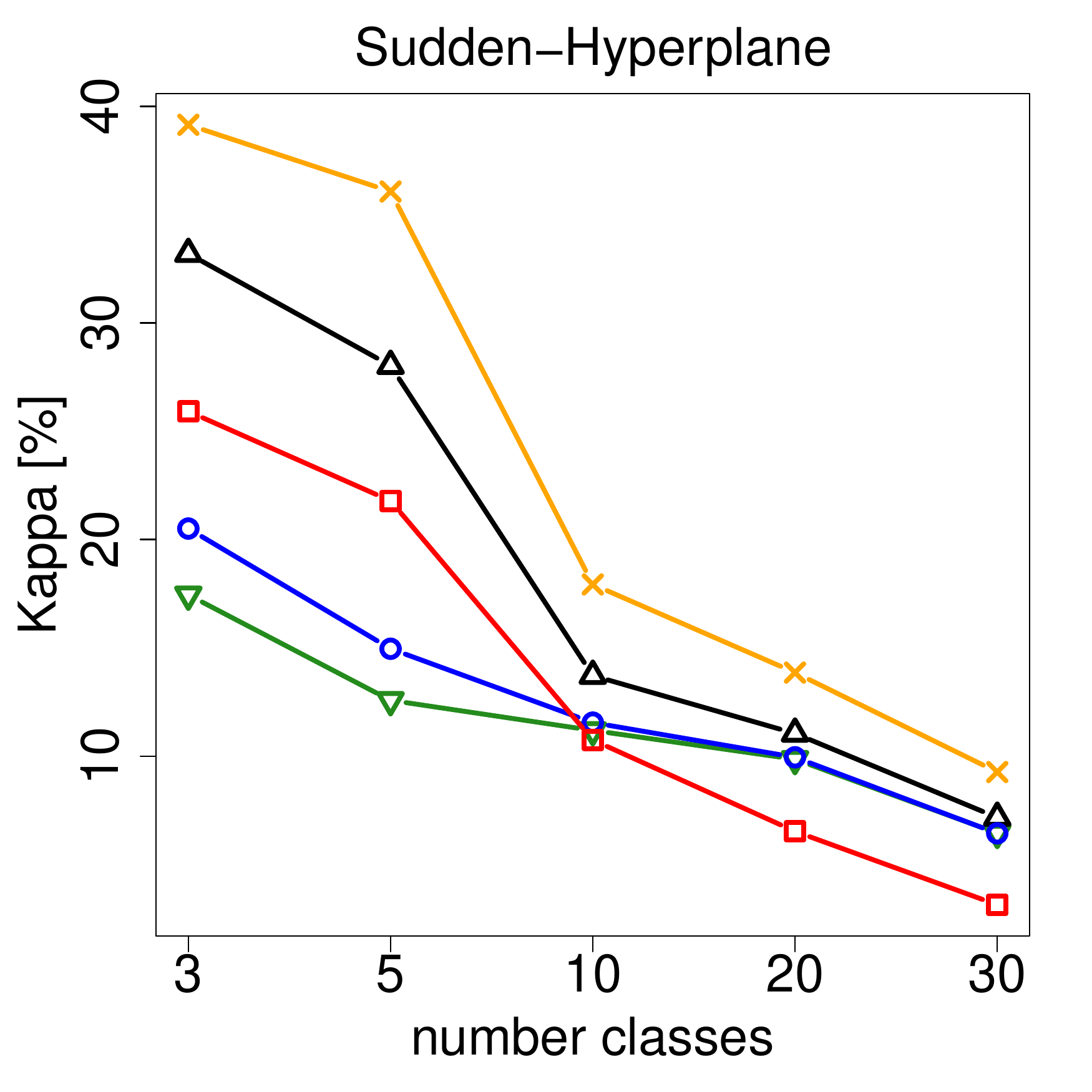}
\includegraphics[width=0.19\columnwidth]{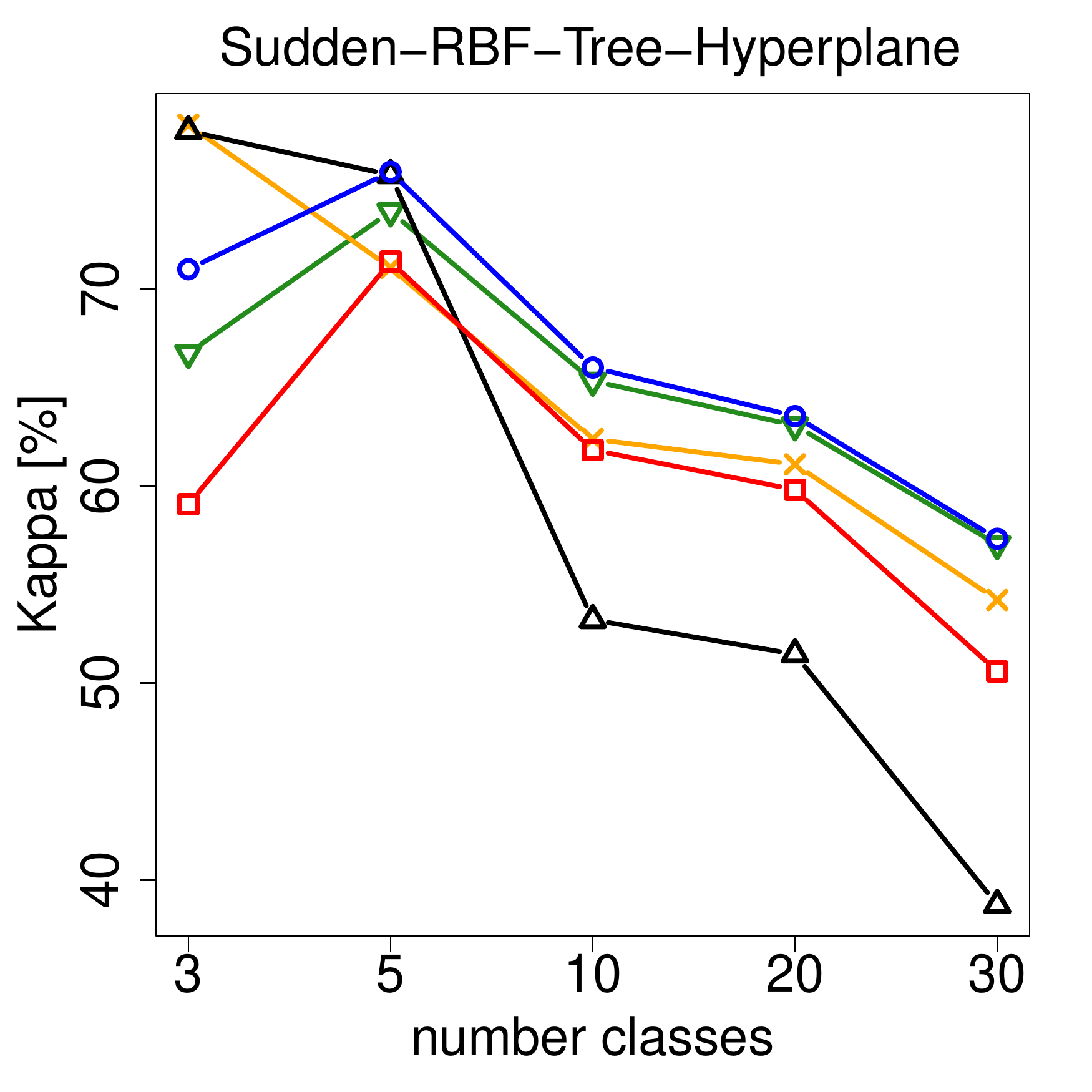}
\caption{Impact of the number of classes on PMAUC and Kappa under concept drift and multi-class shifting imbalance ratio.}
\label{fig:mc_cd_shifting_imbalance_ratio_nclasses}
\end{figure}

\begin{figure}[t!]
\vspace*{0.5cm}
\centering
\includegraphics[width=\columnwidth]{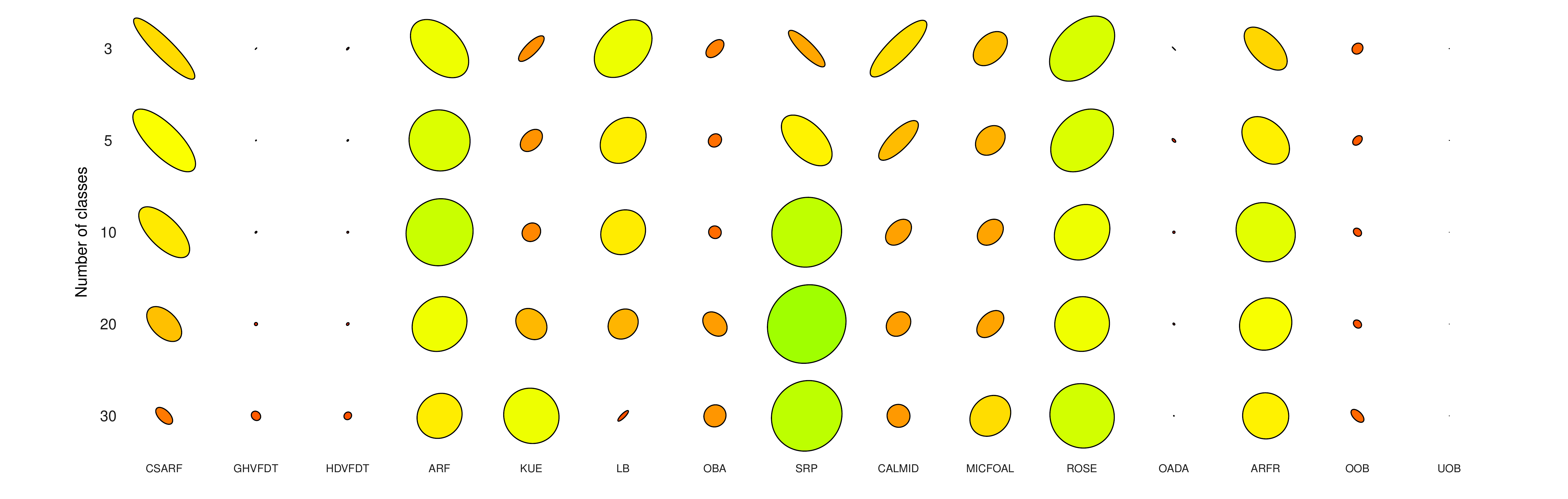}
\caption{Comparison of all 15 algorithms on the number of classes under concept drift and multi-class shifting imbalance ratio. Axes of the ellipse represent PMAUC and Kappa metrics. Color gradient represents the product of both metrics.}
\label{fig:MC_CD_SHF_IR_NCLASSES_ellipse}
\end{figure}

\noindent \textbf{Goal of the experiment.} This experiment was designed to evaluate the robustness of the classifiers to different number of classes under the presence of concept drift and dynamic imbalance ratio. Combining those learning difficulties with different number of classes allow us to evaluate how classifiers deal with higher number of classes and examine if does affect their learning mechanisms or not. To evaluate this, we used the generators in experiment \ref{sec:mc-dynamic-ir}. All these generators were evaluated with the following number of classes \{3, 5, 10, 20, 30\}. Figure~\ref{fig:mc_cd_shifting_imbalance_ratio_nclasses} illustrates the performance of five selected algorithms classifier for each number of classes. Table~\ref{tab:MC_CD_SHF_IR_NCLASSES} summarizes the performance of the top 10 classifiers for each number of classes and their average ranking regarding each metric. For overall comparison, Figure~\ref{fig:MC_CD_SHF_IR_NCLASSES_ellipse} presents the overall aggregated performance of all classifiers. Axes of the ellipse represent PMAUC and Kappa metrics, the more rounded the better, and the color represents the product of both metrics.

\noindent \textbf{Discussion}

\noindent \textit{Impact of class imbalance approach.} We can see that high number of classes pose a significant challenge for most of the examined methods. For resampling-based approaches, we observe that \acrshort{oob} and \acrshort{uob} cannot handle any higher number of classes, returning the worst performance of the same rank as single tree classifiers (\acrshort{ghvfdt} and \acrshort{hdvfdt}). \acrshort{arfr} maintains its performance with the increase in the number of classes, showing that the combination of informed resampling with \acrshort{arf}-based architecture allows for the memorization of more complex decision boundaries, while combating bias using well-placed artificial instances.

When looking at the algorithm-level modifications we can see that \acrshort{csarf}, previously one of the best algorithms, displays no robustness to increasing number of classes. This shows the limitations of cost-sensitive approaches, as with the increased number of classes the cost matrix needs to grow. Large cost matrices lead to loss of meaning behind the penalties and their reduced influence on learning process and no effect on bias towards majority classes. We can conclude that existing cost-sensitive methods are not suitable for handling multi-class imbalanced streams with a high number of classes. \acrshort{calmid}, despite being designed for multi-class problems, cannot handle increasing number of classes and returns performance similar to ensembles based on blind resampling. \acrshort{rose} and \acrshort{micfoal} displayed the best robustness to high number of classes. \acrshort{rose}, especially for the Kappa metric, is a safe choice for scenarios with elevated number of classes.

\noindent \textit{Impact of ensemble architecture.} Our analysis of the ensembles under increasing number of classes showed once again the dominance of bagging-based and hybrid architectures. In this scenario, hybrid approaches became dominant, with \acrshort{rose} displaying best results due to its combination of working on both instance and feature subspaces, combined with per-class memory buffers for balanced class representations. 

\noindent \textit{Impact of the high number of classes.} With the increasing number of classes, we can see a clear break point when the number of classes is $> 20$. This shows that all classifiers could handle the increasing number of classes up to a certain point, after which their capabilities for memorizing new concepts and generalizing over all classes begin to rapidly deteriorate. We can see that for scenarios with $30$ classes most of the methods start returning highly unsatisfactory results. Interestingly, for these cases we can observe a very good performance of standard classifiers, such as \acrshort{srp}. When analyzing ranks, we can see that \acrshort{srp} and \acrshort{rose} are two best performing ensembles when handling high number of classes. While we provided the explanation for the better performance of \acrshort{rose}, it is very surprising to see that \acrshort{srp} performs on par with it. We can explain it by the fact that both \acrshort{rose} and \acrshort{srp} use feature subspaces, which can be seen as lower dimensional projections of a difficult learning task. In such a lower dimensional subspaces the decision boundaries among classes may be simplified, leading to better generalization capabilities. This follows observation made in \citep{korycki2021low}, where it was postulated that low-dimensional representations can overcome class imbalance without any dedicated skew-insensitive mechanisms.

\begin{table*}[t!]
\vspace*{0.5cm}
\centering
\footnotesize
\setlength{\tabcolsep}{4pt}
\caption{PMAUC and Kappa averages on the number of classes under concept drift and multi-class shifting imbalance ratio.}
\label{tab:MC_CD_SHF_IR_NCLASSES}
\begin{tabular}{ll|C{1cm}C{1cm}C{1cm}C{1cm}C{1cm}C{1cm}C{1cm}C{1cm}C{1cm}C{1cm}}
\toprule
& Classes & CSARF & ARF & KUE & LB & SRP & CALMID & MICFOAL & ROSE & ARFR & OOB\\
\midrule
\multirow{5}{*}{\rotatebox[origin=c]{90}{PMAUC}}
& 3 & \textbf{85.81} & 84.88 & 80.07 & 83.19 & 84.06 & 81.26 & 81.62 & 83.99 & 84.30 & 78.45\\
& 5 & \textbf{86.65} & 86.10 & 81.27 & 84.52 & 86.08 & 81.89 & 82.68 & 85.74 & 85.86 & 78.57\\
& 10 & 84.49 & 84.69 & 80.84 & 83.10 & \textbf{84.98} & 80.74 & 81.18 & 84.37 & 84.68 & 77.33\\
& 20 & 81.48 & 81.95 & 80.27 & 79.91 & \textbf{82.92} & 78.89 & 79.11 & 82.32 & 81.99 & 75.96\\
& 30 & 68.71 & 71.13 & 72.19 & 63.77 & \textbf{73.09} & 66.84 & 70.71 & 72.93 & 71.23 & 66.45\\
\midrule
\multirow{5}{*}{\rotatebox[origin=c]{90}{Kappa}}
& 3 & 48.01 & 55.72 & 54.00 & 60.04 & 44.70 & 60.92 & 54.10 & \textbf{62.20} & 52.48 & 43.66\\
& 5 & 55.22 & 58.39 & 53.72 & 58.47 & 54.46 & 58.83 & 54.81 & \textbf{63.08} & 56.11 & 44.43\\
& 10 & 53.11 & 56.71 & 49.11 & 53.30 & \textbf{57.07} & 50.48 & 51.77 & 55.88 & 56.36 & 38.51\\
& 20 & 49.52 & 53.83 & 47.57 & 48.28 & \textbf{56.11} & 46.62 & 49.06 & 51.76 & 53.66 & 36.79\\
& 30 & 24.65 & 34.08 & 33.92 & 22.33 & \textbf{40.06} & 23.88 & 37.08 & 35.19 & 34.06 & 22.43\\
\midrule
\multicolumn{2}{l|}{Avg. PMAUC} & 81.43 & 81.75 & 78.93 & 78.90 & \textbf{82.23} & 77.92 & 79.06 & 81.87 & 81.61 & 75.35\\
\multicolumn{2}{l|}{Avg. Kappa} & 46.10 & 51.75 & 47.66 & 48.48 & 50.48 & 48.15 & 49.37 & \textbf{53.62} & 50.54 & 37.17\\
\midrule
\multicolumn{2}{l|}{Rank PMAUC} & 3.55 &  3.40 &  7.13 &  6.28 &  \textbf{2.93} &  7.75 &  6.93 &  3.90 &  3.88 &  9.28\\
\multicolumn{2}{l|}{Rank Kappa} & 7.30 &  3.70 &  6.45 &  5.18 &  4.15 &  5.30 &  5.70 &  \textbf{2.93} &  4.88 &  9.43\\
\bottomrule
\end{tabular}
\end{table*}

\newpage
\subsubsection{Real-world multi-class imbalanced datasets}
\label{sec:mc-datasets}

\begin{figure}[t!]
\centering
\includegraphics[width=0.19\columnwidth]{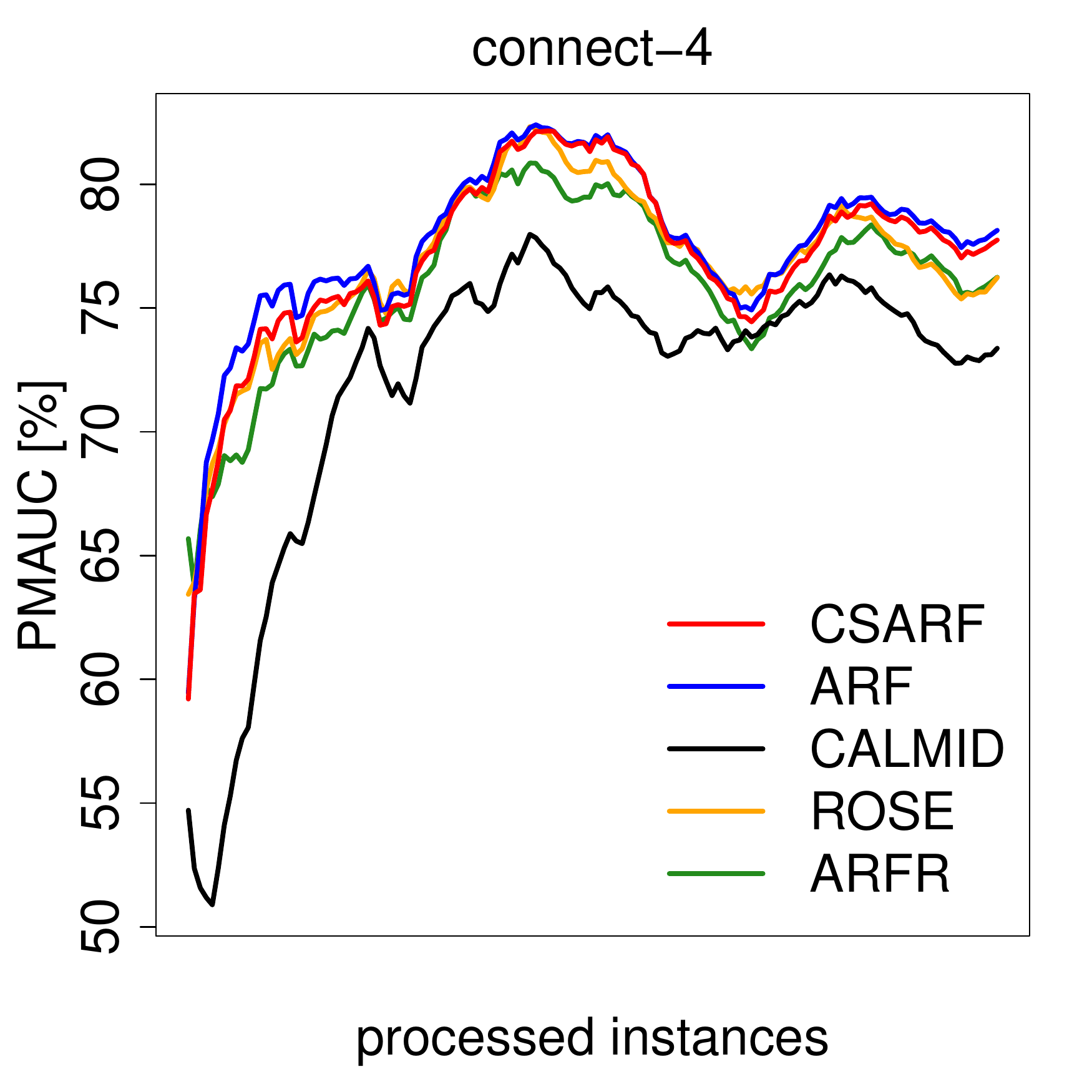}
\includegraphics[width=0.19\columnwidth]{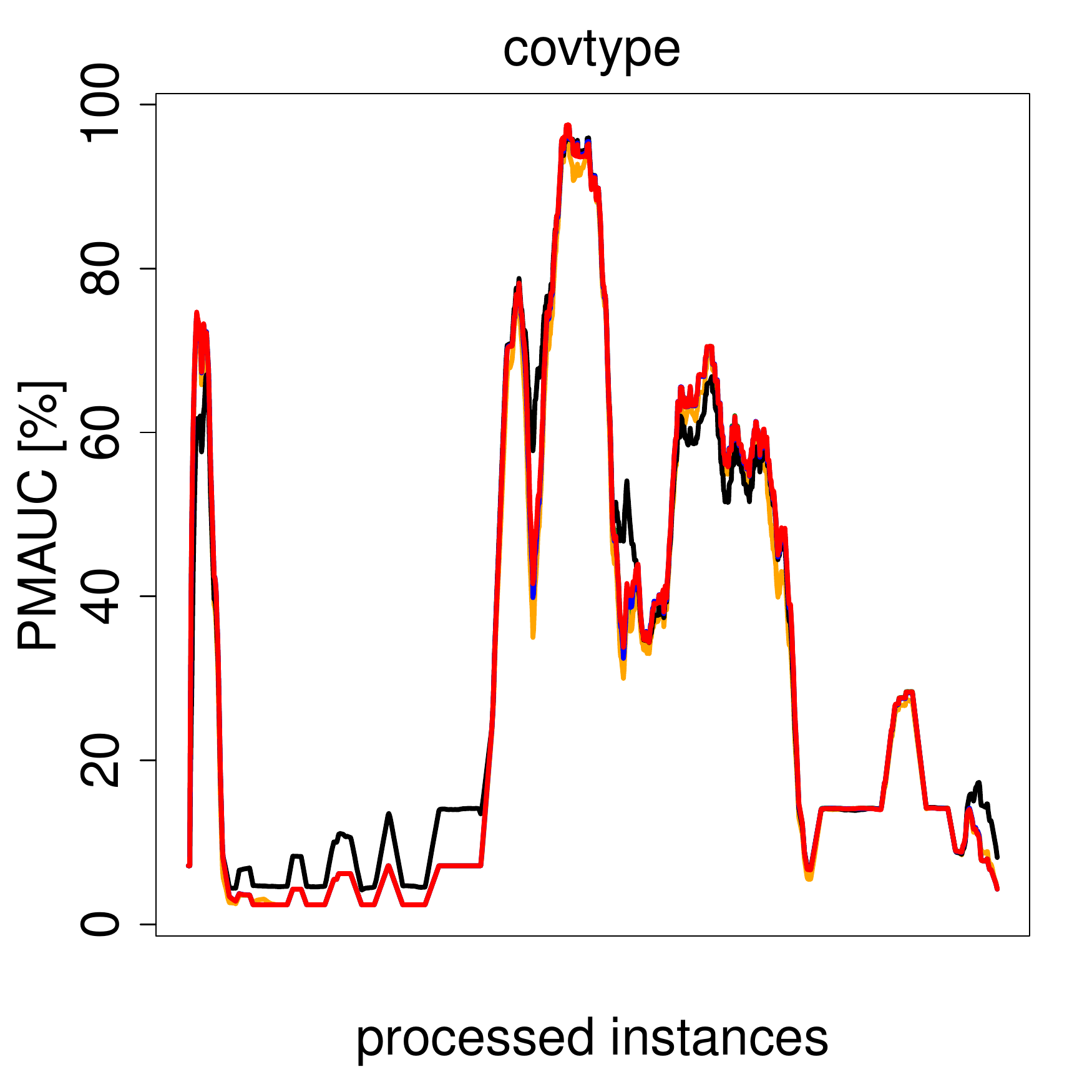}
\includegraphics[width=0.19\columnwidth]{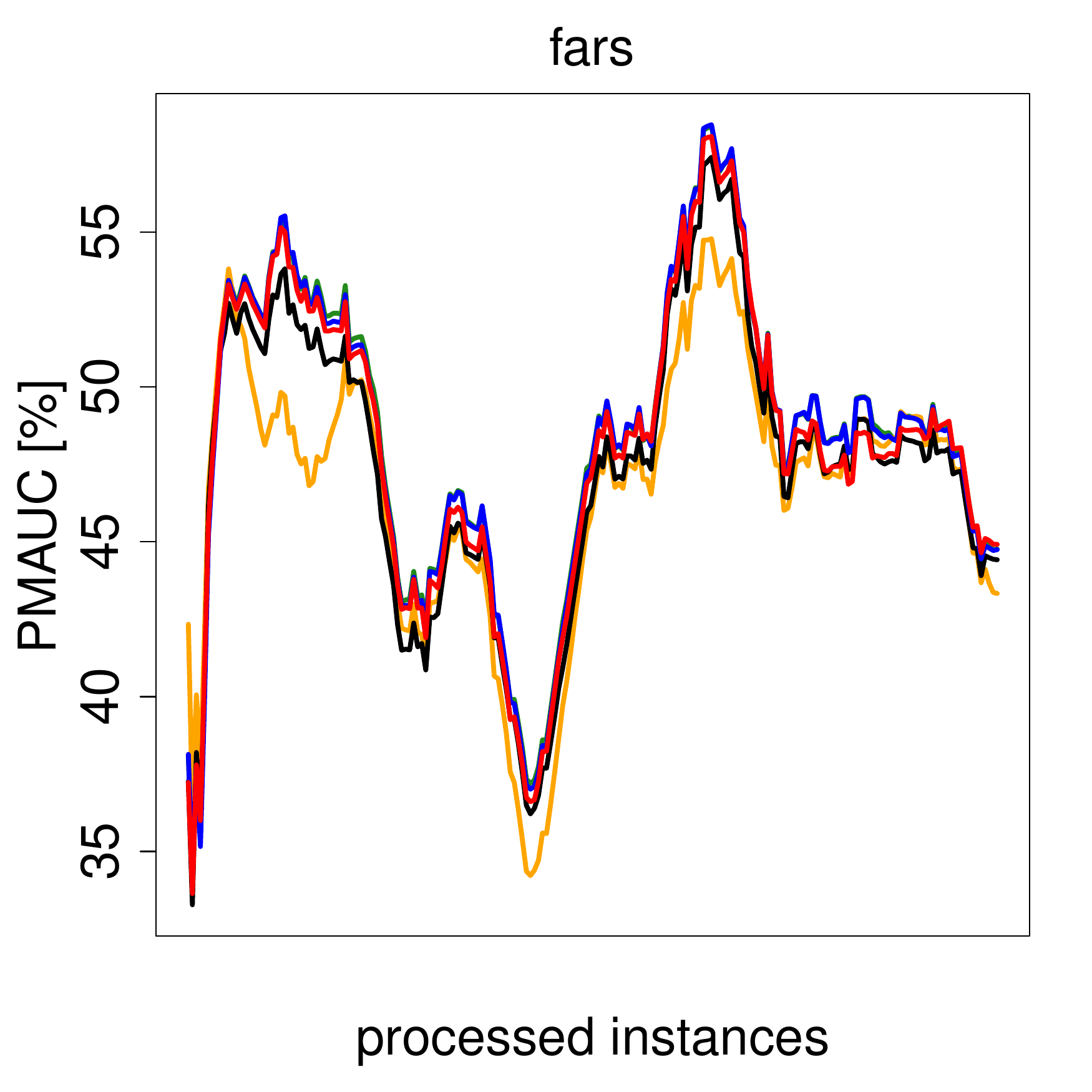}
\includegraphics[width=0.19\columnwidth]{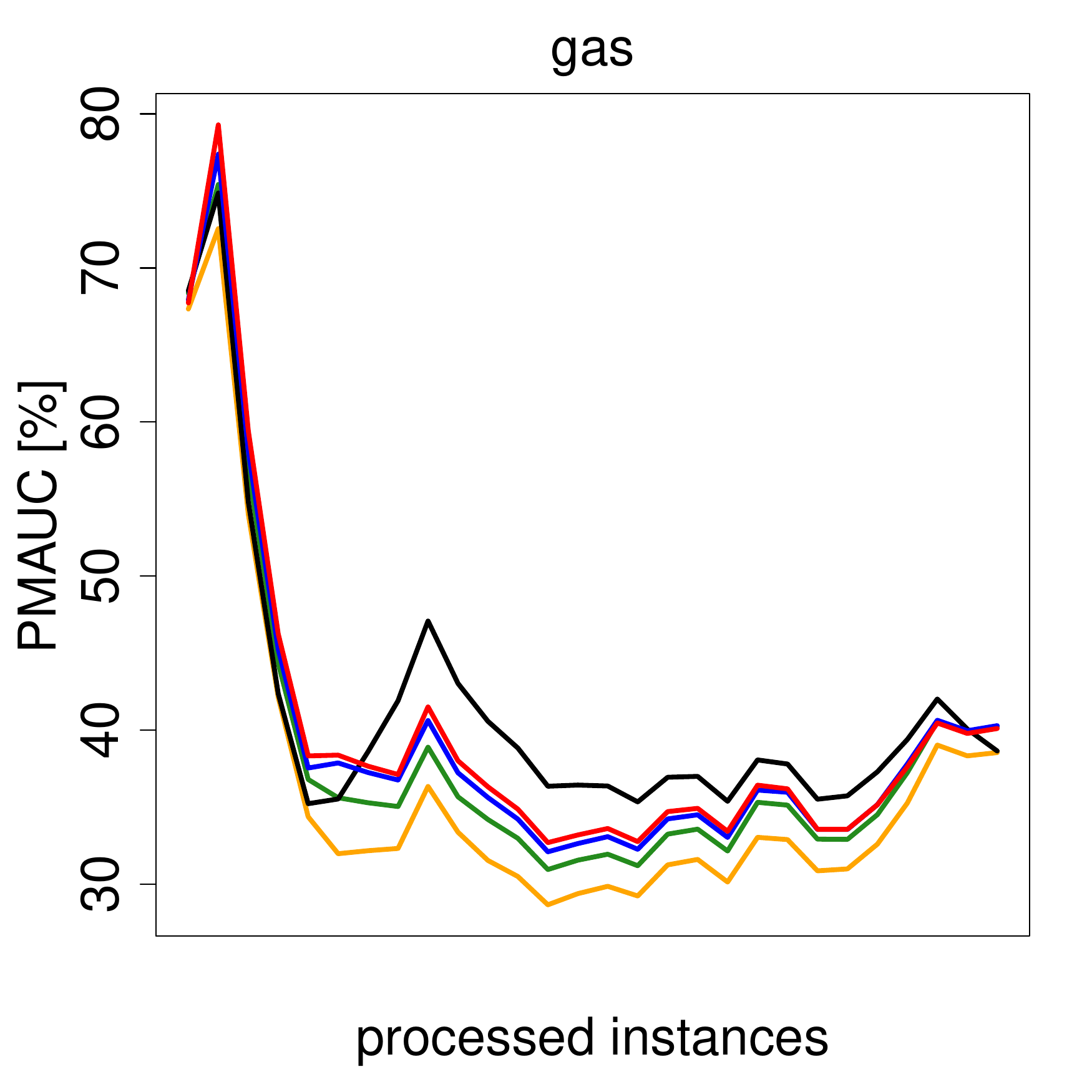}
\includegraphics[width=0.19\columnwidth]{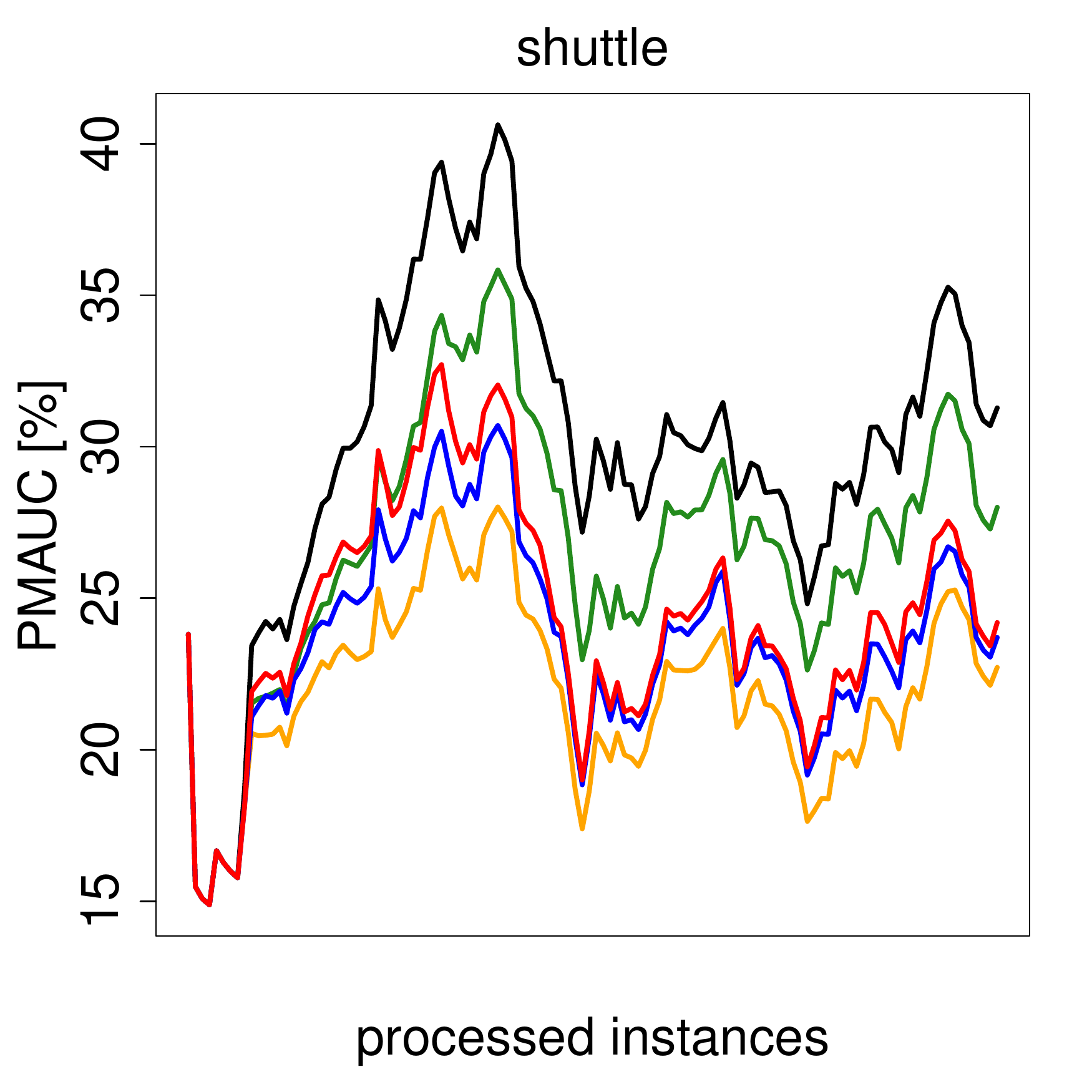}
\includegraphics[width=0.19\columnwidth]{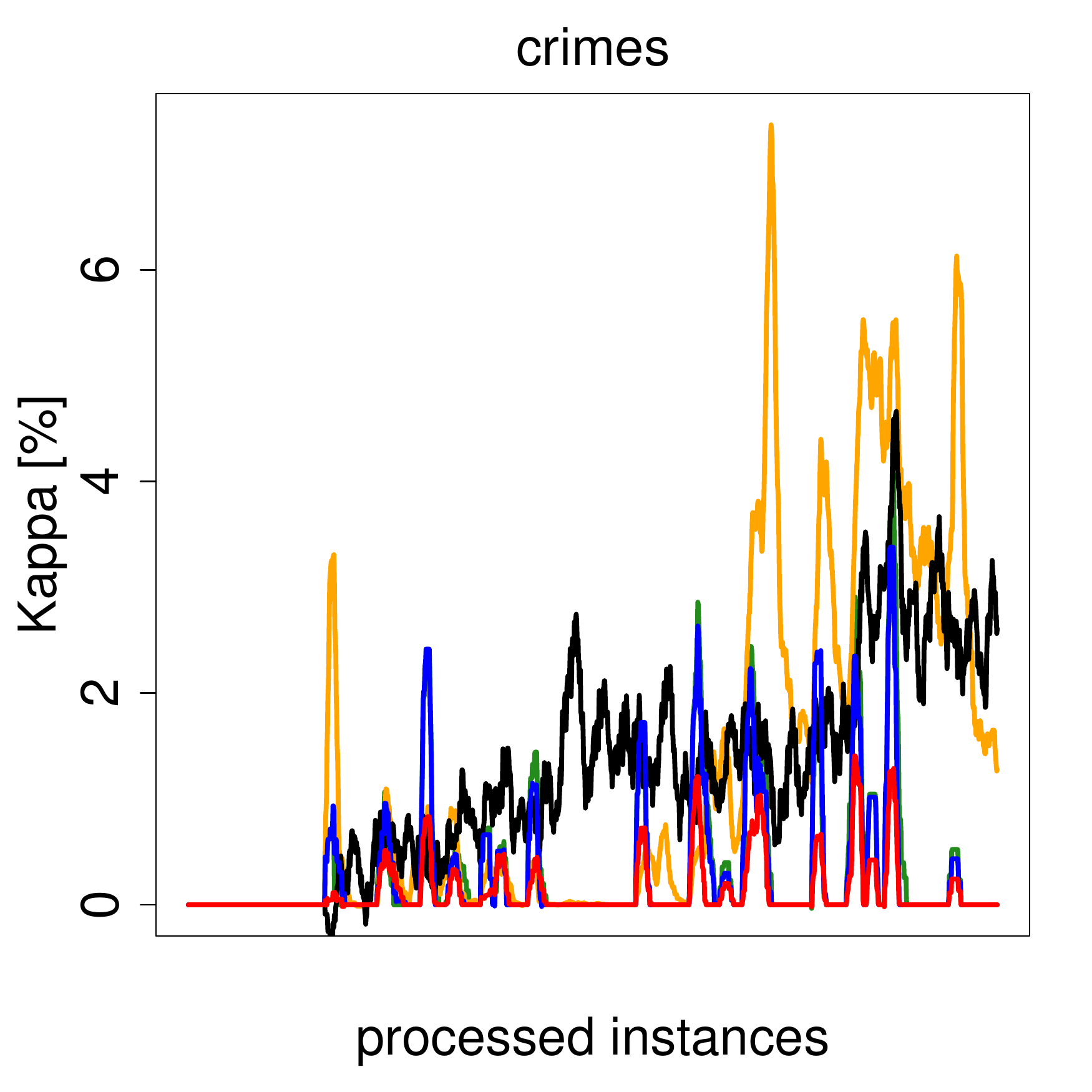}
\includegraphics[width=0.19\columnwidth]{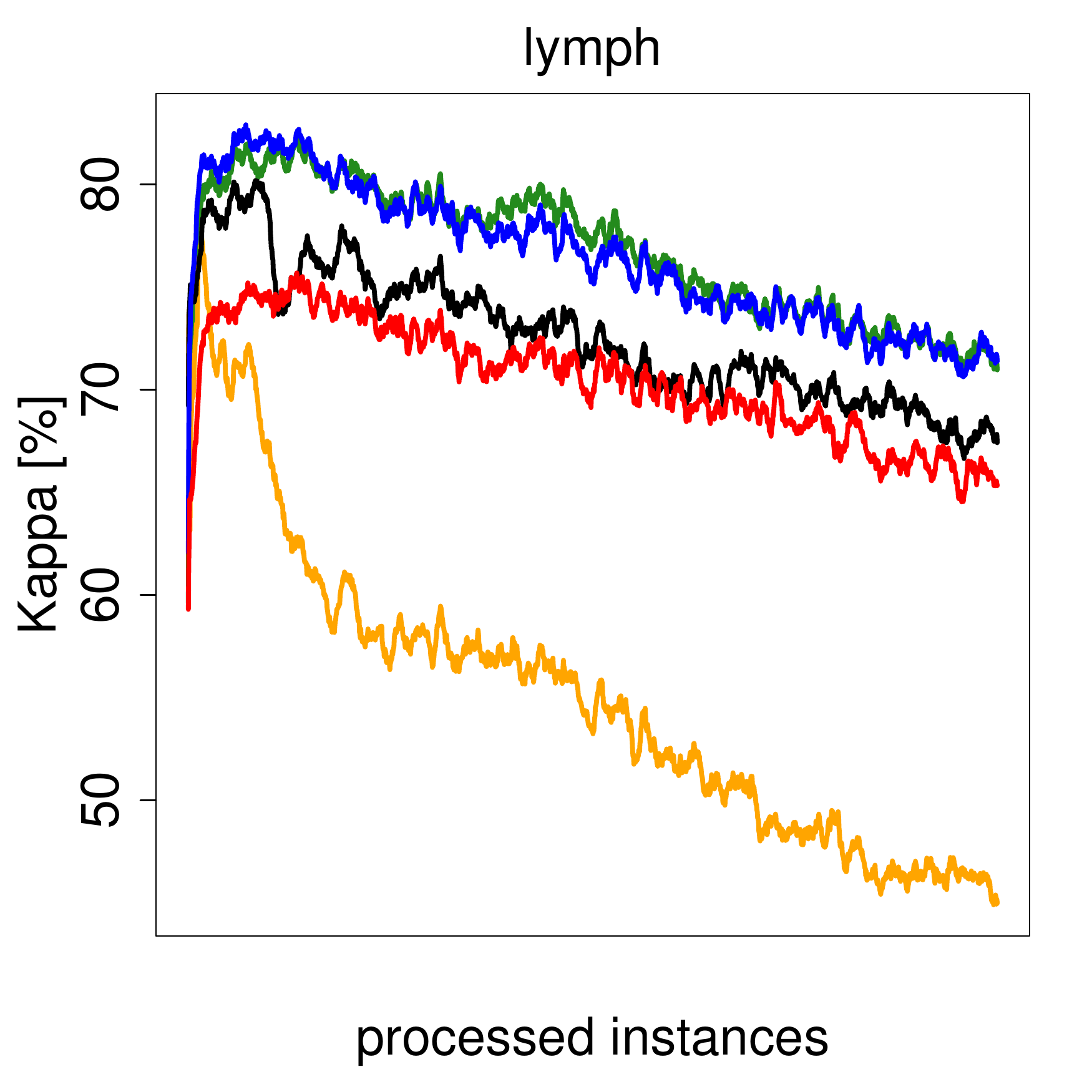}
\includegraphics[width=0.19\columnwidth]{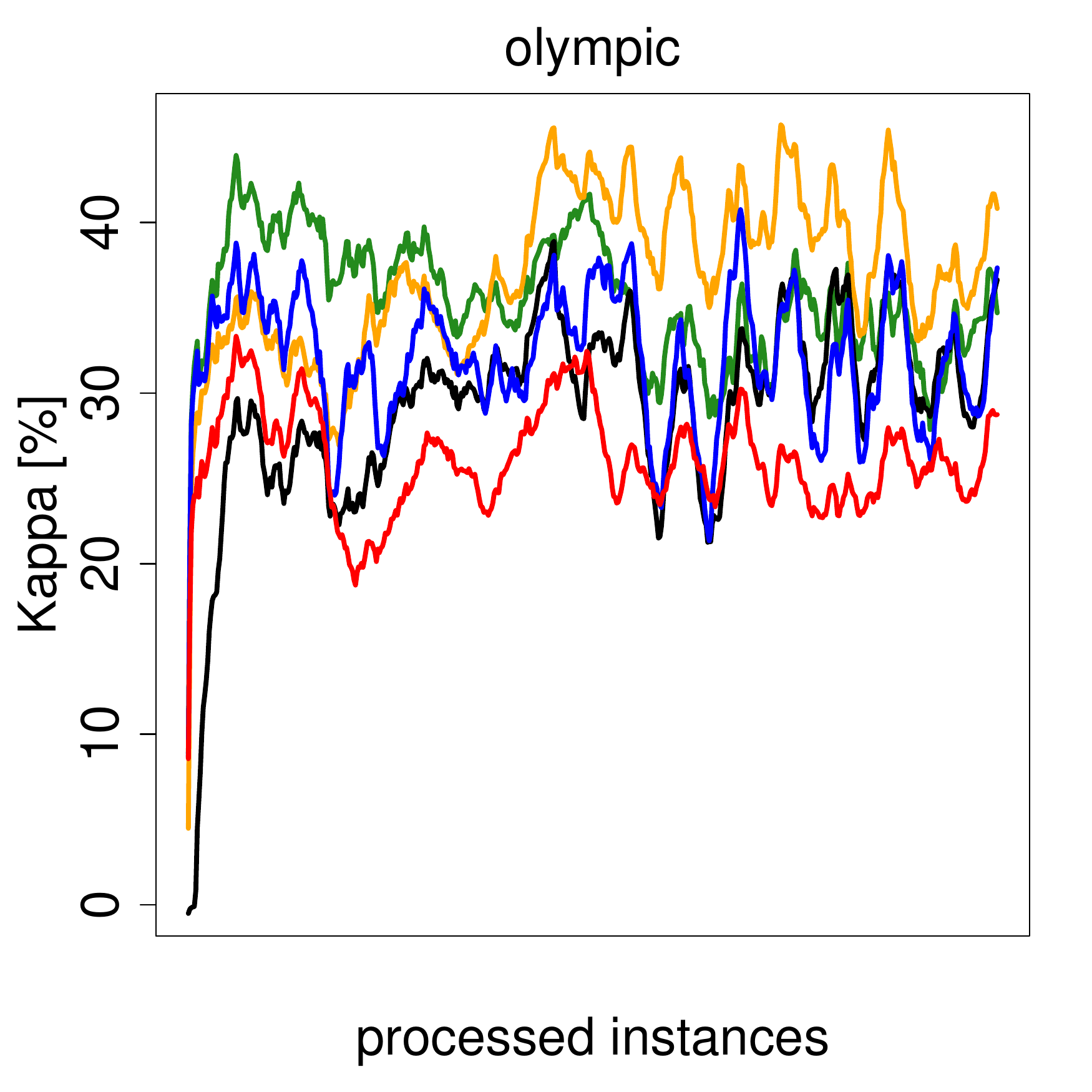}
\includegraphics[width=0.19\columnwidth]{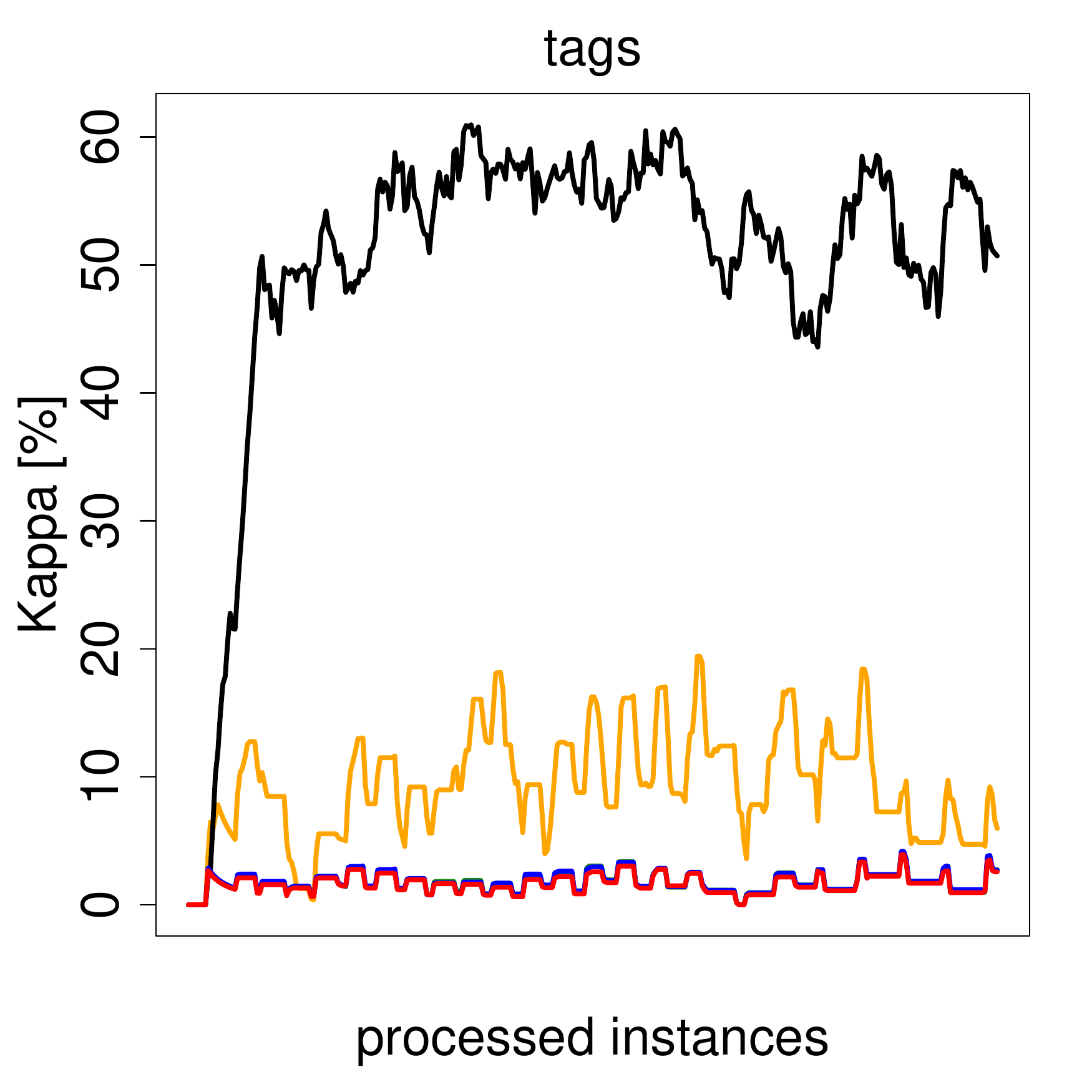}
\includegraphics[width=0.19\columnwidth]{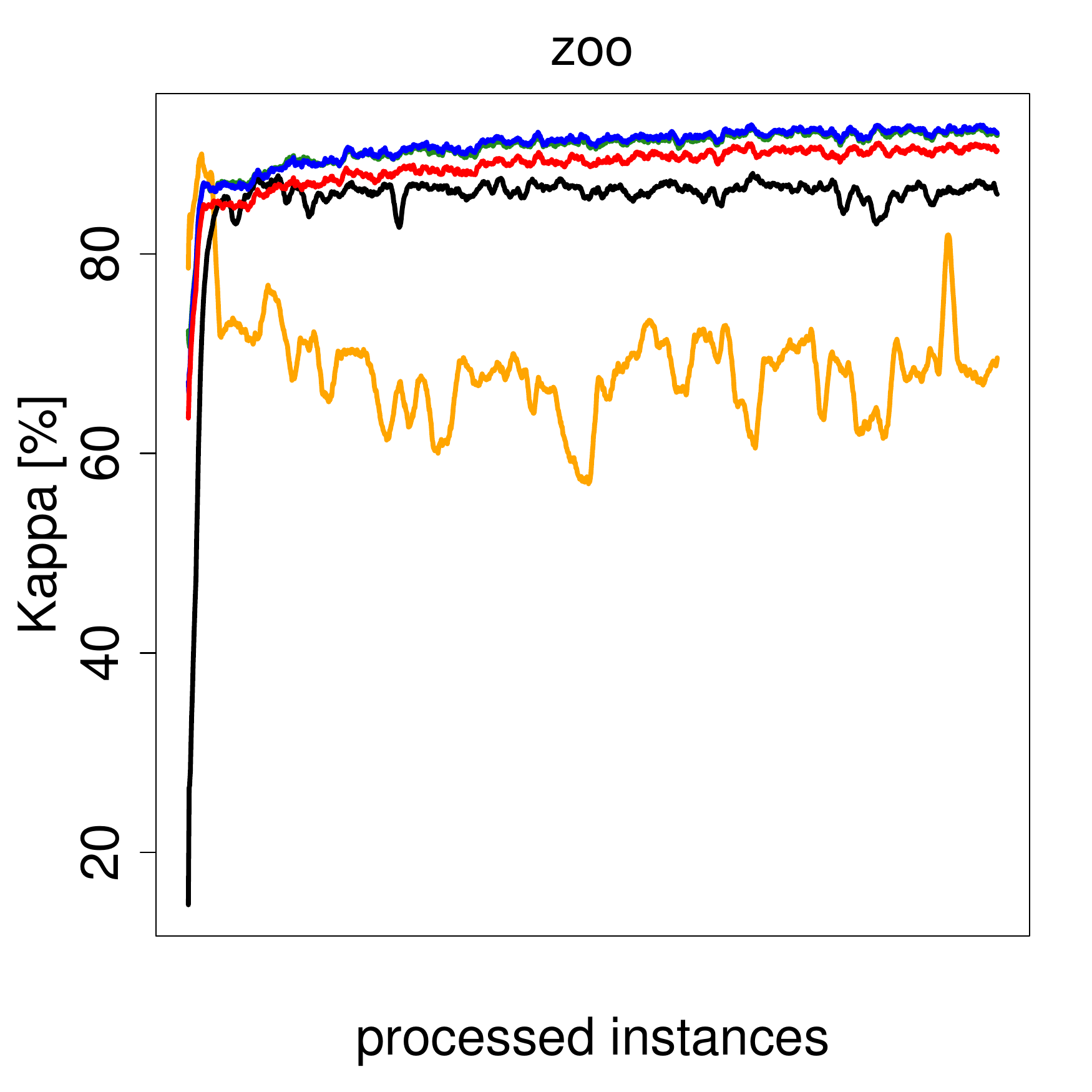}
\caption{PMAUC and Kappa on multi-class imbalanced datasets.}
\label{fig:mc_datasets}
\end{figure}

\begin{figure}[t!]
\centering
\includegraphics[width=0.5\columnwidth]{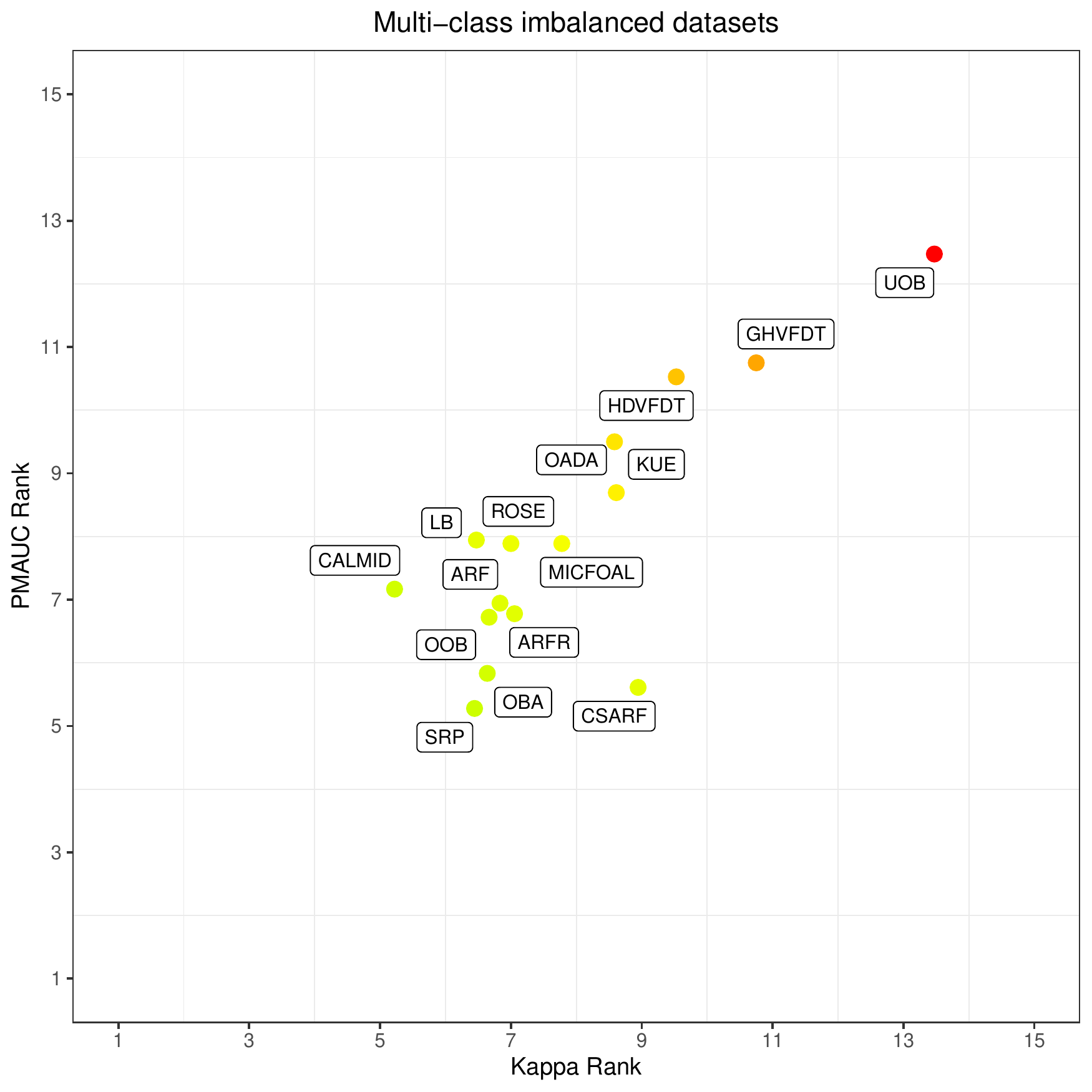}
\caption{Comparison of all 15 algorithms for multi-class imbalanced datasets. Color gradient represents the product of both metrics.}
\label{fig:MC_datasets_scatter}
\end{figure}

\noindent \textbf{Goal of the experiment.}
This experiment was designed to address \textbf{RQ5} and to evaluate the performance of the classifiers on $18$ multi-class real-world imbalanced and drifting data streams. The previous experiments focused on analyzing how the classifiers can deal with multiple learning difficulties in multi-class data streams. This allowed us to examine their behavior in very specific and controlled scenarios. Furthermore, with data stream generators we have full control over the created data, but we cannot generate specific scenarios that are present in real-world scenarios, because they are characterized by merging various learning difficulties at varying frequency and intensity. The real-world data streams employed in the experiment are popular benchmarks for data streams classifiers, and their
specifications are presented in Table~\ref{tab:mc_datasets_spec}. The PMAUC and Kappa for the five selected classifiers are presented in Figure~\ref{fig:mc_datasets}. Table~\ref{tab:MC_datasets} provides the average PMAUC and Kappa for the selected top 10 classifiers for each dataset. Figure~\ref{fig:MC_datasets_scatter} illustrates the overall performance of all classifiers for all real-world datasets.

\noindent \textbf{Characteristics of real-world data streams.} By analyzing the performance of classifiers in real-world datasets it is worth to bring up the difference between artificial streams and real-world imbalance data streams. In real-world datasets data was collected in order to model a specific phenomenon observations and does not hold clear probabilistic mechanisms such as stream generators. Also, in a multi-class real world scenario, relations between features and classes are not so clearly defined as it is on artificial generators. This benchmark allows us to gain insights about the classifiers examining them under real unique and challenging conditions. 

\begin{table}[b!]
\centering
\caption{Real-world multi-class datasets specifications.}
\label{tab:mc_datasets_spec}
\begin{tabular}{@{}lrrr@{}}
\toprule
Dataset & Instances & Features & Classes\\
\midrule
activity & 5,418 & 45 & 6\\
connect-4 & 67,557 & 42 & 3\\
cov-pok-elec & 1,455,525 & 72 & 10\\
covtype & 581,012 & 54 & 7\\
crimes & 878,049 & 3 & 39\\
fars & 100,968 & 29 & 8\\
gas & 13,910 & 128 & 6\\
hypothyroid & 1,000,000 & 29 & 4\\
kddcup & 4,898,431 & 41 & 23\\
kr-vs-k & 28,056 & 6 & 18\\
lymph & 1,000,000 & 18 & 4\\
olympic & 271,116 & 7 & 4\\
poker & 829,201 & 10 & 10\\
sensor & 2,219,803 & 5 & 57\\
shuttle & 57,999 & 9 & 7\\
tags & 164,860 & 4 & 11\\
thyroid & 7,200 & 21 & 3\\
zoo & 1,000,000 & 17 & 7\\
\bottomrule
\end{tabular}
\end{table}

\noindent \textbf{Discussion}

\noindent \textit{Impact of class imbalance approach.} Similar to the previously analyzed binary case, real-world datasets bring a combination of various challenges in addition to the multi-class nature of analyzed streams. However, contrary to our observations from binary experiments, we cannot determine for any of the evaluated classifiers to be better than its peers. Also, it is possible to notice that on average PMAUC was very similar for all classifiers, while Kappa values tend to highlight more differences among algorithms. This shows that Kappa is an effective metric for multi-class imbalanced data streams, allowing us to gain more insight into how each of the algorithms is performing. 

Analyzing the resampling-based approaches, we can see \acrshort{uob} returned unsatisfactory results, confirming our observations regarding its inability to cope with multiple classes. \acrshort{oob} returned much better predictive power, however only for datasets with relatively small number of classes. This confirms our previous observations that blind resampling methods are not suitable for problem characterized by a high number of classes to be learned from. Interestingly, \acrshort{arfr} returned much better results, but on a similar level than standard \acrshort{arf}. This shows that major reason behind the success of \acrshort{arfr} lies not in the chosen informative resampling, but in a good selection of the ensemble architecture.

\newpage
For algorithm-level approaches \acrshort{csarf} remained among the best-performing classifiers, displaying excellent PMAUC metric, but falling behind when it comes to Kappa evaluation. \acrshort{rose}, \acrshort{calmid} and \acrshort{micfoal} presented highly satisfactory results. It is worth to note that \acrshort{rose} did not perform as well as it in previous scenarios, which can be explained by lack of specific learning difficulties in analyzed real world data streams (as \acrshort{rose} excels in very difficult problems). \acrshort{calmid} and \acrshort{micfoal} demonstrated better performance than on artificial domains, showing that their mechanisms lead to good performance over real-world problems.  

\noindent \textit{Impact of ensemble architecture.} While this experiment follows all our previous observations, we should focus on a comparison between general-purpose and skew-insensitive ensembles. Similarly to experiment with high number of classes, we can observe very good performance of general-purpose ensembles on real-world imbalanced benchmarks. \acrshort{csarf} displayed the best results in real-world datasets regarding PMAUC but the worst regarding Kappa. This shows that in the analyzed benchmarks adaptation to change and ability to better separate classes in lower-dimensional subspaces can return at least as good performance as dedicated mechanisms for tackling class imbalance.

\begin{table*}[t!]
\centering
\footnotesize
\setlength{\tabcolsep}{4pt}
\caption{PMAUC and Kappa on multi-class imbalanced datasets.}
\label{tab:MC_datasets}
\begin{tabular}{ll|C{1cm}C{1cm}C{1cm}C{1cm}C{1cm}C{1cm}C{1cm}C{1cm}C{1cm}C{1cm}}
\toprule
& Dataset & CSARF & ARF & KUE & LB & SRP & CALMID & MICFOAL & ROSE & ARFR & OOB\\
\midrule
\multirow{18}{*}{\rotatebox[origin=c]{90}{PMAUC}} 
 &  activity & 71.85 & 72.06 & 58.74 & 71.33 & 71.42 & 61.05 & 60.31 & 68.04 & 71.98 & \textbf{73.01}\\
 &  connect-4 & 77.71 & 78.10 & 70.72 & 76.26 & \textbf{80.01} & 72.79 & 74.17 & 77.32 & 76.60 & 76.06\\
 &  cov-pok-elec & 6.91 & 6.91 & 7.12 & 6.91 & 6.91 & \textbf{9.16} & 6.93 & 7.27 & 6.91 & 7.28\\
 &  covtype & 29.92 & 29.82 & 28.11 & 29.13 & 29.75 & \textbf{31.04} & 29.97 & 28.84 & 29.80 & 27.00\\
 &  crimes & 48.10 & 48.37 & 47.07 & 47.69 & \textbf{48.61} & 47.29 & 47.56 & 46.18 & 48.47 & 44.60\\
 &  fars & \textbf{45.66} & 35.95 & 44.59 & 36.31 & 37.15 & 42.39 & 41.07 & 38.99 & 41.10 & 45.45\\
 &  gas & \textbf{0.01} & \textbf{0.01} & \textbf{0.01} & \textbf{0.01} & \textbf{0.01} & \textbf{0.01} & \textbf{0.01} & \textbf{0.01} & \textbf{0.01} & \textbf{0.01}\\
 &  hypothyroid & \textbf{0.14} & \textbf{0.14} & 0.13 & \textbf{0.14} & \textbf{0.14} & \textbf{0.14} & \textbf{0.14} & \textbf{0.14} & \textbf{0.14} & 0.13\\
 &  kddcup & \textbf{98.12} & 96.77 & 96.58 & 95.91 & 97.47 & 95.75 & 96.92 & 94.28 & 97.06 & 97.39\\
 &  kr-vs-k & 6.41 & 6.40 & 6.66 & 6.47 & 6.38 & \textbf{9.90} & 6.41 & 7.07 & 6.41 & 6.79\\
 &  lymph & 24.29 & 23.50 & 26.66 & 20.97 & 27.13 & \textbf{31.11} & 24.01 & 22.22 & 27.34 & 28.13\\
 &  olympic & 98.64 & 98.36 & 92.39 & 97.38 & \textbf{98.74} & 95.80 & 97.82 & 97.75 & 98.23 & 97.53\\
 &  poker & 98.28 & 98.21 & 98.32 & 95.06 & \textbf{99.12} & 94.87 & 97.58 & 94.30 & 98.15 & 97.95\\
 &  sensor & 23.64 & 23.62 & 24.42 & 24.81 & 25.87 & 23.31 & 23.68 & 24.57 & 23.63 & \textbf{26.42}\\
 &  shuttle & 39.36 & 39.39 & 32.10 & \textbf{41.88} & 41.37 & 39.57 & 40.28 & 36.67 & 38.69 & 36.45\\
 &  tags & \textbf{78.27} & 76.70 & 71.59 & 74.49 & 73.54 & 64.39 & 67.68 & 74.28 & 75.75 & 66.15\\
 &  thyroid & 34.04 & 34.04 & 33.97 & 34.04 & \textbf{35.25} & 34.03 & 34.03 & 33.99 & 34.04 & 33.64\\
 &  zoo & 4.40 & 4.39 & 5.90 & 5.48 & 4.67 & \textbf{6.53} & 6.14 & 4.82 & 4.39 & 6.29\\
\midrule
\multirow{18}{*}{\rotatebox[origin=c]{90}{Kappa}} 
 &  activity & 59.43 & 59.62 & 29.23 & 51.77 & 57.96 & 36.30 & 34.36 & 43.21 & \textbf{61.37} & 53.05\\
 &  connect-4 & 36.88 & 32.23 & 23.75 & 33.63 & 33.53 & 34.79 & 30.81 & \textbf{41.70} & 35.53 & 37.42\\
 &  cov-pok-elec & 0.19 & 0.32 & 1.38 & 0.34 & 0.17 & \textbf{34.38} & 0.32 & 4.95 & 0.33 & 3.99\\
 &  covtype & 47.91 & 52.00 & 38.73 & 43.89 & 52.46 & \textbf{78.25} & 60.98 & 41.61 & 52.25 & 34.60\\
 &  crimes & 50.40 & 66.25 & 62.21 & 65.73 & 68.38 & 66.51 & 62.09 & 47.53 & \textbf{70.31} & 50.91\\
 &  fars & 11.53 & 10.59 & 8.24 & 10.26 & 7.61 & 13.98 & 13.80 & 14.78 & 0.84 & \textbf{21.59}\\
 &  gas & 0.01 & 0.02 & 0.02 & 0.05 & 0.05 & 0.13 & 0.03 & \textbf{0.14} & 0.05 & 0.03\\
 &  hypothyroid & 1.42 & 1.24 & -0.04 & \textbf{1.45} & 1.34 & 1.29 & 1.14 & 1.43 & 1.21 & -0.49\\
 &  kddcup & 70.18 & 76.88 & 77.14 & 69.06 & \textbf{82.22} & 71.68 & 80.04 & 55.16 & 76.80 & 80.75\\
 &  kr-vs-k & 0.46 & 0.67 & 3.28 & 0.87 & 0.20 & \textbf{51.27} & 0.25 & 9.36 & 0.60 & 4.41\\
 &  lymph & -4.15 & 1.63 & 22.86 & 0.33 & 34.71 & \textbf{66.38} & 17.65 & 0.96 & 13.83 & 18.56\\
 &  olympic & 73.92 & 87.08 & 75.49 & \textbf{88.07} & 87.81 & 76.72 & 72.91 & 86.93 & 78.47 & 83.83\\
 &  poker & 88.86 & 90.81 & 90.77 & 84.98 & \textbf{93.65} & 85.41 & 89.35 & 68.97 & 90.59 & 89.63\\
 &  sensor & 0.14 & 0.36 & 1.37 & 2.37 & \textbf{4.40} & 1.18 & 0.01 & 2.01 & 0.37 & 4.35\\
 &  shuttle & 72.19 & 71.84 & 30.42 & 68.24 & \textbf{82.98} & 54.16 & 79.39 & 47.16 & 64.71 & 39.02\\
 &  tags & 26.20 & 32.25 & 24.80 & 34.93 & 8.80 & 29.74 & 30.71 & \textbf{37.29} & 35.57 & 26.97\\
 &  thyroid & 3.08 & 3.21 & 2.05 & \textbf{3.47} & 0.00 & 3.38 & 3.21 & 2.84 & 3.20 & 0.37\\
 &  zoo & 1.79 & 1.94 & 24.51 & 18.20 & 7.27 & \textbf{50.95} & 38.25 & 10.03 & 1.89 & 28.44\\
\midrule
\multicolumn{2}{l|}{Avg. PMAUC} & \textbf{43.65} & 42.93 & 41.39 & 42.46 & 43.53 & 42.17 & 41.93 & 42.04 & 43.26 & 42.79\\
\multicolumn{2}{l|}{Avg. Kappa} &  30.02 & 32.72 & 28.68 & 32.09 & 34.64 & \textbf{42.03} & 34.18 & 28.67 & 32.66 & 32.08\\
\midrule
\multicolumn{2}{l|}{Rank PMAUC} & \textbf{4.06} & 5.22 & 6.89 & 6.11 & 4.17 & 5.56 & 6.11 & 6.17 & 5.11 & 5.61\\
\multicolumn{2}{l|}{Rank Kappa} & 7.00 & 5.39 & 6.83 & 5.06 & 4.94 & \textbf{4.11} & 6.00 & 5.17 & 5.22 & 5.28\\
\bottomrule
\end{tabular}
\end{table*}

\subsubsection{Semi-synthetic multi-class imbalanced datasets}
\label{sec:mc-semisynthetic}

\begin{figure}[t!]
\centering
\includegraphics[width=0.19\columnwidth]{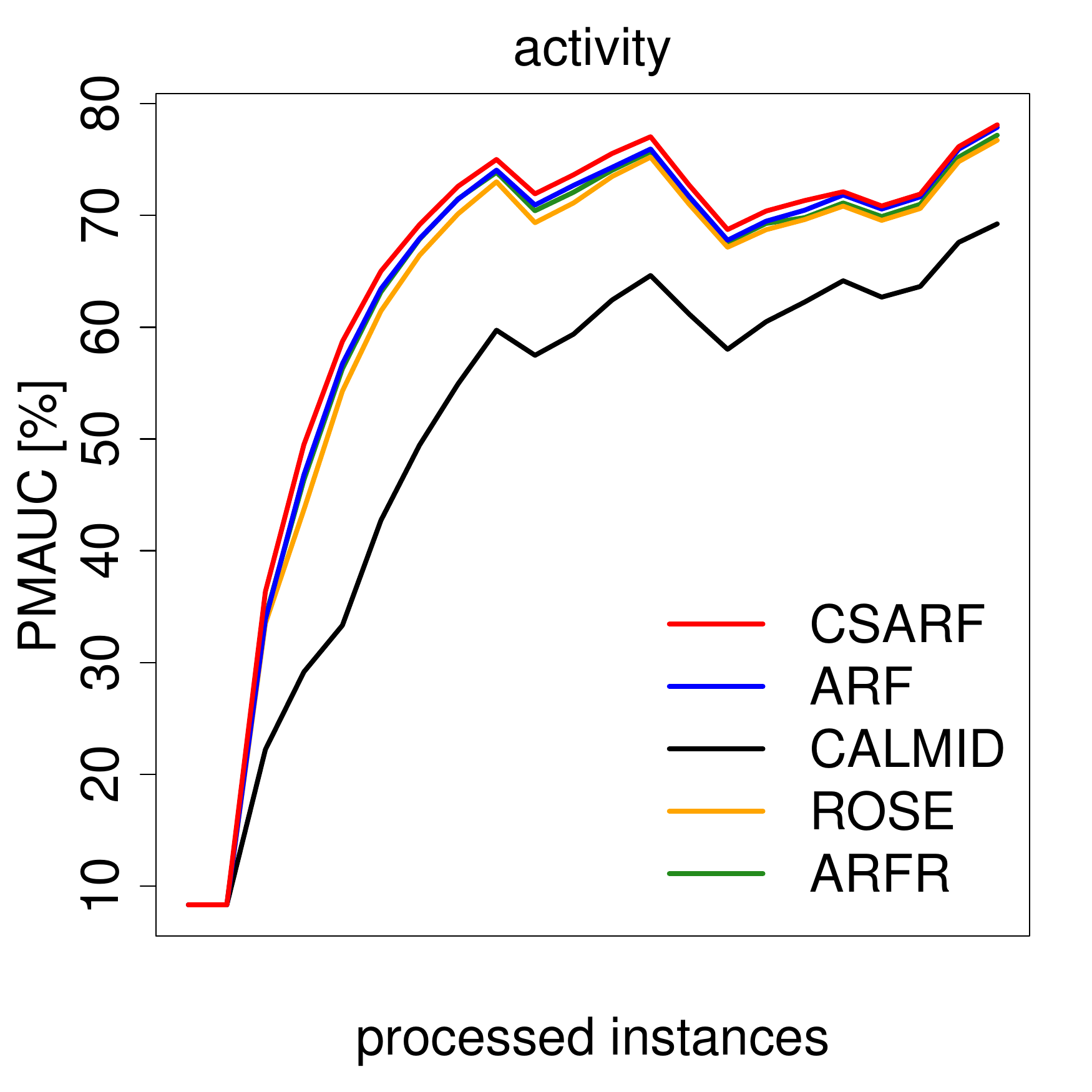}
\includegraphics[width=0.19\columnwidth]{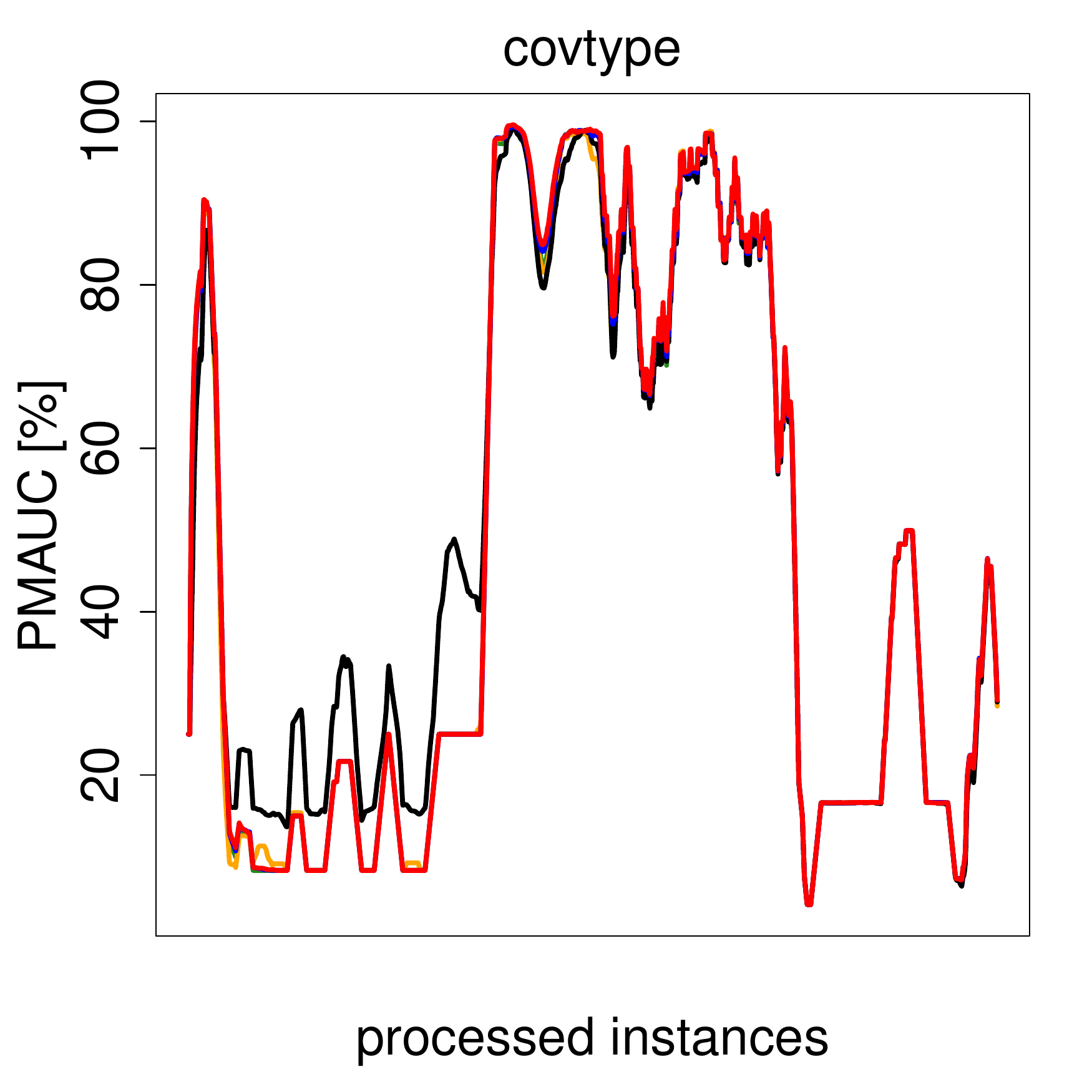}
\includegraphics[width=0.19\columnwidth]{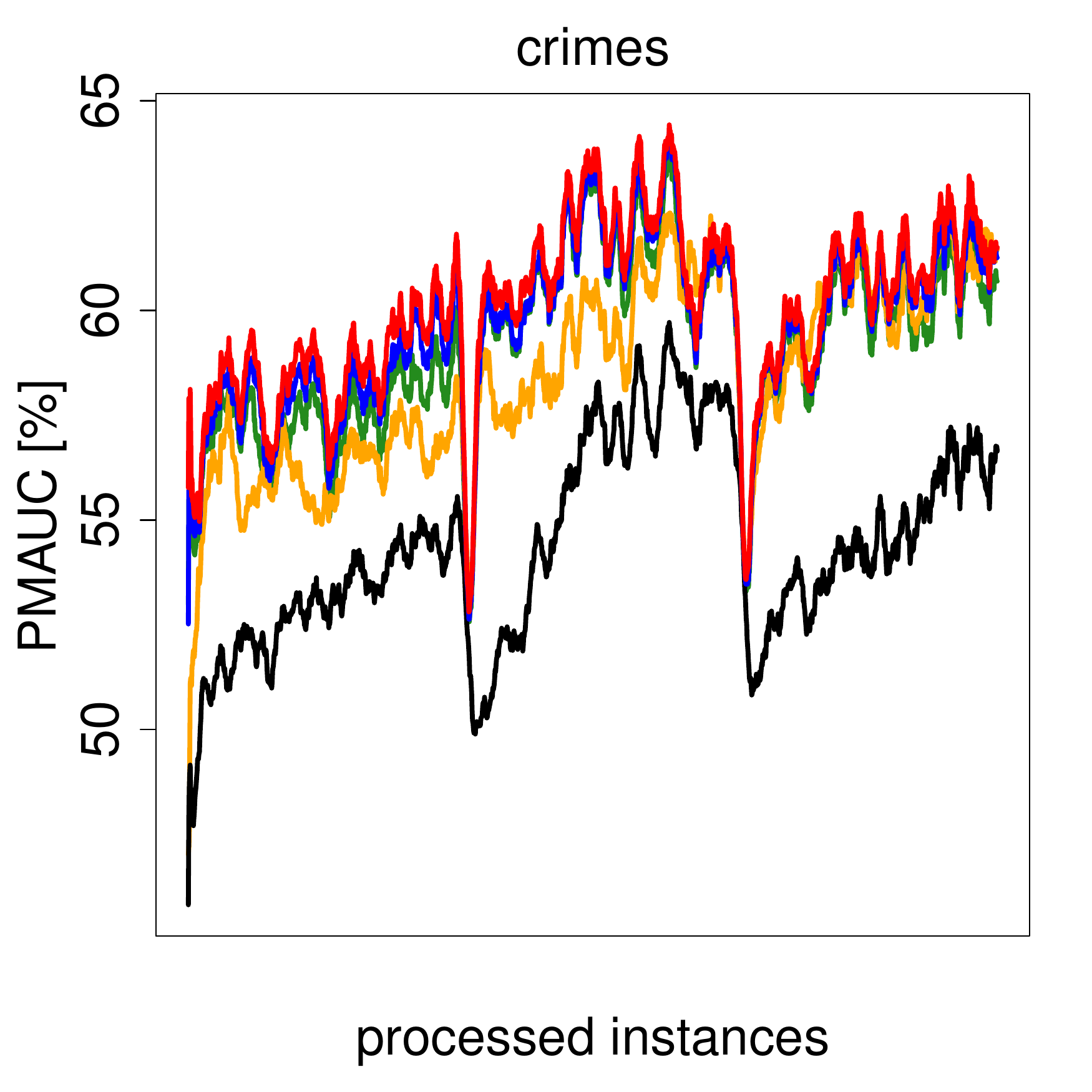}
\includegraphics[width=0.19\columnwidth]{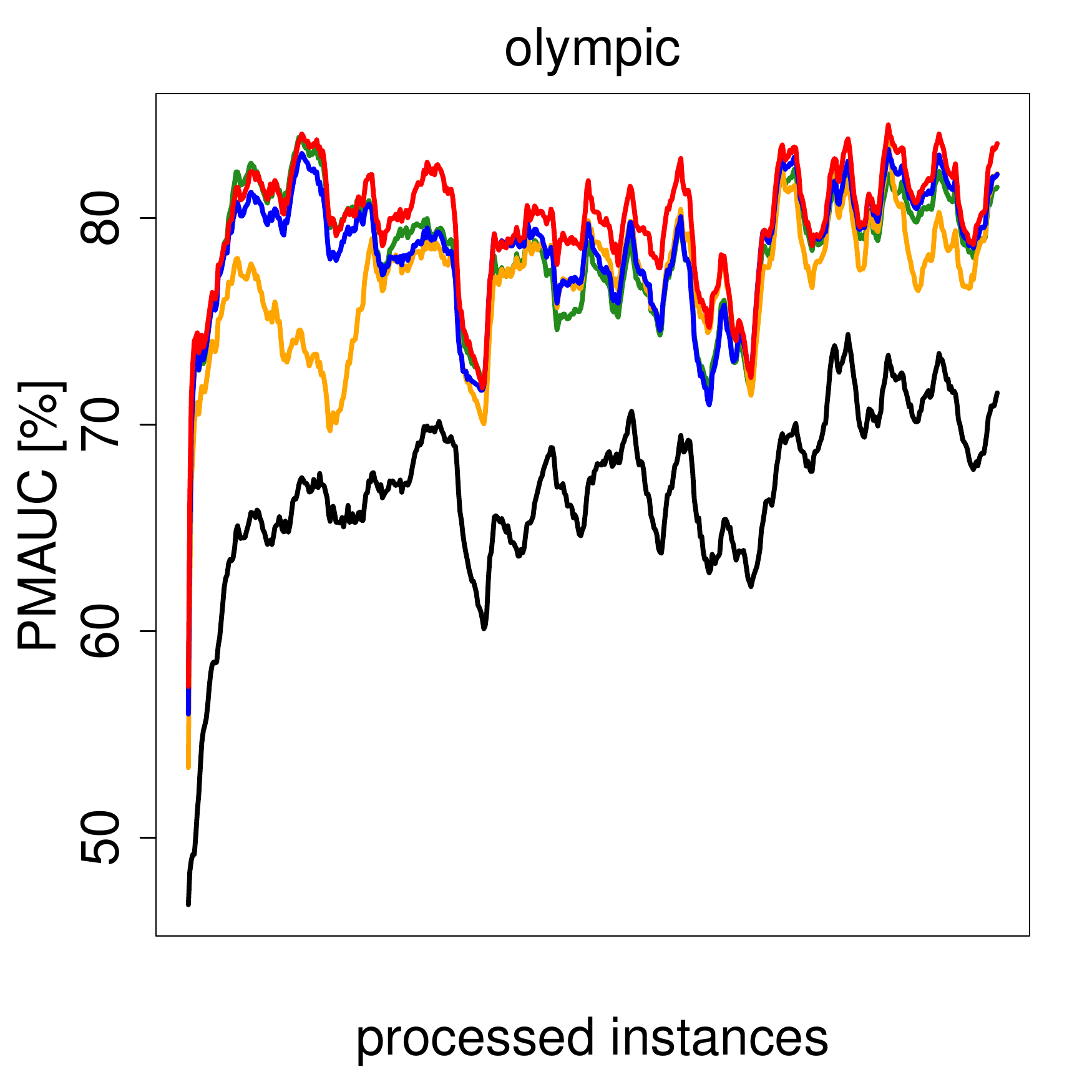}
\includegraphics[width=0.19\columnwidth]{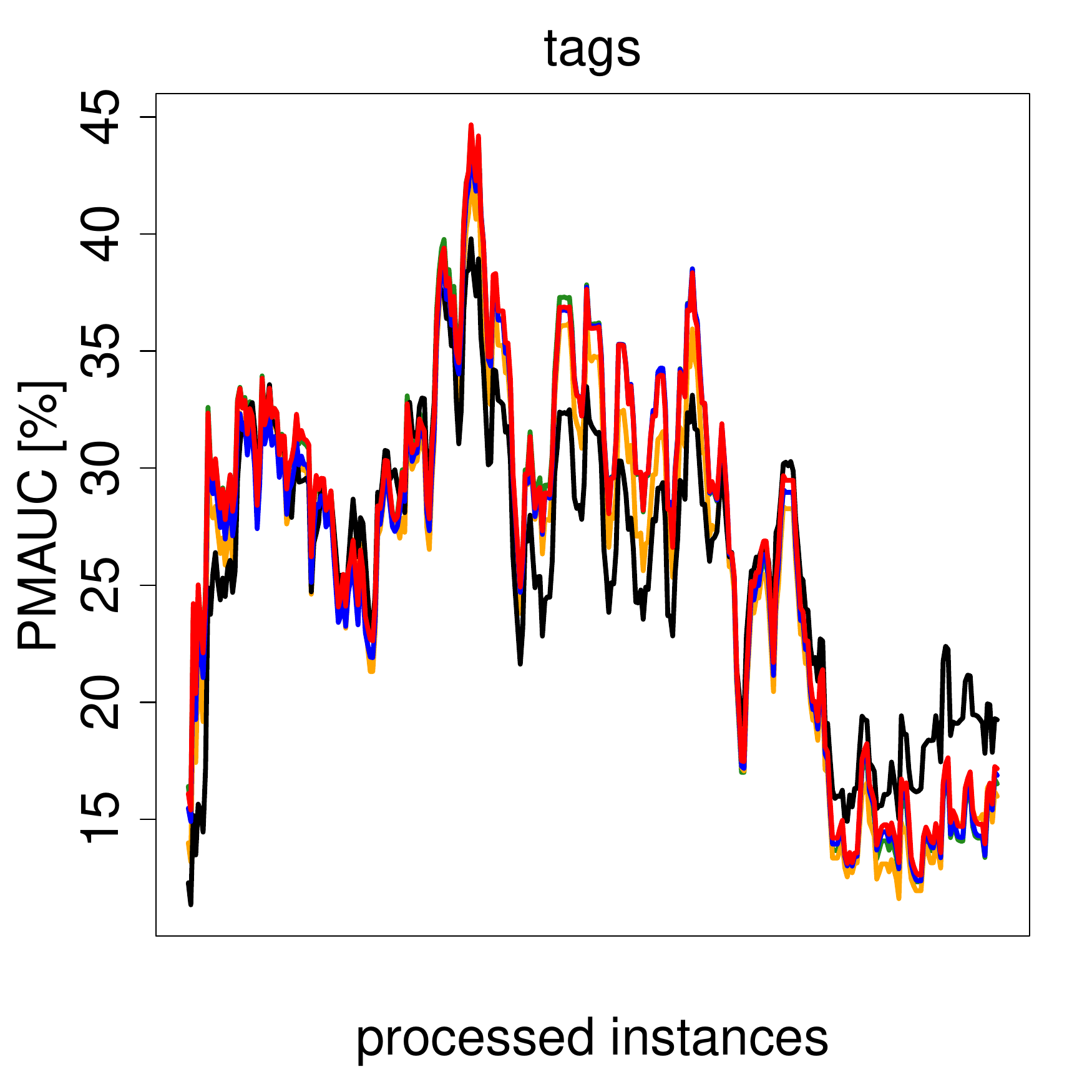}
\includegraphics[width=0.19\columnwidth]{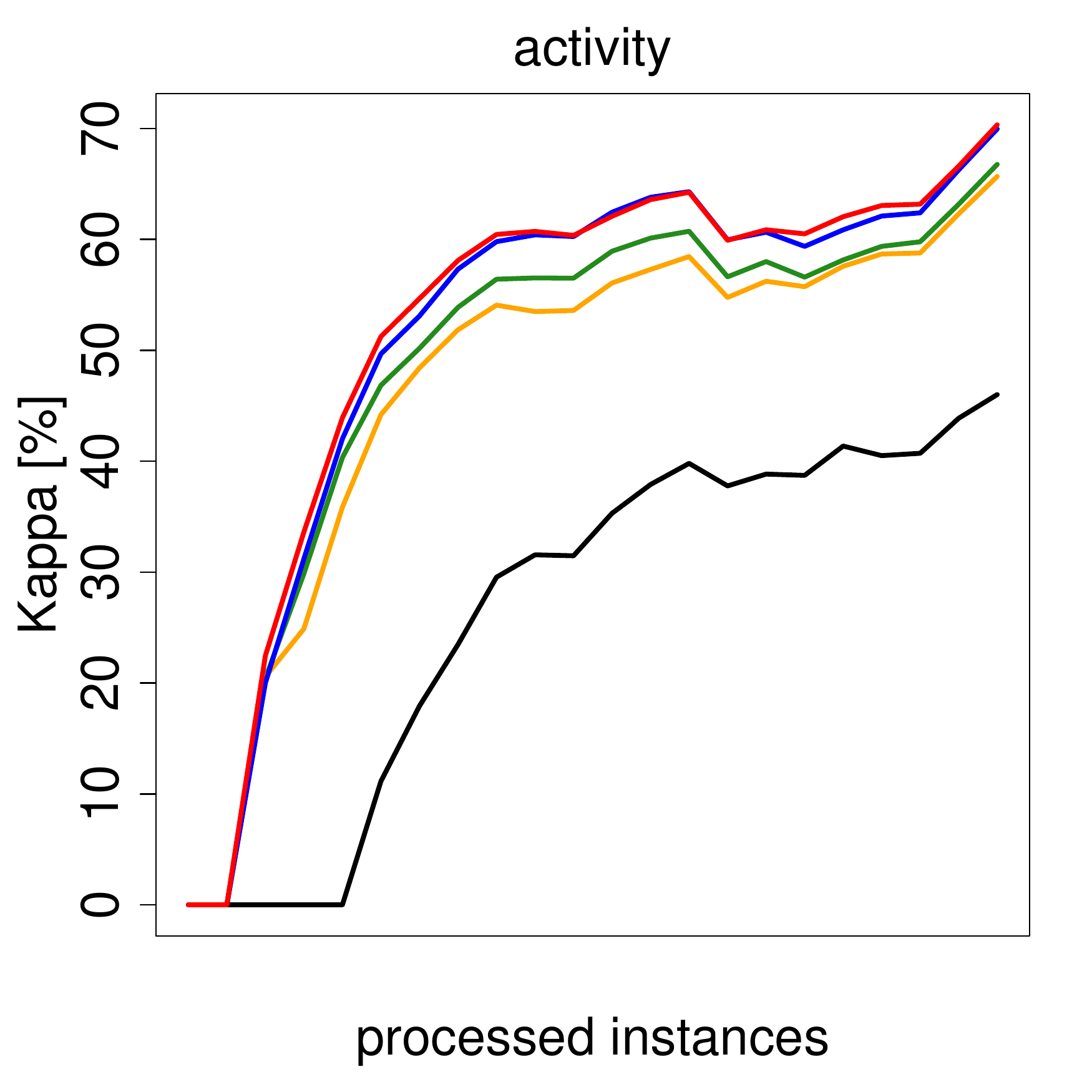}
\includegraphics[width=0.19\columnwidth]{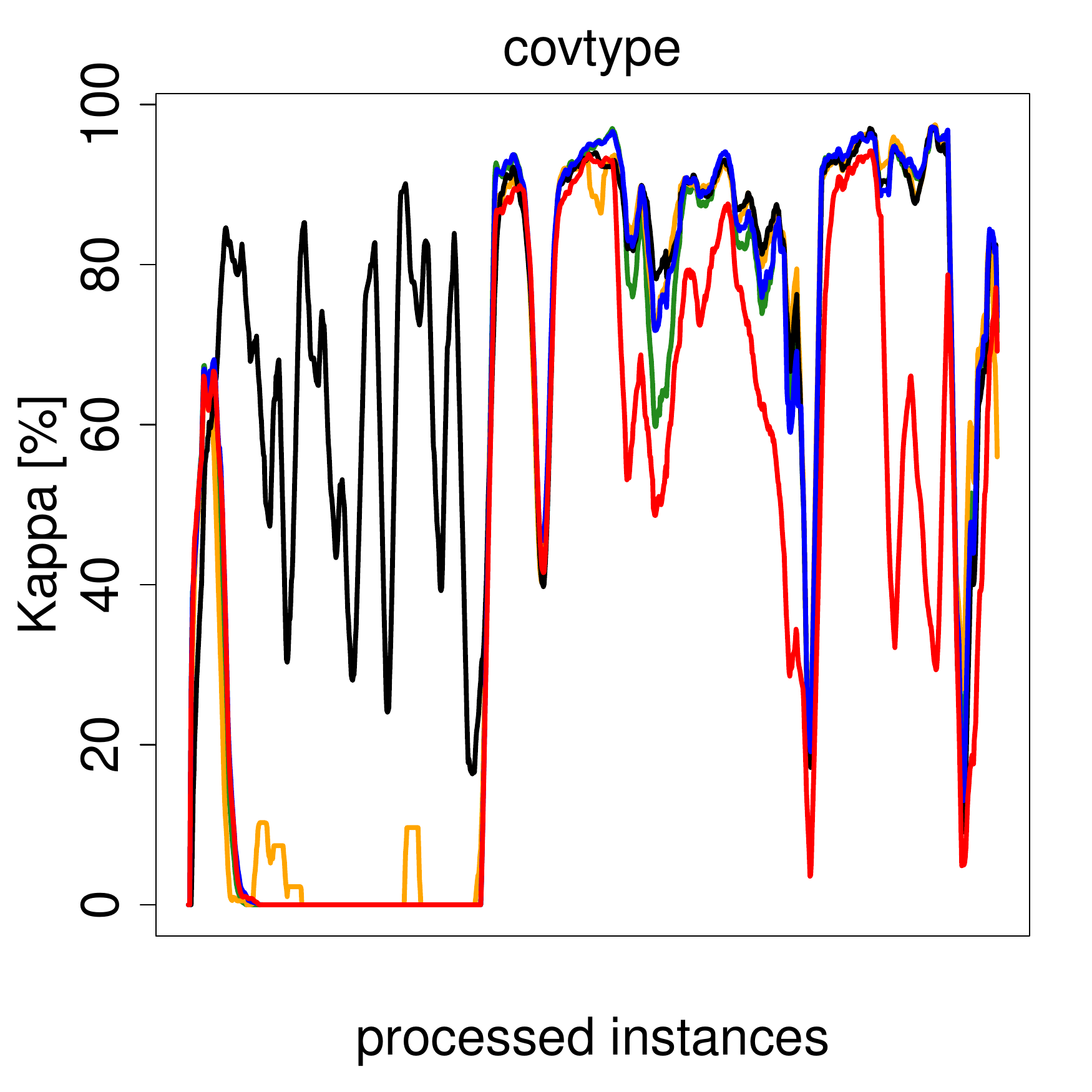}
\includegraphics[width=0.19\columnwidth]{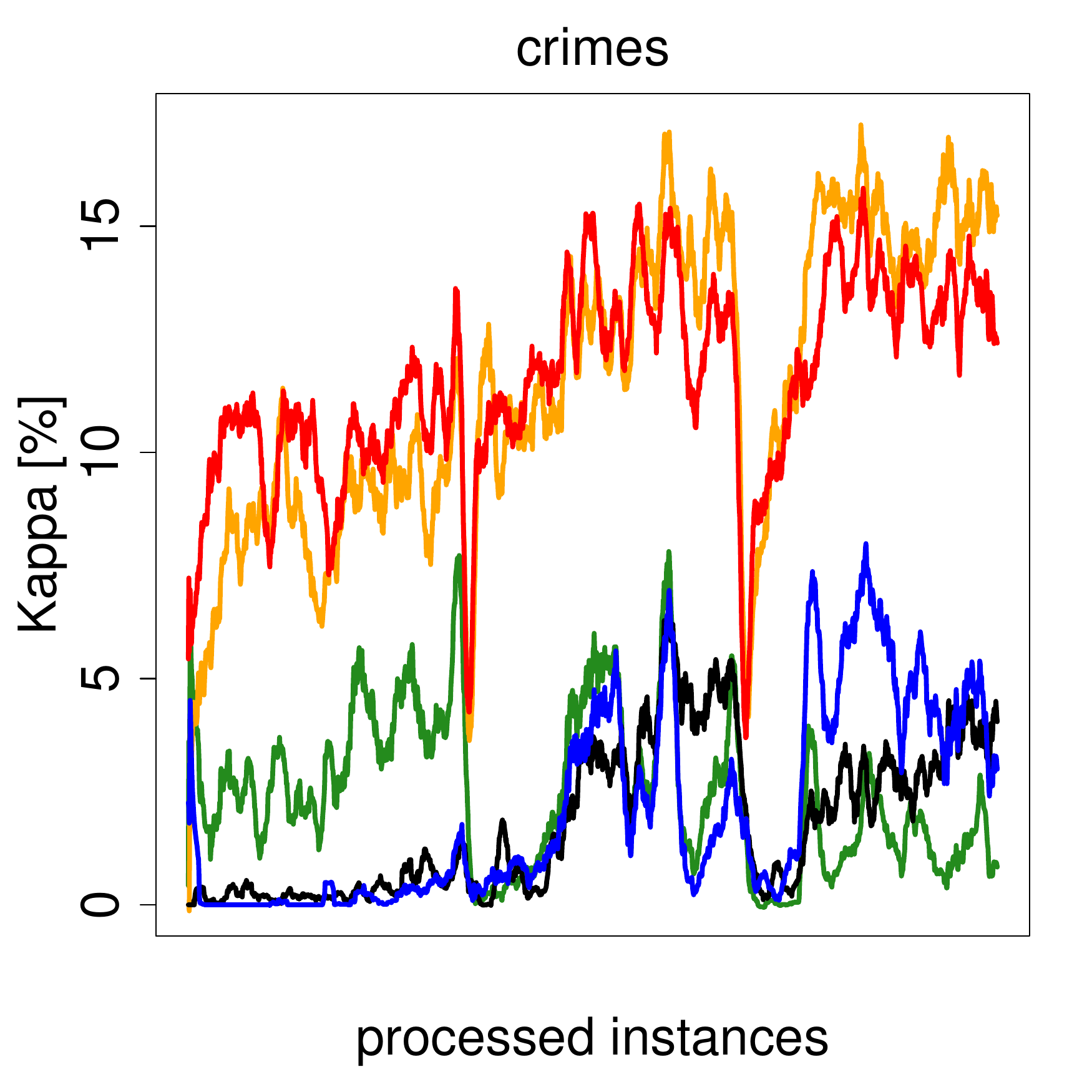}
\includegraphics[width=0.19\columnwidth]{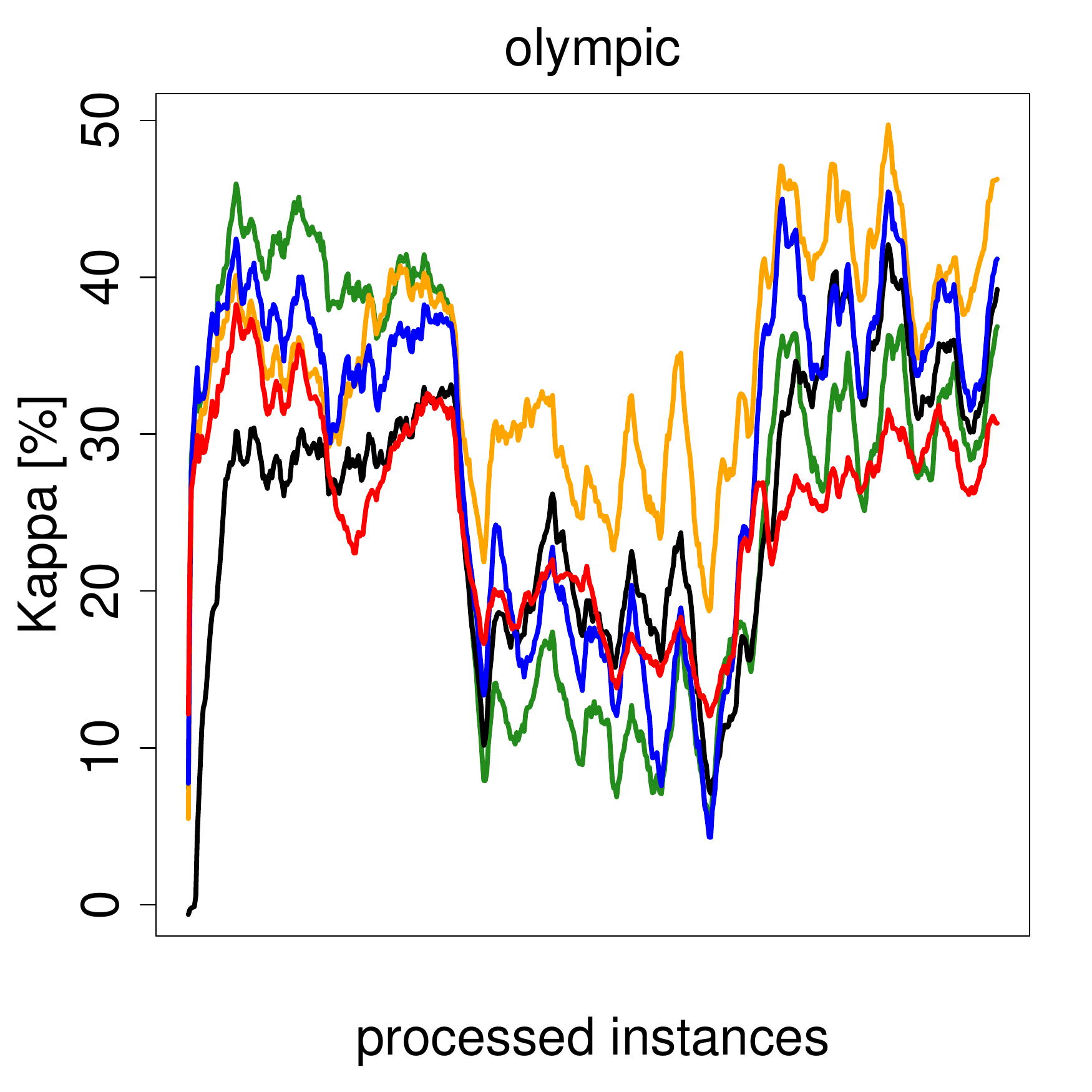}
\includegraphics[width=0.19\columnwidth]{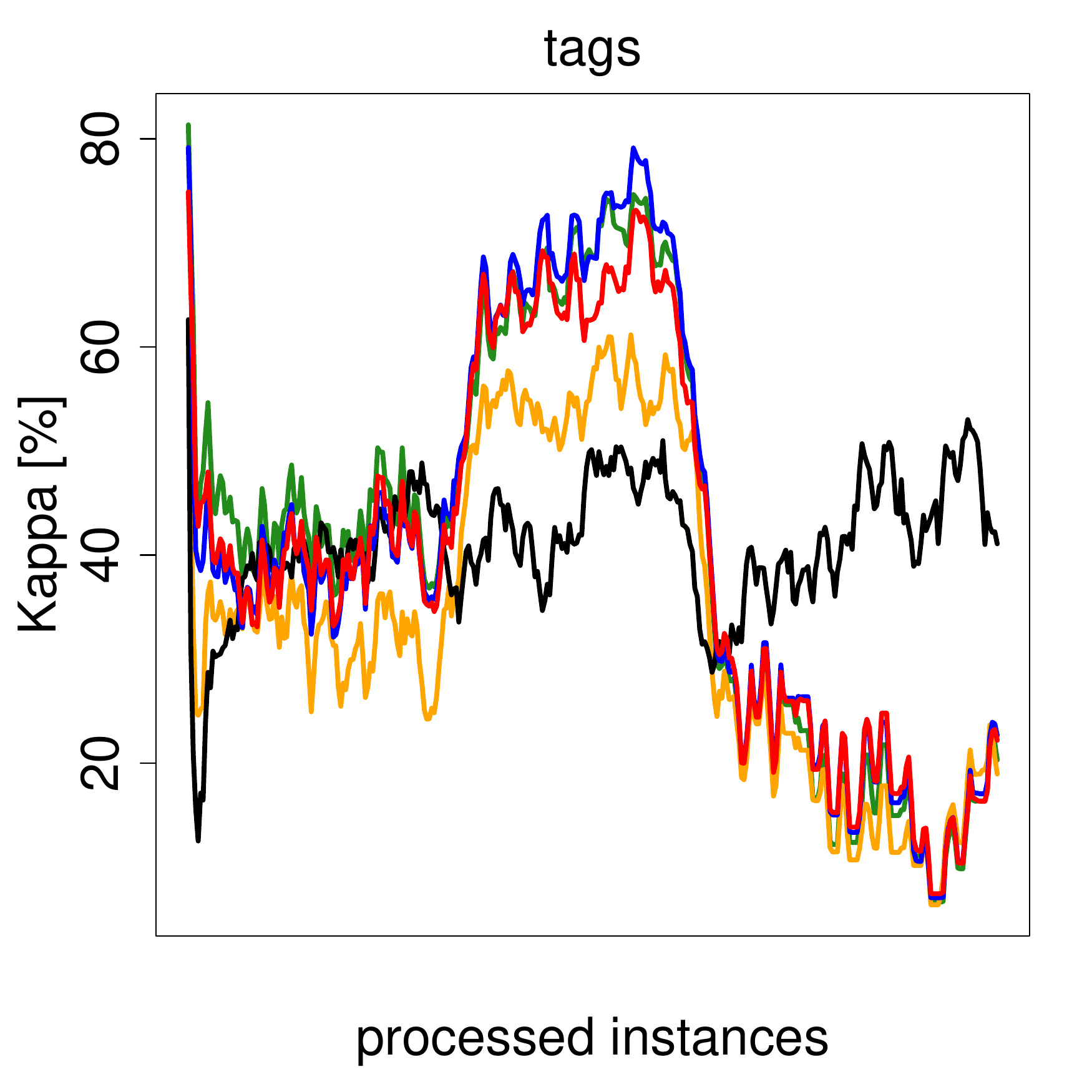}
\caption{PMAUC and Kappa on semi-synthetic multi-class imbalanced datasets.}
\label{fig:mc_semisynthetic}
\end{figure}

\begin{figure}[t!]
\centering
\includegraphics[width=0.5\columnwidth]{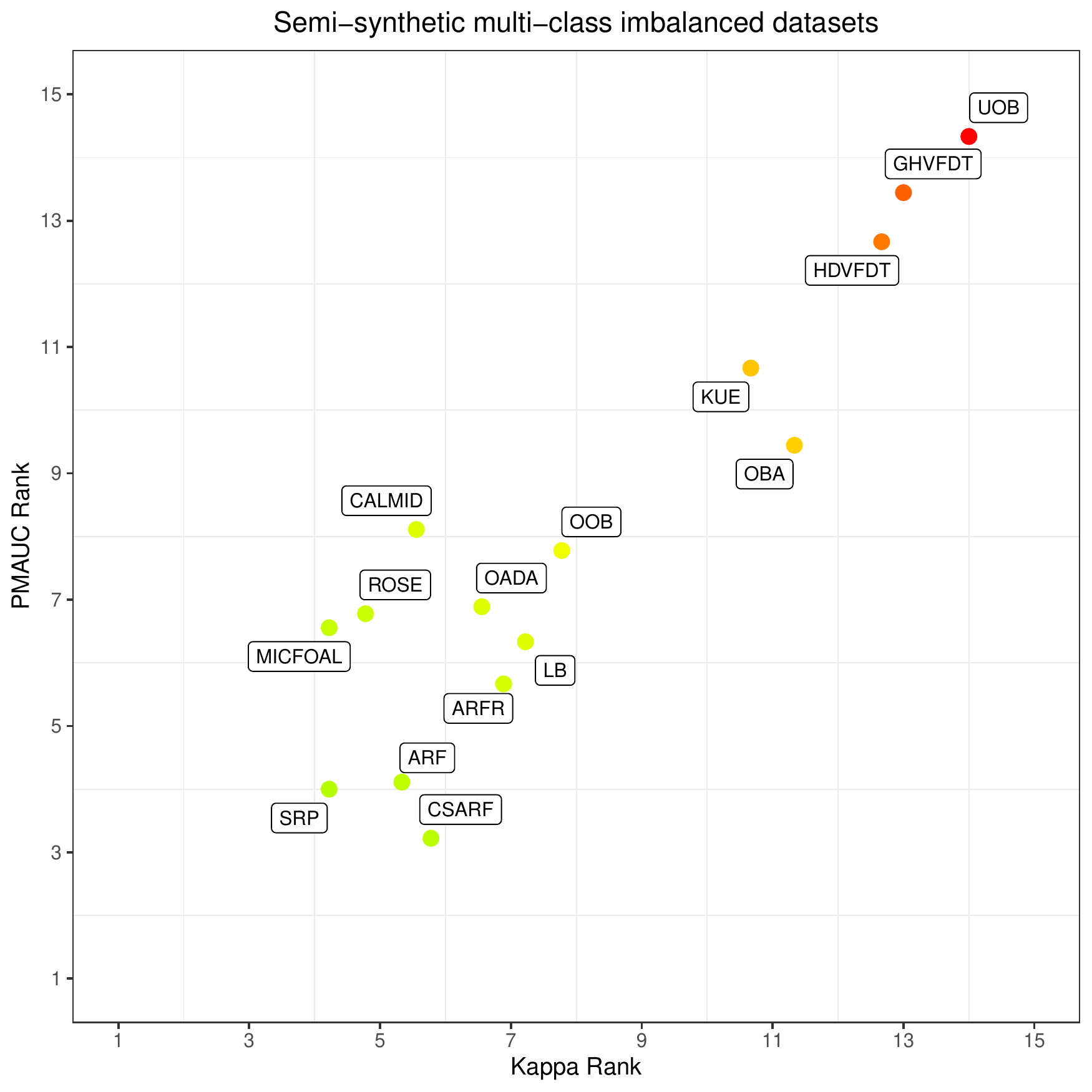}
\caption{Comparison of all 15 algorithms for semi-synthetic multi-class imbalanced datasets. Color gradient represents the product of both metrics.}
\label{fig:MC_semisynthetic_scatter}
\end{figure}

\noindent \textbf{Goal of the experiment.}
This experiment was designed to address more in-depth \textbf{RQ5} and to evaluate the robustness of the classifiers to semi-synthetic data streams \citep{Korycki2020}. We used all $9$ multi-class semi-synthetic data streams proposed in~\footnote{https://github.com/mlrep/imb-drift-20}. These benchmarks simulate critical class ratio changes and concept drifts. This allows us to analyze how the classifiers are able to cope with dynamic changes and concept drifts with real-world data streams, how they are able to adapt to those changes. Figure~\ref{fig:mc_semisynthetic} illustrates the performance of five selected algorithms in the semi-synthetic data streams. Table~\ref{tab:MC_semisynthetic} presents the average PMAUC and Kappa for the top 10 classifiers for each of the evaluated streams and the overall rank of the algorithms. Figure~\ref{fig:MC_semisynthetic_scatter} provides a comparison of all algorithms.

\noindent \textbf{Discussion}

\noindent \textit{Impact of class imbalance approach.} Semi-synthetic benchmarks allowed us to use real-world data to create much more challenging scenarios with rapidly evolving imbalance ratios. Thus, we preserved the desirable characteristics of real-world problems (such as mixed types of drift) but enhanced them with much more challenging problem from the imbalance standpoint. When analyzing the results, we can see that all classifiers formed two clusters when looking at their predictive performance.

For resampling-based methods, we can see that \acrshort{uob} and \acrshort{oob} returned opposite performance, despite them sharing similar core. Here, we can see the superiority of oversampling, which confirms observations found in \citep{Korycki2020}, where authors of these semi-synthetic benchmarks postulated that smart oversampling is the best solution. \acrshort{arfr} again returned very similar performance to standard \acrshort{arfr}, highlighting that its predictive power can mainly be attributed to its robust core design.

For algorithm-level methods \acrshort{csarf} achieved the best-performing classifier regarding PMAUC, while surprisingly displaying good results on Kappa. \acrshort{rose}, \acrshort{calmid} and \acrshort{micfoal} displayed great performance, showing that their hybrid mechanisms are capable of efficient handling of rapid changes in imbalance ratios within real-world datasets.

\noindent \textit{Impact of ensemble architecture.} By adding sudden and extreme changes in real-world benchmarks datasets, we could see an increase in the gap between best and worst performing methods. Similarly, to the previous experiments we can observe an excellent performance returned by \acrshort{srp}, showing a significant potential in using low-dimensional representations for imbalanced data streams, direction so far only explored in \citep{korycki2021low}.

\begin{table*}[t!]
\centering
\footnotesize
\setlength{\tabcolsep}{4pt}
\caption{PMAUC and Kappa on semi-synthetic multi-class imbalanced datasets.}
\label{tab:MC_semisynthetic}
\begin{tabular}{ll|C{1cm}C{1cm}C{1cm}C{1cm}C{1cm}C{1cm}C{1cm}C{1cm}C{1cm}C{1cm}}
\toprule
& Dataset & CSARF & ARF & KUE & LB & SRP & CALMID & MICFOAL & ROSE & ARFR & OOB\\
\midrule
\multirow{9}{*}{\rotatebox[origin=c]{90}{PMAUC}} 
 &  activity & 71.68 & 71.47 & 55.74 & 70.73 & \textbf{71.78} & 63.13 & 62.28 & 70.63 & 70.69 & 69.48\\
 &  connect-4 & 77.56 & 77.95 & 66.96 & 76.00 & \textbf{79.92} & 72.19 & 72.84 & 77.22 & 76.25 & 72.89\\
 &  covtype & 46.79 & 46.65 & 45.64 & 46.14 & 46.99 & \textbf{48.93} & 46.86 & 46.53 & 46.51 & 48.70\\
 &  crimes & 60.08 & 59.70 & 55.82 & 55.83 & \textbf{61.68} & 54.37 & 56.79 & 58.54 & 59.25 & 54.96\\
 &  gas & 44.62 & \textbf{45.47} & 33.15 & 42.60 & 44.97 & 42.31 & 43.06 & 43.73 & 44.16 & 36.55\\
 &  olympic & \textbf{80.05} & 78.74 & 71.48 & 76.39 & 75.54 & 67.43 & 70.23 & 76.98 & 78.62 & 65.77\\
 &  poker & 28.53 & 27.69 & 28.85 & 27.65 & 28.51 & 29.55 & 28.90 & 26.74 & 26.13 & \textbf{29.64}\\
 &  sensor & \textbf{43.81} & 43.68 & 42.03 & 43.64 & 41.72 & 43.38 & 43.63 & 43.11 & 43.69 & 43.50\\
 &  tags & 26.91 & 26.54 & 21.56 & 27.74 & 27.55 & 25.98 & \textbf{28.29} & 25.79 & 26.70 & 23.58\\
\midrule
\multirow{9}{*}{\rotatebox[origin=c]{90}{Kappa}} 
 &  activity & 64.64 & 63.53 & 19.13 & 55.72 & \textbf{67.82} & 41.22 & 45.33 & 57.74 & 60.99 & 45.24\\
 &  connect-4 & 37.68 & 31.92 & 19.68 & 33.64 & 33.49 & 32.97 & 29.85 & \textbf{41.45} & 30.75 & 31.56\\
 &  covtype & 44.59 & 54.99 & 49.18 & 52.09 & 59.29 & \textbf{73.40} & 61.28 & 55.20 & 54.09 & 56.45\\
 &  crimes & 11.61 & 2.13 & 4.20 & 0.59 & \textbf{14.33} & 1.96 & 5.16 & 11.49 & 2.48 & 4.48\\
 &  gas & 68.84 & \textbf{71.70} & 24.94 & 53.97 & 70.92 & 49.34 & 60.89 & 59.49 & 58.22 & 29.68\\
 &  olympic & 25.24 & 29.52 & 22.16 & 28.15 & 10.04 & 26.33 & 29.49 & \textbf{34.48} & 27.28 & 14.25\\
 &  poker & 15.00 & 18.81 & 33.40 & 24.83 & 24.88 & \textbf{48.52} & 46.49 & 33.99 & 13.31 & 39.19\\
 &  sensor & 52.27 & 56.73 & 45.84 & 55.17 & 55.87 & 59.55 & \textbf{60.62} & 50.13 & 56.88 & 56.46\\
 &  tags & 39.63 & 40.73 & 9.18 & 38.09 & 45.27 & 41.28 & \textbf{55.60} & 33.52 & 40.63 & 11.03\\
\midrule
\multicolumn{2}{l|}{Avg. PMAUC} & \textbf{53.34} & 53.10 & 46.80 & 51.86 & 53.18 & 49.70 & 50.32 & 52.14 & 52.44 & 49.45\\
\multicolumn{2}{l|}{Avg. Kappa} & 39.94 & 41.12 & 25.30 & 38.03 & 42.44 & 41.62 & \textbf{43.86} & 41.94 & 38.29 & 32.04\\
\midrule
\multicolumn{2}{l|}{Rank PMAUC} & \textbf{2.89} & 3.67 & 8.67 & 5.78 & 3.67 & 6.78 & 5.56 & 6.22 & 5.11 & 6.67\\
\multicolumn{2}{l|}{Rank Kappa} & 5.44 & 4.67 & 8.67 & 6.56 & 3.89 & 4.89 & \textbf{3.67} & 4.56 & 6.11 & 6.56\\
\bottomrule
\end{tabular}
\end{table*}

\subsection{Overall comparison}
\label{sec:overall}

\begin{figure}[t!]
\centering
\includegraphics[width=0.49\columnwidth]{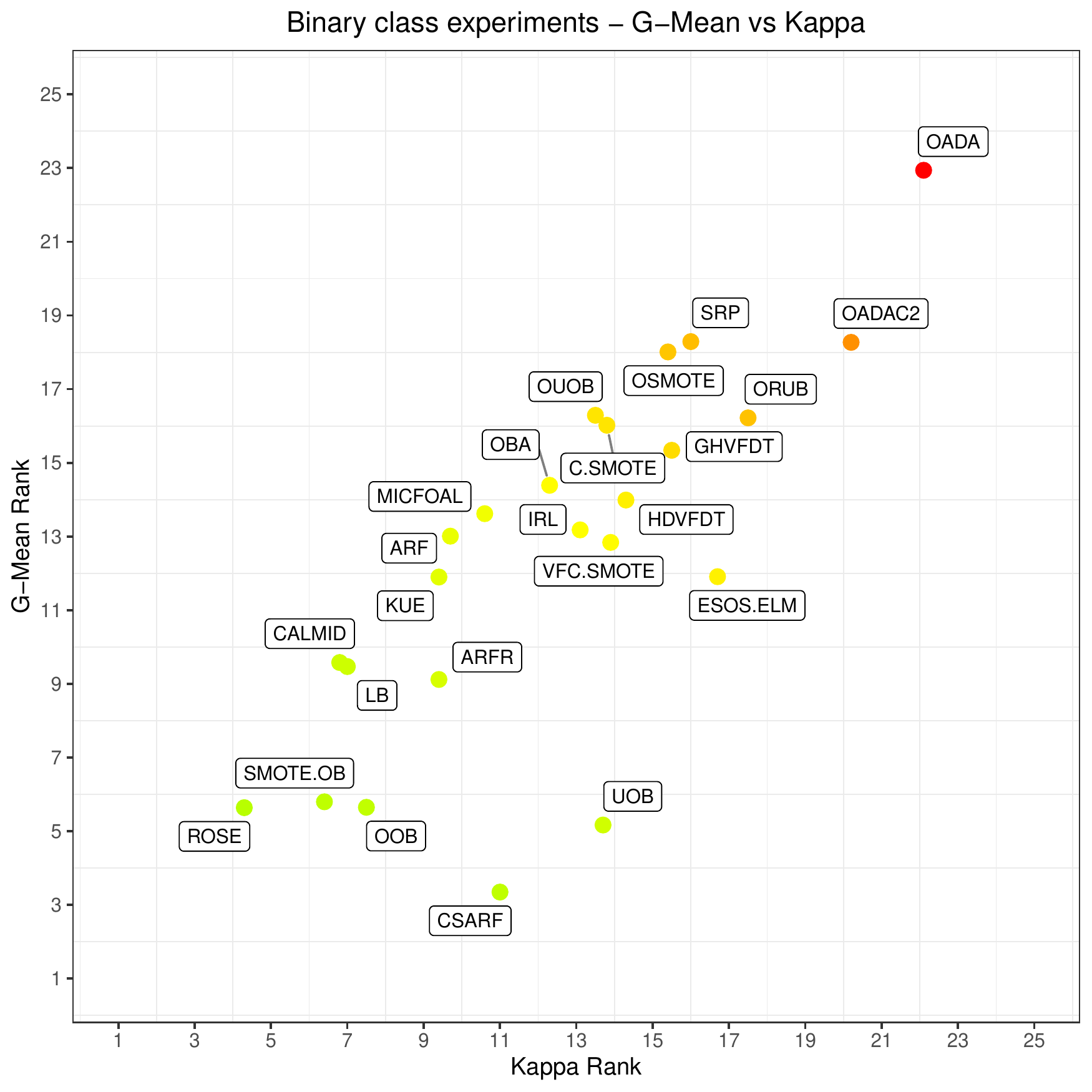}
\includegraphics[width=0.49\columnwidth]{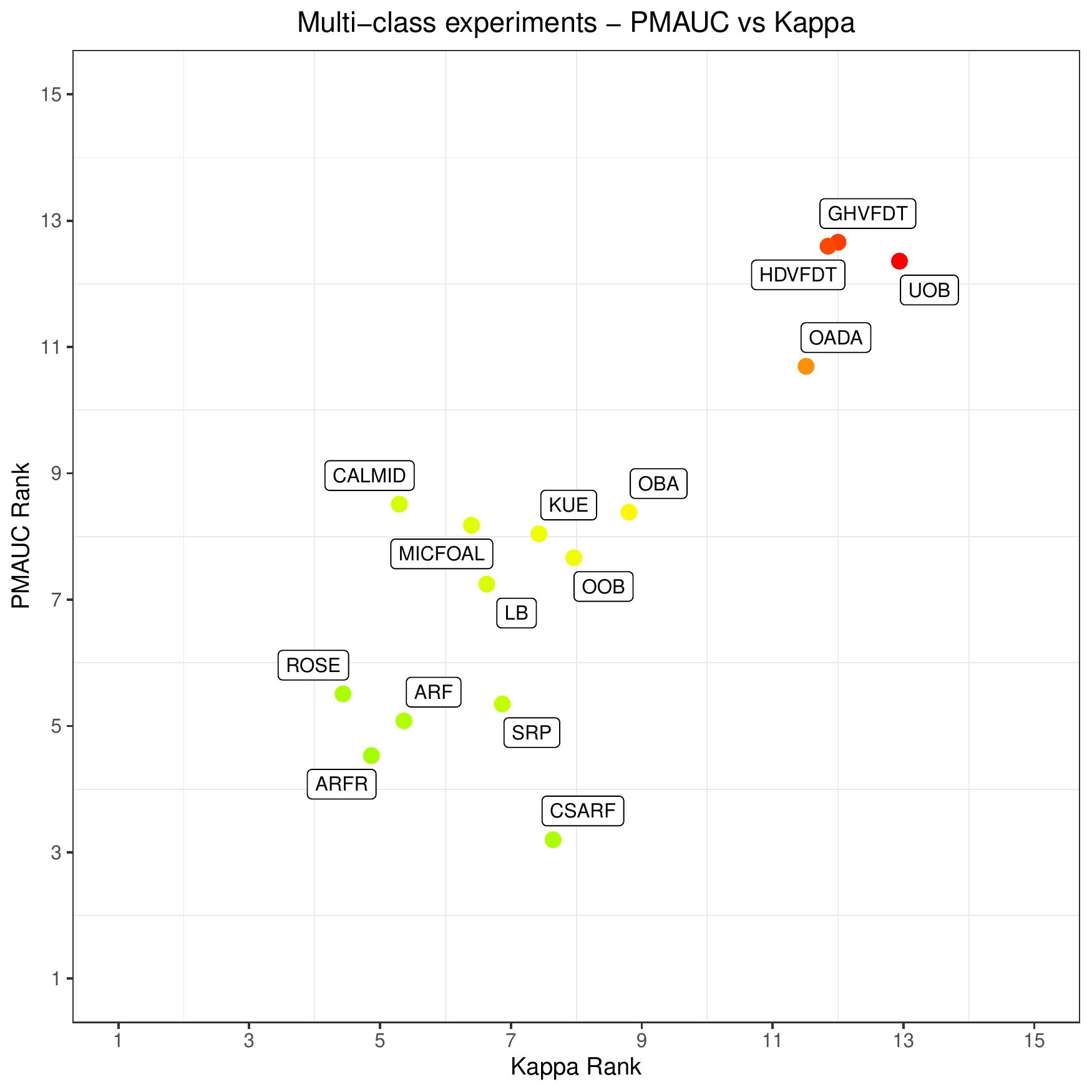}
\includegraphics[width=0.49\columnwidth]{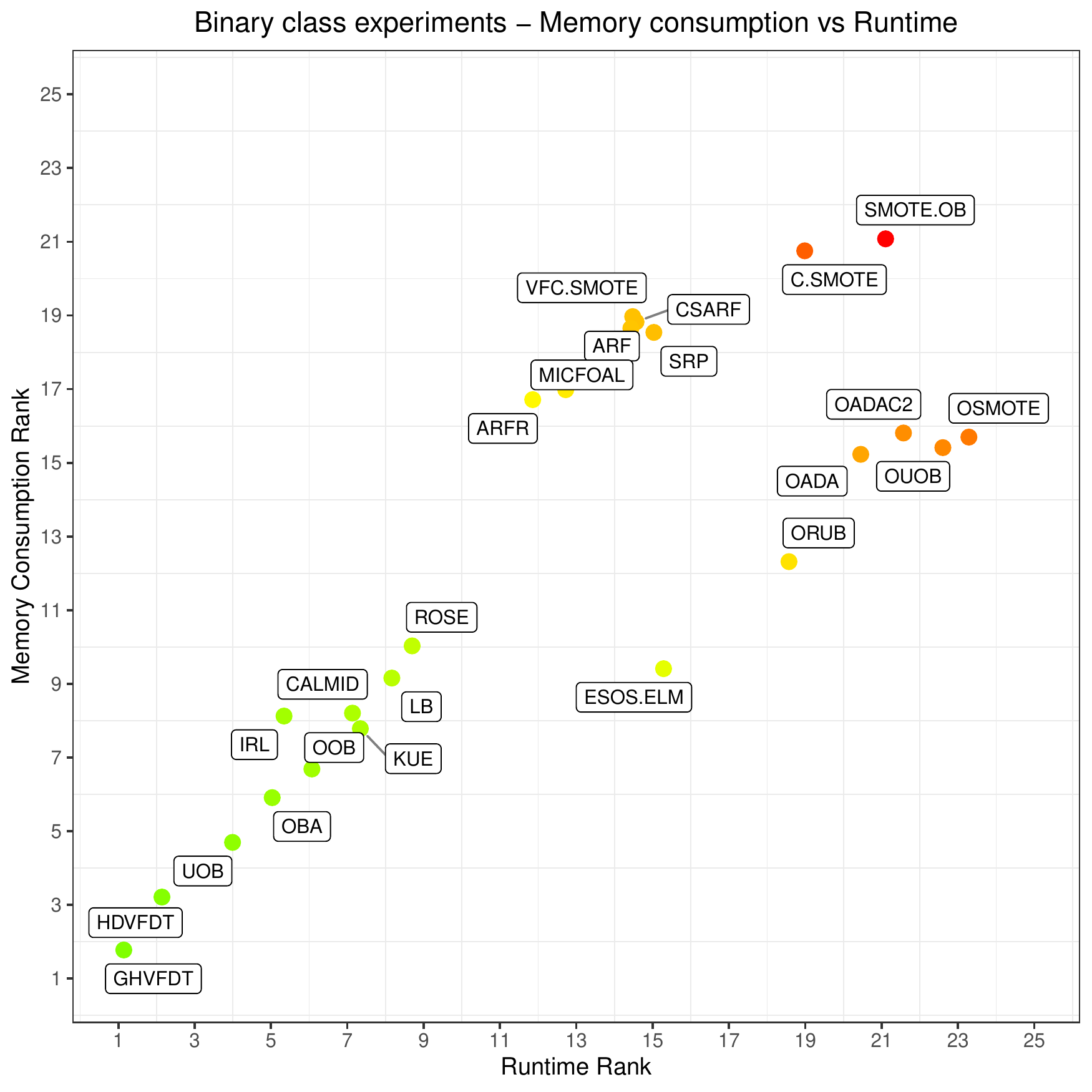}
\includegraphics[width=0.49\columnwidth]{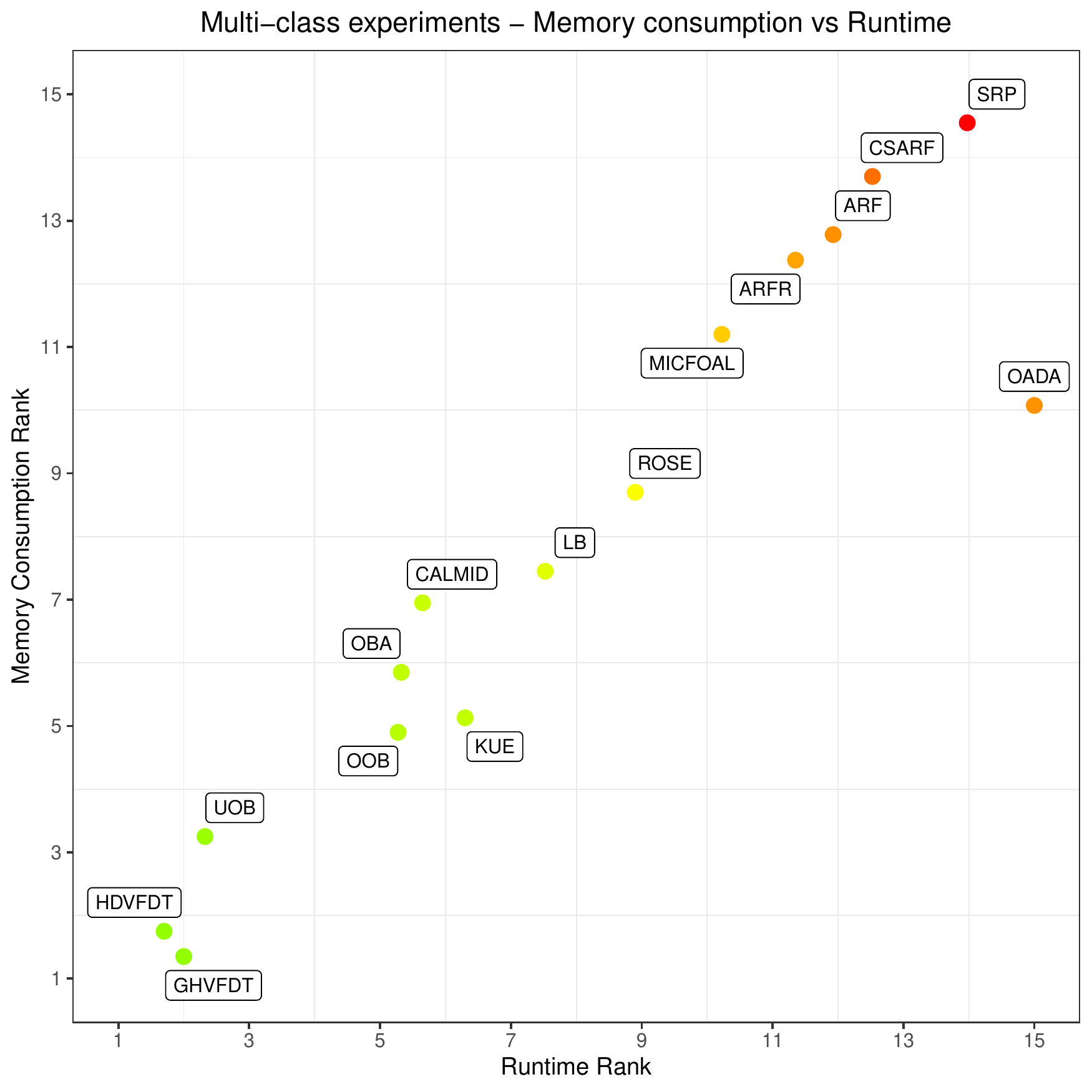}
\caption{Overall comparison of algorithms' ranks for G-Mean/PMAUC vs Kappa and Memory Consumption vs Runtime on binary and multi-class imbalanced benchmarks. Color gradient represents the product of each pair of metrics.}
\label{fig:overall_scatter}
\end{figure}

\begin{figure}[t!]
\centering
\includegraphics[width=0.49\columnwidth]{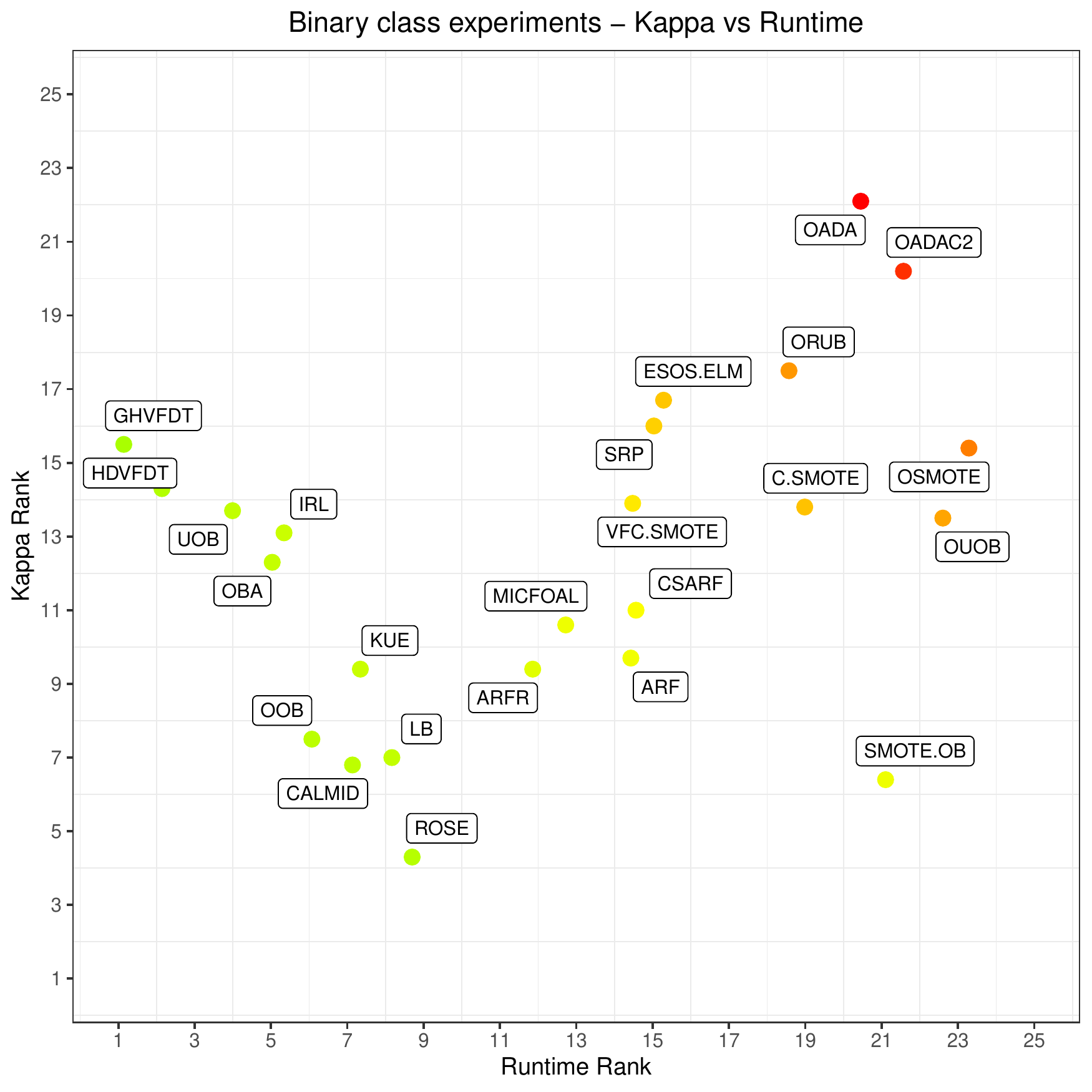}
\includegraphics[width=0.49\columnwidth]{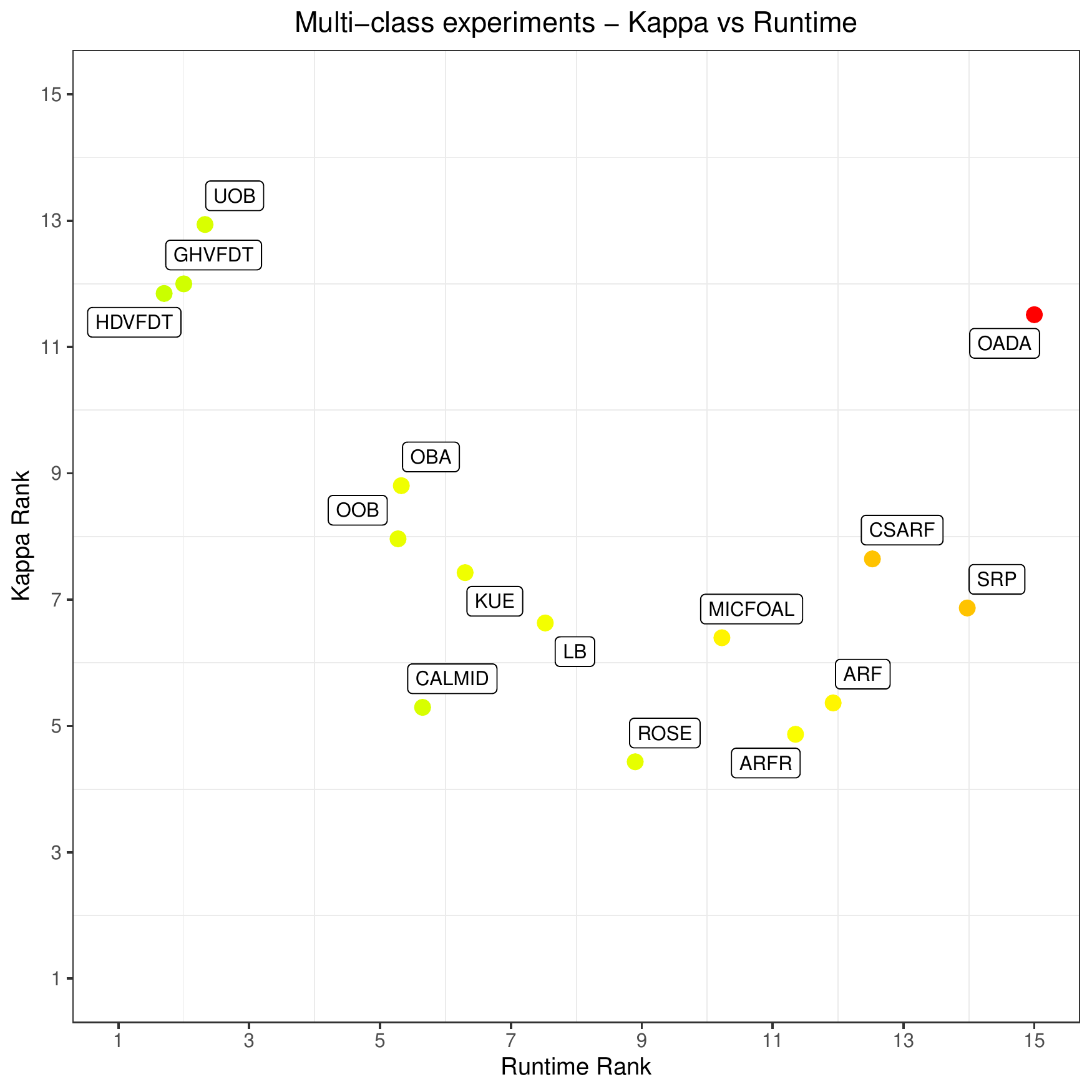}
\includegraphics[width=0.49\columnwidth]{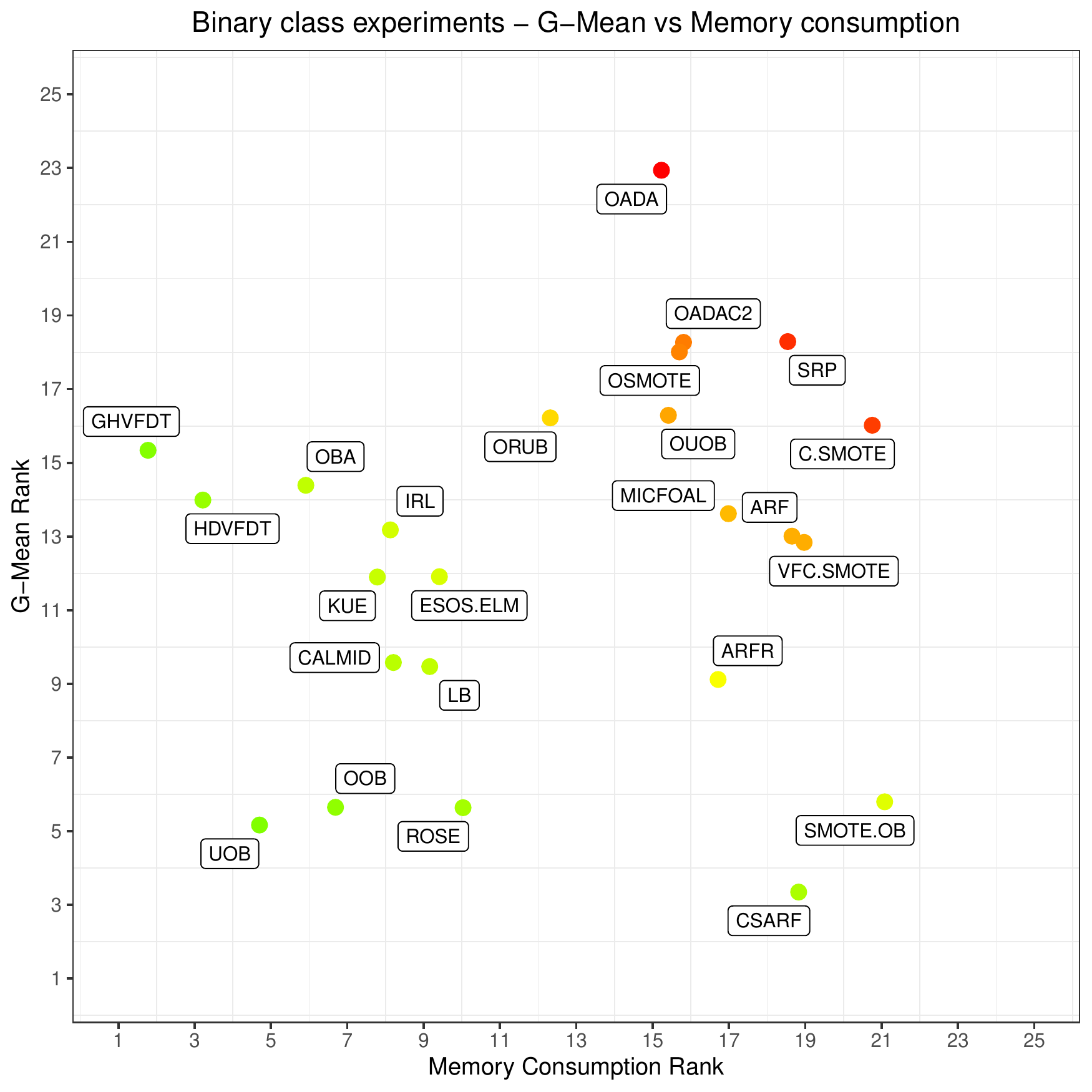}
\includegraphics[width=0.49\columnwidth]{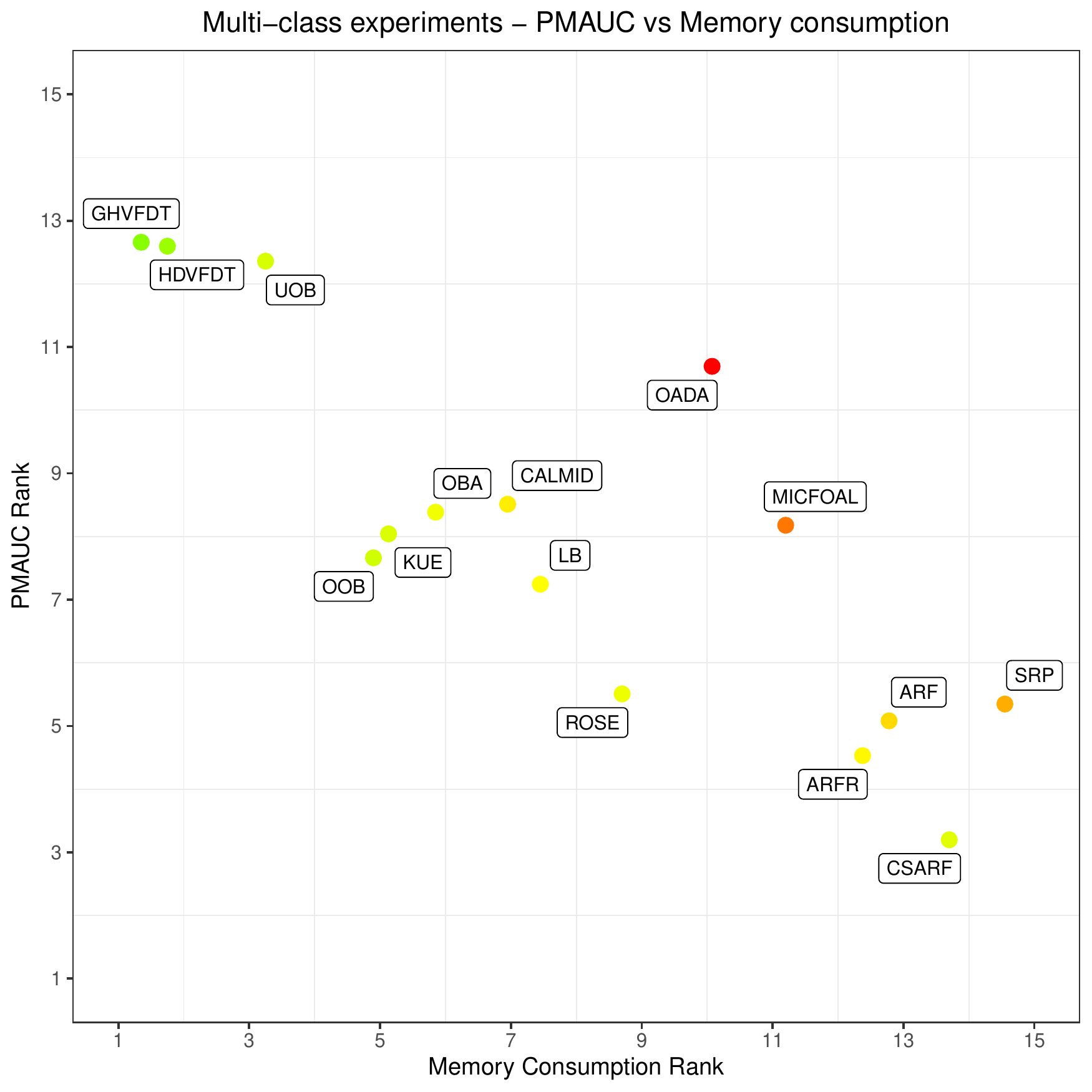}
\caption{Overall comparison of algorithms' ranks for G-Mean/PMAUC/Kappa vs Runtime/Memory Consumption on binary and multi-class imbalanced benchmarks. Color gradient represents the product of each pair of metrics.}
\label{fig:overall_scatter2}
\end{figure}

\begin{figure}[t!]
\centering
\includegraphics[width=0.49\columnwidth]{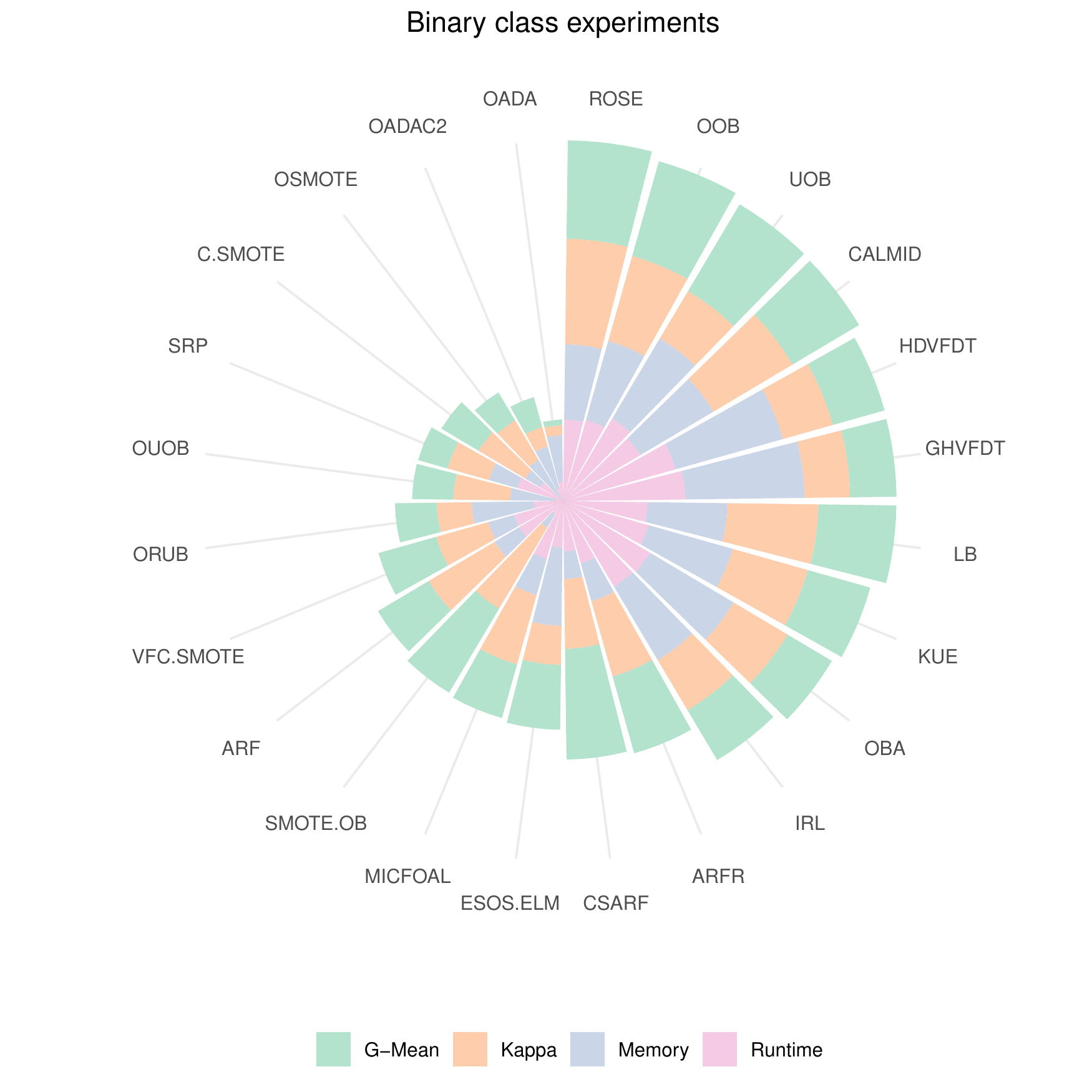}
\includegraphics[width=0.49\columnwidth]{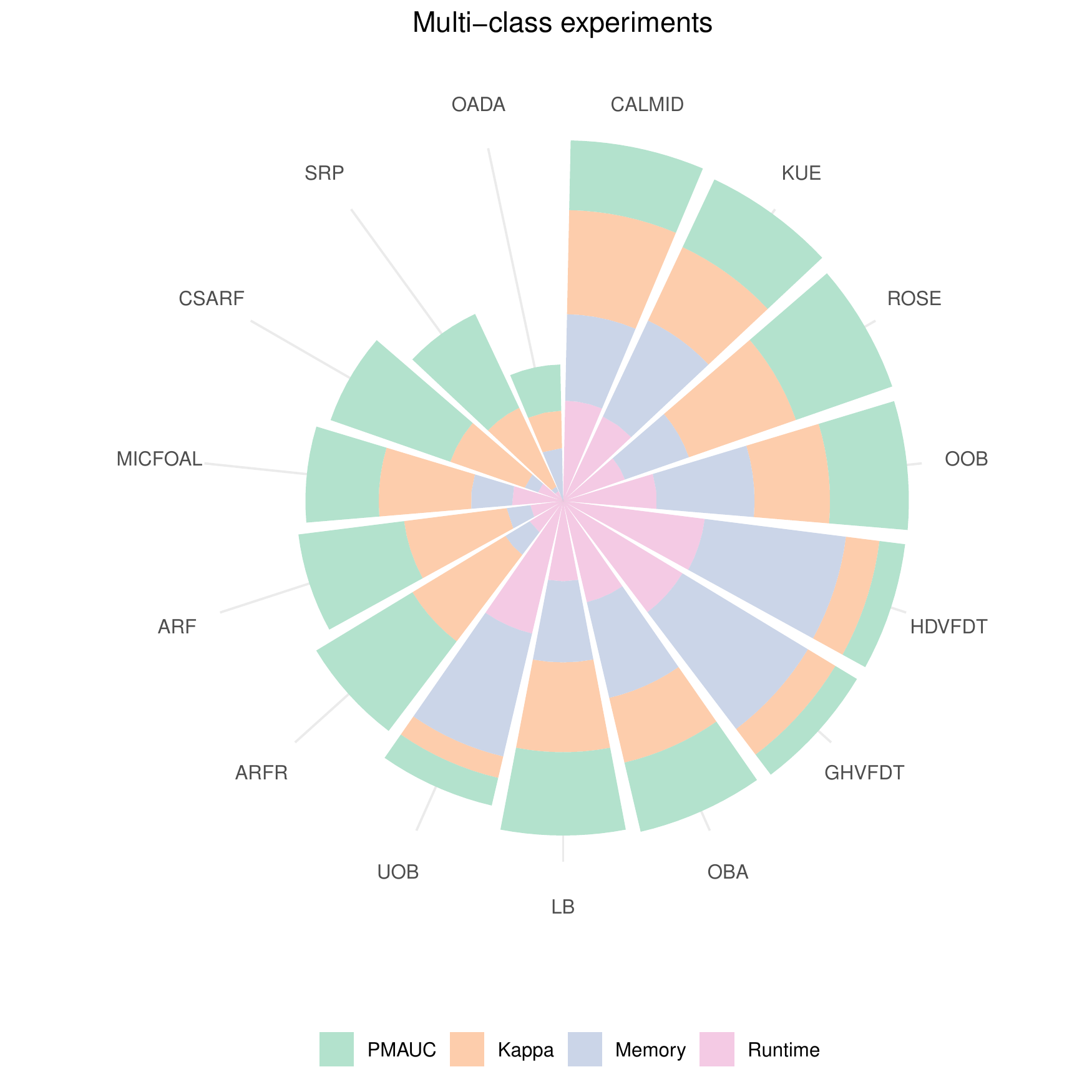}
\caption{Overall comparison of stacked algorithms' ranks for G-Mean/PMAUC, Kappa, runtime, and memory on binary and multi-class imbalanced benchmarks. Algorithms are sorted clockwise by the stacked ranks best to worst. Equal weight for the four metrics.}
\label{fig:overall_barplot}
\end{figure}

\noindent \textbf{Goal of the experiment.}
The previous experiments discussed how different individual underlying data properties affected the performance of the classifiers. The goal of this experiment is to perform a joint comparison of the algorithms, identify performance trends and divergences, that will allow us to make recommendations to end-users. Moreover, we analyze the computational and memory complexity of the algorithms to address \textbf{RQ6}. The goal of any algorithm for data streams is to simultaneously maximize the classification metrics while minimizing the runtime and memory consumption \citep{krempl2014open}. However, these are often conflicting objectives and highly accurate methods often require long runtimes, which is not acceptable for real-time high-speed data streams. Table~\ref{tab:computational_complexity} shows the runtime and memory consumption of the 24 algorithms both for binary and multi-class imbalanced streams. Figures~\ref{fig:overall_scatter} and~\ref{fig:overall_scatter2} present a pairwise joint comparison of the algorithm's ranks on G-Mean, PMAUC, Kappa, runtime and memory consumption across all experiments. Figure~\ref{fig:overall_barplot} shows a circular stacked barplot with the ranks for the four metrics. The bigger the stack the better aggregated performance. The circular barplot displays the algorithms sorted clockwise based on the stack size.

\noindent \textbf{Discussion}

\noindent \textit{Classification metrics.} All the above experiments showcased the importance of using not only more than a single metric for evaluating classifiers for imbalanced data streams, but also the importance of using diverse and complimentary metrics. G-mean and PMAUC are strongly correlated with each other and follow the same trends, thus making using both redundant. However, by adding Kappa metric we gained an additional insight into specific characteristics of evaluated classifiers, thus allowing us to better understand which of the classifiers favor only minority classes and which return balanced performance over all analyzed classes. 

Two best performing classifiers across all of experiments were \acrshort{rose} and \acrshort{csarf}. \acrshort{rose} returned single best performance regarding Kappa metric and one of the best for the other metrics. This allows us to conclude that \acrshort{rose} is a well-rounded classifier that demonstrates robustness to various learning difficulties embedded in imbalanced and drifting data streams, both binary and multi-class. \acrshort{csarf} returned excellent results in both types of experiments for G-Mean (for binary tasks) and PMAUC (for multi-class tasks) metrics. However, its rank dropped significantly under Kappa evaluation, showing that \acrshort{csarf} is driven by its performance on minority classes, not balanced performance on all of them. Furthermore, \acrshort{csarf} becomes unsuitable for scenarios with very high number of classes. 

\begin{table*}[t!]
\centering
\scriptsize
\caption{Comparison of runtime (seconds per 1,000 instances) and memory consumption (RAM-Hours).}
\label{tab:computational_complexity}
\begin{tabular}{l|R{1cm}R{1.8cm}|R{1cm}R{1cm}|R{1cm}R{1.8cm}|R{1cm}R{1cm}}
\cmidrule{2-9}
& \multicolumn{4}{c|}{Binary class experiments} & \multicolumn{4}{c}{Multi-class experiments}\\
\midrule
\multirow{2}{*}{Algorithm} & Runtime & Memory & Runtime & Memory & Runtime & Memory & Runtime & Memory\\
& (seconds) & (RAM-Hours) & (Rank) & (Rank) & (seconds) & (RAM-Hours) & (Rank) & (Rank)\\
\midrule
IRL & 0.15 & 3.93E-05 & 5.34 & 8.13 & -- & -- & -- & --\\
C-SMOTE & 18.01 & 2.39E-02 & 18.98 & 20.75 & -- & -- & -- & --\\
VFC-SMOTE & 2.51 & 2.52E-03 & 14.48 & 18.97 & -- & -- & -- & --\\
CSARF & 3.97 & 1.12E-02 & 14.56 & 18.83 & 3.35 & 5.94E-06 & 12.53 & 13.70\\
GHVFDT & \textbf{0.01} & \textbf{1.16E-08} & \textbf{1.14} & \textbf{1.78} & 0.09 & \textbf{9.92E-11} & 2.00 & \textbf{1.35}\\
HDVFDT & \textbf{0.01} & 2.85E-08 & 2.14 & 3.21 & \textbf{0.08} & 2.14E-10 & \textbf{1.70} & 1.75\\
ARF & 3.65 & 8.42E-03 & 14.43 & 18.65 & 3.14 & 1.31E-06 & 11.93 & 12.78\\
KUE & 0.11 & 4.93E-06 & 7.34 & 7.79 & 0.28 & 1.64E-08 & 6.30 & 5.13\\
LB & 0.13 & 7.30E-06 & 8.17 & 9.16 & 0.33 & 1.21E-08 & 7.53 & 7.45\\
OBA & 0.05 & 2.60E-07 & 5.03 & 5.91 & 0.25 & 2.25E-08 & 5.65 & 6.95\\
SRP & 3.51 & 6.27E-03 & 15.03 & 18.54 & 5.34 & 8.56E-06 & 13.98 & 14.55\\
ESOS-ELM & 2.61 & 1.11E-06 & 15.29 & 9.41 & -- & -- & -- & --\\
CALMID & 0.09 & 2.80E-06 & 7.13 & 8.21 & 0.25 & 2.25E-08 & 5.65 & 6.95\\
MICFOAL & 1.91 & 3.37E-03 & 12.72 & 16.98 & 1.95 & 3.72E-06 & 10.23 & 11.20\\
ROSE & 0.13 & 6.45E-06 & 8.70 & 10.03 & 0.49 & 2.82E-08 & 8.90 & 8.70\\
OADA & 24.48 & 2.10E-02 & 20.45 & 15.23 & 97.24 & 1.54E-04 & 15.00 & 10.08\\
OADAC2 & 31.61 & 2.11E-02 & 21.57 & 15.81 & -- & -- & -- & --\\
ARFR & 1.25 & 8.73E-04 & 11.86 & 16.71 & 2.94 & 1.13E-06 & 11.35 & 12.38\\
SMOTE-OB & 46.71 & 2.78E-01 & 21.10 & 21.08 & -- & -- & -- & --\\
OSMOTE & 113.38 & 3.66E-01 & 23.29 & 15.70 & -- & -- & -- & --\\
OOB & 0.07 & 1.94E-06 & 6.07 & 6.69 & 0.25 & 6.05E-08 & 5.28 & 4.90\\
UOB & 0.03 & 3.55E-07 & 3.99 & 4.70 & 0.08 & 1.08E-09 & 2.33 & 3.25\\
ORUB & 14.18 & 4.20E-03 & 18.57 & 12.32 & -- & -- & -- & --\\
OUOB & 52.58 & 3.18E-01 & 22.60 & 15.41 & -- & -- & -- & --\\
\bottomrule
\end{tabular}
\end{table*}

Other highly ranked classifiers included \acrshort{smoteob} and \acrshort{oob} for binary scenarios and \acrshort{arfr} for multi-class ones. \acrshort{smoteob} was the only classifier based on \acrshort{smote} that ranked among top performers, showing that \acrshort{smote}-based resampling for drifting streams needs to be further developed to achieve success, especially under instance-level difficulties. \acrshort{oob} did not get good results on multi-class scenarios, with close to average performance. Since \acrshort{smoteob} does not support multi-class problems we could not evaluate it in this scenario. For multi-class imbalanced data streams, \acrshort{arfr} returned excellent results. This can be explained by the ability of its architecture to deal with multi-class scenarios and adapt to changes on multiple classes. This combined with a informed resampling approach lead to an effective classifier capable of handling multiple skewed classes in the stream. 

\acrshort{oada} can be pointed out as the worst classifiers regarding classification metrics for both settings. This gives us insights about limitations of boosting-based ensembles for imbalanced data streams, where various learning difficulties destabilize the Boosting procedure and lead to low predictive power. 

\noindent \textit{Computational and memory complexity.} When evaluating a classifier for data stream mining, we have to take into account how much resources are needed to run it. In the streaming setting we often deal with a situation where memory or computational power is limited, thus we may not choose the best classifier, but the one that fits our scenario. \acrshort{hdvfdt} and \acrshort{ghvfdt} are characterized by a very small memory usage, and fast runtime. This happens because they are tree-based classifiers which are naturally lightweight, with simple structure and low-cost prediction mechanisms. \acrshort{uob} can also be seen as a relatively low-cost classifier, which is justified by its nature of removing samples from the data streams in order to balance it. Therefore, we reduce the size of each batch and obtain more compact base classifiers. 

When analyzing the classifiers that require the highest computational resources, we can see that they are dominated by oversampling-based approaches. This comes as an obvious observation, as oversampling increases the size of the already big data stream by generating a high number of artificial instances. Additionally, the increase in computational cost lies in the oversampling method itself. All \acrshort{smote}-based approaches rely on nearest neighbor computation to generate artificial instances, which leads to significant increases in their complexity. Out of approaches relying on blind oversampling (and thus free of nearest neighbor computations), \acrshort{ouob} and \acrshort{osmote} consumed the highest amount of resources. This can be explained by it employing resampling mechanisms combined with dynamic switching between them and drift detectors. Out of the classifiers that do not rely on resampling, \acrshort{oada} was the most computational heavy one. Although its memory consumption is similar to other ensemble approaches, the runtime was bigger that its peers. This is another motivation against using current boosting-based algorithms for imbalanced data streams.

\noindent \textit{Relationship between predictive power and computational and memory complexity. Is there a trade-off?.} We can see that both analyzed criteria are often in a direct opposition to each other, the most lightweight classifiers are also among the worst performing ones. Therefore, how one can strike a balance between predictive power and computational complexity? How to select the best trade-off for imbalanced data streams? To select the most suitable classifier for a given data stream we cannot always get the best-performing regarding classification metrics, due to resources restrictions. For example, \acrshort{smoteob} got excellent results regarding classification metrics, but often required more than $256$GB of RAM per run, a prohibitive number for many real-world scenarios. On the other hand, we cannot also choose the lightweight classifier if it does not present good predictive power for the problem (e.g. single tree-based classifiers are non-competitive to most of ensembles for imbalanced data streams).

Analyzing all our experiments, we aim to select such classifiers that balance both sides. We can clearly see that \acrshort{oob}, \acrshort{uob}, \acrshort{rose} and \acrshort{calmid} presented the best trade-off between their predictive performance and computational complexity for binary and multi-class experiments.
\acrshort{rose} presented the best overall performance when using equal weights for the predictive performance and complexity metrics. While second, the oversampling method in \acrshort{oob} demanded more memory and runtime when building the classifier. \acrshort{uob} undersamples the majority class which reduces the runtime complexity of the classifier learning. However, \acrshort{uob} has shown that undersampling in multi-class imbalanced data did not perform as in the binary scenario.
\acrshort{rose} and \acrshort{calmid} rely on highly efficient hybrid architectures and do not employ any costly mechanisms such as oversampling or adaptive cost-sensitive matrix. When focusing on the predictive metrics only, \acrshort{rose}, \acrshort{smoteob}, and \acrshort{csarf} perform the best on binary class while \acrshort{arfr}, \acrshort{rose} and \acrshort{csarf} perform the best on multi-class.

\section{Lessons learned}
\label{sec:lessons}

In order to address \textbf{RQ7} and summarize the knowledge we extracted through the extensive experimental evaluation, this section presents the lessons learned and recommendations for future researchers.

\noindent \textbf{Design of the experimental study.} To gain insights into the performance of classifiers and fairly evaluate them for imbalanced data streams, a properly designed experimental testbed is crucial. The experimental evaluation must be done in a holistic and comprehensive manner that will assess the robustness of the classifiers to the most important challenges embedded in imbalanced data streams. These must include: (i) static and dynamic imbalance ratios with switching class roles; (ii) instance-level difficulties; (iii) various types and speeds of concept drift; (iv) binary and multi-class scenarios; (v) increasing number of classes; and (vi) real-world datasets. Only such a comprehensive evaluation will allow for comparing new classifiers to existing state-of-the-art. For the sake of reproducible research, this paper offers a ready to use testbed available on GitHub that allows for easy and reproducible evaluation of new classifiers designed for imbalanced data streams. 

\noindent \textbf{Class imbalance approach.} Our experiments showed that among the top performing methods we had two approaches based on training modifications (\acrshort{rose} and \acrshort{calmid}), two approaches based on resampling (\acrshort{arfr} and \acrshort{smoteob}), and one cost-sensitive method (\acrshort{csarf}). This is a very interesting outcome, as it shows that any of existing approaches to class imbalance can achieve excellent robustness and thus confirms the no-free-lunch theory – there is no single best way of tackling class imbalance in drifting data streams. Each of these solutions has their merits and works best in slightly different settings. In the next section we will formulate recommendations on what algorithms should be used in which scenarios. For future research it is important to understand what characteristics of each successful algorithm led to its superior performance, as those characteristics should be preserved and further developed when designing new classifiers.

\noindent \textbf{Desirable properties of data-level solutions.} When analyzing the resampling-based algorithms, we can see the dominance of oversampling approaches, both in their blind and informative versions. Blind oversampling has much lower computational cost and good reactivity to concept drift. However, it fails in multi-class scenarios, especially with high number of classes. Informative oversampling based on \acrshort{smote}, when combined with ensembles, offer a very high predictive power, being able to handle instance-level difficulties and adapt to various types of non-stationary stream characteristics. This came at the price of extremely high computational complexity (mainly due to the distance calculations), as well as being currently designed only for binary problems. 

\noindent \textbf{Desirable properties of algorithm-level solutions.} When analyzing the algorithm-level solutions, we can see that two main dominant approaches were based either on modifying training method or using cost-sensitive classification. \acrshort{rose} stands as a primary example of effective training modification, as it offers top performance over a plethora of analyzed scenarios and the best robustness to various learning difficulties. This can be contributed to combination of diversity assurance for base classifiers (on both instance and feature levels), effective classifier replacement scheme (where pruning can replace multiple classifiers at once), and not relying on any resampling scheme (instead using class-specific buffers that allow for handling high number of classes). Those modifications allowed \acrshort{rose} to strike a balance between predictive power (across all metrics) and its computational complexity. Cost-sensitive solution realized within \acrshort{csarf} showed that the combination of efficient design with cost matrix leads to a highly competitive classifier that offers great adaptation to concept drift and do not rely on any resampling. However, current limitations of cost-sensitive approaches include bias towards G-mean/PMAUC (while underperforming on Kappa) and inability to effectively handle higher number of classes. 

\noindent \textbf{Ensemble architectures.} All experiments pointed out to the dominance of bagging-based and hybrid ensemble architectures (please note that most successful hybrid architectures were also rooted in bagging). Both static and dynamic ensemble setups worked well with bagging initialization, showing that this leads to creation of diverse base learners that can perform well under concept drift and various learning challenges. Furthermore, ensembles that added a feature space diversification on top of bagging, such as \acrshort{rose} or \acrshort{arfr} were among the top performers. This shows that the feature space manipulation is a highly promising direction. Boosting proved to be the least efficient, not being able to cope with high imbalance ratios or data-level difficulties. 

\noindent \textbf{Adaptation to concept drift vs robustness to class imbalance.} We can see that the most challenging scenarios where when dynamic class imbalance was combined with concept drift. Here we could observe that the classifiers either focused on drift adaptation, or handling bias towards majority classes. Interestingly, classifiers with very good adaptation mechanisms tend to perform slightly better in these complex scenarios than their counterparts that focus mainly on robustness to imbalance. 

\noindent \textbf{Data-level difficulties.} Instance-level characteristics can be very disruptive to existing algorithms for imbalanced data streams. They should be analyzed not only as individual instances, but also as subconcepts within minority class that can evolve over time (e.g. merge or split). We can see that resampling-based solutions tend to perform well under these difficulties, mirroring observations for static data. However, none of the algorithms could explicitly use the instance-level characteristics to their advantage, as suggested by \citep{krawczyk2017cost}. 

\noindent \textbf{Handling high number of classes.} When analyzing the robustness of classifiers to very high number of classes, we observed that \acrshort{srp}, a general-purpose ensemble with no skew-insensitive mechanisms, returns one of the best performances. This, combined with very good performance of \acrshort{rose}, allows us to conclude that for multi-class imbalanced problems with very high number of classes using lower dimensional representations may lead to simplification of learning tasks. Using feature subspaces may lead to more diverse capturing of relationships among classes. This confirms observations made by \citep{korycki2021low} that discussed the merit of low-dimensional embeddings for extremely imbalanced and difficult data streams. 

\noindent \textbf{Classifier evaluation.} To evaluate a classifier in imbalanced data streams, we require the use of multiple diverse and complimentary metrics. In our testbed we argue for the use of Kappa and G-Mean/PMAUC. These metrics assess different and complementary perspectives, thus if only one is provided the evaluation of a classifier is biased towards measuring how it performs on minority class under highly imbalance ratio (Kappa) or on how it balances majority and minority class performance (PMAUC/G-mean). We showed how under high imbalance ratios, Kappa significantly penalizes the false positives whereas G-Mean tolerates a larger proportion of false positives.

\noindent \textbf{Computational and memory complexity.} One must take into an account the trade-off between predictive power and computational complexity. Algorithms requiring lowest resource consumption are among the weakest ones (such as skew-insensitive versions of Adaptive Very Fast Decision Trees). On the other hand, some of the best performing classifiers are characterized by almost prohibitive computational complexity (e.g. \acrshort{smoteob}). \acrshort{rose}, \acrshort{calmid} and \acrshort{oob} presented the best trade-off between computational resources consumption and predictive power.

\section{Recommendations}
\label{sec:rec}

After analyzing all the scenarios and evaluating different approaches to class imbalance, we could summarize some recommendations to help future researchers when designing their own algorithms to tackle imbalanced data streams and other learning difficulties:

\noindent \textbf{Choose the best off-the-shelf algorithms.} If you are looking for efficient classifiers for solving your real-world imbalanced data streams, or you are looking for effective reference methods for your experiments, it is important to be aware of the most efficient off-the shelf solutions. Based on our exhaustive experimental study, we can recommend \acrshort{rose}, \acrshort{csarf}, \acrshort{oob}, \acrshort{arfr}, and \acrshort{calmid} as the ready to use and effective classifiers. We especially recommend using \acrshort{rose} due to its balanced performance, great trade-off between predictive power and computational cost, excellent robustness in all analyzed scenarios, as well as ease of use due to its autonomously self-adaptive parameters. 

\noindent \textbf{Analyze the dynamics of imbalance ratio.} In data streams where imbalance ratio is static, oversampling and training modification methods return excellent performance. Ensembles based on bagging and hybrid architecture are a good choice. When it comes to evolving imbalance ratios, we need a more sophisticated mechanism adapting to the changing imbalance ratio. Here we can see a dominance of algorithm-level solutions that offer dynamic ensemble line-up with effective pruning, such as \acrshort{rose}.

\noindent \textbf{Consider the presence of concept drift.} Our experiments showed that many skew-insensitive classifiers suffer due to their lackluster adaptation mechanisms. On the other hand, general-purpose classifiers can display surprisingly good performance in specific cases, showing the impact of recovery from concept drift. This allows us to recommend paying close attention to embedding an efficient concept drift adaptation mechanism into your method. Regardless of how robust your skew-insensitive mechanism will be, it will not be sufficient to cope with the drifting nature of imbalanced data streams. 

\noindent \textbf{Check for instance-level difficulties.} Instance-level difficulties in data streams pose significant difficulties to most of the classifiers \citep{Brzezinski2021}. It is crucial to analyze your stream to understand if such factors are present. We noticed that methods based on oversampling tend to handle instance-level difficulties particularly well. However, none of them can directly take an advantage of such challenging instances to improve adaptation and robustness. Existing research suggest that incorporating such information during learning from imbalanced streams may be highly beneficial \citep{krawczyk2017cost}. Therefore, we recommend to truly understand the nature of streams you are working with and focusing on how you can leverage this information to make your classifiers more robust. 

\noindent \textbf{Consider the number of classes.} There is a significant difference in developing methods for binary and multi-class imbalanced data streams. While some of algorithms work well regardless of the number of classes (e.g. \acrshort{rose}), other are very sensitive to it and their performance deteriorates significantly with increase in the number of classes (e.g. \acrshort{csarf}). Multi-class data streams will require the development of dedicated resampling algorithms, just like in the static scenarios \citep{Krawczyk:2020rbo}. Existing resampling methods work well mainly in binary cases and do not translate well to a higher number of classes. Finally, most of the existing classifiers work under fixed number of classes. This should be considered when dealing with emerging and disappearing classes, as existing classifiers need to be extended with dedicated mechanisms to handle this phenomenon \citep{masud2009integrating,masud2010classification,masud2010addressing}. 

\noindent \textbf{Think outside of the box.} While data-level and algorithm-level solutions are the most popular approaches to handling class imbalance, there are other promising directions to explore. Instead of focusing on another online resampling method or cost-sensitive modification, explore alternative solutions. Our experiments showed the high promise behind low-dimensional representations for imbalanced data streams, as firstly explored by \citep{korycki2021low}. This is just the tip of an iceberg in developing novel techniques tailored to imbalanced data streams that do not follow these two most popular directions. 

\noindent \textbf{Use fair and holistic evaluation.} New classifiers for imbalanced data streams should always be compared with both the popular methods (e.g. \acrshort{oob} or \acrshort{uob}), as well as with the most recently published and top performing ones (as of the time of this study these will include \acrshort{rose}, \acrshort{csarf}, or \acrshort{oob}). It is important to use an established experimental setup and follow the best practices in this field. This paper provides reproducible code for the entire testbed, along with all examined classifiers and datasets. This is the first standardized approach for evaluating classifiers for imbalanced data streams. We recommend for future researchers to simply plug-in their new methods into our framework to ensure fair and holistic evaluation of newly proposed methods. 

\noindent \textbf{Do not neglect using general-purpose ensembles as reference.} Our experiments showed that general-purpose ensembles can return surprisingly good performance for non-stationary imbalanced data streams, due to their well-designed drift adaptation mechanisms. Therefore, it is important to use them as a point of reference to see if the proposed skew-insensitive mechanism actually contributes significantly to the performance of a new classifier.

\noindent \textbf{Use multiple performance metrics.} There are many performance metrics for evaluating imbalanced data streams including Kappa, G-Mean, PMAUC, WMAUC, EWMAUC. Section~\ref{sec:metrics} presented the different aspects these performance metrics assess, and acknowledged the different biases in individual metrics. We recommend using multiple metrics exhibiting complementary behavior rather than picking a single metric.

\noindent \textbf{Ensure reproducible research.} Reproducible research is the key towards the advancement of the machine learning community. If you want your method to have an impact, always provide the source code on GitHub and use popular frameworks such as MOA \citep{bifet2010moa}, River \citep{2020river}, Stream-learn \citep{KSIENIEWICZ202211}, and Scikit-multiflow \citep{montiel2018scikit}. This will make sure that other researchers can use your classifier, as well as that it can be easily embedded in existing frameworks, for comparison with other methods.

\noindent \textbf{One size does not fit all.} This survey paper presents a very large experimental evaluation of as many imbalanced data scenarios as possible in order to compare existing methods in the state of the art. It is not our intention nor realistic that every study from now on is required to always use the full set of benchmarks. Our goal is that future works can build on our recommendations to include some of the benchmarks proposed as appropriate in each work, acknowledging that not all of them are necessary nor suitable for all studies.

\section{Open challenges and future directions}
\label{sec:open}

After formulating recommendations regarding the currently available algorithms, we will now present and discuss open challenges and future directions for learning from imbalanced data streams.  

\noindent \textbf{Informative and fast resampling.} Our experimental study showed that current undersampling-based methods underperform for imbalanced data streams, especially when faced with multiple classes. There is a need to develop novel and informative undersampling approaches that can adapt to concept drift and allow to efficiently tackle dynamic class imbalance, while preserving the desirable low computational complexity. Current informative oversampling methods are rooted in SMOTE, offering good improvements in predictive power at the high computational cost. We should develop novel oversampling methods that do not rely on a nearest neighbor approach, thus reducing the computational complexity and alleviating SMOTE limitations \citep{Krawczyk:2020rbo}.

\noindent \textbf{Proactive instead of reactive tackling of dynamic class imbalance.} Existing methods focus on adaptation to both concept drift and dynamic class imbalance after the change has taken place. But is there a possibility to anticipate the change? Can we predict how the class imbalance will evolve over time and offer proactive approach? This would significantly reduce the recovery time after changes in data streams and lead to more robust classifiers. 

\noindent \textbf{Improving boosting-based ensembles.} We have discussed how existing boosting-based ensembles perform poorly for imbalanced data streams. Yet boosting is one of the most successful ensemble architectures and deserves a second chance. We hope that the weaknesses of boosting identified in this paper will help other researchers develop more suitable classifiers based on this architecture, capable of fast adaptation to changes and overcoming small sample size in minority classes. 

\noindent \textbf{Handling evolving number of classes.} While we investigated the impact of the number of classes on imbalanced problems, we have not touched upon dynamic changes in class numbers \citep{masud2009integrating,masud2010classification,masud2010addressing}. In data stream scenarios classes may emerge, disappear, and recur over time. An evolving number of classes combined with dynamic imbalance ratio creates an extremely challenging scenario that requires new and flexible models capable of detecting and incorporating new classes into their structures, as well as forgetting the outdated classes and remembering recurring classes \citep{masud2011detecting,masud2012classification,al2012stream,sun2016online}. We envision strong parallels with continual and lifelong learning approaches \citep{Korycki:2021cl}.

\noindent \textbf{Fairness in imbalanced data streams.} Algorithmic fairness is a subject of intense research \citep{iosifidis2021online}, aiming at creating non-biased classifiers that do not rely on protected attributes. Recent works by suggest that algorithmic fairness and class imbalance are the two sides of the same coin, as protected information is often displayed by underrepresented, minority groups. Fairness in data stream mining could benefit from enhancing existing methods with skew-insensitive approaches, as both domains aim at countering bias in data.

\noindent \textbf{Online skew-insensitive feature selection.} We have noticed a superior performance of ensembles based on reduced feature subspaces, especially for difficult multi-class problems. While existing methods are based on randomized approaches, there is a need to develop efficient online feature selection methods insensitive to class imbalance. This will allow not only to create more compact classifier, filter irrelevant features, but also eliminate features that increase bias towards the majority class. This could further be expanded into scenarios where the feature space size evolves over time.

\noindent \textbf{Beyond binary and multi-class imbalanced data streams.} Most of the existing research in imbalanced data streams focuses on binary and multi-class classification. However, multiple other tasks in data streams may be subject to data imbalance. Multi-label data is inherently imbalanced and calls for dedicated methods capable of handling multi-target outputs \citep{2022-NEUCOM-AESAKNNS}. Regression from streams is also frequently subject to imbalance in the form of rare values, as frequencies of specific ground truths may evolve over time \citep{branco2017smogn,aminian2021chebyshev}. Finally, streaming times series also require dedicated resampling and skew-insensitive methods to facilitate robust predictions. 

\section{Conclusions}
\label{sec:conclusion}

\noindent \textbf{Summary.} In this paper, we offered an exhaustive and informative experimental review of classification methods for imbalanced data streams. We designed a robust experimental framework, publicly available for reproducibility, to evaluate state-of-the-art classifiers in varied scenarios and understand how each aspect of imbalanced data streams affects the performance of classifiers, and provide a template for future researchers to evaluate their newly classifiers with the state of the art. With this experimental framework, we performed an experimental comparison with $24$ algorithms in multiple scenarios to analyze their behavior and discuss their performance trends and divergences. The classifiers were evaluated on 515 benchmarks with different difficulties such as dynamic and static imbalance ratio, with and without concept drift, the presence of data-level difficulty factors, and real-world problems. All these settings were evaluated isolated and combined, in a binary and multi-class scenario, to gain insights and understand how they would affect the underlying learning mechanisms of data-streams classifiers. Throughout the experiments, we could demonstrate which approaches work or do not work for each scenario, such as undersampling techniques were undermined in multi-class scenarios, and dynamic ensemble methods such as \acrshort{rose} could do better in many different settings, demonstrating robustness. Our proposed experimental framework allowed us to get insights into all the classifiers and how would they perform in different scenarios, therefore future researchers can follow the same standard of evaluation when proposing their classifier for imbalanced data streams, in order to achieve the most transparent and complete results possible. 

\noindent \textbf{Towards the future of reproducible research in data stream mining.} We proposed a standardized and holistic framework for evaluating imbalanced data streams. We strongly believe that this is a crucial step towards unifying the community working in this domain, offering a flexible tool for long-time practitioners, and an easy way to get started for newcomers. Guidelines and recommendations formulated in this paper should allow more streamlined and effective improvement of existing algorithms and development of new solutions. Only as a community working together, we can truly advance our understanding of data streams and design truly impactful, well-rounded, and thoroughly evaluated algorithms that will be used in both academia and industry.

We hope that our framework will begin to grow over time with new algorithms, problems, and benchmarks being added by the community. There are still many questions unanswered in this domain and many open challenges for the future. We look forward to discovering new knowledge together.

\begin{acknowledgements}
High Performance Computing resources provided by the High Performance Research Computing (HPRC) Core Facility at Virginia Commonwealth University (https://hprc.vcu.edu) were used for conducting the research reported in this work.
\end{acknowledgements}

\newpage
\section*{Declarations}

\noindent \textbf{Funding.} This research was partially supported by the 2018 VCU Presidential Research Quest Fund (Alberto Cano) and an Amazon AWS Machine Learning Research award (Alberto Cano \& Bartosz Krawczyk).

\noindent \textbf{Conflict of interest.} The authors declare that they have no conflict of interest.

\noindent \textbf{Ethics approval.} Not Applicable.

\noindent \textbf{Consent to participate.} Not Applicable.

\noindent \textbf{Consent for publication.} Not Applicable.

\noindent \textbf{Availability of data and material.} Data \& materials available at \href{https://github.com/canoalberto/imbalanced-streams}{https://github.com/canoalberto/imbalanced-streams}

\noindent \textbf{Code availability.} Source code is available at \href{https://github.com/canoalberto/imbalanced-streams}{https://github.com/canoalberto/imbalanced-streams}

\noindent \textbf{Authors' contributions.} Gabriel Aguiar contributed to the manuscript preparation. Alberto Cano contributed to the experimental evaluation and manuscript preparation. Bartosz Krawczyk contributed to the manuscript preparation. All authors read and approved the final manuscript.

\setglossarystyle{listdotted}
\setlength{\glslistdottedwidth}{.2\linewidth}
\printnoidxglossary[type=\acronymtype]

\newpage
\bibliographystyle{spbasic} 
\bibliography{references}

\end{document}